%% file: main.tex
\documentclass{article}

\PassOptionsToPackage{numbers, compress}{natbib}

\usepackage[final]{style/neurips_2025}

\input{style/neurips_2025}

\title{IntrinsiX: \\ High-Quality PBR Generation using Image Priors}

\author{
Peter Kocsis \quad\quad
Lukas H{\"o}llein \quad\quad
Matthias Nie{\ss}ner \\
\vspace{6pt}
Technical University of Munich \\\\
\tt\href{https://peter-kocsis.github.io/IntrinsiX/}{peter-kocsis.github.io/IntrinsiX/}
}

\input{figures/teaser}

\makeatletter
\renewcommand{\settitle}{\@maketitle}
\makeatother

\begin{document}
\maketitle

\input{sections/0-abstract}       

\input{sections/document}           

\input{sections/acknowledgements}


\bibliographystyle{style/ieeenat_fullname}
\bibliography{sections/bibliography}

\input{sections/checklist}

\input{sections/supplementary}     

\end{document}

%% file: style/neurips_2025.tex
\usepackage[utf8]{inputenc} 
\usepackage[T1]{fontenc}    
\usepackage{hyperref}       
\usepackage{url}            
\usepackage{booktabs}       
\usepackage{amsfonts}       
\usepackage{nicefrac}       
\usepackage{microtype}      
\usepackage{xcolor}         

\input{src/header} 

%% file: src/header.tex
\input{src/published-packages} 
\input{src/published-commands} 

\input{src/dev-commands} 

%% file: src/published-packages.tex
\usepackage[export]{adjustbox}

\usepackage{subcaption}  
\usepackage{makecell}   
\usepackage{amsmath}
\usepackage{nicefrac}
\usepackage{multirow}

\usepackage{tikz}
\usetikzlibrary{tikzmark}
\usepackage{enumitem}

\usepackage{mathtools}
\usepackage{caption}

\usepackage{cleveref}
\usepackage{wrapfig}
\usepackage{subcaption}

%% file: src/published-commands.tex
\makeatletter
\apptocmd\@maketitle{{\myfigure{}\par}}{}{}
\makeatother

\makeatletter
\newcommand{\settitle}{\@maketitle}
\makeatother

%% file: src/dev-commands.tex
\definecolor{LIGHTPINK}{RGB}{237,157,202}
\definecolor{LIGHTRED}{RGB}{210,121,121}
\definecolor{LIGHTORANGE}{RGB}{230,170,50}
\definecolor{LIGHTGOLD}{RGB}{210,194,121}
\definecolor{LIGHTGREEN}{RGB}{121,210,121}
\definecolor{LIGHTAQUA}{RGB}{121,206,210}
\definecolor{LIGHTBLUE}{RGB}{121,124,210}
\definecolor{LIGHTPURPLE}{RGB}{153,102,255}
\definecolor{RED}{RGB}{178,34,34}
\definecolor{GRAY}{RGB}{166,166,166}
\definecolor{WHITE}{RGB}{255,255,255}

\newcommand{\showrevise}[1]{}  



%% file: figures/teaser.tex
\newcommand{\myfigure}{
    \centering
    \vspace{-18pt}
    \includegraphics[width=\textwidth]{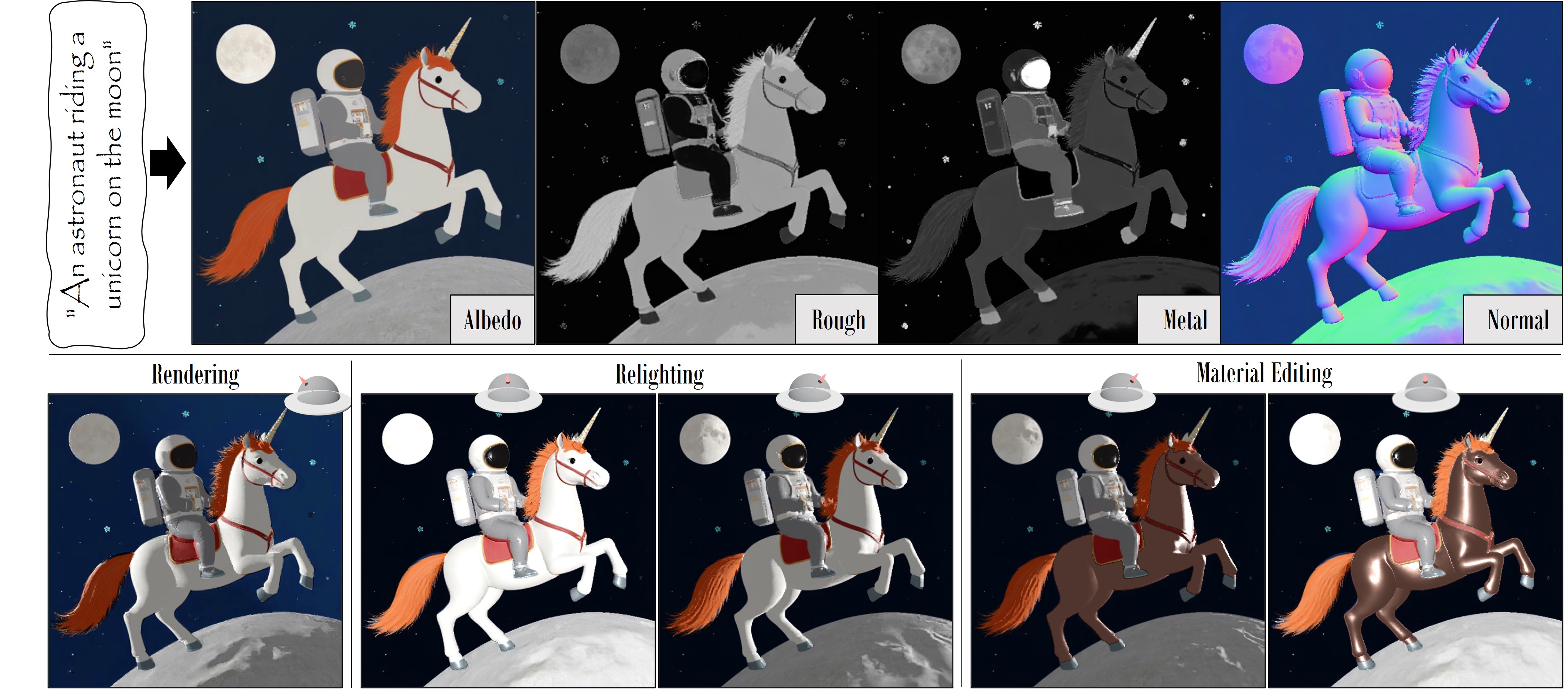}
  \vspace{-9pt}
\captionsetup{type=figure}\caption{\textbf{IntrinsiX.} 
    We present a text-guided intrinsic image generator. 
    Given a text prompt, our method produces high-quality albedo, roughness, metallic, and normal maps which can be rerendered under any lighting conditions. 
    Our model enables downstream applications, such as relightable object or scene generation, and material or lighting editing. 
        }
    \label{fig:teaser}
    \vspace{6pt}
}

%% file: sections/0-abstract.tex
\begin{abstract} 

We introduce IntrinsiX, a novel method that generates high-quality intrinsic images from text description.
In contrast to existing text-to-image models whose outputs contain baked-in scene lighting, our approach predicts physically-based rendering (PBR) maps.
This enables the generated outputs to be used for content creation scenarios in core graphics applications that facilitate re-lighting, editing, and texture generation tasks. 
In order to train our generator, we exploit strong image priors, and pre-train separate models for each PBR material component (albedo, roughness, metallic, normals).
We then align these models with a new cross-intrinsic attention formulation that concatenates key and value features in a consistent fashion. 
This allows us to exchange information between each output modality and to obtain semantically coherent PBR predictions.
To ground each intrinsic component, we propose a rendering loss which provides image-space signals to constrain the model, thus facilitating sharp details also in the output BRDF properties. 
Our results demonstrate detailed intrinsic generation with strong generalization capabilities that outperforms existing intrinsic image decomposition methods used with generated images by a significant margin.
Finally, we show a series of applications, including re-lighting, editing, and for the first time text-conditioned room-scale PBR texture generation.
We will release the code and the pre-trained model weights. 

\end{abstract}

%% file: sections/document.tex

\input{sections/1-introduction}
\input{sections/2-relatedwork}

\input{sections/3-method}

\input{sections/4-experiments}
\input{sections/5-conclusion}

%% file: sections/1-introduction.tex

\section{Introduction}
\label{sec:introduction}

Text-to-image (T2I) models have revolutionized 2D content creation, by generating high-quality RGB images from just a text description \cite{rombach2022high, SaharCSLWDGAMLSHFN2022, ramesh2022hierarchical}.
They are used in widespread applications, including extensions for controllable generation beyond text \cite{zhang2023controlnet,ye2023ip,mou2024t2i}, personalization and stylization of generated images \cite{DreamBooth, LoRA}, and 3D asset or scene generation \cite{DreamFusion, SceneTex, hollein2023text2room}.
However, in all cases the content is typically generated in shaded RGB space, that contains baked-in lighting effects (e.g., reflections, shadows, specular highlights).
This limits the usability of T2I models for many content creation scenarios such as gaming or VR applications, that requires PBR maps (albedo, roughness, metallic, normal) to render or relight scenes realistically.

Existing methods perform intrinsic image decomposition on RGB images \cite{InteriorVerse, IID, RGBX, careagaColorful}.
However, finding the correct decomposition to a given input image is a constrained task, usually causing over-smoothed or simplified predictions on out-of-domain samples. 
These methods are trained with synthetic conditioning input \cite{InteriorVerse, Openrooms2021}, leading to low-quality decompositions for out-of-distribution inputs, limiting their effectiveness on diverse real-world images.
Similarly, methods that generate 3D PBR content from T2I models \cite{AssetGen, PBRGen, qiu2024RichDreamer, huang2024MaterialAnything} are trained on object-scale datasets \cite{ObjaVerse, ObjaVerseXL}, making them unsuitable for large-scale 3D scenes.

We take a different approach for PBR map generation.
For the first time, we \textit{directly} generate PBR maps from text as input in a probabilistic diffusion process.
Since our method does not rely on an input image, it is more self-contained, enabling better generalization capabilities. 
We can use the generated PBR maps for downstream tasks, such as physically-based rendering, relighting, or material editing (\Cref{fig:teaser}).
We also showcase that our method can generate PBR textures for entire 3D scenes, for the first time to the best of our knowledge, making it directly usable for gaming/VR applications (\Cref{fig:exp:scenetex}).
Our method leverages the strong image prior of pretrained T2I models and converts it into a PBR map generator.
This way, our model can generate PBR content from diverse, out-of-distribution text prompts, similar to existing T2I models that operate in RGB space.
Concretely, we first train intrinsic priors for each material property and for normal map generation separately (\Cref{subsec:lora_training}).
We leverage small, curated datasets and the established LoRA \cite{LoRA} extension for T2I models.
Then, we fine-tune all priors jointly by employing cross-intrinsic attention in the diffusion transformer network (\Cref{subsec:alignment}).
This allows intrinsic properties to interact, enabling their joint and coherent generation.
We also introduce a rendering objective with importance-based lighting sampling to ground the intrinsic components.
This image-space signal encourages sharp and semantically meaningful decompositions.
In summary, our contributions:
\begin{itemize}[leftmargin=*,topsep=1pt, noitemsep]
    \item We introduce the first method, that \textit{directly} generates PBR images from text as input in a probabilistic diffusion process.
    In comparison to baselines, our PBR maps are of higher quality and can be used for various downstream tasks, including physically-based rendering, editing/relighting, and room-scale 3D scene PBR texturing.
    \item We decompose the strong image prior of pretrained T2I models into intrinsic components in a two-stage training process.
    This allows us to generate PBR maps from diverse text prompts, that are not limited to the distribution of existing, synthetic datasets.
    \item We combine cross-intrinsic attention with a novel rendering objective using importance-based light sampling to jointly generate semantically coherent PBR maps.
\end{itemize}

%% file: sections/2-relatedwork.tex


\section{Related Work}
\label{sec:relatedwork}

\paragraph{Text-to-Image Models}
Text-to-image (T2I) models have emerged as powerful tools for 2D content creation; they create high-quality, diverse images from only text as input \cite{rombach2022high, SaharCSLWDGAMLSHFN2022, ramesh2022hierarchical}.
Since their inception, several models further increased the visual quality of generated images \cite{podell2023sdxl, flux2023, xie2024sana, zhang2023text}.
These models are trained on datasets consisting of billions of images, like \cite{schuhmann2022laion}.
This makes them a strong 2D prior for arbitrary content generation.
They typically model the diffusion process following \citet{ho2020denoising} or \citet{lipman2022flow} with U-Net \cite{ronneberger2015u} or diffusion transformer (DiT) \cite{peebles2023scalable, vaswani2017attention} architectures.
Many downstream applications leverage T2I models, including controllable content generation \cite{zhang2023controlnet, ye2023ip, mou2024t2i, kocsis2024lightit, sharma2024alchemist} as well as personalization and stylization of generated images \cite{DreamBooth, LoRA, wang2023styleadapter, sohn2023styledrop}.
We leverage pretrained T2I models as prior for our task, the generation of PBR maps from text.

\vspace{-6pt}
\paragraph{Task-specific Finetuning of Text-to-Image Models}
In order to use T2I models for downstream tasks, different modifications to the model architecture exist and can be applied \cite{zhang2023controlnet, mou2024t2i, ye2023ip, liu2023zero, kocsis2024lightit}.
In particular, LoRA layers \cite{LoRA} can be used to teach T2I models about specific ``styles'' (e.g., artistic paintings).
Additional low-rank linear layers are trained in every attention block, which keeps the generalized prior of the T2I model, while finetuning on smaller-scale datasets.

We similarly finetune multiple LoRAs to teach a T2I model about the distribution of intrinsic images.

Other tasks generate multi-view image outputs, such as video generation \cite{wu2022tune, yang2023rerender} multi-view image generation \cite{hollein2024viewdiff, liu2023syncdreamer, Tang2023mvdiffusion} or multi-modal generation \cite{MTFormer}.
They augment the attention operation in the transformer architecture to jointly process multiple images in a batch with the same model.
Related to these tasks, we perform cross-intrinsic attention to generate aligned PBR maps in a single denoising forward pass with our finetuned model.

T2I models are also applied to 3D tasks, like object generation \cite{chen2023text2tex} or scene generation \cite{hollein2023text2room, SceneTex}.
Some methods finetune T2I models on synthetic 3D objects datasets, like \cite{ObjaVerse, ObjaVerseXL}, to generate object-scale 3D assets \cite{bensadoun2024meta, AssetGen, ARM2025}.
In contrast, we utilize score distillation \cite{DreamFusion, li2024flowdreamer} to generate PBR textures of entire 3D scenes following \citet{SceneTex}.

\vspace{-6pt}
\paragraph{Material Reconstruction}
Completely decoupling lighting from material properties requires multiple surface observations under different lighting conditions. 
Pret-trained models can enable material acquisition from sparse observations. 
\cite{SVBRDF} uses a feed-forward model to predict the texture of a single material and similarly use lighting sampling with the reflected view direction. Orthogonal to this, we aim to generate the PBR properties of complex scenes in image space to enable downstream applications. We introduce a rendering loss to the diffusion framework using a roughness-weighted importance sampling for the lighting direction.

The field of intrinsic image decomposition focuses on obtaining PBR maps from a single RGB image.
Early approaches focus on separating the reflectance from shading \cite{land1971lightness, horn1974determining, wu2023MAW} using various heuristics, such as sparsity in reflectance properties \cite{Entropy2004, shen2008NonLocal, grosse2009MITIntrinsics, zhang22SelfSimilarity}, or smoothness \cite{bell2014IIW}.
Later, deep-learning methods \cite{ComplexInvIndoor, IntrinsicLoRA, IID, RGBX, PBRGen, DiffusionRenderer2025} train decomposition networks on synthetic datasets, such as \cite{InteriorVerse}.
However, the decomposition of an RGB image into its intrinsic properties is a constrained task, making it hard to generalize to out-of-distribution input images.
In contrast, we directly generate all PBR components from text as input.
This drastically improves the performance on in-the-wild settings.

\vspace{-6pt}
\paragraph{Material Generation}
Recent works use diffusion models for single material generation, conditioned on text or image inputs \cite{MaterialPalette2024, MaterialPicker2025}. The work of \cite{Xue2025} concurrently addresses text-conditional complex PBR generation. They train a shared ControlNet \cite{zhang2023controlnet} and use a diffusion renderer for editability. In contrast, our method first learns a prior over the PBR properties independently, then aligns them with cross-intrinsic attention using a fixed renderer to ensure compatibility with standard rendering engines.

%% file: sections/3-method.tex
\vspace{-6pt}
\section{Method}
\label{sec:method}
\vspace{-6pt}

Our method generates the intrinsic properties of an image given a text prompt as input (\Cref{fig:teaser} top).
Specifically, we leverage the strong prior of a pretrained text-to-image model and turn it into a PBR map generator.
First, we learn the distribution of intrinsic properties (albedo, roughness, metallic, normal) by finetuning LoRA layers on each modality separately (\Cref{subsec:lora_training}).
Then, we learn the joint distribution by leveraging cross-intrinsic attention and by minimizing a novel rendering objective (\Cref{subsec:alignment}).
Our method generates multiple images corresponding to the different PBR maps, allowing for various downstream applications (\Cref{fig:exp:editable_image_generation}, \Cref{fig:exp:scenetex}).
We summarize our method in \Cref{fig:method:pipeline}.

\input{figures/method/pipeline}
\subsection{PBR Prior Training}
\label{subsec:lora_training}
In order to generate PBR maps of an image, we model the distribution of the individual intrinsic image properties.
Specifically, we model the probability distribution $p_{\theta}(\mathbf{X}_0)$ over data $\mathbf{X}_0 {\sim} q(\mathbf{X}_0)$, where $\mathbf{X}_0 {=} \{\mathbf{x}_{a} {\in} \mathbb{R}^{3 \times P}, \mathbf{x}_r {\in} \mathbb{R}^{P}, \mathbf{x}_m {\in} \mathbb{R}^{P}, \mathbf{x}_n {\in} \mathbb{R}^{3 \times P}\}$, $P {\coloneq} H \times W$ is shorthand for the image size, and the suffixes $a, r, m, n$ refer to the albedo, roughness, metallic, and normal intrinsic properties, respectively.
In other words, we learn the \textit{joint} probability distribution of all intrinsic properties through the parameters $\theta$ of a neural network.

Unfortunately, existing datasets, such as Openrooms \cite{Openrooms2021}, InteriorVerse \cite{InteriorVerse} or Hypersim \cite{roberts2021hypersim}, contain either only synthetic examples of intrinsic decompositions or are limited in size.
Thus, models trained on such datasets exhibit limited generalization to arbitrary, real-world examples.
On the other side, recent text-to-image diffusion models \cite{rombach2022high, podell2023sdxl, xie2024sana} are able to generate high-quality and diverse image samples.
These models learn the probability distribution $p_{\phi}(\mathbf{x}_0 | \mathbf{c}) {=} \int p_{\phi}(\mathbf{x}_{0{:}T} | \mathbf{c})d\mathbf{x}_{1{:}T}$ where $\mathbf{c}$ is a text condition, $\mathbf{x}_0 {\in} \mathbb{R}^{3 \times P} \sim q_{rgb}(\mathbf{x}_0)$ is sampled from billions of RGB images \cite{schuhmann2022laion}, and the latent variables $\mathbf{x}_{1{:}T} {=} \mathbf{x}_1, \ldots, \mathbf{x}_T$ gradually add more Gaussian noise to the data, following \cite{ho2020denoising}.
We leverage this strong image prior by turning pretrained diffusion models into PBR map generators.

In the first stage, we model the intrinsic image properties separately.
That is, we learn $p_{\phi, \theta_a}(\mathbf{x}_a)$ and $p_{\phi, \theta_n}(\mathbf{x}_n)$ corresponding to the albedo and normal maps, respectively.
Since roughness and metallic are both 1-channel properties, we concatenate them together with an additional 0-channel and learn $p_{\phi, \theta_{r,m}}(\mathbf{x}_r, \mathbf{x}_m)$.
This concatenation makes our samples compatible with the VAE, similarly as in \cite{IID}. 
Here, $\phi$ are the pretrained weights of the Flux.1-dev~\footnote{\url{https://huggingface.co/black-forest-labs/FLUX.1-dev}} model \cite{flux2023} and $\theta$ are the parameters of LoRA layers \cite{LoRA} injected into the MLP layers before and after the attention (to\_q, to\_k, to\_v, to\_out) module of all DiT blocks of the diffusion transformer model architecture \cite{peebles2023scalable}.
This is an established way to teach large text-to-image models about new concepts (e.g., our PBR map distribution), while retaining the ability to generate diverse samples \cite{han2024parameter}.
To this end, we curate a paired dataset of prompts and intrinsic properties and train the LoRA layers, while keeping the rest of the pretrained model frozen.
Precisely, we minimize the conditional flow matching loss \cite{lipman2022flow}:
\begin{align}
\label{eq:loss-cfm}
    \mathcal{L}_{\text{CFM}}(\theta_a) = \mathbb{E}_{t {\sim} \mathcal{U}(0,1), \epsilon {\sim} \mathcal{N}(\mathbf{0}, \mathbf{I})} \left[ || \hat{\mathbf{u}}_t(\mathbf{z}_t; t) - \mathbf{u}_t(\mathbf{x}_a; \epsilon) ||^2_2 \right]
\end{align}
where $\mathbf{x}_a {\sim} q(\mathbf{x}_a)$, $\mathbf{z}_t {=} (1 {-} t)\mathbf{x}_a {+} t\epsilon$ the noisy data at timestep $t$, $\mathbf{u}_t {=} \epsilon {-} \mathbf{x}_a$ the ground-truth vector field, and $\hat{\mathbf{u}}_t {=} \hat{\epsilon} {-} \hat{\mathbf{x}}_a$ its network prediction.

\paragraph{Dataset for albedo and normals}
Thanks to utilizing a pre-trained image prior, our method does not require extensive PBR datasets, which are generally not available. 
We collect as little as 20 synthetic examples of albedo and normal maps from the InteriorVerse dataset \cite{InteriorVerse}.
Then, we generate captions for each image with the Florence-2 model \cite{xiao2024florence} using the respective rgb renderings.
We train the LoRAs $\theta_a$ and $\theta_n$ on these text-image pairs and obtain high-quality results for diverse, out-of-distribution prompts.
This follows previous works, in which text-to-image models learn a new ``style'' of generated images given only a few example images \cite{wang2023styleadapter, sohn2023styledrop, ruiz2023dreambooth}.
We refer to the supplementary material for more details.

\paragraph{Dataset for roughness and metallic}
Similarly, we collect and caption samples for roughness and metallic properties.
However, we observe that training on a small dataset does not teach the model intricate details about the distribution of these PBR maps.
We hypothesize that this is because the data distribution of roughness/metallic is drastically different from RGB images and therefore requires more observations to learn.
To this end, we curate a large dataset of 20K roughness/metallic samples using the InteriorVerse dataset \cite{InteriorVerse}.
The resulting LoRA $\theta_{r,m}$ exhibits worse generalization capabilities than $\theta_a$ and $\theta_n$, i.e., it overfits to the indoor scene setup.
However, in \Cref{subsec:alignment}, we show how we can still turn $\theta_{r,m}$ into a generalized PBR generator by combining it with $\theta_a$ and $\theta_n$.

\subsection{PBR Prior Alignment}
\label{subsec:alignment}

After training the LoRAs separately in the first stage, we finetune all LoRA parameters together to learn the \textit{joint} distribution $p_{\phi, \theta_a, \theta_r, \theta_m, \theta_n}(\mathbf{X}_0)$.
At inference time, this allows us to sample aligned PBR maps across all modalities.
First, we replace self-attention with cross-intrinsic attention in every DiT block to facilitate communication between the different PBR maps.
Second, we propose a novel rendering objective that uses all generated PBR maps to create an RGB output image.
In the following, we detail both components.

\subsubsection{Cross-Intrinsic Attention}
\label{subsubsec:cia}
Inspired by multi-view diffusion methods \cite{liu2023syncdreamer, hollein2024viewdiff, Tang2023mvdiffusion, gao2024cat3d}, we leverage cross-attention in the DiT blocks to facilitate communication between batch elements.
We employ a batch-size of 3 and use one of the intrinsic LoRAs from the first stage training for each of the images, while sharing weights for all the other parts of the model.
We denote $\mathbf{q}_a^i, \mathbf{k}_a^i, \mathbf{v}_a^i$ as the query, key, and value features of the $i$-th DiT block for the batch element corresponding to the albedo image and similarly for the other intrinsic properties.
Then, we calculate cross-intrinsic attention as:
\begin{align}
\label{eq:cfa}
\mathbf{h}_a^i = \text{softmax}(\frac{\mathbf{q}_a^i [\mathbf{k}_{r,m}^i, \mathbf{k}_n^i, \mathbf{k}_a^i]^T}{\sqrt{d}})[\mathbf{v}_{r,m}^i, \mathbf{v}_n^i, \mathbf{v}_a^i]
\end{align}
where $[\cdot,\cdot]$ denotes concatenation along the sequence dimension and we omit the text feature for clarity.
We similarly calculate $\mathbf{h}_{r,m}^i$ and $\mathbf{h}_n^i$.
Finetuning all LoRA layers \textit{jointly} with cross-intrinsic attention allows us to generate aligned PBR maps of the same image.

Additionally, we employ dropout regularization to preserve the learned prior of the intrinsic LoRAs.
That is, with probability $p_i=0.25$ we calculate regular self-attention instead of cross-intrinsic attention in the $i$-th DiT block during training.
We show in \Cref{fig:exp:ablations}, that this yields PBR maps of higher quality with sharper details.

\subsubsection{RGB Rendering Loss}
\label{subsubsec:rend-loss}

\input{figures/method/light_sampling}

Cross-intrinsic attention allows us to generate aligned PBR maps of the same image.
However, individual intrinsics can still be of low quality (see \Cref{fig:exp:ablations}).
This is because all LoRAs are finetuned jointly, which encourages similar feature distributions during attention, i.e., the differences between the PBR maps are ``averaged out''.
To this end, we incorporate a novel rendering loss in the finetuning stage.
Its goal is to provide semantic guidance to the intrinsic properties, that is, it teaches how the PBR maps are combined, encouraging their distinct feature distributions.

Concretely, we render an RGB image from the predicted PBR maps.
First, we obtain the clean data samples as $\hat{\mathbf{z}}_0 {=} \mathbf{z}_t {-} t \hat{\mathbf{u}}(\mathbf{z}_t; t)$, where $\mathbf{z}$ denotes the batched data of all PBR maps.
Then, we decode them from the latent space with the VAE to obtain $\hat{\mathbf{X}}_0$.
We use the simplified Disney BRDF model \cite{burley2012physically} and interpret $\hat{\mathbf{X}}_0$ as the screen-space buffers of albedo, roughness, metallic, and normal properties.
Assuming a single directional light source, we can use deferred shading to obtain an RGB image as:
\begin{align}
\label{eq:rendering}
\hat{\mathbf{I}} = f(\omega_o; \omega_i; \hat{\mathbf{X}}_0) \cdot L_i \cdot(\hat{x}_n^T \omega_i)
\end{align}
where $f$ is the BRDF evaluation value, $\omega_o$ the viewing direction, and $\{L_i$, $\omega_i\}$ the intensity and direction of a single light source.
We determine the viewing direction $\omega_0$ using the camera intrinsics of the dataset (we find this still yields good results during inference).
Similarly, we obtain the ground-truth RGB image $\mathbf{I}$ by using the same light, but the PBR maps of the dataset.
Then, we calculate the rendering loss: 
\begin{align}
\label{eq:loss-rgb}
\mathcal{L}_{\text{rgb}}(\hat{\mathbf{I}}, \mathbf{I}) = || \hat{\mathbf{I}} - \mathbf{I} ||^2_2 + 0.1 \cdot\text{LPIPS}(\hat{\mathbf{I}}, \mathbf{I})
\end{align}
where $\text{LPIPS}$ denotes the perceptual loss \cite{johnson2016perceptual}.

We require light samples $\{L_i, \omega_i\}$ to render RGB images, following \Cref{eq:rendering}.
In practice, we sample a single directional light source per image and always use constant intensity $L_i {=} e^2$.
We employ importance sampling to obtain the direction of the light $\omega_i$ (see \Cref{fig:method:light_sampling}).
That is, we invert the generated roughness $\hat{\mathbf{x_r}} {\in} [0, 1]$ and use it as the weights for multinomial sampling of a pixel in the image.
Thus, pixels with \textit{lower} roughness are selected more often.
Then, we obtain the light direction as the reflected view direction $\omega_i {=} 2 \hat{\mathbf{x}}_n \langle\ \hat{\mathbf{x}}_n, \omega_o \rangle {-} \omega_o$, where $\omega_o$ is the viewing direction and $\hat{\mathbf{x}}_n$ the normal vector corresponding to the sampled pixel.
This way, we produce RGB images that contain specular highlights and therefore we obtain better gradients for the roughness and metallic LoRAs.
This helps to increase the quality of those PBR maps (\Cref{fig:exp:ablations}).

During the second finetuning stage, we sample 5 directional light sources in every iteration and render a separate RGB image with each of them.
The final loss then becomes $\mathcal{L} {=} \mathcal{L}_{\text{CFM}} {+} \sum_{i=1}^5 \mathcal{L}_{\text{rgb}}(\hat{\mathbf{I}}_i, \mathbf{I}_i)$.
We do not backpropagate $\mathcal{L}_{\text{rgb}}$ to the parameters $\theta_n$ of the normal LoRA, as we find it stabilizes the rendering quality by avoiding ambiguities between material and geometry.

%% file: figures/method/pipeline.tex
\begin{figure*}[t]
    \centering
    \includegraphics[width=\textwidth]{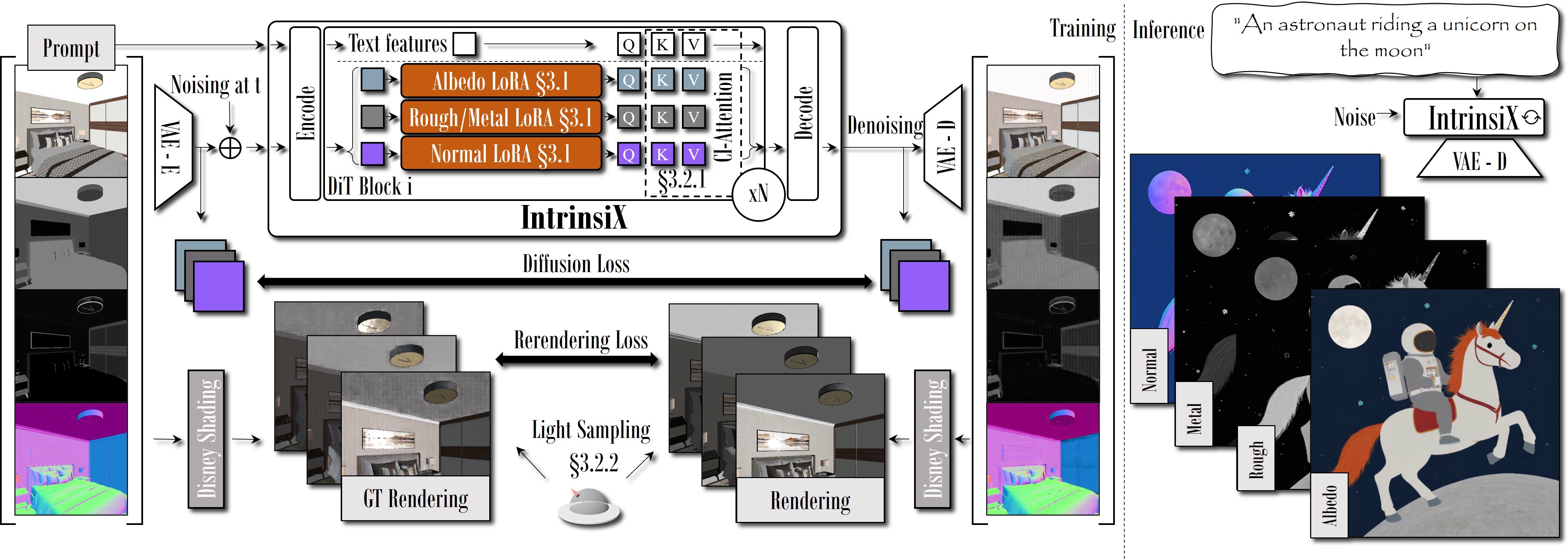}

    \caption{\textbf{Method Overview}. 
    We generate the intrinsic properties of an image given text as input.
    \textbf{Left:} we train 3 different LoRAs for a pretrained, latent text-to-image model, corresponding to the intrinsic properties (albedo, normal, and roughness + metallic) on curated synthetic datasets (\Cref{subsec:lora_training}).
    We facilitate communication between all 4 modalities through cross-intrinsic attention to predict PBR maps corresponding to the same image (\Cref{subsubsec:cia}).
    A novel rendering loss using importance-based light sampling ensures that we can render high-quality RGB images from physically realistic PBR maps (\Cref{subsubsec:rend-loss}).
    \textbf{Right:} after training, we jointly denoise and decode all 4 PBR maps and can prompt our model with diverse, out-of-distribution descriptions.
    }
    \label{fig:method:pipeline}
\end{figure*}

%% file: figures/method/light_sampling.tex
\begin{wrapfigure}{r}{0.5\textwidth}
    \centering
    \vspace{-36pt}
    \includegraphics[width=0.5\textwidth]{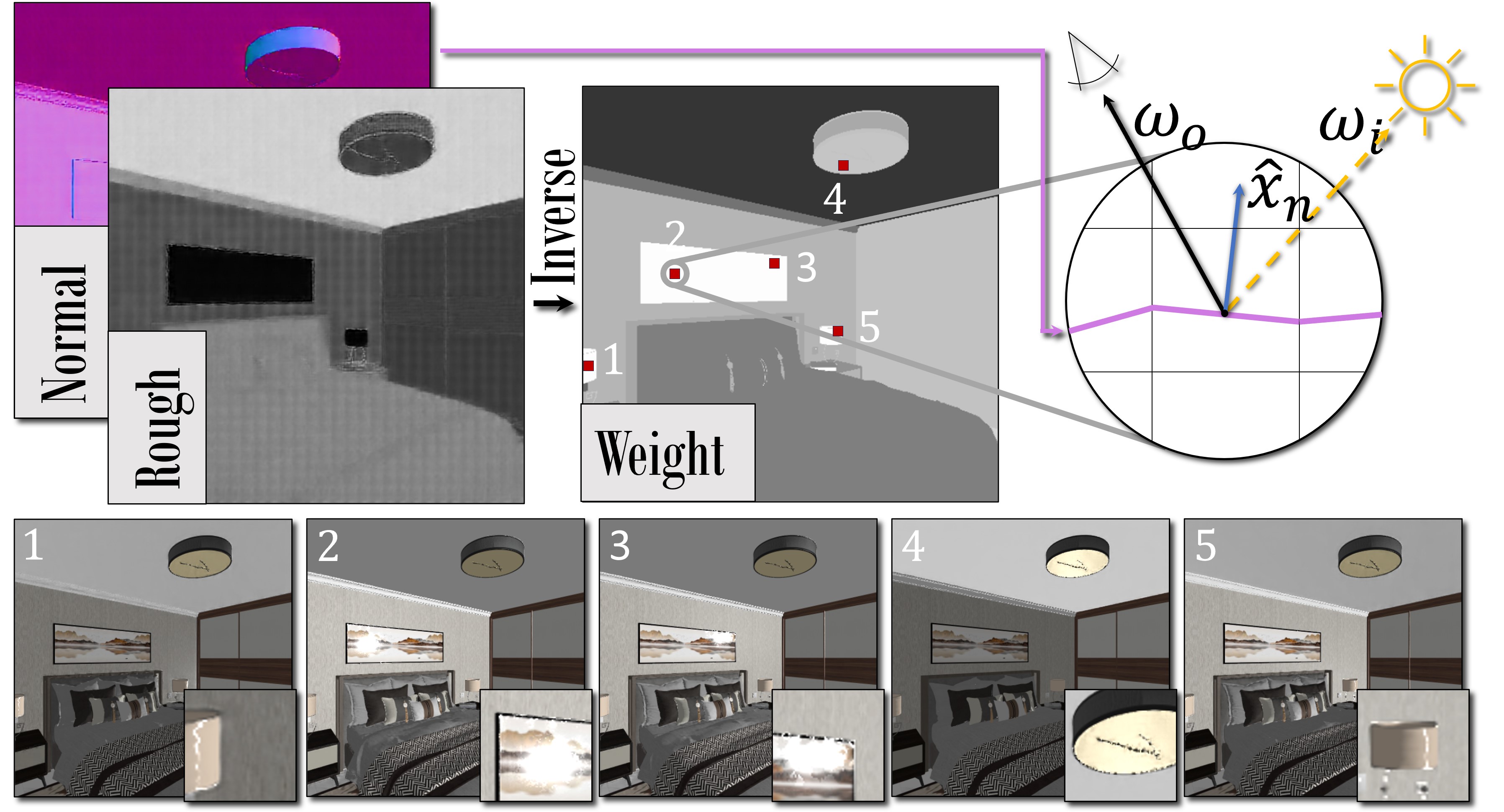}
    \caption{\textbf{Importance-based Light Sampling}. 
    We render RGB images (bottom) from our generated PBR maps and a sampled light source as input (top).
    We employ multinomial importance sampling based using the inverse roughness to select \textit{less} rough pixels more often (red squares).
    The light direction is then the viewing direction to the pixel reflected by its normal.
    The rendered images thus contain more specular effects, which provides better gradients during training.
    }
    \label{fig:method:light_sampling}
    \vspace{-24pt}
\end{wrapfigure}

%% file: sections/4-experiments.tex


\section{Applications}
\label{sec:experiments:applications}

\input{figures/applications/editable_image_generation}
\input{figures/applications/scenetex}

Since we directly generate PBR maps, we can utilize standard computer graphics pipelines for physically-based rendering to produce RGB renderings.
This allows for various downstream tasks.

\vspace{24pt}
\paragraph{Editable Image Generation}
We select a directional light source during rendering of an RGB image from our PBR maps (see \Cref{eq:rendering}).
Since our model produces PBR maps, we can vary the direction of the light source arbitrarily and render them under numerous lighting conditions. 
Similarly, we can manually edit the individual PBR maps, e.g., by changing the albedo color of individual objects or by making them more specular.
We show two examples in \Cref{fig:exp:editable_image_generation} and \Cref{fig:teaser}.
Note that our PBR maps are not restricted to a single lighting direction. 
This can enable artists to precisely tune the appearance of our generated images to their individual needs and therefore make the generations more useful for practical applications.

\paragraph{PBR Scene Texturing}
We can use our method to perform 3D scene PBR texturing (\Cref{fig:exp:scenetex}).
Recently, pretrained text-to-image models have been used as prior to distill information in 3D \cite{DreamFusion, AssetGen, SceneTex}.
We apply the SceneTex approach \cite{SceneTex}, but use our fine-tuned PBR model instead of an RGB model.
This enables us to distill \textit{uv}-textures for a given geometry corresponding to the individual intrinsic components.
We can then render, relight, and edit an entire 3D scene according to physically-based rendering frameworks (see \Cref{fig:exp:scenetex}).
This shows the potential of \textit{direct} PBR map generation using AI-generated environments for games or VR applications.

SceneTex \cite{SceneTex} requires a conditional generator. 
To this end, we finetune our model for 4K iterations after the first stage as described in \Cref{subsec:alignment}.
Additionally, we randomly (with probability $p {=} 0.25$) set one of the PBR maps to the ground truth and the corresponding timestep to $t {=} 0$.
This enables our model to be conditioned on any PBR input, similar to \cite{hollein2024viewdiff}.
We render normal maps of the geometry from different viewpoints to condition our PBR generation.
In the first stage, we optimize for the material properties, conditioned on the rendered normal. 
Since our model is based on Flow Matching \cite{flux2023}, we also modify the distillation in SceneTex \cite{SceneTex}.
Concretely, we use the VFDS loss \cite{li2024flowdreamer} in image space.
To avoid over-smoothed results, we use $\text{CFG}{=}10$ and normalize the flow direction. 
We backpropagate to separate \textit{uv}-textures for each property and follow the weighting scheme of \cite{huang2023dreamtime}.
We weight the loss with the observation frequency and represent the textures with a regularized multi-resolution Laplacian-pyramid to stabilize the updates for sparsely observed regions. 
In the second stage, we similarly optimize normal textures for fine geometric details, conditioned on the already obtained material properties. 
We represent the normal map in tangent space and regularize with the original geometry.
For more samples and details, see the supplement.

\section{Experiments}
\label{sec:experiments}

\paragraph{Training and Testing Details}
In the first stage, we train the LoRAs separately for 2K iterations with a batch size of 10, which takes 5h on a single NVIDIA A100 (80GB) GPU.
In the second stage, we finetune for another 2.5K iterations with a batch size of 30 (10 aligned PBR maps), which takes 21h.
We employ the Prodigy optimizer \cite{mishchenko2023prodigy} in both stages.
The LoRA layers use a rank of 64, which gives a total of 224M additional parameters.
For inference, we use a single NVIDIA A6000 (48GB) GPU. Sampling a single image takes around 12 seconds. 

\paragraph{Rendering Images}

\input{figures/experiments/comparisons}

\input{tables/experiments/baselines}

During inference, we render RGB images following \Cref{eq:rendering} to obtain $\hat{\mathbf{I}}$.
We use a slightly higher lighting intensity ($L_i {=} e^3$) than during training.
Then we add an ambient color term: $\hat{\mathbf{I}}_{\text{amb}} = (1 {-} \alpha) \hat{\mathbf{I}} + \alpha \hat{\mathbf{x}}_a$ with $\alpha {=} 0.2$.
Afterwards, we apply the tone mapping from \cite{kalantari2017deep}: $\hat{\mathbf{I}}_{\text{tone}} {=} \text{log}(1 + \mu \hat{\mathbf{I}}_{\text{amb}}) {/} \text{log}(1 + \mu)$ with $\mu {=} 64$.
We empirically find this creates visually more pleasing RGB images.
This also demonstrates the advantages of generating intrinsic image properties, i.e., we can arbitrarily render them post-generation.
We list the used text prompts in the supplemental. 

\paragraph{Baselines}
To the best of our knowledge, we are the first method to perform \textit{direct} PBR map generation (from only text as input).
Therefore, we compare our method against recent methods that perform intrinsic image decomposition, namely \textit{IID} \cite{IID}, \textit{RGBX} \cite{RGBX}, and \textit{ColorfulShading} \cite{careagaColorful}.
In contrast to our method, these works require an RGB image as input from which the PBR maps are generated.
Unless noted otherwise, we generate the RGB image for the baselines by prompting our pretrained text-to-image model \cite{flux2023}.
We only compare albedo quality against \textit{ColorfulShading} \cite{careagaColorful}, since they are decomposing an image into albedo and shading components, which does not allow for complete relighting (including specular effects) or editing effects.

\paragraph{Metrics}
We measure the quality of generated PBR maps through various metrics.
First, we calculate the FID score \cite{heusel2017gans} on in-distribution and out-of-distribution albedo images.
For in-distribution (A-ID-FID), we use all $2595$ albedo images from the InteriorVerse \cite{InteriorVerse} test set and caption them based on the corresponding renderings with Florence-2 \cite{xiao2024florence}.
For each caption, we generate an albedo image, creating a total of $2595$ generated albedo images.
For out-of-distribution (A-OOD-FID), we evaluate on the pre-processed G-Buffer renderings \cite{qiu2024RichDreamer} of ObjaVerse \cite{ObjaVerse} (GObjaVerse). We take 1000 samples from the diverse ``Daily-Used" category.
As before, we generate an albedo map for each of the prompts, creating a total of 1000 generated albedo images. 
In both cases, we calculate FID against the respective ground-truth albedos.

Evaluating generated PBR maps remains a hard problem.
To this end, we also conduct a user study and ask to rate the quality of albedo (A-PQ), specularity (S-PQ), rendered images (R-PQ), and the prompt coherence (PC).
In total, we collect 2,274 data points from 36 participants and report averaged results (we refer to the supplementary material for more details).

\subsection{Intrinsic Image Generation}
\label{sec:experiments:intrinsic_image_generation}

\input{figures/experiments/albedo_comparisons} 

We show qualitative comparisons against IID \cite{IID}, RGBX \cite{RGBX} and \cite{careagaColorful} in \Cref{fig:exp:comparisons} using text prompts from \cite{gao2024cat3d}, LLM-generated ones, and our own prompts.
The baselines receive an RGB image as input, which was created with our pretrained text-to-image model, whereas we directly generate the PBR maps from only text as input.
For a fair comparison, we also train another variant of IID on the same small dataset and same architecture as ours, i.e. we use LoRA fine-tuning of FLUX \cite{flux2023}. 
All methods showcase similar diversity, i.e., the generated images align well with the out-of-distribution text prompts.
This showcases that our finetuned model still retains the generalized prior, which is also confirmed in the user study (\Cref{tab:exp:comparisons}, PC).
Furthermore, our generated PBR maps are of higher quality, semantically more meaningful, and they closer resemble the expected distribution for physically-based rendering.
This is because the baseline methods are trained on synthetic, indoor scenes \cite{InteriorVerse} and are not designed to generalize their decomposition to out-of-distribution setups.
Furthermore, intrinsic image decomposition is constrained to match the appearance of the input image, making it difficult to rather focus on the PBR distribution for out-of-domain samples. 
Additional albedo comparisons in \Cref{fig:exp:albedo_comparisons} as well as the quantitative comparisons in \Cref{tab:exp:comparisons} confirm this observation.
Our generated albedos are not oversmoothed, showing sharp details with flat colors. 
We provide more samples in the supplemental.

\subsection{Ablations}
\label{sec:experiments:ablations}

The main technical contributions of our method are the cross-intrinsic attention (\Cref{subsubsec:cia}) and the rendering loss (\Cref{subsubsec:rend-loss}).
In the following, we highlight the importance of each component.
We provide additional ablations in our supplementary material. 

%
%
\paragraph{How important is the rendering loss?}
The rendering loss improves the quality of \textit{all} PBR maps (see \Cref{fig:exp:ablations} and \Cref{tab:exp:comparisons}).
The additional supervision of \Cref{eq:loss-rgb} provides more diverse gradients to the LoRA weights than the L2 loss of \Cref{eq:loss-cfm}.
This way, the influence of the loss on the individual PBR maps is different and becomes grounded in image space through the rendering function, \Cref{eq:rendering}.
This leads to a better separation of the intrinsic properties, giving meaningful normal maps, detailed albedos without baked-in lighting effects, and sharper roughness/metallic maps without undesired texture or lighting patterns.
Our importance-based light sampling strategy further improves the sharpness of roughness and metallic maps.
In comparison, sampling light directions uniformly renders specular effects less often.
This results in less pronounced PBR maps in \Cref{fig:exp:ablations}.

\paragraph{How important is the dropout in Cross-Intrinsic Attention?}

\input{figures/experiments/ablations}

Without cross-intrinsic attention, we cannot create aligned PBR maps, because then there is no communication between batch elements during inference (see supplementary material).
Additionally, we utilize dropout regularization on our cross-intrinsic attention.
This technique motivates the model to preserve the prior of each intrinsic component during the 2nd stage alignment training. 
As can be seen in \Cref{fig:exp:ablations} and \Cref{tab:exp:comparisons}, this increases the quality of both the rendered images and the PBR maps.
The generated samples are sharper and do not suffer from noisy artifacts. 

\paragraph{Can a generalized PBR prior help intrinsic image decomposition?}

We believe that intrinsic image decomposition methods are constrained to match the appearance of the input image. Thus, the model can learn to rely more on the input, instead of focusing on staying in the PBR distribution; therefore, facing an out-of-distribution input image poses a challenge to these models to produce faithful PBR maps, corresponding to the respective distributions.
On the contrary, our model is trained with the key constraint to produce faithful PBR maps enabling better generalization.
To verify this hypothesis, we fine-tune our model to make it image-conditional. 
We extend the set of input modalities with the rgb rendering and apply dropout similarly how we achieve a normal conditional variant for the room-scale scene texturing \cref{sec:experiments:applications}. 
The resulting model can be conditioned on an input image during inference, similar to the baselines. 
This variant achieves $70.51$ FID on the out-of-domain GObjaverse evaluation set (A-OOD-FID), outperforming direct intrinsic image decomposition training (IID w/ FLUX-LoRA), and even our base model.
This shows that a pre-trained prior that directly models the underlying distribution is beneficial for generalization.


%% file: figures/applications/editable_image_generation.tex
\begin{figure*}[t]
    \centering
        \centering
    \setlength\tabcolsep{1.25pt}
    \resizebox{\textwidth}{!}{
    \fboxsep=0pt
    \begin{tabular}{ccccccccc}
         &
         0\textdegree \tikzmark{a} & 
         & 
         & 
         & 
         & 
         & 
         & 
         \tikzmark{b} 315\textdegree \\ 

        \rotatebox{90}{Relighting}
        &
        \fbox{\includegraphics[width=0.15\textwidth]{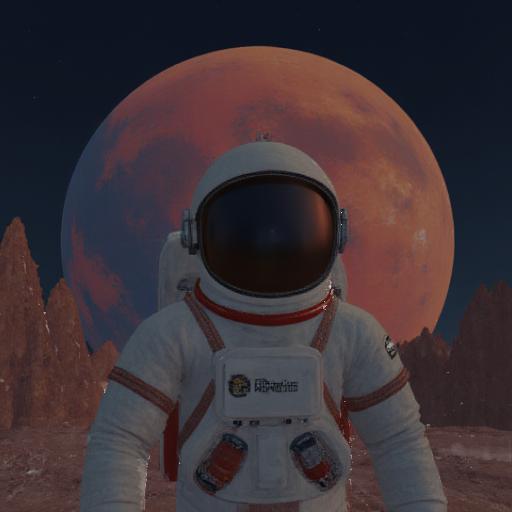}} 
        &
        \fbox{\includegraphics[width=0.15\textwidth]{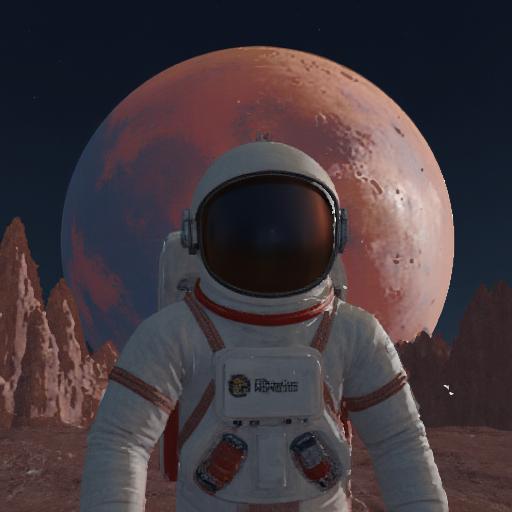}} 
        &
        \fbox{\includegraphics[width=0.15\textwidth]{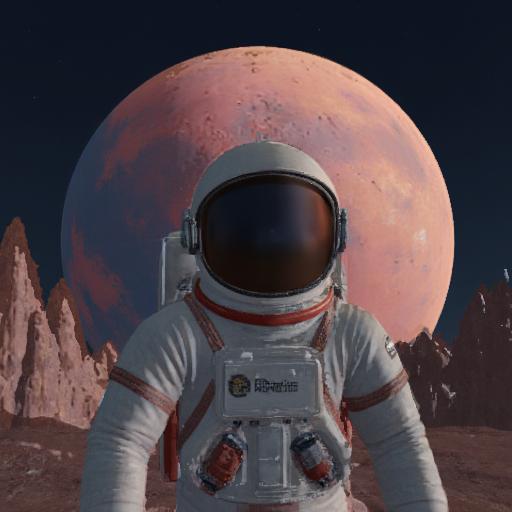}} 
        &
        \fbox{\includegraphics[width=0.15\textwidth]{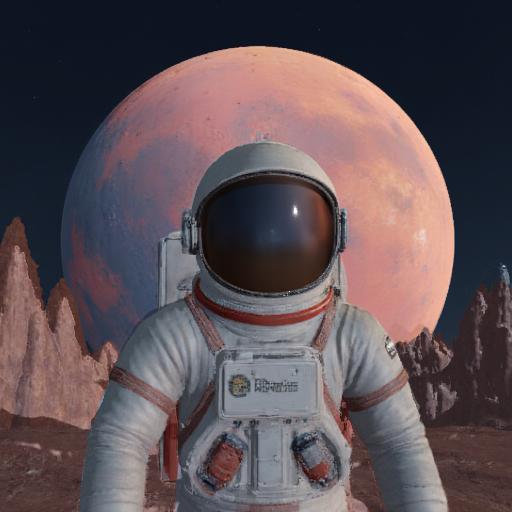}} 
        &
        \fbox{\includegraphics[width=0.15\textwidth]{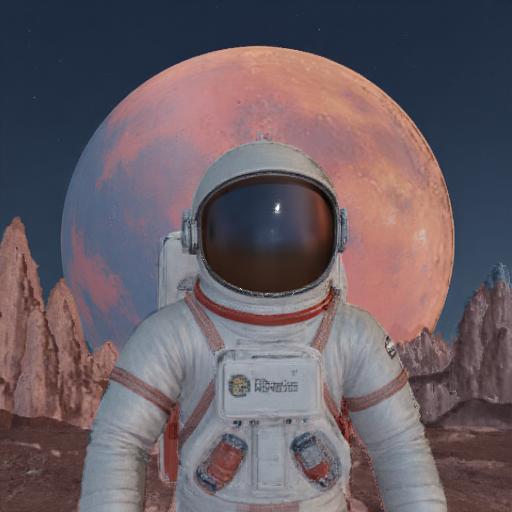}} 
        &
        \fbox{\includegraphics[width=0.15\textwidth]{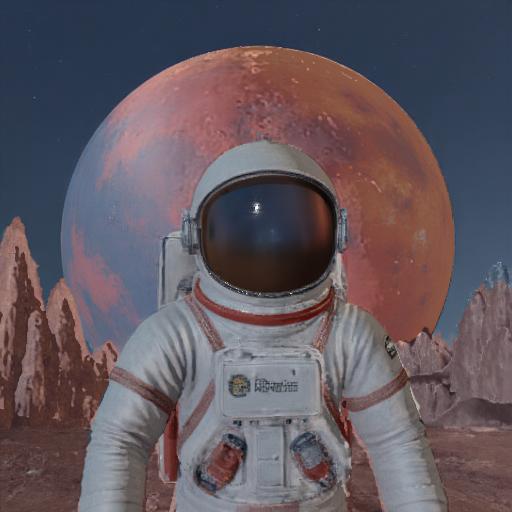}} 
        &
        \fbox{\includegraphics[width=0.15\textwidth]{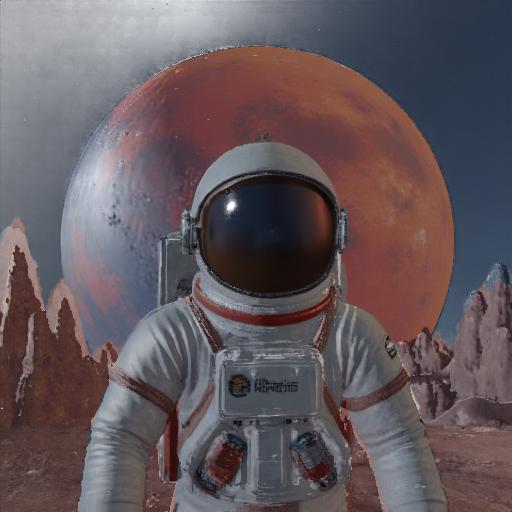}} 
        &
        \fbox{\includegraphics[width=0.15\textwidth]{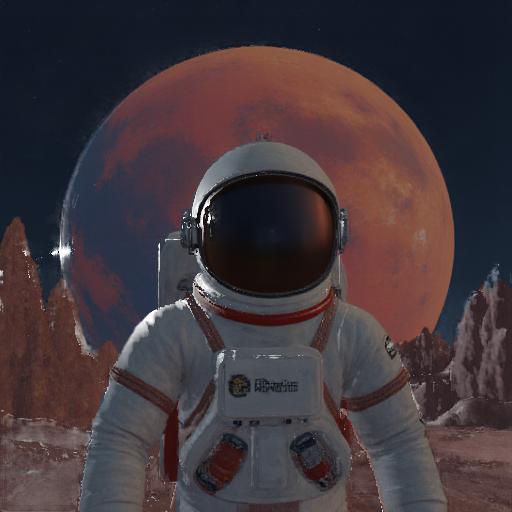}} 
        \\

        \rotatebox{90}{Albedo Edit}
        &
        \fbox{\includegraphics[width=0.15\textwidth]{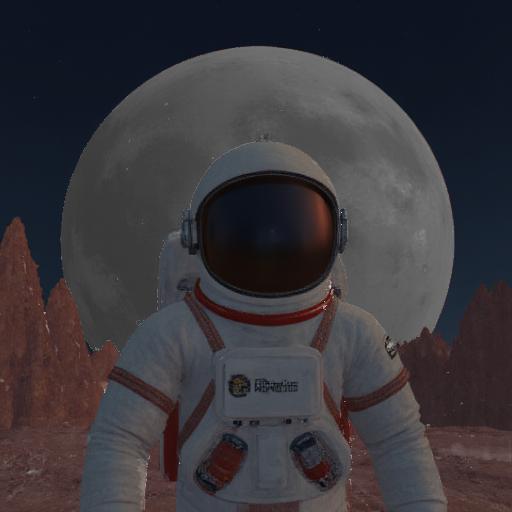}} 
        &
        \fbox{\includegraphics[width=0.15\textwidth]{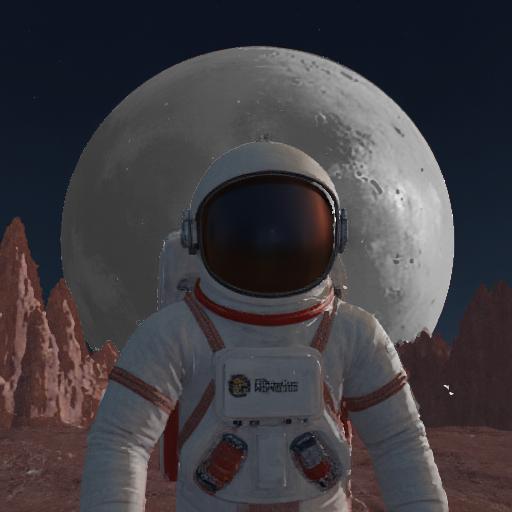}} 
        &
        \fbox{\includegraphics[width=0.15\textwidth]{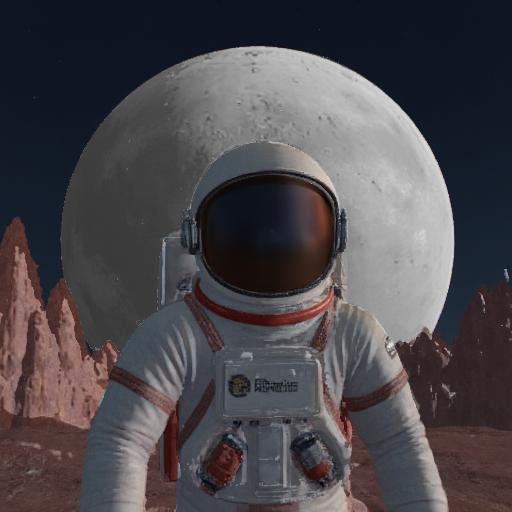}} 
        &
        \fbox{\includegraphics[width=0.15\textwidth]{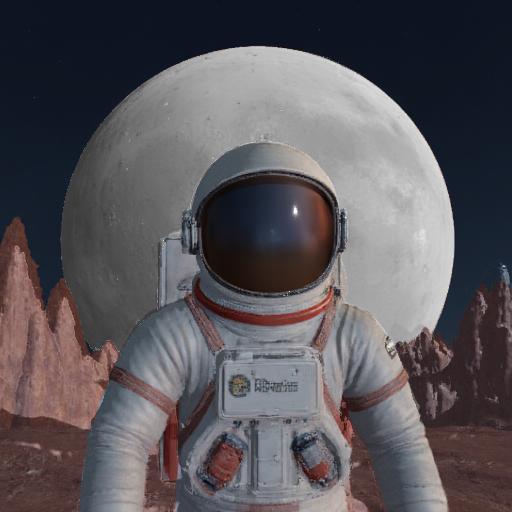}} 
        &
        \fbox{\includegraphics[width=0.15\textwidth]{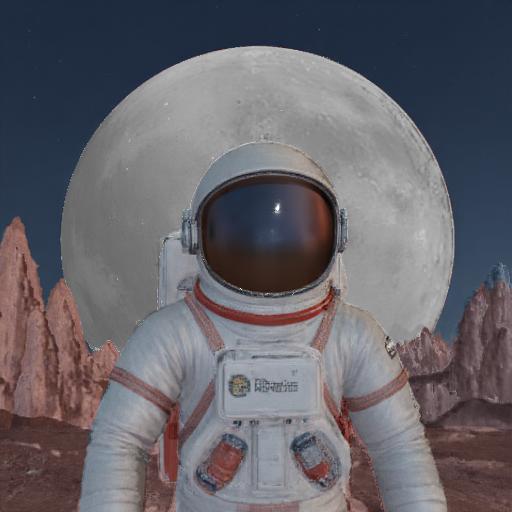}} 
        &
        \fbox{\includegraphics[width=0.15\textwidth]{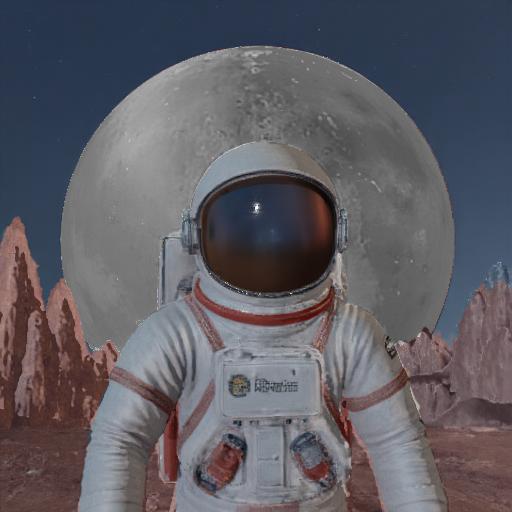}} 
        &
        \fbox{\includegraphics[width=0.15\textwidth]{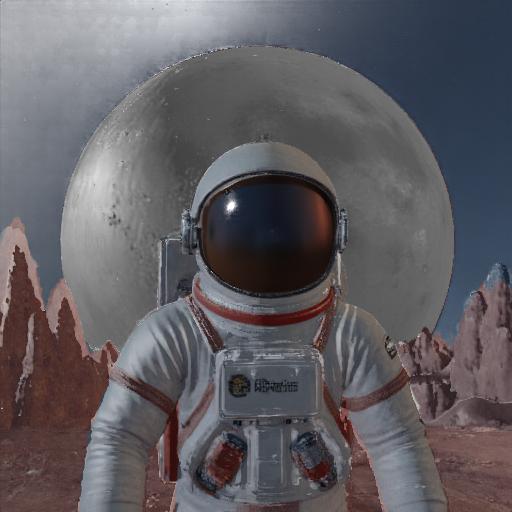}} 
        &
        \fbox{\includegraphics[width=0.15\textwidth]{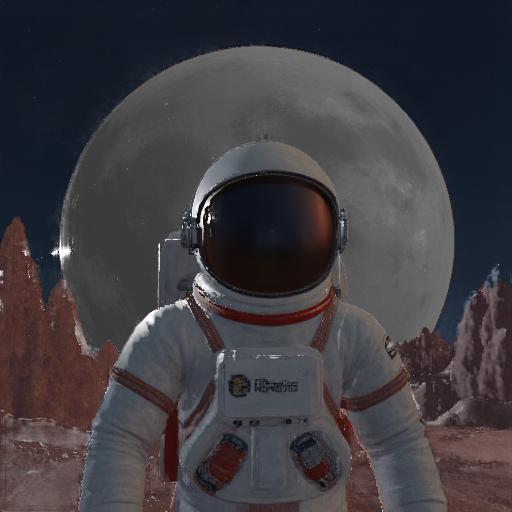}} 
        \\


        \rotatebox{90}{Specular Edit}
        &
        \fbox{\includegraphics[width=0.15\textwidth]{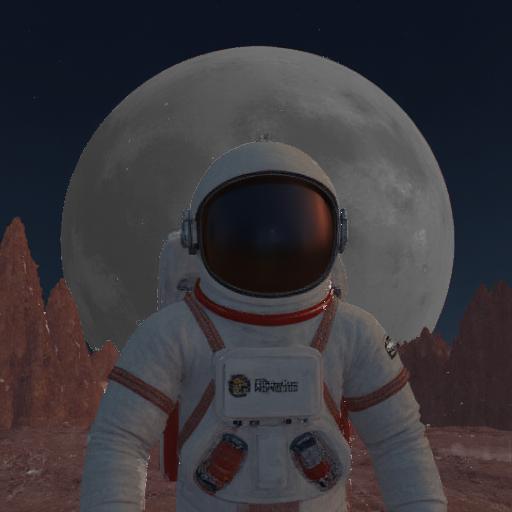}} 
        &
        \fbox{\includegraphics[width=0.15\textwidth]{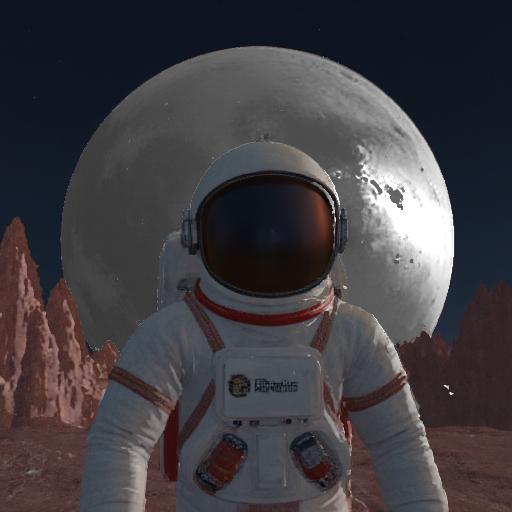}} 
        &
        \fbox{\includegraphics[width=0.15\textwidth]{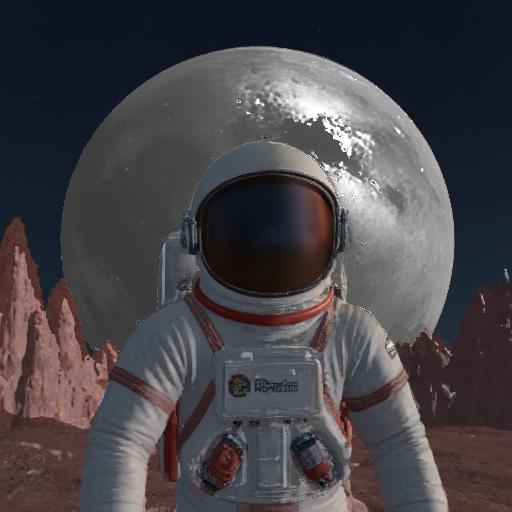}} 
        &
        \fbox{\includegraphics[width=0.15\textwidth]{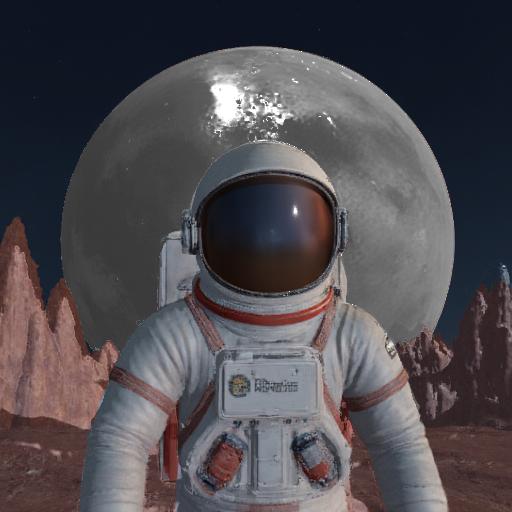}} 
        &
        \fbox{\includegraphics[width=0.15\textwidth]{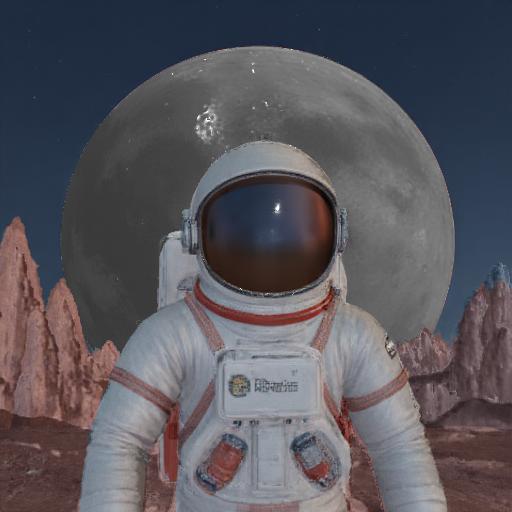}} 
        &
        \fbox{\includegraphics[width=0.15\textwidth]{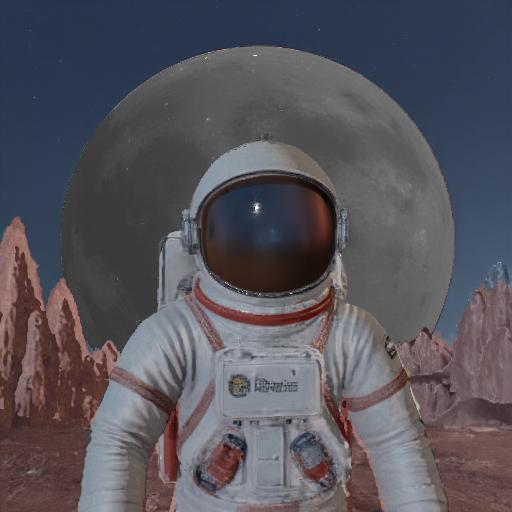}} 
        &
        \fbox{\includegraphics[width=0.15\textwidth]{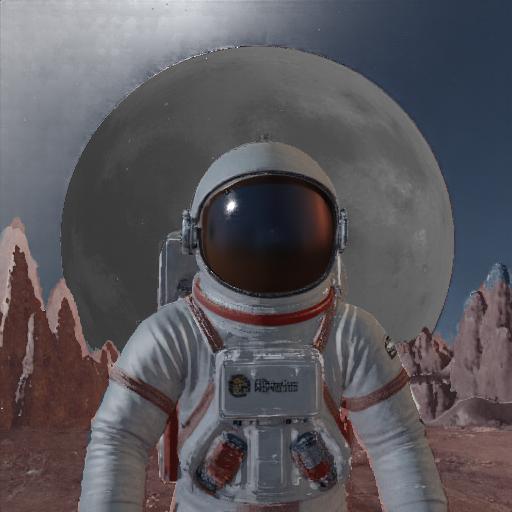}} 
        &
        \fbox{\includegraphics[width=0.15\textwidth]{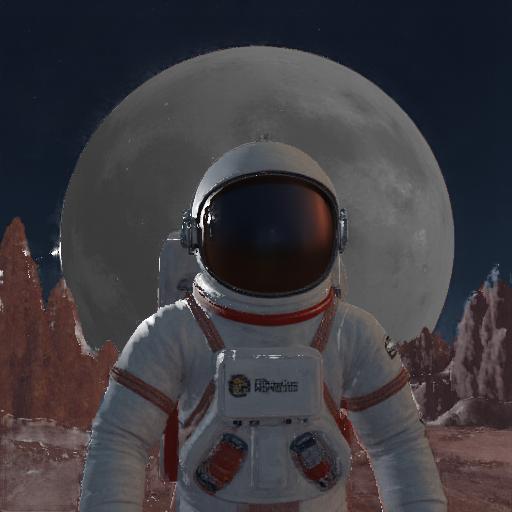}} 
        
    \end{tabular}
    \begin{tikzpicture}[overlay,remember picture]
    \draw[->,black,thick] ([yshift=0.15cm]pic cs:a) -- ([yshift=0.15cm]pic cs:b);
    \end{tikzpicture}
    }
    \vspace{12pt}
    \caption{\textbf{Editable Image Generation}. 
    Our generated PBR maps can be edited and utilized in standard physically-based rendering frameworks to produce RGB renderings.
    Here, we place a light source on top of the scene at constant elevation and rotate it around the vertical axis.
    From top to bottom we show,
    \textbf{(1):} RGB renderings with different light source positions;
    \textbf{(2):} manual edit of the albedo (desaturate the moon color);
    \textbf{(4):} lower roughness and higher metallic value (more glossy, mirror-like reflections).
    }
    \label{fig:exp:editable_image_generation}
\end{figure*}

%% file: figures/applications/scenetex.tex
\begin{figure*}[t]
    \centering
    \setlength\tabcolsep{0.5pt}
    \renewcommand{\arraystretch}{0.75} 
    \resizebox{\textwidth}{!}{
    \fboxsep=0pt
    \begin{tabular}{ccccc|c}
        \multirow{2}{*}[48pt]{\includegraphics[width=0.24\textwidth]{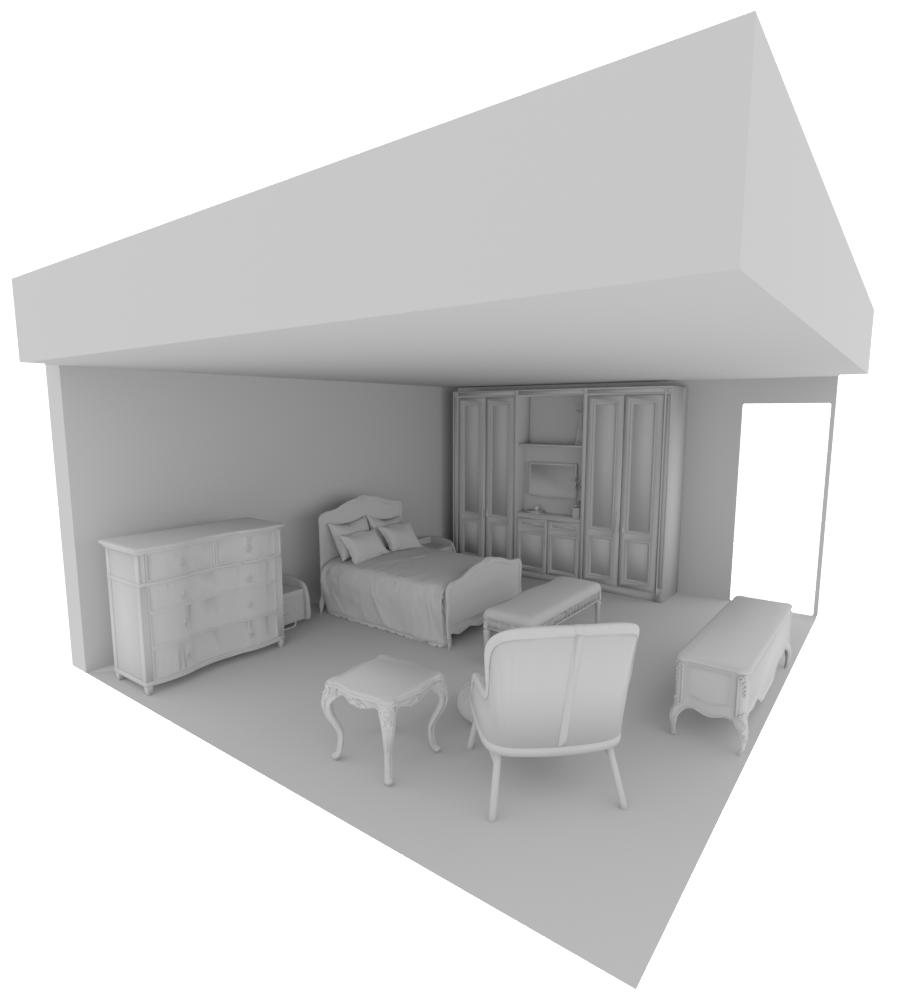}}
        &
        \rotatebox{90}{{\footnotesize View 1}}
        &
        \fbox{\includegraphics[width=0.24\textwidth]{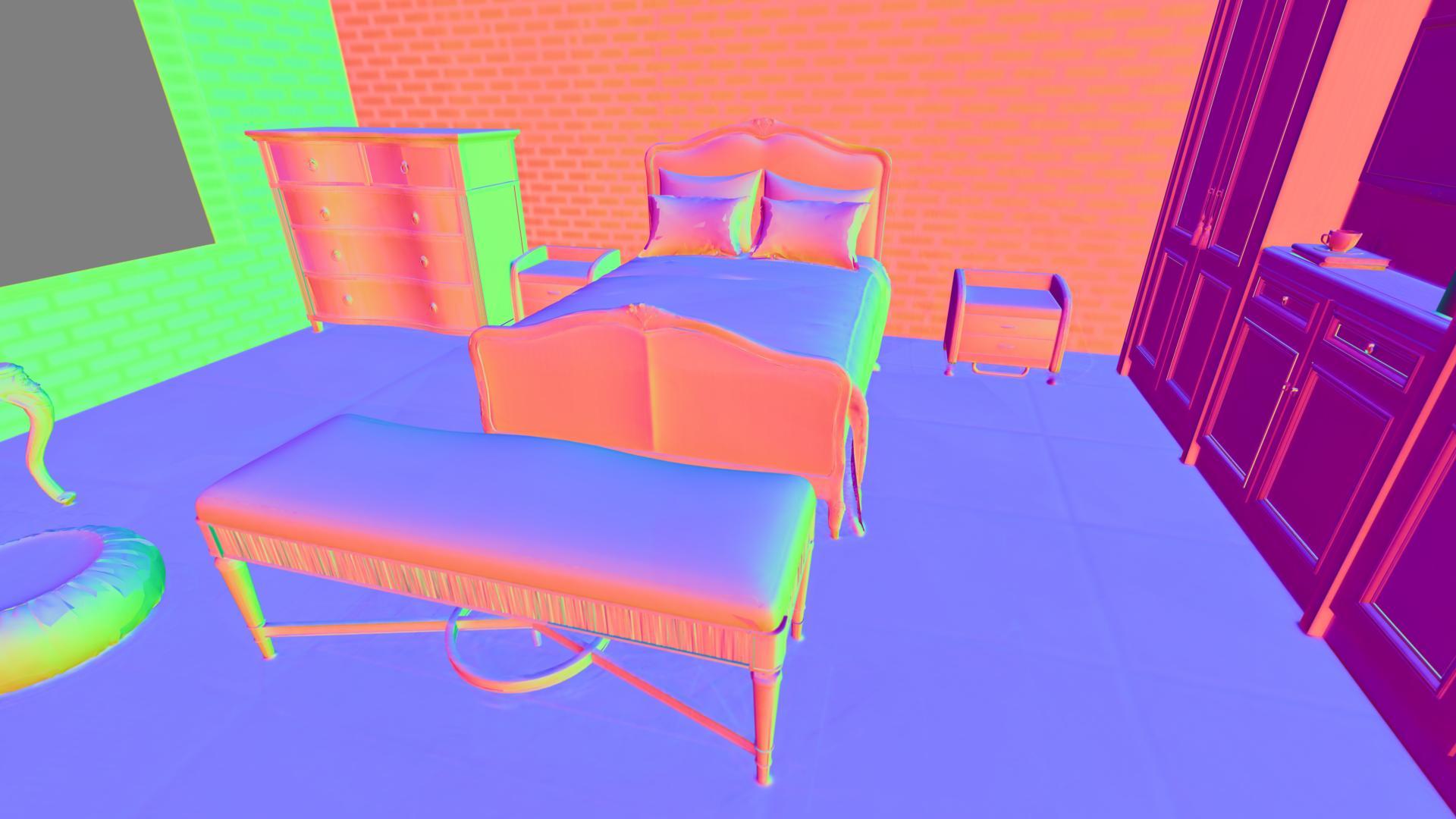}} 
        &
        \fbox{\includegraphics[width=0.24\textwidth]{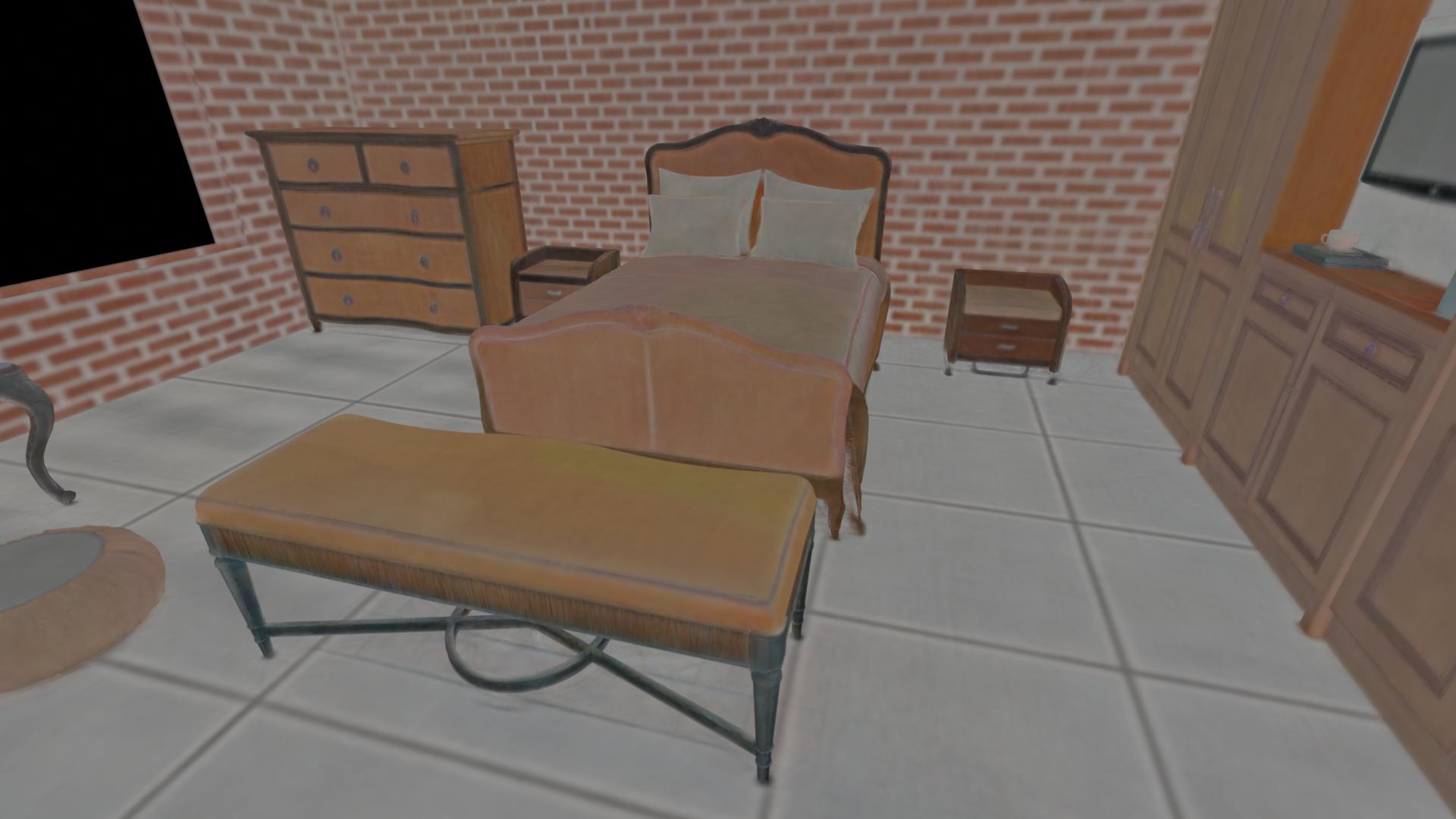}} 
        &
        \begin{tabular}[b]{@{}c@{}}
            \fbox{\includegraphics[width=0.1135\textwidth]{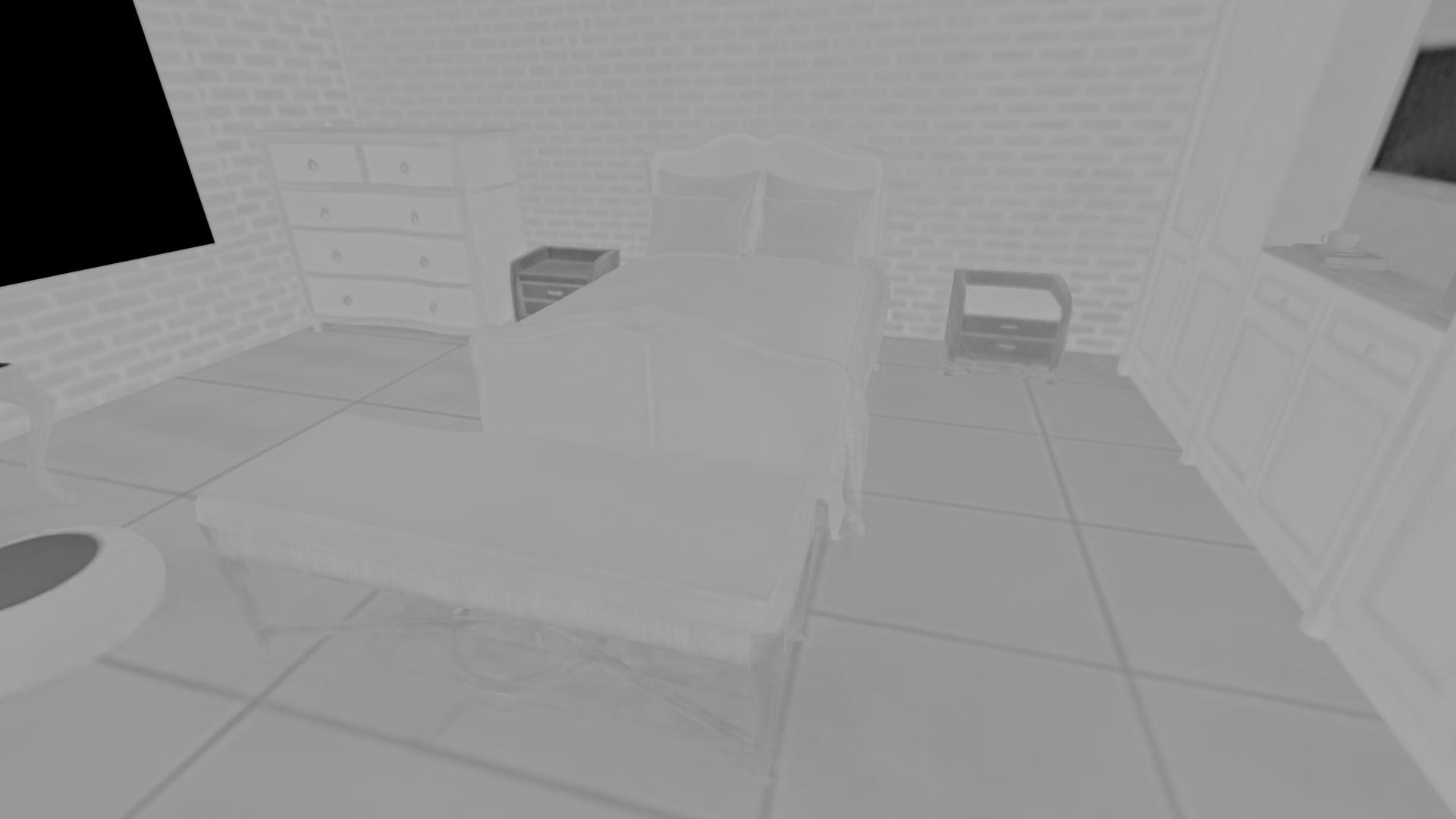}} 
            \\
            \fbox{\includegraphics[width=0.1135\textwidth]{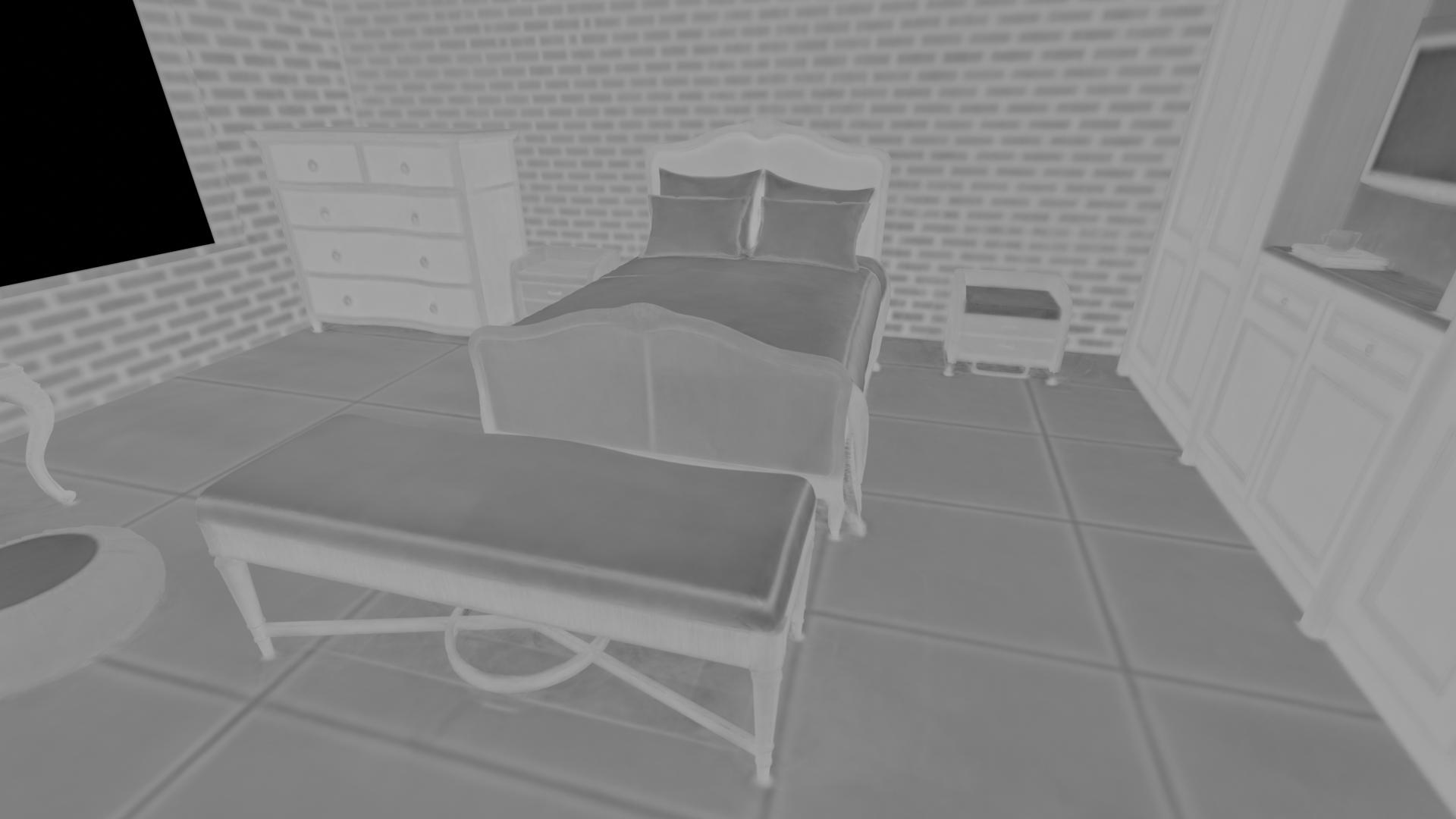}} 
        \end{tabular}        
        &
        \fbox{\includegraphics[width=0.24\textwidth]{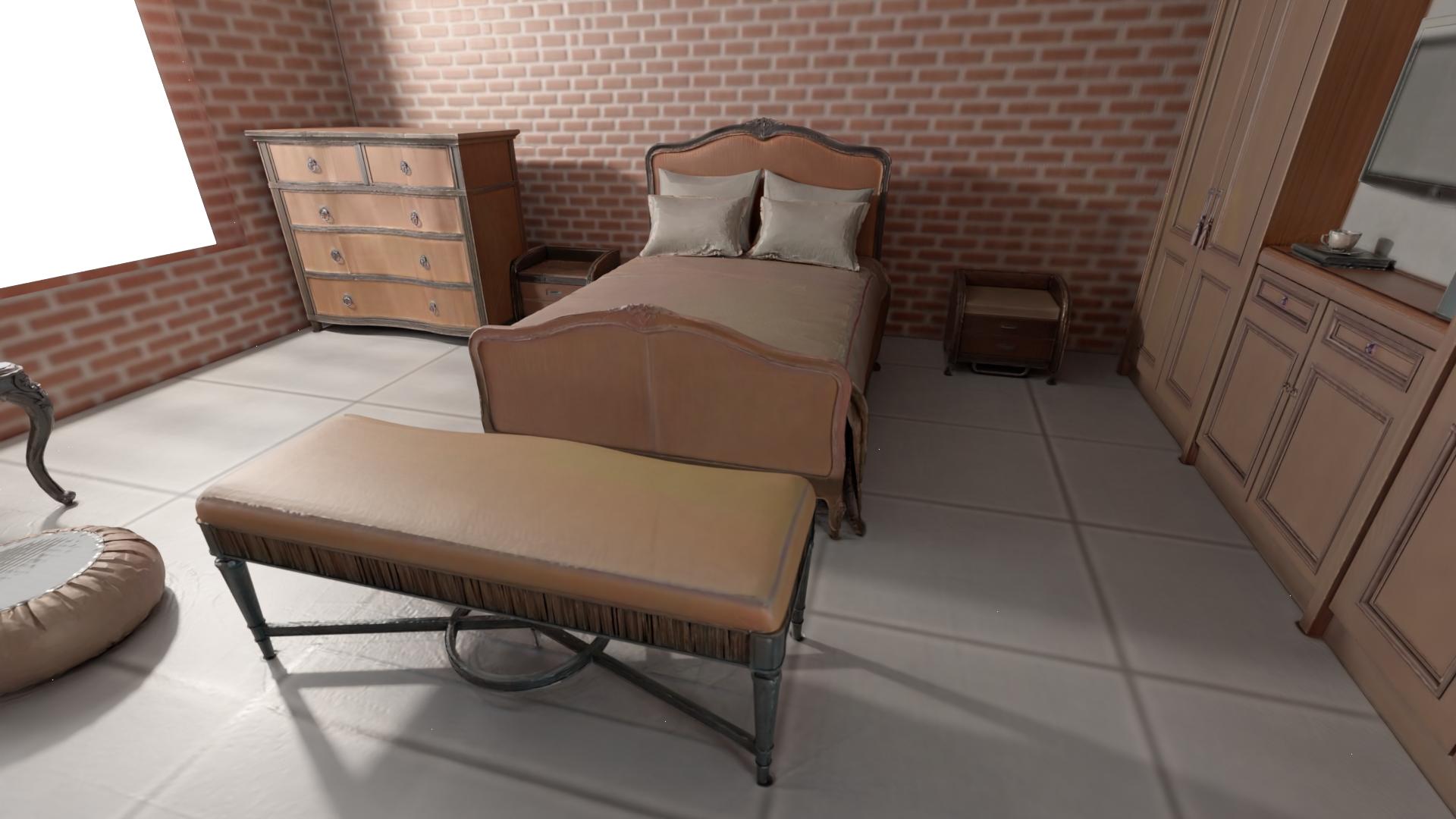}} 
        \\
        

        &
        \rotatebox{90}{{\footnotesize View 2}}
        &
        \fbox{\includegraphics[width=0.24\textwidth]{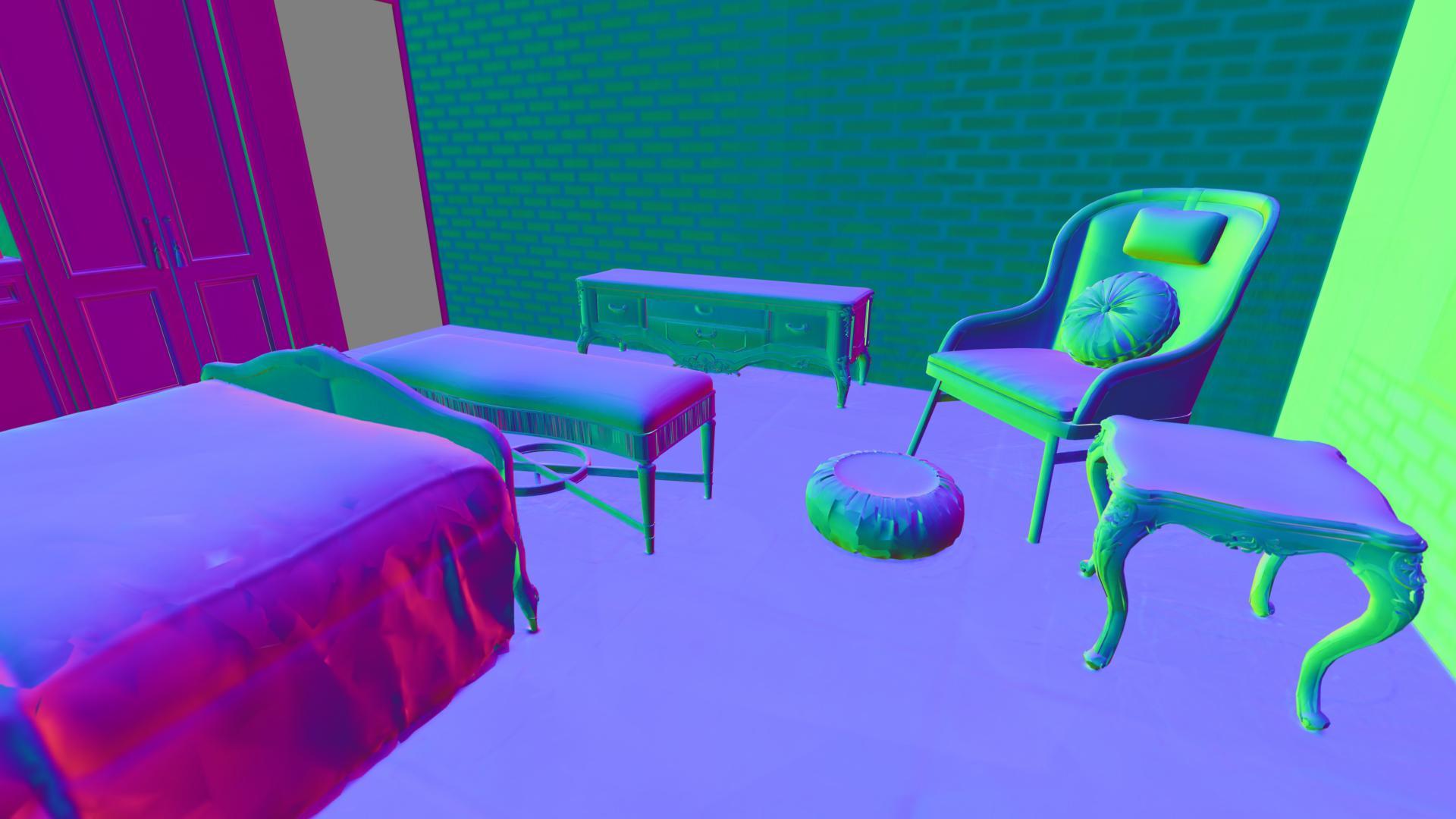}} 
        &
        \fbox{\includegraphics[width=0.24\textwidth]{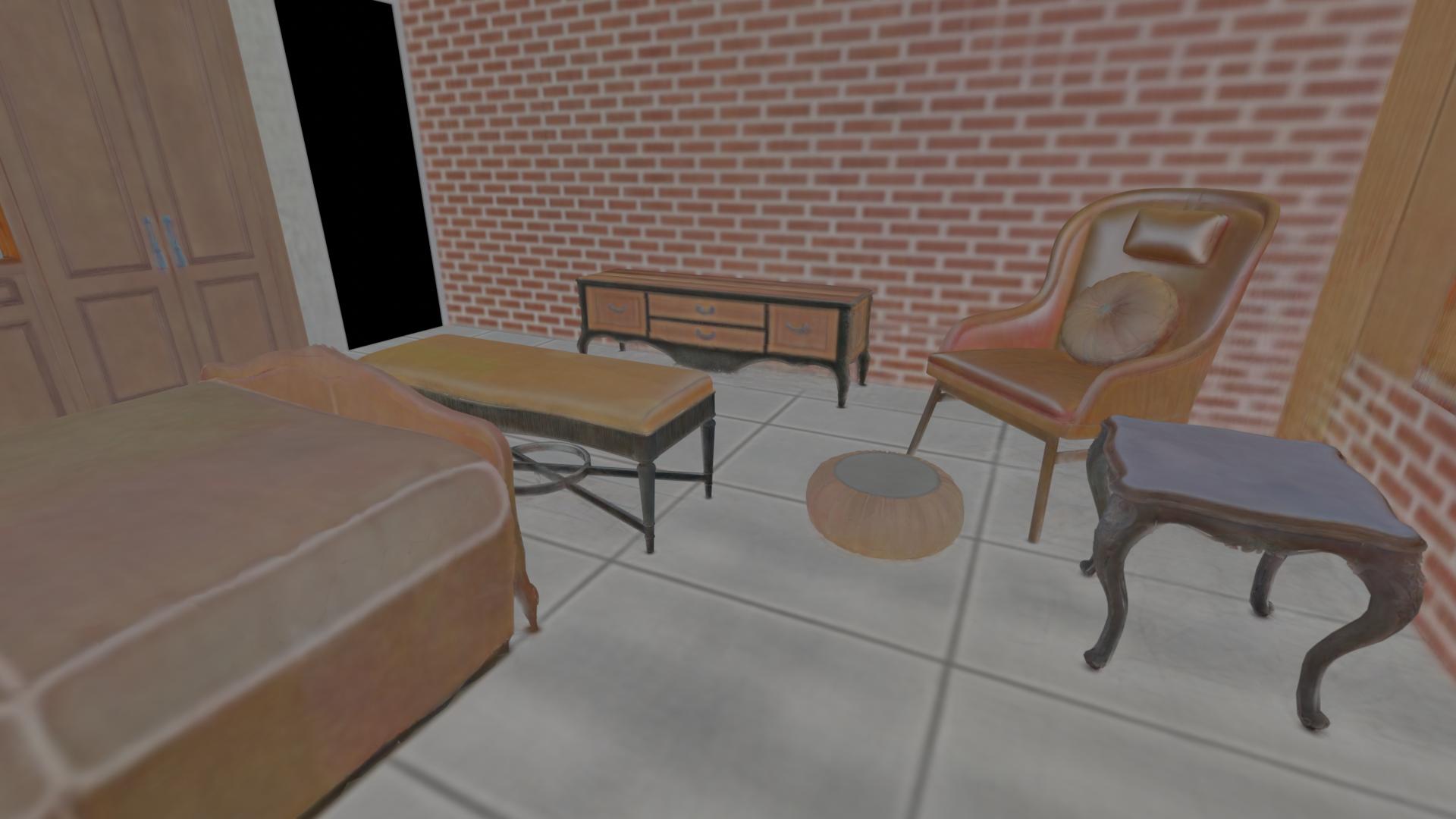}} 
        &
        \begin{tabular}[b]{@{}c@{}}
            \fbox{\includegraphics[width=0.1135\textwidth]{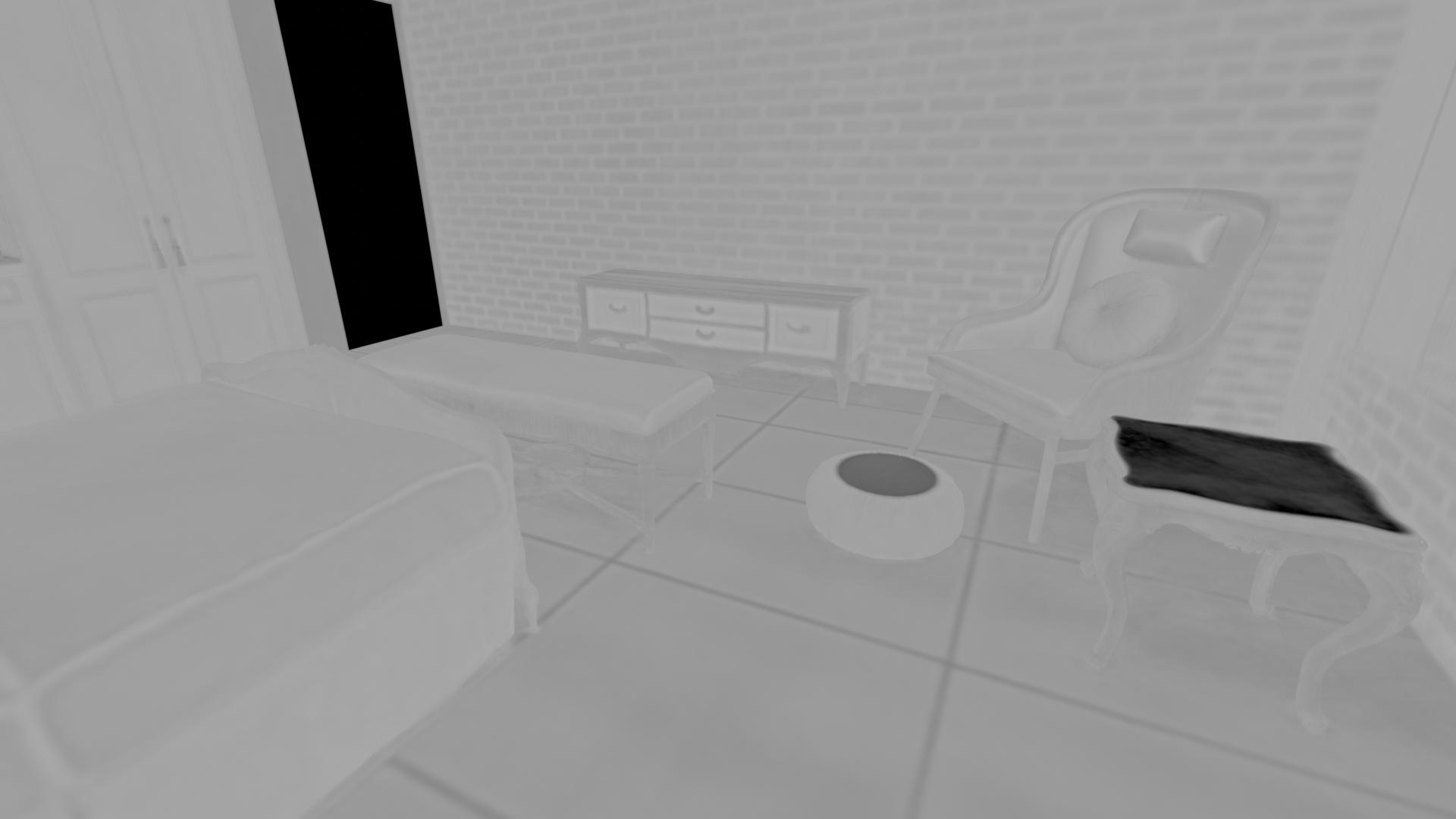}} 
            \\
            \fbox{\includegraphics[width=0.1135\textwidth]{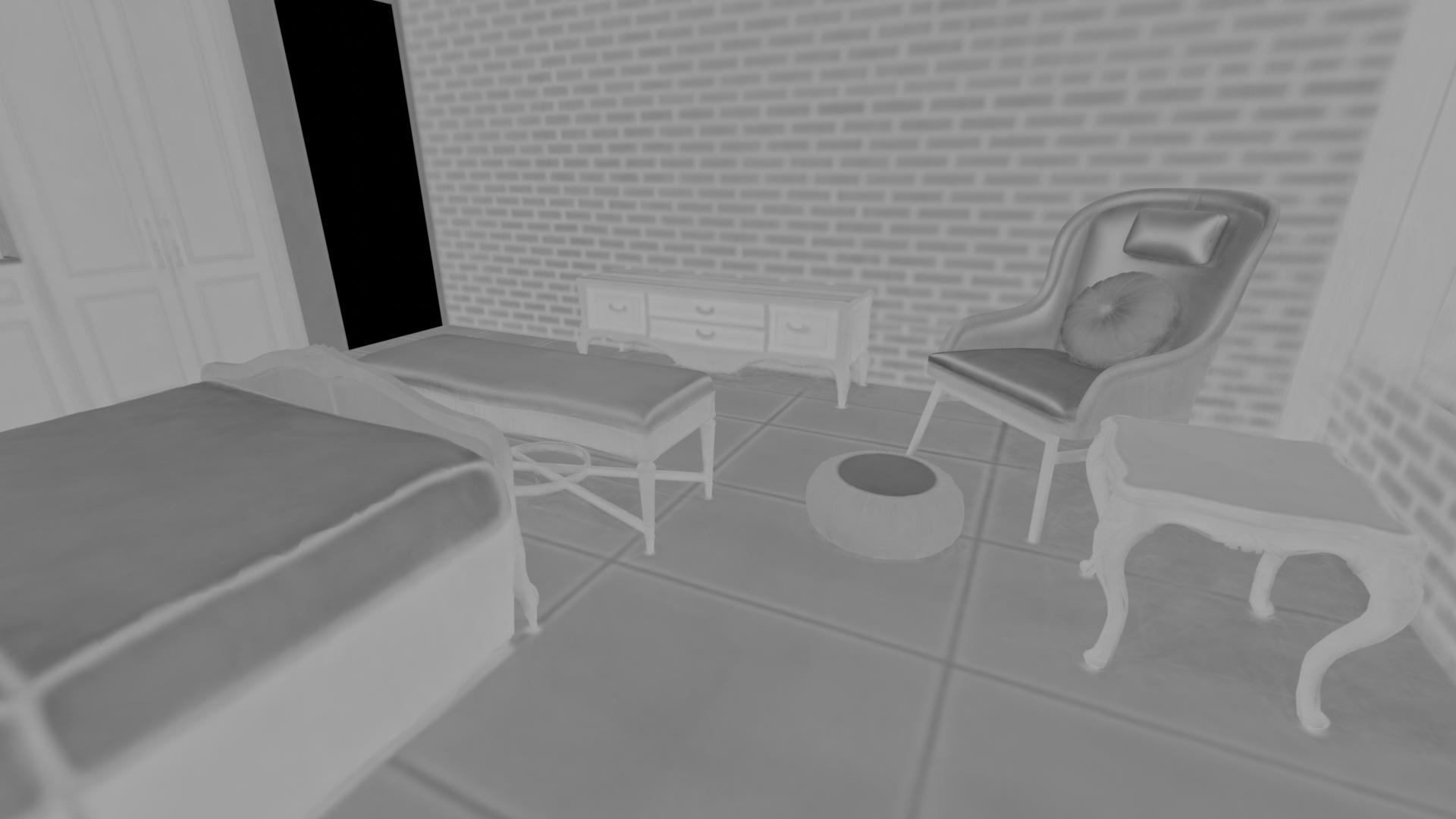}} 
        \end{tabular}   
        &
        \fbox{\includegraphics[width=0.24\textwidth]{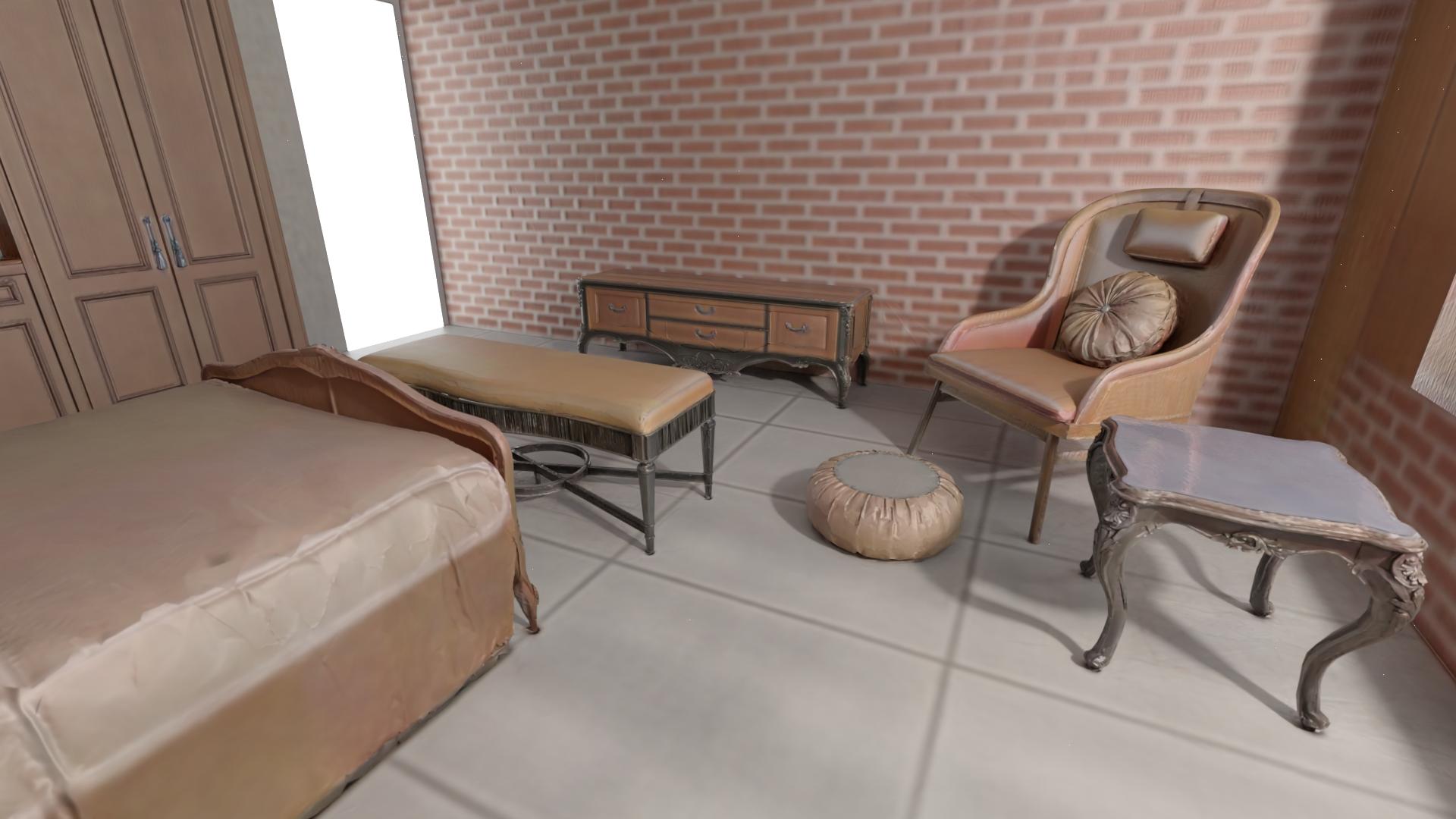}} 
        \\
   
        {\footnotesize Input Scene Mesh} &
        &
        {\footnotesize Displaced Normal} &
        {\footnotesize Albedo} &
        {\footnotesize R/M} &
        {\footnotesize Rendering}
        
    \end{tabular}}
    \vspace{6pt}
    \caption{\textbf{Scene Texturing}. 
    We can use our method for scene texturing using score distillation \cite{SceneTex}.
    Given a scene geometry, first, we condition our method on the rendered normal maps to produce the remaining PBR maps.
    Through iterative optimization, we obtain realistic PBR textures for the whole scene.
    Then, we similarly optimize for normal map textures to obtain fine geometric details, conditioned on rendered material maps. 
    This showcases the potential of \textit{direct} PBR map generation to democratize scene texturing from only text as input.
    }
    \label{fig:exp:scenetex}
\end{figure*}

%% file: figures/experiments/comparisons.tex
\begin{figure*}[t]
    \centering
    \setlength\tabcolsep{1.25pt}
    \resizebox{\textwidth}{!}{
    \fboxsep=0pt
        \begin{tabular}{ccc|cc||cc|cc}
    
        \rotatebox{90}{IID \cite{IID}}
        &
        \fbox{\includegraphics[width=0.14\textwidth]{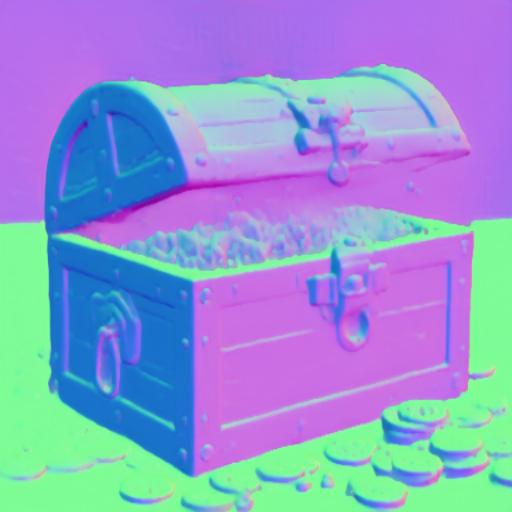}} 
        &
        \begin{tikzpicture}[every node/.style={anchor=north west,inner sep=0pt},x=1pt, y=-1pt,]  
             \node (fig1) at (0,0)
               {\fbox{\includegraphics[width=0.14\textwidth]{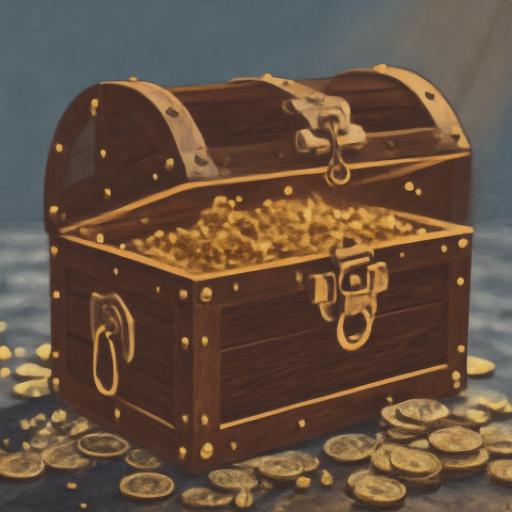}}};   
             \node (fig2) at (59,0)
               {\fbox{\includegraphics[width=0.065\textwidth]{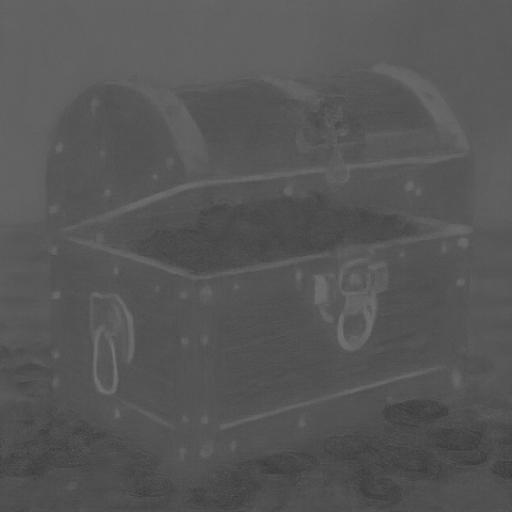}}};
             \node (fig3) at (59,30)
               {\fbox{\includegraphics[width=0.065\textwidth]{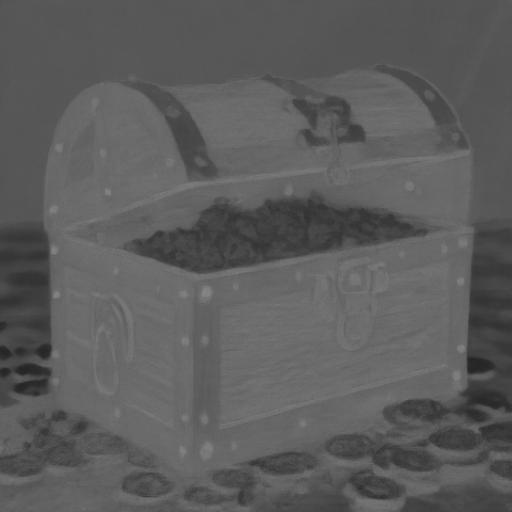}}};
        \end{tikzpicture}
        &
        \fbox{\includegraphics[width=0.14\textwidth]{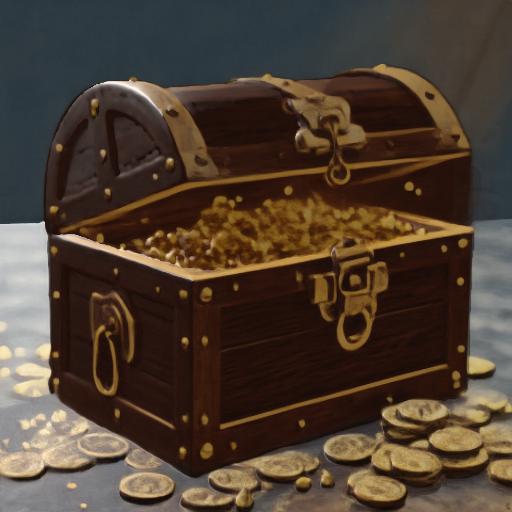}} 
        &
        \fbox{\includegraphics[width=0.14\textwidth]{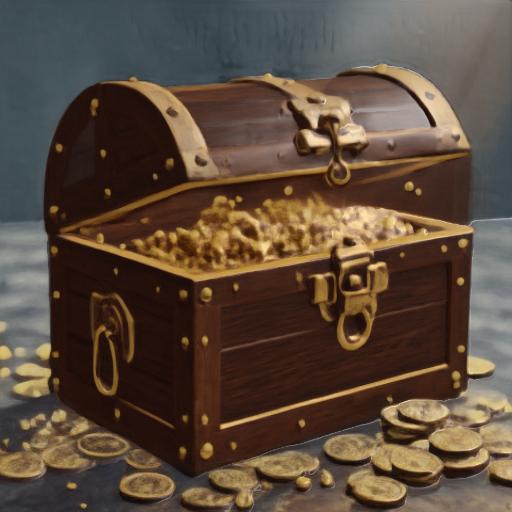}} 
        
        \hspace{6pt}
        &
        \hspace{6pt}
        
        \fbox{\includegraphics[width=0.14\textwidth]{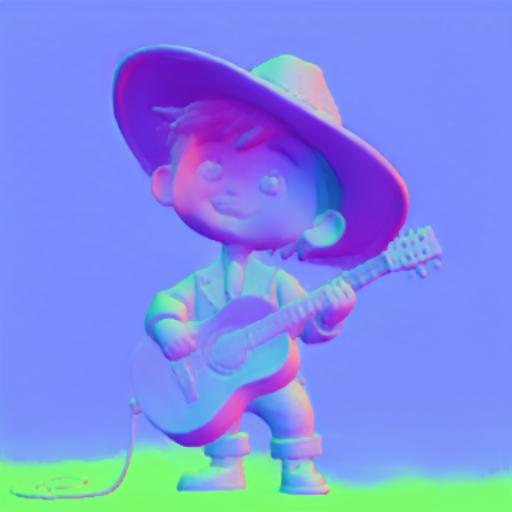}} 
        &
        \begin{tikzpicture}[every node/.style={anchor=north west,inner sep=0pt},x=1pt, y=-1pt,]  
             \node (fig1) at (0,0)
               {\fbox{\includegraphics[width=0.14\textwidth]{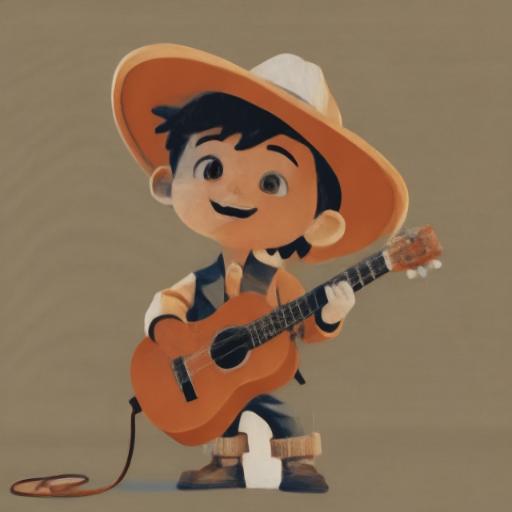}}};   
             \node (fig2) at (59,0)
               {\fbox{\includegraphics[width=0.065\textwidth]{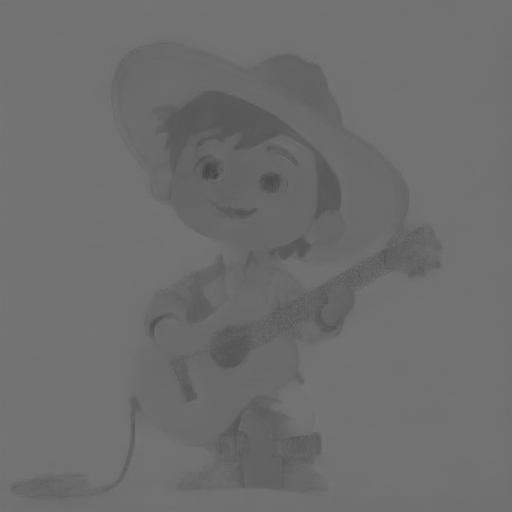}}};
             \node (fig3) at (59,30)
               {\fbox{\includegraphics[width=0.065\textwidth]{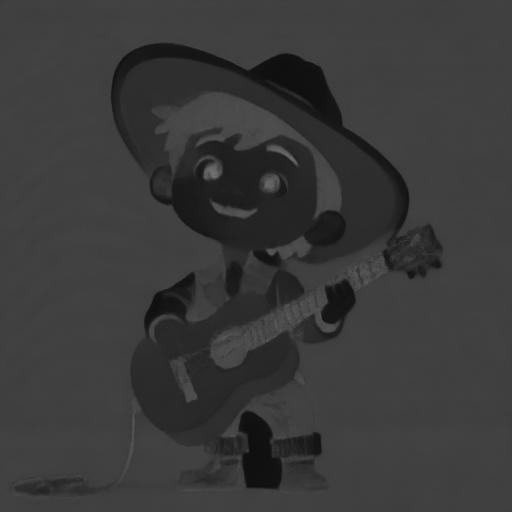}}};
        \end{tikzpicture}
        &
        \fbox{\includegraphics[width=0.14\textwidth]{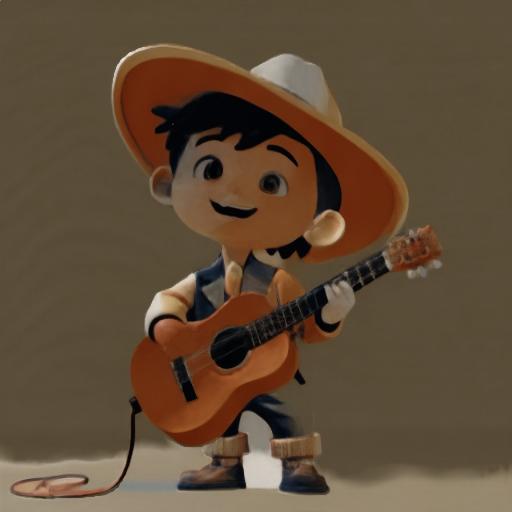}} 
        &
        \fbox{\includegraphics[width=0.14\textwidth]{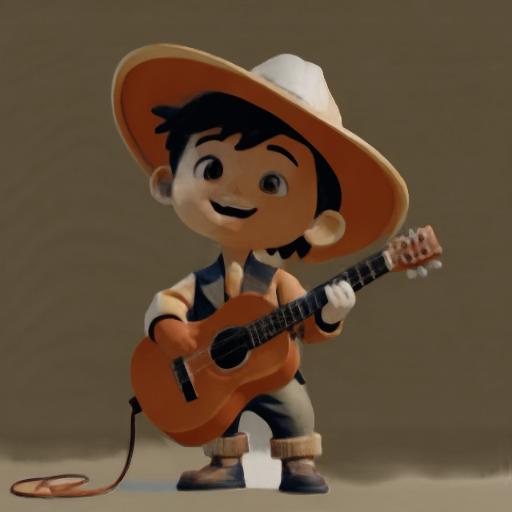}} 
        \\
        
        \rotatebox{90}{RGBX \cite{RGBX}}
        &
        \fbox{\includegraphics[width=0.14\textwidth]{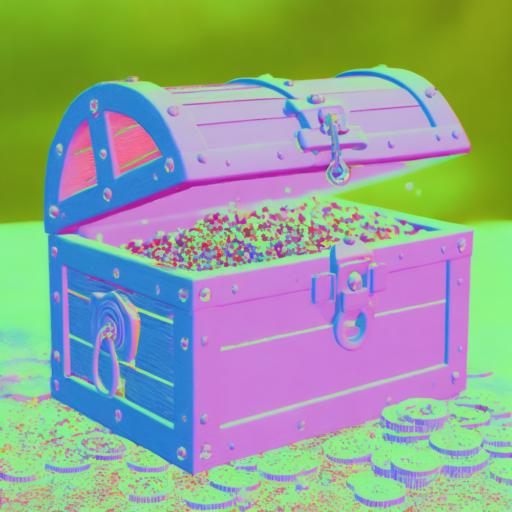}} 
        &
        \begin{tikzpicture}[every node/.style={anchor=north west,inner sep=0pt},x=1pt, y=-1pt,]  
             \node (fig1) at (0,0)
               {\fbox{\includegraphics[width=0.14\textwidth]{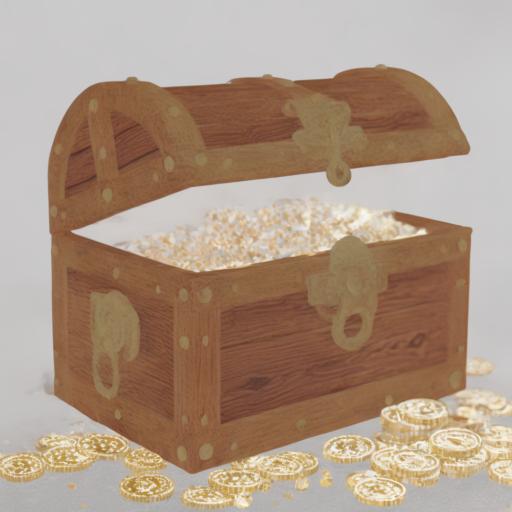}}};   
             \node (fig2) at (59,0)
               {\fbox{\includegraphics[width=0.065\textwidth]{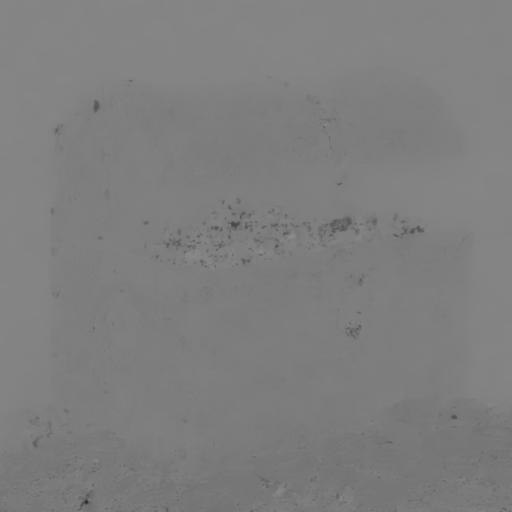}}};
             \node (fig3) at (59,30)
               {\fbox{\includegraphics[width=0.065\textwidth]{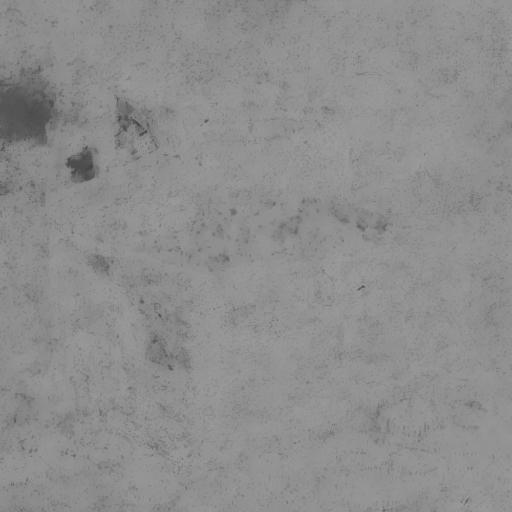}}};
        \end{tikzpicture}
        &
        \fbox{\includegraphics[width=0.14\textwidth]{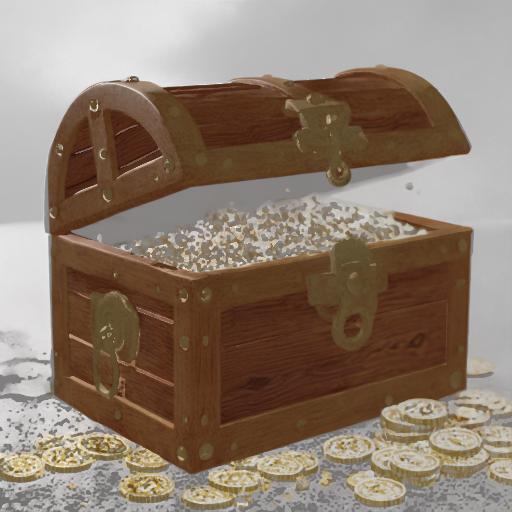}} 
        &
        \fbox{\includegraphics[width=0.14\textwidth]{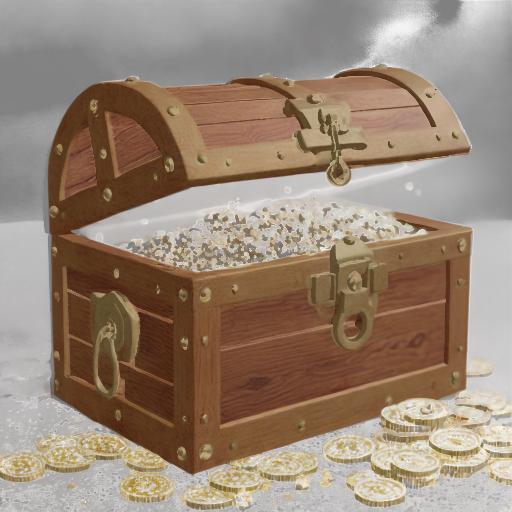}} 
        
        \hspace{6pt}
        &
        \hspace{6pt}
        
        \fbox{\includegraphics[width=0.14\textwidth]{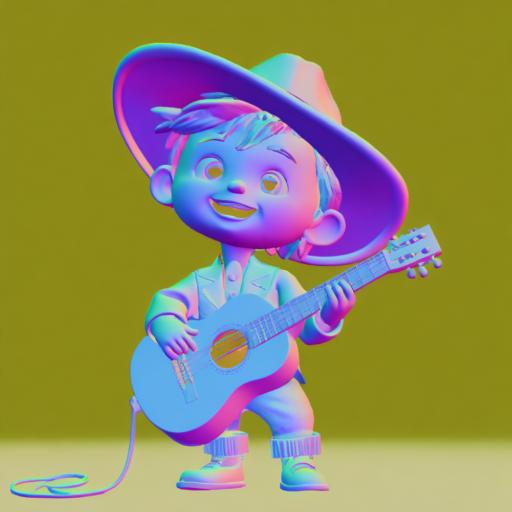}} 
        &
        \begin{tikzpicture}[every node/.style={anchor=north west,inner sep=0pt},x=1pt, y=-1pt,]  
             \node (fig1) at (0,0)
               {\fbox{\includegraphics[width=0.14\textwidth]{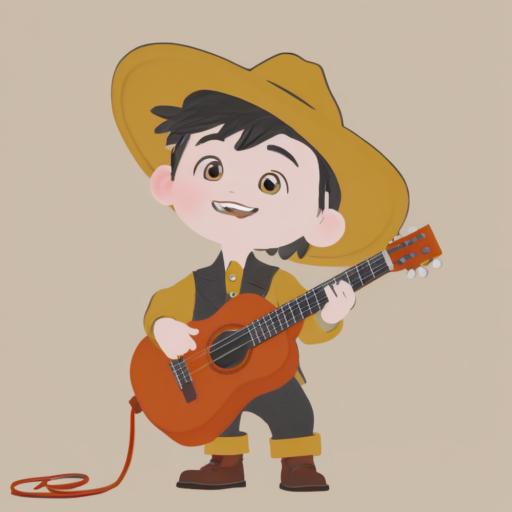}}};   
             \node (fig2) at (59,0)
               {\fbox{\includegraphics[width=0.065\textwidth]{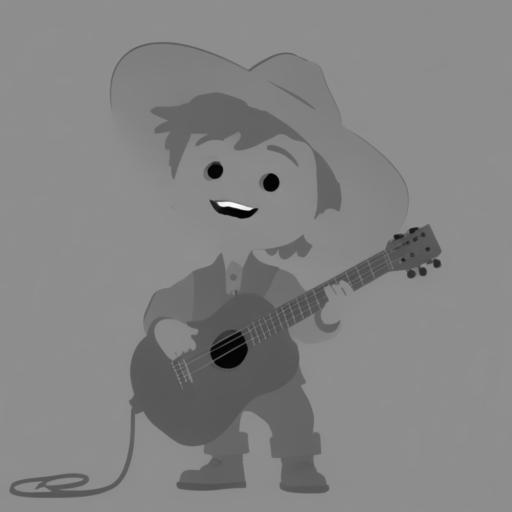}}};
             \node (fig3) at (59,30)
               {\fbox{\includegraphics[width=0.065\textwidth]{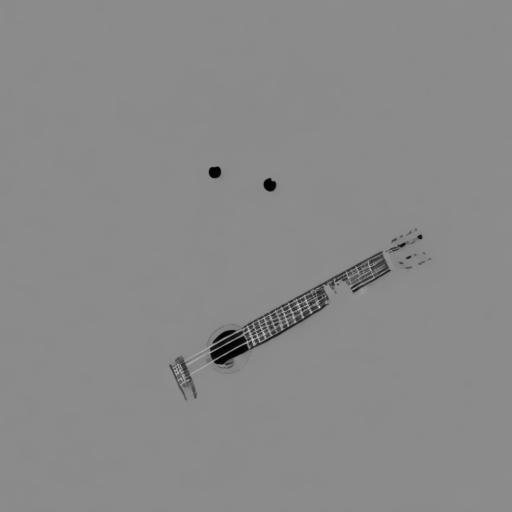}}};
        \end{tikzpicture}
        &
        \fbox{\includegraphics[width=0.14\textwidth]{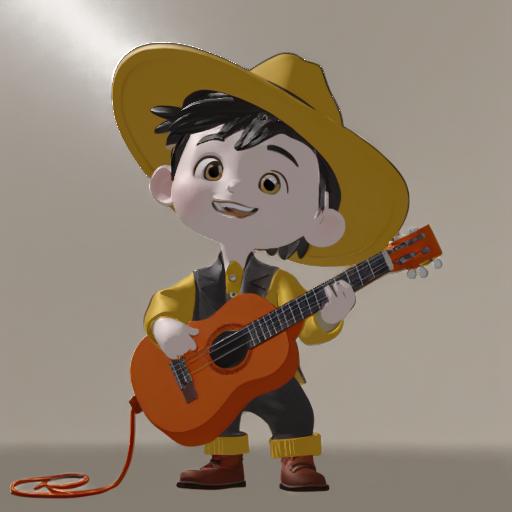}} 
        &
        \fbox{\includegraphics[width=0.14\textwidth]{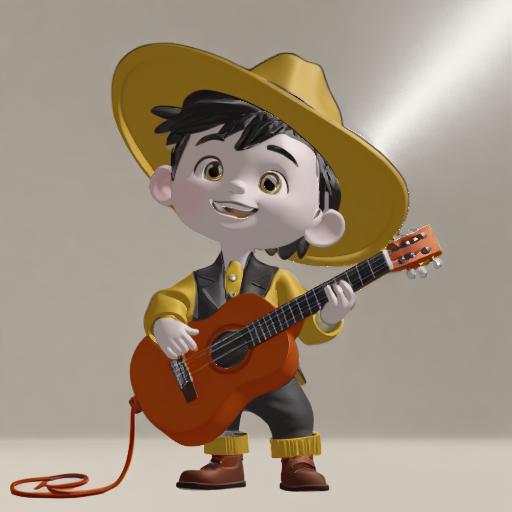}} 
        \\

        \midrule
        
        \rotatebox{90}{Ours}
        &
        \fbox{\includegraphics[width=0.14\textwidth]{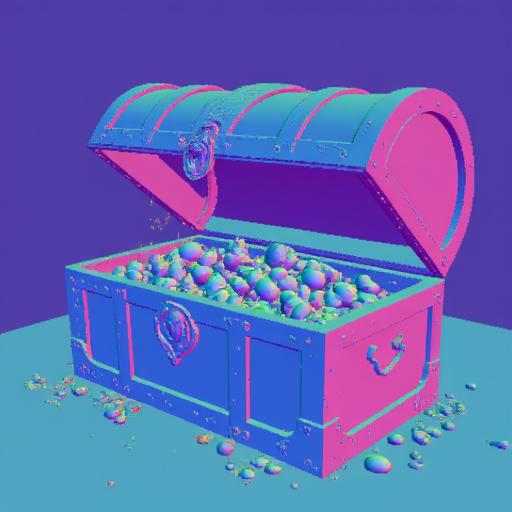}}
        &
        \begin{tikzpicture}[every node/.style={anchor=north west,inner sep=0pt},x=1pt, y=-1pt,]  
             \node (fig1) at (0,0)
               {\fbox{\includegraphics[width=0.14\textwidth]{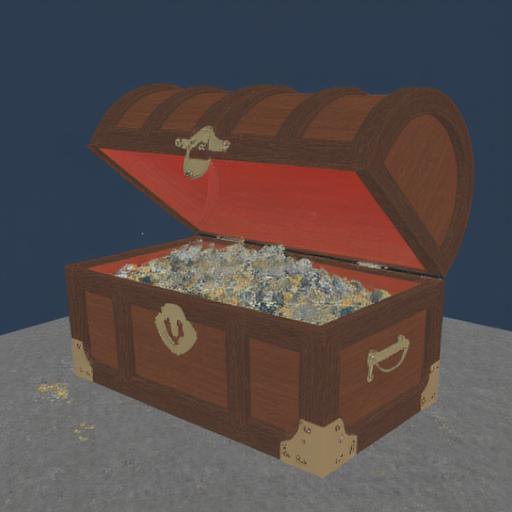}}};   
             \node (fig2) at (59,0)
               {\fbox{\includegraphics[width=0.065\textwidth]{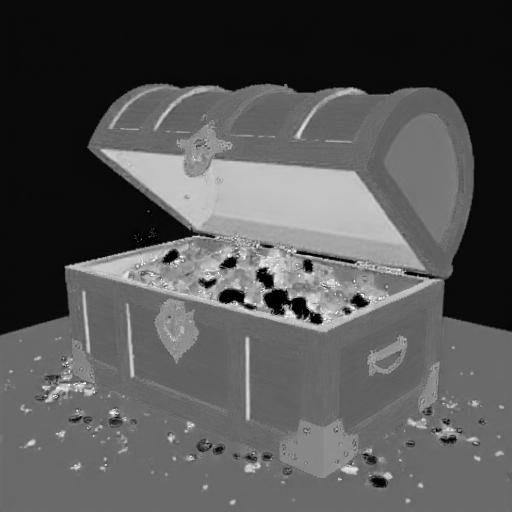}}};
             \node (fig3) at (59,30)
               {\fbox{\includegraphics[width=0.065\textwidth]{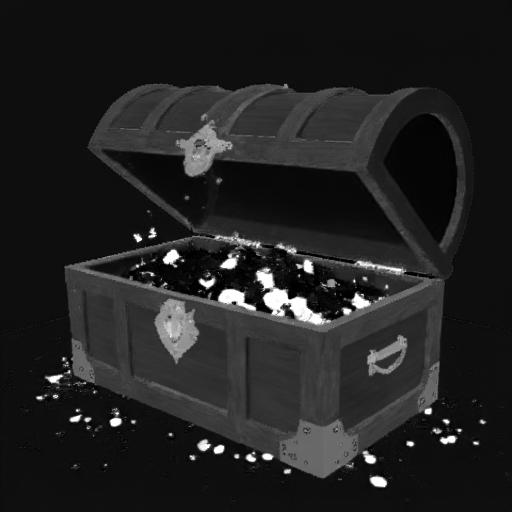}}};
        \end{tikzpicture}
        &
        \fbox{\includegraphics[width=0.14\textwidth]{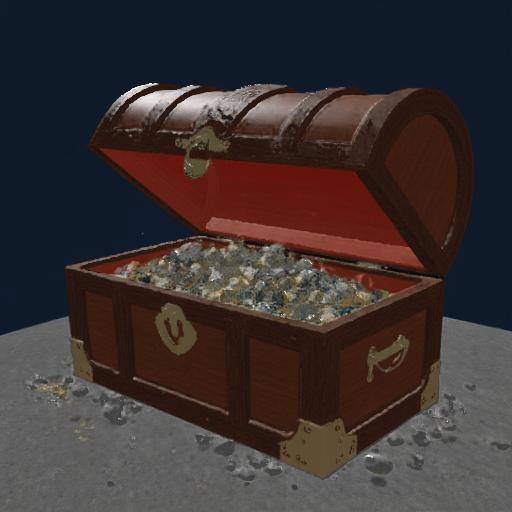}}
        &
        \fbox{\includegraphics[width=0.14\textwidth]{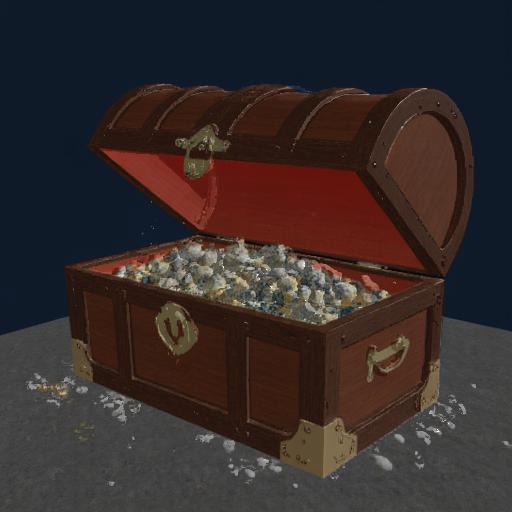}}
        
        \hspace{6pt}
        &
        \hspace{6pt}
        
        \fbox{\includegraphics[width=0.14\textwidth]{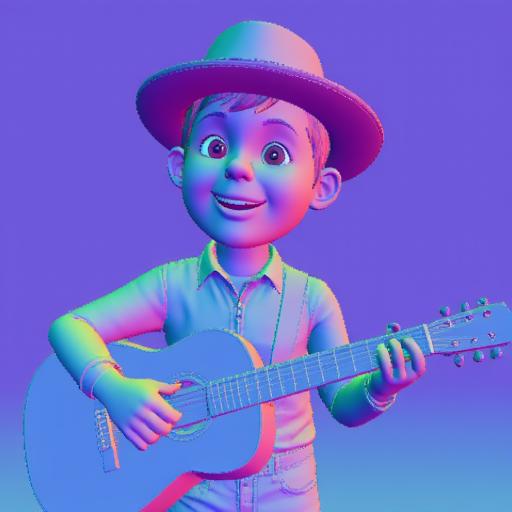}}
        &
        \begin{tikzpicture}[every node/.style={anchor=north west,inner sep=0pt},x=1pt, y=-1pt,]  
             \node (fig1) at (0,0)
               {\fbox{\includegraphics[width=0.14\textwidth]{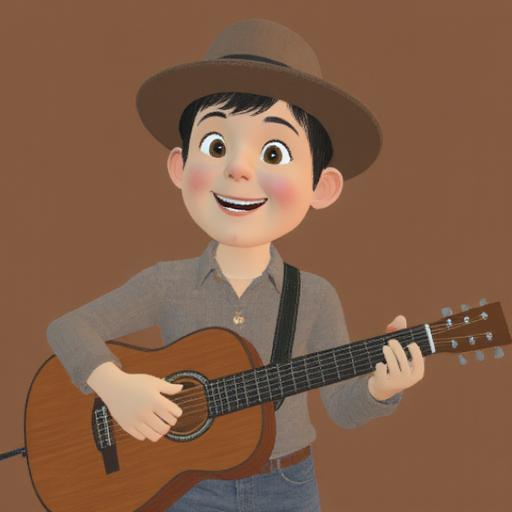}}};   
             \node (fig2) at (59,0)
               {\fbox{\includegraphics[width=0.065\textwidth]{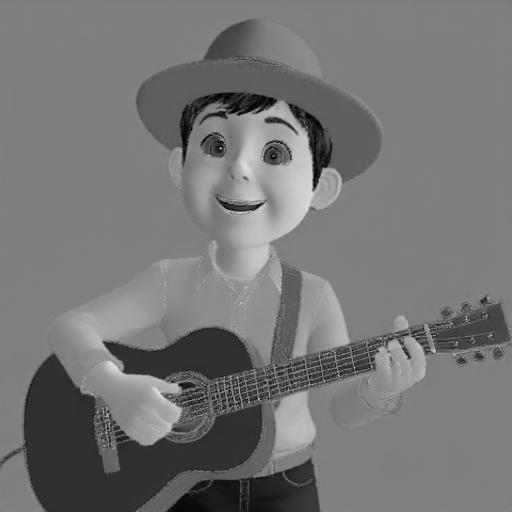}}};
             \node (fig3) at (59,30)
               {\fbox{\includegraphics[width=0.065\textwidth]{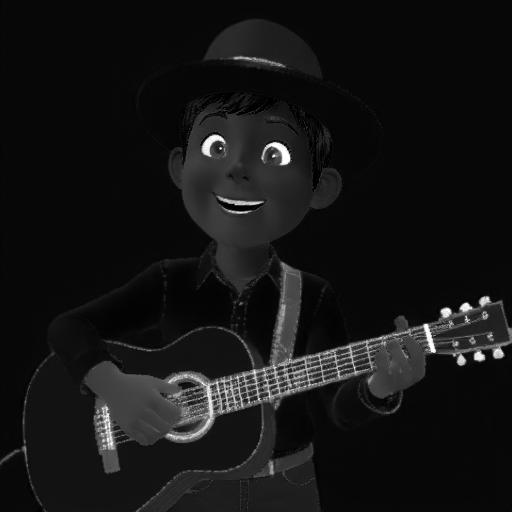}}};
        \end{tikzpicture}
        &
        \fbox{\includegraphics[width=0.14\textwidth]{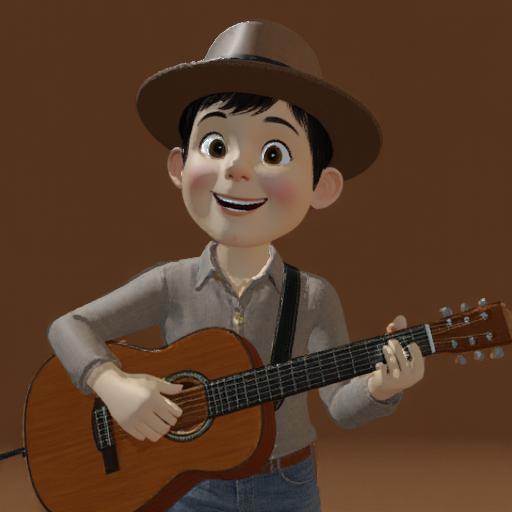}}
        &
        \fbox{\includegraphics[width=0.14\textwidth]{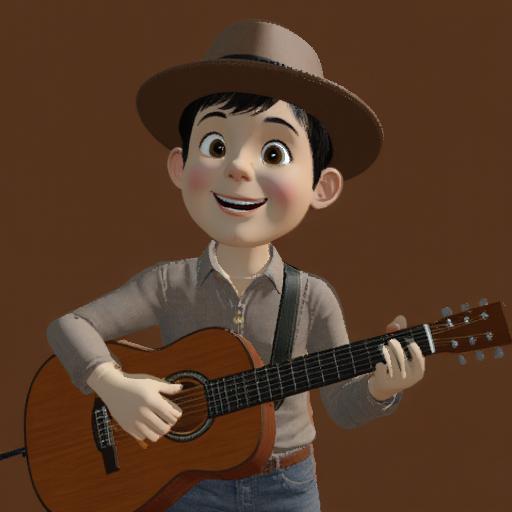}}
        \\

         &
        Normal &
        Material &
        Lighting 1 &
        Lighting 2 &
        
        Normal &
        Material &
        Lighting 1 &
        Lighting 2 \\
        
    \end{tabular}}
    \vspace{12pt}
    \caption{\textbf{Rendering comparisons}. 
    We show sample PBR maps of our method and baselines as well as rendered RGB images under two different lighting conditions.
    We use a diverse set of text prompts to produce our PBR maps, as well as the input RGB images for the baseline methods.
    This highlights our models' capability to retain the generalized prior of the pretrained text-to-image model.
    Our method better captures the semantic meaning of the individual intrinsic properties.
    For example, there are no baked-in lighting effects in the albedo, and the metallic/roughness maps are sharper with more intricate details.
    This leads to more realistic renderings and lighting effects.
    }
    \label{fig:exp:comparisons}
\end{figure*}

%% file: tables/experiments/baselines.tex
\begin{table}
    \setlength\tabcolsep{2pt}
    \begin{center}
    \caption{\textbf{Baseline comparisons.}
    We compare the albedo quality for in-distribution (A-ID-FID) and out-of-distribution (A-OOD-FID) settings as well as perceptually with a user study (A-PQ).
    We evaluate the material quality with a user study focusing on the rendering quality (R-PQ), specularity quality (S-PQ), and prompt coherence (PC). 
    Our method produces the best quality and it is preferred by most of the participants.
    }
    \label{tab:exp:comparisons}
    \vspace{12pt}
    \begin{tabular}{l{c}{c}{c}{c}{c}{c}}
      \toprule
        & A-ID-FID$\downarrow$ & A-OOD-FID$\downarrow$ & A-PQ$\uparrow$ & R-PQ$\uparrow$ & S-PQ$\uparrow$ & PC$\uparrow$  \\
       \midrule 
      IID \cite{IID}  & 78.77 & 98.77 & 14.24\% & 2.95\tiny{{$\pm$}1.03} & 2.82\tiny{{$\pm$}1.13} & 4.47\tiny{{$\pm$}0.89} \\
      RGBX \cite{RGBX} & \textbf{61.36} & 90.12 & 15.63\% & 2.96\tiny{{$\pm$}0.98} & 2.57\tiny{{$\pm$}1.07} & 4.33\tiny{{$\pm$}0.93} \\
      ColorfulShading \cite{careagaColorful} & 91.10 & 86.48 & 2.77\% & N/A & N/A & N/A \\
      \midrule 
      IID\cite{IID} w/ FLUX-LoRA  & 103.36 & 79.29 & N/A & N/A & N/A & N/A \\
      \midrule 
      w/o Rendering & 78.77 & 72.23 & N/A & 3.42\tiny{{$\pm$}0.92} & 2.73\tiny{{$\pm$}0.93} & 4.52\tiny{{$\pm$}0.78}\\
      w/o CIA-Dropout & 71.47  & 75.54 & N/A & 3.68\tiny{{$\pm$}0.87} & 3.21\tiny{{$\pm$}1.18} & 4.52\tiny{{$\pm$}0.76} \\
      Ours  & 72.09 & \textbf{71.39} & \textbf{67.36\%} & \textbf{3.93\tiny{{$\pm$}0.88}} & \textbf{3.62\tiny{{$\pm$}0.96}} & \textbf{4.62\tiny{{$\pm$}0.67}} \\
      \bottomrule
    \end{tabular}
    \end{center}
\end{table}

%% file: figures/experiments/albedo_comparisons.tex
\begin{wrapfigure}{r}{0.5\textwidth}
    \centering
    \vspace{-36pt}
    \setlength\tabcolsep{1.25pt}
    \resizebox{0.5\textwidth}{!}{
    \fboxsep=0pt
        \begin{tabular}{ccccccccccc}
        \rotatebox{90}{IID \cite{IID}}
        &
        \fbox{\includegraphics[width=0.13\textwidth]{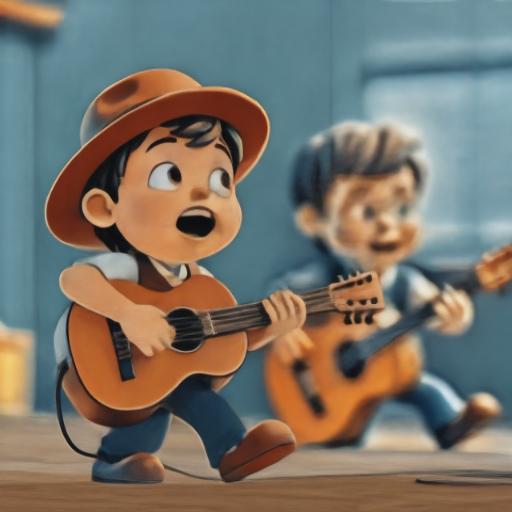}} 
        &
        \fbox{\includegraphics[width=0.13\textwidth]{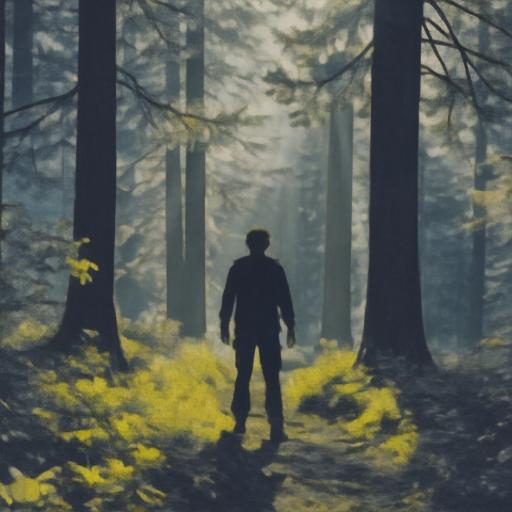}} 
        &
        \fbox{\includegraphics[width=0.13\textwidth]{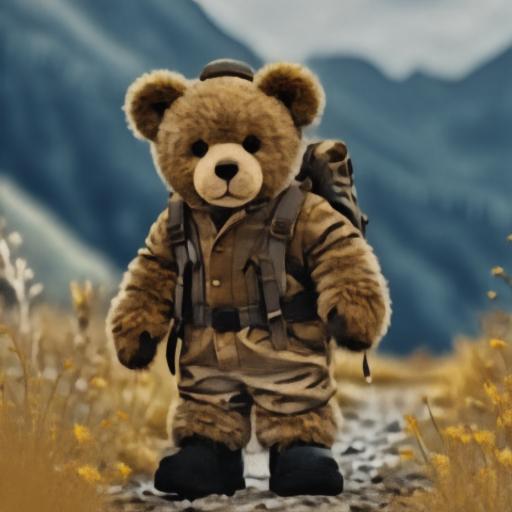}} 
        &
        \fbox{\includegraphics[width=0.13\textwidth]{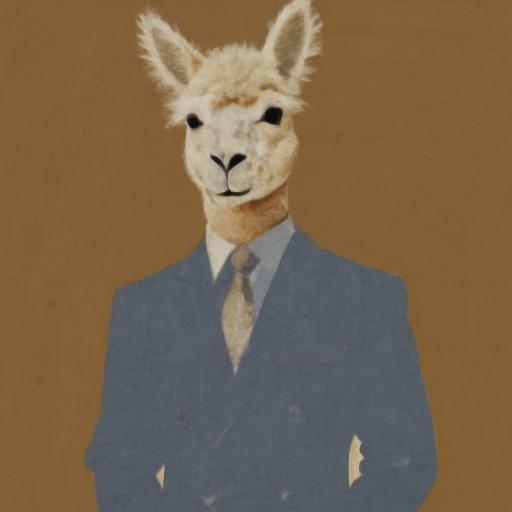}} 
        &
        \fbox{\includegraphics[width=0.13\textwidth]{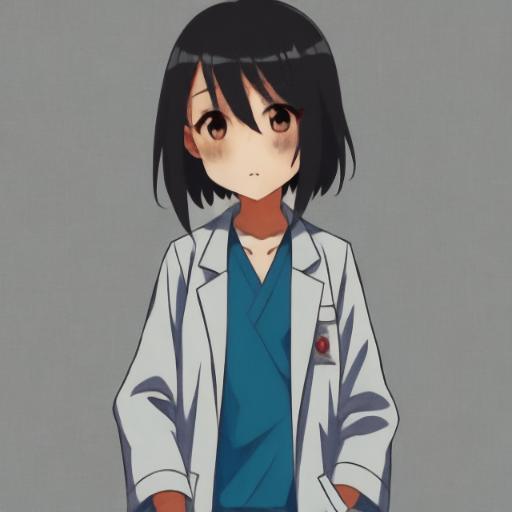}} 
        \\
        
        \rotatebox{90}{RGBX \cite{RGBX}}
        &
        \fbox{\includegraphics[width=0.13\textwidth]{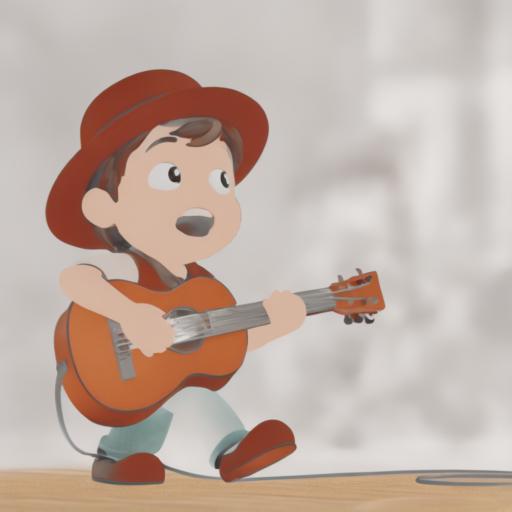}} 
        &
        \fbox{\includegraphics[width=0.13\textwidth]{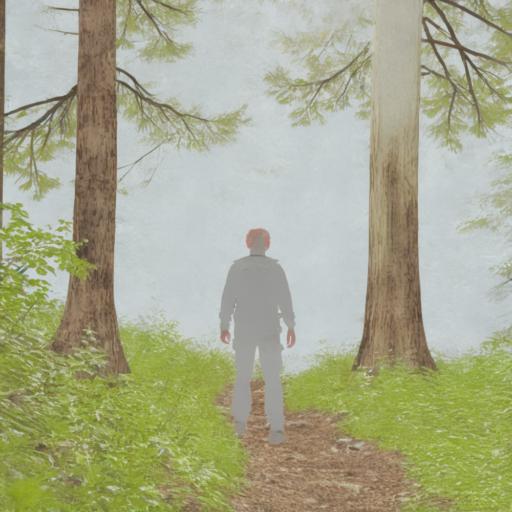}} 
        &
        \fbox{\includegraphics[width=0.13\textwidth]{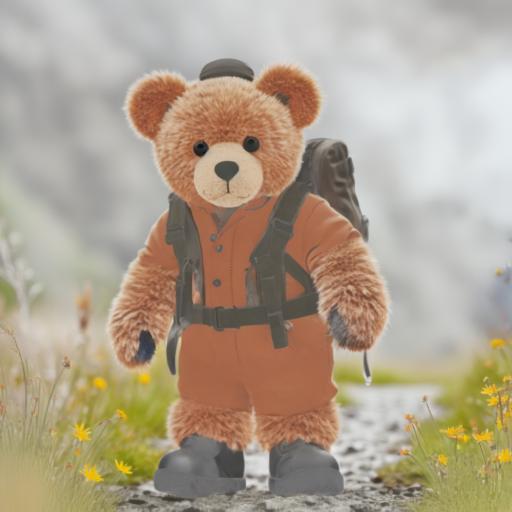}} 
        &
        \fbox{\includegraphics[width=0.13\textwidth]{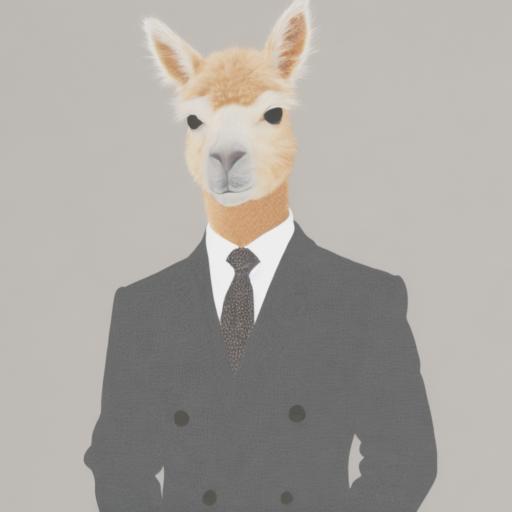}} 
        &
        \fbox{\includegraphics[width=0.13\textwidth]{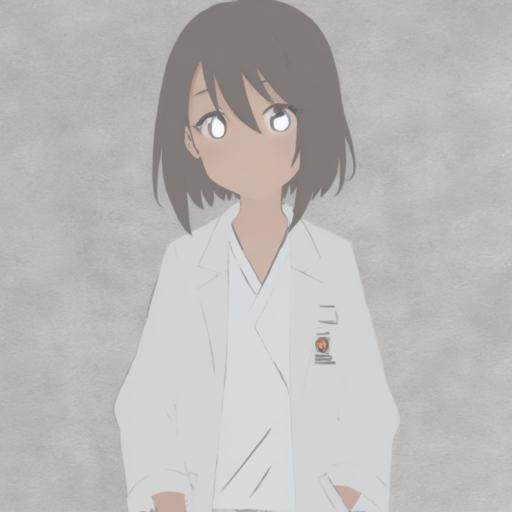}} 
        \\
        
        \rotatebox{90}{\begin{tabular}[b]{@{}c@{}}Colorful\\Shading \cite{careagaColorful}\end{tabular}}
        &
        \fbox{\includegraphics[width=0.13\textwidth]{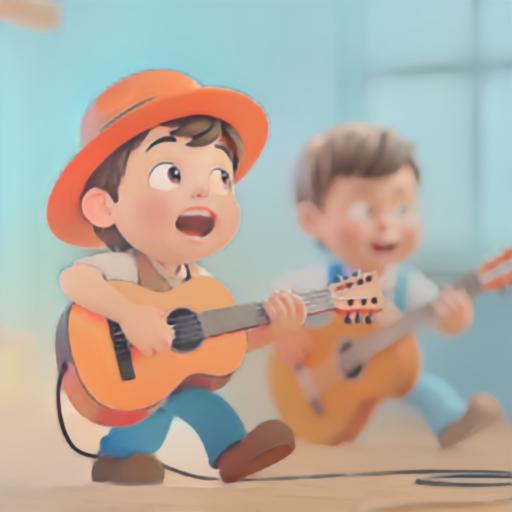}} 
        &
        \fbox{\includegraphics[width=0.13\textwidth]{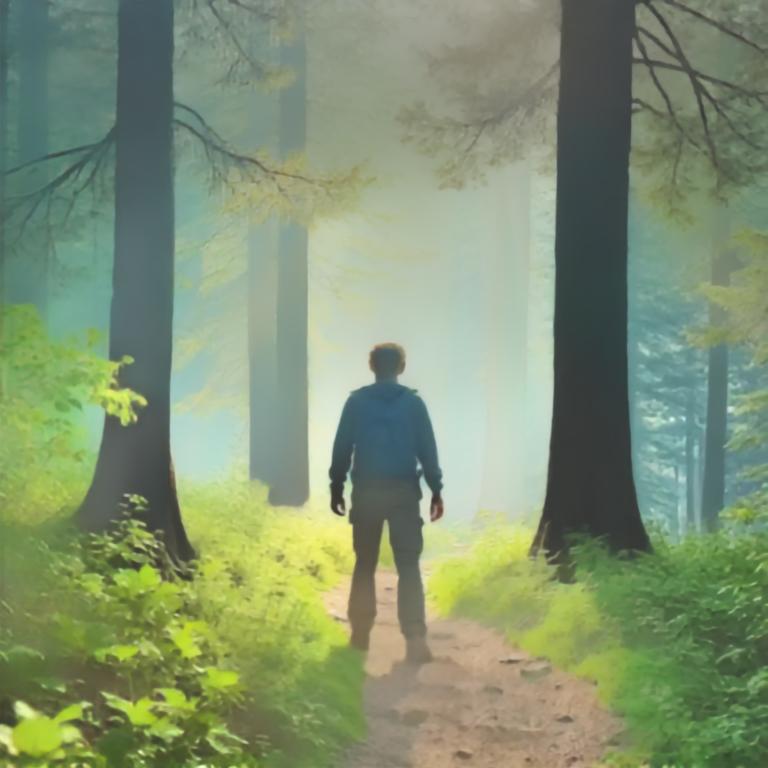}} 
        &
        \fbox{\includegraphics[width=0.13\textwidth]{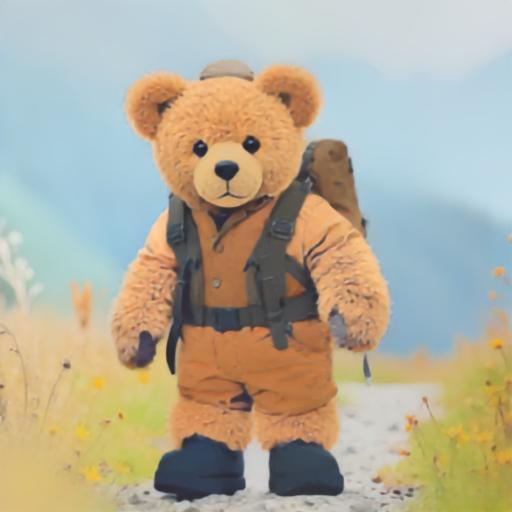}} 
        &
        \fbox{\includegraphics[width=0.13\textwidth]{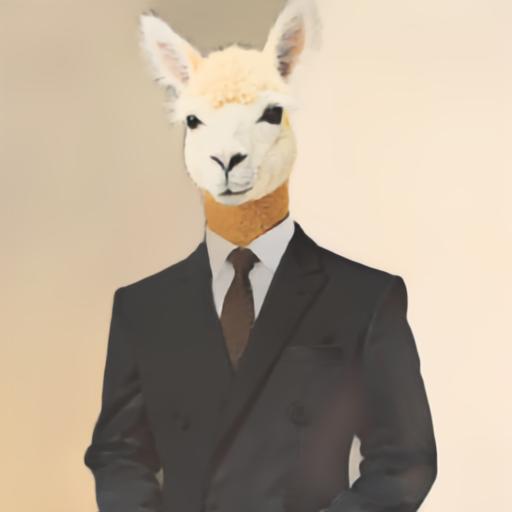}} 
        &
        \fbox{\includegraphics[width=0.13\textwidth]{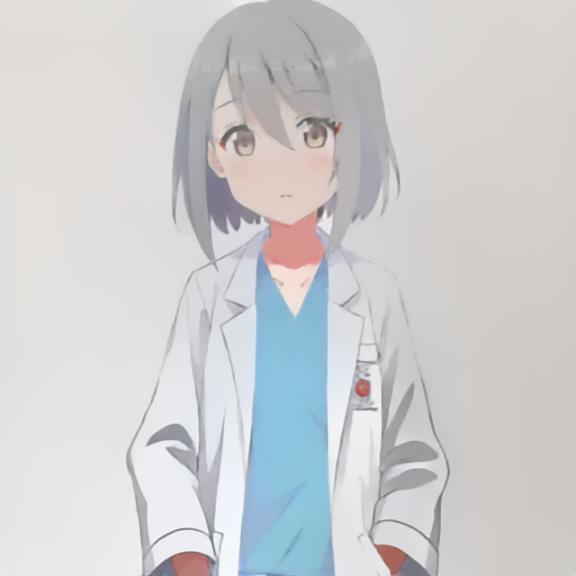}} 
        \\
        
        \rotatebox{90}{Ours}
        &
        \fbox{\includegraphics[width=0.13\textwidth]{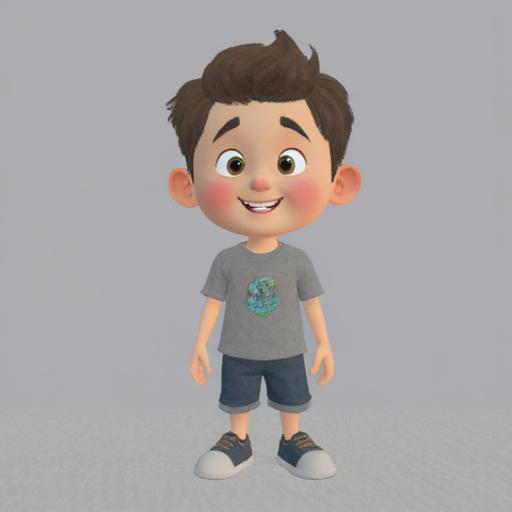}} 
        &
        \fbox{\includegraphics[width=0.13\textwidth]{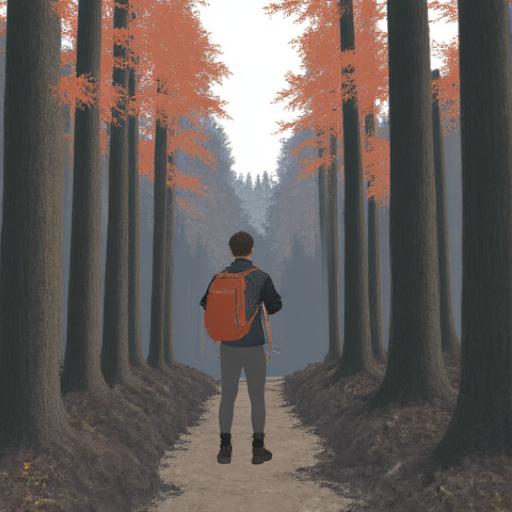}} 
        &
        \fbox{\includegraphics[width=0.13\textwidth]{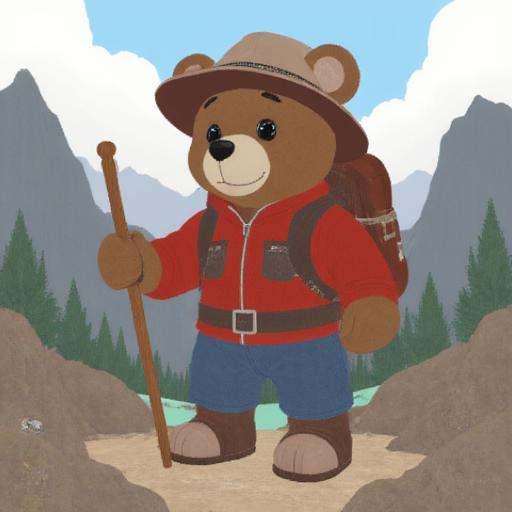}} 
        &
        \fbox{\includegraphics[width=0.13\textwidth]{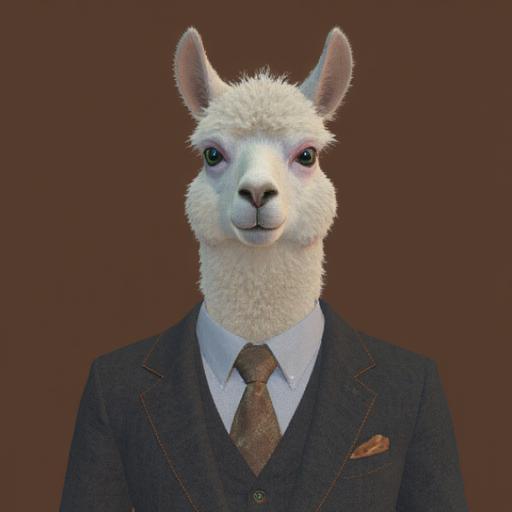}} 
        &
        \fbox{\includegraphics[width=0.13\textwidth]{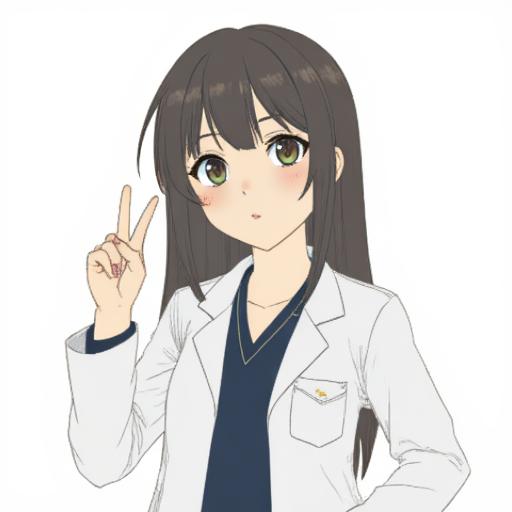}} 
        \\
    \end{tabular}}
    \caption{\textbf{Albedo comparisons}. 
    We show albedo images of our method and baselines corresponding to the same text prompt in each column.
    Our albedo images have less baked-in shadows and reflections, which is desirable for downstream tasks, such as physically-based rendering.
    We provide more samples in the supplemental.
    }
    \label{fig:exp:albedo_comparisons}
\end{wrapfigure}

%% file: figures/experiments/ablations.tex
\begin{wrapfigure}{r}{0.5\textwidth}
    \centering
    \setlength\tabcolsep{1.25pt}
    \resizebox{0.5\textwidth}{!}{
    \fboxsep=0pt
    \begin{tabular}[b]{cc}
        \rotatebox{90}{\large{w/o Rendering}}
        &
        \begin{tabular}[b]{@{}c@{}}
            \resizebox{0.64\textwidth}{!}{
            \begin{tabular}[b]{ccccc}
                Normal &
                Albedo & 
                R/M &
                Lighting 1 &
                Lighting 2
                \\
                \fbox{\includegraphics[width=0.10\textwidth]{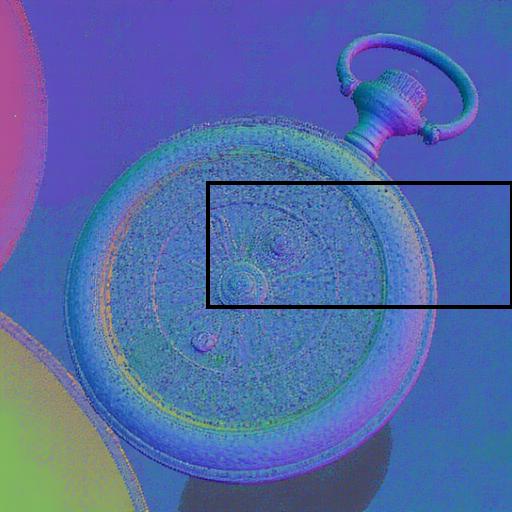}} 
                &
                \fbox{\includegraphics[width=0.10\textwidth]{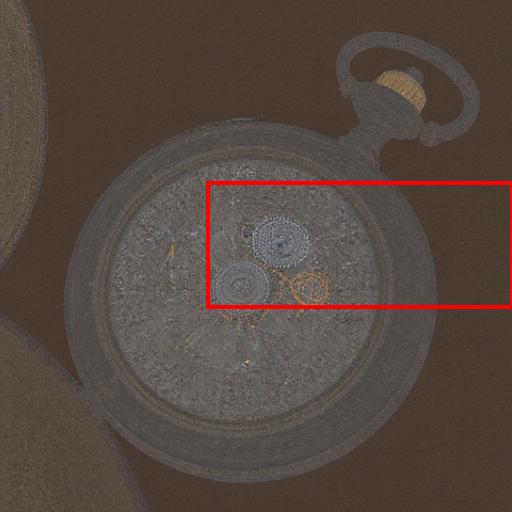}} 
                &
                \begin{tabular}[b]{@{}c@{}}
                    \fbox{\includegraphics[width=0.046\textwidth]{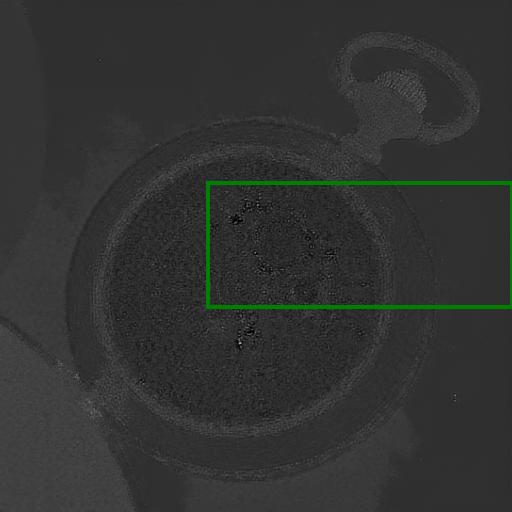}}
                    \\
                    \fbox{\includegraphics[width=0.046\textwidth]{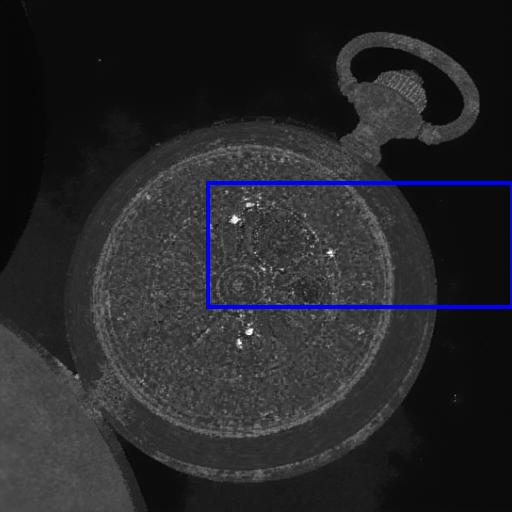}}
                \end{tabular}
                &
                \fbox{\includegraphics[width=0.10\textwidth]{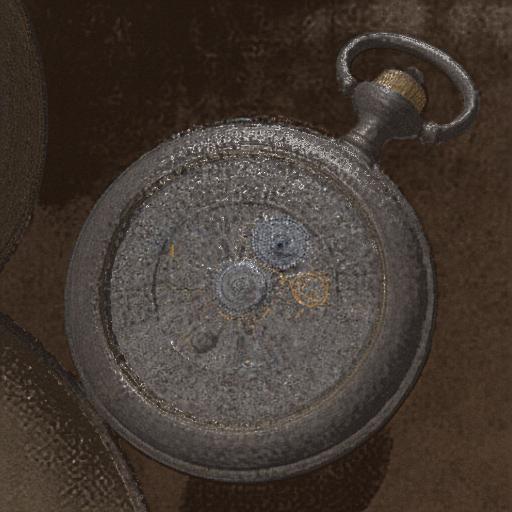}} 
                &
                \fbox{\includegraphics[width=0.10\textwidth]{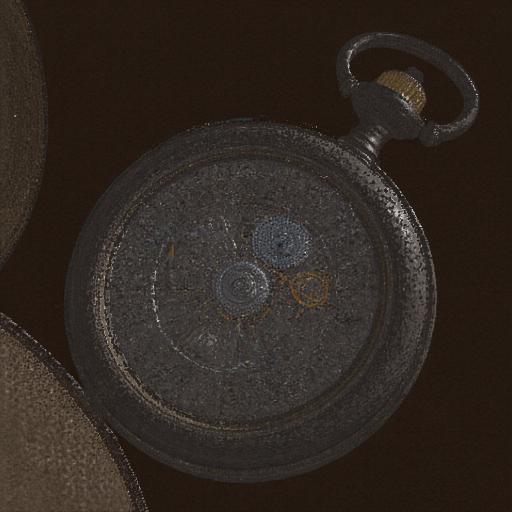}} 
            \end{tabular}}
            \\
            \resizebox{0.64\textwidth}{!}{
            \begin{tabular}[b]{cccc}
                \fbox{\includegraphics[width=0.10\textwidth]{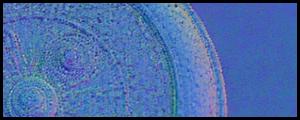}} 
                &
                \fbox{\includegraphics[width=0.10\textwidth]{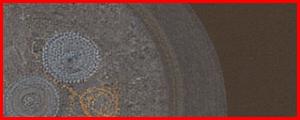}} 
                &
                \fbox{\includegraphics[width=0.10\textwidth]{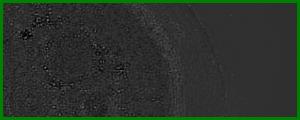}} 
                &
                \fbox{\includegraphics[width=0.10\textwidth]{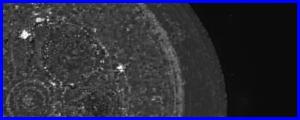}} 
            \end{tabular}}
        \end{tabular} \\
                
        \rotatebox{90}{w/o Light Sampling}
        &
        \begin{tabular}[b]{@{}c@{}}
            \resizebox{0.64\textwidth}{!}{
            \begin{tabular}[b]{ccccc}
            \fbox{\includegraphics[width=0.10\textwidth]{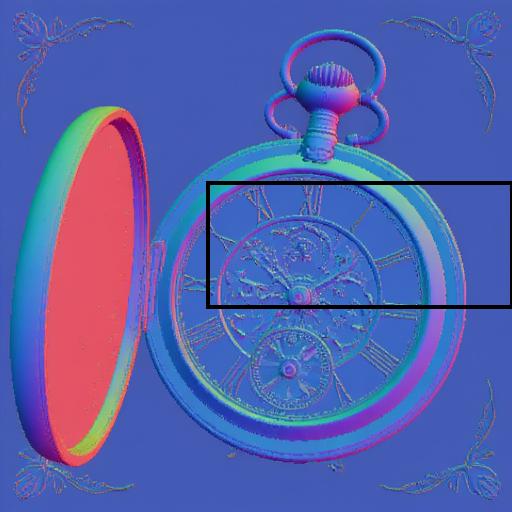}} 
            &
            \fbox{\includegraphics[width=0.10\textwidth]{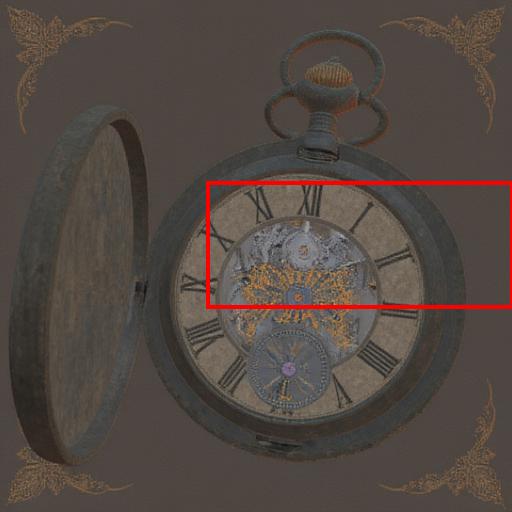}} 
            &
            \begin{tabular}[b]{@{}c@{}}
                \fbox{\includegraphics[width=0.046\textwidth]{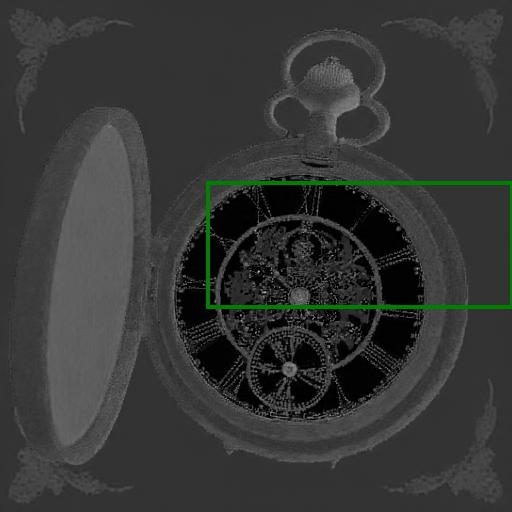}}
                \\
                \fbox{\includegraphics[width=0.046\textwidth]{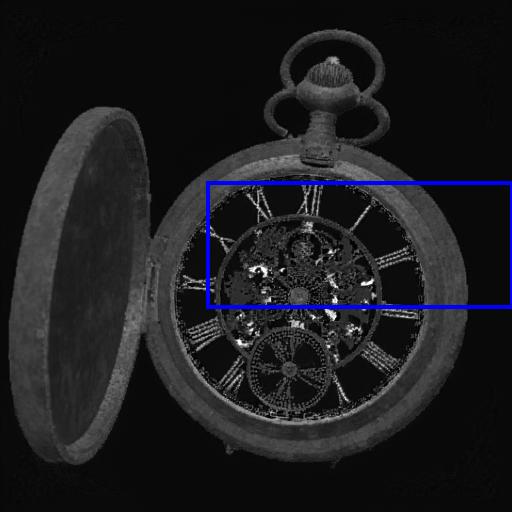}}
            \end{tabular}
            &
            \fbox{\includegraphics[width=0.10\textwidth]{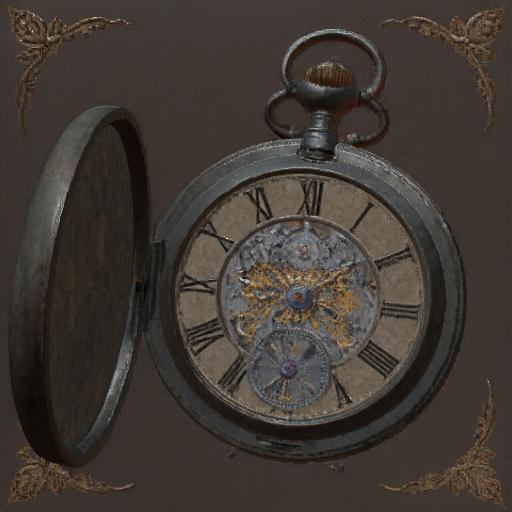}} 
            &
            \fbox{\includegraphics[width=0.10\textwidth]{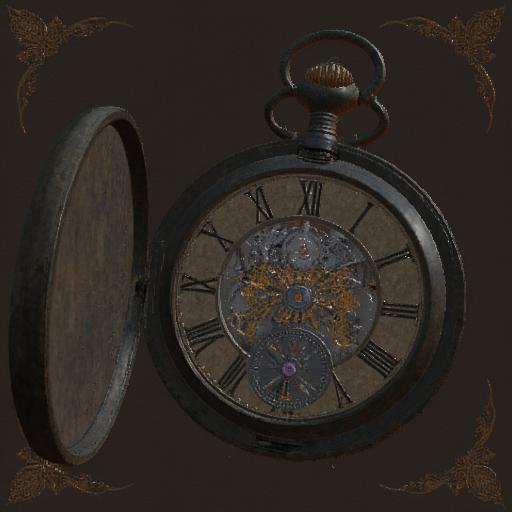}} 
        \end{tabular}}
            \\
            \resizebox{0.64\textwidth}{!}{
            \begin{tabular}[b]{cccc}
                \fbox{\includegraphics[width=0.10\textwidth]{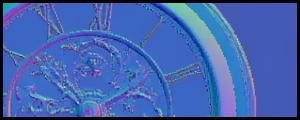}} 
                &
                \fbox{\includegraphics[width=0.10\textwidth]{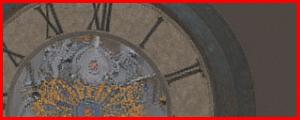}} 
                &
                \fbox{\includegraphics[width=0.10\textwidth]{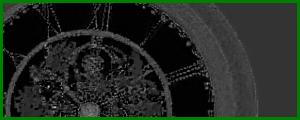}} 
                &
                \fbox{\includegraphics[width=0.10\textwidth]{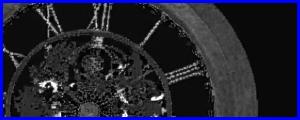}} 
            \end{tabular}}
        \end{tabular} \\
                
        \rotatebox{90}{w/o CIA-Dropout}
        &
        \begin{tabular}[b]{@{}c@{}}
            \resizebox{0.64\textwidth}{!}{
            \begin{tabular}[b]{ccccc}
            \fbox{\includegraphics[width=0.10\textwidth]{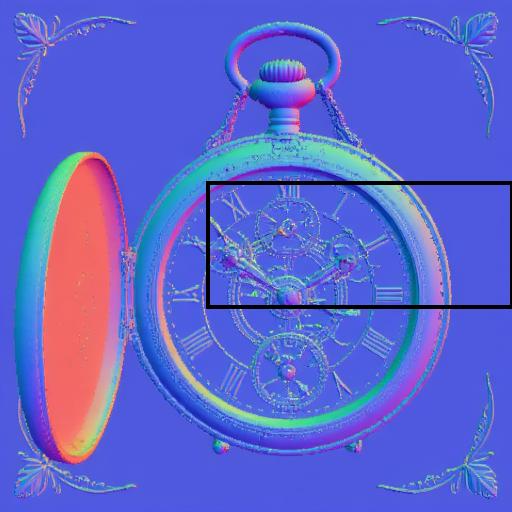}} 
            &
            \fbox{\includegraphics[width=0.10\textwidth]{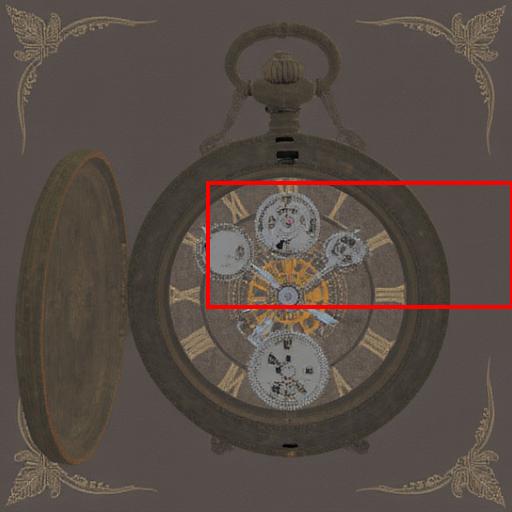}} 
            &
            \begin{tabular}[b]{@{}c@{}}
                \fbox{\includegraphics[width=0.046\textwidth]{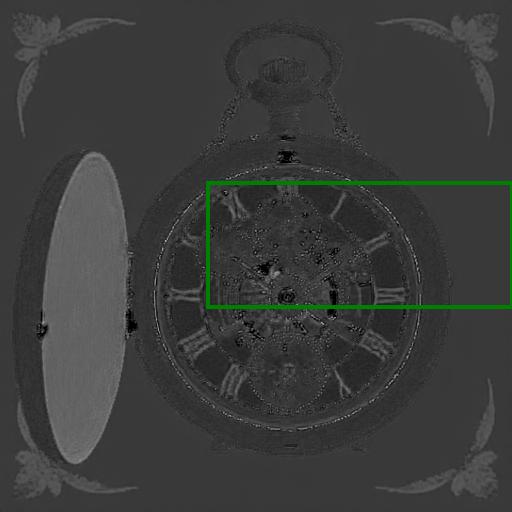}}
                \\
                \fbox{\includegraphics[width=0.046\textwidth]{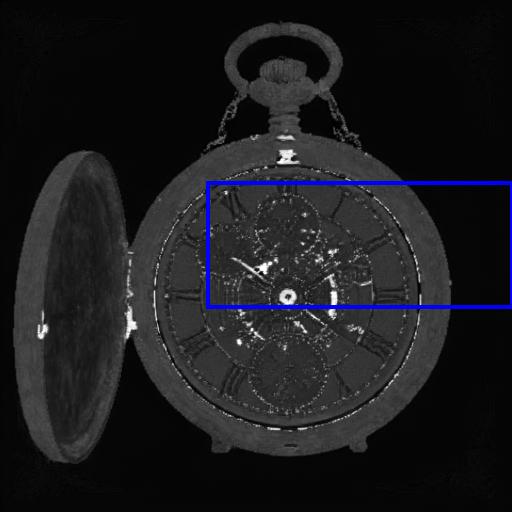}}
            \end{tabular}
            &
            \fbox{\includegraphics[width=0.10\textwidth]{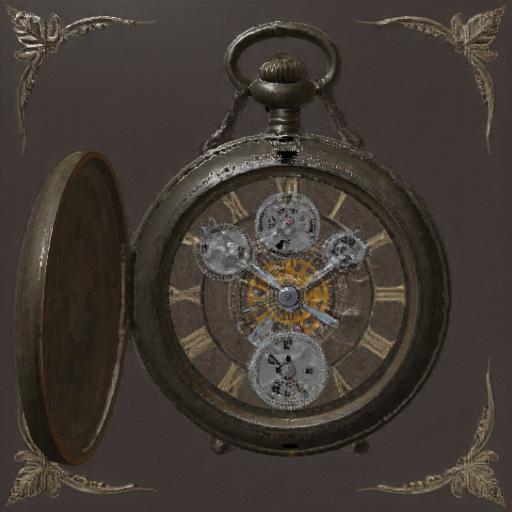}} 
            &
            \fbox{\includegraphics[width=0.10\textwidth]{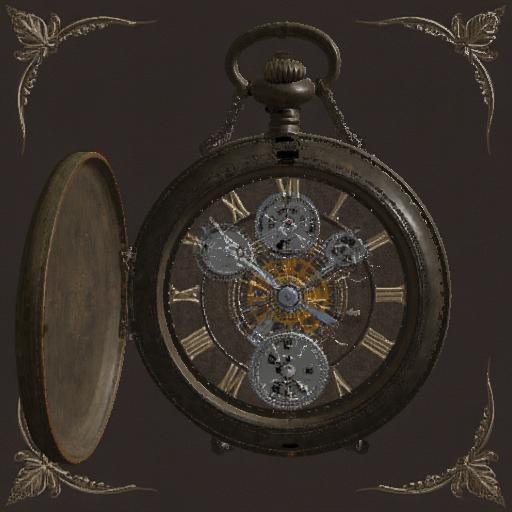}} 
        \end{tabular}}
            \\
            \resizebox{0.64\textwidth}{!}{
            \begin{tabular}[b]{cccc}
                \fbox{\includegraphics[width=0.10\textwidth]{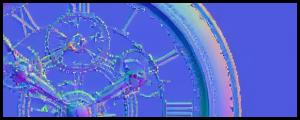}} 
                &
                \fbox{\includegraphics[width=0.10\textwidth]{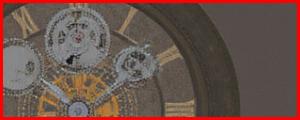}} 
                &
                \fbox{\includegraphics[width=0.10\textwidth]{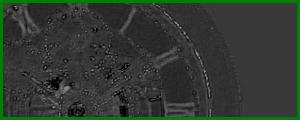}} 
                &
                \fbox{\includegraphics[width=0.10\textwidth]{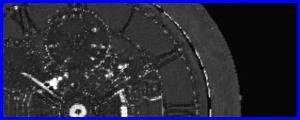}} 
            \end{tabular}}
        \end{tabular} \\

        \midrule
        \rotatebox{90}{Ours}
        &
        \begin{tabular}[b]{@{}c@{}}
            \resizebox{0.64\textwidth}{!}{
            \begin{tabular}[b]{ccccc}
            \fbox{\includegraphics[width=0.10\textwidth]{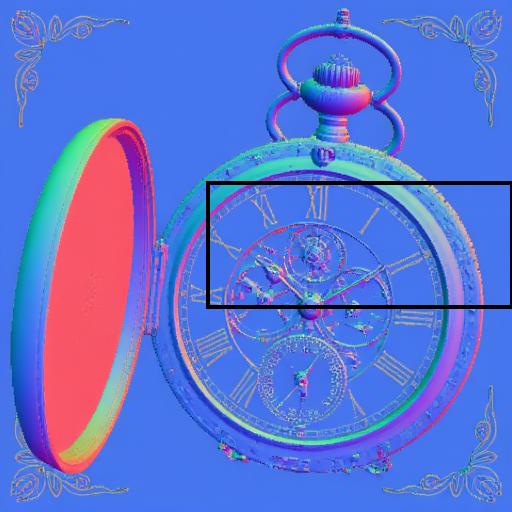}} 
            &
            \fbox{\includegraphics[width=0.10\textwidth]{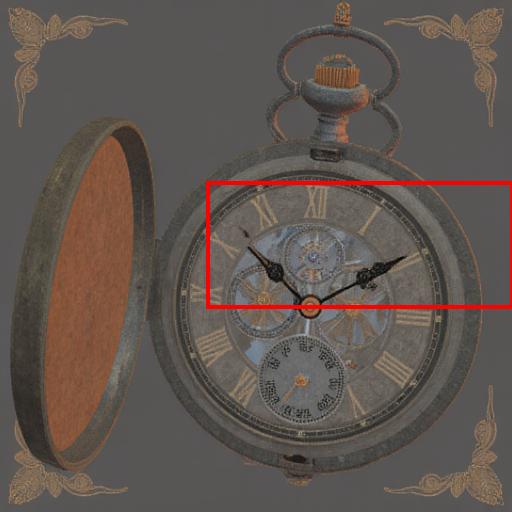}} 
            &
            \begin{tabular}[b]{@{}c@{}}
                \fbox{\includegraphics[width=0.046\textwidth]{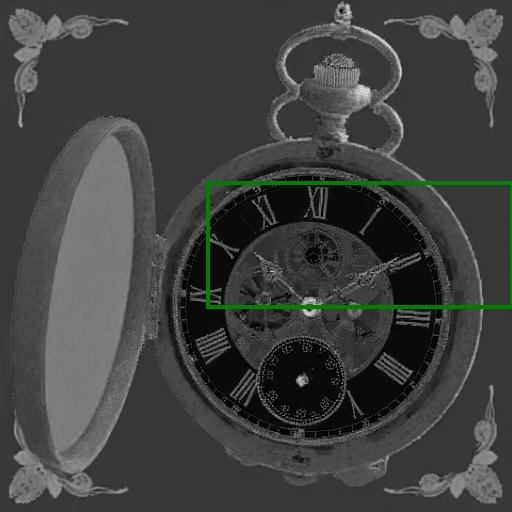}}
                \\
                \fbox{\includegraphics[width=0.046\textwidth]{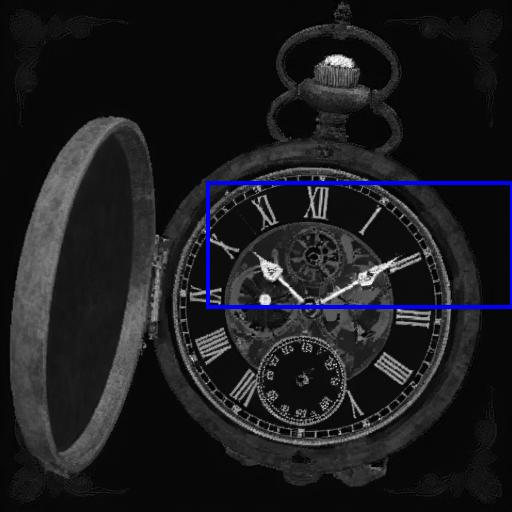}}
            \end{tabular}
            &
            \fbox{\includegraphics[width=0.10\textwidth]{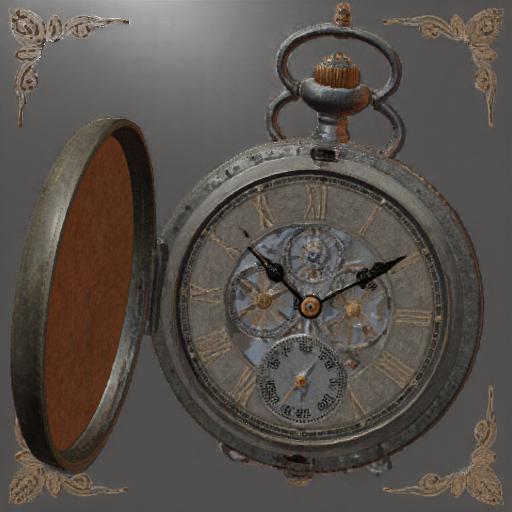}} 
            &
            \fbox{\includegraphics[width=0.10\textwidth]{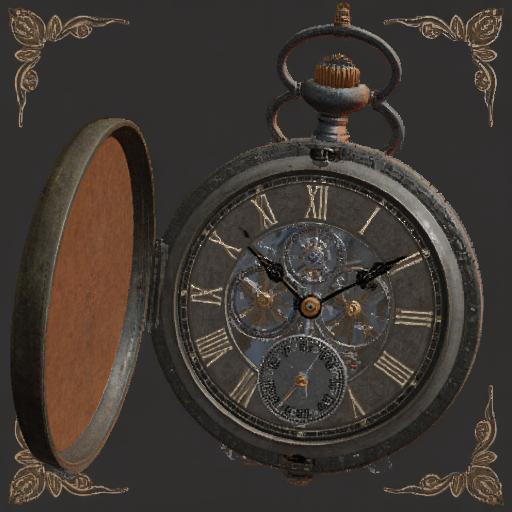}} 
        \end{tabular}}
            \\
            \resizebox{0.64\textwidth}{!}{
            \begin{tabular}[b]{cccc}
                \fbox{\includegraphics[width=0.10\textwidth]{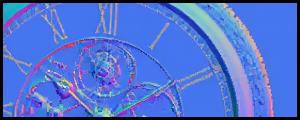}} 
                &
                \fbox{\includegraphics[width=0.10\textwidth]{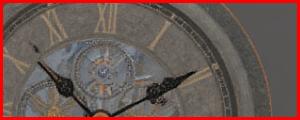}} 
                &
                \fbox{\includegraphics[width=0.10\textwidth]{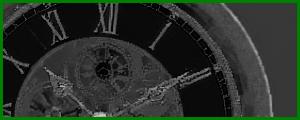}} 
                &
                \fbox{\includegraphics[width=0.10\textwidth]{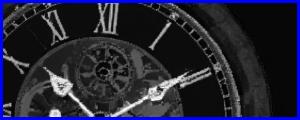}} 
            \end{tabular}}
        \end{tabular} \\

    \end{tabular}}
    \caption{\textbf{Ablations}. 
    We compare our full method against ablations that do not use the rendering loss (w/o Rendering), use uniform light sampling instead of importance-based light sampling (w/o Light Sampling), and do not use dropout in the cross-intrinsic attention (w/o CIA-Dropout).
    Without the rendering loss (\Cref{subsubsec:rend-loss}), the PBR maps lose their semantic meaning, e.g., there are baked-in shadows in the albedo and the generated images appear ``averaged out''.
    Importance-based light sampling (\Cref{subsubsec:rend-loss}) and CIA dropout (\Cref{subsubsec:cia}) both increase the sharpness of individual PBR maps, e.g., the roughness/metallic images have realistic details \textit{without} baked-in textures.
    Overall, all components improve the quality of rendered images under varied lighting conditions.
    We provide more samples in the supplement. 
    }
    \label{fig:exp:ablations}
    \vspace{-24pt}
\end{wrapfigure}

%% file: sections/5-conclusion.tex

\section{Conclusion}
\label{sec:conclusion}

We have presented IntrinsiX, the first method for \textit{direct} generation of intrinsic image properties from text as input.
We leverage the strong image prior of pretrained text-to-image models and convert it into a PBR map generator.
We have introduced cross-intrinsic attention to produce semantically aligned PBR maps. 
Furthermore, we have shown that using our novel rendering loss with tailored light sampling provides important signal for the model to better ground each intrinsic component.
Our approach allows us to generate high-quality, diverse results that go beyond the distribution of existing, synthetic datasets.
Our method enables several downstream applications, such as physically-based rendering, material editing, relighting, and for the first time 3D scene PBR texture generation.
We believe this showcases the potential that text-to-image models like ours can have on gaming and VR applications.
Instead of generating content in shaded RGB space, we produce the PBR maps that can be directly used in standard computer graphics pipelines.


%% file: sections/acknowledgements.tex
\begin{ack}
This work was supported by the ERC Consolidator Grant Gen3D (101171131) of Matthias Nießner, the German Research Foundation (DFG) Grant ``Making Machine Learning on Static and Dynamic 3D Data Practical'', and the German Research Foundation (DFG) Research Unit ``Learning and Simulation in Visual Computing''. We thank Angela Dai for the video voice-over. 
\end{ack}

%% file: sections/checklist.tex
\clearpage
\section*{NeurIPS Paper Checklist}

\begin{enumerate}

\item {\bf Claims}
    \item[] Question: Do the main claims made in the abstract and introduction accurately reflect the paper's contributions and scope?
    \item[] Answer: \answerYes{} 
    \item[] Justification: The key contributions are summarized in the abstract as well as in the last paragraph of the introduction. 

\item {\bf Limitations}
    \item[] Question: Does the paper discuss the limitations of the work performed by the authors?
    \item[] Answer: \answerYes{} 
    \item[] Justification: \Cref{sec:limitations} describes the limitations of our method. 

\item {\bf Theory assumptions and proofs}
    \item[] Question: For each theoretical result, does the paper provide the full set of assumptions and a complete (and correct) proof?
    \item[] Answer: \answerNA{} 
    \item[] Justification: The paper does not include theoretical results. 

    \item {\bf Experimental result reproducibility}
    \item[] Question: Does the paper fully disclose all the information needed to reproduce the main experimental results of the paper to the extent that it affects the main claims and/or conclusions of the paper (regardless of whether the code and data are provided or not)?
    \item[] Answer: \answerYes{} 
    \item[] Justification: We will release the training and testing codes along with our trained model weights upon acceptance. It should also be possible to reproduce the model based on our description in the paper and the supplementary. 

\item {\bf Open access to data and code}
    \item[] Question: Does the paper provide open access to the data and code, with sufficient instructions to faithfully reproduce the main experimental results, as described in supplemental material?
    \item[] Answer: \answerYes{} 
    \item[] Justification: We will release the training and testing codes along with our trained model weights upon acceptance.

\item {\bf Experimental setting/details}
    \item[] Question: Does the paper specify all the training and test details (e.g., data splits, hyperparameters, how they were chosen, type of optimizer, etc.) necessary to understand the results?
    \item[] Answer: \answerYes{} 
    \item[] Justification: \Cref{sec:experiments,sec:method} and \Cref{sec:user_study,sec:prompts} describe the the necessary training and testing details. 

\item {\bf Experiment statistical significance}
    \item[] Question: Does the paper report error bars suitably and correctly defined or other appropriate information about the statistical significance of the experiments?
    \item[] Answer: \answerYes{} 
    \item[] Justification: We report the standard deviation for our user study results (\cref{tab:exp:comparisons}). The FID metrics are evaluating a statistical similarity between two distributions (between a large amount of generated samples), thus already providing information about the statistical significance of the experiments. Additionally, we show multiple generated samples with different seeds in \Cref{fig:exp:diversity}.

\item {\bf Experiments compute resources}
    \item[] Question: For each experiment, does the paper provide sufficient information on the computer resources (type of compute workers, memory, time of execution) needed to reproduce the experiments?
    \item[] Answer: \answerYes{} 
    \item[] Justification: The first section of \Cref{sec:experiments} describes the required resources. 
    
\item {\bf Code of ethics}
    \item[] Question: Does the research conducted in the paper conform, in every respect, with the NeurIPS Code of Ethics \url{https://neurips.cc/public/EthicsGuidelines}?
    \item[] Answer: \answerYes{} 
    \item[] Justification: Our work is in accordance with the NeurIPS Code of Ethics. 

\item {\bf Broader impacts}
    \item[] Question: Does the paper discuss both potential positive societal impacts and negative societal impacts of the work performed?
    \item[] Answer: \answerYes{} 
    \item[] Justification: We provide a discussion about the societal impact in \Cref{sec:societal_impact}. 
    
\item {\bf Safeguards}
    \item[] Question: Does the paper describe safeguards that have been put in place for responsible release of data or models that have a high risk for misuse (e.g., pretrained language models, image generators, or scraped datasets)?
    \item[] Answer: \answerYes{} 
    \item[] Justification: We will add a section about ethical and out-of-scope usage in our code release similarly to \cite{flux2023} to ensure responsible usage. 

\item {\bf Licenses for existing assets}
    \item[] Question: Are the creators or original owners of assets (e.g., code, data, models), used in the paper, properly credited and are the license and terms of use explicitly mentioned and properly respected?
    \item[] Answer: \answerYes{} 
    \item[] Justification: We describe the licenses of all the assets in \Cref{sec:licenses}.

\item {\bf New assets}
    \item[] Question: Are new assets introduced in the paper well documented and is the documentation provided alongside the assets?
    \item[] Answer: \answerYes{} 
    \item[] Justification: We will release the code and the trained model weights with training and testing scripts. We provide details about the training in \Cref{sec:experiments}. We describe the licenses of all the assets in \Cref{sec:licenses}.

\item {\bf Crowdsourcing and research with human subjects}
    \item[] Question: For crowdsourcing experiments and research with human subjects, does the paper include the full text of instructions given to participants and screenshots, if applicable, as well as details about compensation (if any)? 
    \item[] Answer: \answerYes{} 
    \item[] Justification: We conduct a user-study with human participants. We describe the details in \Cref{sec:user_study} and provide one example of the provided instructions in \Cref{fig:user-study:sample-questions}. 

\item {\bf Institutional review board (IRB) approvals or equivalent for research with human subjects}
    \item[] Question: Does the paper describe potential risks incurred by study participants, whether such risks were disclosed to the subjects, and whether Institutional Review Board (IRB) approvals (or an equivalent approval/review based on the requirements of your country or institution) were obtained?
    \item[] Answer: \answerYes{} 
    \item[] Justification: We conduct a user-study with human participants. We describe the details in \Cref{sec:user_study}. Our user-study is anonymous and does not raise any risks for the participants. 

\item {\bf Declaration of LLM usage}
    \item[] Question: Does the paper describe the usage of LLMs if it is an important, original, or non-standard component of the core methods in this research? Note that if the LLM is used only for writing, editing, or formatting purposes and does not impact the core methodology, scientific rigorousness, or originality of the research, declaration is not required.
    \item[] Answer: \answerNA{} 
    \item[] Justification: The core method development in this research does not involve LLMs as any important, original, or non-standard components.

\end{enumerate}

%% file: sections/supplementary.tex
\newpage
\title{IntrinsiX: \\ High-Quality PBR Generation using Image Priors \\\vspace{2.0mm} --- Supplementary material ---}
\renewcommand\myfigure{%
    \centering
    \vbox{%
    	\vskip -0.2in
    	\hsize\textwidth
    	\linewidth\hsize
    	\centering
    	\normalsize
    	\vskip 0.2in
    }}
\settitle

\appendix
\setcounter{page}{1}

\vspace{-24pt}
\section{Additional Ablations}

\paragraph{How important is the dataset size and diversity?}

In the first stage of training, we train 3 separate LoRAs, corresponding to the different intrinsic properties.
We curate synthetic indoor scene examples from the InteriorVerse dataset \cite{InteriorVerse}.
We empirically find that we need a large dataset size for the roughness/metallic PBR maps to achieve reasonable understanding of the corresponding intrinsic distribution.
In contrast, the albedo/normal maps can be learned from a much smaller dataset of only 20 samples.
This is important to retain the generalizable prior of the pretrained text-to-image model (see \Cref{sec:suppl:first-stage}).
We confirm this with additional experiments in \Cref{tab:exp:datset}, that compare the quality and diversity of generated albedo images for different dataset sizes.
The in-distribution FID (A-ID-FID) measures the quality of the albedo (calculated on a subset of 100 test images of InteriorVerse \cite{InteriorVerse}, similar as in the main paper).
The diversity metric (A-Diversity) compares the FID between the generated set of all images and the mean of the generated set.
This measures if the distribution is collapsed and therefore signals how diverse the generated samples are.
We can see that a dataset consisting of 20 samples does the best in terms of diversity, while still having reasonable albedo quality.
Importantly, albedos trained on larger datasets also start to include baked-in lighting effects (see \Cref{fig:stage1:dataset_size}).
This motivates our choice to not increase the dataset size further.
The final dataset consists of sampled images from the InteriorVerse dataset \cite{InteriorVerse}.
We sample images from the following room-types to curate 20 samples: 5 bedrooms, 5 kitchens, 5 livingrooms, 1 kidroom, 2 offices, 1 cabinet, 1 bathroom.

\input{figures/stage1/lora_rank}

\paragraph{LoRA Rank}

We ablate the rank of the LoRA \cite{LoRA} modules in \cref{fig:stage1:rank}. 
The rank determines the total number of trainable parameters. 
Therefore, with a too low rank, the number of trainable parameters are too low in order to achieve the domain shift. 
On the other hand, with a too high rank, we are introducing too many parameters, which will lead to forgetting; thus, negatively effecting the generalization. 
As a middle-ground, we chose rank 64.

\paragraph{Can we maintain sample diversity?}

\input{tables/experiments/dataset}
\input{figures/stage1/dataset_size}

\input{figures/experiments/diversity}

We show multiple samples using the same text prompt in \Cref{fig:exp:diversity}.
Our method manages to maintain the generalization capabilities of the T2I model and generates diverse samples even for out-of-distribution prompts (see also \Cref{fig:exp:comparisons} and the supplementary material).

\paragraph{More samples}
We show additional qualitative comparisons in \Cref{fig:supp:ablations}. 
\input{figures/experiments/ablations_more}

\paragraph{Individual PBR Priors}
\label{sec:suppl:first-stage}
\input{figures/stage1/samples}

In the first stage of training, we train 3 separate LoRAs, corresponding to the different intrinsic properties.
We curate synthetic indoor scene examples from the InteriorVerse dataset \cite{InteriorVerse}.
We show in \Cref{fig:stage1:samples}~(top) that this leads to high-quality and diverse albedo and normal map generations.
This confirms our choice of training these PBR maps on small-scale datasets, i.e., we retain the generalized prior of the pretrained text-to-image model during the first stage finetuning.

In contrast, the roughness/metallic LoRAs fail to generalize to out-of-distribution scenarios.
This is because we use a larger dataset for training this LoRA.
However, \Cref{fig:stage1:samples}~(bottom) shows that the second stage alignment training turns this LoRA to an equally-well generalizable PBR map generator.
In other words, the generalizability of the albedo/normal LoRAs can be combined with the understanding of the intrinsic distribution of the roughness/metallic LoRA.
Together, we can still produce high-quality, diverse PBR maps.

\section{Additional Results}

\paragraph{Lighting Direction Sampling}
We sample light directions that maximize specular highlights to provide strong gradients about reflectance. 
As an alternative, we train a BRDF-sampled variant (Disney model) for more diverse lighting directions. This yields slightly worse results on the in-domain dataset ($75.26$ A-ID-FID), but further improves generalization ($68.87$ A-OOD-FID), showing that lighting direction sampling is crucial for our task. Exploring other sampling strategies is a great avenue for future research. 

\paragraph{Baseline comparisons}
We show additional comparisons to the baselines in \Cref{fig:supp:comparisons}. 
\input{figures/experiments/comparisons_more}

\paragraph{Albedo comparisons}
We show additional albedo comparisons to the baselines in \Cref{fig:supp:albedo_comparisons}. 
\input{figures/experiments/albedo_comparisons_more}

\paragraph{Scene Texturing Results}
\label{sec:suppl:scenetex}
\input{figures/applications/scenetex_more}
We show more scene texturing results in \Cref{fig:supp:scenetex_more}. 
We used Blender \cite{Blender} to render the scene with uniform white environment map lighting and a single spherical light source. 
To enhance geometric details, we used an approximation of the displacement map by thresholding the normal textures. 

To achieve room-scale scene texturing, we apply the SceneTex method \cite{SceneTex} in a two-stage manner with the conditional variant of our model.
First, we render normal maps from multiple views and generate material (albedo, roughness and metallic) textures, conditioned on the rendered normals. 
Then, we render the material properties for the given views and generate fine-grained normal details, conditioned on the rendered material maps. 
We use VFDS \cite{li2024flowdreamer} \cite{li2024flowdreamer} loss in image space. 
To balance the updates between over- and under-sampled texels, we weight the lostt with the inverse obervations frequency. 
As a pre-processing step, we create a texture, which stores the texel observation frequency. During the optimization, we render this texture together with the other components apply the weighting pixel-wise. 
We found that too low CFG value causes over-smoothed results, while too high values can break the generated images in case of Flux \cite{flux2023}. 
To solve this issue, we normalize the flow direction to keep the norm of the text-conditional prediction, but use the direction towards the extrapolated flow direction.

\section{Limitations}
\input{figures/experiments/limitations}

\label{sec:limitations}
We use a screen-space renderer similar to \cite{IID,InteriorVerse}. For better 2D results, a neural/diffusion renderer can optionally be trained on top of our method. 
Our method maintains generalization far beyond the minimal training set, thanks to formulating the task directly in PBR space. Since FLUX does not inherently know about intrinsic components, we sacrifice some compositional diversity to enable our task using LoRA modules (see \Cref{fig:limitations}). Ultimately, training with a diverse billion-scale PBR dataset would further improve generalization, but it does not exist.

\section{Societal Impact}
\label{sec:societal_impact}
Generative models impact the society in general. Next to accelerating and democratizing creative content creation, they can also raise ethical concerns. Potential misuse for generating misinformation or deepfakes can become a major threat for naive users. These risks needs to be discussed and made public as soon as possible with open-sourcing and publishing results in the field to show the limits of current state-of-the-art. These challenges have been widely discussed in the recent years, such as in \cite{DBLP:journals/corr/abs-2403-04667}. Our method enables decomposed generation, which coupled with a photo-realistic rendering can produce realistic-looking results.

\section{Licenses}
\label{sec:licenses}

\input{tables/licenses}

\Cref{tab:supp:licenses} shows a summary about the licenses of the used components. 

\section{User Study}
\label{sec:user_study}
\input{figures/user_study/sample_questions}

To better evaluate the quality of our generated PBR maps, we conduct a user study. 
The participation is anonymous and no personal data is collected. 
We summarize in \Cref{fig:user-study:sample-questions} the questions we asked by the participants.
In the following, we explain how each metric is calculated.

\begin{itemize}[leftmargin=*,topsep=1pt, noitemsep]
\item A-PP: we calculate the perceptual preference of albedo images (see \Cref{fig:user-study:sample-questions}~top). Users choose one of the images and we calculate in percentage how often each method was preferred.
\item S-PQ: we calculate the quality of specularity of the rendered video under varying lighting conditions (see \Cref{fig:user-study:sample-questions}~bottom). Users rate on a scale of 1-5 how good the specular quality is.
\item R-PQ: we calculate the general quality of the rendered video under varying lighting conditions (see \Cref{fig:user-study:sample-questions}~bottom). Users rate on a scale of 1-5 how good the general quality is.
\item PC: we calculate the prompt coherence, i.e, how well the text prompt matches the rendered video (see \Cref{fig:user-study:sample-questions}~bottom). Users rate on a scale of 1-5 how good the coherence is.
\end{itemize}

\section{Prompts}
\label{sec:prompts}
We used the following prompts in our main results. 
We used our own, LLM-generated prompts, and prompts from \citet{gao2024cat3d}:
\begin{itemize}[leftmargin=*,topsep=1pt, noitemsep]
\item \Cref{fig:teaser}: ``An astronaut riding a unicorn on the moon''
\item \Cref{fig:method:pipeline}: ``An astronaut riding a unicorn on the moon''
\item \Cref{fig:exp:editable_image_generation}: ``Astronaut in front of landscape space alien planet''
\item \Cref{fig:exp:scenetex}: ``An industrial-style room with exposed brick walls, and reclaimed wood furniture, The room features a leather sofa, a coffee table made from a metal frame, and modern decor that complements its raw, edgy vibe''
\item \Cref{fig:exp:comparisons} from left to right and top to bottom: 
``A wooden treasure chest reinforced with golden bands, its lid slightly ajar to reveal glittering jewels and coins, with faint beams of light spilling out from inside'', 
``3d cartoon folk singers character music guitar animation'', 
\item \Cref{fig:exp:albedo_comparisons} from left to right: 
``3d cartoon boy character animation'', 
``Adventurer standing in forest exploration nature trees hiking woodland outdoor'', 
``Adventurous teddy bear explorer travel outdoor'', 
``Alpaca wearing a suit animal clothing formal wool'', 
``Anime character in lab coat scientist cartoon drawing japanese style'', 
\item \Cref{fig:exp:ablations}: ``A vintage pocket watch with its cover open, revealing a complex arrangement of gears and springs, some of which are glowing faintly, surrounded by engraved floral patterns.''
\item \Cref{fig:stage1:rank}: ``An astronaut riding a unicorn on the moon''
\item \Cref{fig:stage1:dataset_size}: ``An astronaut riding a unicorn on the moon''
\item \Cref{fig:exp:diversity}: ``A wooden chair''
\item \Cref{fig:supp:ablations}: ``A sportcar''
\item \Cref{fig:stage1:samples} from left to right: 
``An astronaut riding a unicorn on the moon'', 
``3d cartoon folk singers character music guitar animation'',
``Alien merchant extraterrestrial market fantasy science fiction'',
``Alley city urban narrow passage architecture outdoor'',
``Astronaut in front of landscape space alien planet'',
``A majestic castle made entirely of ice, perched atop a snowy hill with shimmering pink and golden light reflecting off its towers. Below, a frozen lake mirrors the grandeur of the scene'',
``A sprawling library with towering bookshelves reaching to the ceiling, glowing orbs floating mid-air to provide light, books that seem to fly on their own, and a spiral staircase made of golden wood.'',
``A house in a forest'',
``New York'',
``A wooden chair''
\item \Cref{fig:supp:comparisons} from left to right and top to bottom: 
``A rusted sword with a glowing blue rune etched into the blade, its hilt wrapped in weathered leather, and a faint aura of light surrounding it as if imbued with ancient magic'', 
``An astronaut riding a unicorn on the moon'', 
``A large, ornate key made of silver, with intricate vine-like patterns etched along the shaft and a glowing emerald embedded in the handle'', 
``Taj Mahal''
\item \Cref{fig:supp:albedo_comparisons} from left to right: 
``Arches national park nature rock formations desert travel outdoor'', 
``Astronaut in colorful cave exploration adventure discovery geology outdoor'', 
``An epic battlefield where knights in shining armor clash with dragon-riding warriors under a stormy sky. A massive fire-breathing dragon is mid-flight, casting shadows over the chaos below'', 
``A massive sea turtle with a forest on its back swims through crystal-clear waters, accompanied by schools of colorful fish. A small sailing ship navigates beside it, dwarfed by the turtle's size'', 
``A sleek, metallic helmet with a reflective visor that glows neon blue, featuring angular designs and small vents that emit a soft, white mist''
\item \Cref{fig:limitations}: ``A very strange burger food unusual creative''
\item \Cref{fig:supp:scenetex_more} from top to bottom: 
``An opulent Baroque-style room with intricate details, Walls are decorated with elaborate molding, in shades of cream, gold, and soft pastels, A plush velvet sofa, A richly patterned Persian rug covers the marble floor'', 
``An industrial-style room with exposed brick walls, and reclaimed wood furniture, The room features a leather sofa, a coffee table made from a metal frame, and modern decor that complements its raw, edgy vibe'', 
``A Tuscan-style room with warm earthy tones, terracotta tiles, and wrought iron details, The furniture features rich wood frames and soft cushions, complemented by Mediterranean-inspired decor'', 
``A breathtaking Greek-style room with intricate details, featuring a serene blue-and-white color scheme, Majestic marble columns with ornate Corinthian capitals support a high, coffered ceiling adorned with classical frescoes, The walls showcase elegant friezes and gold-accented moldings, reflecting the grandeur of ancient Greece, Large arched windows allow soft, natural light to flood the space, enhancing the contrast between crisp white walls and rich blue decorative elements, A luxurious chaise lounge with blue  upholstery sits, accompanied by a marble-topped table with delicate carvings, The floor is adorned with intricate mosaic patterns''
\end{itemize}

%% file: figures/stage1/lora_rank.tex
\begin{figure}[b]
    \centering
    \setlength\tabcolsep{1.25pt}
    \resizebox{0.8\textwidth}{!}{
    \fboxsep=0pt
        \begin{tabular}{cccccc} 
            &\fbox{\includegraphics[width=0.15\columnwidth]{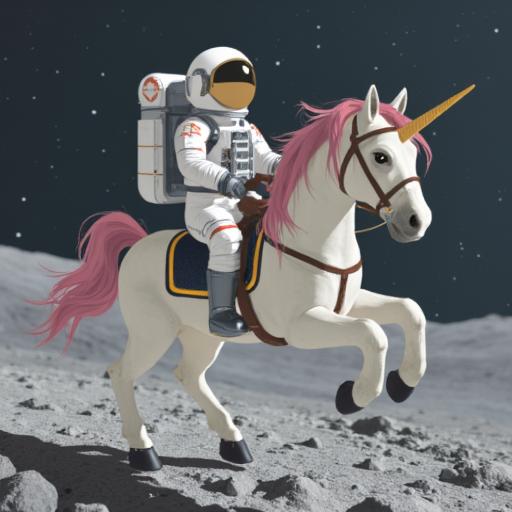}}
            &
            \fbox{\includegraphics[width=0.15\columnwidth]{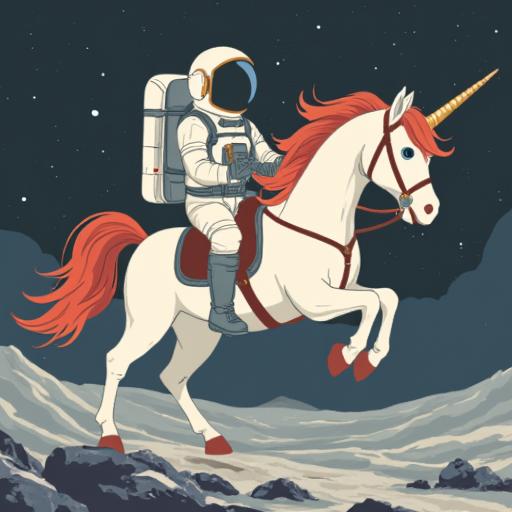}}
            &
            \fbox{\includegraphics[width=0.15\columnwidth]{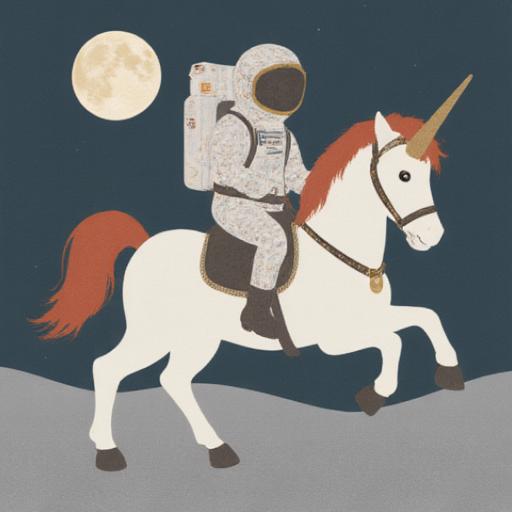}}
            &
            \fbox{\includegraphics[width=0.15\columnwidth]{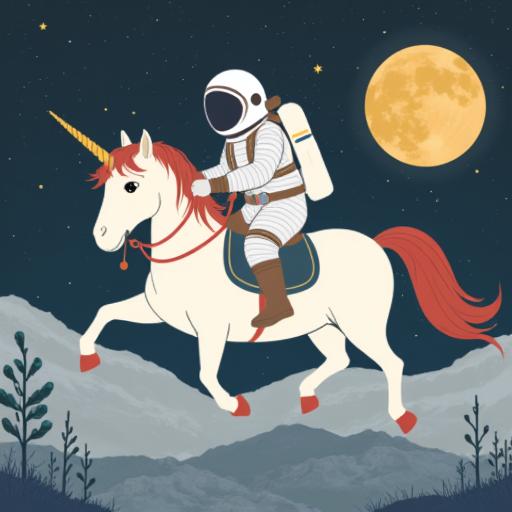}}
            &
            \fbox{\includegraphics[width=0.15\columnwidth]{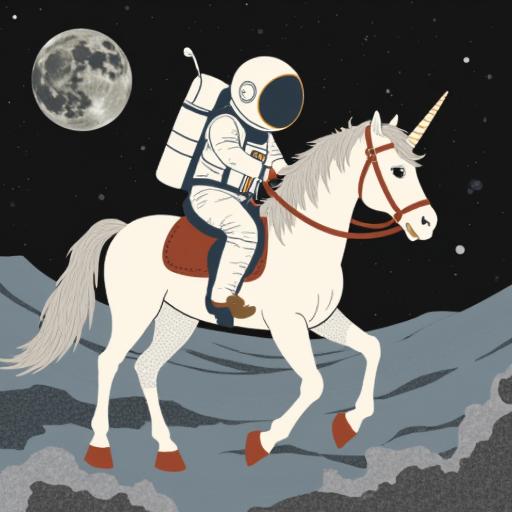}}
            \\
            
            Rank & 16 & 32 & 64 (Ours) & 128 & 256
            \\

            A-OOD-FID & 68.52 & 70.07 & \textbf{67.25} & 68.48 & 71.13
        \end{tabular}}
    \caption{\textbf{Qualitative and quantitative comparison of albedo quality for different LoRA \cite{LoRA} ranks.}
    Too low rank fails to change the domain from rgb to the target albedo modality. On the other hand, a too high rank negatively impacts the generalization, resulting in unrealistic composition. 
    }
    \label{fig:stage1:rank} 
\end{figure}

%% file: tables/experiments/dataset.tex
\begin{table}
    \begin{center}
    \caption{\textbf{Quantitative comparison of albedo quality for different dataset sizes.}
    We observe that training with larger dataset might lead to slightly better albedo quality (A-ID-FID); however, the diversity (A-Diversity) and thus the generalization capabilities degrade.
    This motivates our choice for a small, curated dataset of 20 samples for the first stage finetuning of the albedo/normal LoRAs.
    }
    \label{tab:exp:datset}
    \vspace{9pt}
    \begin{tabular}{lccc}
      \toprule
       Dataset size & A-ID-FID $\downarrow$ & A-Diversity $\uparrow$  \\
       \midrule 
      10  & 220.28 & 284.93 \\
      20 (Ours) & 187.83 & 
      \textbf{398.36}\\
      100  & 161.51 & 369.43 \\
      1k  & \textbf{154.58} & 366.35 \\
      20k  & 155.64 & 352.04  \\
      \bottomrule
    \end{tabular}
    \end{center}
\end{table}

%% file: figures/stage1/dataset_size.tex
\begin{figure}
    \centering
    \setlength\tabcolsep{1.25pt}
    \resizebox{0.5\textwidth}{!}{
    \fboxsep=0pt
        \begin{tabular}{ccccc} 
            \fbox{\includegraphics[width=0.15\columnwidth]{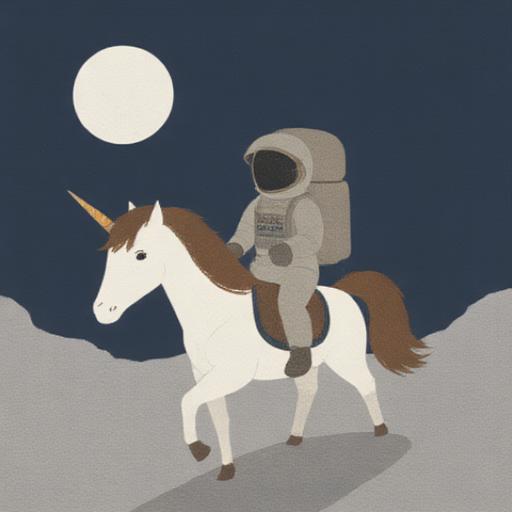}}
            &
            \fbox{\includegraphics[width=0.15\columnwidth]{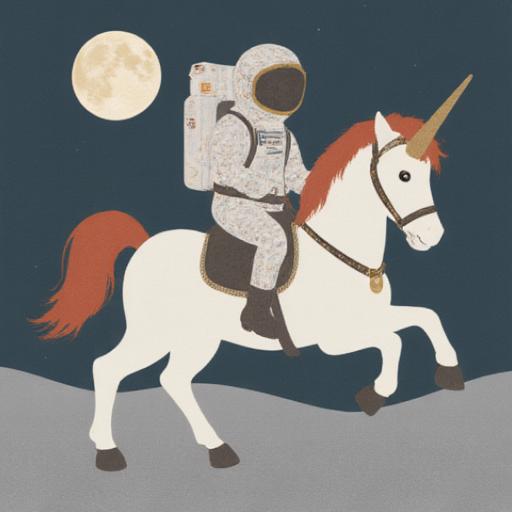}}
            &
            \fbox{\includegraphics[width=0.15\columnwidth]{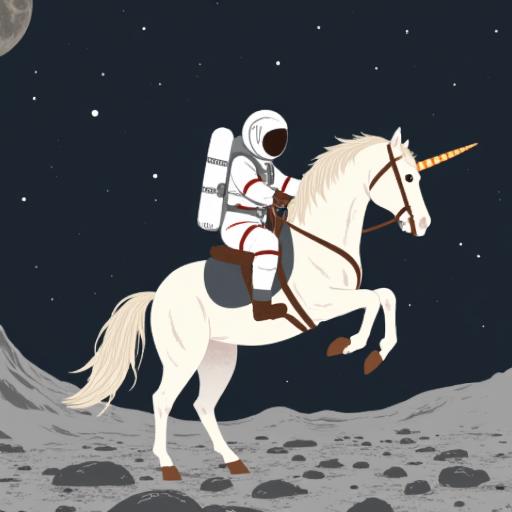}}
            &
            \fbox{\includegraphics[width=0.15\columnwidth]{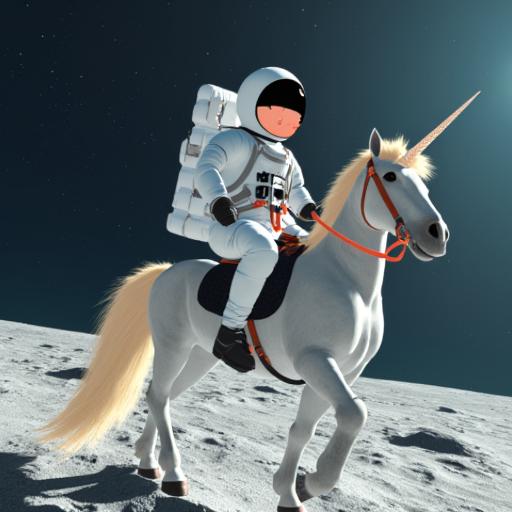}}
            &
            \fbox{\includegraphics[width=0.15\columnwidth]{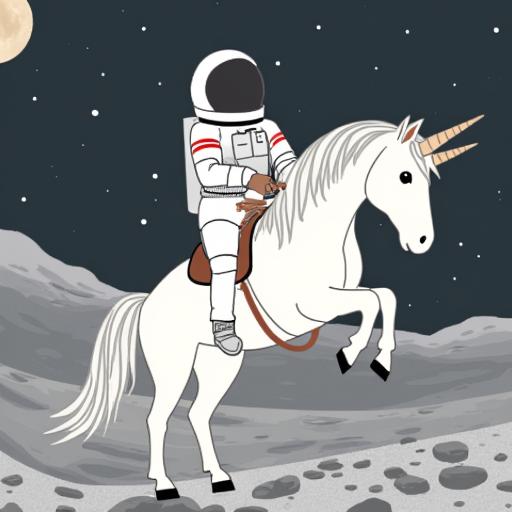}}
            \\
            \fbox{\includegraphics[width=0.15\columnwidth]{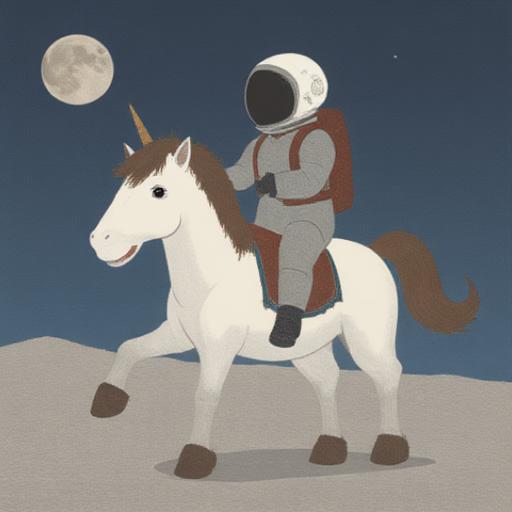}}
            &
            \fbox{\includegraphics[width=0.15\columnwidth]{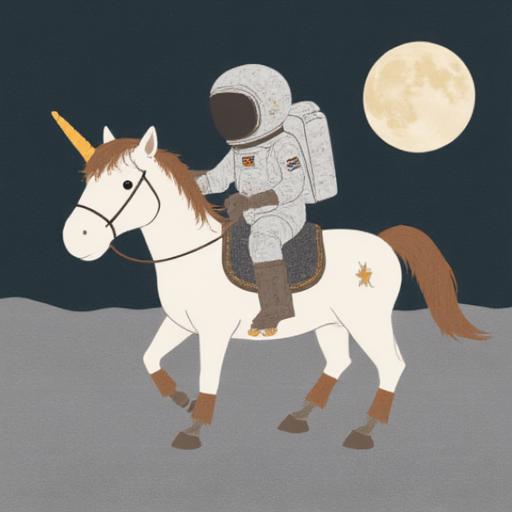}}
            &
            \fbox{\includegraphics[width=0.15\columnwidth]{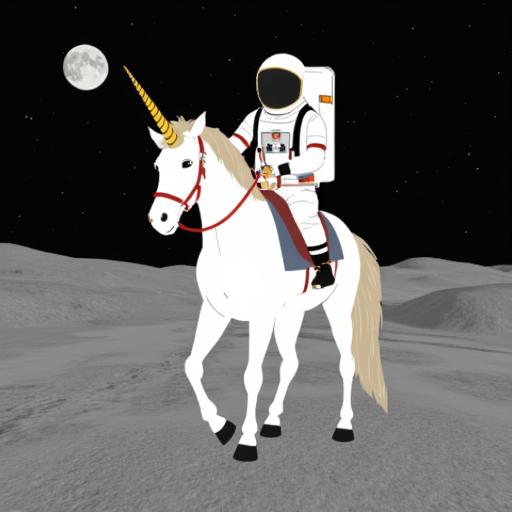}}
            &
            \fbox{\includegraphics[width=0.15\columnwidth]{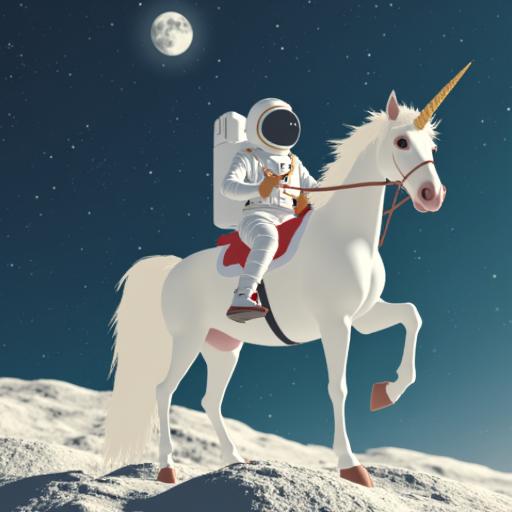}}
            &
            \fbox{\includegraphics[width=0.15\columnwidth]{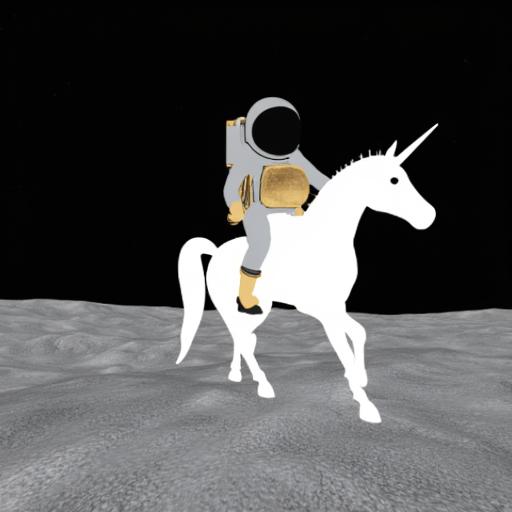}}
            \\
            \fbox{\includegraphics[width=0.15\columnwidth]{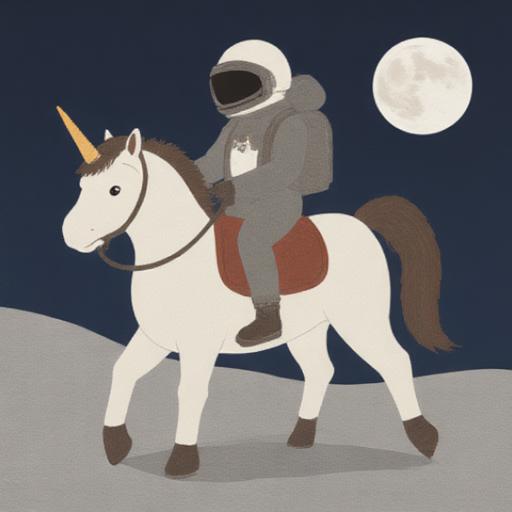}}
            &
            \fbox{\includegraphics[width=0.15\columnwidth]{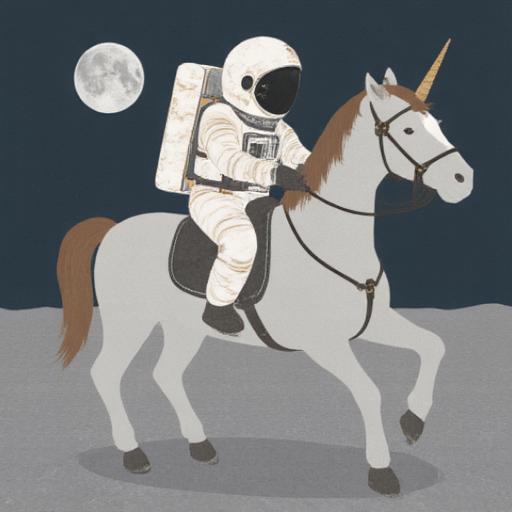}}
            &
            \fbox{\includegraphics[width=0.15\columnwidth]{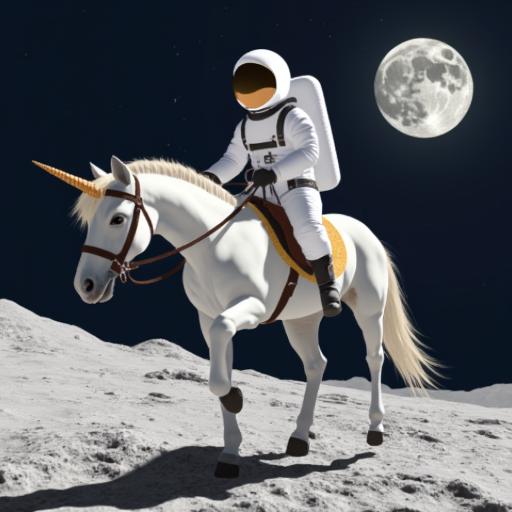}}
            &
            \fbox{\includegraphics[width=0.15\columnwidth]{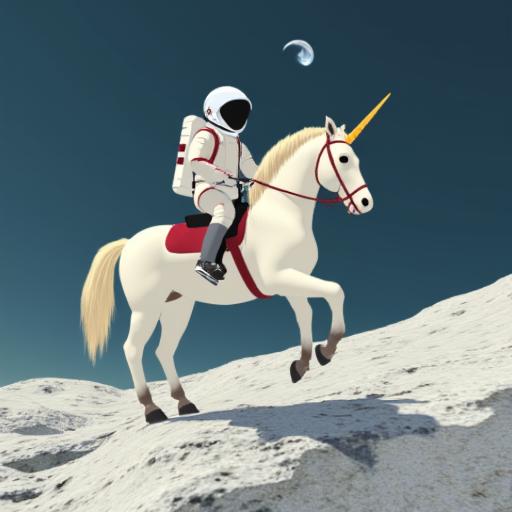}}
            &
            \fbox{\includegraphics[width=0.15\columnwidth]{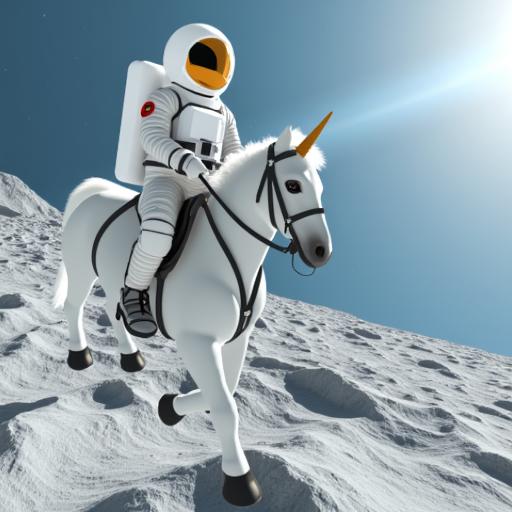}}
            \\
            \midrule
            \fbox{\includegraphics[width=0.15\columnwidth]{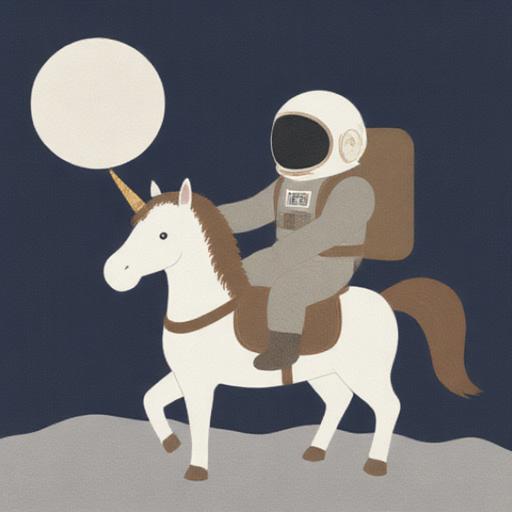}}
            &
            \fbox{\includegraphics[width=0.15\columnwidth]{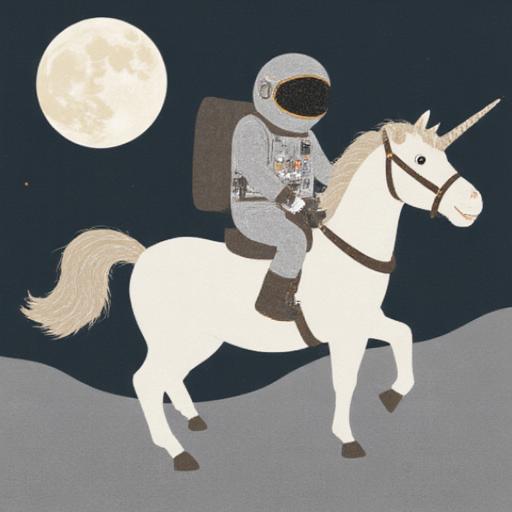}}
            &
            \fbox{\includegraphics[width=0.15\columnwidth]{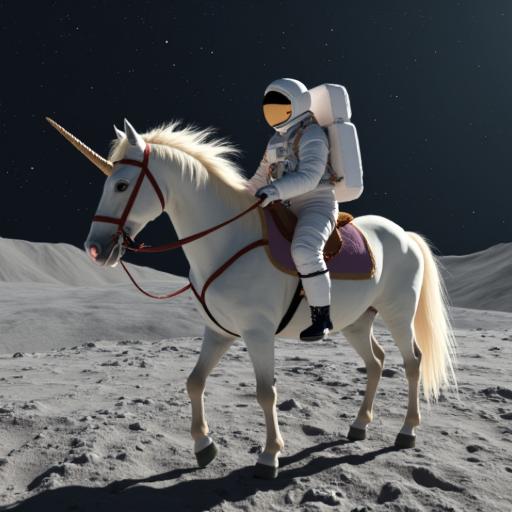}}
            &
            \fbox{\includegraphics[width=0.15\columnwidth]{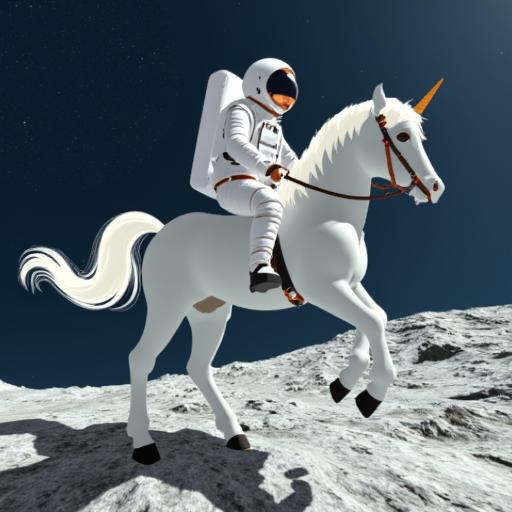}}
            &
            \fbox{\includegraphics[width=0.15\columnwidth]{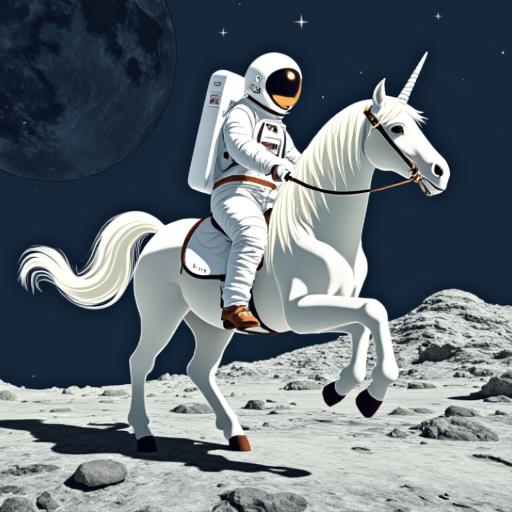}}
            \\

            10 & 20 (Ours) & 100 & 1k & 20k
        \end{tabular}}
    \caption{\textbf{Qualitative comparison of albedo quality for different dataset sizes.}
    Dataset sizes of 100 or more images tend to generate albedos with baked-in lighting effects, which is undesirable for physically-based rendering.
    A dataset that only consists of 10 images shows less details in generated albedos.
    This motivates our usage of 20 curated samples in the albedo/normal LoRA training, which balances both extrema.
    We show multiple samples per column, corresponding to different generations from the same text prompts.
    This highlights, that our model creates diverse images.
    }
    \label{fig:stage1:dataset_size} 
\end{figure}

%% file: figures/experiments/diversity.tex
\begin{wrapfigure}{r}{0.5\textwidth}
    \centering
    \setlength\tabcolsep{1.25pt}
    \resizebox{0.5\textwidth}{!}{
    \fboxsep=0pt
        \begin{tabular}{cccc}

        \fbox{\includegraphics[width=0.13\textwidth]{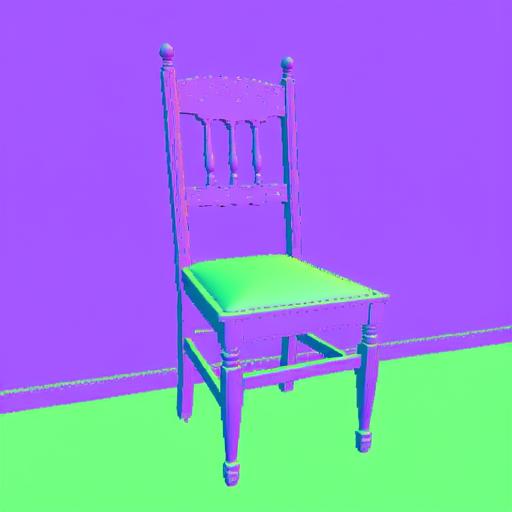}} 
        &
        \fbox{\includegraphics[width=0.13\textwidth]{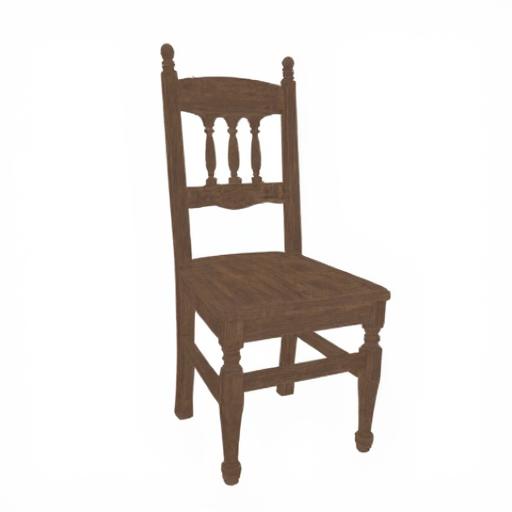}} 
        &
        \fbox{\includegraphics[width=0.13\textwidth]{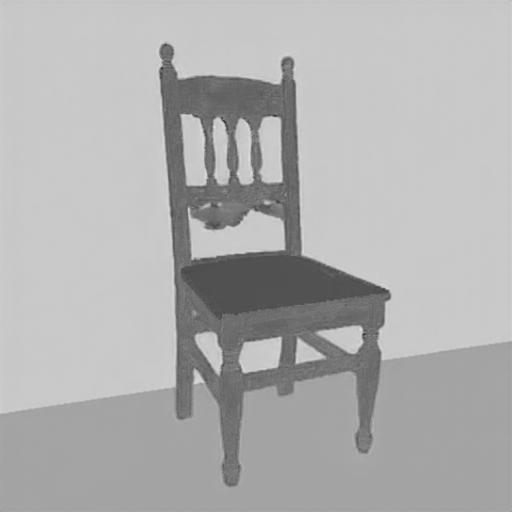}} 
        &
        \fbox{\includegraphics[width=0.13\textwidth]{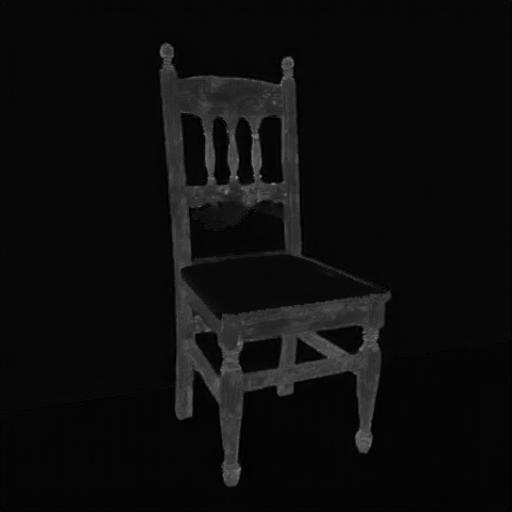}} 
        \\

        \fbox{\includegraphics[width=0.13\textwidth]{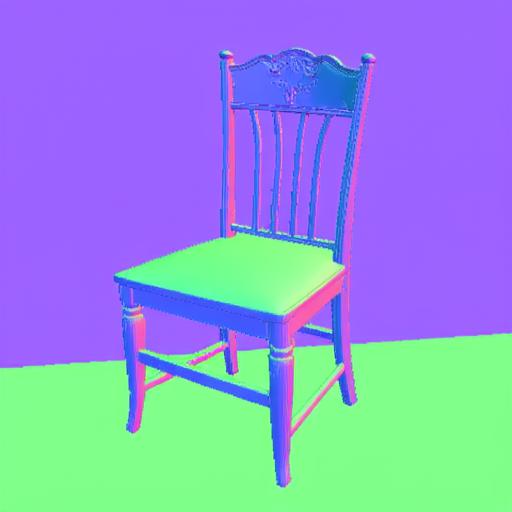}} 
        &
        \fbox{\includegraphics[width=0.13\textwidth]{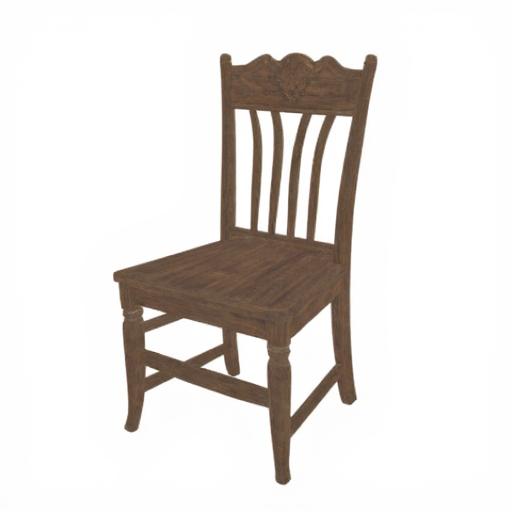}} 
        &
        \fbox{\includegraphics[width=0.13\textwidth]{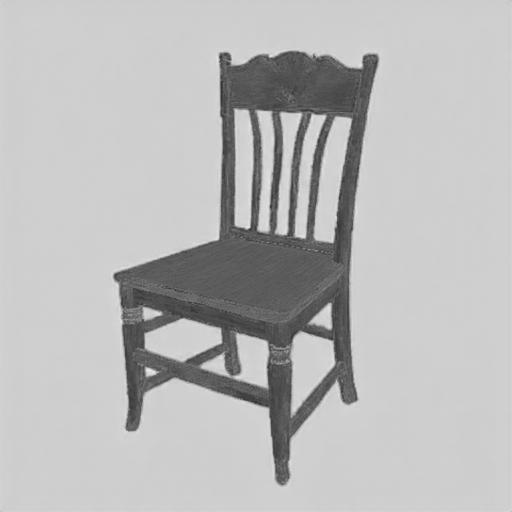}} 
        &
        \fbox{\includegraphics[width=0.13\textwidth]{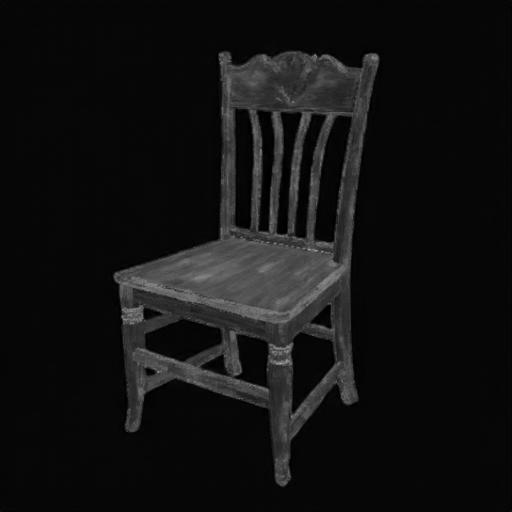}} 
        \\
        
        \fbox{\includegraphics[width=0.13\textwidth]{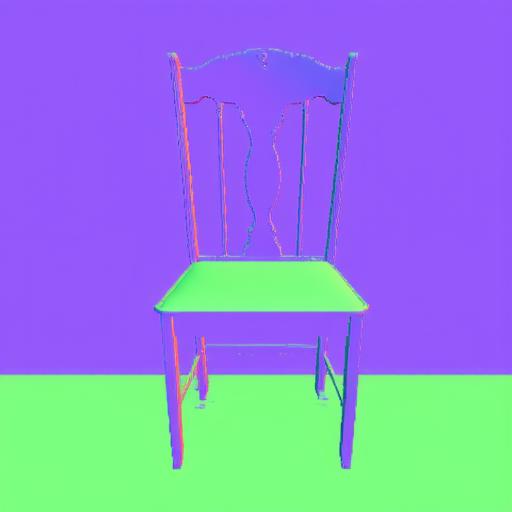}} 
        &
        \fbox{\includegraphics[width=0.13\textwidth]{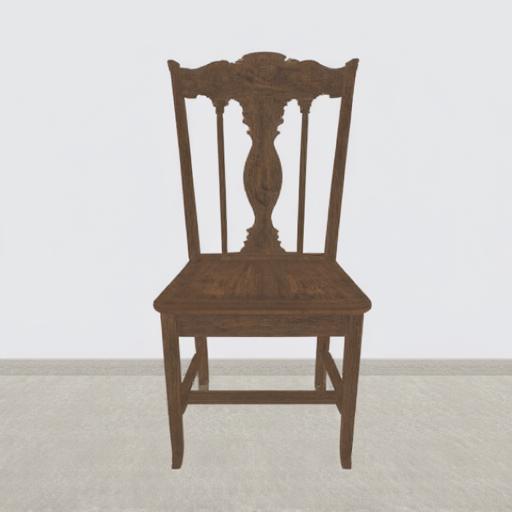}} 
        &
        \fbox{\includegraphics[width=0.13\textwidth]{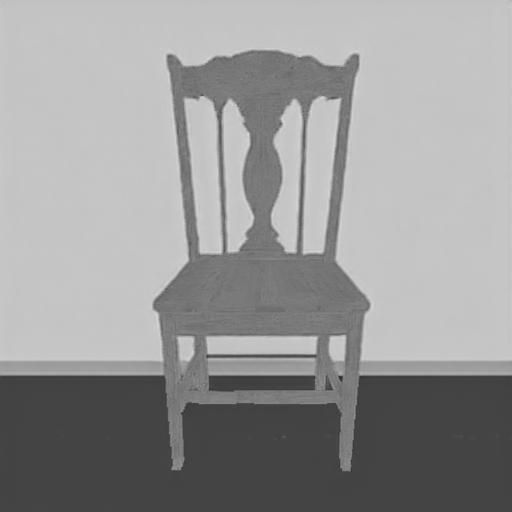}} 
        &
        \fbox{\includegraphics[width=0.13\textwidth]{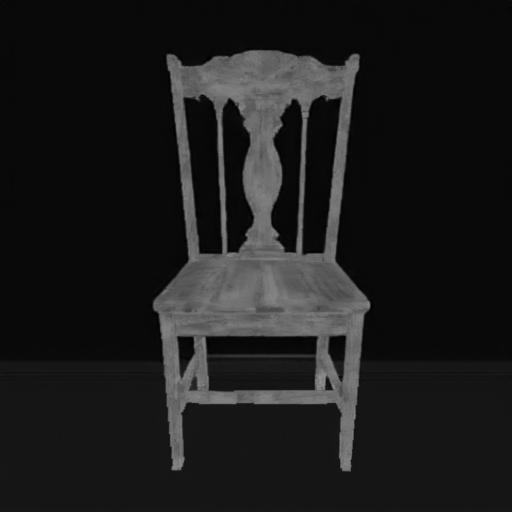}} 
        \\

        Normal &
        Albedo &
        Roughness &
        Metallic 
    \end{tabular}}
    \caption{\textbf{Sample diversity}. 
    We show 3 generated samples using the same text prompt.
    Our model predicts different samples and maintains the diversity of the T2I backbone (numerous chairs were not seen during training).
    }
    \label{fig:exp:diversity}
    \vspace{-36pt}
\end{wrapfigure}

%% file: figures/experiments/ablations_more.tex
\begin{figure}[t]
    \centering
    \setlength\tabcolsep{1.25pt}
    \resizebox{0.7\columnwidth}{!}{
    \fboxsep=0pt
    \begin{tabular}[b]{cc}
        \rotatebox{90}{w/o Rendering}
        &
        \begin{tabular}[b]{@{}c@{}}
            \resizebox{0.7\textwidth}{!}{
            \begin{tabular}[b]{ccccc}
            \tiny{Normal} &
            \tiny{Albedo} & 
            \tiny{R/M} &
            \tiny{Lighting 1} &
            \tiny{Lighting 2}
            \\
            \fbox{\includegraphics[width=0.10\textwidth]{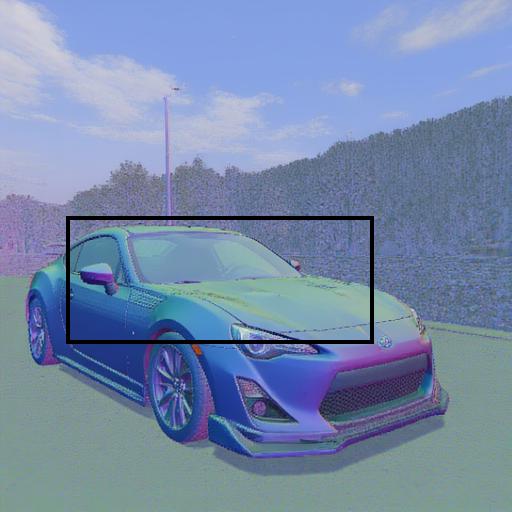}} 
            &
            \fbox{\includegraphics[width=0.10\textwidth]{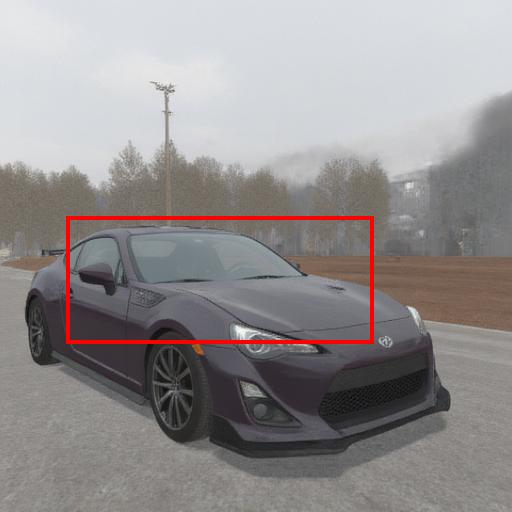}} 
            &
            \begin{tabular}[b]{@{}c@{}}
                \fbox{\includegraphics[width=0.046\textwidth]{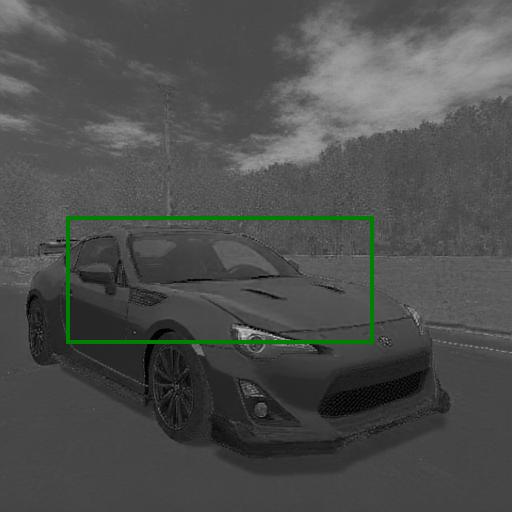}}
                \\
                \fbox{\includegraphics[width=0.046\textwidth]{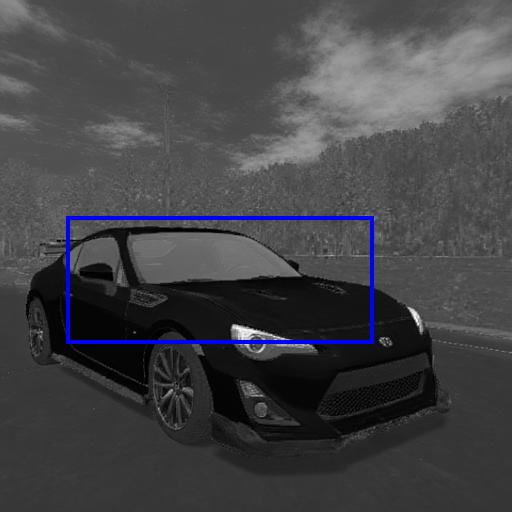}}
            \end{tabular}
            &
            \fbox{\includegraphics[width=0.10\textwidth]{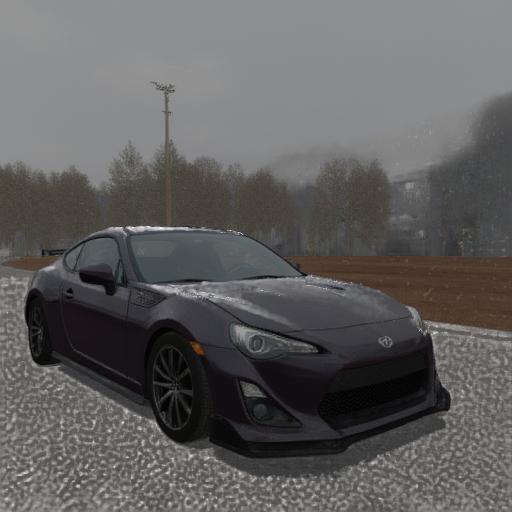}} 
            &
            \fbox{\includegraphics[width=0.10\textwidth]{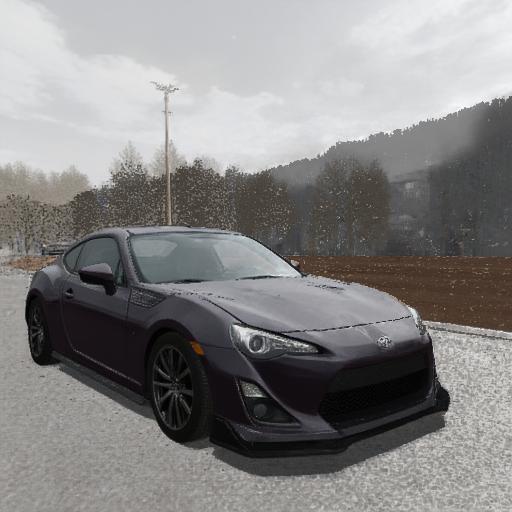}} 
        \end{tabular}}
            \\
            \resizebox{0.7\textwidth}{!}{
            \begin{tabular}[b]{cccc}
                \fbox{\includegraphics[width=0.10\textwidth]{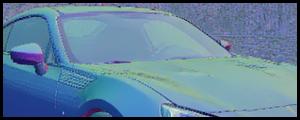}} 
                &
                \fbox{\includegraphics[width=0.10\textwidth]{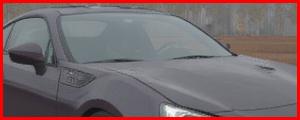}} 
                &
                \fbox{\includegraphics[width=0.10\textwidth]{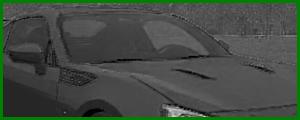}} 
                &
                \fbox{\includegraphics[width=0.10\textwidth]{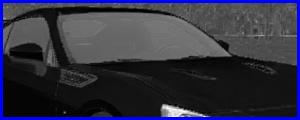}} 
            \end{tabular}}
        \end{tabular}
        \\
                
        \rotatebox{90}{w/o Light Sampling}
        &
        \begin{tabular}[b]{@{}c@{}}
            \resizebox{0.7\textwidth}{!}{
            \begin{tabular}[b]{ccccc}
            \fbox{\includegraphics[width=0.10\textwidth]{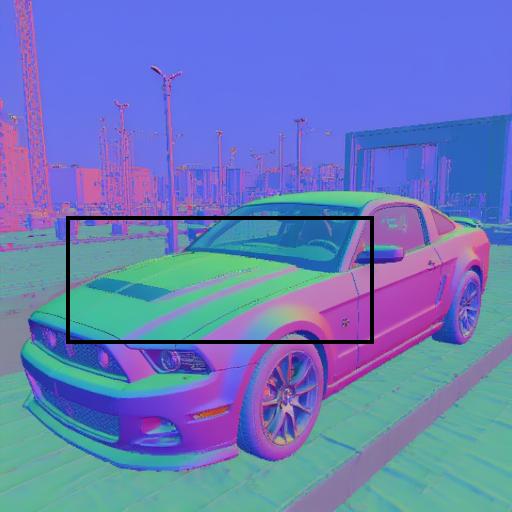}} 
            &
            \fbox{\includegraphics[width=0.10\textwidth]{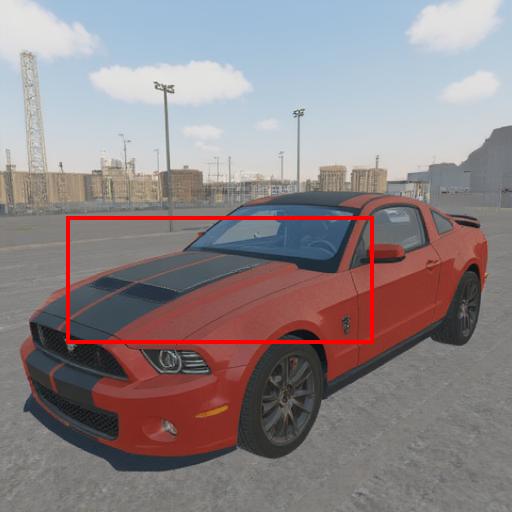}} 
            &
            \begin{tabular}[b]{@{}c@{}}
                \fbox{\includegraphics[width=0.046\textwidth]{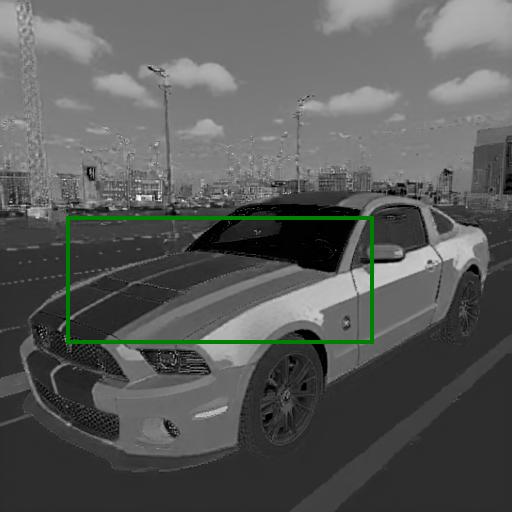}}
                \\
                \fbox{\includegraphics[width=0.046\textwidth]{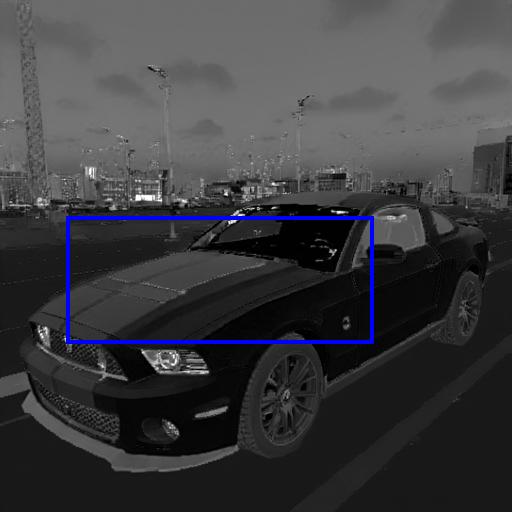}}
            \end{tabular}
            &
            \fbox{\includegraphics[width=0.10\textwidth]{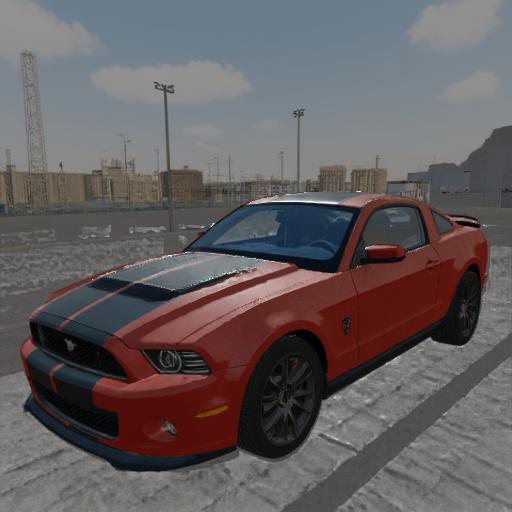}} 
            &
            \fbox{\includegraphics[width=0.10\textwidth]{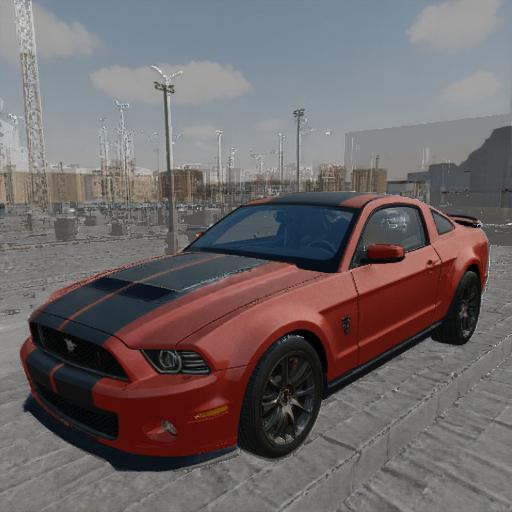}} 
        \end{tabular}}
            \\
            \resizebox{0.7\textwidth}{!}{
            \begin{tabular}[b]{cccc}
                \fbox{\includegraphics[width=0.10\textwidth]{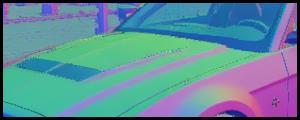}} 
                &
                \fbox{\includegraphics[width=0.10\textwidth]{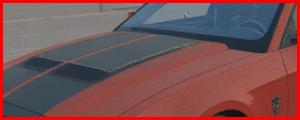}} 
                &
                \fbox{\includegraphics[width=0.10\textwidth]{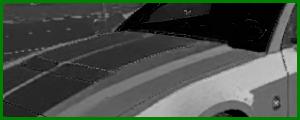}} 
                &
                \fbox{\includegraphics[width=0.10\textwidth]{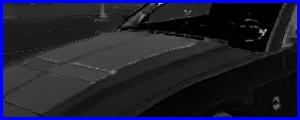}} 
            \end{tabular}}
        \end{tabular}
        \\
                
        \rotatebox{90}{w/o CIA-Dropout}
        &
        \begin{tabular}[b]{@{}c@{}}
            \resizebox{0.7\textwidth}{!}{
            \begin{tabular}[b]{ccccc}
            \fbox{\includegraphics[width=0.10\textwidth]{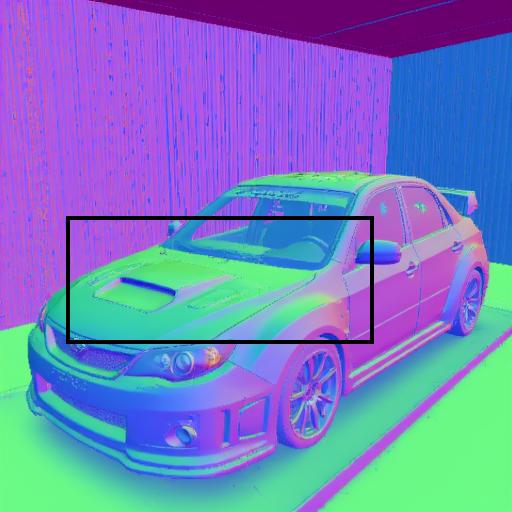}} 
            &
            \fbox{\includegraphics[width=0.10\textwidth]{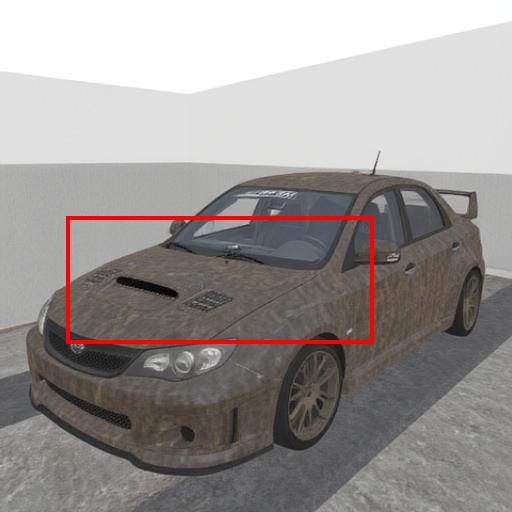}} 
            &
            \begin{tabular}[b]{@{}c@{}}
                \fbox{\includegraphics[width=0.046\textwidth]{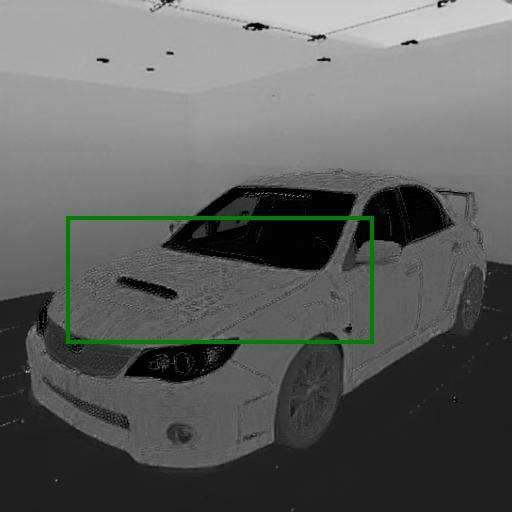}}
                \\
                \fbox{\includegraphics[width=0.046\textwidth]{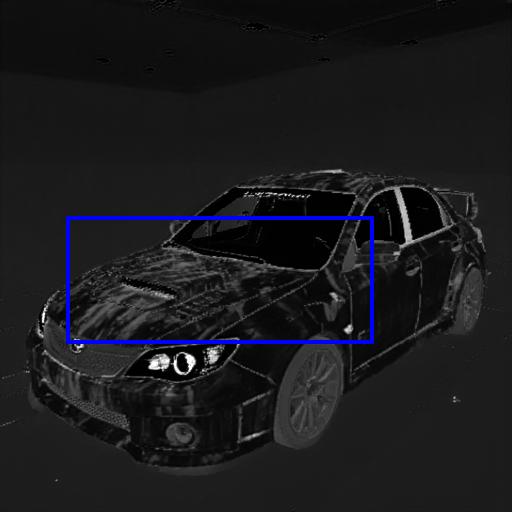}}
            \end{tabular}
            &
            \fbox{\includegraphics[width=0.10\textwidth]{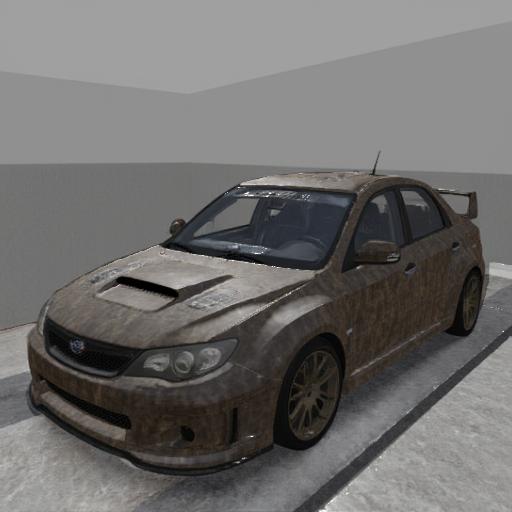}} 
            &
            \fbox{\includegraphics[width=0.10\textwidth]{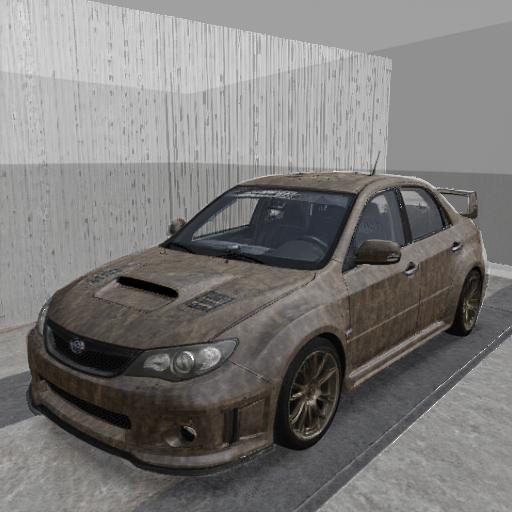}} 
        \end{tabular}}
            \\
            \resizebox{0.7\textwidth}{!}{
            \begin{tabular}[b]{cccc}
                \fbox{\includegraphics[width=0.10\textwidth]{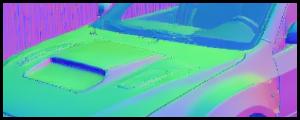}} 
                &
                \fbox{\includegraphics[width=0.10\textwidth]{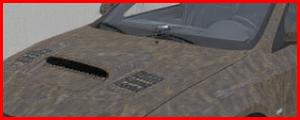}} 
                &
                \fbox{\includegraphics[width=0.10\textwidth]{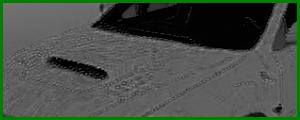}} 
                &
                \fbox{\includegraphics[width=0.10\textwidth]{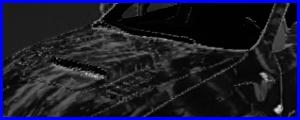}} 
            \end{tabular}}
        \end{tabular}
        \\
                
        \rotatebox{90}{Ours}
        &
        \begin{tabular}[b]{@{}c@{}}
            \resizebox{0.7\textwidth}{!}{
            \begin{tabular}[b]{ccccc}
            \fbox{\includegraphics[width=0.10\textwidth]{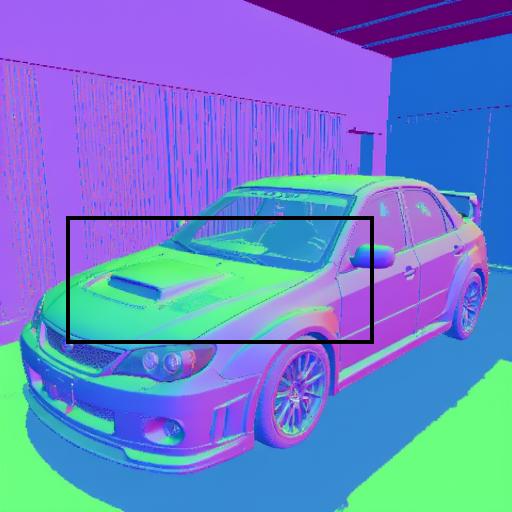}} 
            &
            \fbox{\includegraphics[width=0.10\textwidth]{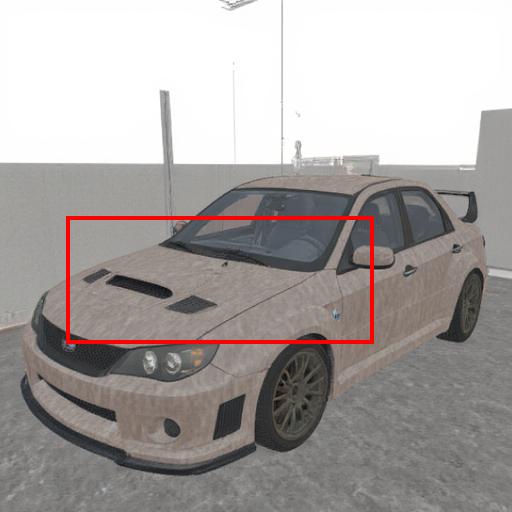}} 
            &
            \begin{tabular}[b]{@{}c@{}}
                \fbox{\includegraphics[width=0.046\textwidth]{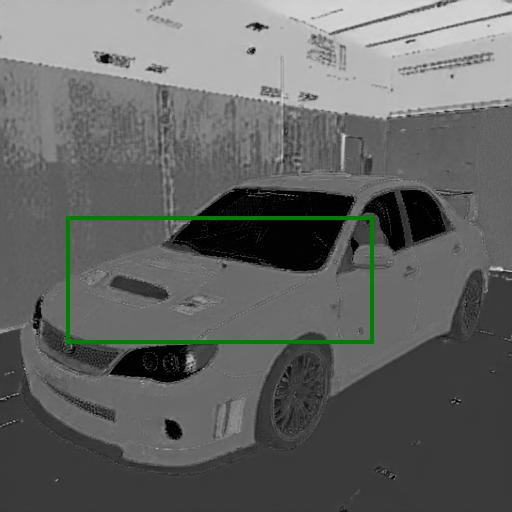}}
                \\
                \fbox{\includegraphics[width=0.046\textwidth]{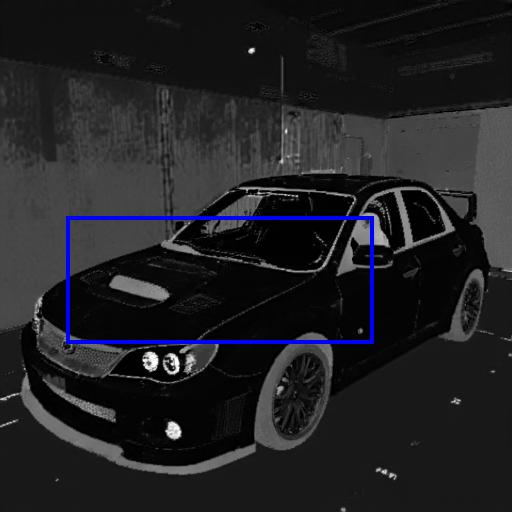}}
            \end{tabular}
            &
            \fbox{\includegraphics[width=0.10\textwidth]{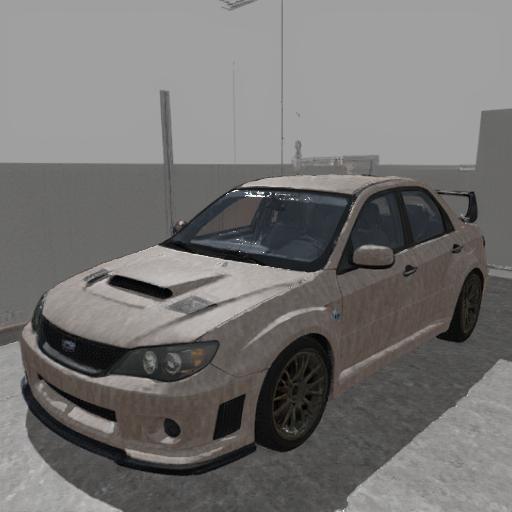}} 
            &
            \fbox{\includegraphics[width=0.10\textwidth]{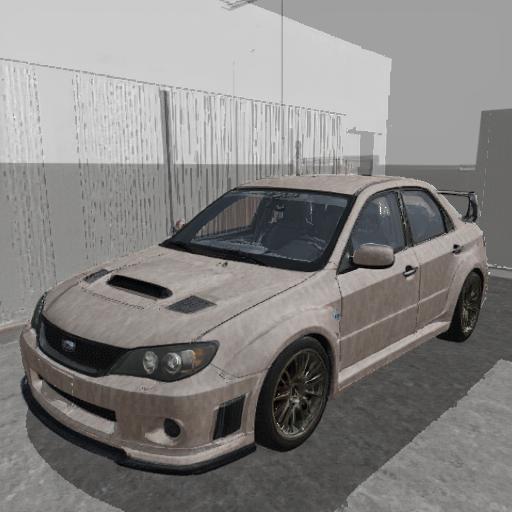}} 
        \end{tabular}}
            \\
            \resizebox{0.7\textwidth}{!}{
            \begin{tabular}[b]{cccc}
                \fbox{\includegraphics[width=0.10\textwidth]{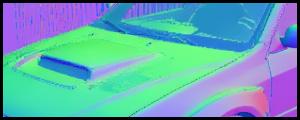}} 
                &
                \fbox{\includegraphics[width=0.10\textwidth]{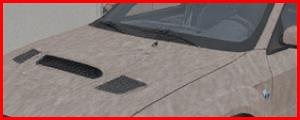}} 
                &
                \fbox{\includegraphics[width=0.10\textwidth]{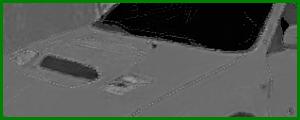}} 
                &
                \fbox{\includegraphics[width=0.10\textwidth]{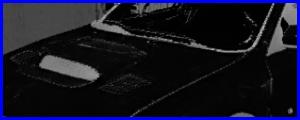}} 
            \end{tabular}}
        \end{tabular}
        \\

    \end{tabular}}
    \caption{\textbf{Additonal ablations}. 
    We compare our full method against ablations that do not use the rendering loss (w/o Rendering), use uniform light sampling instead of importance-based light sampling (w/o Light Sampling), and do not use dropout in the cross-intrinsic attention (w/o CIA-Dropout).
    Without the rendering loss (\Cref{subsubsec:rend-loss}), the PBR maps lose their semantic meaning, e.g., there are baked-in shadows in the albedo and the generated images appear ``averaged out''.
    Importance-based light sampling (\Cref{subsubsec:rend-loss}) and CIA dropout (\Cref{subsubsec:cia}) both increase the sharpness of individual PBR maps, e.g., the roughness/metallic images have realistic details \textit{without} baked-in textures.
    Overall, all components improve the quality of rendered images under varied lighting conditions.
    }
    \label{fig:supp:ablations}
\end{figure}

%% file: figures/stage1/samples.tex
\begin{figure*}[t]
    \centering
    \setlength\tabcolsep{1.25pt}
    \resizebox{\textwidth}{!}{
    \fboxsep=0pt
        \begin{tabular}{ccccccccccc}
        \rotatebox{90}{Normal}
        &
        \fbox{\includegraphics[width=0.11\textwidth]{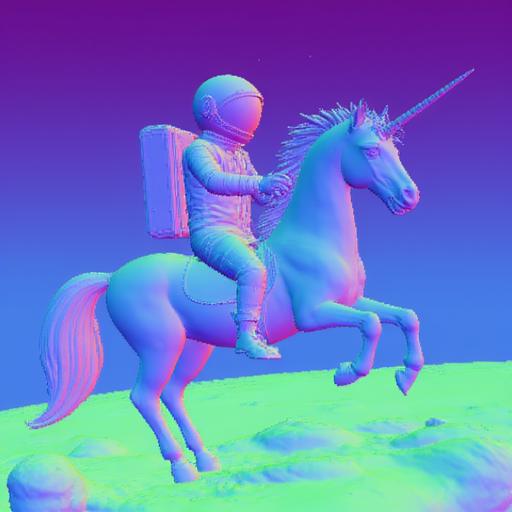}} 
        &
        \fbox{\includegraphics[width=0.11\textwidth]{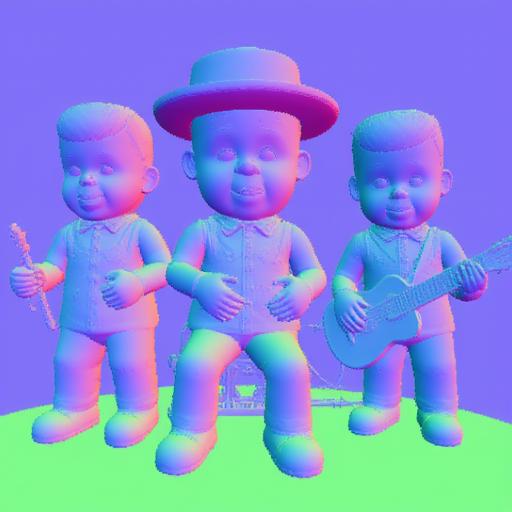}} 
        &
        \fbox{\includegraphics[width=0.11\textwidth]{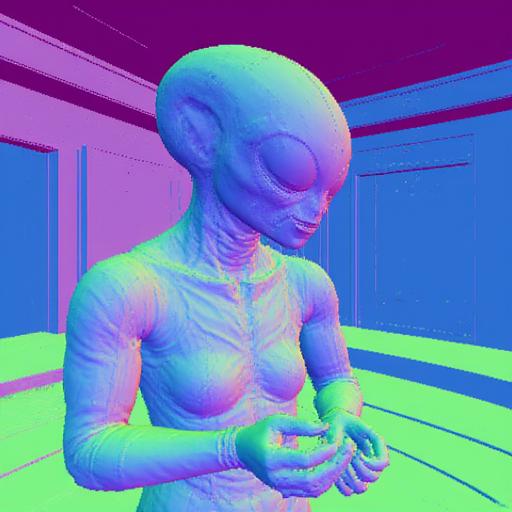}} 
        &
        \fbox{\includegraphics[width=0.11\textwidth]{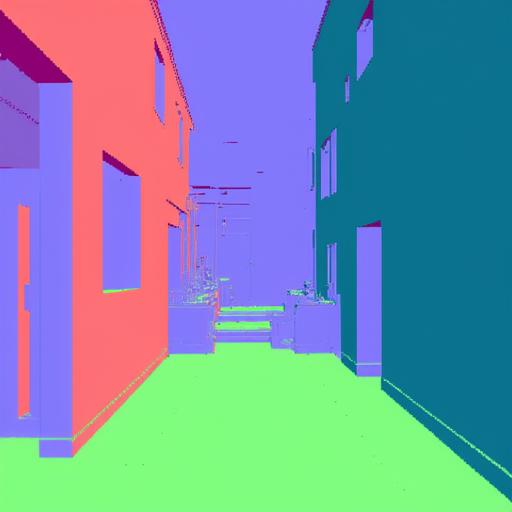}} 
        &
        \fbox{\includegraphics[width=0.11\textwidth]{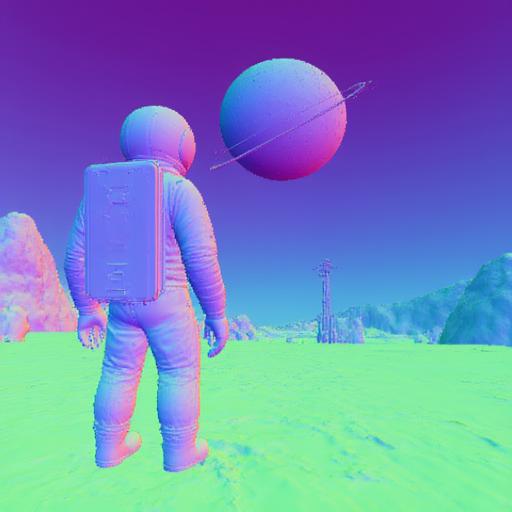}} 
        &
        \fbox{\includegraphics[width=0.11\textwidth]{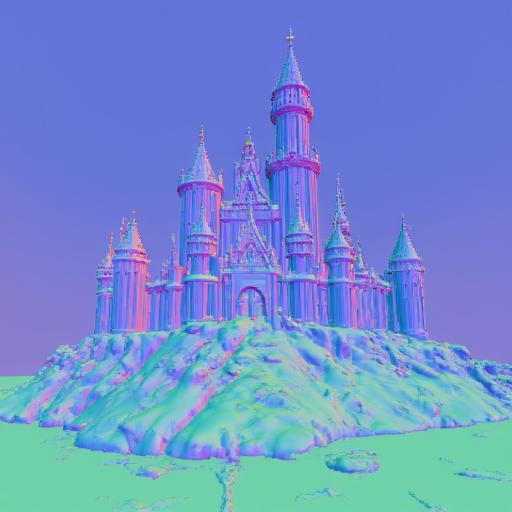}} 
        &
        \fbox{\includegraphics[width=0.11\textwidth]{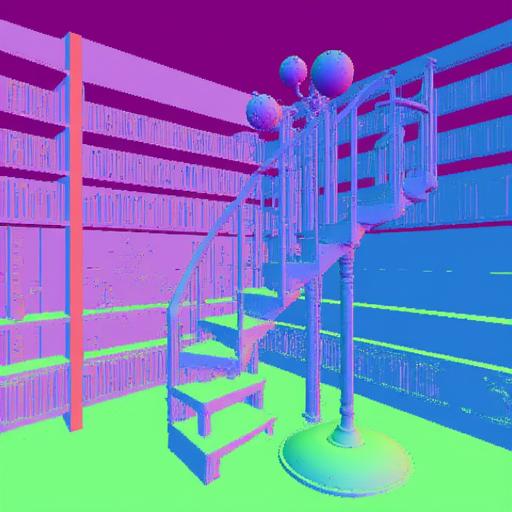}} 
        &
        \fbox{\includegraphics[width=0.11\textwidth]{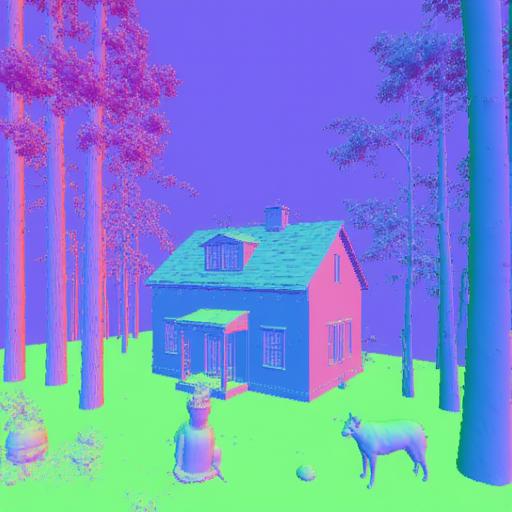}} 
        &
        \fbox{\includegraphics[width=0.11\textwidth]{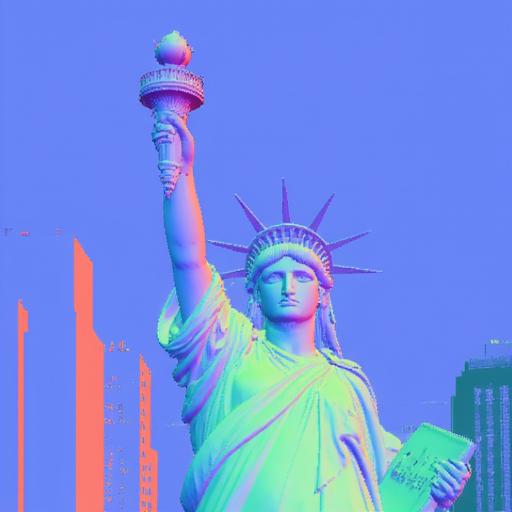}} 
        &
        \fbox{\includegraphics[width=0.11\textwidth]{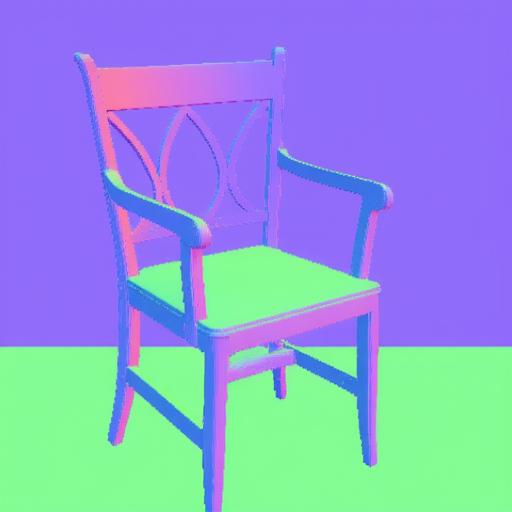}} 
        \\
        
        \rotatebox{90}{Albedo}
        &
        \fbox{\includegraphics[width=0.11\textwidth]{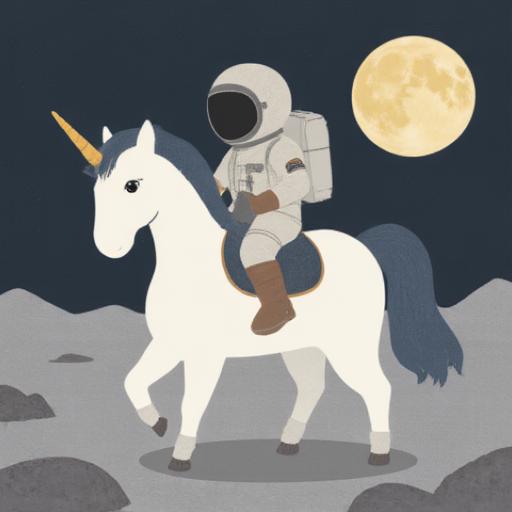}} 
        &
        \fbox{\includegraphics[width=0.11\textwidth]{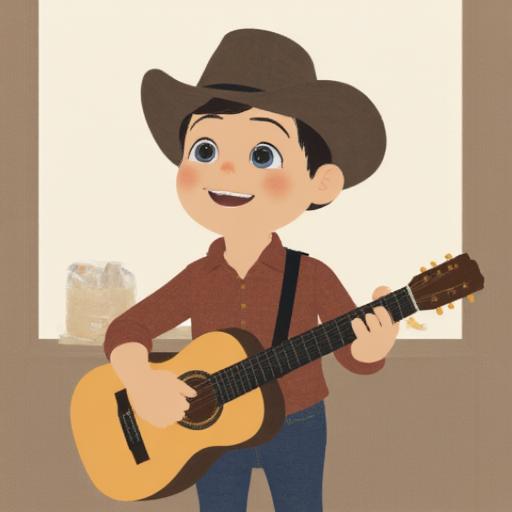}} 
        &
        \fbox{\includegraphics[width=0.11\textwidth]{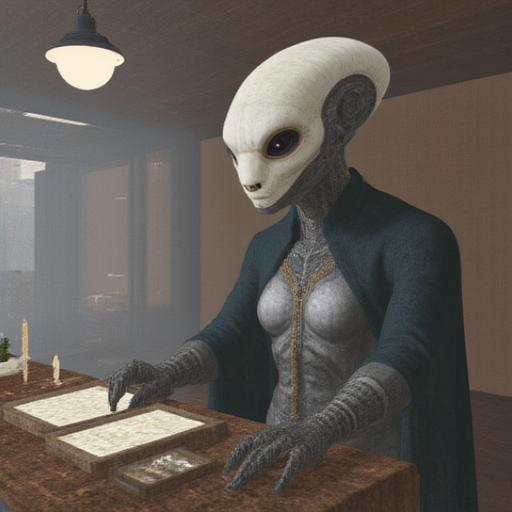}} 
        &
        \fbox{\includegraphics[width=0.11\textwidth]{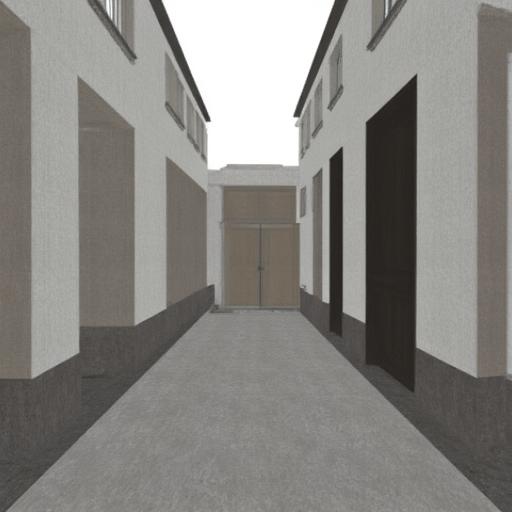}} 
        &
        \fbox{\includegraphics[width=0.11\textwidth]{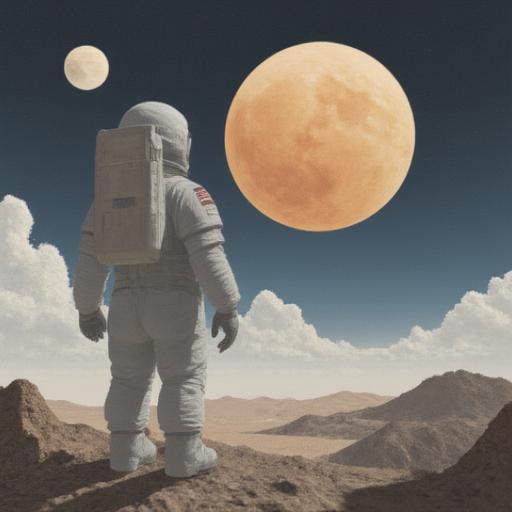}} 
        &
        \fbox{\includegraphics[width=0.11\textwidth]{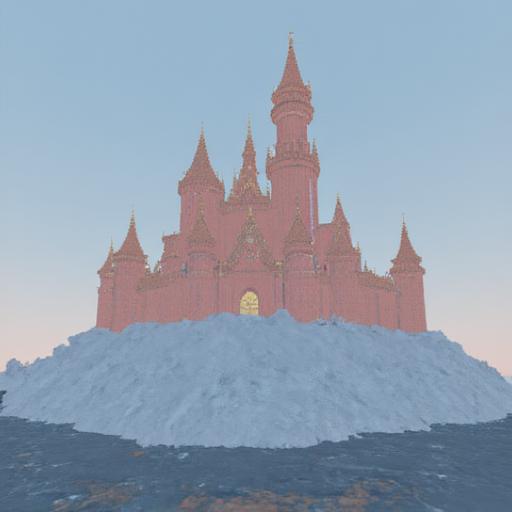}} 
        &
        \fbox{\includegraphics[width=0.11\textwidth]{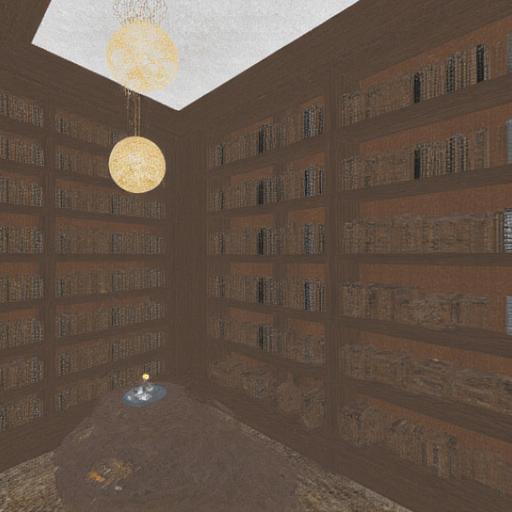}} 
        &
        \fbox{\includegraphics[width=0.11\textwidth]{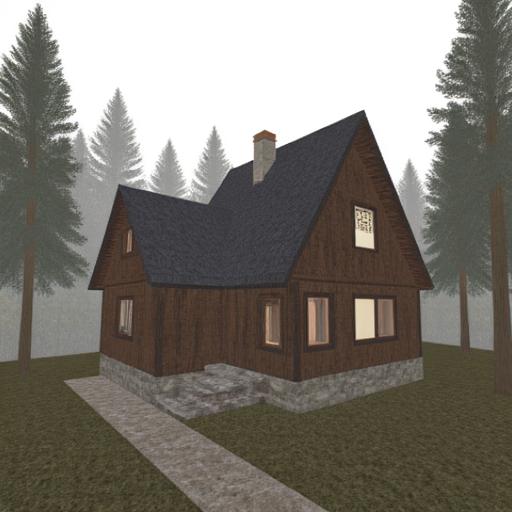}} 
        &
        \fbox{\includegraphics[width=0.11\textwidth]{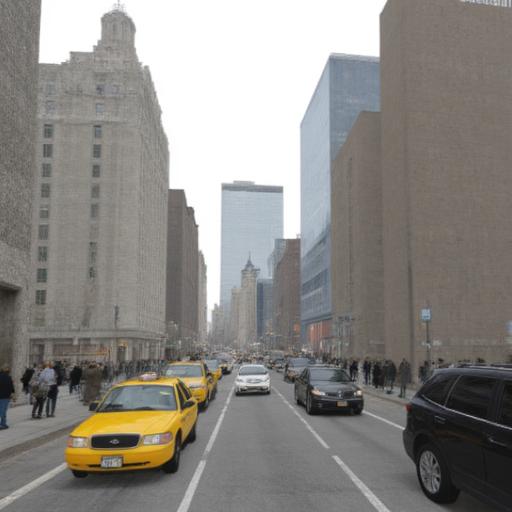}} 
        &
        \fbox{\includegraphics[width=0.11\textwidth]{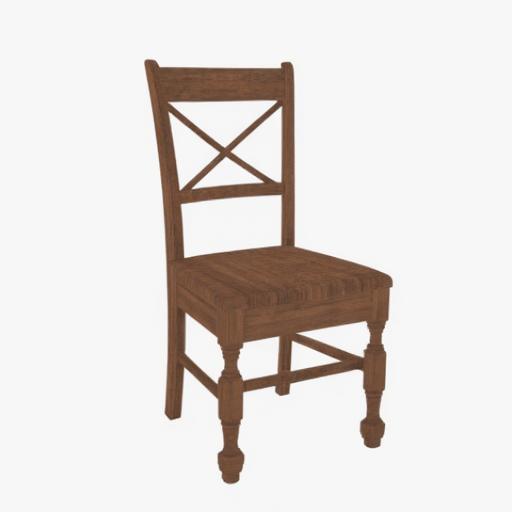}} 
        \\
        
        \rotatebox{90}{Roughness}
        &
        \fbox{\includegraphics[width=0.11\textwidth]{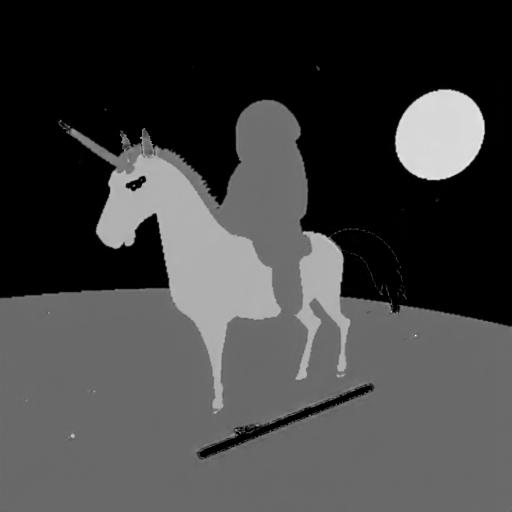}} 
        &
        \fbox{\includegraphics[width=0.11\textwidth]{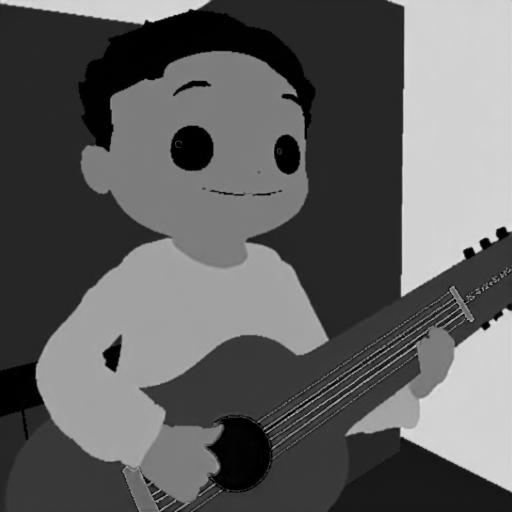}} 
        &
        \fbox{\includegraphics[width=0.11\textwidth]{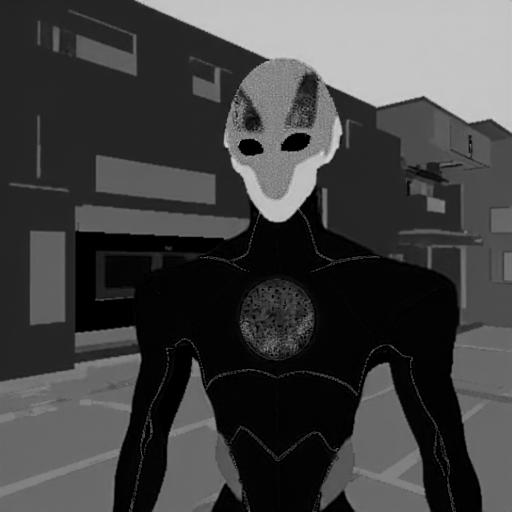}} 
        &
        \fbox{\includegraphics[width=0.11\textwidth]{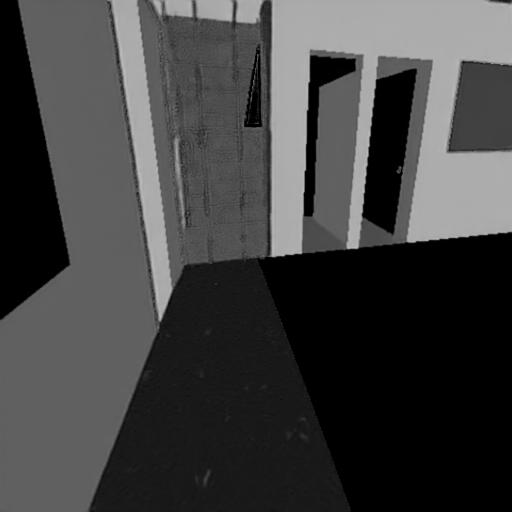}} 
        &
        \fbox{\includegraphics[width=0.11\textwidth]{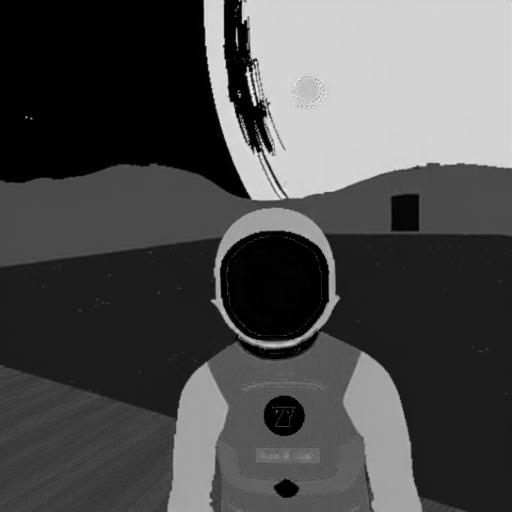}} 
        &
        \fbox{\includegraphics[width=0.11\textwidth]{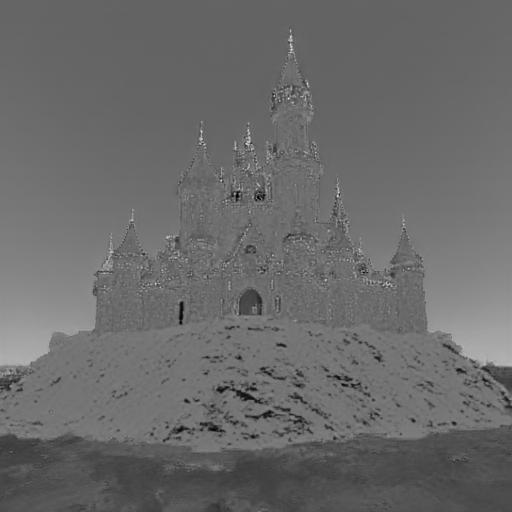}} 
        &
        \fbox{\includegraphics[width=0.11\textwidth]{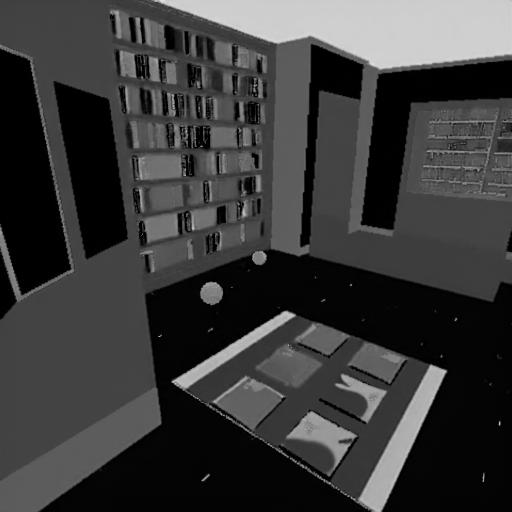}} 
        &
        \fbox{\includegraphics[width=0.11\textwidth]{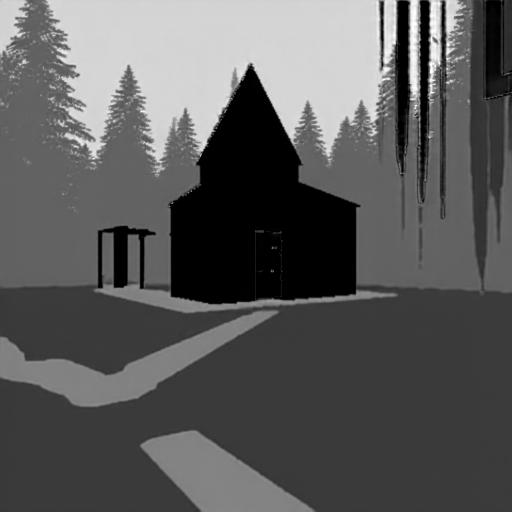}} 
        &
        \fbox{\includegraphics[width=0.11\textwidth]{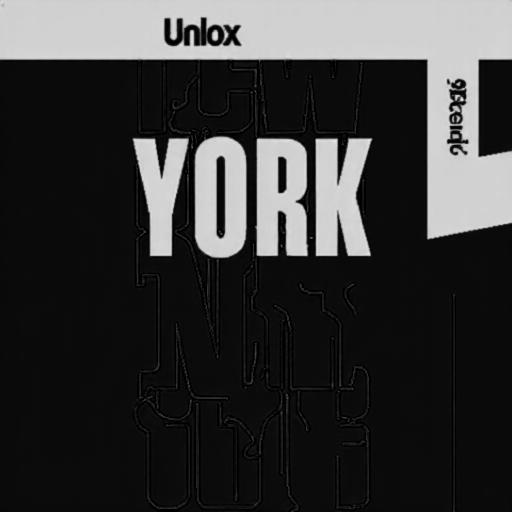}} 
        &
        \fbox{\includegraphics[width=0.11\textwidth]{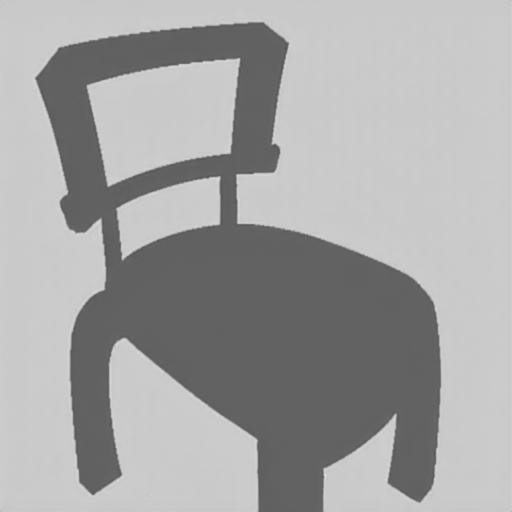}} 
        \\
        
        \rotatebox{90}{Metallic}
        &
        \fbox{\includegraphics[width=0.11\textwidth]{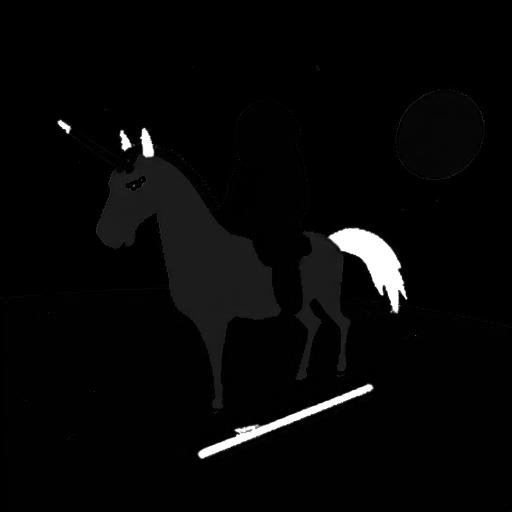}} 
        &
        \fbox{\includegraphics[width=0.11\textwidth]{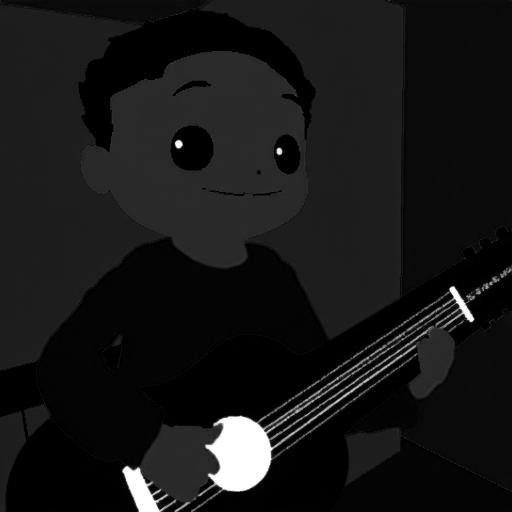}} 
        &
        \fbox{\includegraphics[width=0.11\textwidth]{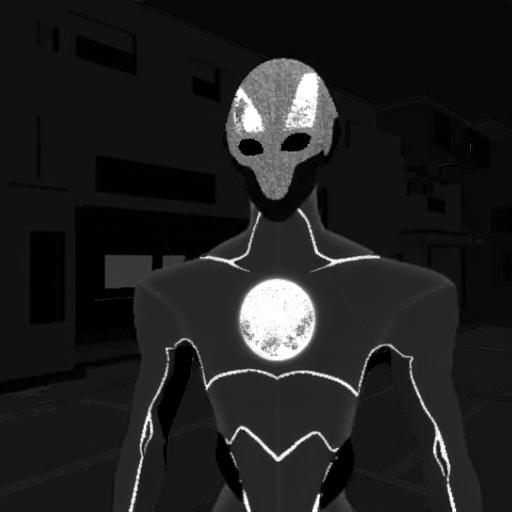}} 
        &
        \fbox{\includegraphics[width=0.11\textwidth]{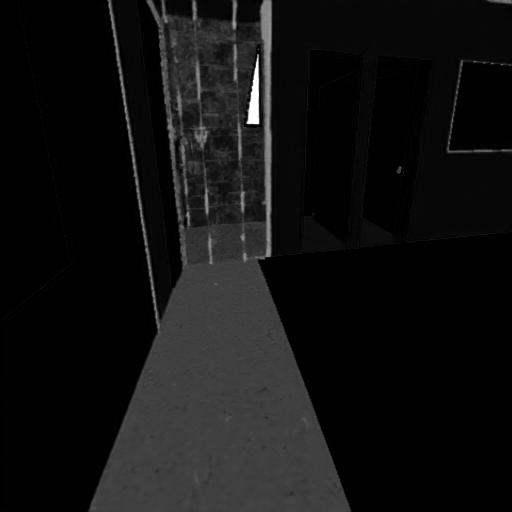}} 
        &
        \fbox{\includegraphics[width=0.11\textwidth]{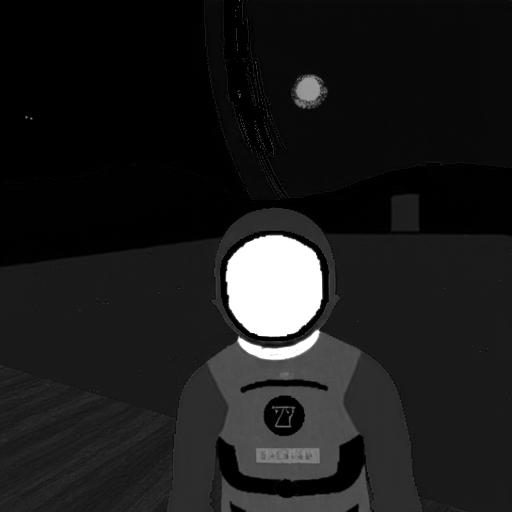}} 
        &
        \fbox{\includegraphics[width=0.11\textwidth]{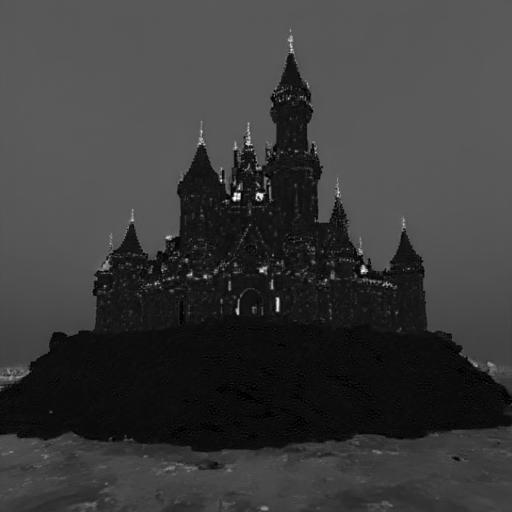}} 
        &
        \fbox{\includegraphics[width=0.11\textwidth]{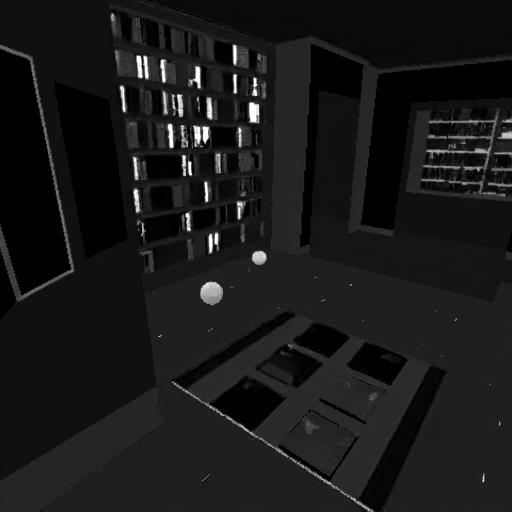}} 
        &
        \fbox{\includegraphics[width=0.11\textwidth]{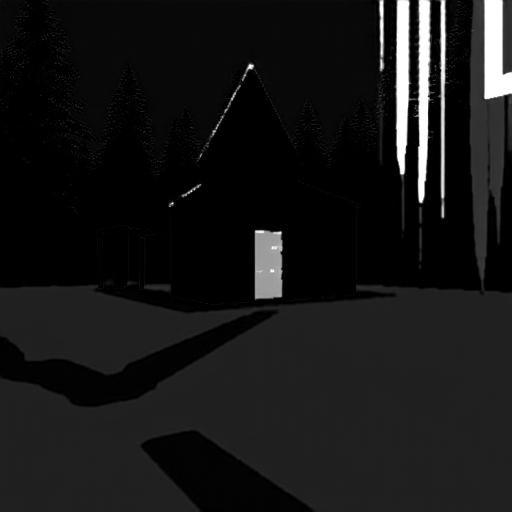}} 
        &
        \fbox{\includegraphics[width=0.11\textwidth]{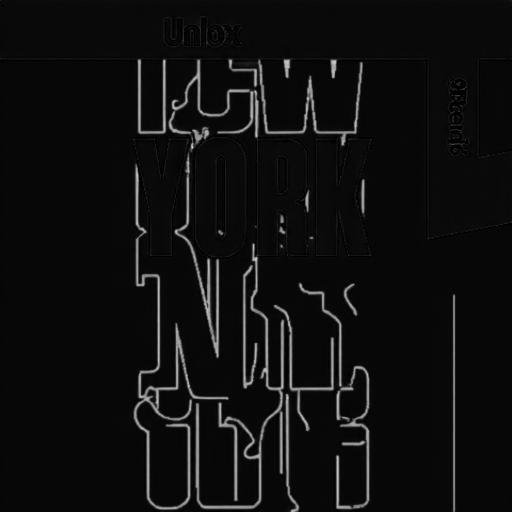}} 
        &
        \fbox{\includegraphics[width=0.11\textwidth]{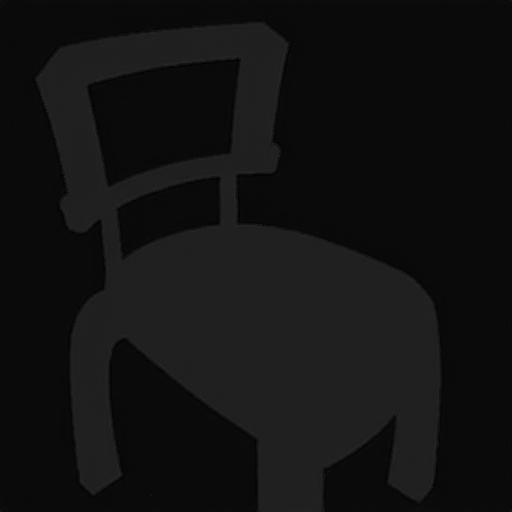}} 
        \\

        \midrule

        \rotatebox{90}{Normal}
        &
        \fbox{\includegraphics[width=0.11\textwidth]{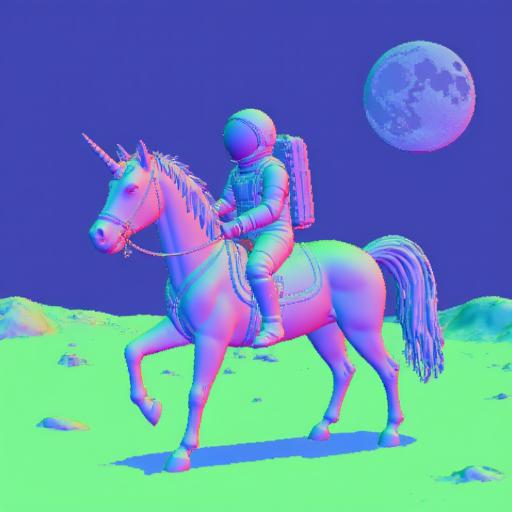}} 
        &
        \fbox{\includegraphics[width=0.11\textwidth]{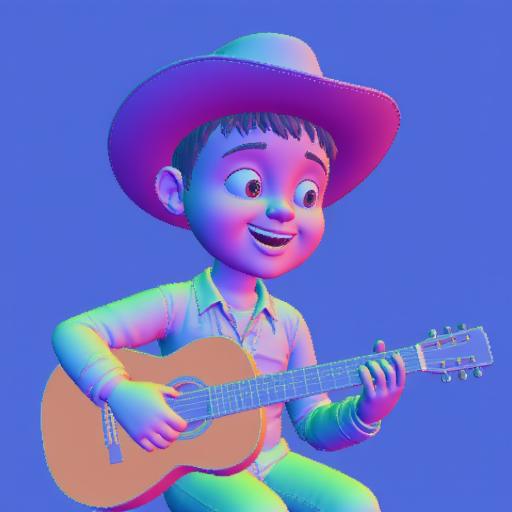}} 
        &
        \fbox{\includegraphics[width=0.11\textwidth]{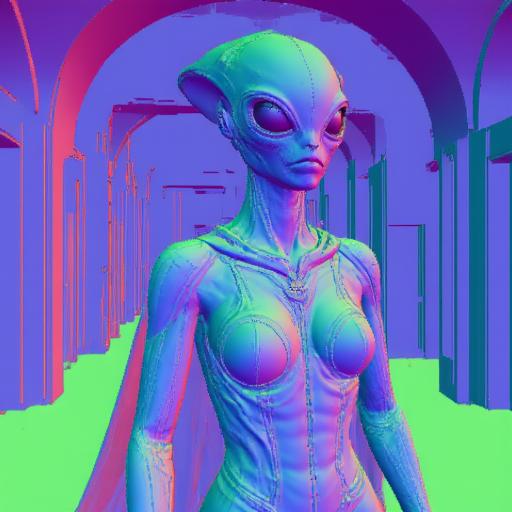}} 
        &
        \fbox{\includegraphics[width=0.11\textwidth]{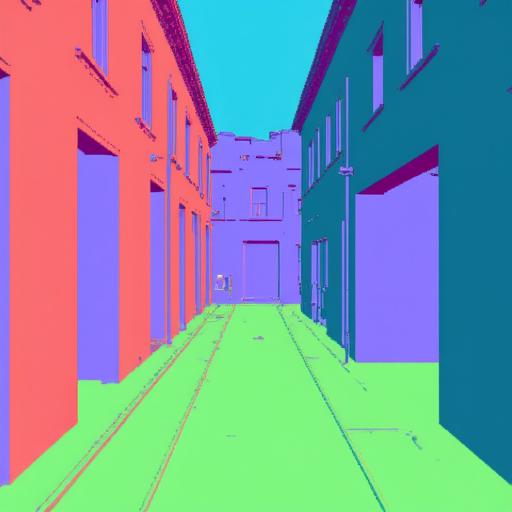}} 
        &
        \fbox{\includegraphics[width=0.11\textwidth]{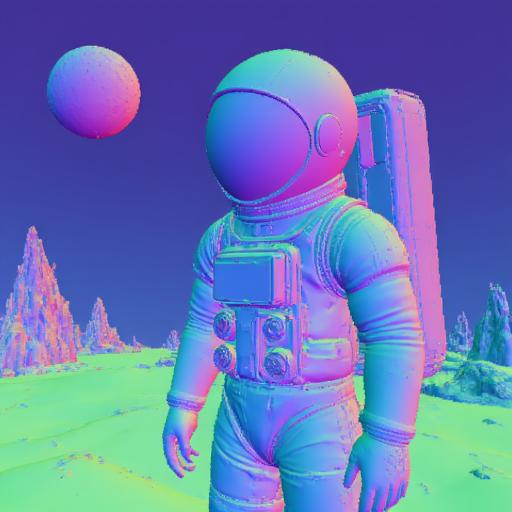}} 
        &
        \fbox{\includegraphics[width=0.11\textwidth]{res/stage1/samples/stage2/021/001_normal.jpg}} 
        &
        \fbox{\includegraphics[width=0.11\textwidth]{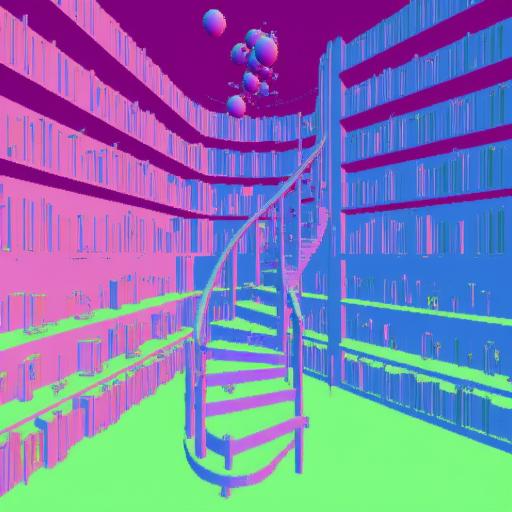}} 
        &
        \fbox{\includegraphics[width=0.11\textwidth]{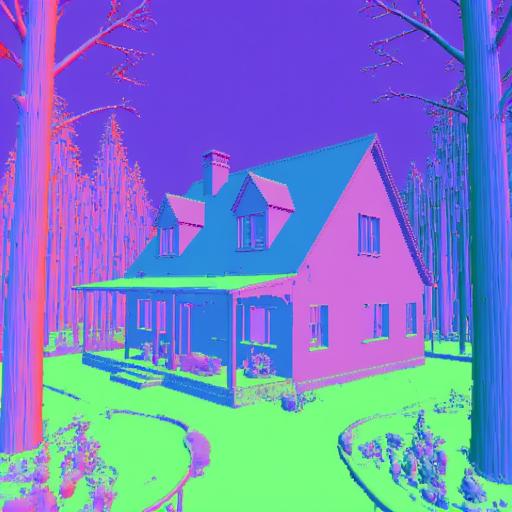}} 
        &
        \fbox{\includegraphics[width=0.11\textwidth]{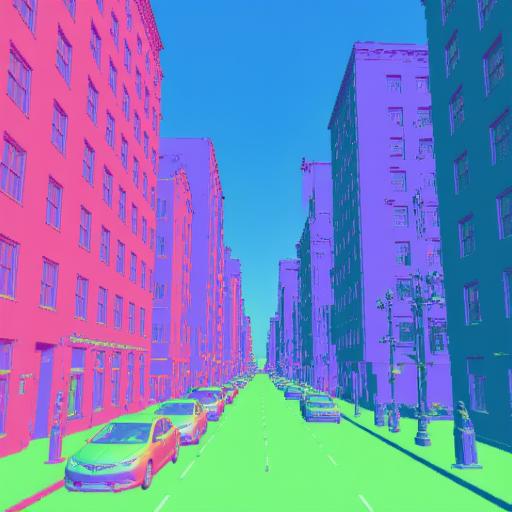}} 
        &
        \fbox{\includegraphics[width=0.11\textwidth]{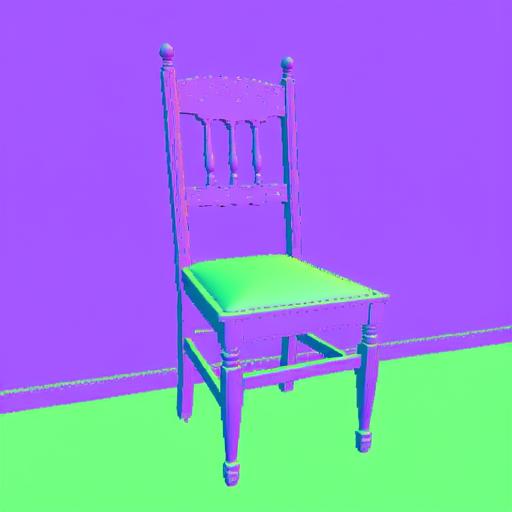}} 
        \\
        
        \rotatebox{90}{Albedo}
        &
        \fbox{\includegraphics[width=0.11\textwidth]{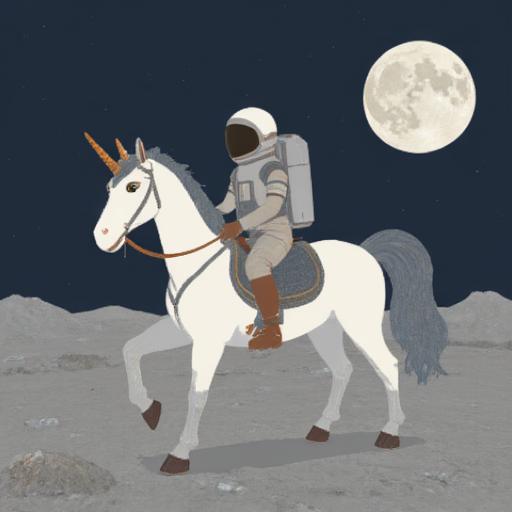}} 
        &
        \fbox{\includegraphics[width=0.11\textwidth]{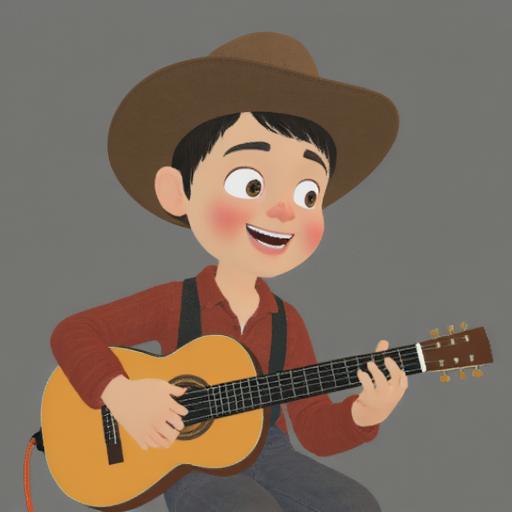}} 
        &
        \fbox{\includegraphics[width=0.11\textwidth]{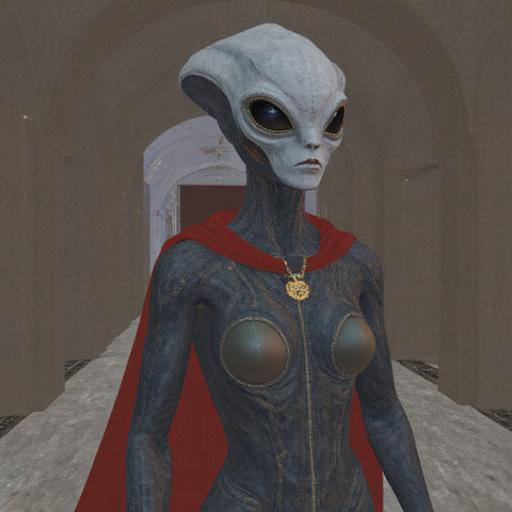}} 
        &
        \fbox{\includegraphics[width=0.11\textwidth]{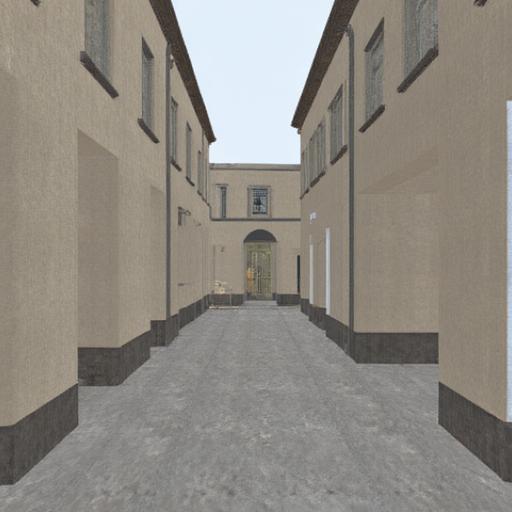}} 
        &
        \fbox{\includegraphics[width=0.11\textwidth]{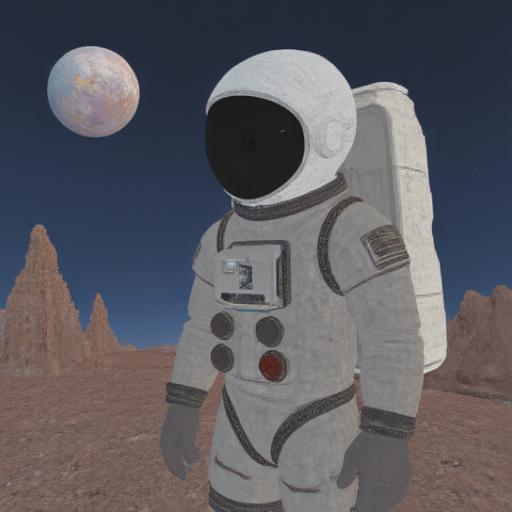}} 
        &
        \fbox{\includegraphics[width=0.11\textwidth]{res/stage1/samples/stage2/021/001_albedo.jpg}} 
        &
        \fbox{\includegraphics[width=0.11\textwidth]{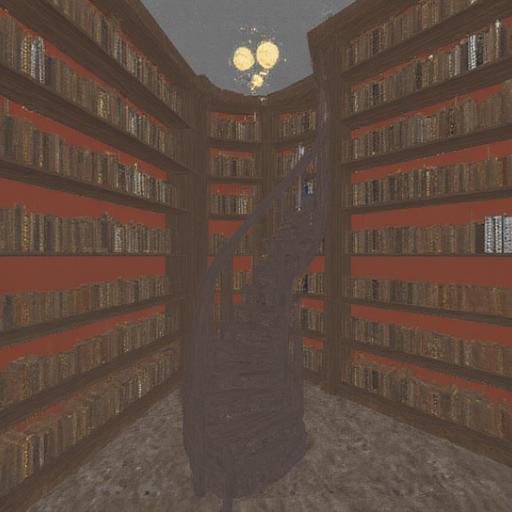}} 
        &
        \fbox{\includegraphics[width=0.11\textwidth]{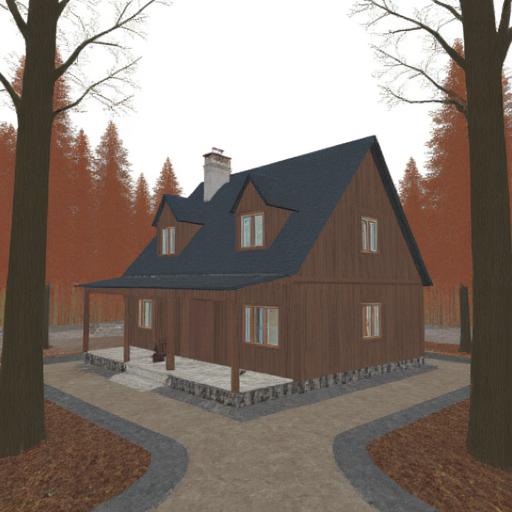}} 
        &
        \fbox{\includegraphics[width=0.11\textwidth]{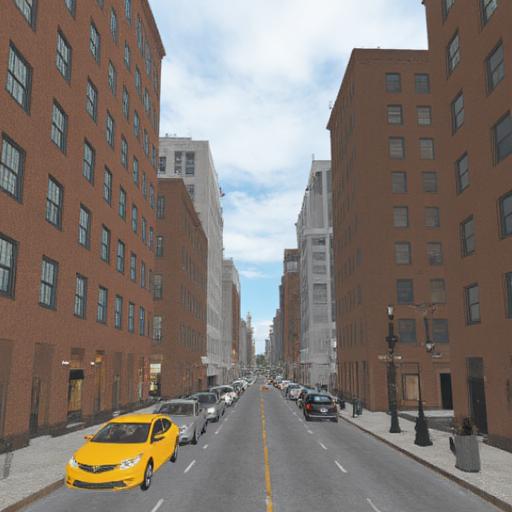}} 
        &
        \fbox{\includegraphics[width=0.11\textwidth]{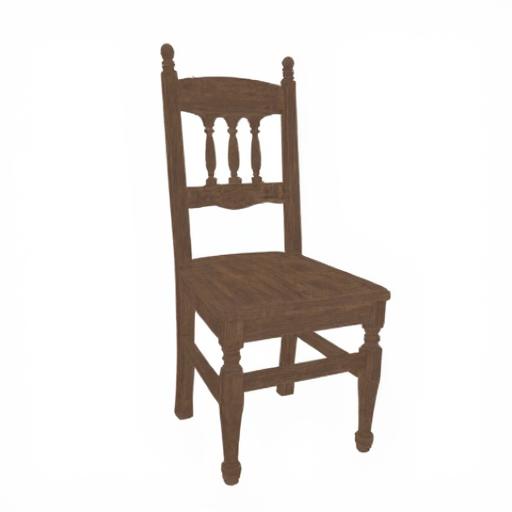}} 
        \\
        
        \rotatebox{90}{Roughness}
        &
        \fbox{\includegraphics[width=0.11\textwidth]{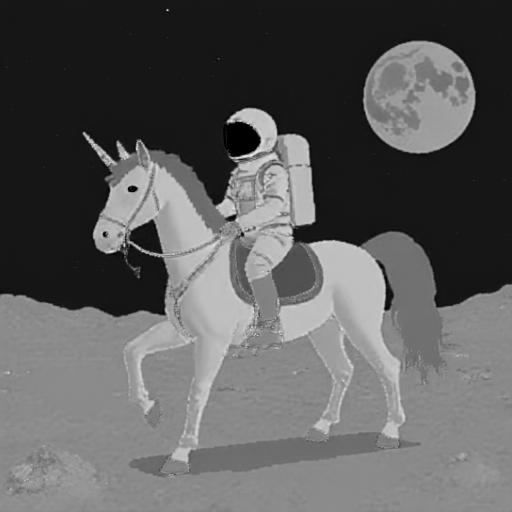}} 
        &
        \fbox{\includegraphics[width=0.11\textwidth]{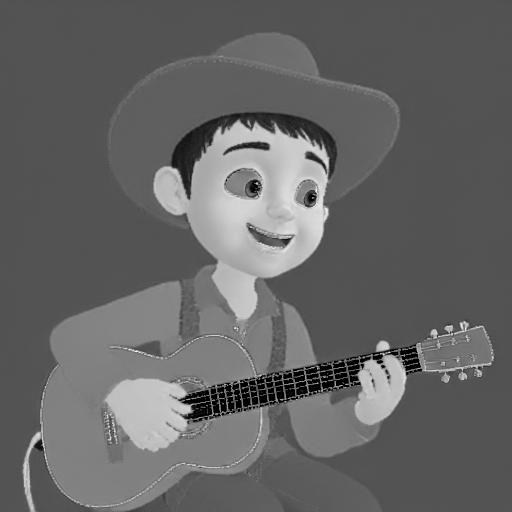}} 
        &
        \fbox{\includegraphics[width=0.11\textwidth]{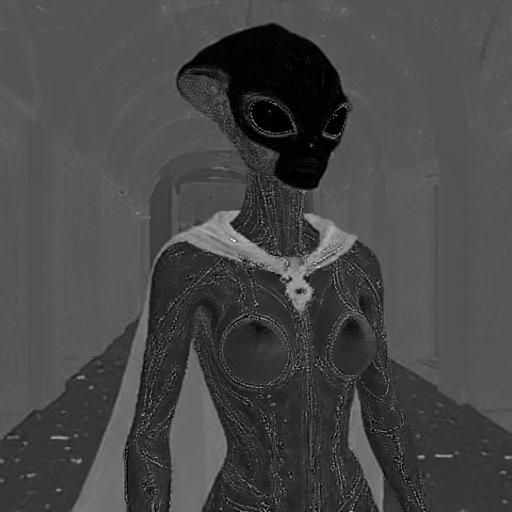}} 
        &
        \fbox{\includegraphics[width=0.11\textwidth]{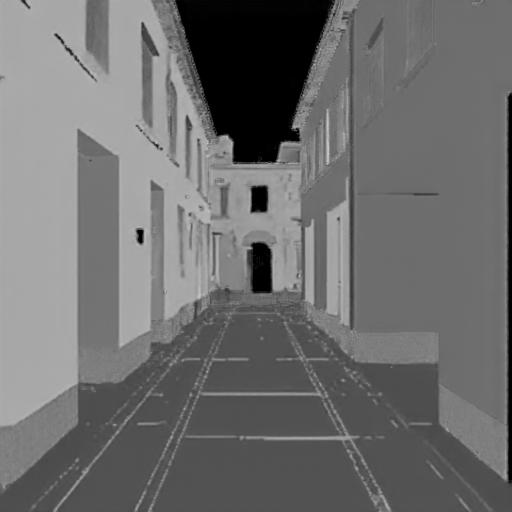}} 
        &
        \fbox{\includegraphics[width=0.11\textwidth]{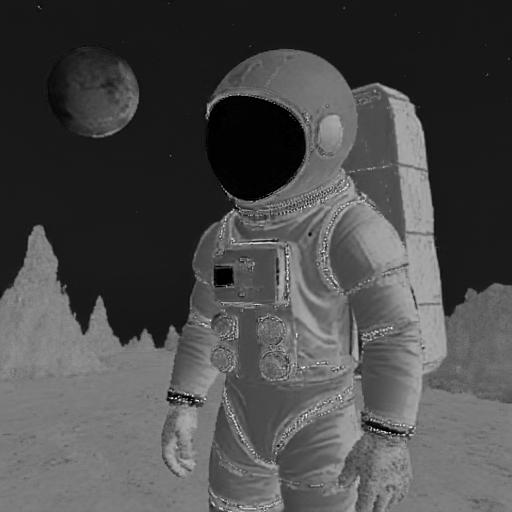}} 
        &
        \fbox{\includegraphics[width=0.11\textwidth]{res/stage1/samples/stage2/021/001_roughness.jpg}} 
        &
        \fbox{\includegraphics[width=0.11\textwidth]{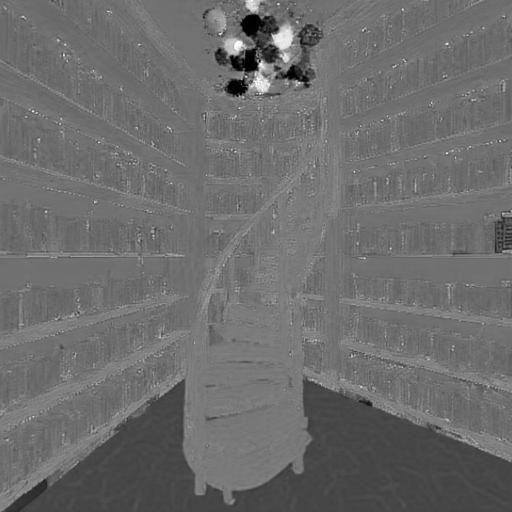}} 
        &
        \fbox{\includegraphics[width=0.11\textwidth]{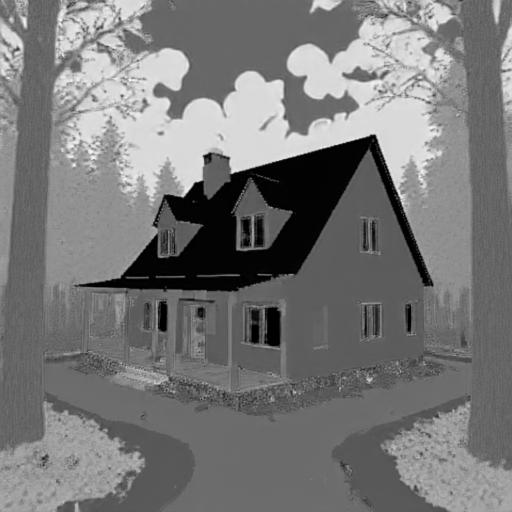}} 
        &
        \fbox{\includegraphics[width=0.11\textwidth]{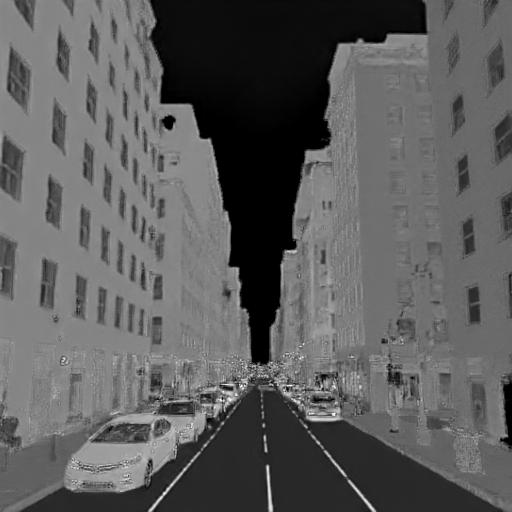}} 
        &
        \fbox{\includegraphics[width=0.11\textwidth]{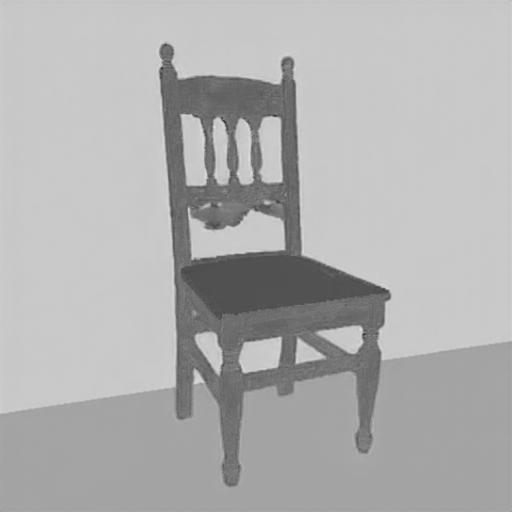}} 
        \\
        
        \rotatebox{90}{Metallic}
        &
        \fbox{\includegraphics[width=0.11\textwidth]{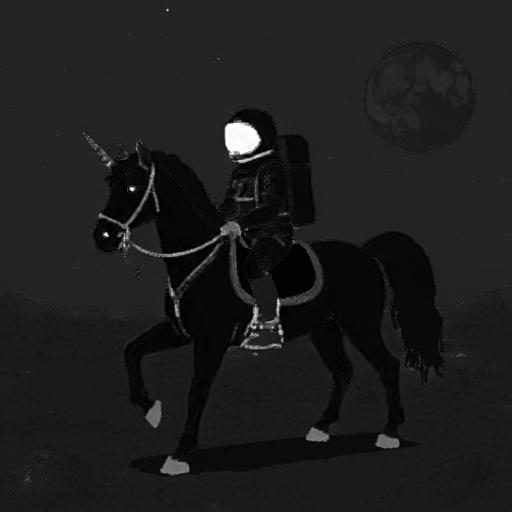}} 
        &
        \fbox{\includegraphics[width=0.11\textwidth]{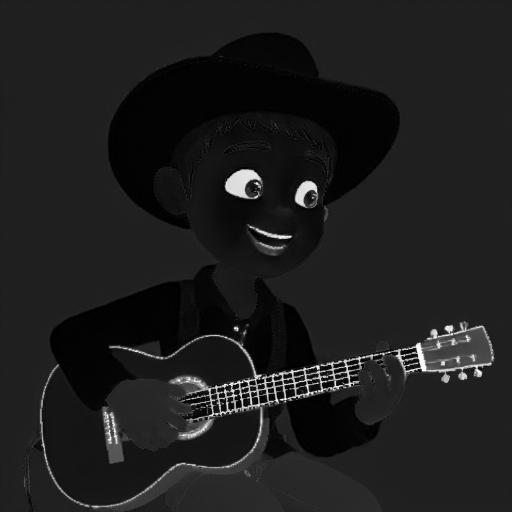}} 
        &
        \fbox{\includegraphics[width=0.11\textwidth]{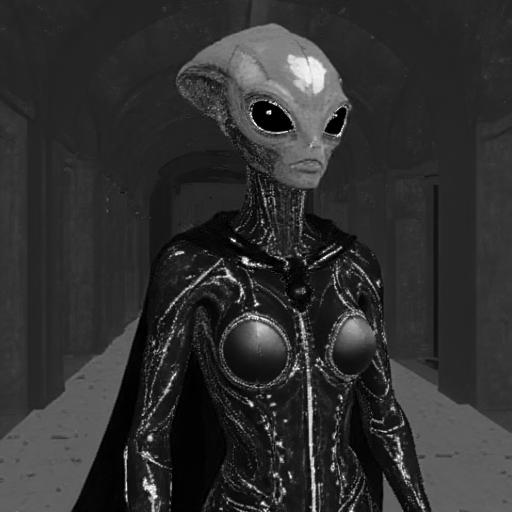}} 
        &
        \fbox{\includegraphics[width=0.11\textwidth]{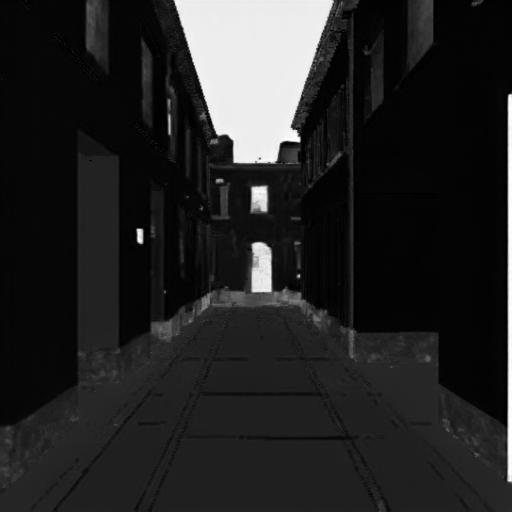}} 
        &
        \fbox{\includegraphics[width=0.11\textwidth]{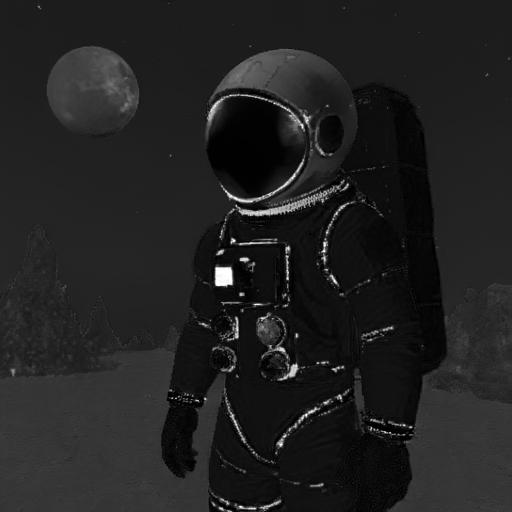}} 
        &
        \fbox{\includegraphics[width=0.11\textwidth]{res/stage1/samples/stage2/021/001_metallic.jpg}} 
        &
        \fbox{\includegraphics[width=0.11\textwidth]{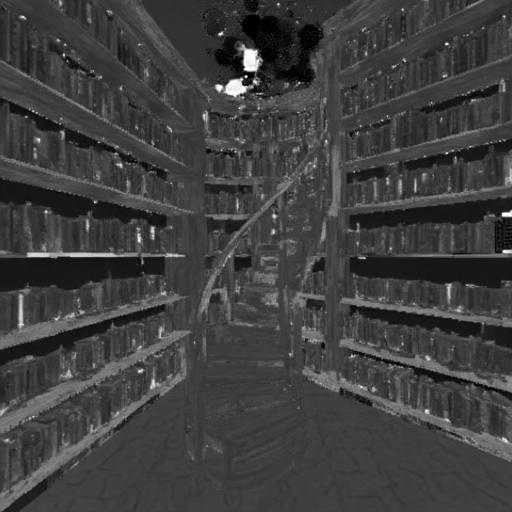}} 
        &
        \fbox{\includegraphics[width=0.11\textwidth]{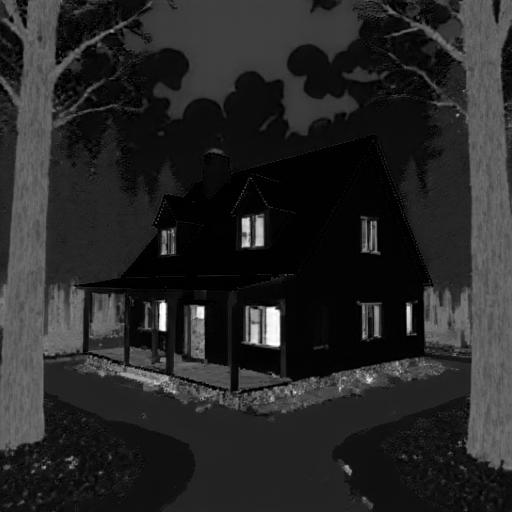}} 
        &
        \fbox{\includegraphics[width=0.11\textwidth]{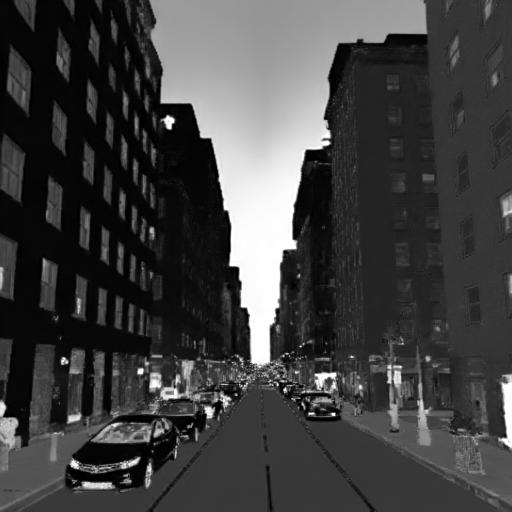}} 
        &
        \fbox{\includegraphics[width=0.11\textwidth]{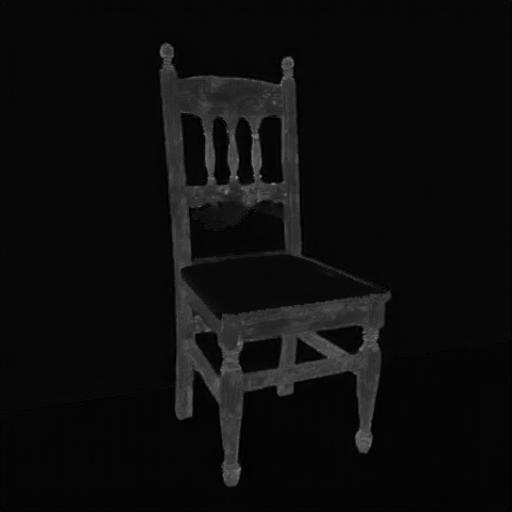}} 
        \\
    \end{tabular}}

    \vspace{-6pt}
    \caption{\textbf{Comparison between stage 1 and stage 2 samples}. 
    In the first stage, we train 3 LoRAs separately corresponding to the different PBR maps (albedo, normal, roughness+metallic) on synthetic indoor-scene examples.
    In the second stage, we align these PBR priors through cross-intrinsic attention and the rendering loss.
    \textbf{Top:} generated images in the first stage (independently for each modality) show good quality for the albedo and normal maps. 
    However, the roughness/metallic predictions are only reasonable for in-distribution scenarios (e.g. the 4th column) and become less detailed for out-of-distribution prompts.
    \textbf{Bottom:} after alignment training, all PBR maps have meaningful structure and exhibit sharp, high-quality content.
    }
    \label{fig:stage1:samples}
    \vspace{-6pt}
\end{figure*}

%% file: figures/experiments/comparisons_more.tex
\begin{figure*}[t]
    \centering
    \setlength\tabcolsep{1.25pt}
    \resizebox{\textwidth}{!}{
    \fboxsep=0pt
        \begin{tabular}{ccc|cc||cc|cc}
        \rotatebox{90}{IID}
        &
        \fbox{\includegraphics[width=0.14\textwidth]{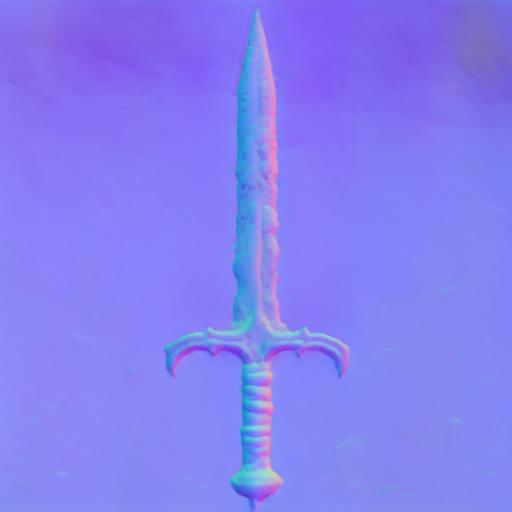}} 
        &
        \begin{tikzpicture}[every node/.style={anchor=north west,inner sep=0pt},x=1pt, y=-1pt,]  
             \node (fig1) at (0,0)
               {\fbox{\includegraphics[width=0.14\textwidth]{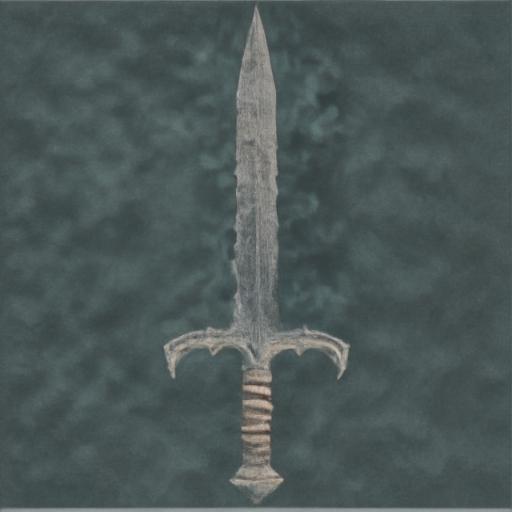}}};   
             \node (fig2) at (59,0)
               {\fbox{\includegraphics[width=0.065\textwidth]{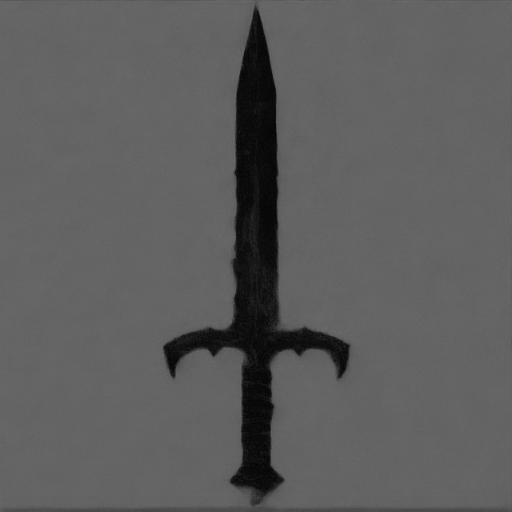}}};
             \node (fig3) at (59,30)
               {\fbox{\includegraphics[width=0.065\textwidth]{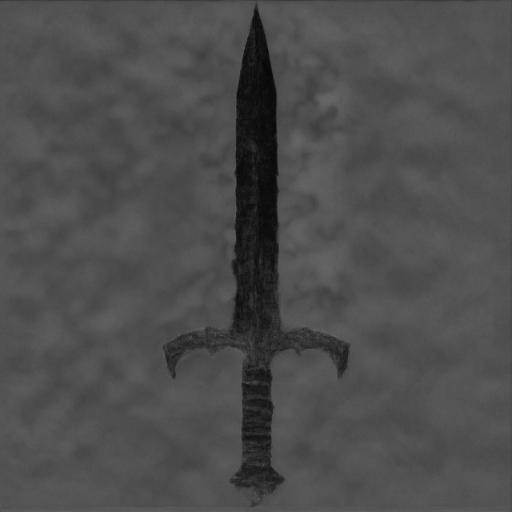}}};
        \end{tikzpicture}
        &
        \fbox{\includegraphics[width=0.14\textwidth]{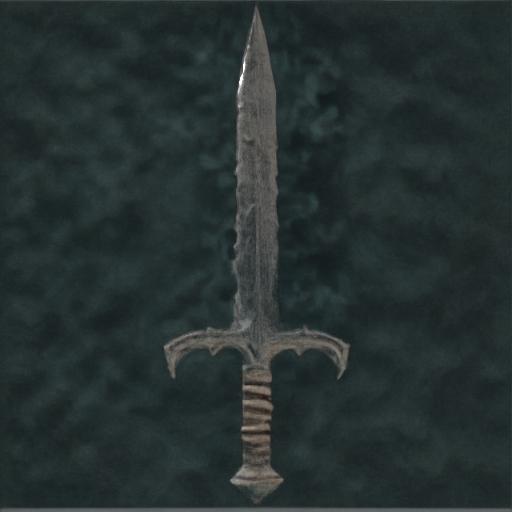}} 
        &
        \fbox{\includegraphics[width=0.14\textwidth]{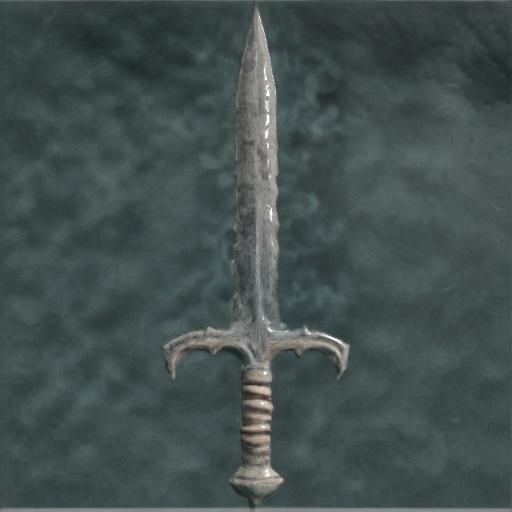}} 
        
        \hspace{6pt}
        &
        \hspace{6pt}
        
        \fbox{\includegraphics[width=0.14\textwidth]{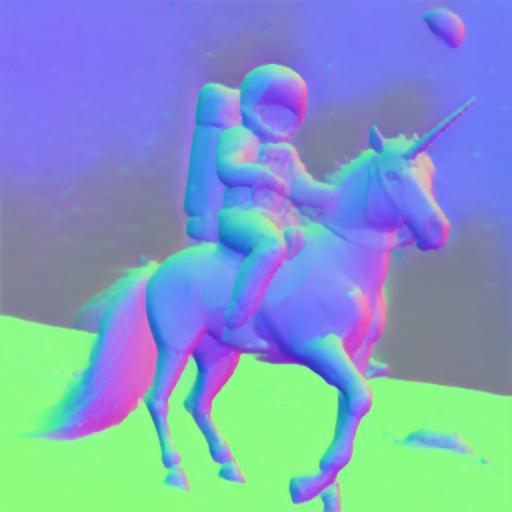}} 
        &
        \begin{tikzpicture}[every node/.style={anchor=north west,inner sep=0pt},x=1pt, y=-1pt,]  
             \node (fig1) at (0,0)
               {\fbox{\includegraphics[width=0.14\textwidth]{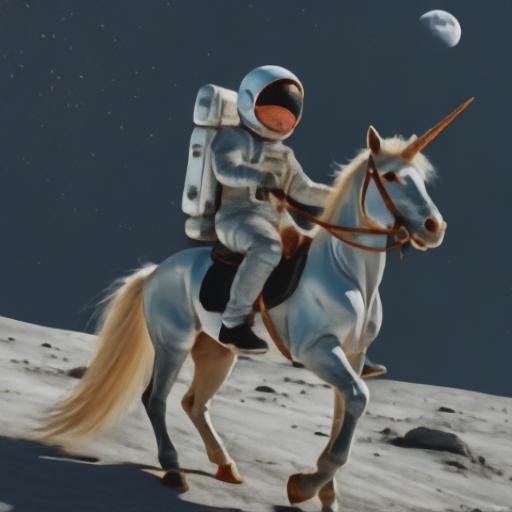}}};   
             \node (fig2) at (59,0)
               {\fbox{\includegraphics[width=0.065\textwidth]{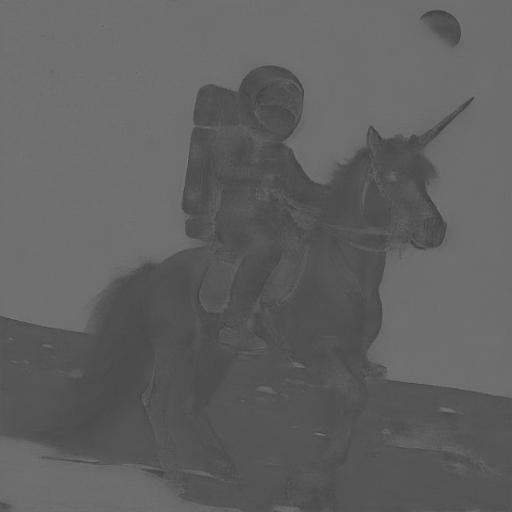}}};
             \node (fig3) at (59,30)
               {\fbox{\includegraphics[width=0.065\textwidth]{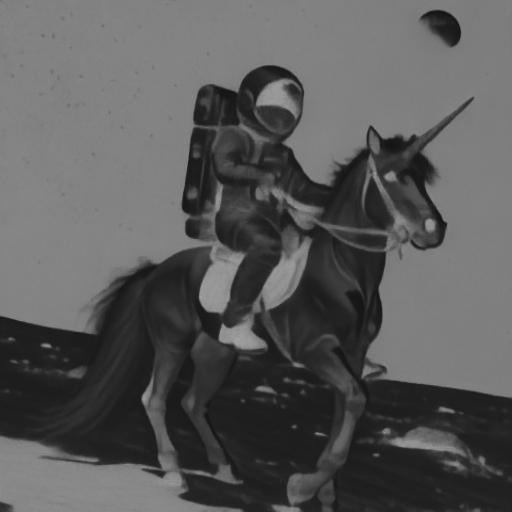}}};
        \end{tikzpicture}
        &
        \fbox{\includegraphics[width=0.14\textwidth]{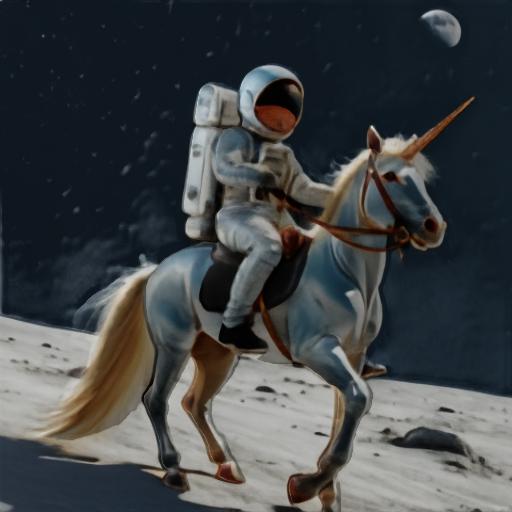}} 
        &
        \fbox{\includegraphics[width=0.14\textwidth]{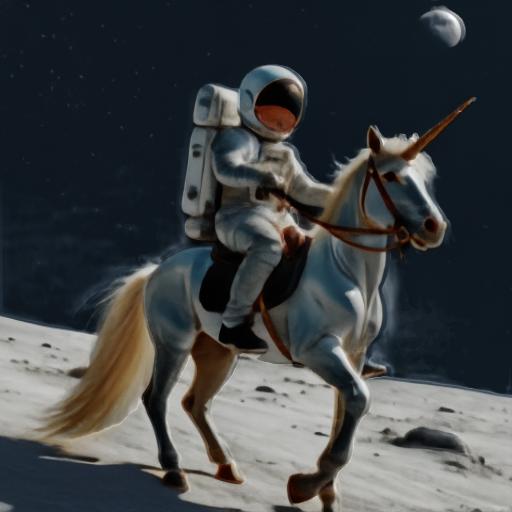}} 
        \\
        
        \rotatebox{90}{RGBX}
        &
        \fbox{\includegraphics[width=0.14\textwidth]{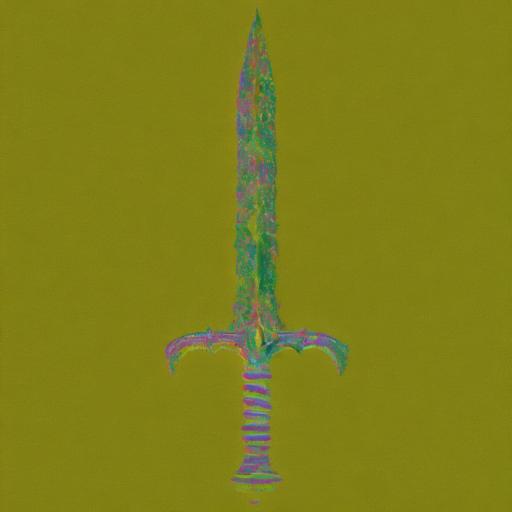}} 
        &
        \begin{tikzpicture}[every node/.style={anchor=north west,inner sep=0pt},x=1pt, y=-1pt,]  
             \node (fig1) at (0,0)
               {\fbox{\includegraphics[width=0.14\textwidth]{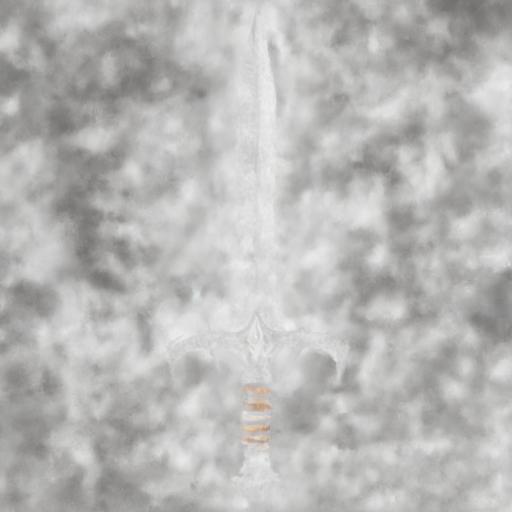}}};   
             \node (fig2) at (59,0)
               {\fbox{\includegraphics[width=0.065\textwidth]{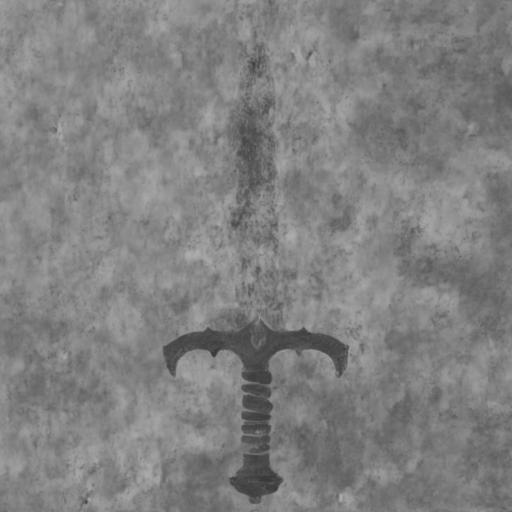}}};
             \node (fig3) at (59,30)
               {\fbox{\includegraphics[width=0.065\textwidth]{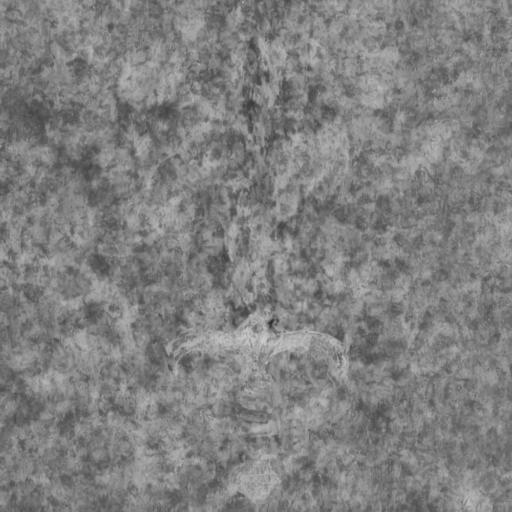}}};
        \end{tikzpicture}
        &
        \fbox{\includegraphics[width=0.14\textwidth]{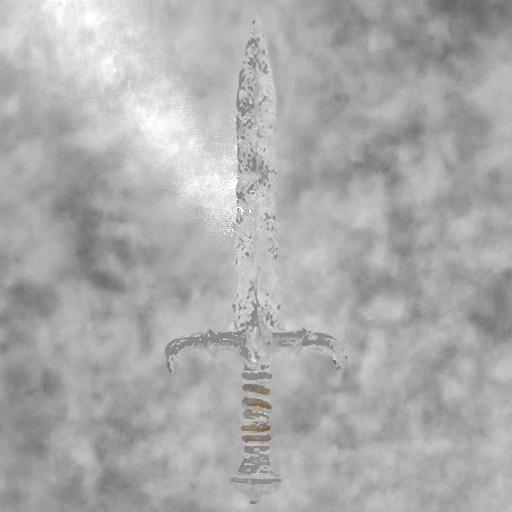}} 
        &
        \fbox{\includegraphics[width=0.14\textwidth]{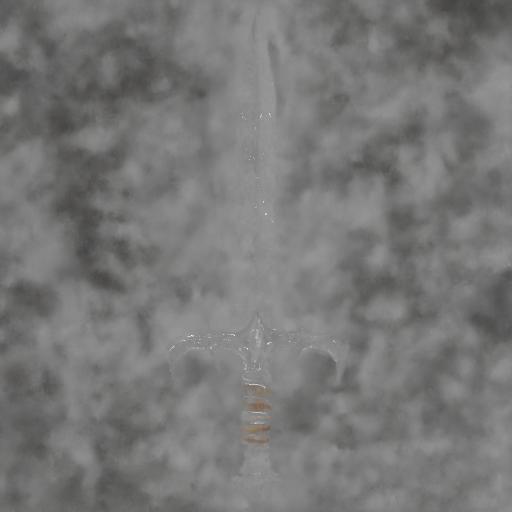}} 
        
        \hspace{6pt}
        &
        \hspace{6pt}
        
        \fbox{\includegraphics[width=0.14\textwidth]{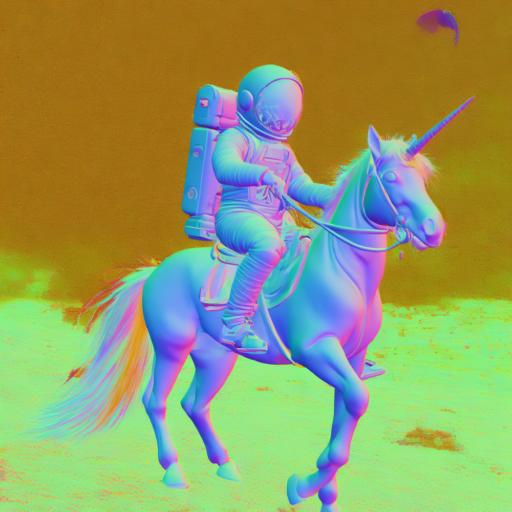}} 
        &
        \begin{tikzpicture}[every node/.style={anchor=north west,inner sep=0pt},x=1pt, y=-1pt,]  
             \node (fig1) at (0,0)
               {\fbox{\includegraphics[width=0.14\textwidth]{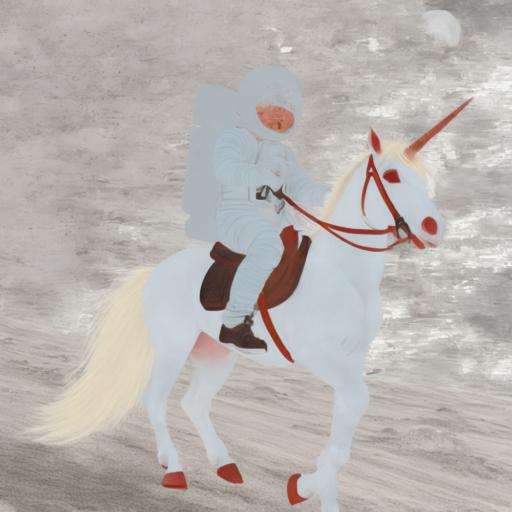}}};   
             \node (fig2) at (59,0)
               {\fbox{\includegraphics[width=0.065\textwidth]{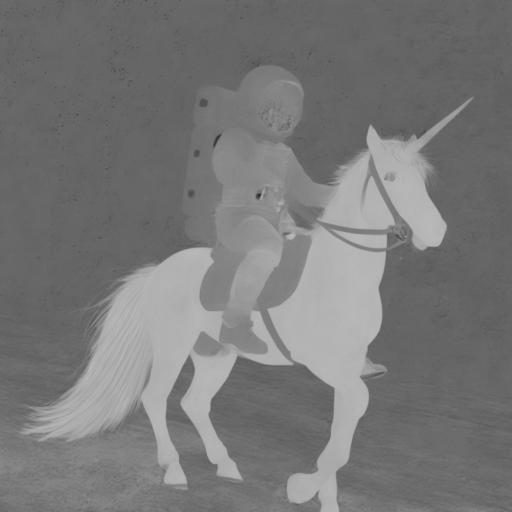}}};
             \node (fig3) at (59,30)
               {\fbox{\includegraphics[width=0.065\textwidth]{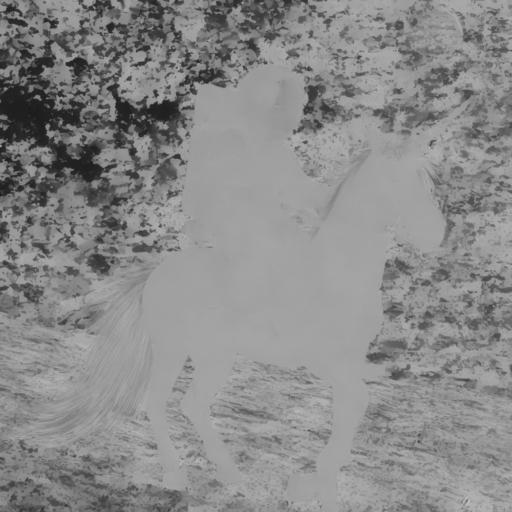}}};
        \end{tikzpicture}
        &
        \fbox{\includegraphics[width=0.14\textwidth]{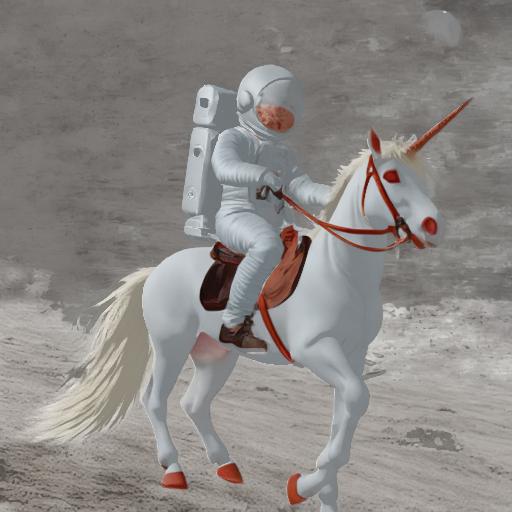}} 
        &
        \fbox{\includegraphics[width=0.14\textwidth]{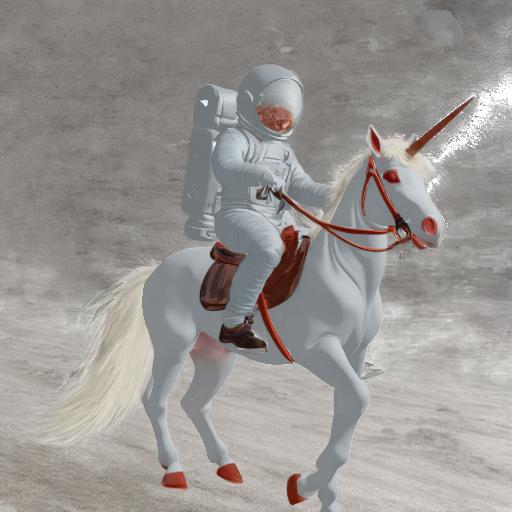}} 
        \\

        \midrule
        
        \rotatebox{90}{Ours}
        &
        \fbox{\includegraphics[width=0.14\textwidth]{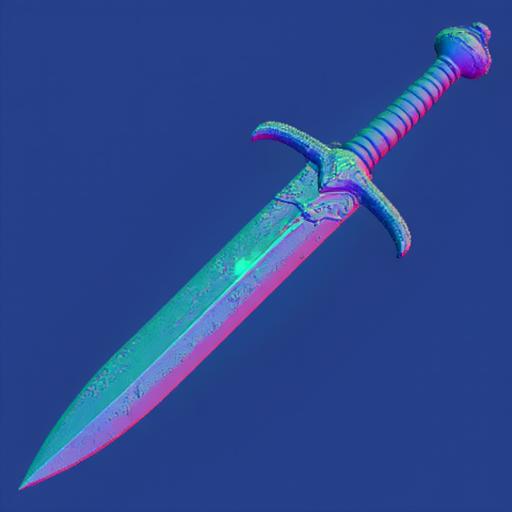}}
        &
        \begin{tikzpicture}[every node/.style={anchor=north west,inner sep=0pt},x=1pt, y=-1pt,]  
             \node (fig1) at (0,0)
               {\fbox{\includegraphics[width=0.14\textwidth]{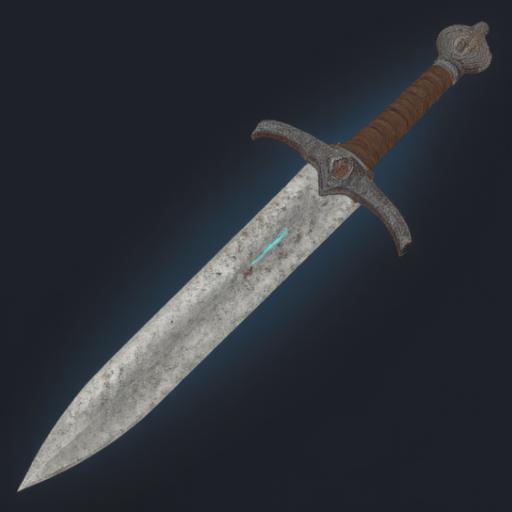}}};   
             \node (fig2) at (59,0)
               {\fbox{\includegraphics[width=0.065\textwidth]{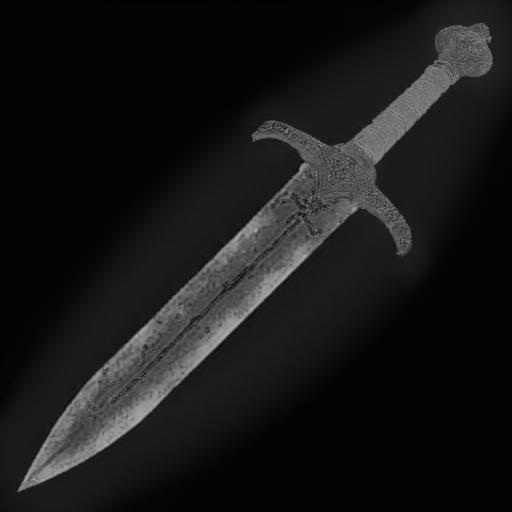}}};
             \node (fig3) at (59,30)
               {\fbox{\includegraphics[width=0.065\textwidth]{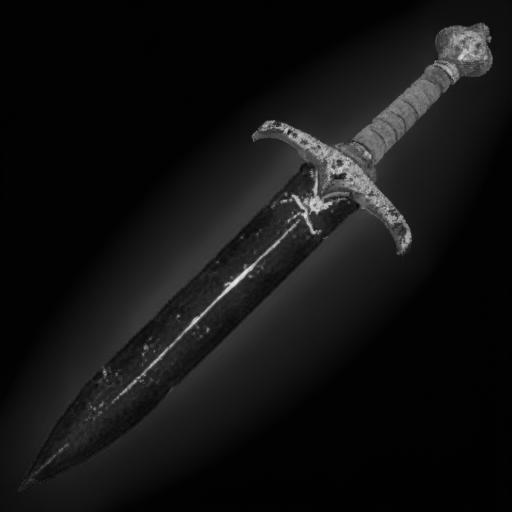}}};
        \end{tikzpicture}
        &
        \fbox{\includegraphics[width=0.14\textwidth]{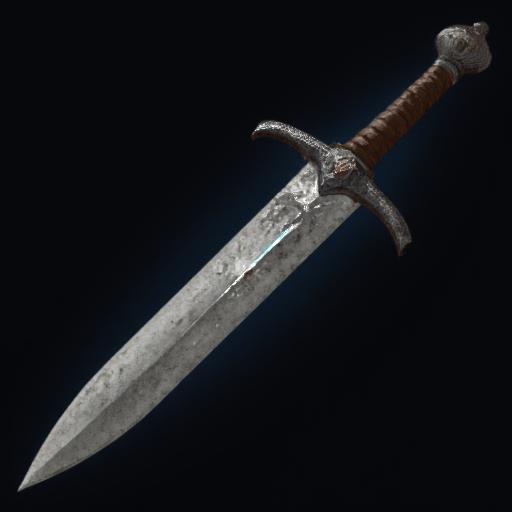}}
        &
        \fbox{\includegraphics[width=0.14\textwidth]{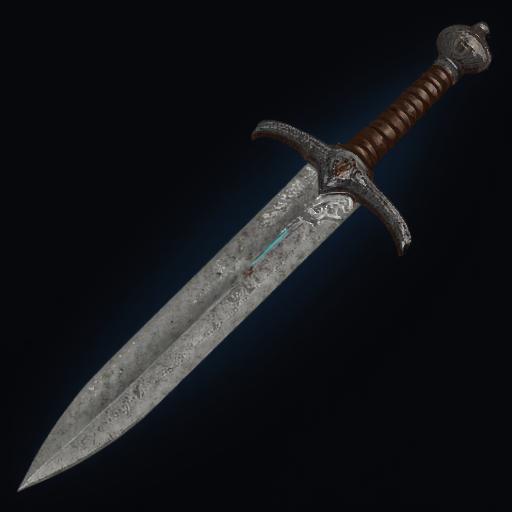}}
        
        \hspace{6pt}
        &
        \hspace{6pt}
        
        \fbox{\includegraphics[width=0.14\textwidth]{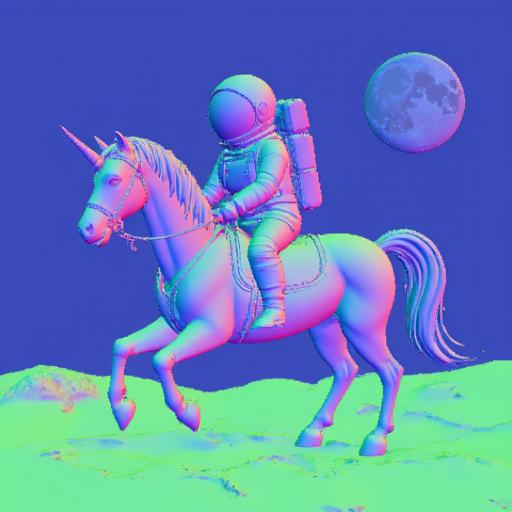}}
        &
        \begin{tikzpicture}[every node/.style={anchor=north west,inner sep=0pt},x=1pt, y=-1pt,]  
             \node (fig1) at (0,0)
               {\fbox{\includegraphics[width=0.14\textwidth]{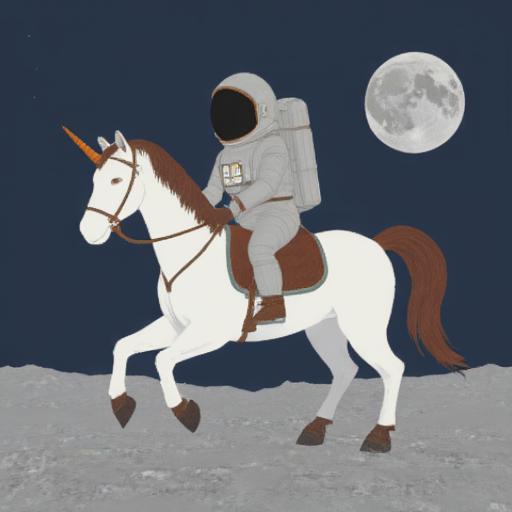}}};   
             \node (fig2) at (59,0)
               {\fbox{\includegraphics[width=0.065\textwidth]{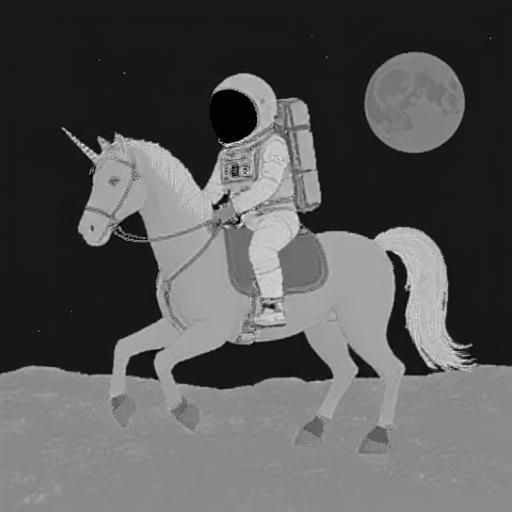}}};
             \node (fig3) at (59,30)
               {\fbox{\includegraphics[width=0.065\textwidth]{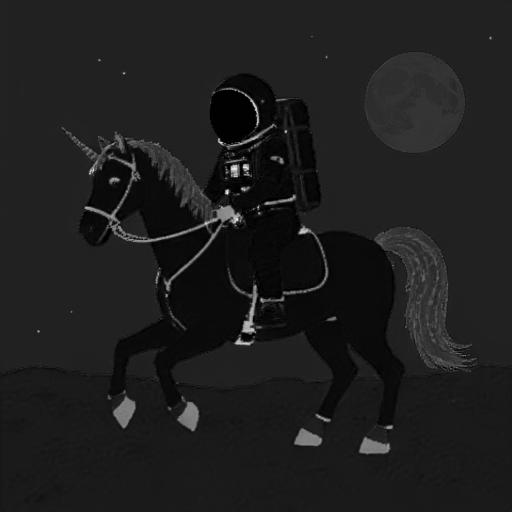}}};
        \end{tikzpicture}
        &
        \fbox{\includegraphics[width=0.14\textwidth]{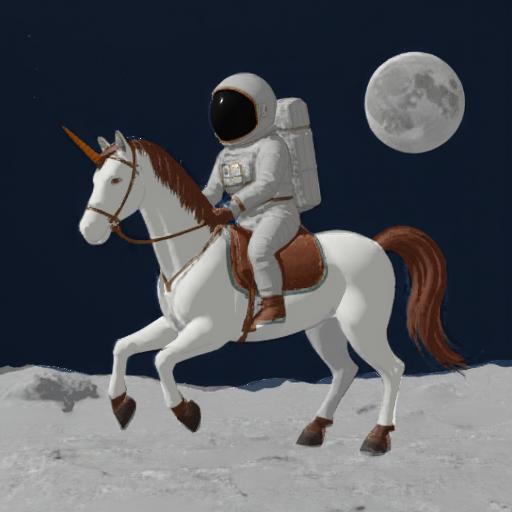}}
        &
        \fbox{\includegraphics[width=0.14\textwidth]{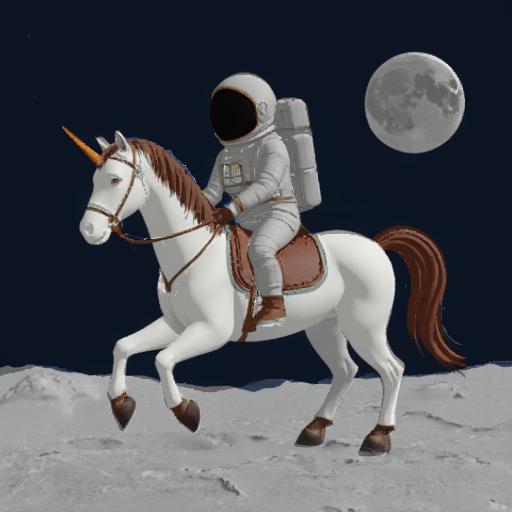}}
        \\

         &
        Normal &
        Material &
        Lighting 1 &
        Lighting 2 &
        
        Normal &
        Material &
        Lighting 1 &
        Lighting 2 \\

        \midrule
        \midrule
        
        \rotatebox{90}{IID}
        &
        \fbox{\includegraphics[width=0.14\textwidth]{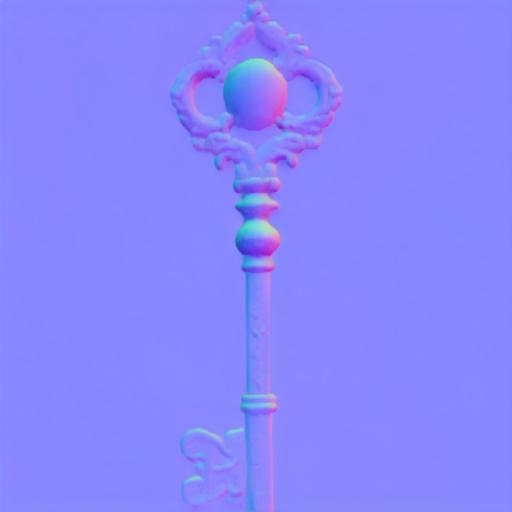}} 
        &
        \begin{tikzpicture}[every node/.style={anchor=north west,inner sep=0pt},x=1pt, y=-1pt,]  
             \node (fig1) at (0,0)
               {\fbox{\includegraphics[width=0.14\textwidth]{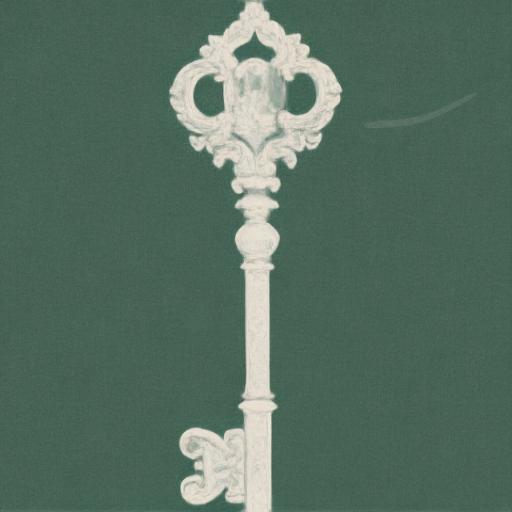}}};   
             \node (fig2) at (59,0)
               {\fbox{\includegraphics[width=0.065\textwidth]{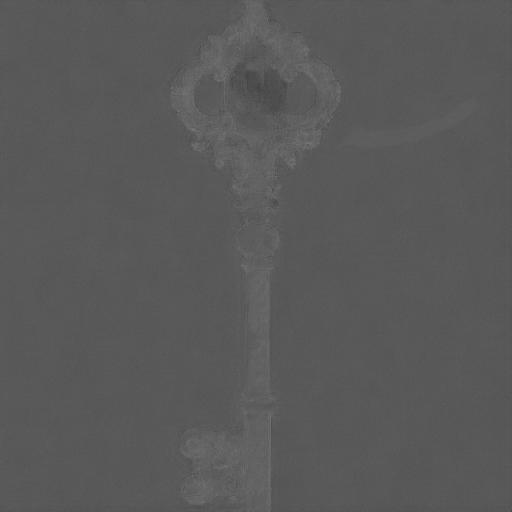}}};
             \node (fig3) at (59,30)
               {\fbox{\includegraphics[width=0.065\textwidth]{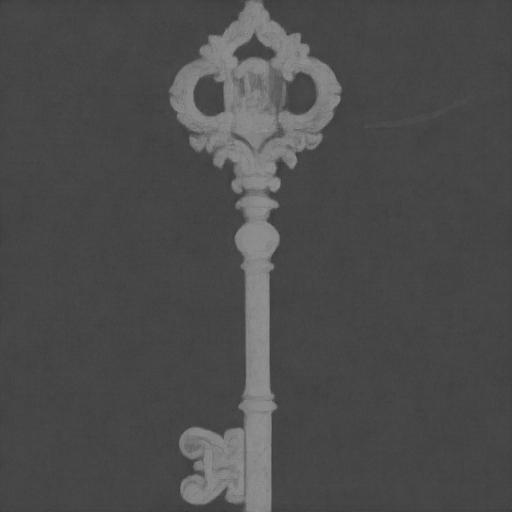}}};
        \end{tikzpicture}
        &
        \fbox{\includegraphics[width=0.14\textwidth]{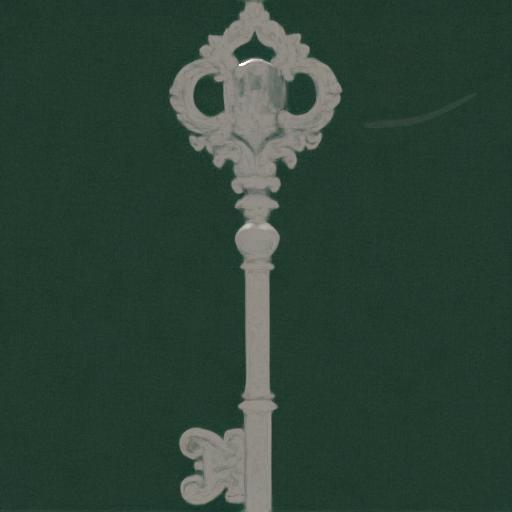}} 
        &
        \fbox{\includegraphics[width=0.14\textwidth]{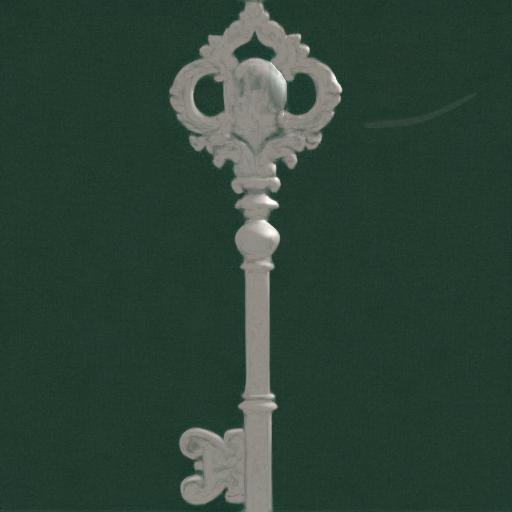}} 
        
        \hspace{6pt}
        &
        \hspace{6pt}
        
        \fbox{\includegraphics[width=0.14\textwidth]{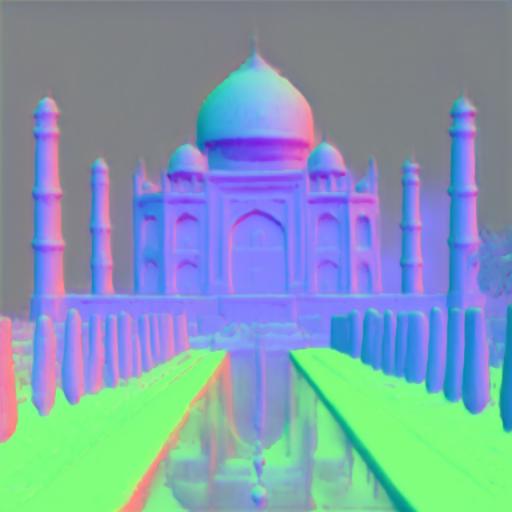}} 
        &
        \begin{tikzpicture}[every node/.style={anchor=north west,inner sep=0pt},x=1pt, y=-1pt,]  
             \node (fig1) at (0,0)
               {\fbox{\includegraphics[width=0.14\textwidth]{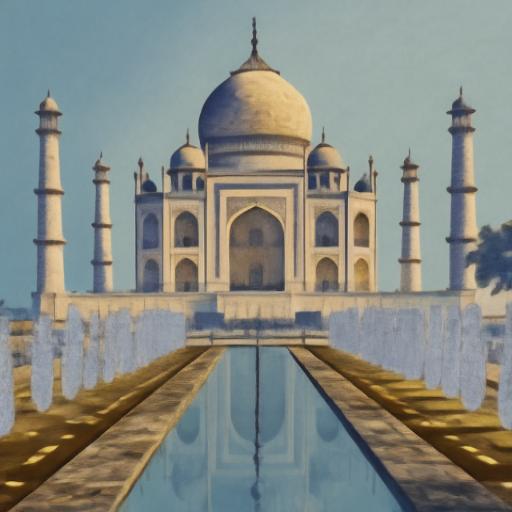}}};   
             \node (fig2) at (59,0)
               {\fbox{\includegraphics[width=0.065\textwidth]{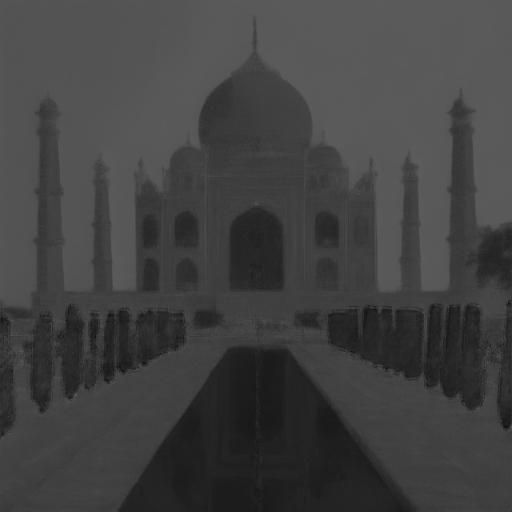}}};
             \node (fig3) at (59,30)
               {\fbox{\includegraphics[width=0.065\textwidth]{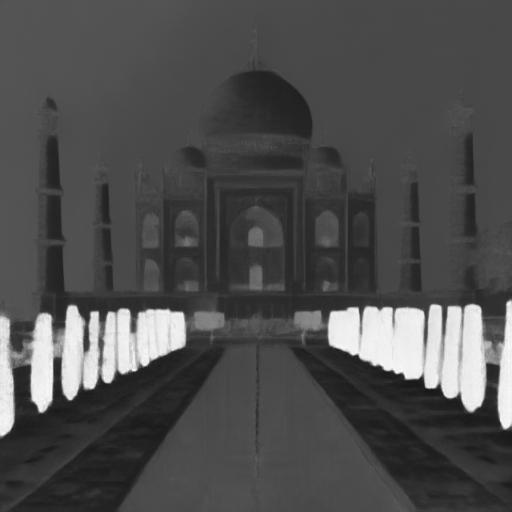}}};
        \end{tikzpicture}
        &
        \fbox{\includegraphics[width=0.14\textwidth]{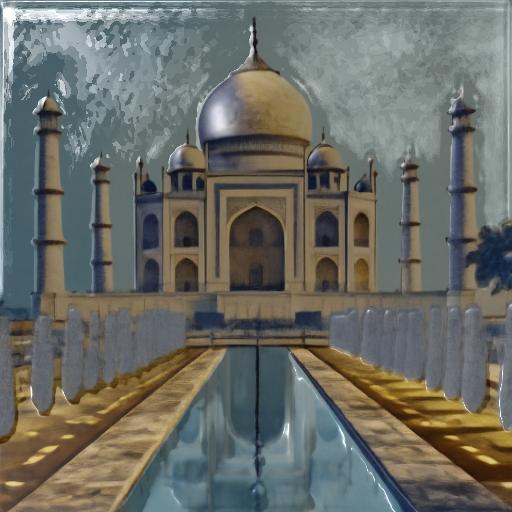}} 
        &
        \fbox{\includegraphics[width=0.14\textwidth]{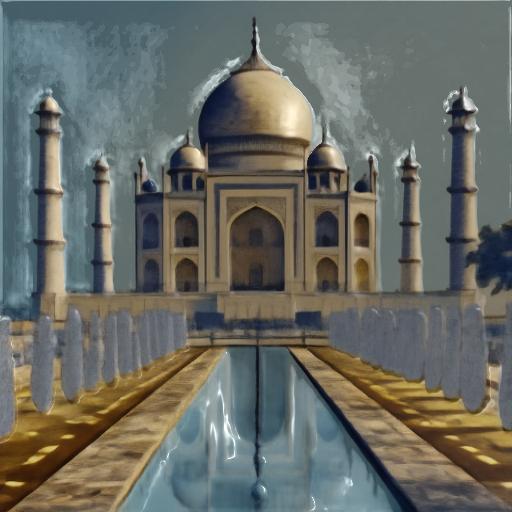}} 
        \\
        
        \rotatebox{90}{RGBX}
        &
        \fbox{\includegraphics[width=0.14\textwidth]{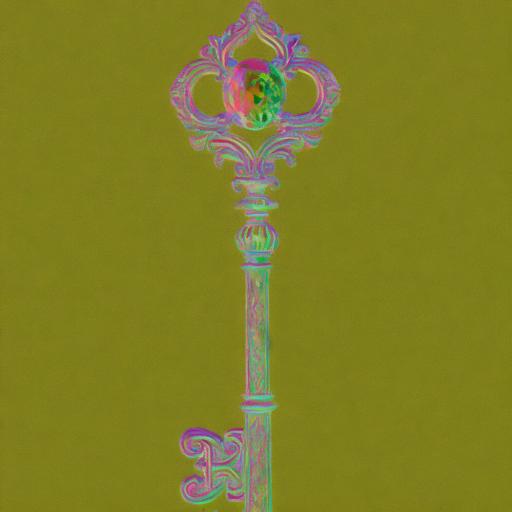}} 
        &
        \begin{tikzpicture}[every node/.style={anchor=north west,inner sep=0pt},x=1pt, y=-1pt,]  
             \node (fig1) at (0,0)
               {\fbox{\includegraphics[width=0.14\textwidth]{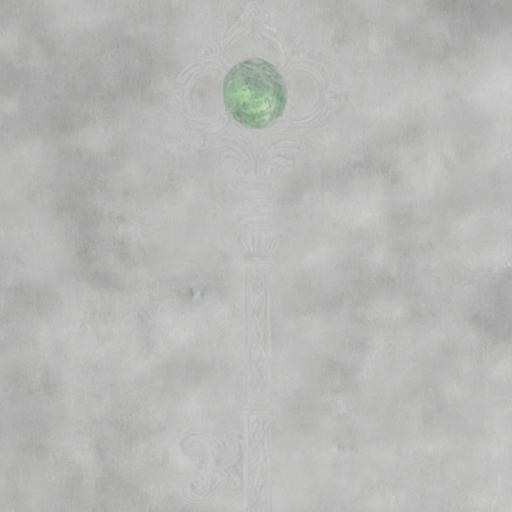}}};   
             \node (fig2) at (59,0)
               {\fbox{\includegraphics[width=0.065\textwidth]{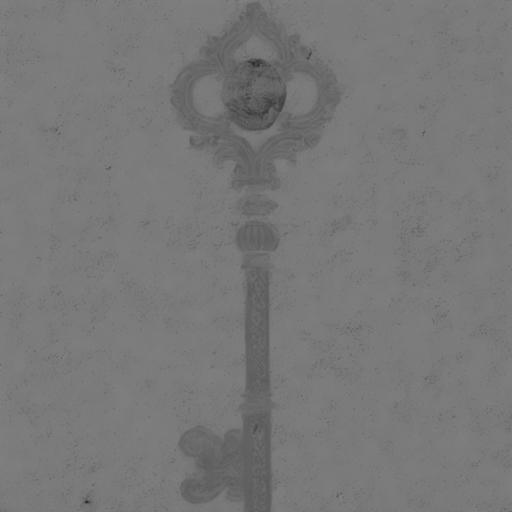}}};
             \node (fig3) at (59,30)
               {\fbox{\includegraphics[width=0.065\textwidth]{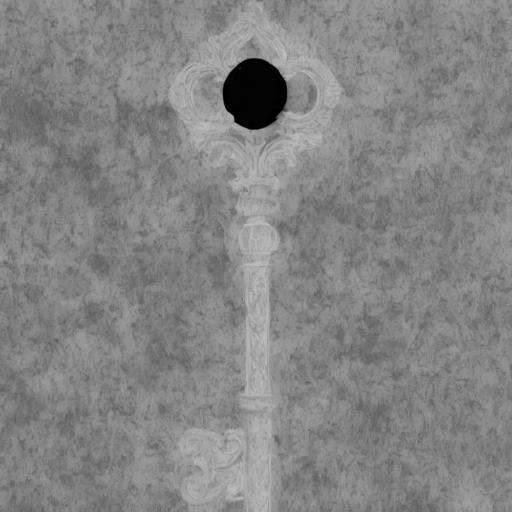}}};
        \end{tikzpicture}
        &
        \fbox{\includegraphics[width=0.14\textwidth]{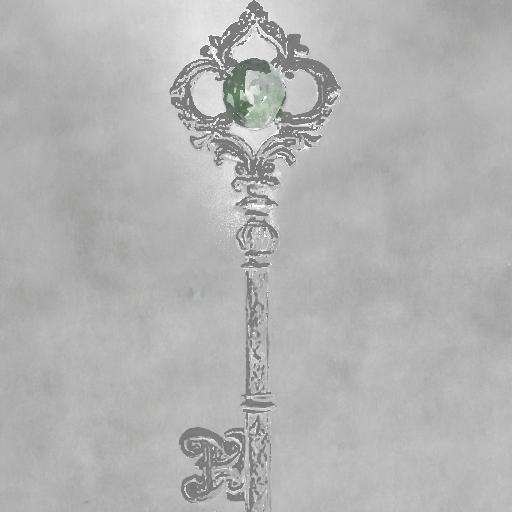}} 
        &
        \fbox{\includegraphics[width=0.14\textwidth]{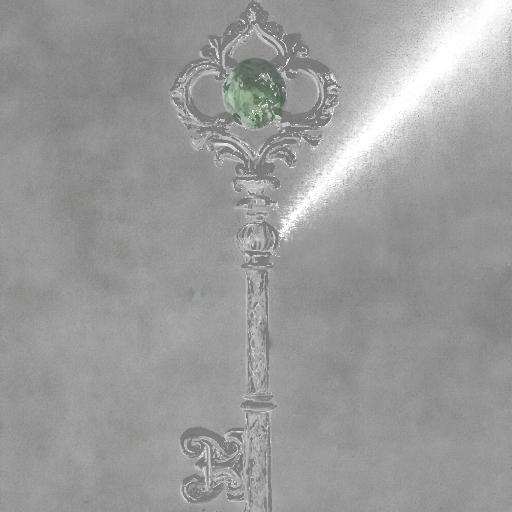}} 
        
        \hspace{6pt}
        &
        \hspace{6pt}
        
        \fbox{\includegraphics[width=0.14\textwidth]{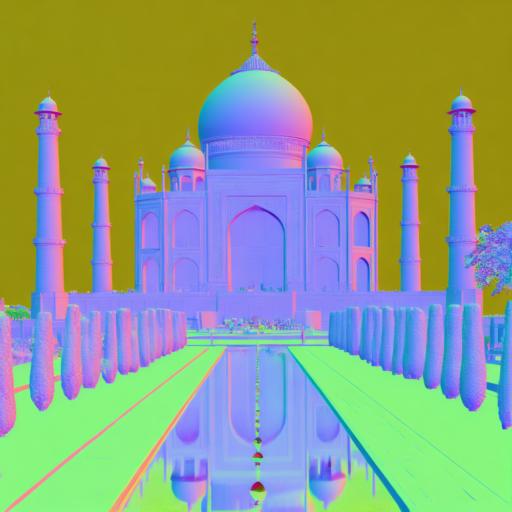}} 
        &
        \begin{tikzpicture}[every node/.style={anchor=north west,inner sep=0pt},x=1pt, y=-1pt,]  
             \node (fig1) at (0,0)
               {\fbox{\includegraphics[width=0.14\textwidth]{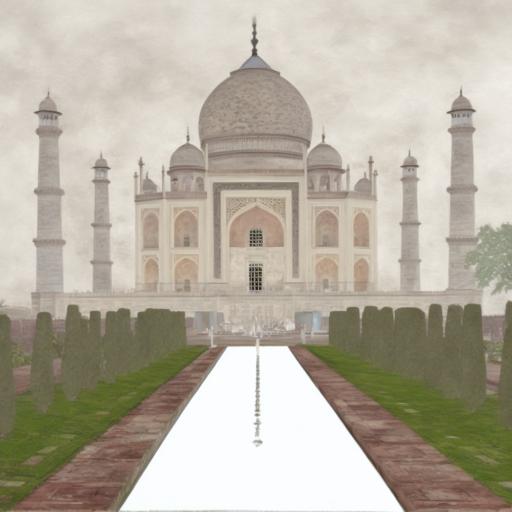}}};   
             \node (fig2) at (59,0)
               {\fbox{\includegraphics[width=0.065\textwidth]{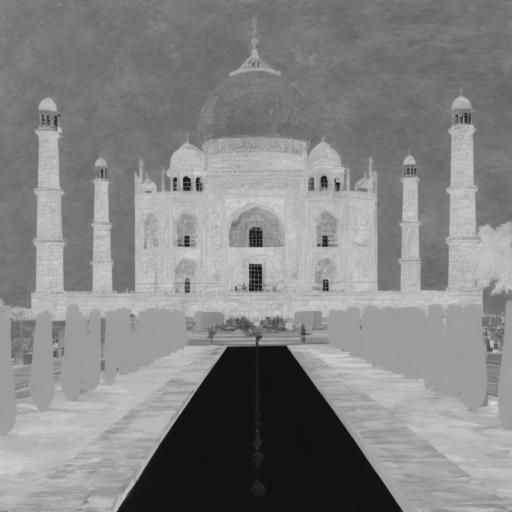}}};
             \node (fig3) at (59,30)
               {\fbox{\includegraphics[width=0.065\textwidth]{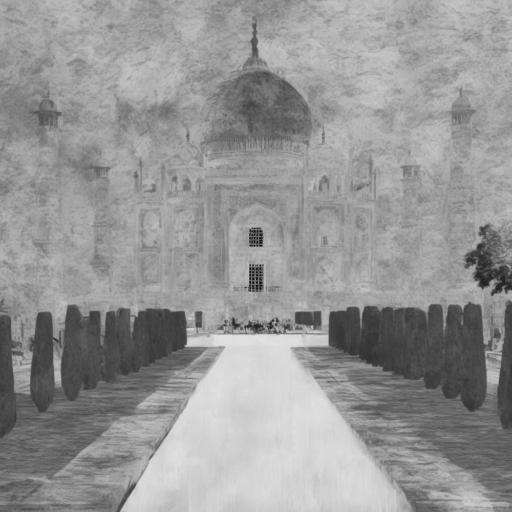}}};
        \end{tikzpicture}
        &
        \fbox{\includegraphics[width=0.14\textwidth]{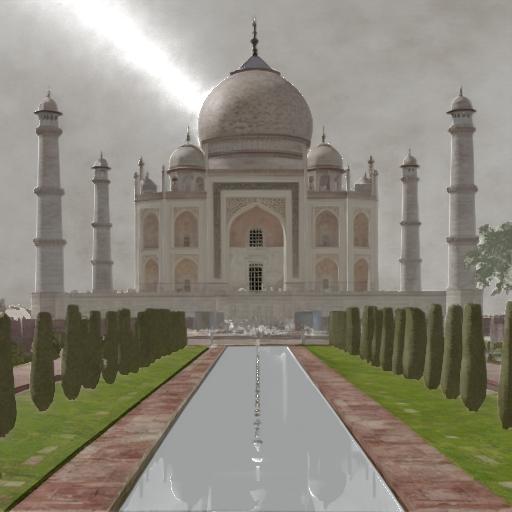}} 
        &
        \fbox{\includegraphics[width=0.14\textwidth]{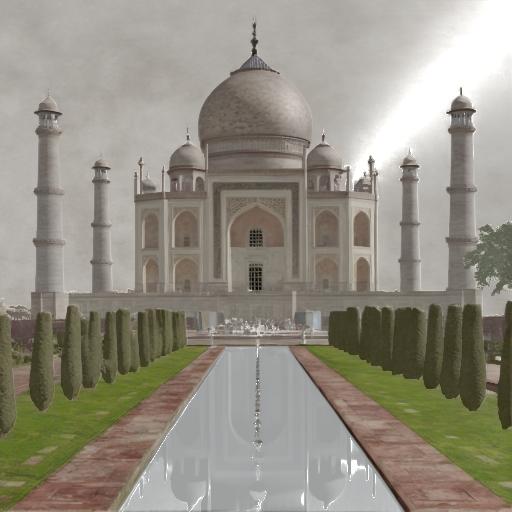}} 
        \\

        \midrule
        
        \rotatebox{90}{Ours}
        &
        \fbox{\includegraphics[width=0.14\textwidth]{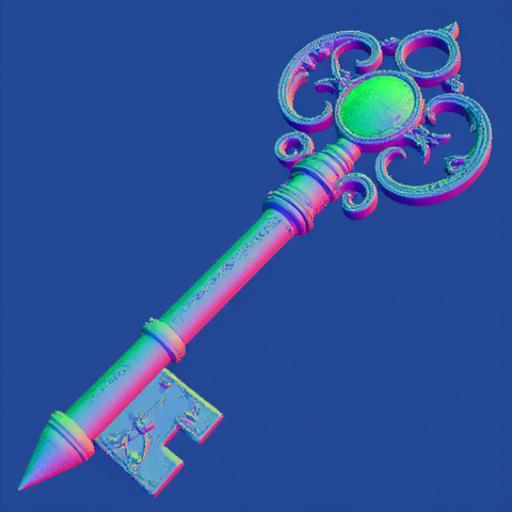}}
        &
        \begin{tikzpicture}[every node/.style={anchor=north west,inner sep=0pt},x=1pt, y=-1pt,]  
             \node (fig1) at (0,0)
               {\fbox{\includegraphics[width=0.14\textwidth]{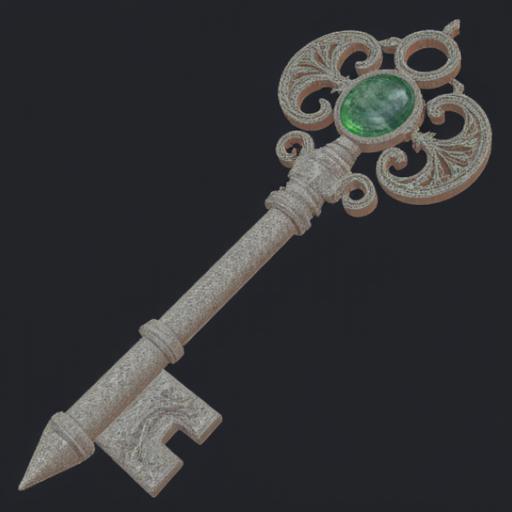}}};   
             \node (fig2) at (59,0)
               {\fbox{\includegraphics[width=0.065\textwidth]{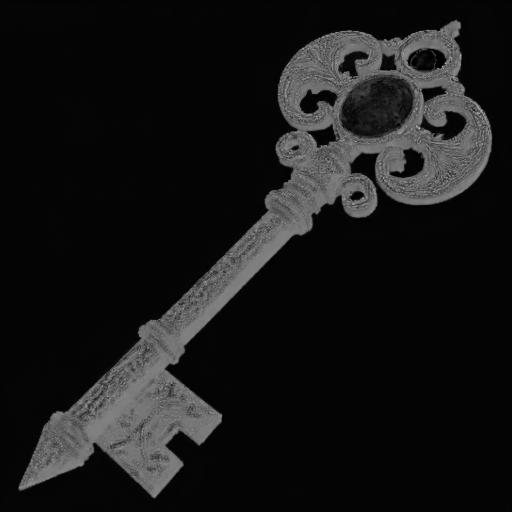}}};
             \node (fig3) at (59,30)
               {\fbox{\includegraphics[width=0.065\textwidth]{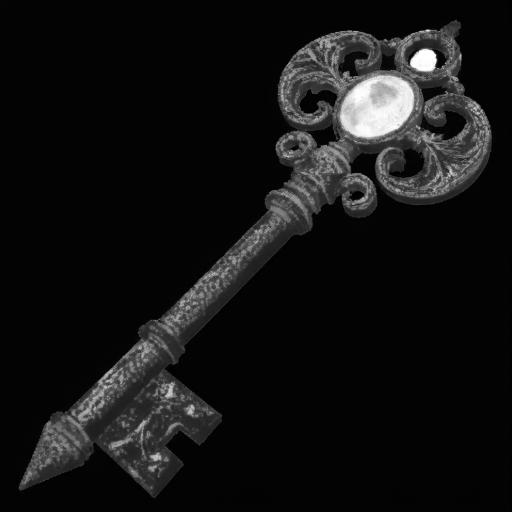}}};
        \end{tikzpicture}
        &
        \fbox{\includegraphics[width=0.14\textwidth]{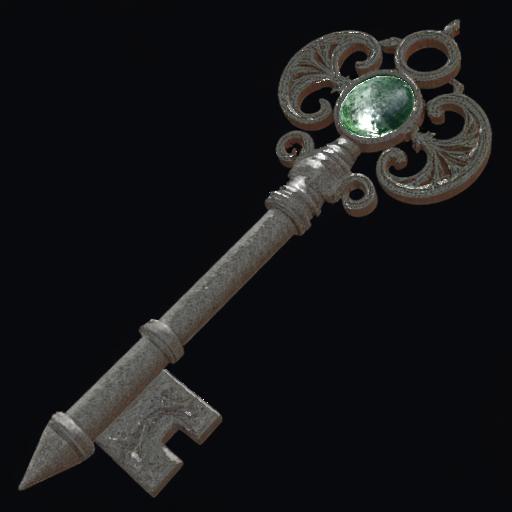}}
        &
        \fbox{\includegraphics[width=0.14\textwidth]{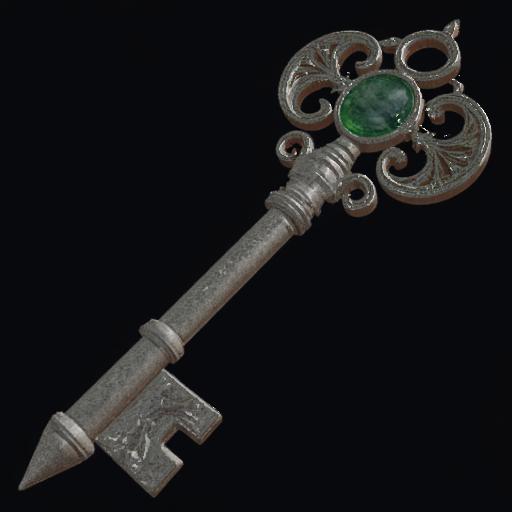}}
        
        \hspace{6pt}
        &
        \hspace{6pt}
        
        \fbox{\includegraphics[width=0.14\textwidth]{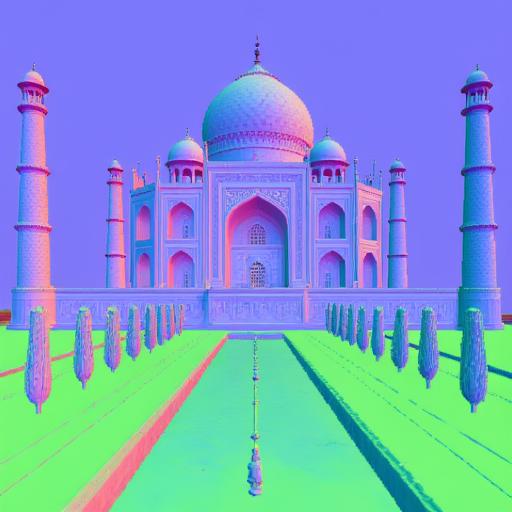}}
        &
        \begin{tikzpicture}[every node/.style={anchor=north west,inner sep=0pt},x=1pt, y=-1pt,]  
             \node (fig1) at (0,0)
               {\fbox{\includegraphics[width=0.14\textwidth]{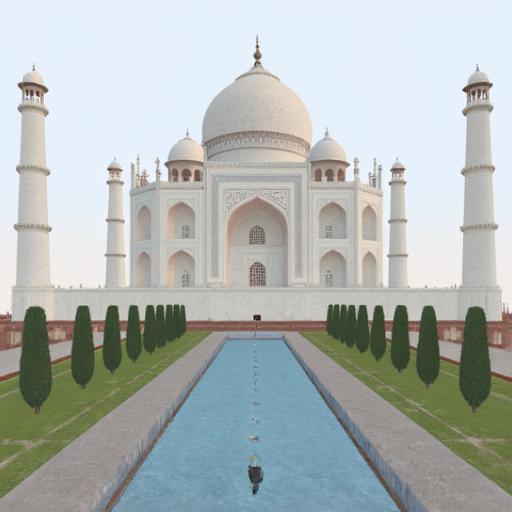}}};   
             \node (fig2) at (59,0)
               {\fbox{\includegraphics[width=0.065\textwidth]{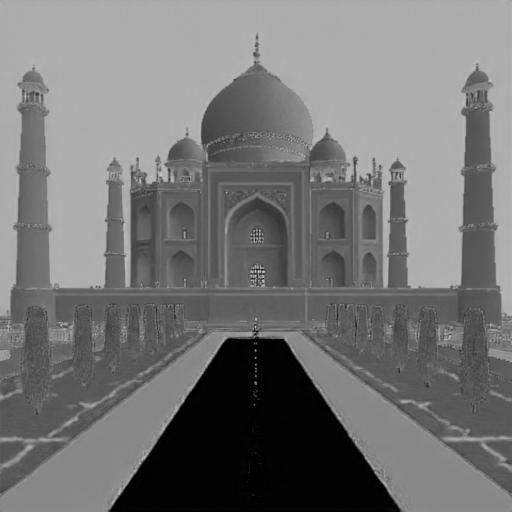}}};
             \node (fig3) at (59,30)
               {\fbox{\includegraphics[width=0.065\textwidth]{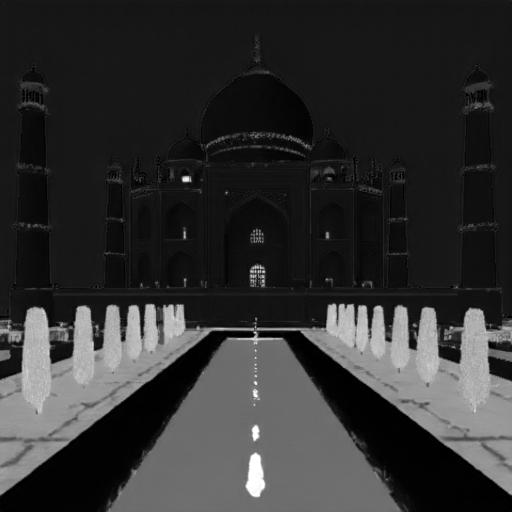}}};
        \end{tikzpicture}
        &
        \fbox{\includegraphics[width=0.14\textwidth]{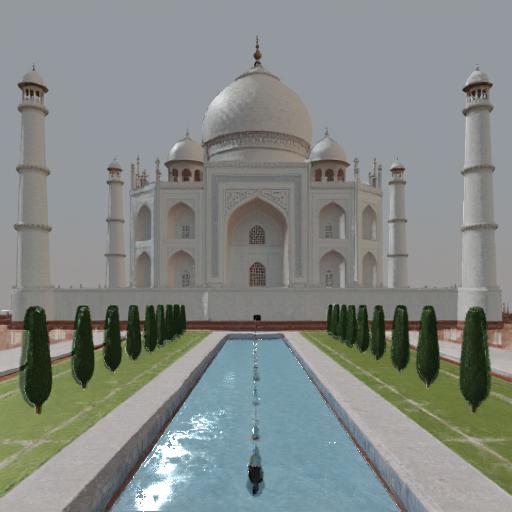}}
        &
        \fbox{\includegraphics[width=0.14\textwidth]{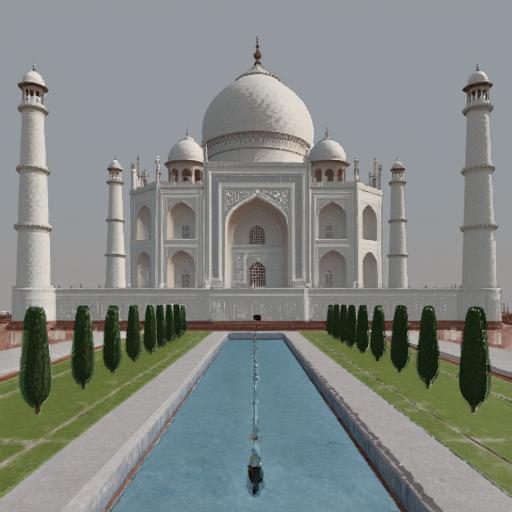}}
        \\
        
    \end{tabular}}

    \caption{\textbf{Additional rendering comparisons}. 
    We show sample PBR maps of our method and baselines as well as rendered RGB images under two different lighting conditions.
    We use a diverse set of text prompts to produce our PBR maps, as well as the input RGB images for the baseline methods.
    This highlights our models' capability to retain the generalized prior of the pretrained text-to-image model.
    Our method better captures the semantic meaning of the individual intrinsic properties.
    For example, there are no baked-in lighting effects in the albedo, and the metallic/roughness maps are sharper with more intricate details.
    This leads to more realistic renderings and lighting effects.
    }
    \label{fig:supp:comparisons}
\end{figure*}

%% file: figures/experiments/albedo_comparisons_more.tex
\begin{figure}[t]
    \centering
    \setlength\tabcolsep{1.25pt}
    \resizebox{0.7\columnwidth}{!}{
    \fboxsep=0pt
        \begin{tabular}{cccccc}
        IID \cite{IID}
        &
        \fbox{\includegraphics[width=0.13\textwidth]{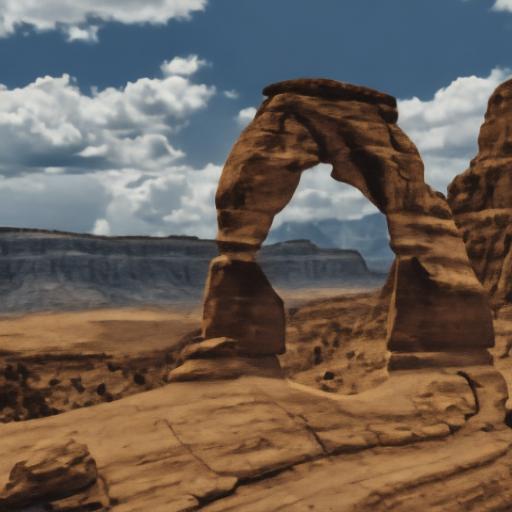}} 
        &
        \fbox{\includegraphics[width=0.13\textwidth]{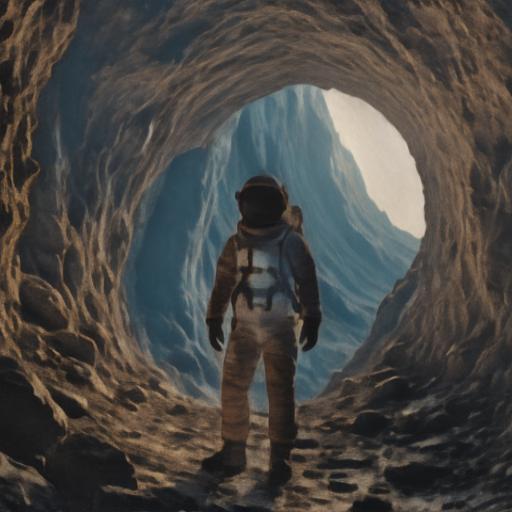}} 
        &
        \fbox{\includegraphics[width=0.13\textwidth]{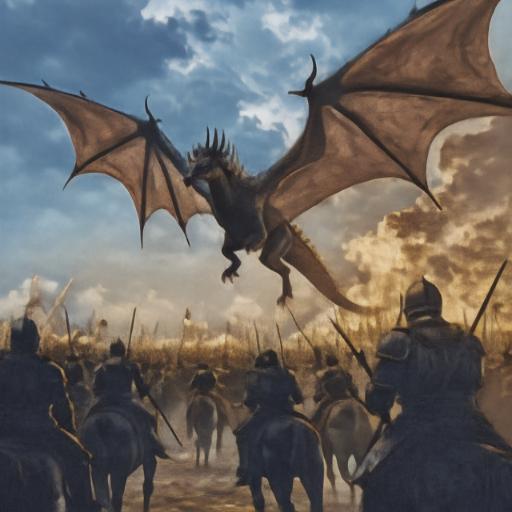}} 
        &
        \fbox{\includegraphics[width=0.13\textwidth]{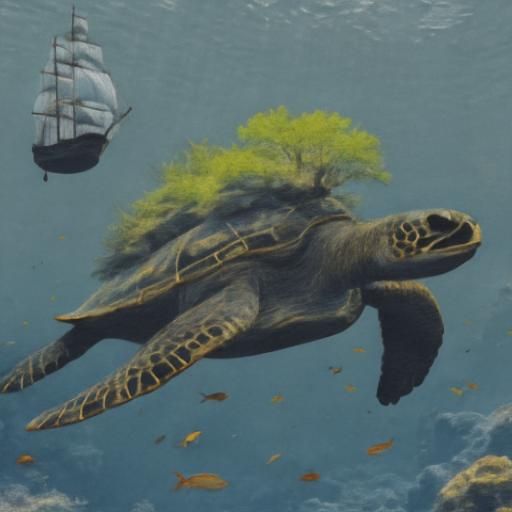}} 
        &
        \fbox{\includegraphics[width=0.13\textwidth]{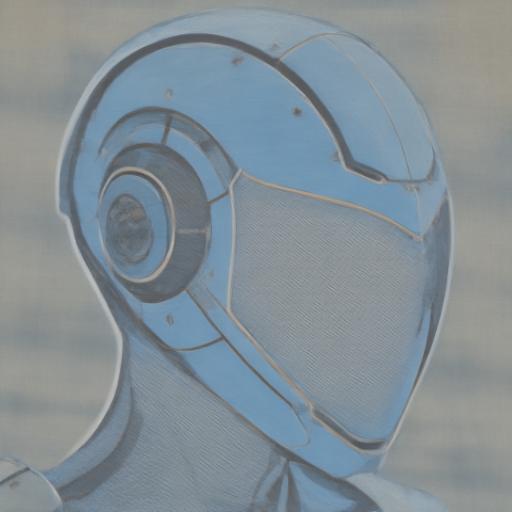}} 
        \\
        
        RGBX \cite{RGBX}
        &
        \fbox{\includegraphics[width=0.13\textwidth]{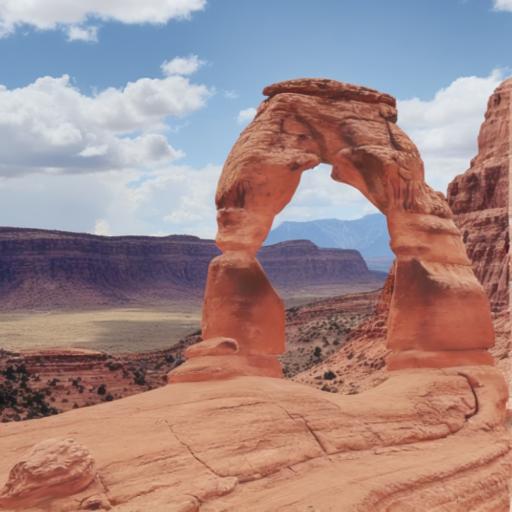}} 
        &
        \fbox{\includegraphics[width=0.13\textwidth]{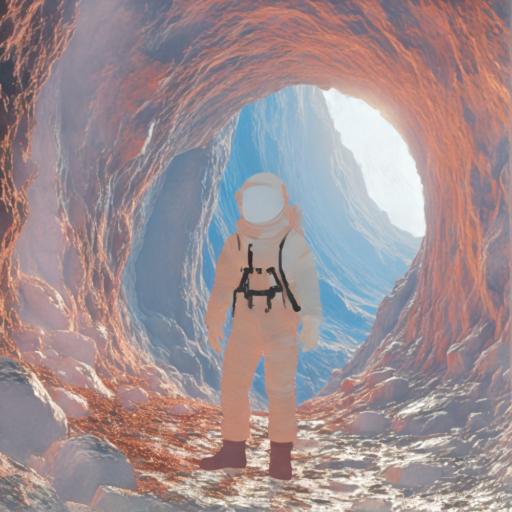}} 
        &
        \fbox{\includegraphics[width=0.13\textwidth]{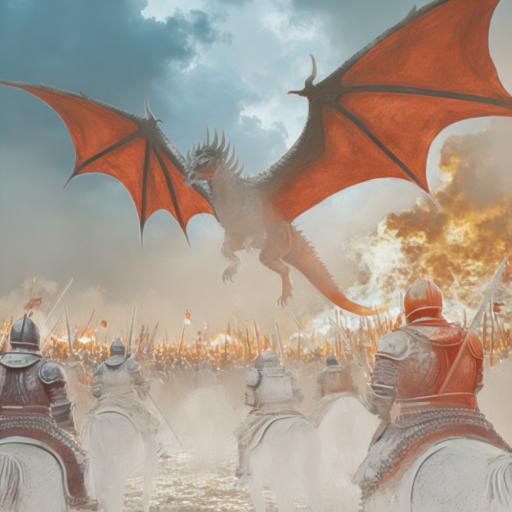}} 
        &
        \fbox{\includegraphics[width=0.13\textwidth]{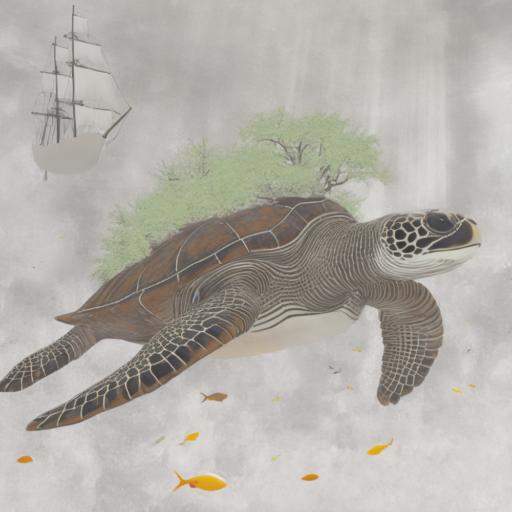}} 
        &
        \fbox{\includegraphics[width=0.13\textwidth]{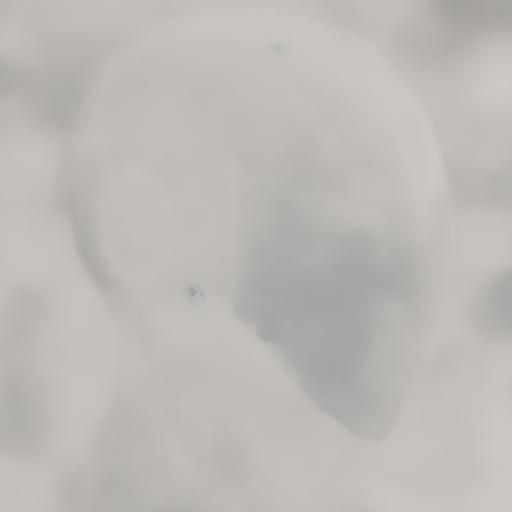}} 
        \\
        
        \begin{tabular}[b]{@{}c@{}}Colorful\\Shading \cite{careagaColorful}\end{tabular}
        &
        \fbox{\includegraphics[width=0.13\textwidth]{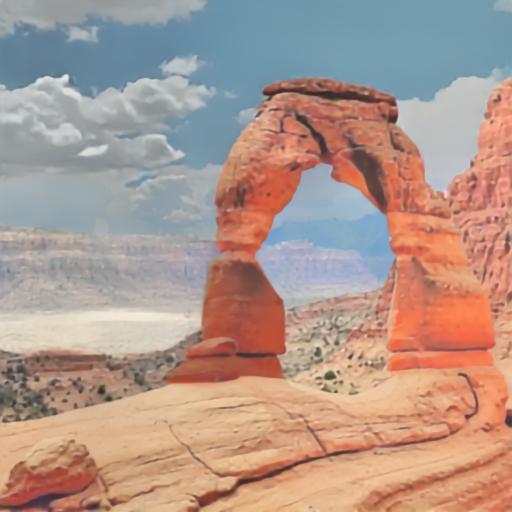}} 
        &
        \fbox{\includegraphics[width=0.13\textwidth]{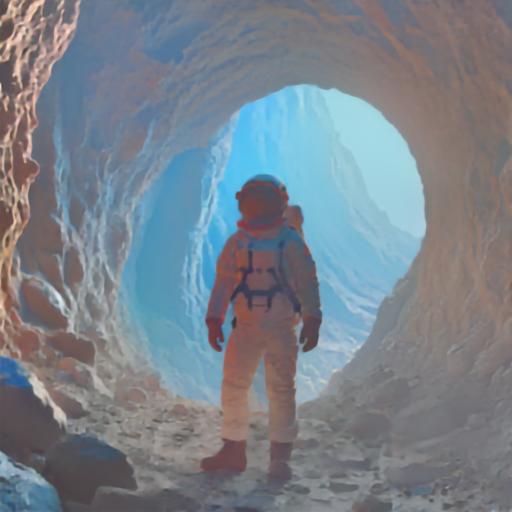}} 
        &
        \fbox{\includegraphics[width=0.13\textwidth]{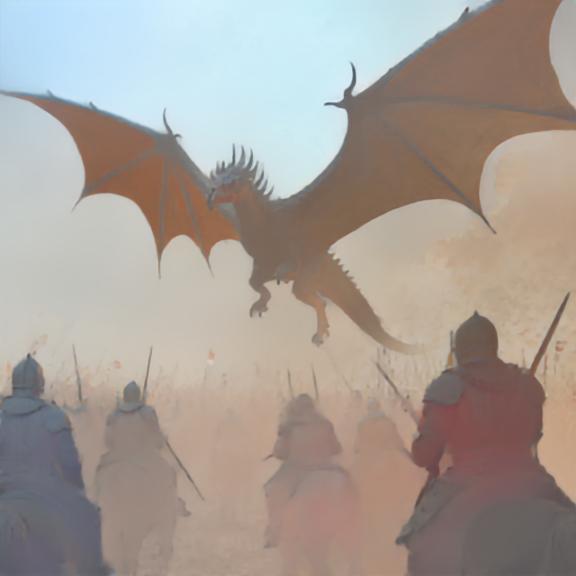}} 
        &
        \fbox{\includegraphics[width=0.13\textwidth]{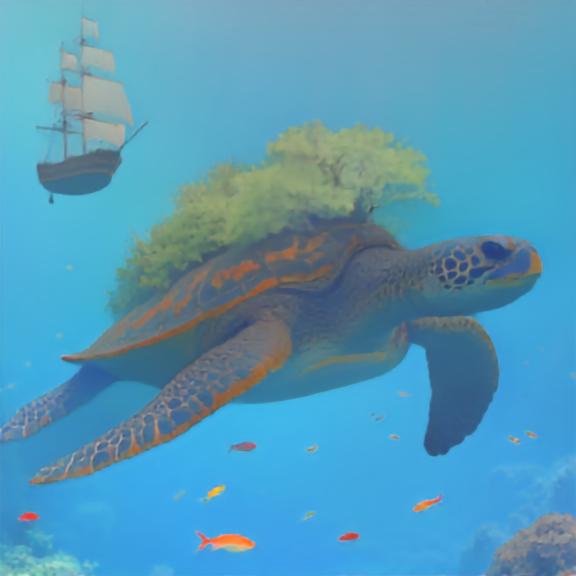}} 
        &
        \fbox{\includegraphics[width=0.13\textwidth]{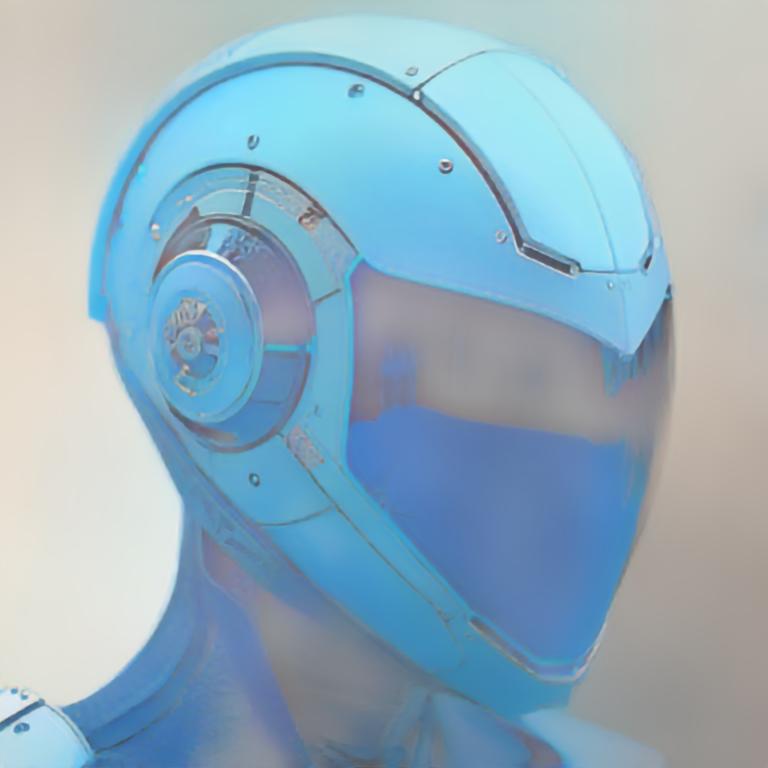}} 
        \\

        \midrule
        \begin{tabular}[b]{@{}c@{}}IID \cite{IID}\\w/ FLUX-LoRA \cite{flux2023, LoRA}\\SmallData\end{tabular}
        &
        \fbox{\includegraphics[width=0.13\textwidth]{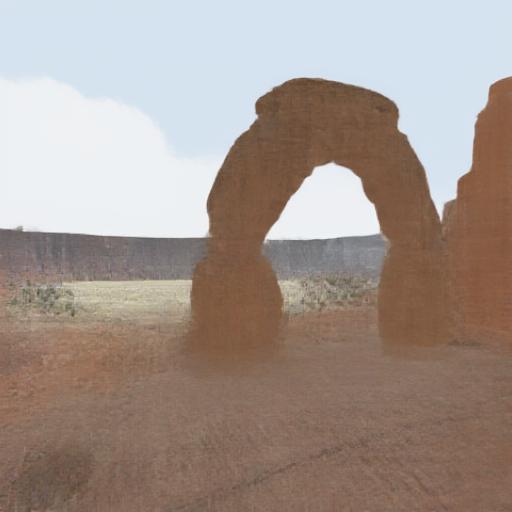}} 
        &
        \fbox{\includegraphics[width=0.13\textwidth]{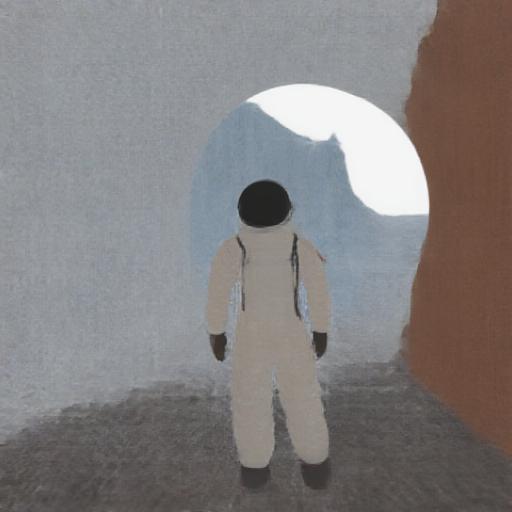}} 
        &
        \fbox{\includegraphics[width=0.13\textwidth]{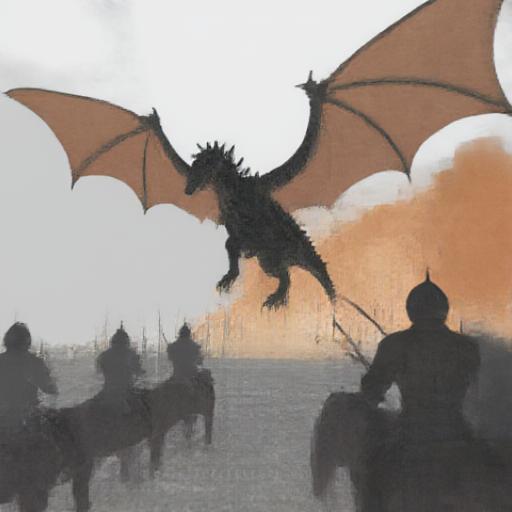}} 
        &
        \fbox{\includegraphics[width=0.13\textwidth]{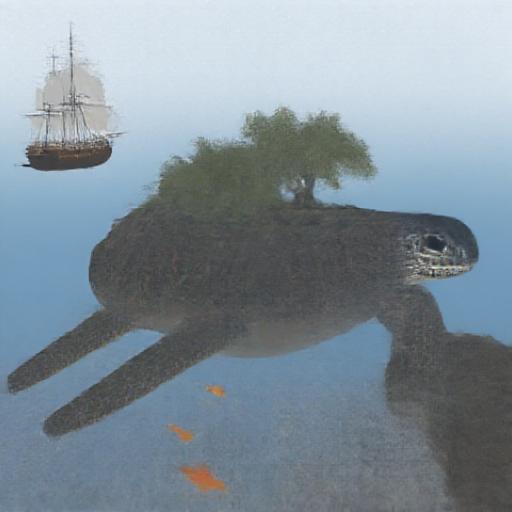}} 
        &
        \fbox{\includegraphics[width=0.13\textwidth]{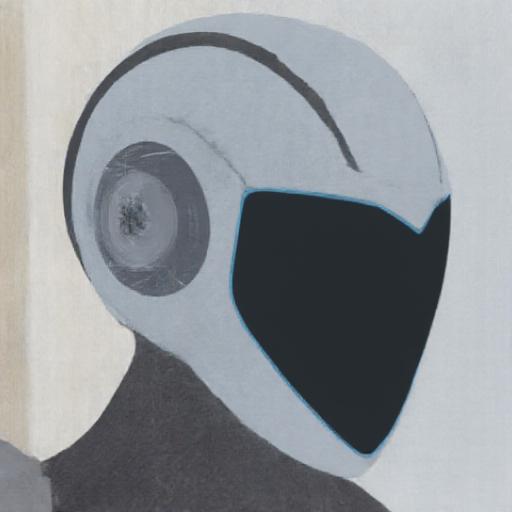}} 
        \\

        \midrule
        \begin{tabular}[b]{@{}c@{}}Ours\\w/ Im-Cond \end{tabular}
        &
        \fbox{\includegraphics[width=0.13\textwidth]{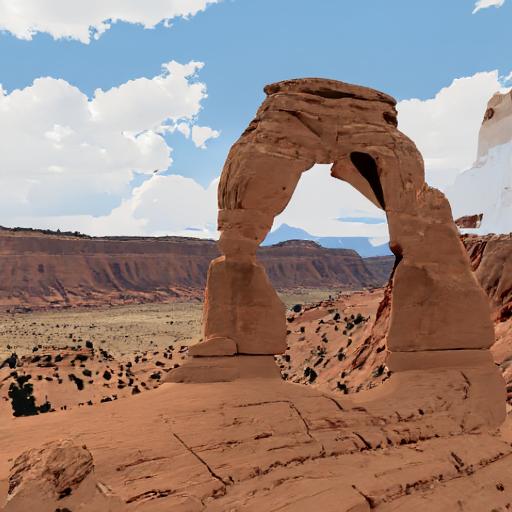}} 
        &
        \fbox{\includegraphics[width=0.13\textwidth]{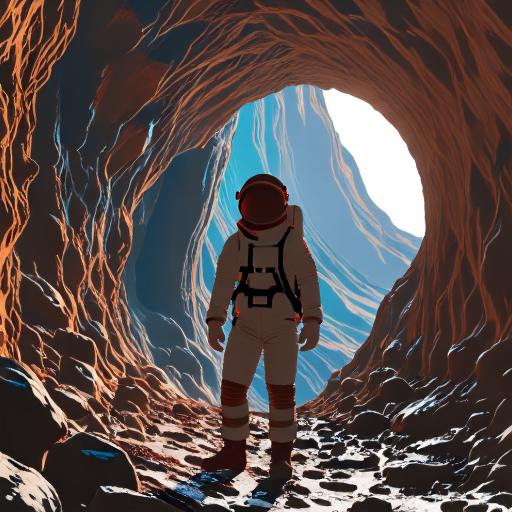}} 
        &
        \fbox{\includegraphics[width=0.13\textwidth]{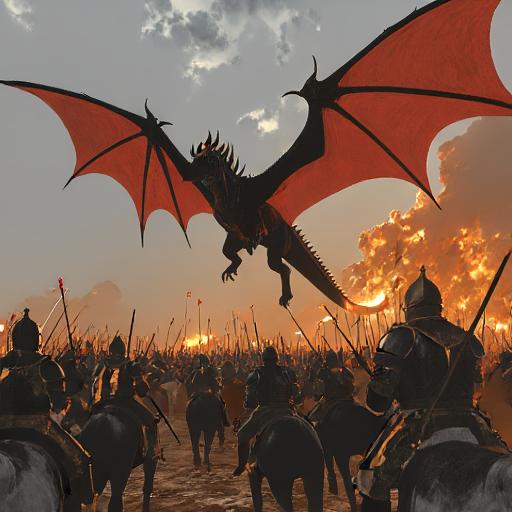}} 
        &
        \fbox{\includegraphics[width=0.13\textwidth]{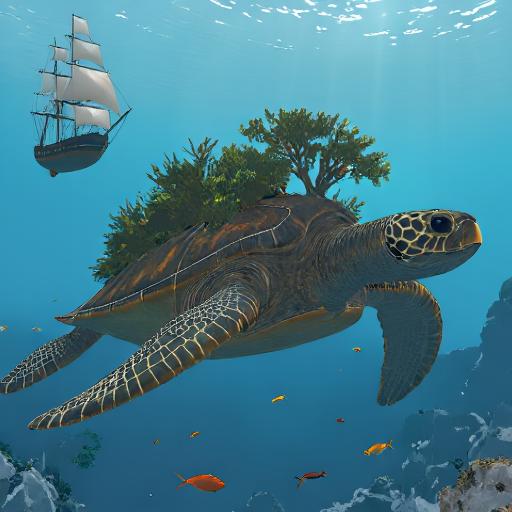}} 
        &
        \fbox{\includegraphics[width=0.13\textwidth]{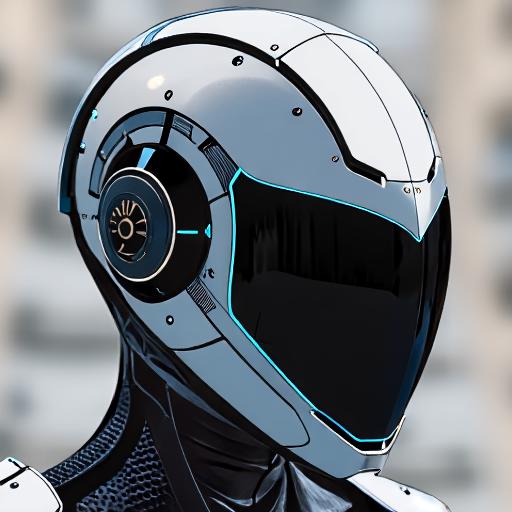}} 
        \\
        
        Ours
        &
        \fbox{\includegraphics[width=0.13\textwidth]{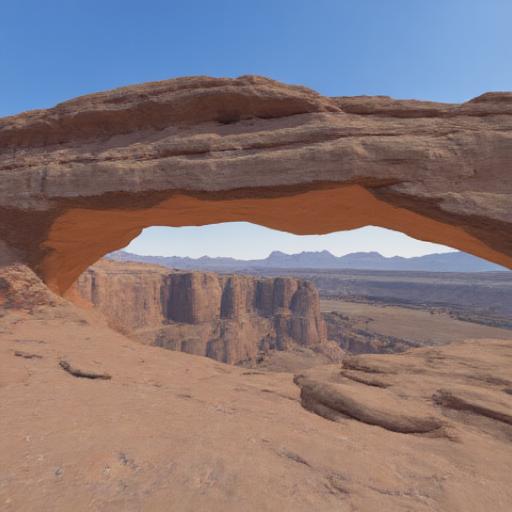}} 
        &
        \fbox{\includegraphics[width=0.13\textwidth]{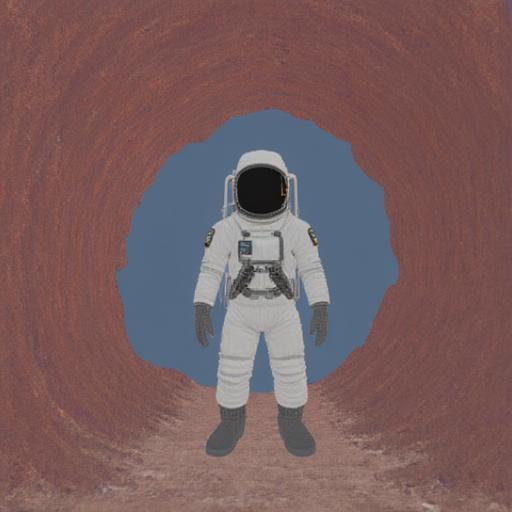}} 
        &
        \fbox{\includegraphics[width=0.13\textwidth]{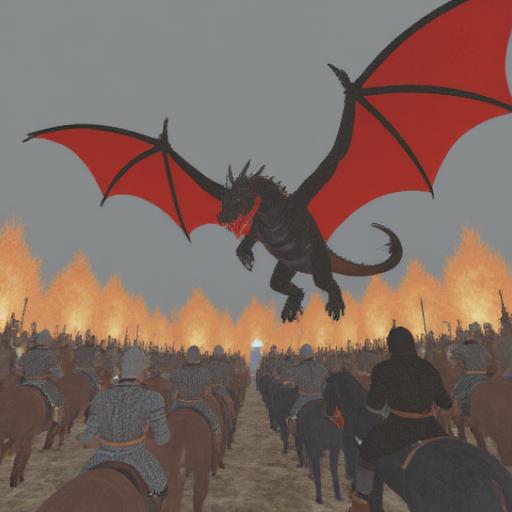}} 
        &
        \fbox{\includegraphics[width=0.13\textwidth]{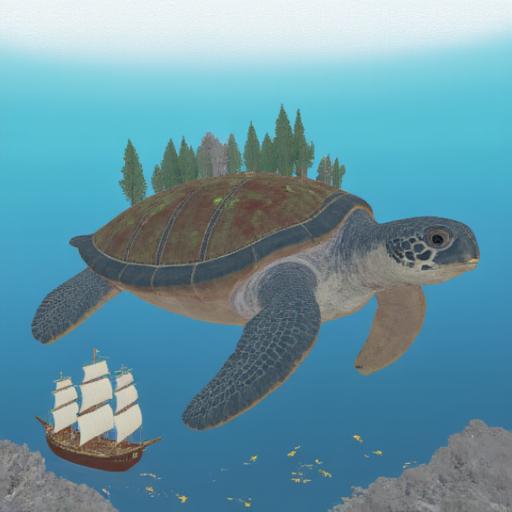}} 
        &
        \fbox{\includegraphics[width=0.13\textwidth]{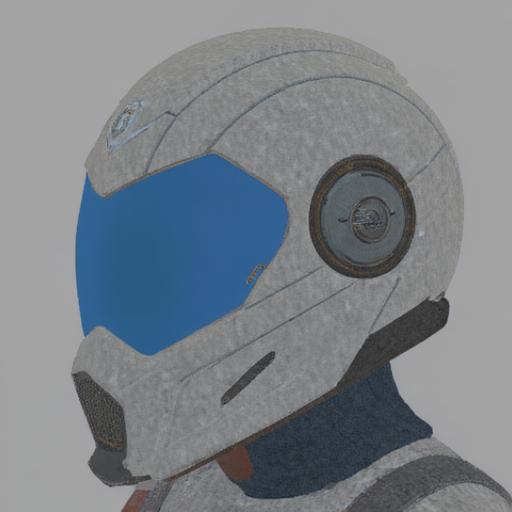}} 
        \\
    \end{tabular}}

    \caption{\textbf{Additional albedo comparisons}. 
    We show albedo images of our method and baselines corresponding to the same text prompt in each column.
    Our albedo images have less baked-in shadows and reflections, which is desirable for downstream tasks, such as physically-based rendering.
    }
    \label{fig:supp:albedo_comparisons}
\end{figure}

%% file: figures/applications/scenetex_more.tex
\begin{figure*}[p]
    \centering
    \setlength\tabcolsep{1.25pt}
    \resizebox{\textwidth}{!}{
    \fboxsep=0pt
    \begin{tabular}{ccccc|c}
        \rotatebox{90}{{\footnotesize View 1}}
        &
        \fbox{\includegraphics[width=0.15\textwidth,trim={10cm 0 10cm 0},clip]{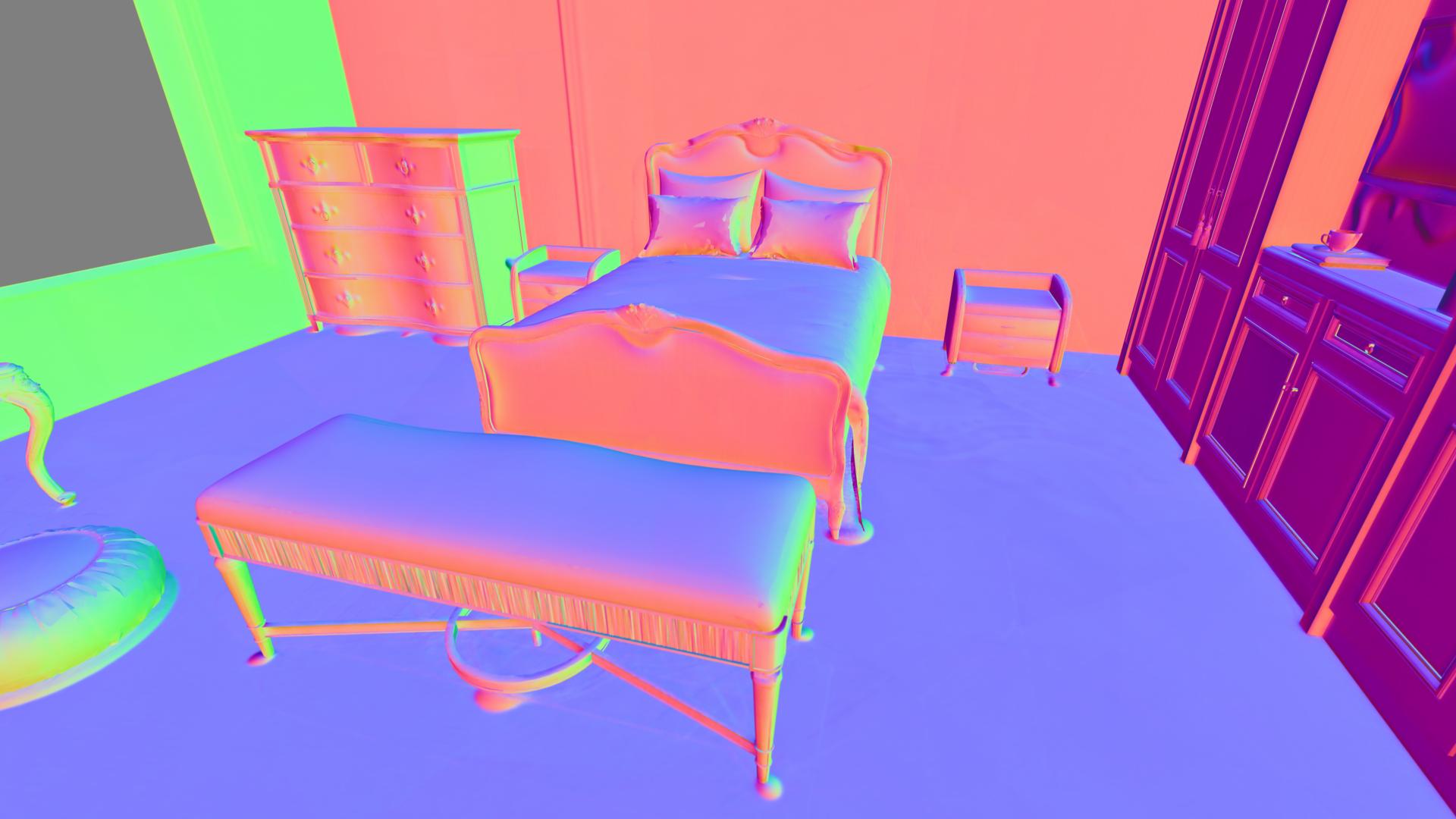}}
        &
        \fbox{\includegraphics[width=0.15\textwidth,trim={10cm 0 10cm 0},clip]{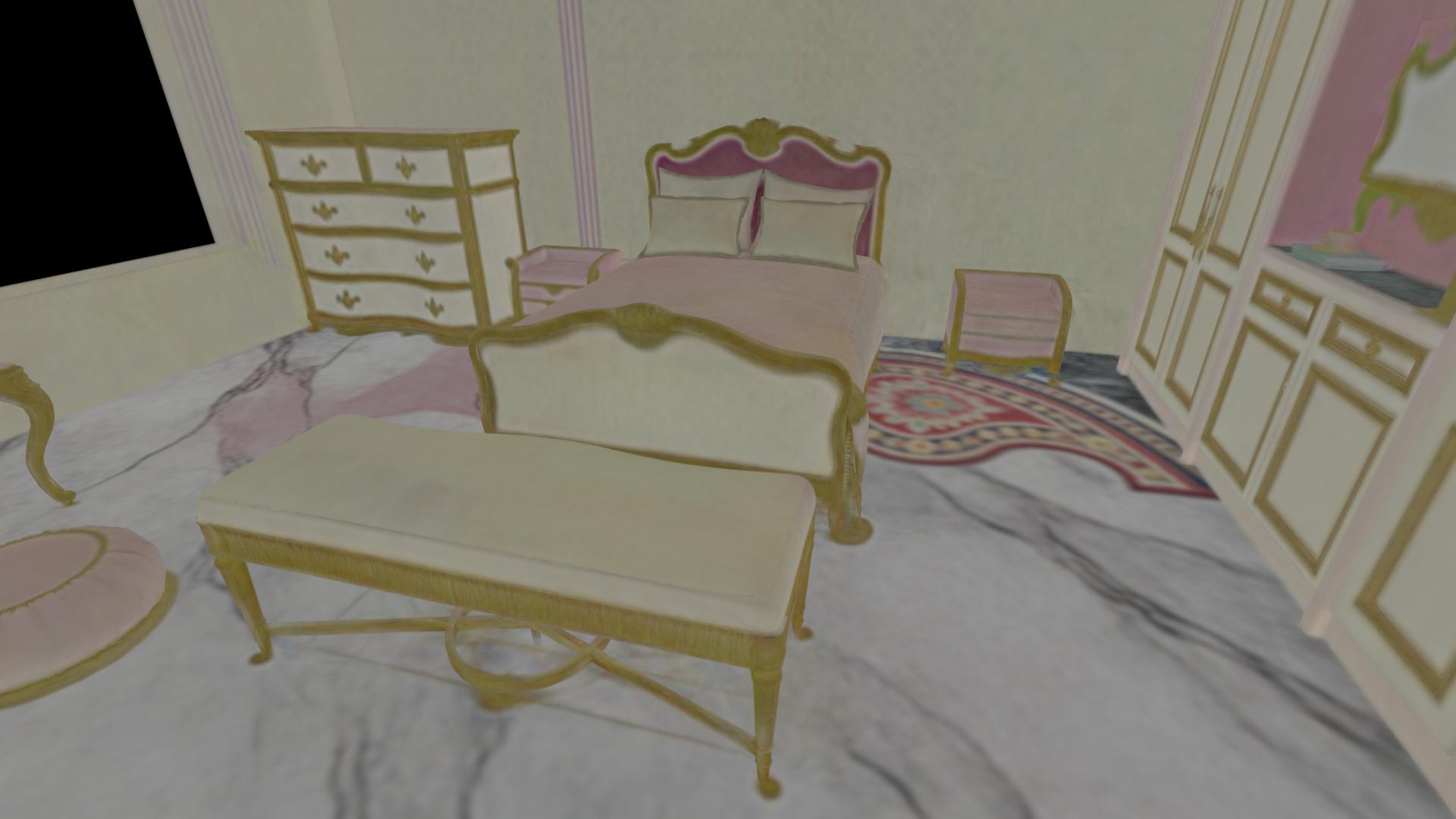}} 
        &
        \fbox{\includegraphics[width=0.15\textwidth,trim={10cm 0 10cm 0},clip]{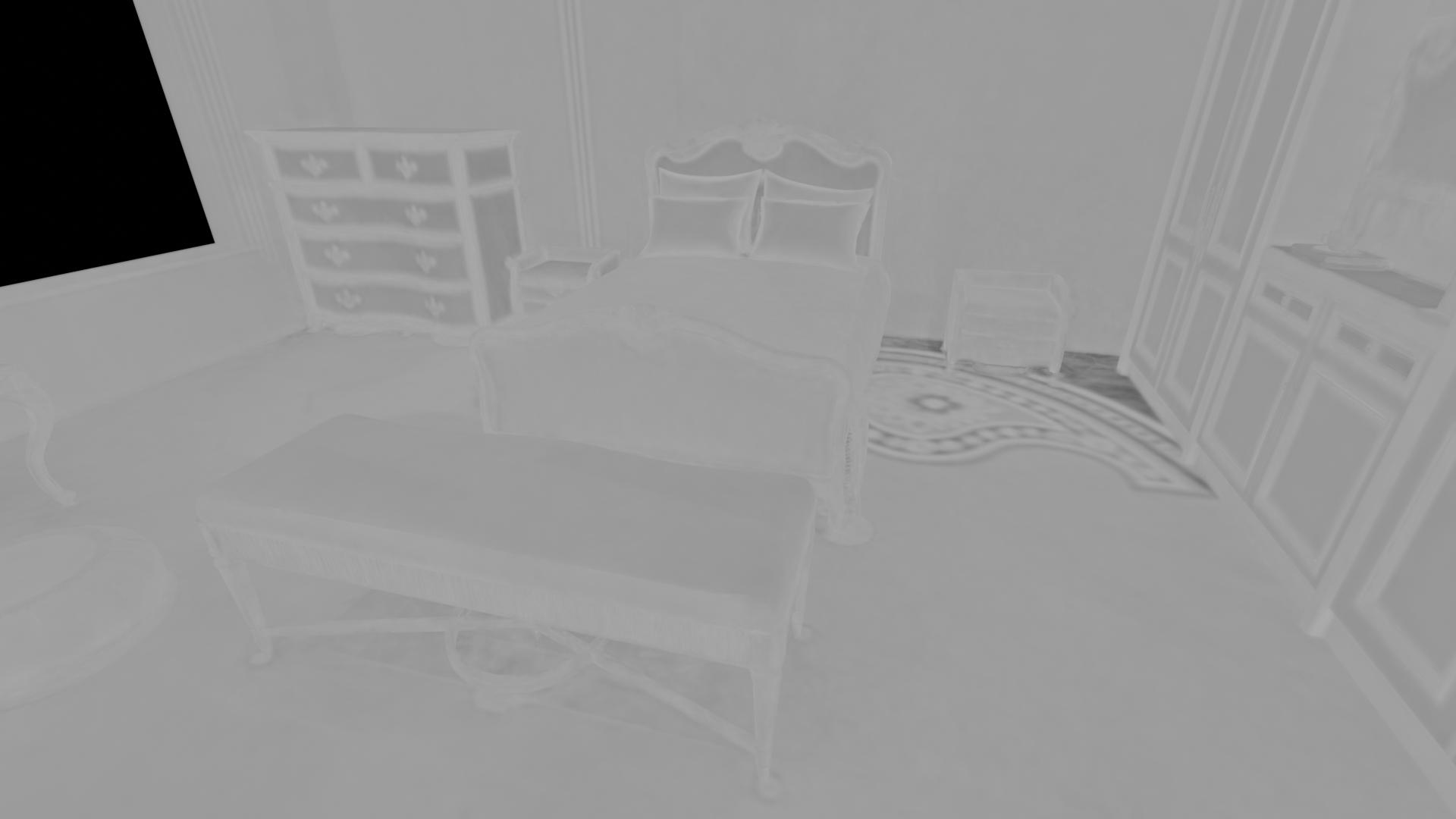}} 
        &
        \fbox{\includegraphics[width=0.15\textwidth,trim={10cm 0 10cm 0},clip]{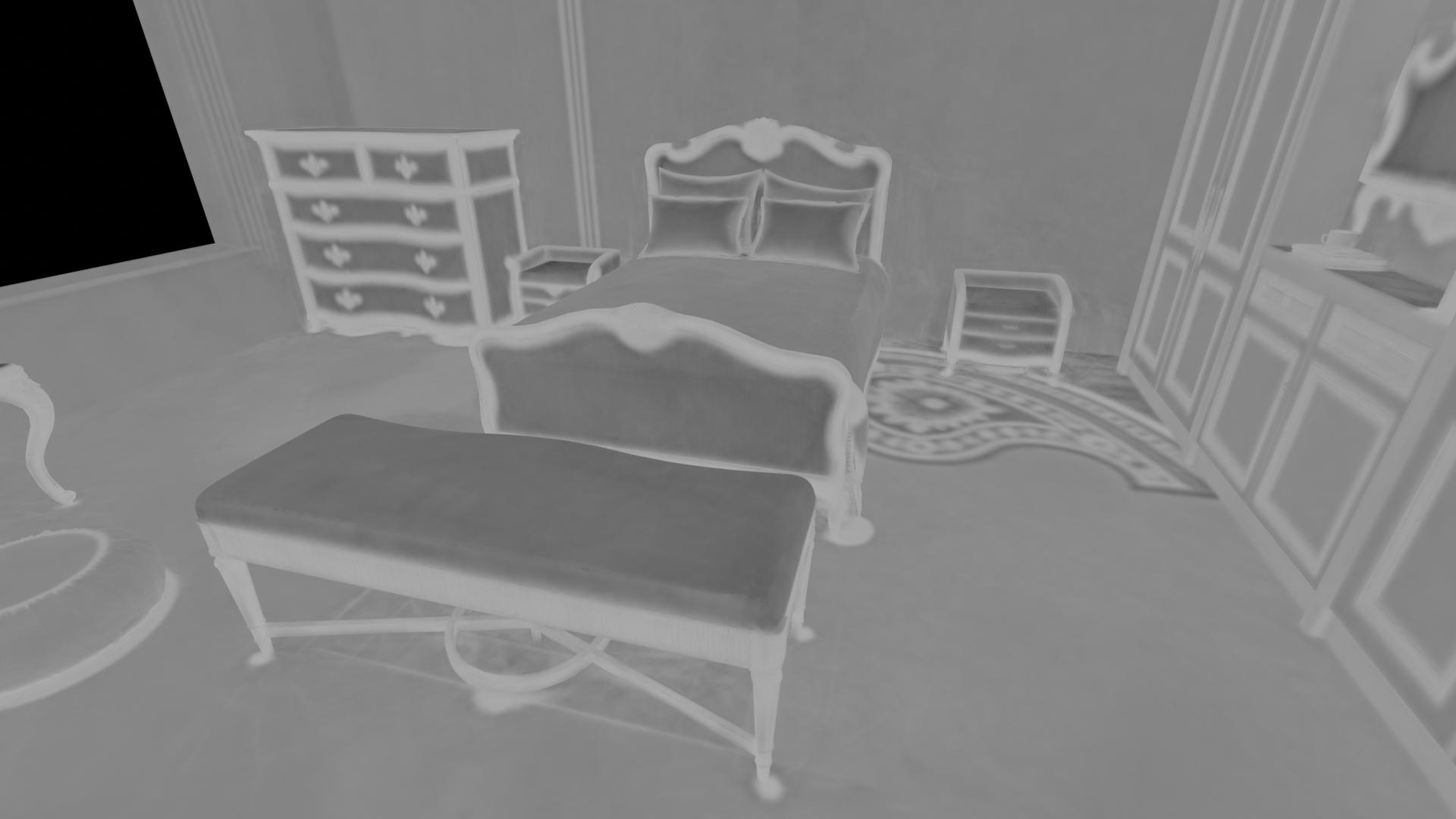}} 
        &
        \fbox{\includegraphics[width=0.15\textwidth,trim={10cm 0 10cm 0},clip]{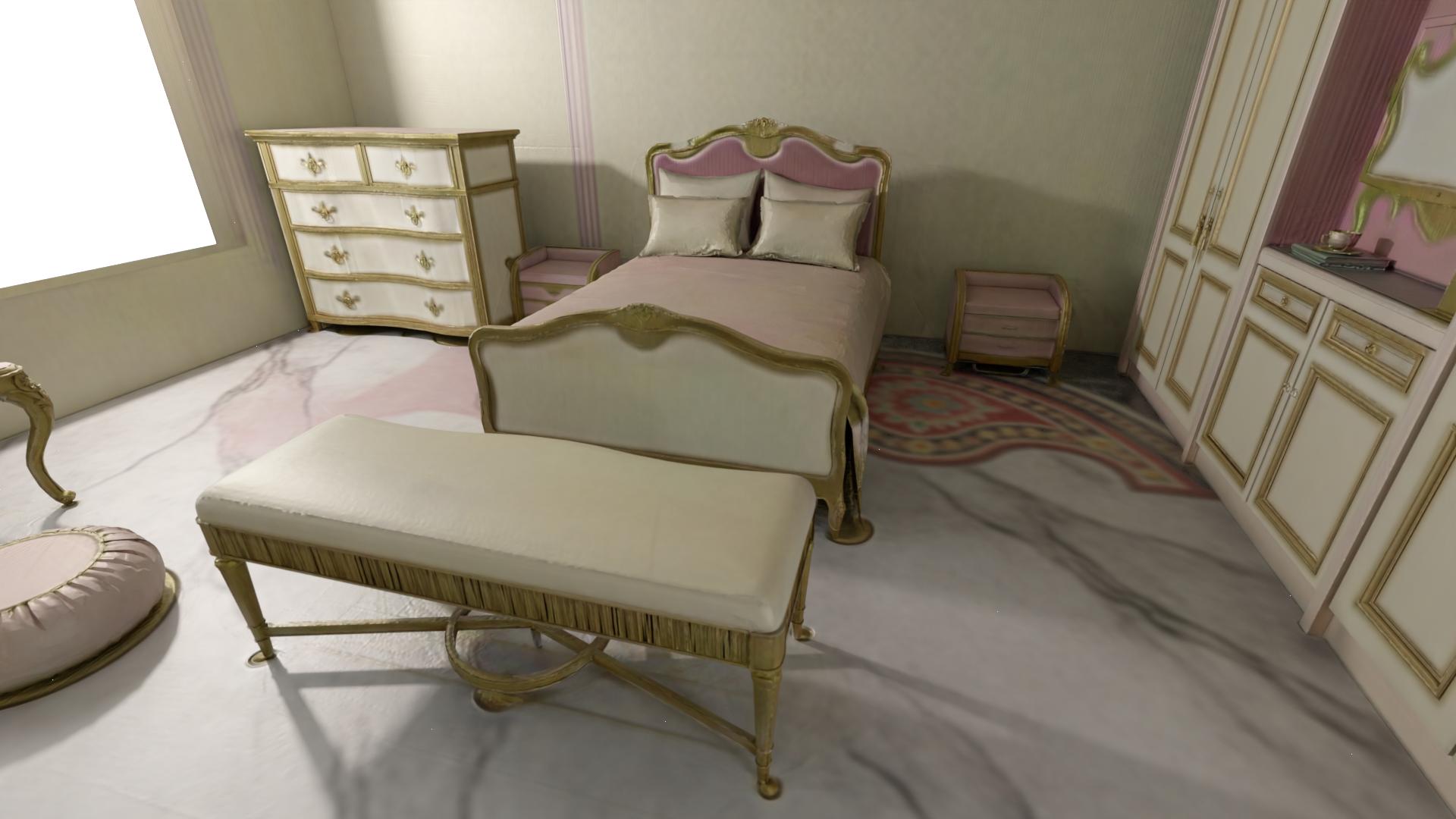}} 
        \\
        
        
        \rotatebox{90}{{\footnotesize View 3}}
        &
        \fbox{\includegraphics[width=0.15\textwidth,trim={10cm 0 10cm 0},clip]{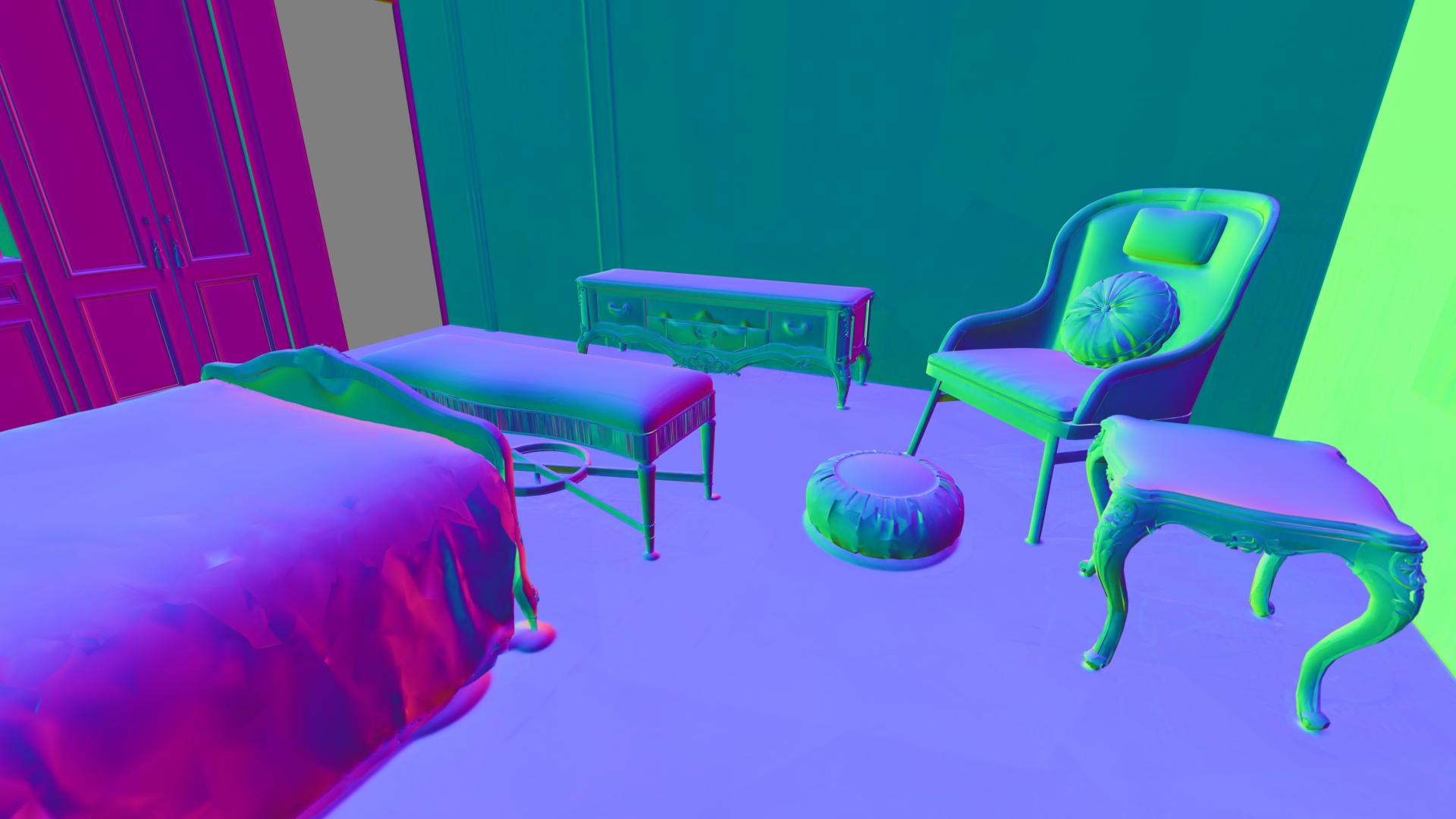}} 
        &
        \fbox{\includegraphics[width=0.15\textwidth,trim={10cm 0 10cm 0},clip]{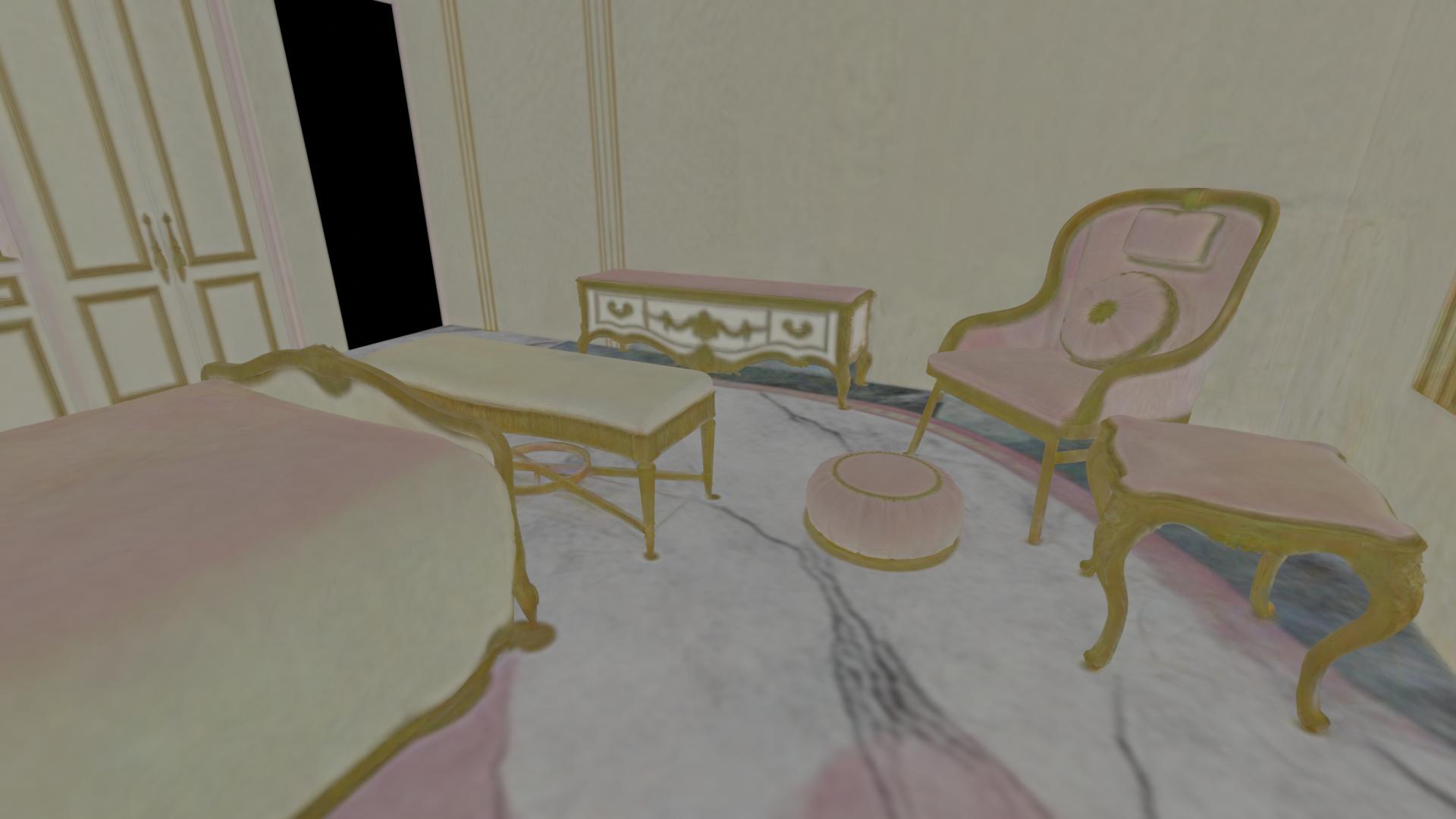}} 
        &
        \fbox{\includegraphics[width=0.15\textwidth,trim={10cm 0 10cm 0},clip]{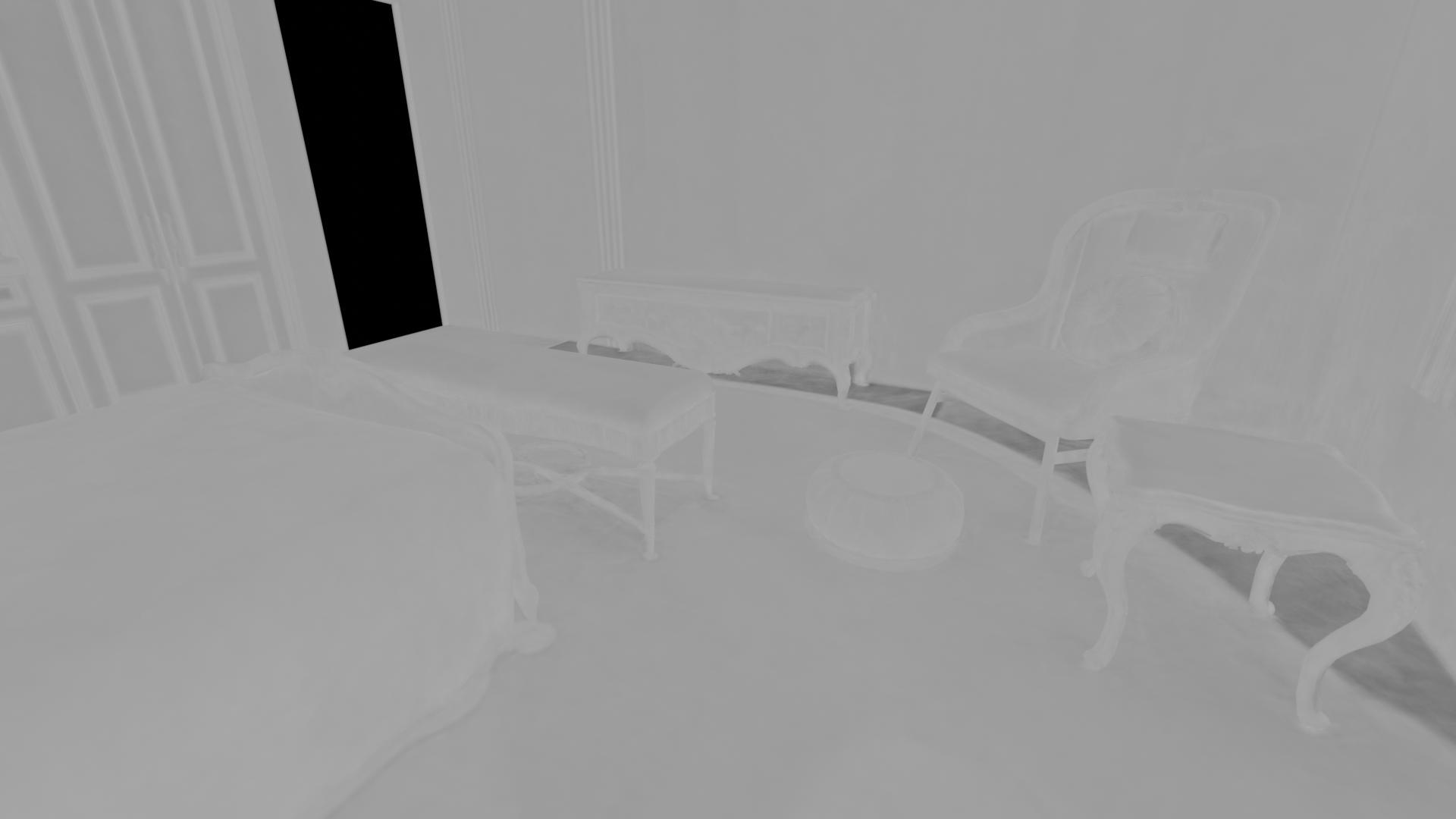}} 
        &
        \fbox{\includegraphics[width=0.15\textwidth,trim={10cm 0 10cm 0},clip]{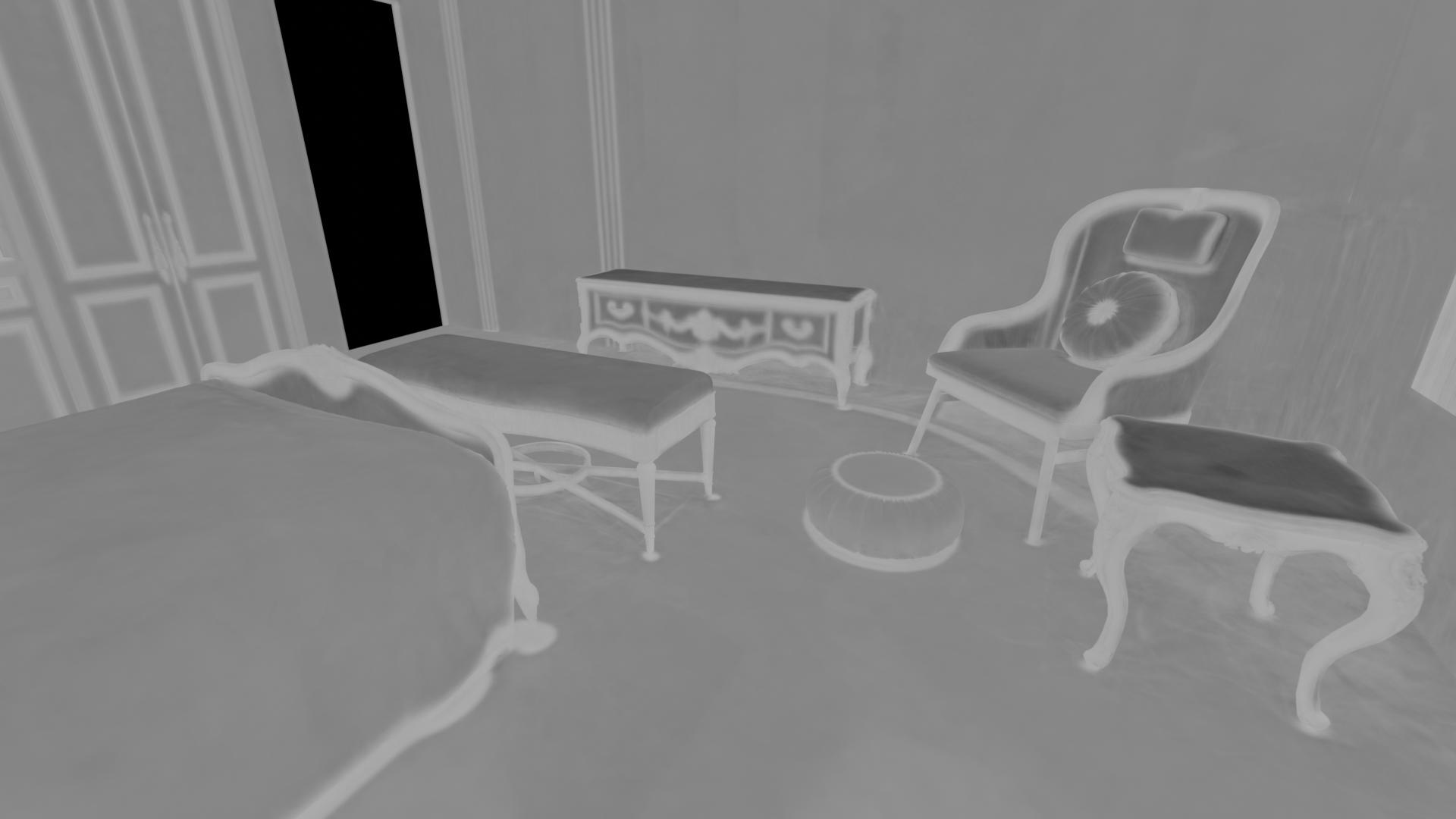}} 
        &
        \fbox{\includegraphics[width=0.15\textwidth,trim={10cm 0 10cm 0},clip]{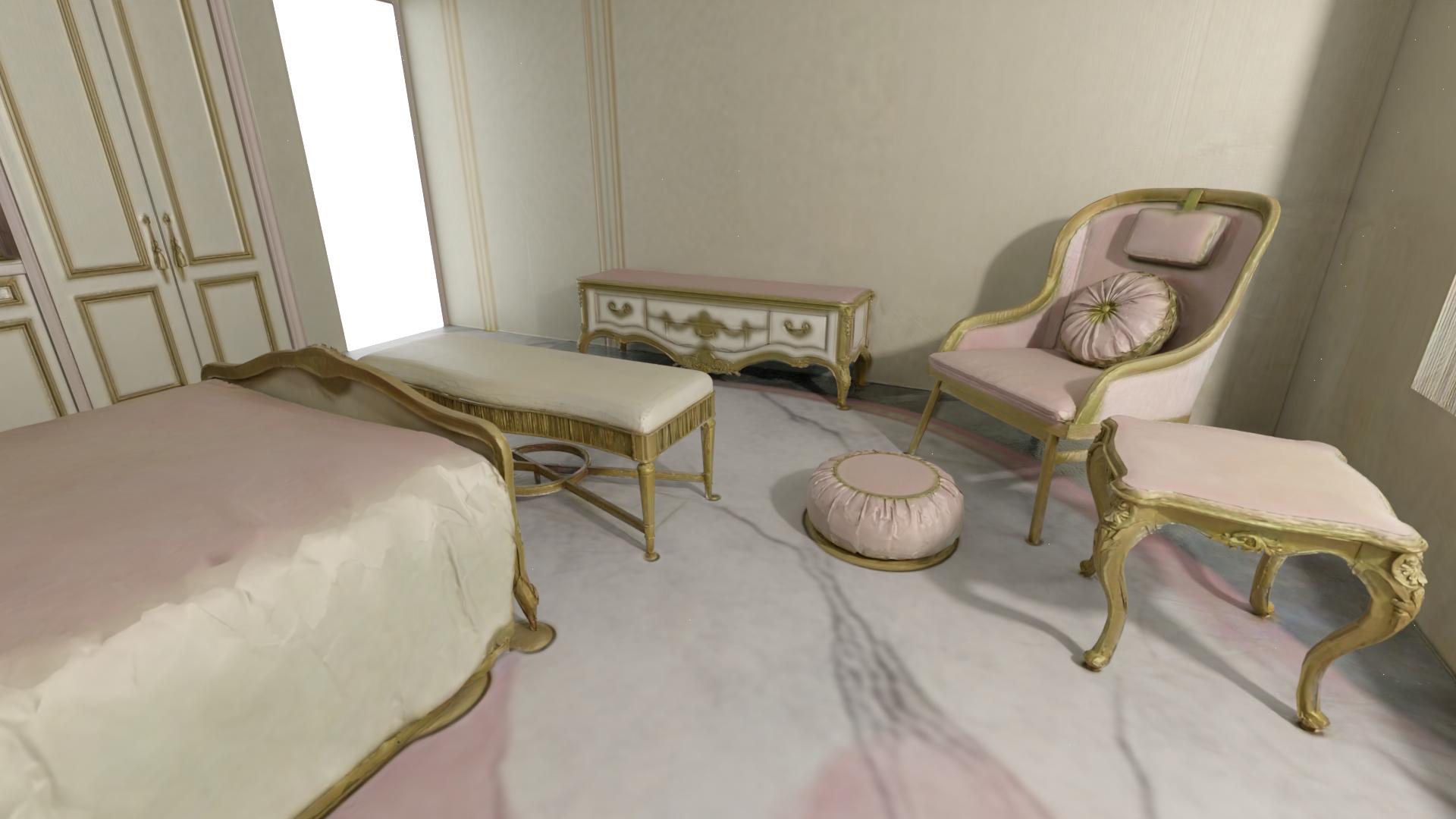}} 
        \\

        &
        {\footnotesize Normal} &
        {\footnotesize Albedo} &
        {\footnotesize Roughness} &
        {\footnotesize Metallic} &
        {\footnotesize Rendering} \\

        \midrule
        
        \rotatebox{90}{{\footnotesize View 1}}
        &
        \fbox{\includegraphics[width=0.15\textwidth,trim={10cm 0 10cm 0},clip]{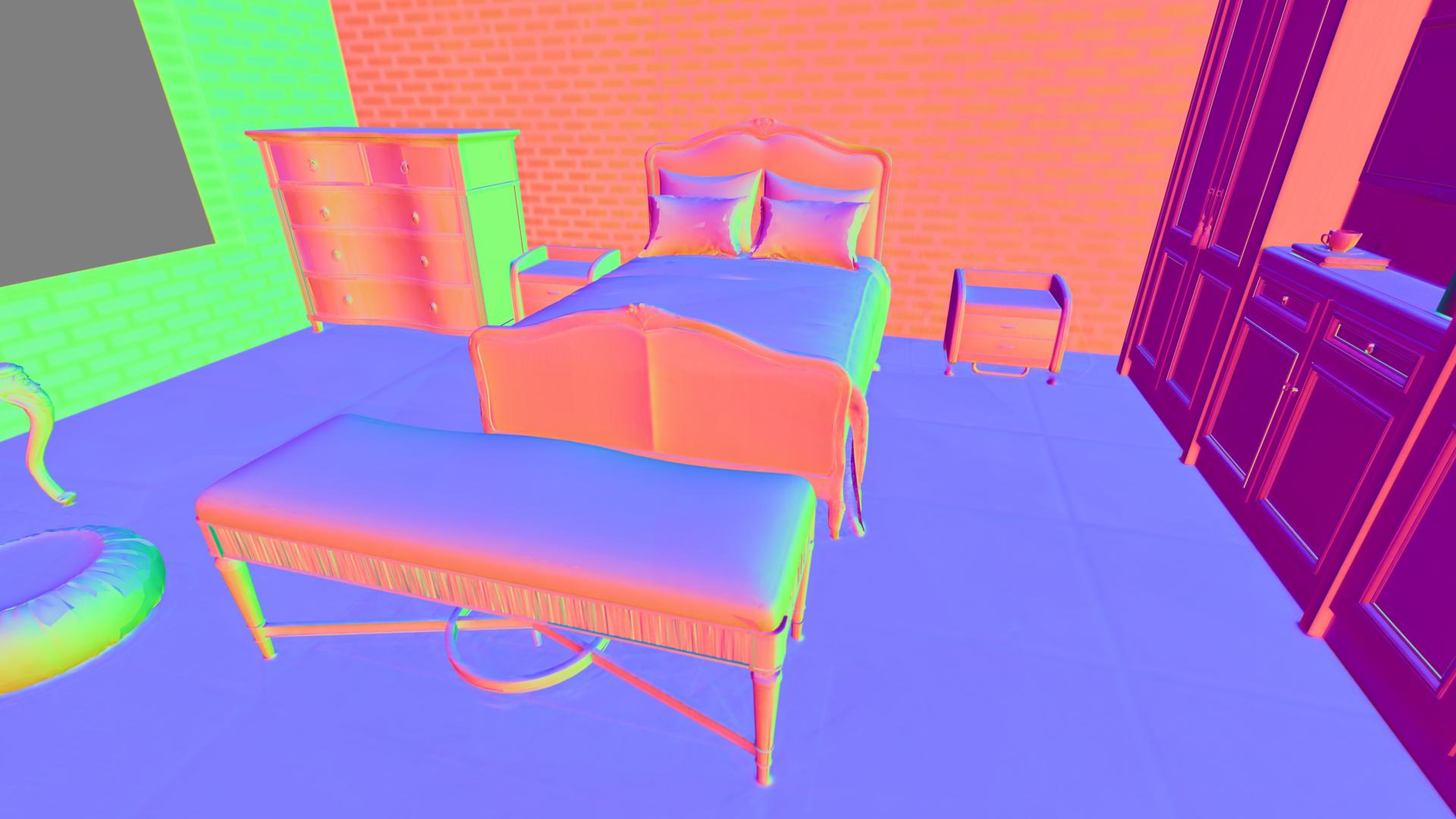}}
        &
        \fbox{\includegraphics[width=0.15\textwidth,trim={10cm 0 10cm 0},clip]{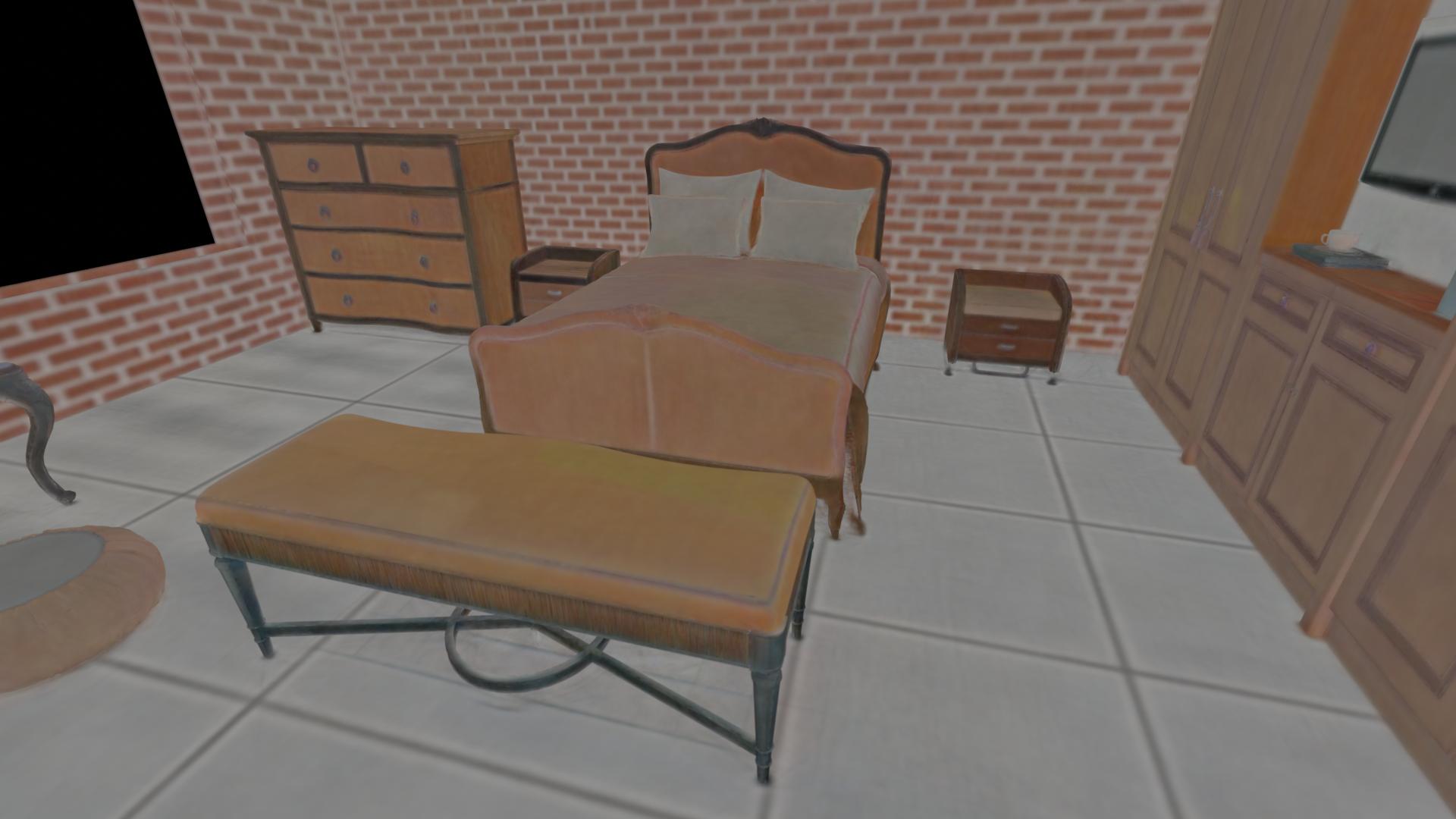}} 
        &
        \fbox{\includegraphics[width=0.15\textwidth,trim={10cm 0 10cm 0},clip]{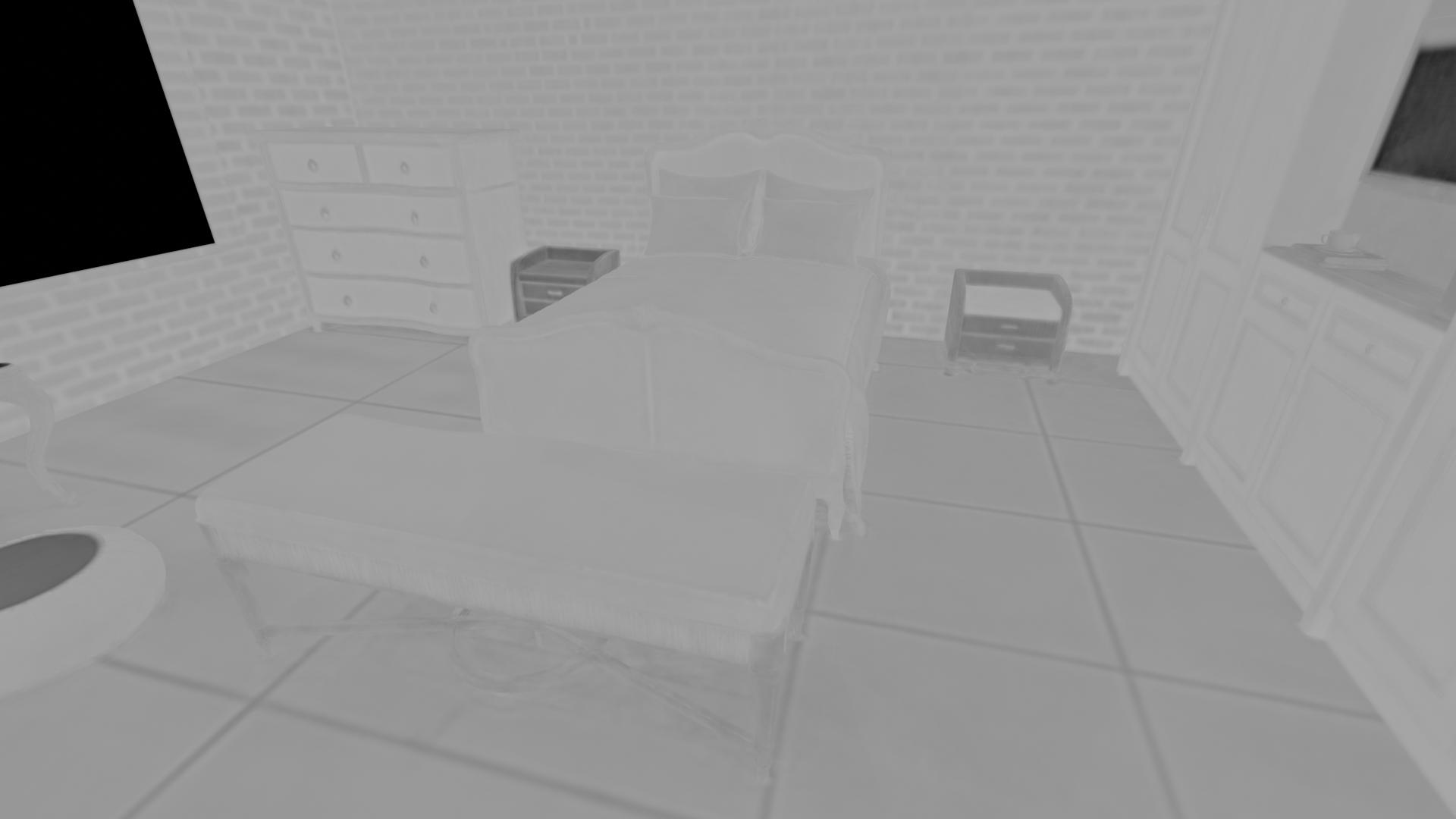}} 
        &
        \fbox{\includegraphics[width=0.15\textwidth,trim={10cm 0 10cm 0},clip]{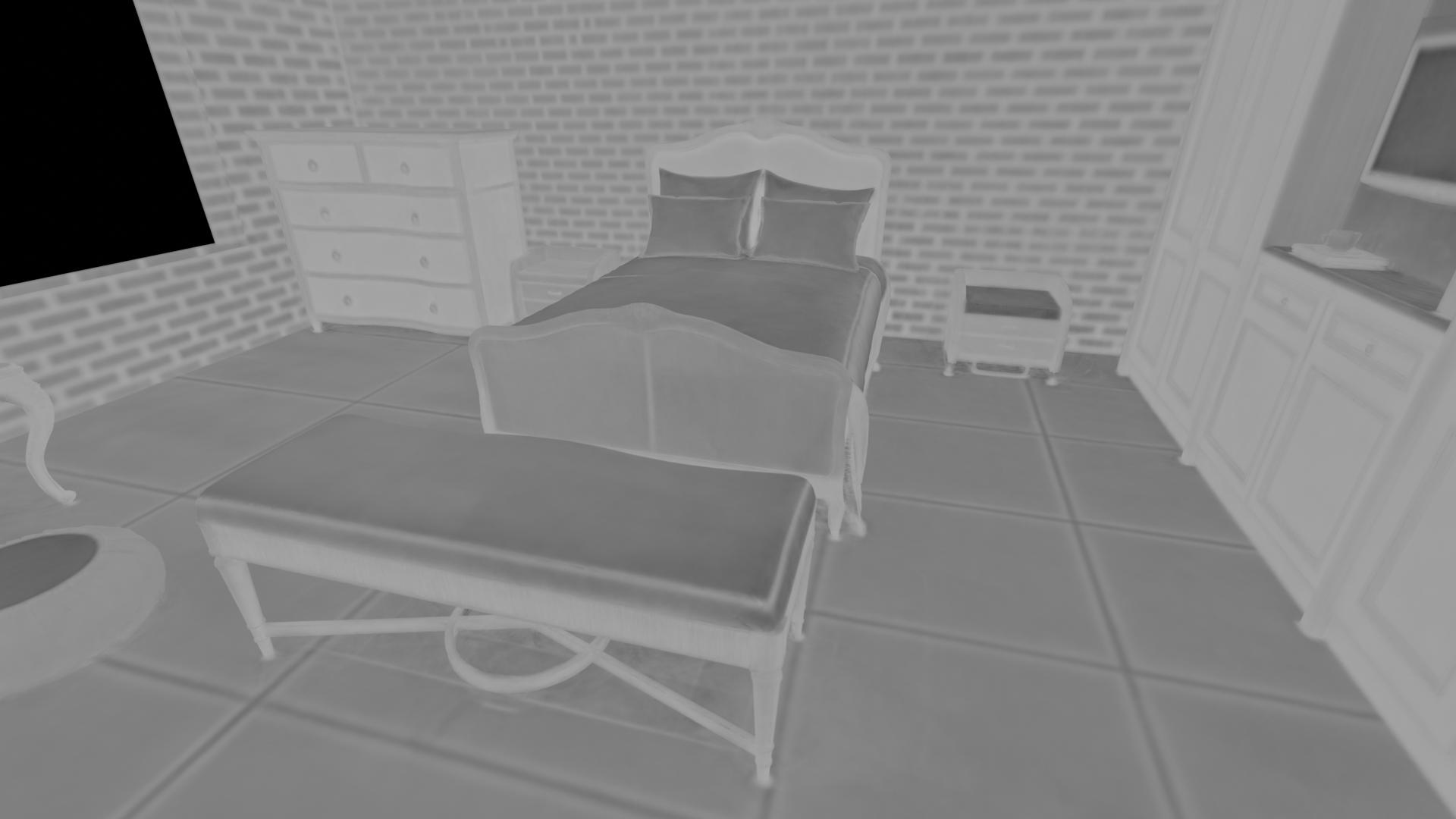}} 
        &
        \fbox{\includegraphics[width=0.15\textwidth,trim={10cm 0 10cm 0},clip]{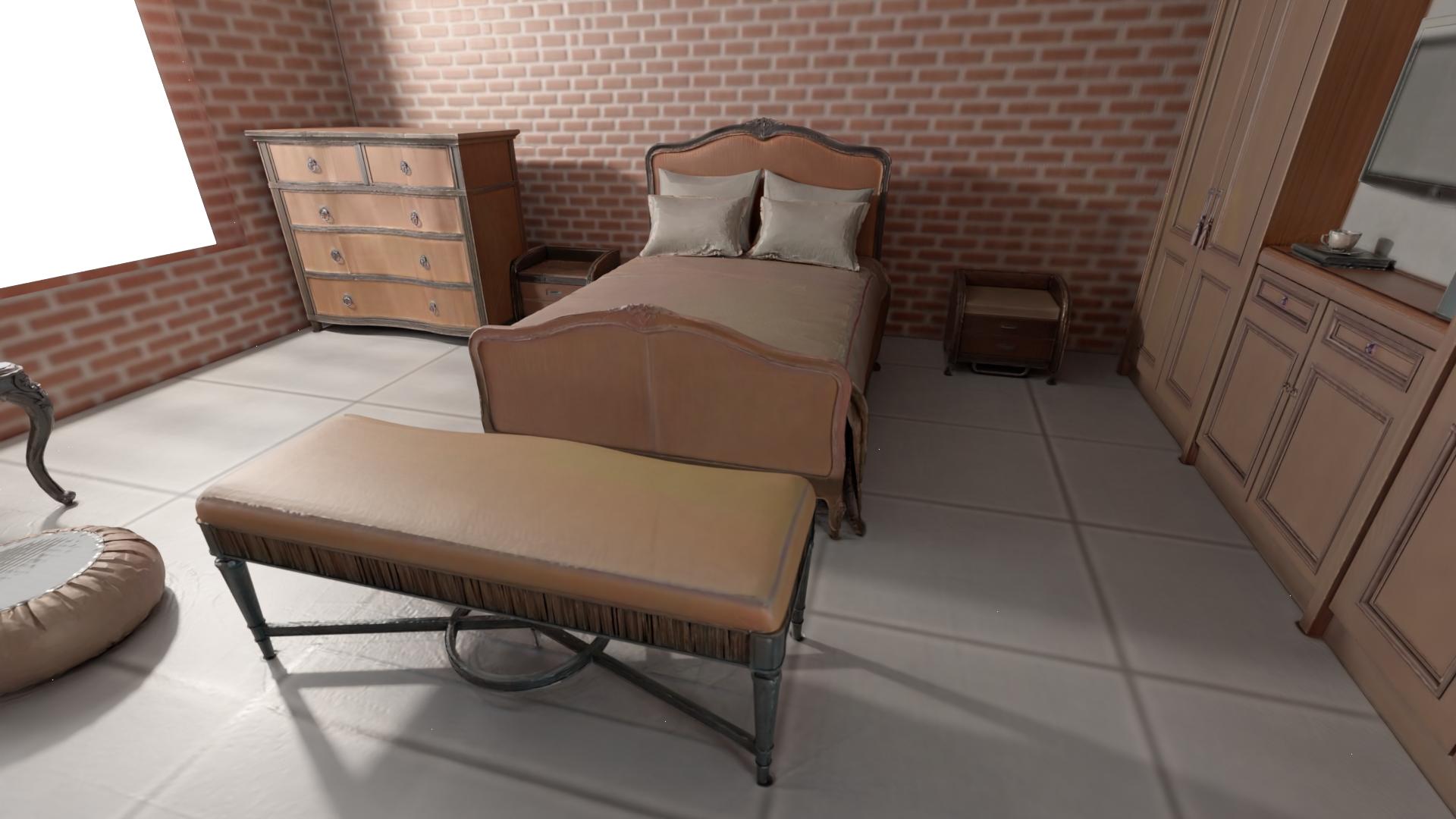}} 
        \\
        
        
        \rotatebox{90}{{\footnotesize View 3}}
        &
        \fbox{\includegraphics[width=0.15\textwidth,trim={10cm 0 10cm 0},clip]{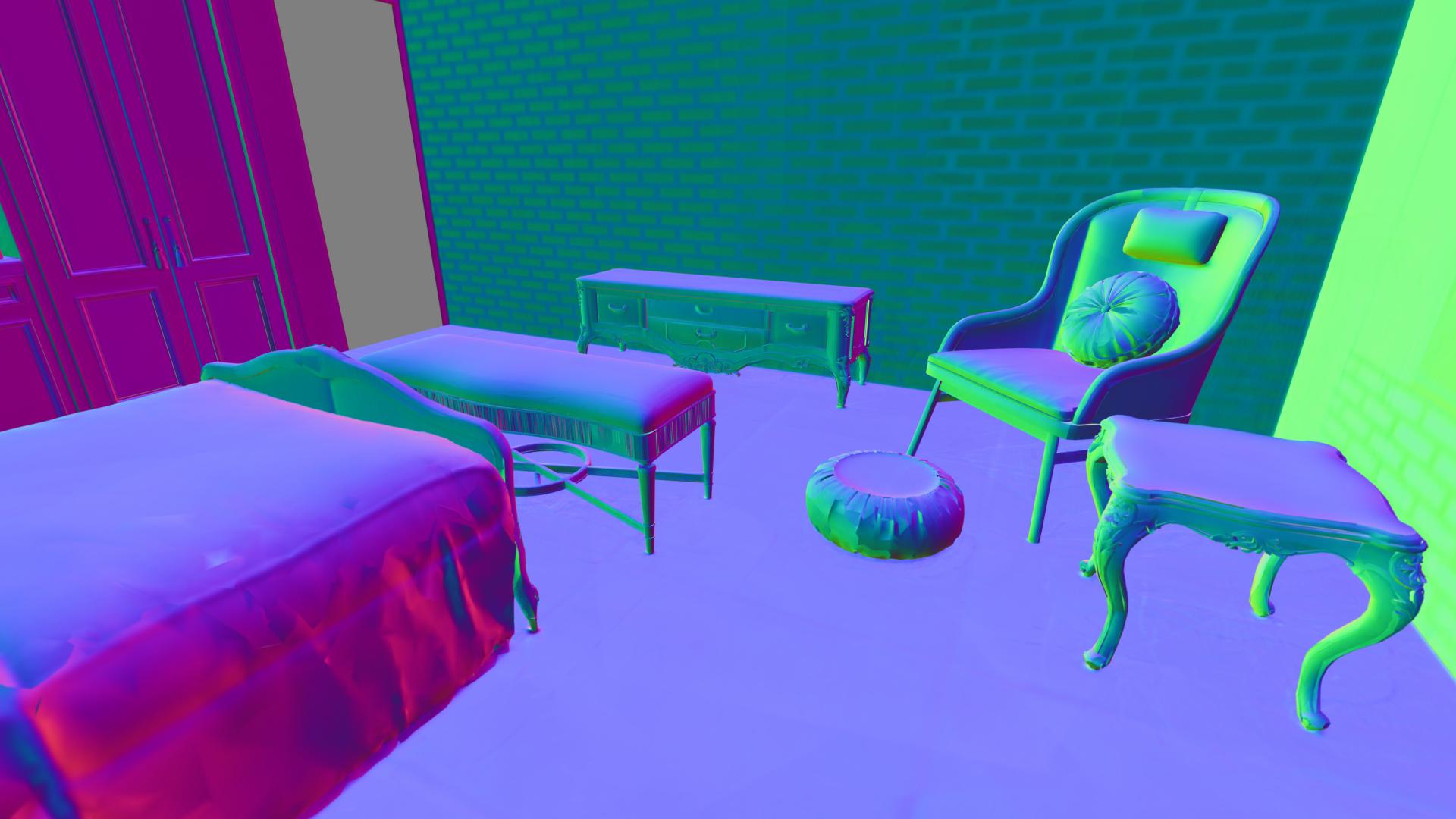}} 
        &
        \fbox{\includegraphics[width=0.15\textwidth,trim={10cm 0 10cm 0},clip]{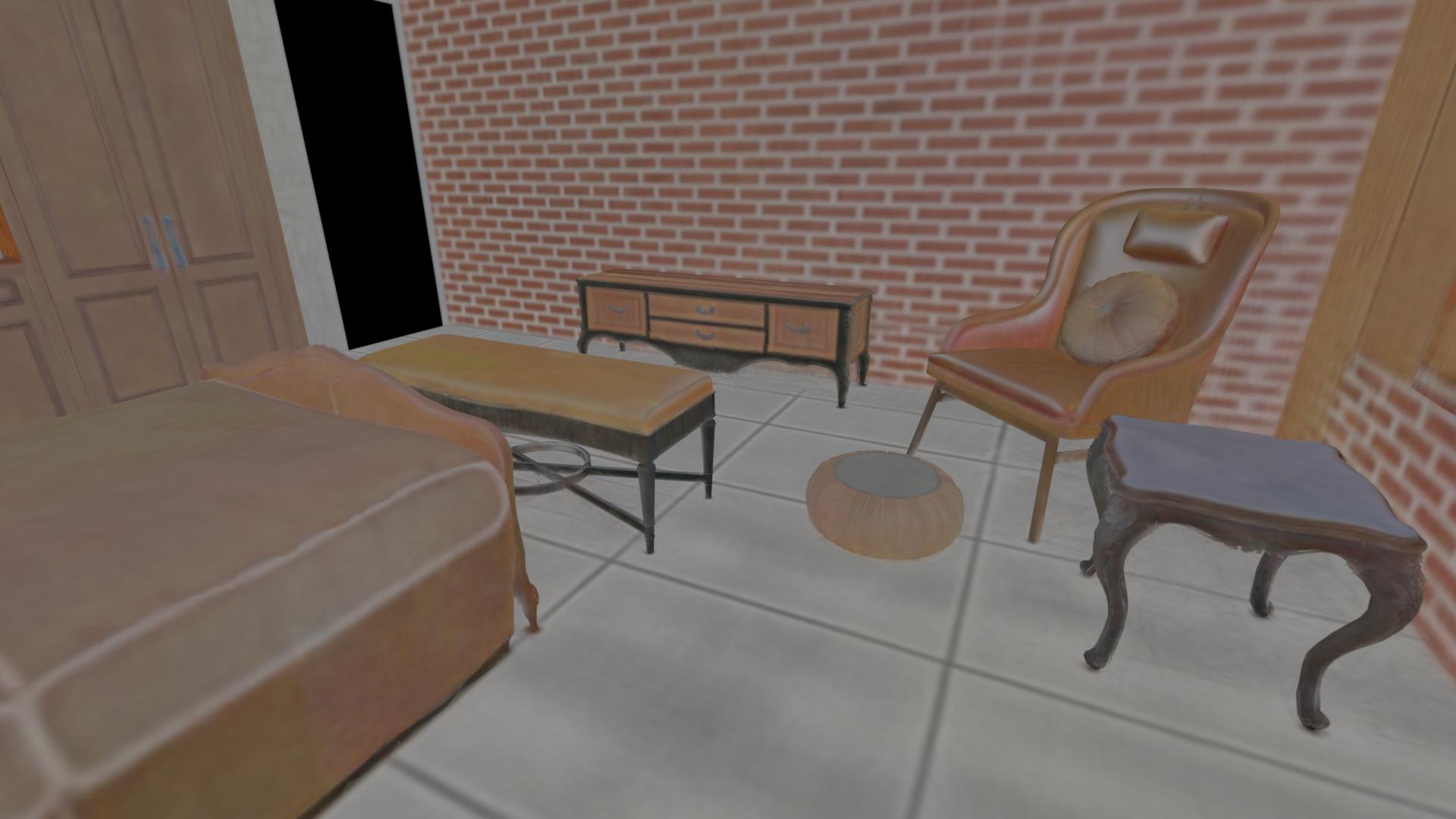}} 
        &
        \fbox{\includegraphics[width=0.15\textwidth,trim={10cm 0 10cm 0},clip]{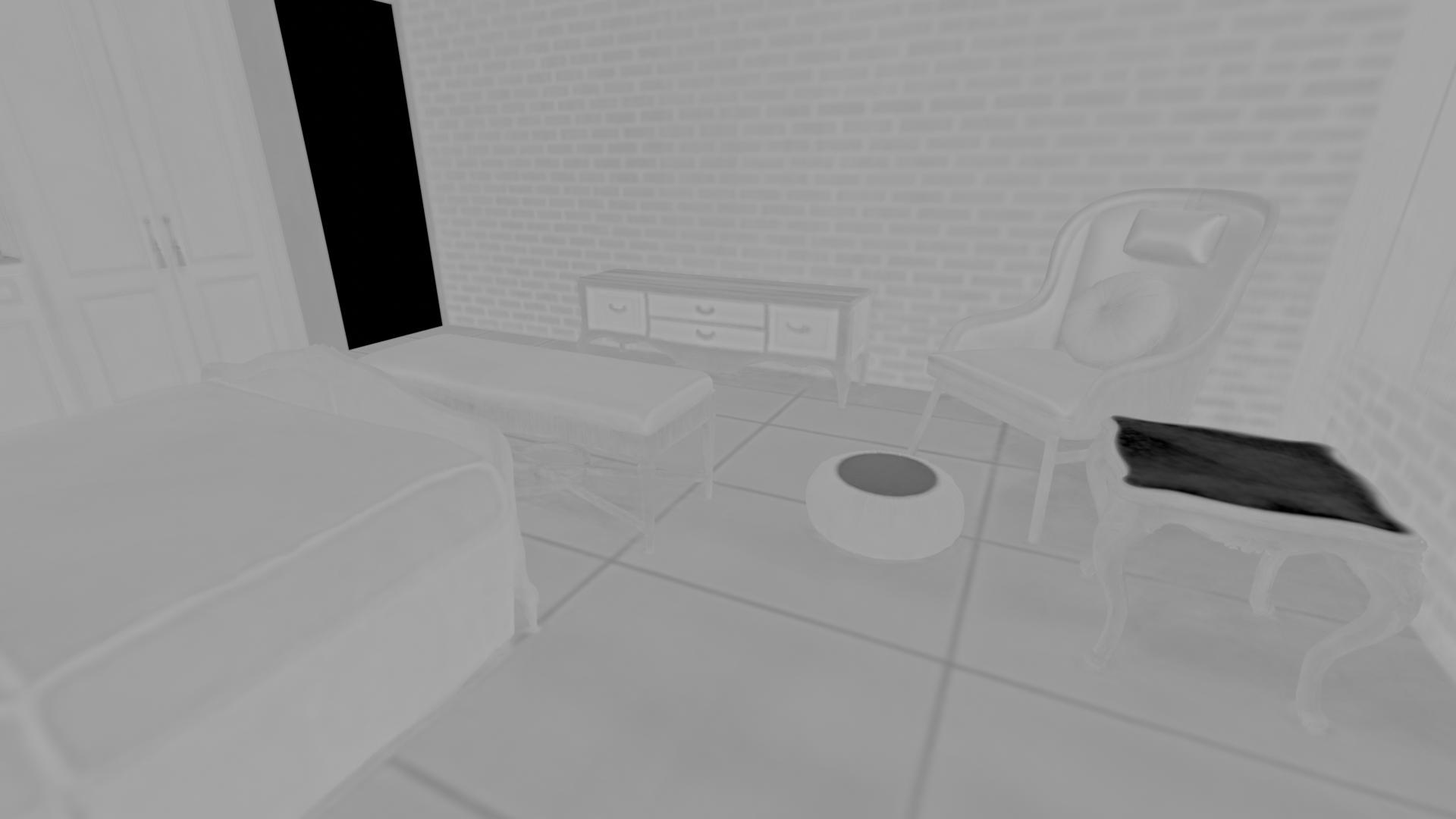}} 
        &
        \fbox{\includegraphics[width=0.15\textwidth,trim={10cm 0 10cm 0},clip]{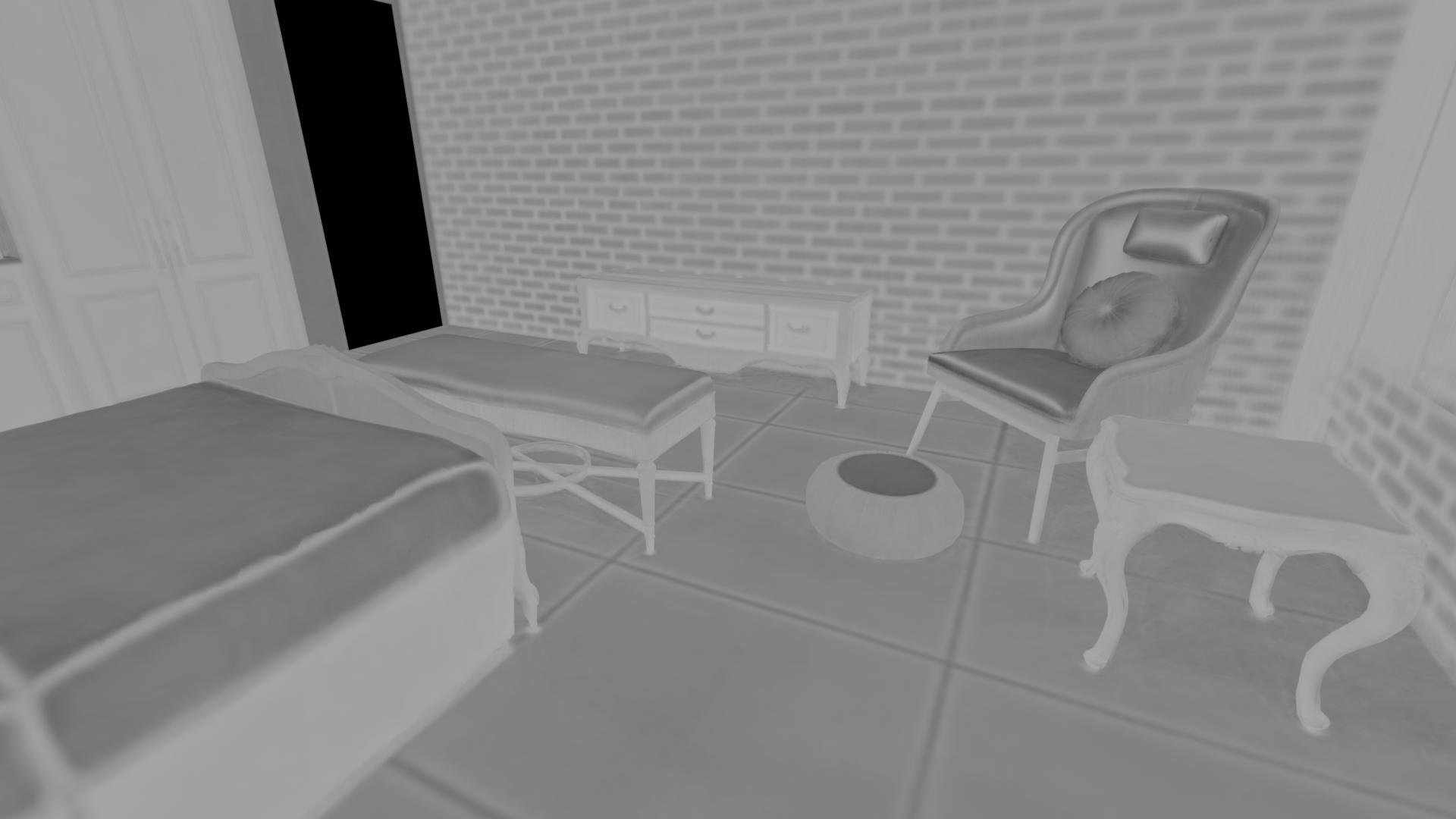}} 
        &
        \fbox{\includegraphics[width=0.15\textwidth,trim={10cm 0 10cm 0},clip]{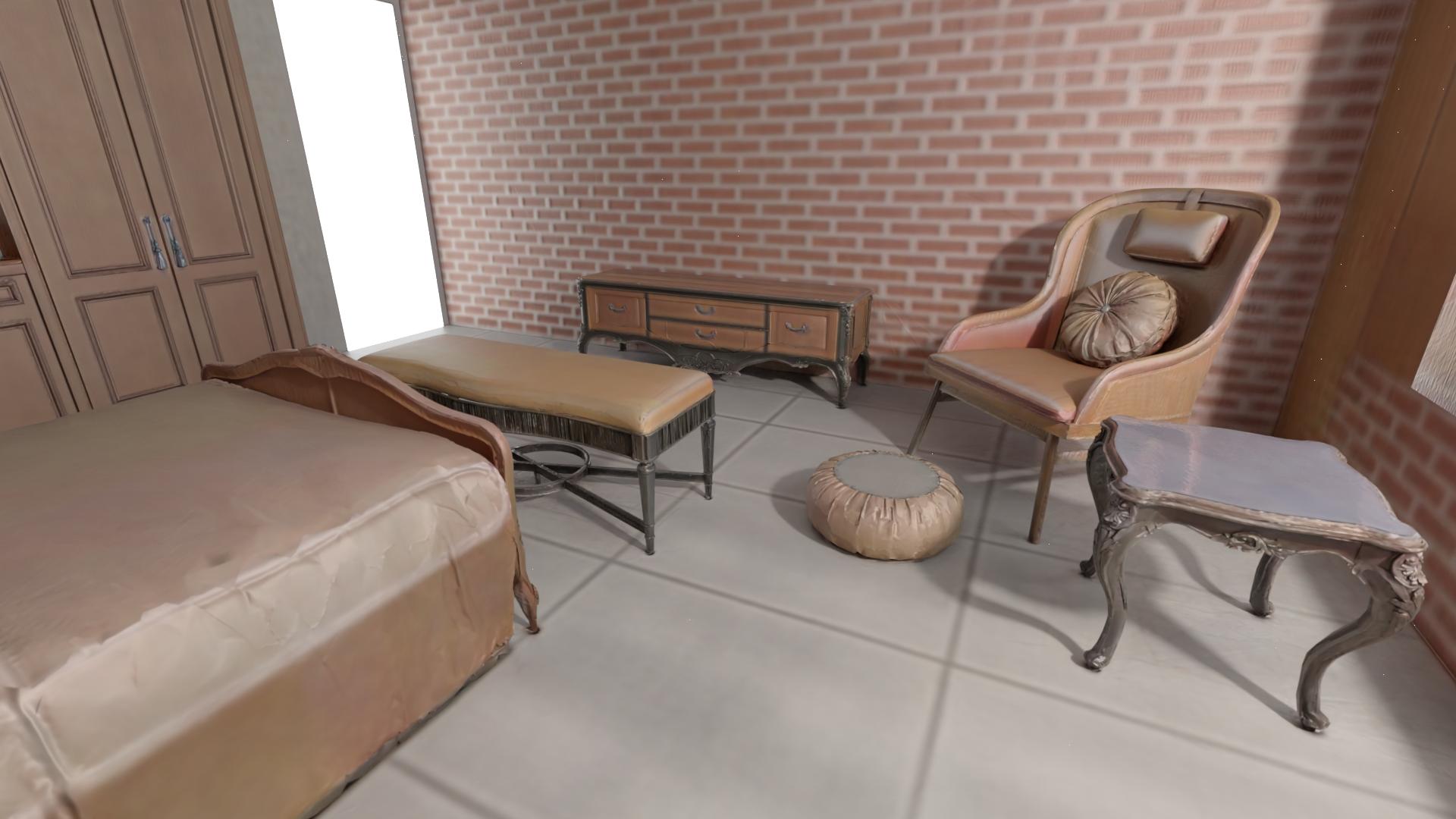}} 
        \\

        &
        {\footnotesize Normal} &
        {\footnotesize Albedo} &
        {\footnotesize Roughness} &
        {\footnotesize Metallic} &
        {\footnotesize Rendering} \\

        \midrule
        
        \rotatebox{90}{{\footnotesize View 1}}
        &
        \fbox{\includegraphics[width=0.15\textwidth,trim={10cm 0 10cm 0},clip]{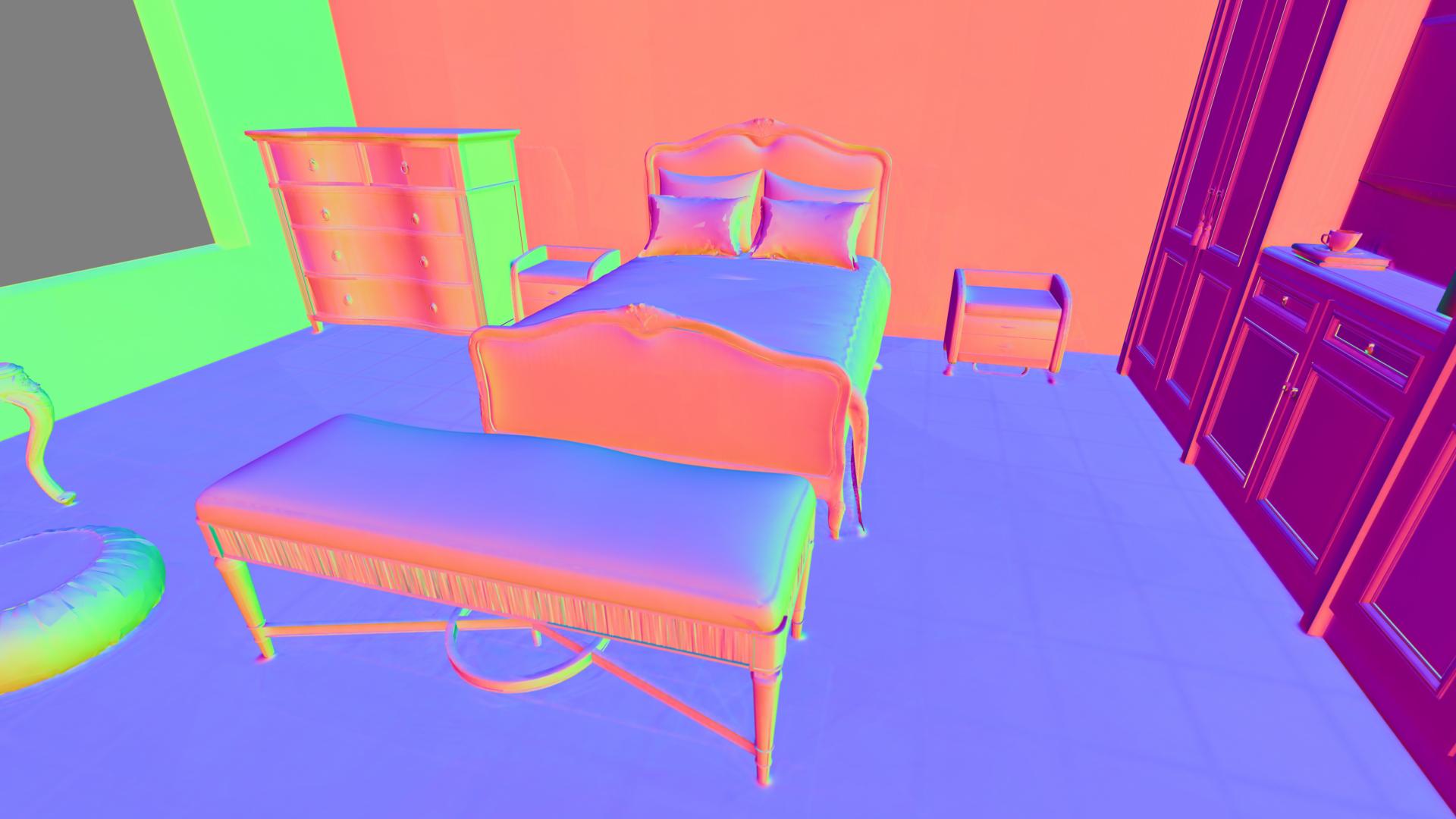}}
        &
        \fbox{\includegraphics[width=0.15\textwidth,trim={10cm 0 10cm 0},clip]{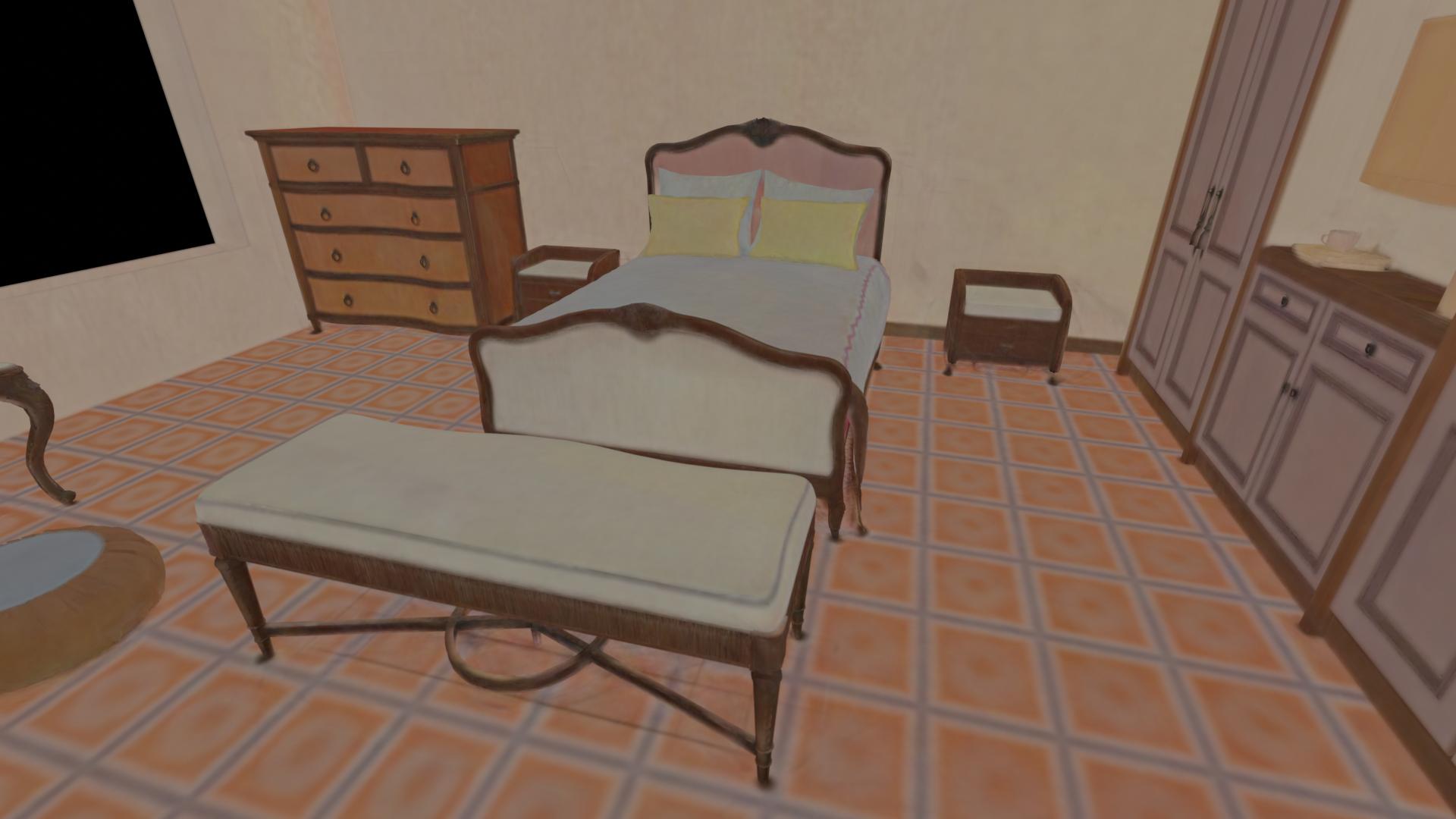}} 
        &
        \fbox{\includegraphics[width=0.15\textwidth,trim={10cm 0 10cm 0},clip]{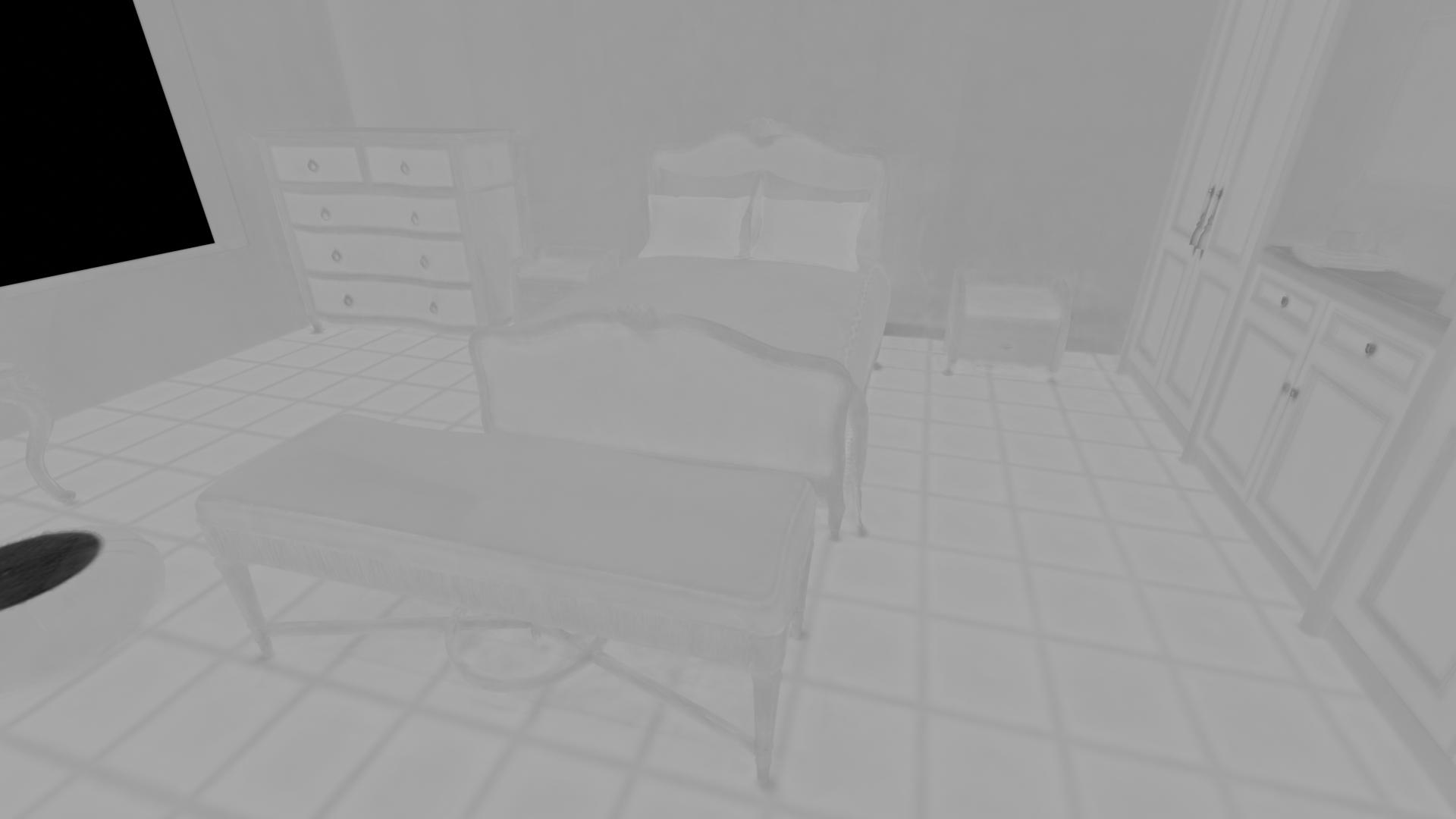}} 
        &
        \fbox{\includegraphics[width=0.15\textwidth,trim={10cm 0 10cm 0},clip]{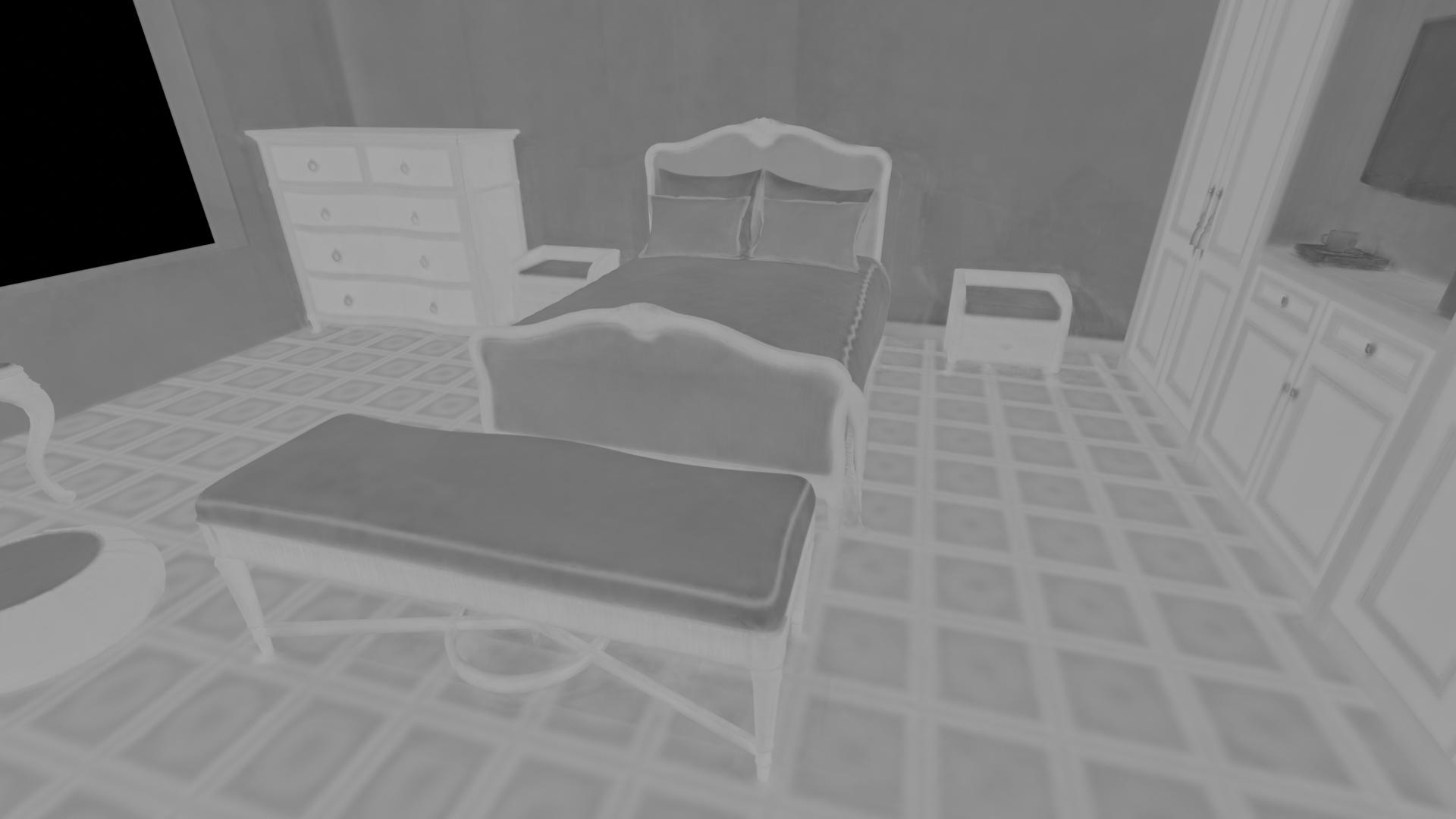}} 
        &
        \fbox{\includegraphics[width=0.15\textwidth,trim={10cm 0 10cm 0},clip]{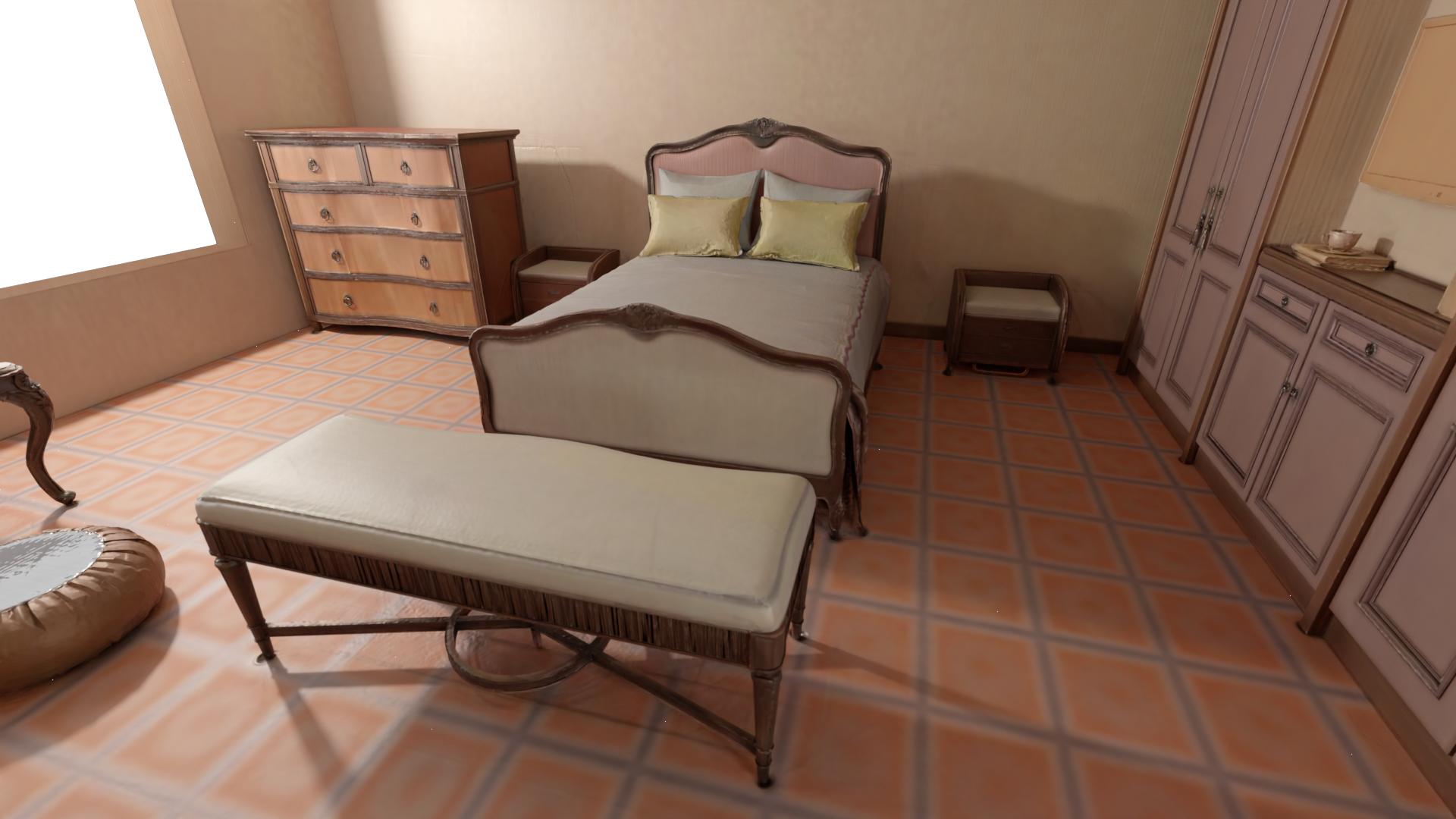}} 
        \\
        
        
        \rotatebox{90}{{\footnotesize View 3}}
        &
        \fbox{\includegraphics[width=0.15\textwidth,trim={10cm 0 10cm 0},clip]{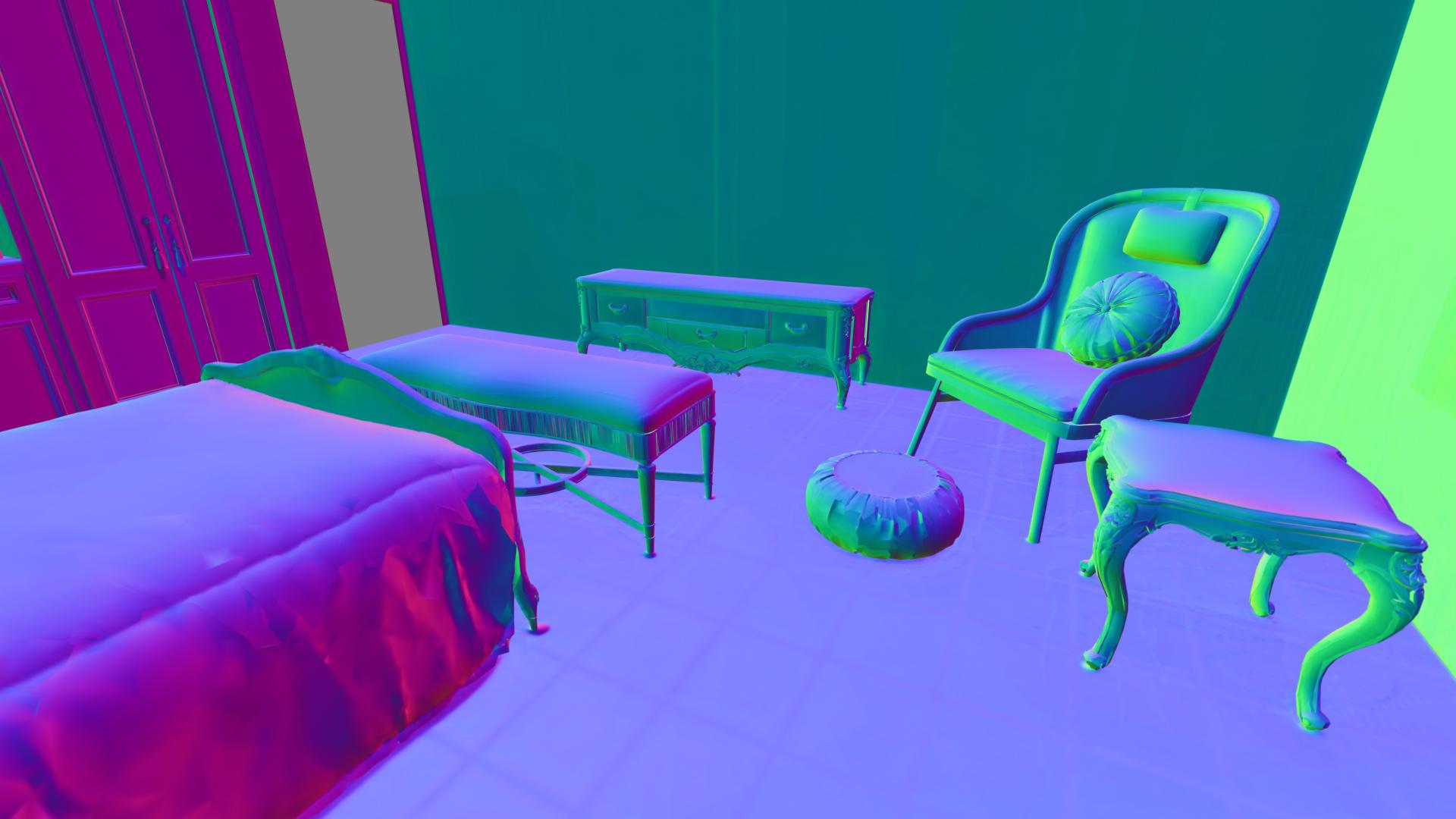}} 
        &
        \fbox{\includegraphics[width=0.15\textwidth,trim={10cm 0 10cm 0},clip]{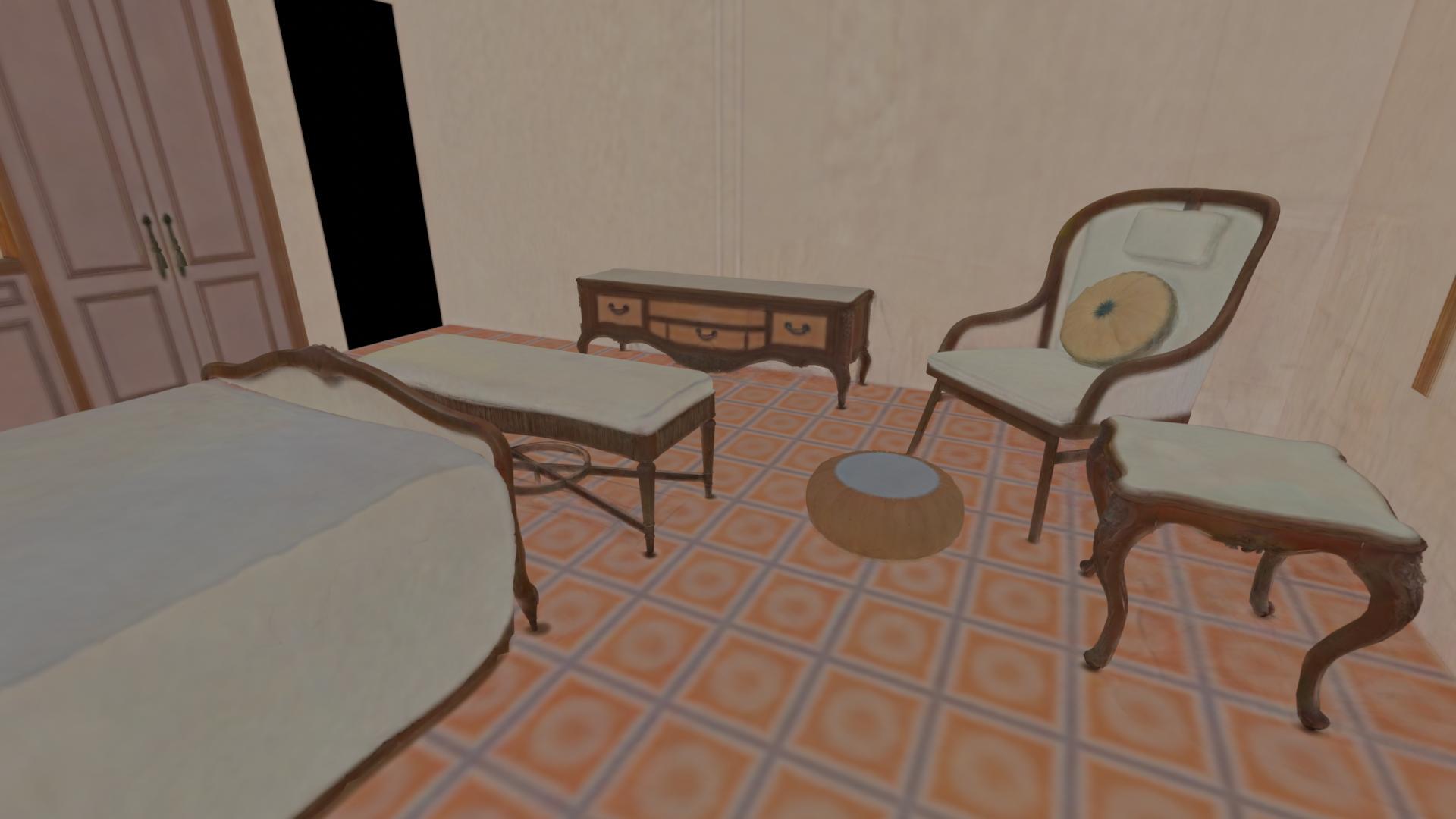}} 
        &
        \fbox{\includegraphics[width=0.15\textwidth,trim={10cm 0 10cm 0},clip]{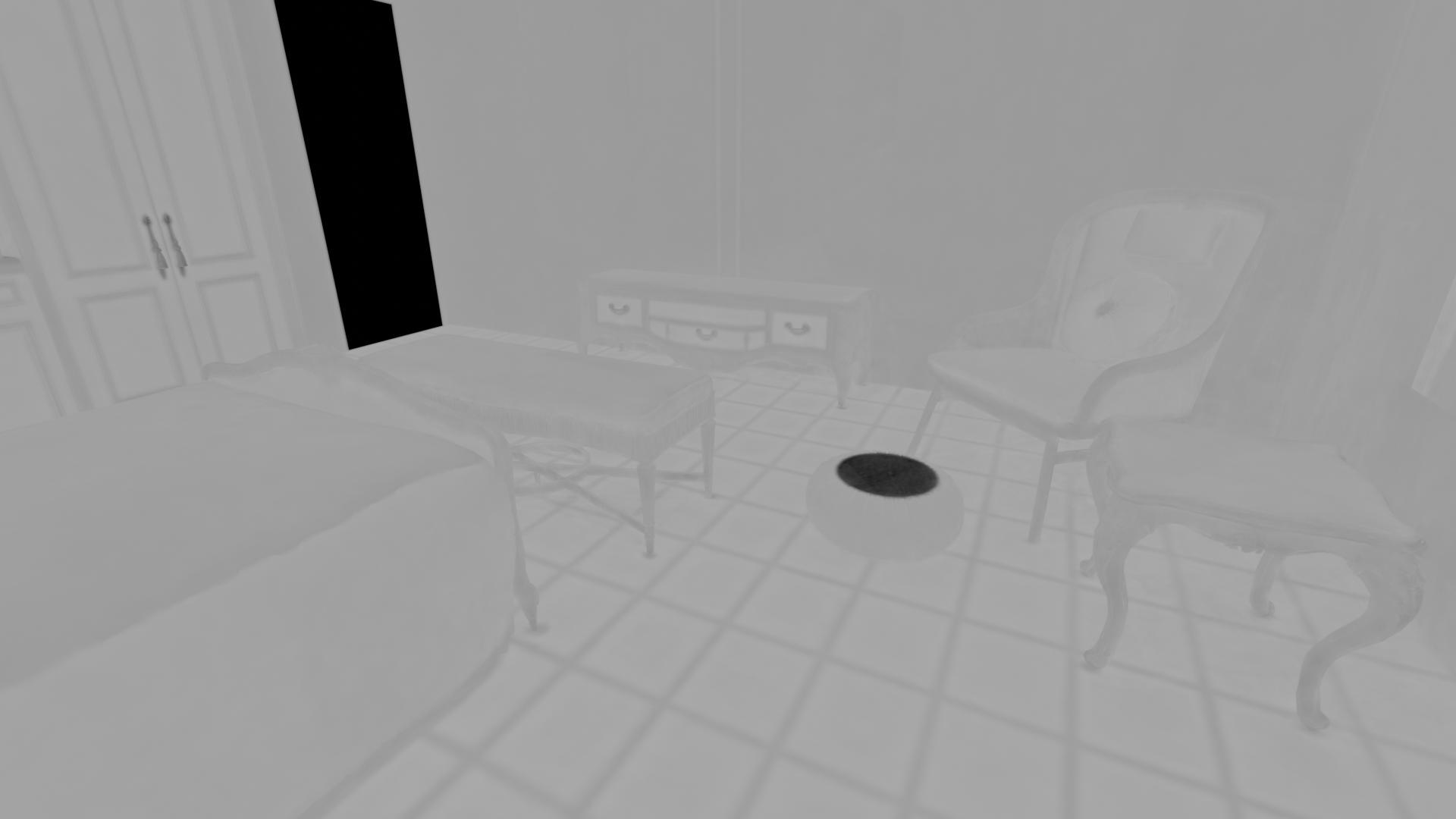}} 
        &
        \fbox{\includegraphics[width=0.15\textwidth,trim={10cm 0 10cm 0},clip]{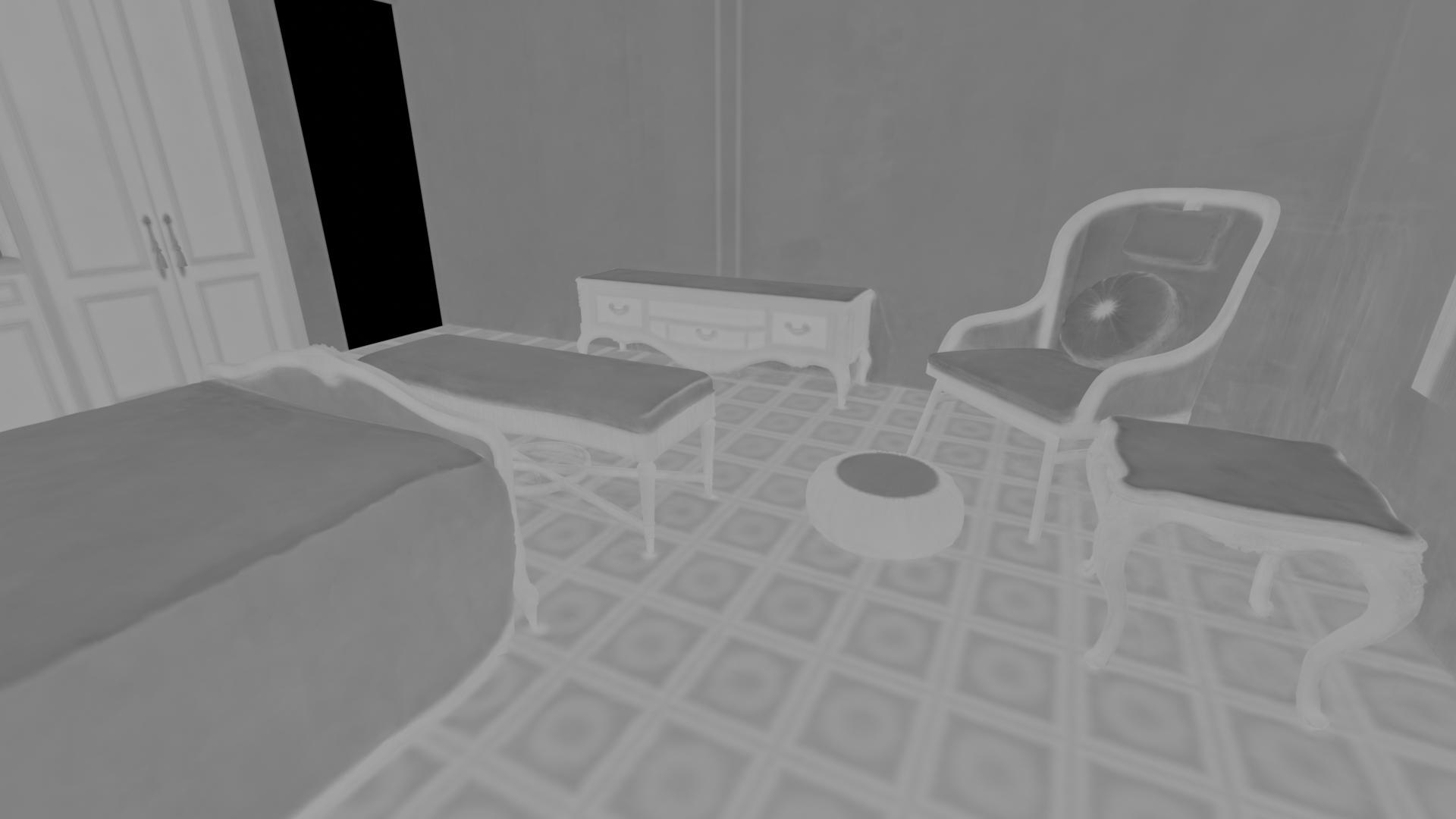}} 
        &
        \fbox{\includegraphics[width=0.15\textwidth,trim={10cm 0 10cm 0},clip]{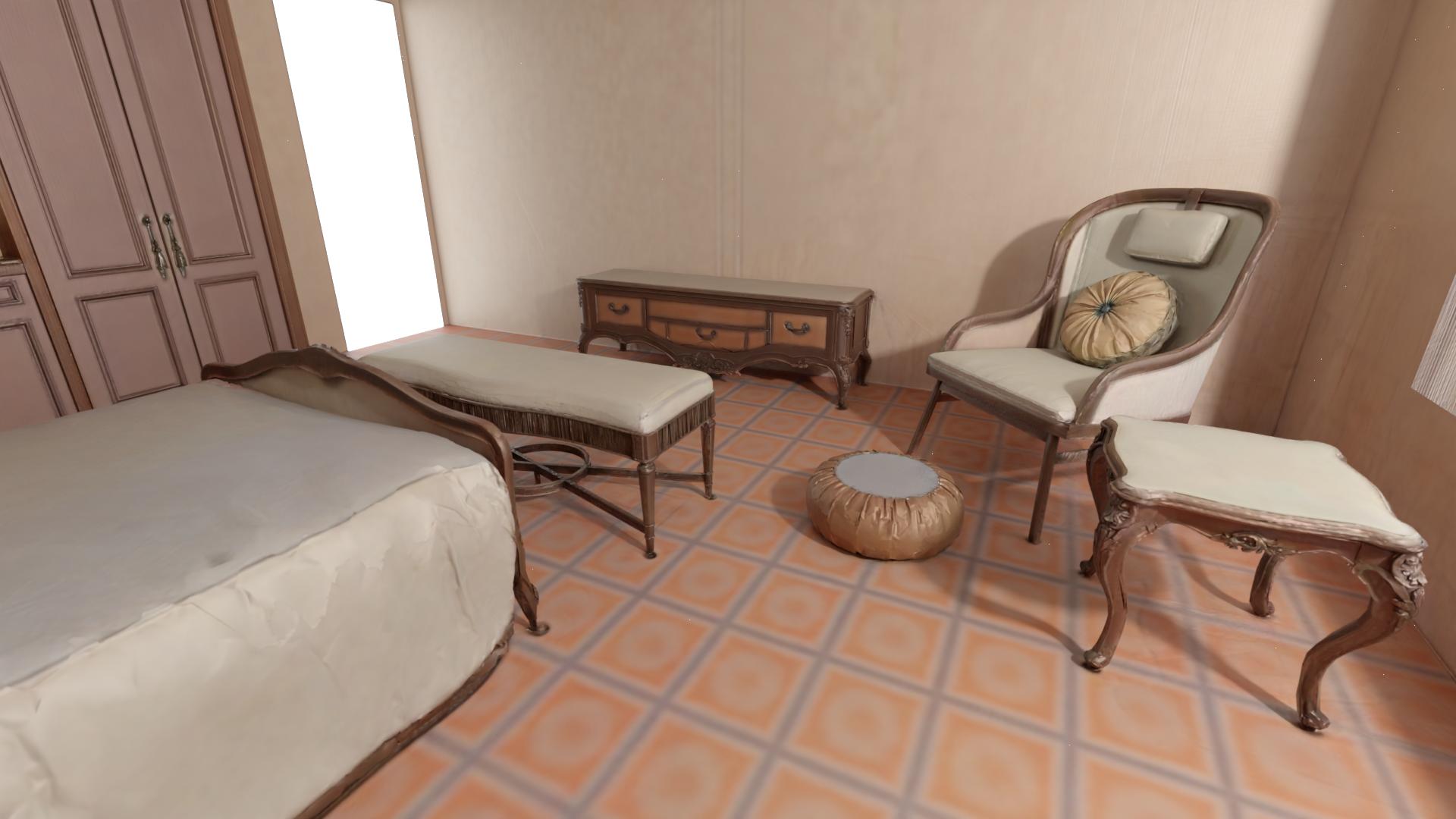}} 
        \\

        &
        {\footnotesize Normal} &
        {\footnotesize Albedo} &
        {\footnotesize Roughness} &
        {\footnotesize Metallic} &
        {\footnotesize Rendering} \\

        \midrule
        
        \rotatebox{90}{{\footnotesize View 1}}
        &
        \fbox{\includegraphics[width=0.15\textwidth,trim={10cm 0 10cm 0},clip]{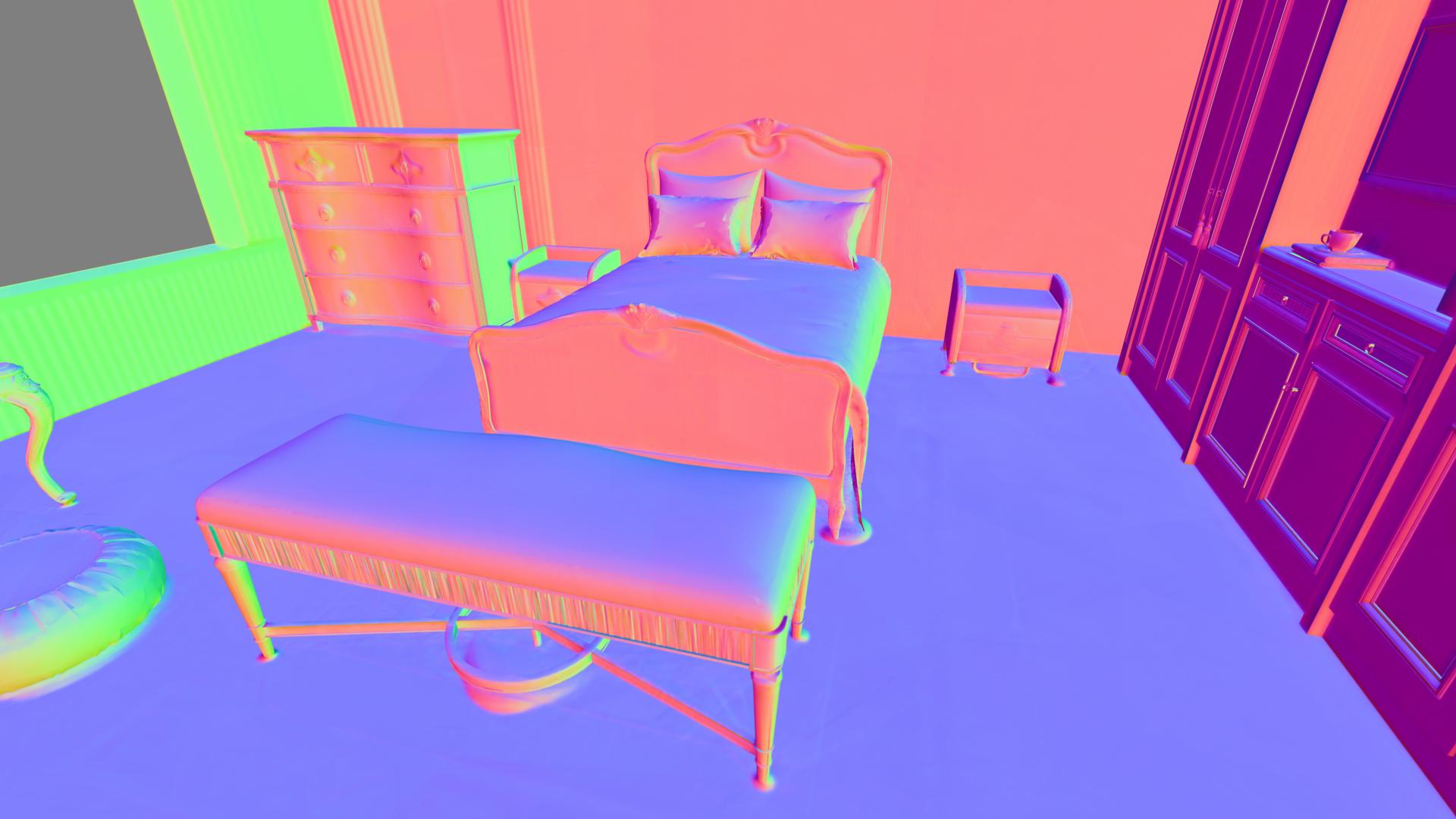}}
        &
        \fbox{\includegraphics[width=0.15\textwidth,trim={10cm 0 10cm 0},clip]{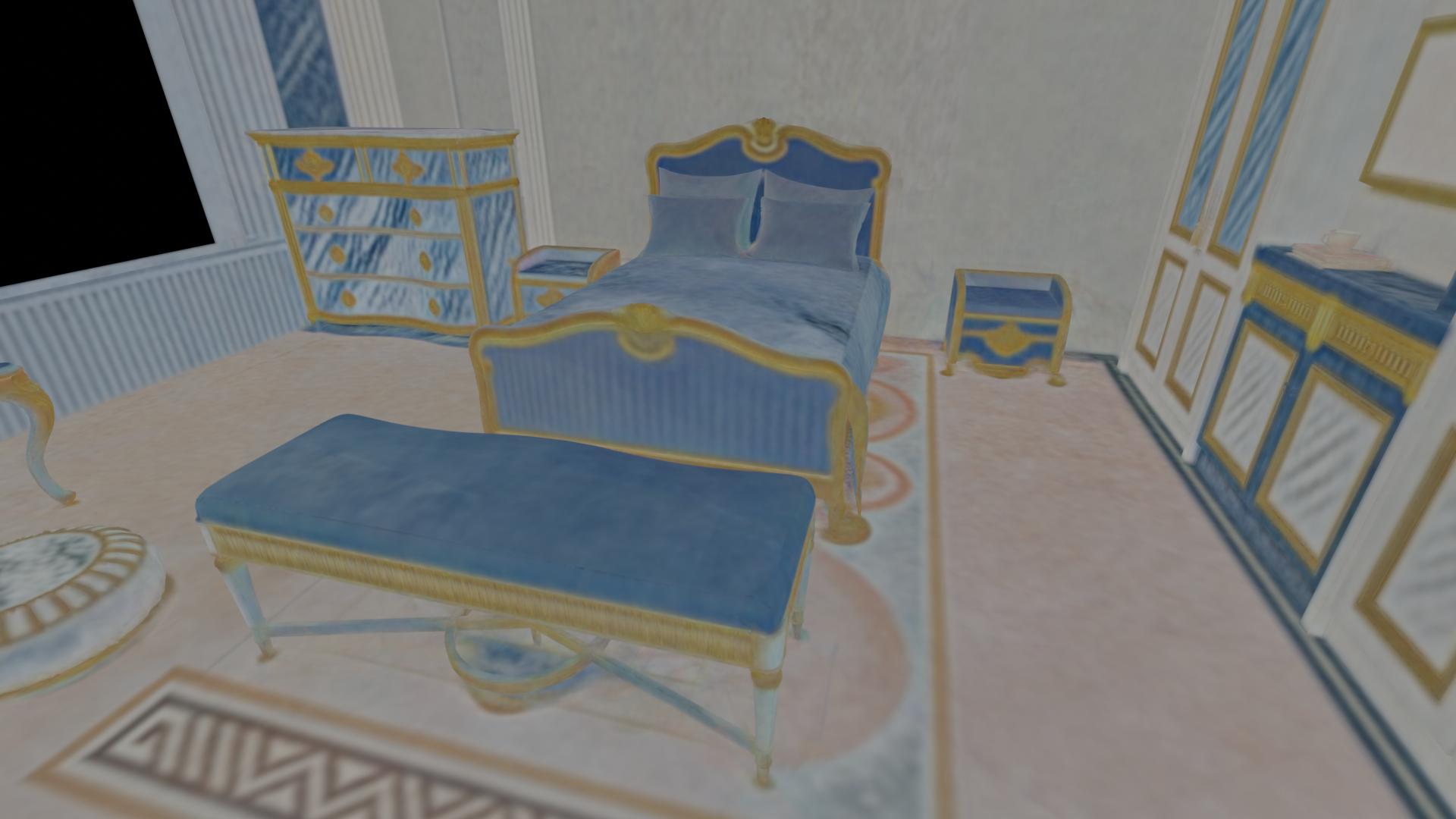}} 
        &
        \fbox{\includegraphics[width=0.15\textwidth,trim={10cm 0 10cm 0},clip]{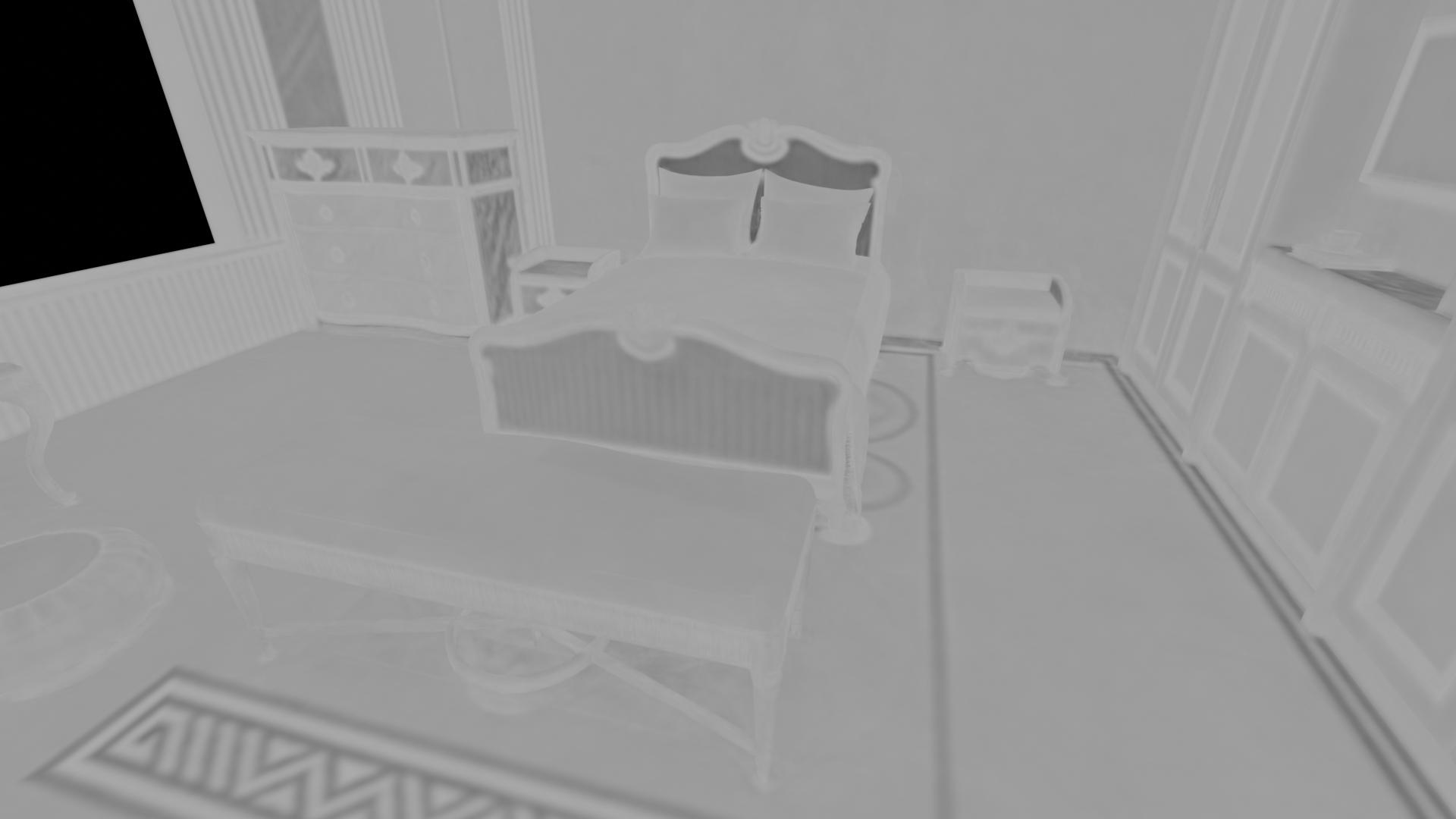}} 
        &
        \fbox{\includegraphics[width=0.15\textwidth,trim={10cm 0 10cm 0},clip]{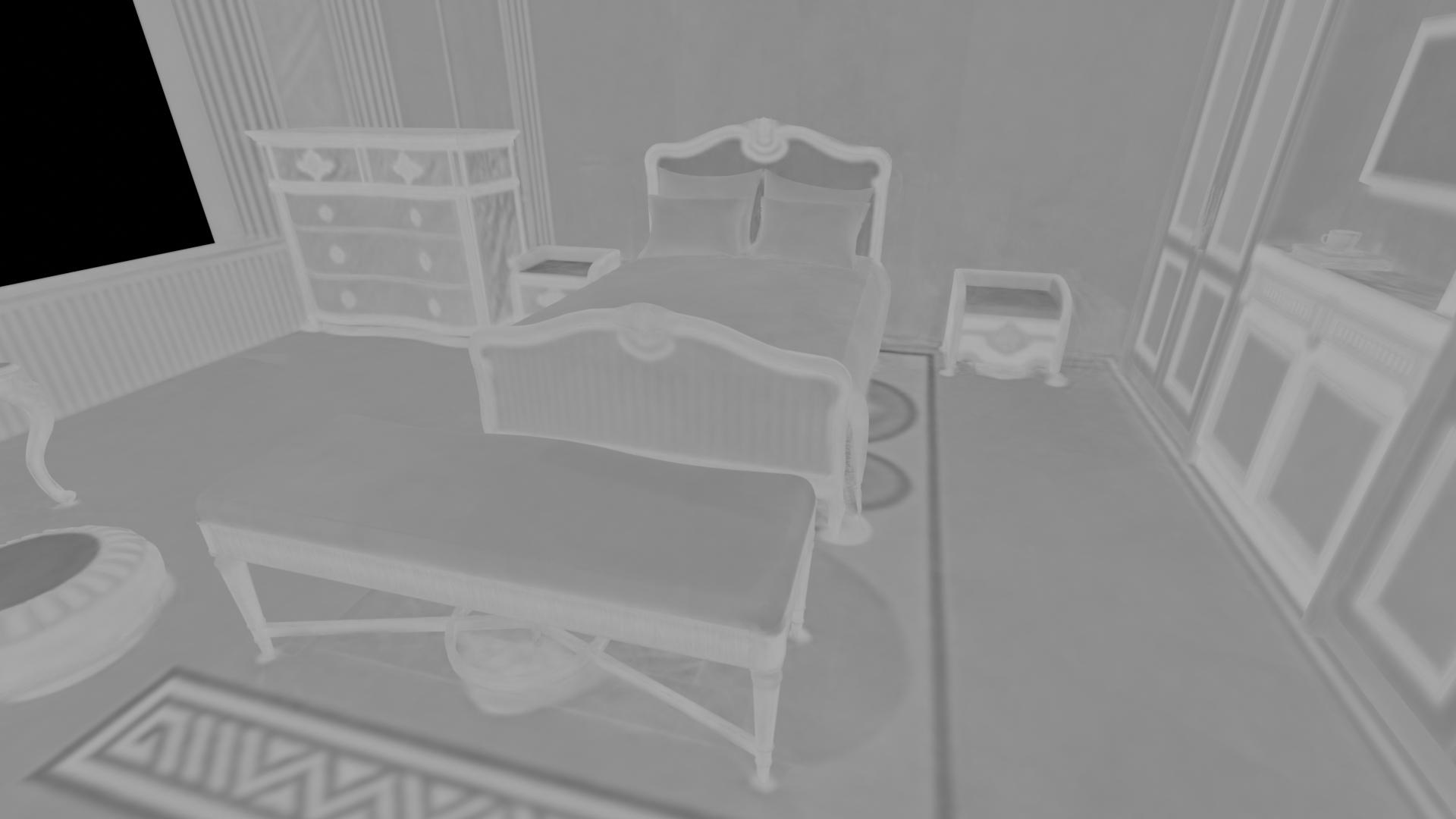}} 
        &
        \fbox{\includegraphics[width=0.15\textwidth,trim={10cm 0 10cm 0},clip]{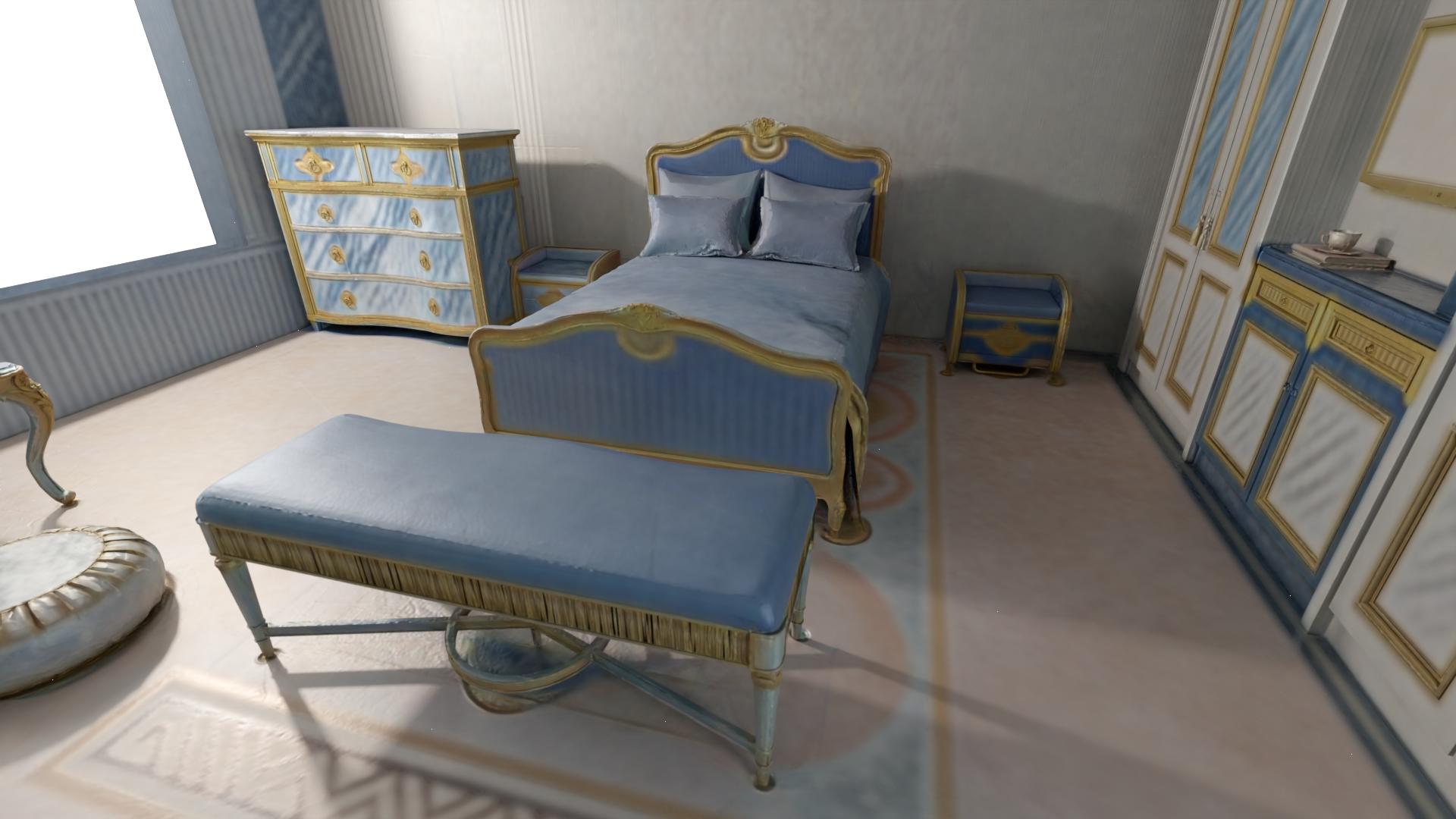}} 
        \\
        
        
        \rotatebox{90}{{\footnotesize View 3}}
        &
        \fbox{\includegraphics[width=0.15\textwidth,trim={10cm 0 10cm 0},clip]{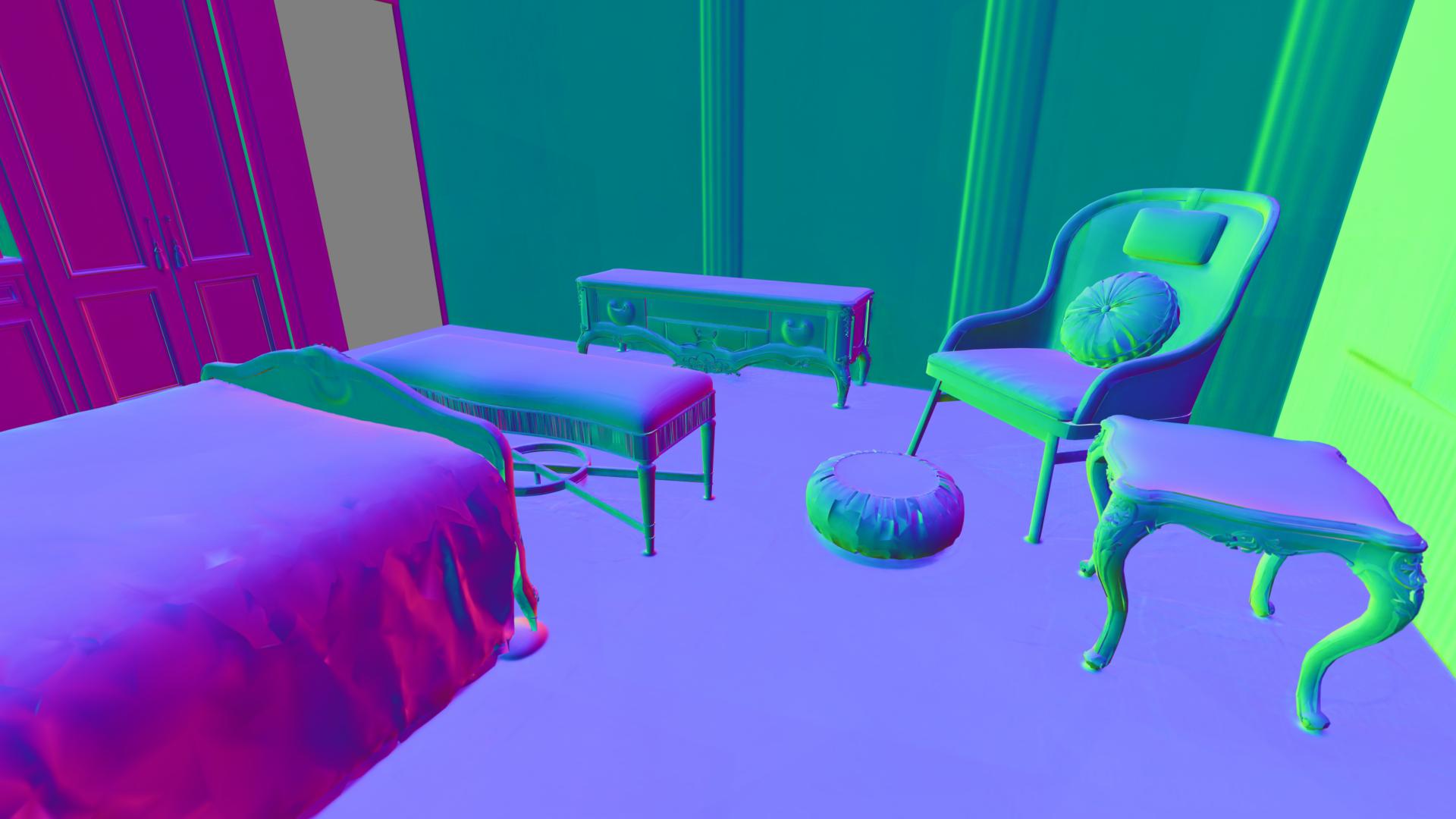}} 
        &
        \fbox{\includegraphics[width=0.15\textwidth,trim={10cm 0 10cm 0},clip]{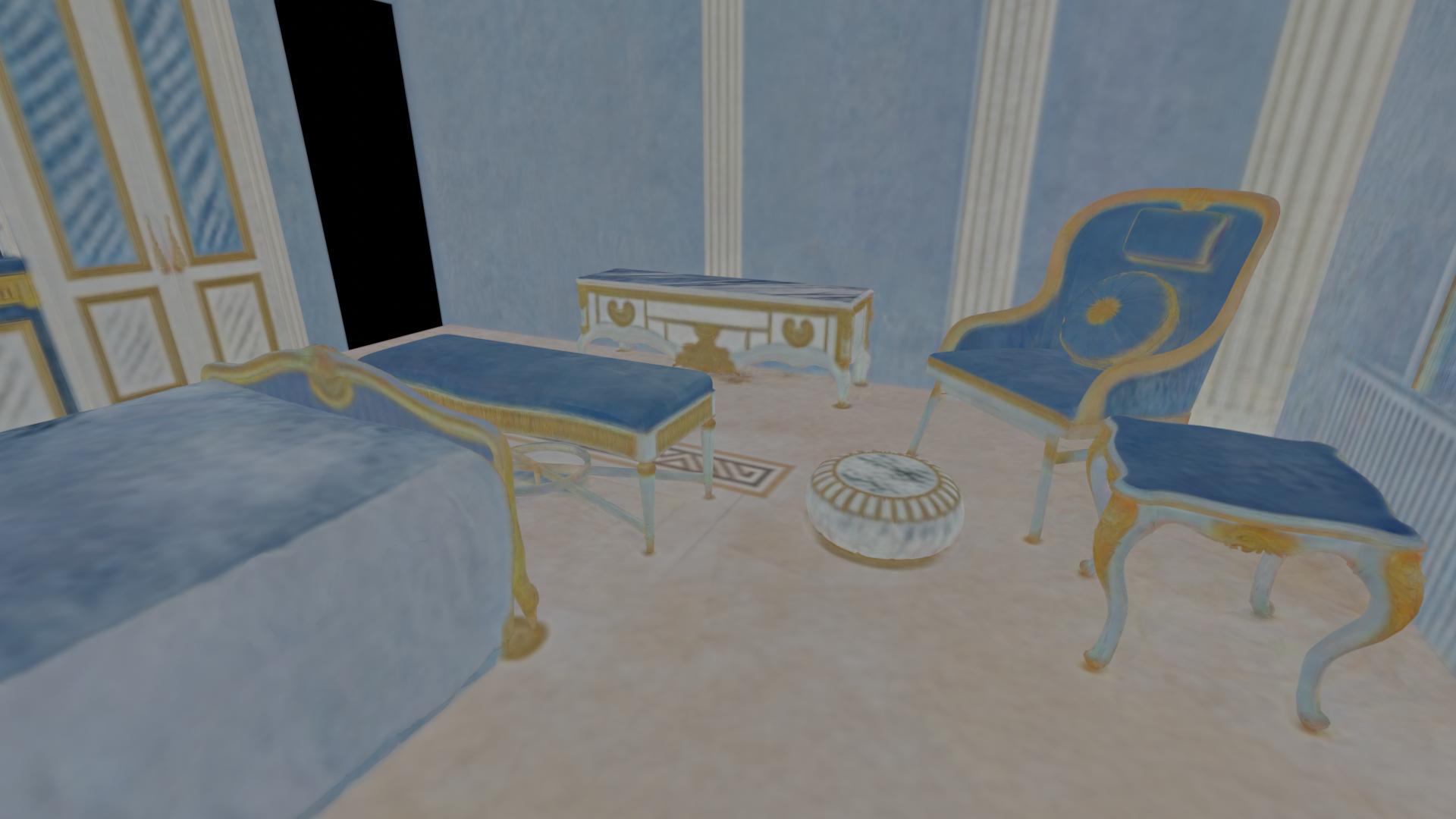}} 
        &
        \fbox{\includegraphics[width=0.15\textwidth,trim={10cm 0 10cm 0},clip]{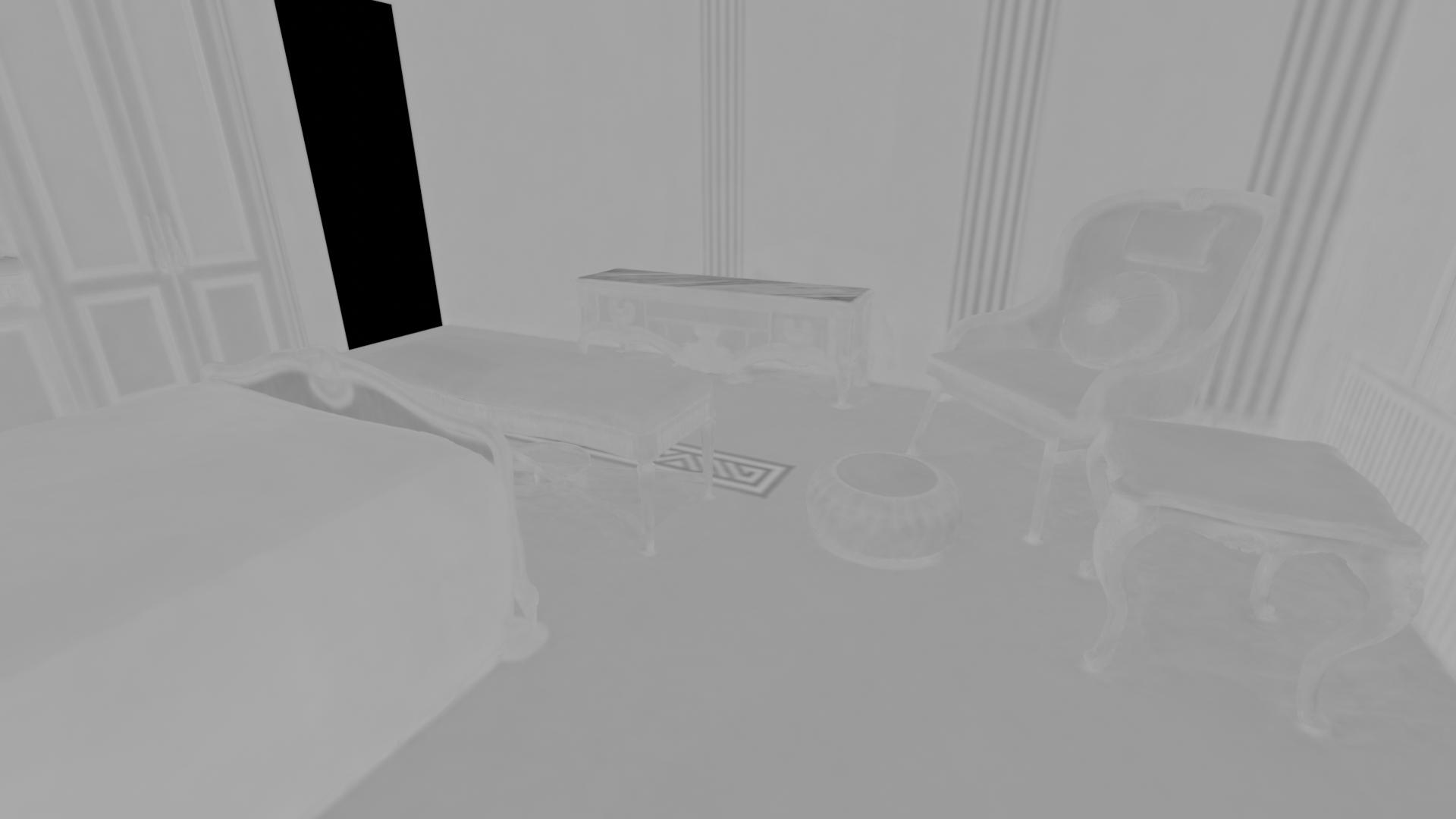}} 
        &
        \fbox{\includegraphics[width=0.15\textwidth,trim={10cm 0 10cm 0},clip]{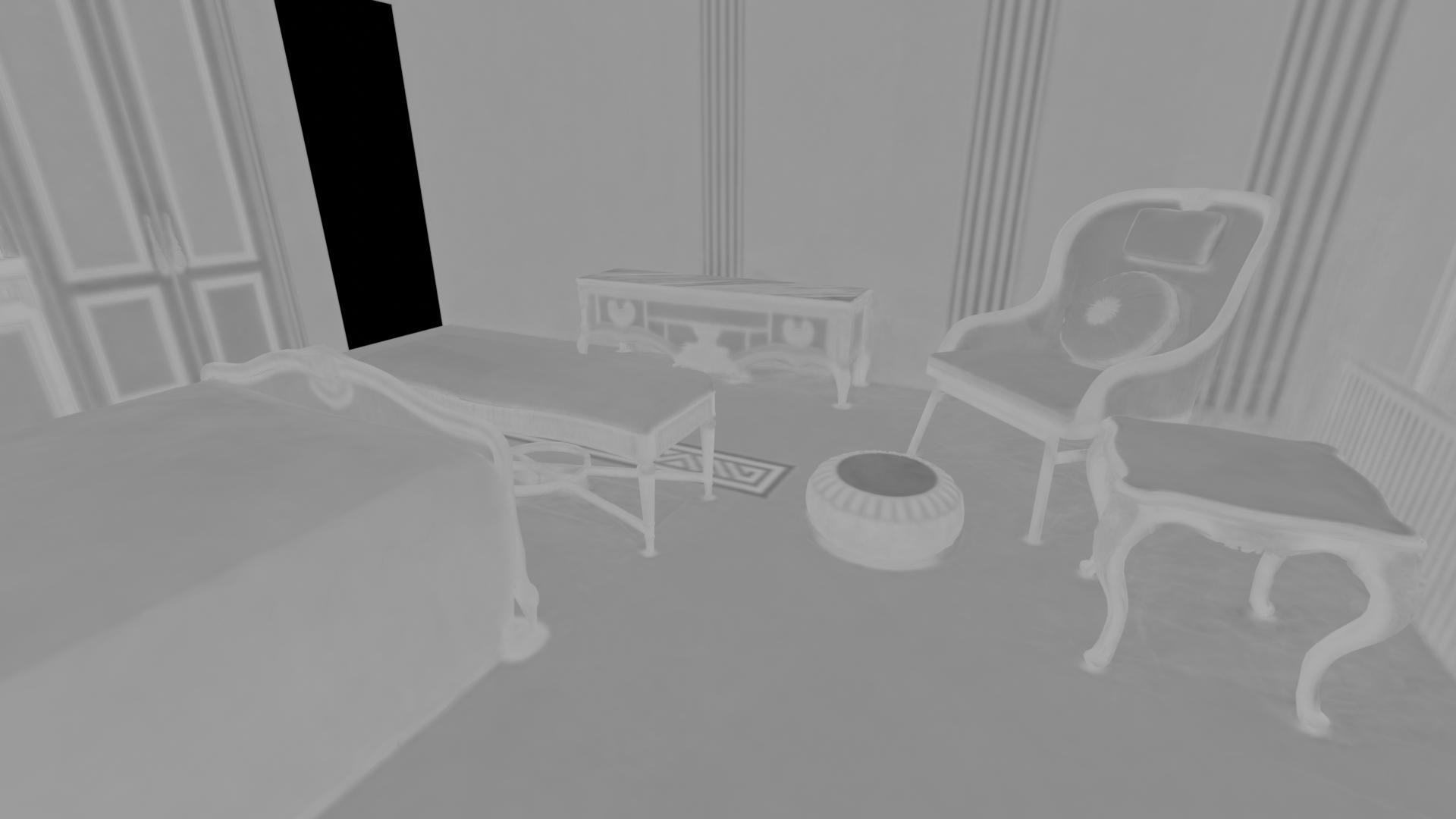}} 
        &
        \fbox{\includegraphics[width=0.15\textwidth,trim={10cm 0 10cm 0},clip]{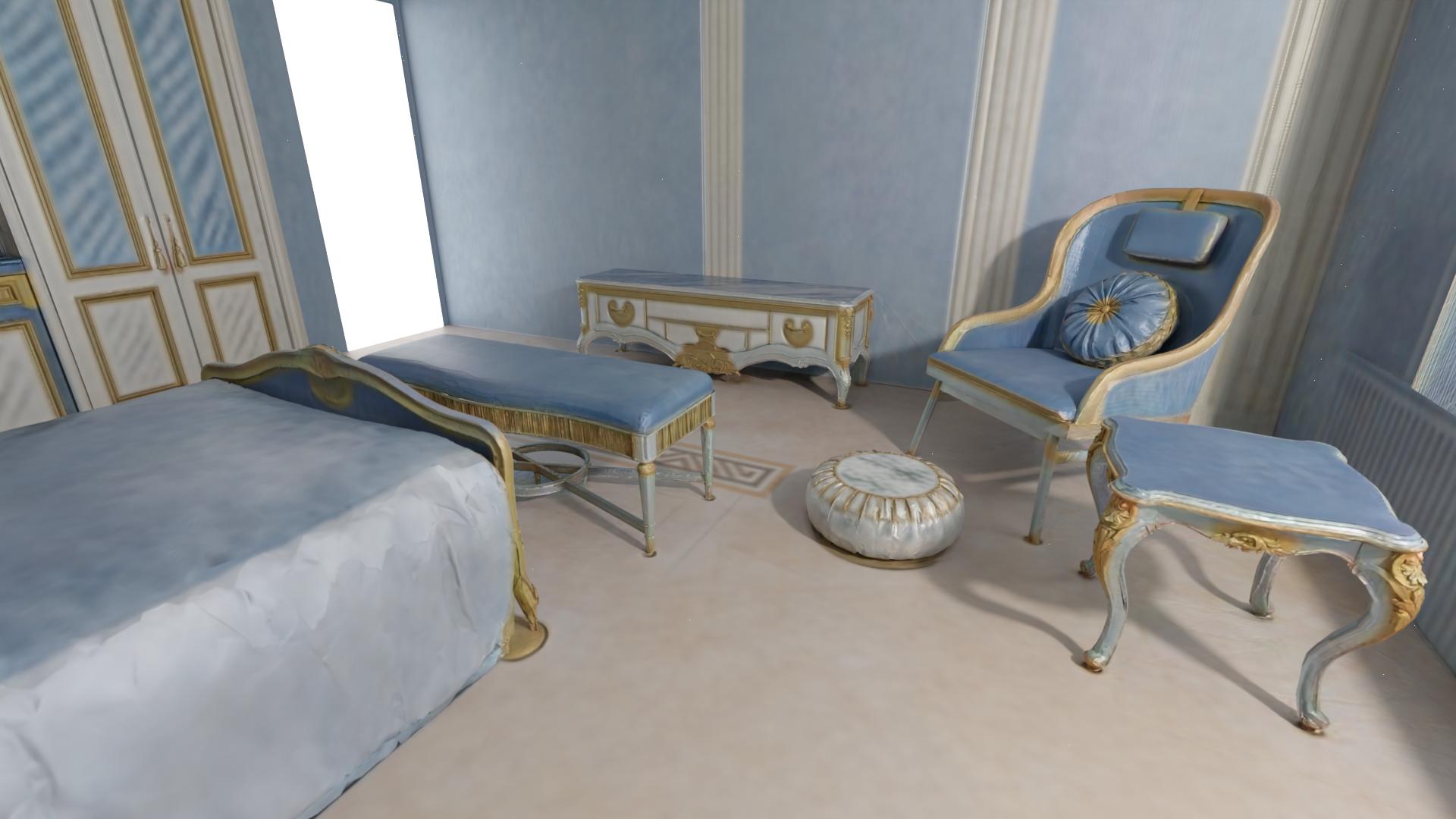}} 
        \\

        &
        {\footnotesize Normal} &
        {\footnotesize Albedo} &
        {\footnotesize Roughness} &
        {\footnotesize Metallic} &
        {\footnotesize Rendering} \\
    \end{tabular}}
    \caption{\textbf{Scene Texturing}. 
    We show more scene texturing results on multiple 3D-Front scenes \cite{Front3d} with multiple prompts. Continues on the next page.
    }
\end{figure*}
\begin{figure*}[p]
    \ContinuedFloat
    \centering
    \setlength\tabcolsep{1.25pt}
    \resizebox{\textwidth}{!}{
    \fboxsep=0pt
    \begin{tabular}{ccccc|c}
        \rotatebox{90}{{\footnotesize View 1}}
        &
        \fbox{\includegraphics[width=0.15\textwidth,trim={10cm 0 10cm 0},clip]{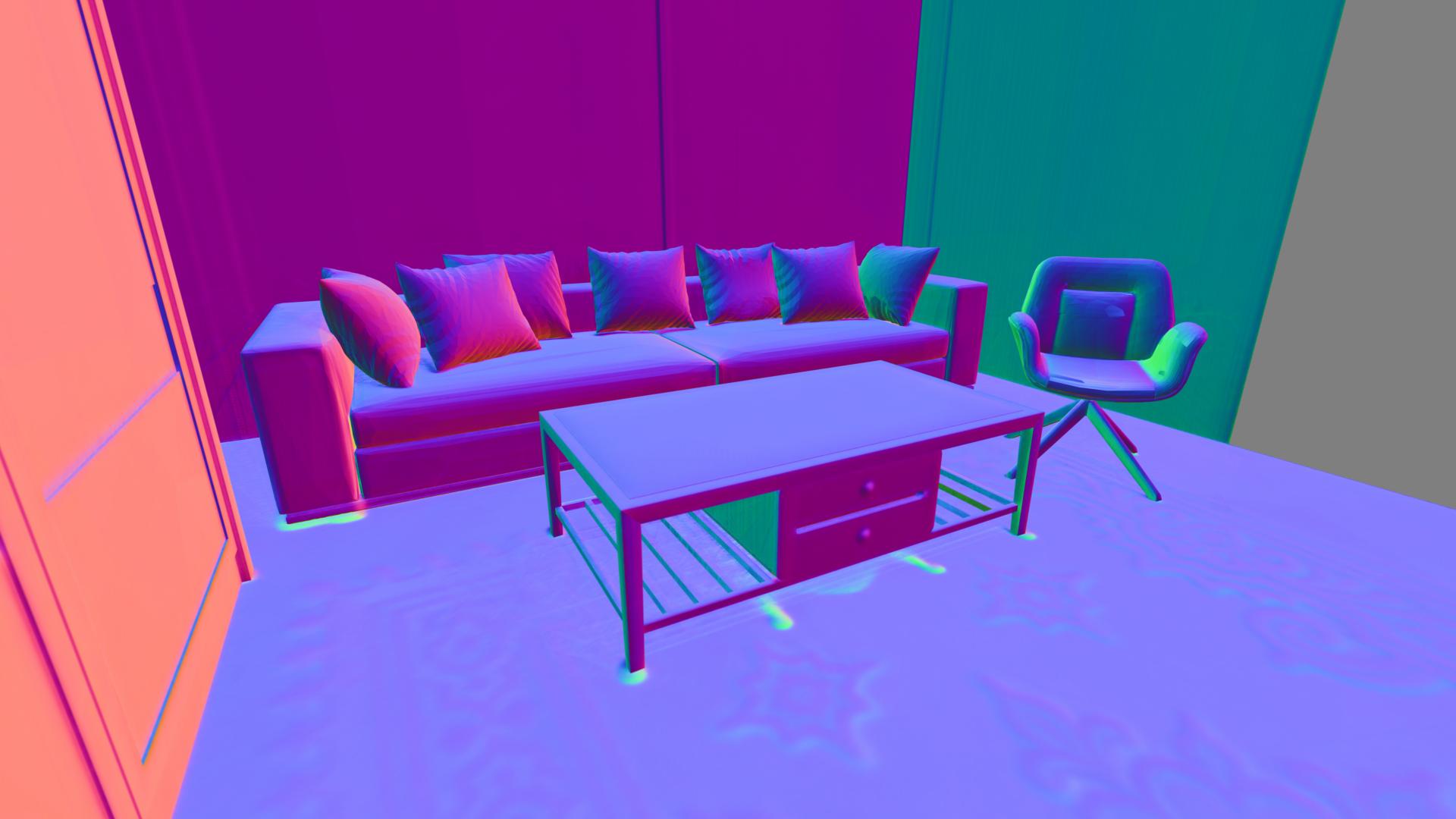}}
        &
        \fbox{\includegraphics[width=0.15\textwidth,trim={10cm 0 10cm 0},clip]{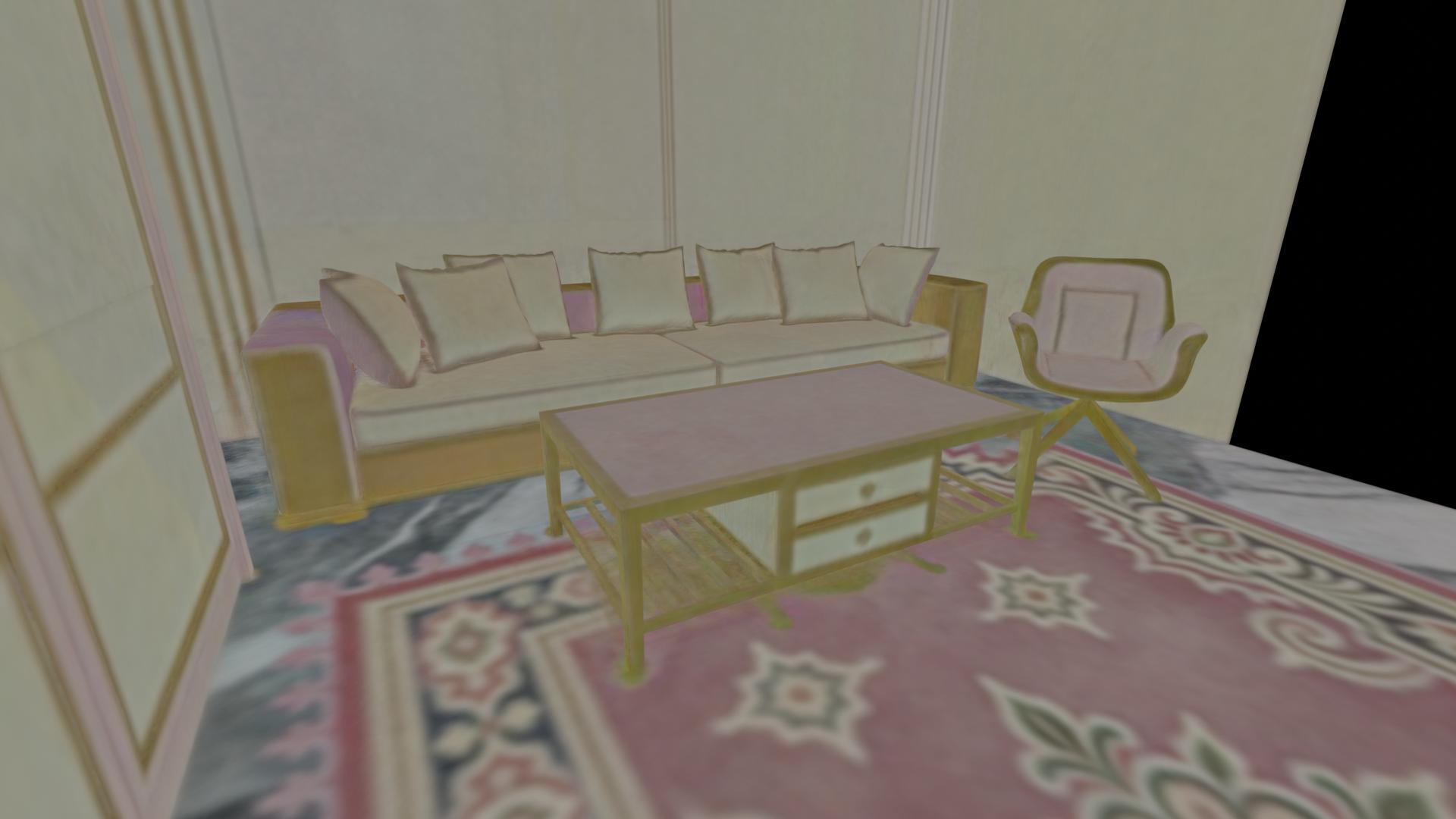}} 
        &
        \fbox{\includegraphics[width=0.15\textwidth,trim={10cm 0 10cm 0},clip]{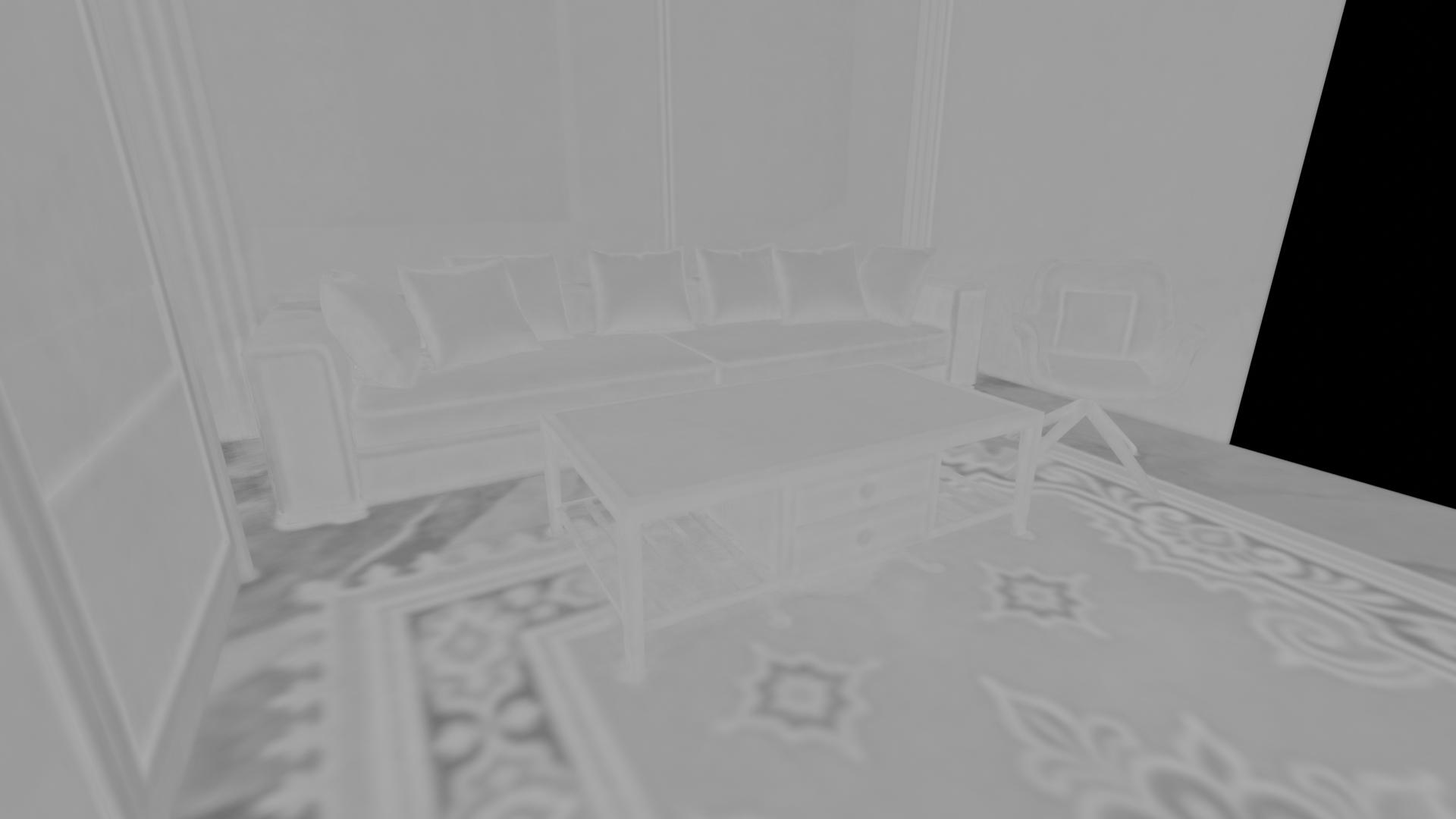}} 
        &
        \fbox{\includegraphics[width=0.15\textwidth,trim={10cm 0 10cm 0},clip]{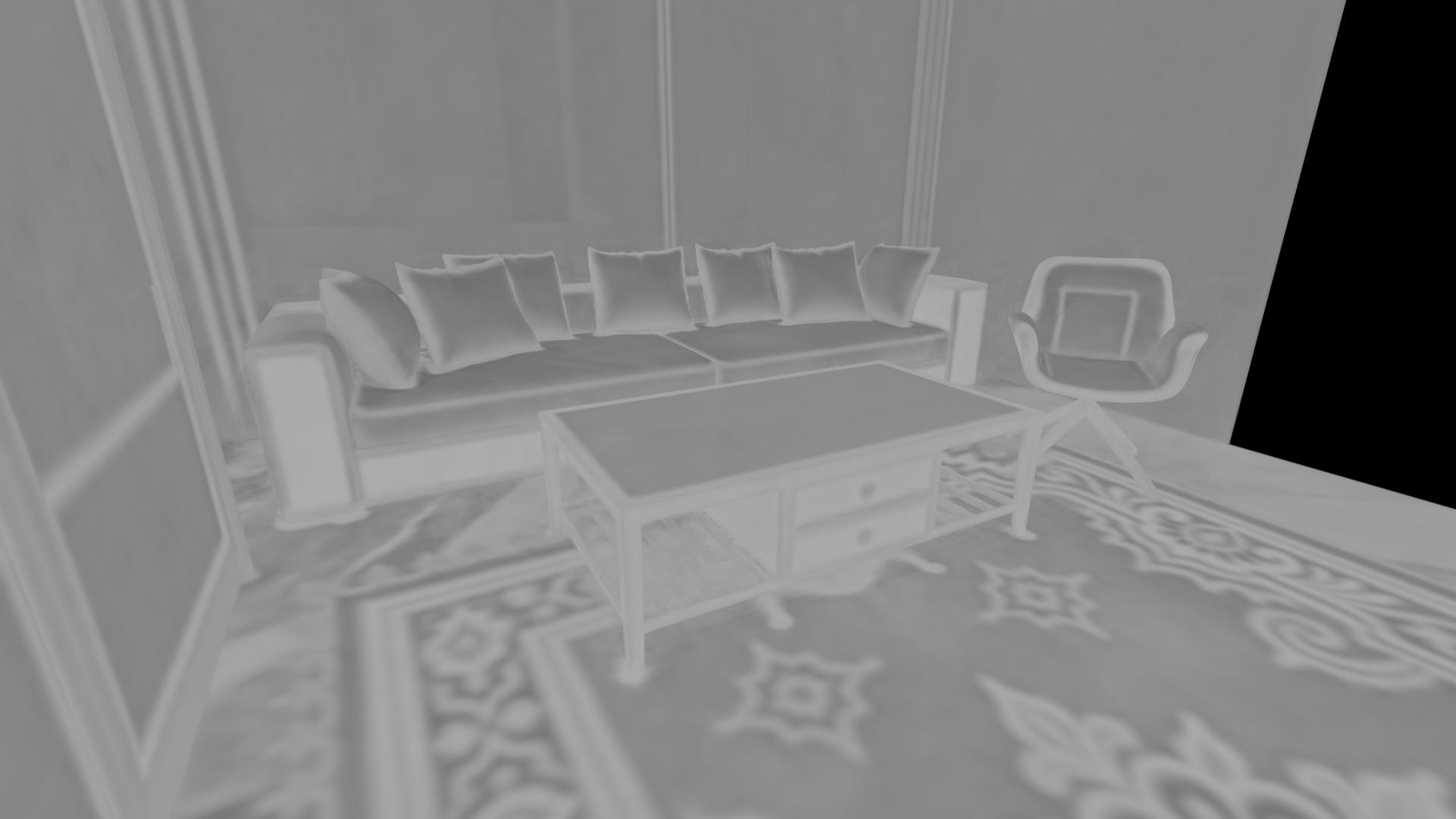}} 
        &
        \fbox{\includegraphics[width=0.15\textwidth,trim={10cm 0 10cm 0},clip]{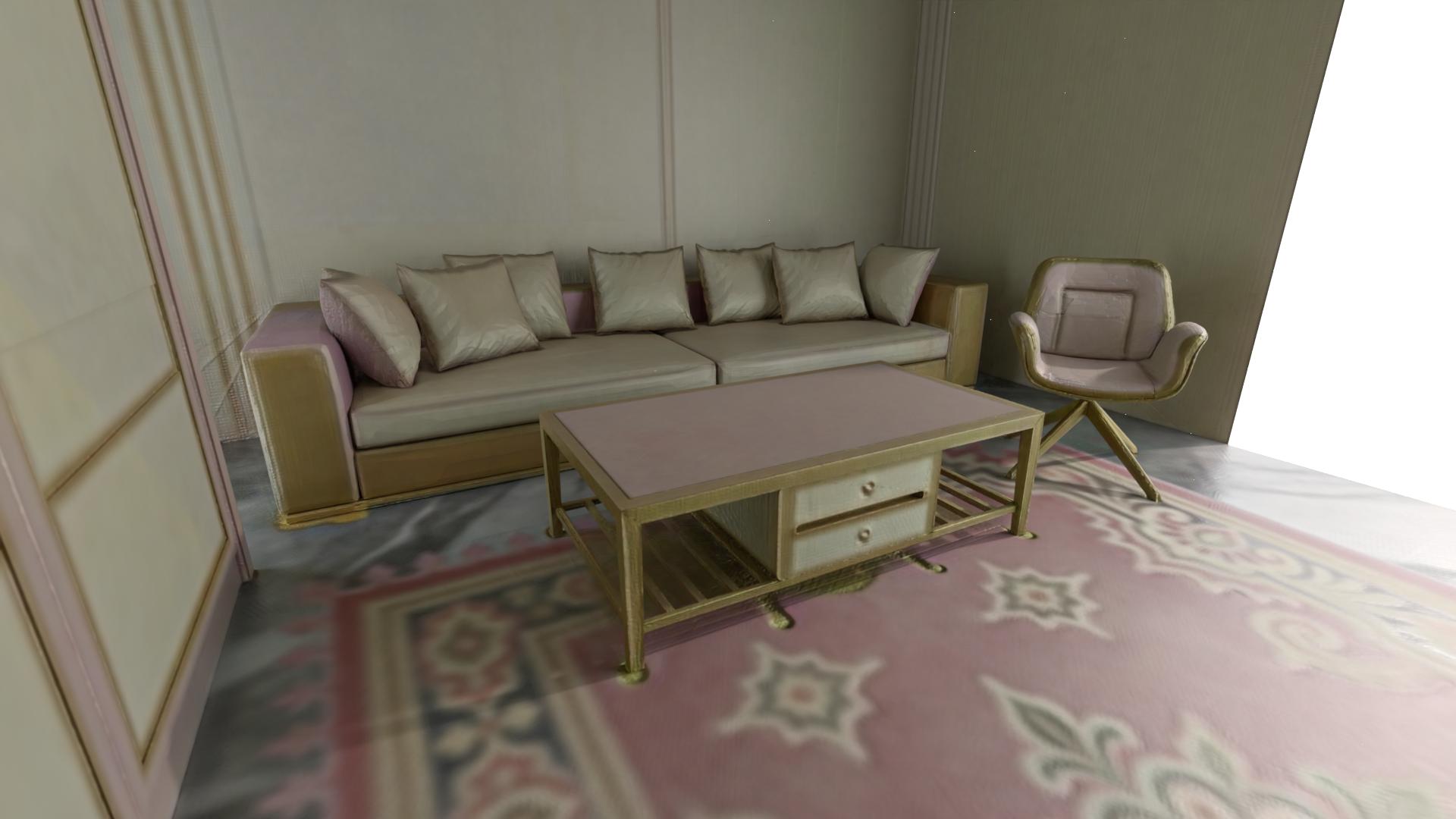}} 
        \\
        
        
        \rotatebox{90}{{\footnotesize View 3}}
        &
        \fbox{\includegraphics[width=0.15\textwidth,trim={10cm 0 10cm 0},clip]{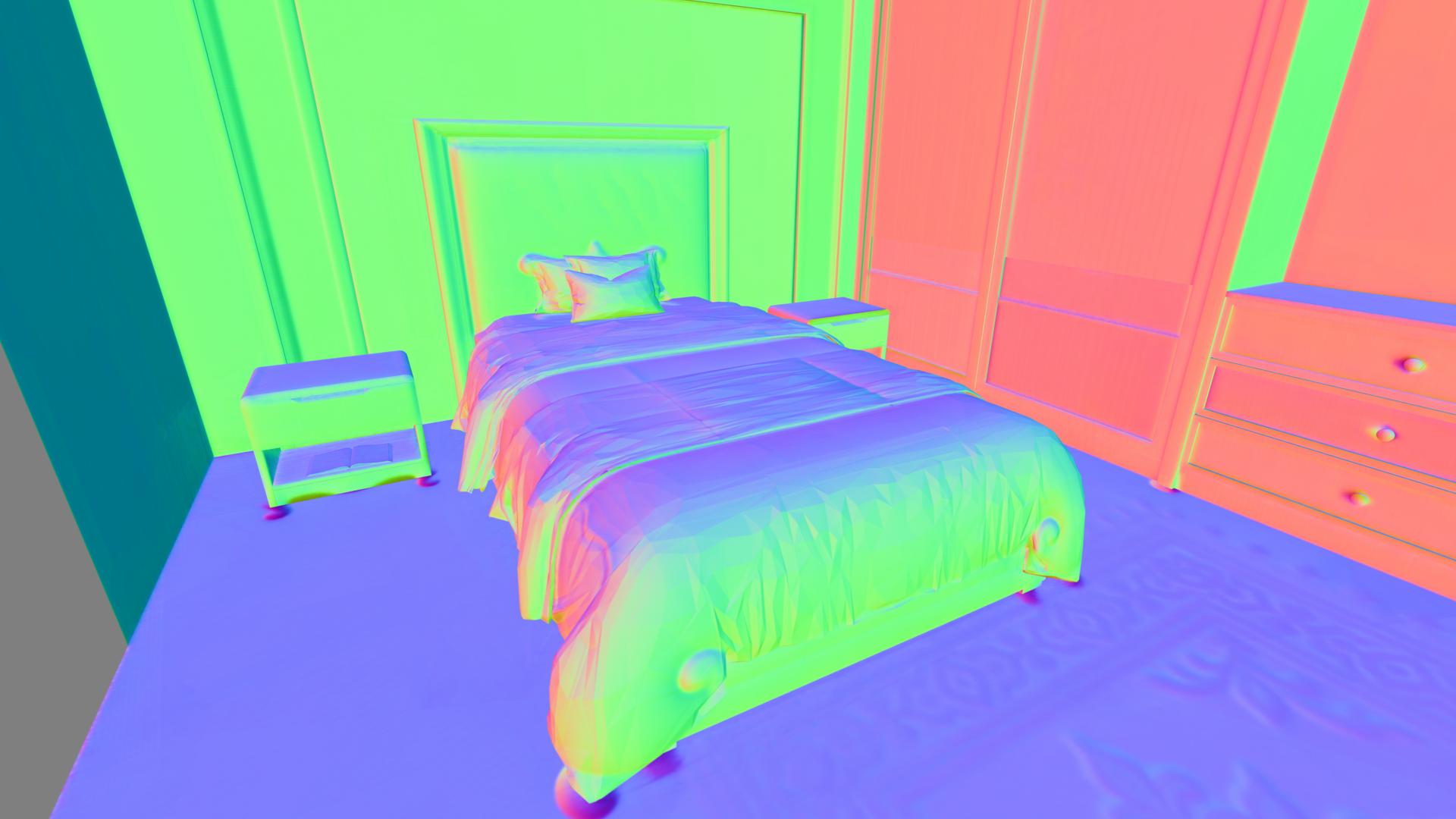}} 
        &
        \fbox{\includegraphics[width=0.15\textwidth,trim={10cm 0 10cm 0},clip]{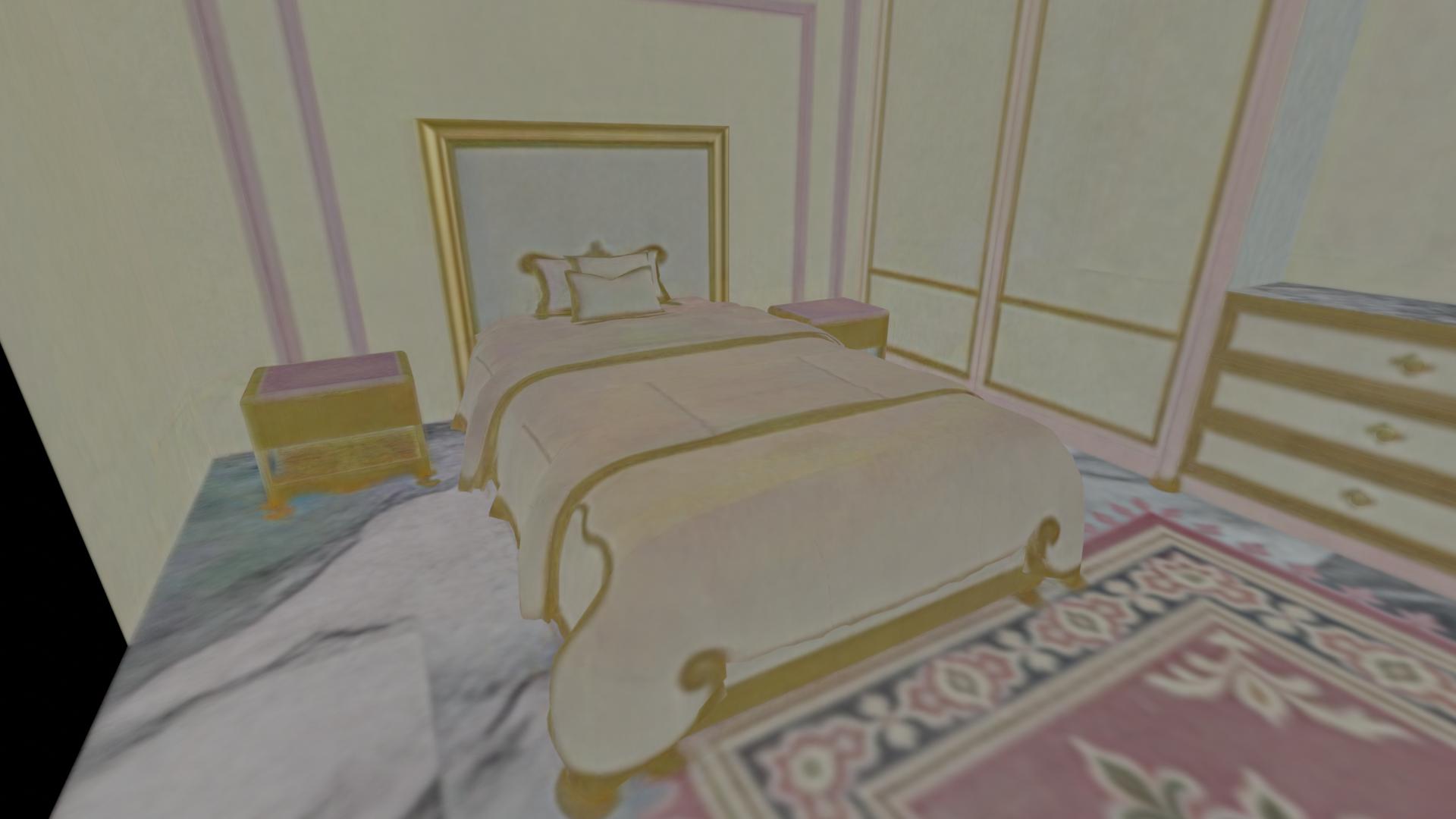}} 
        &
        \fbox{\includegraphics[width=0.15\textwidth,trim={10cm 0 10cm 0},clip]{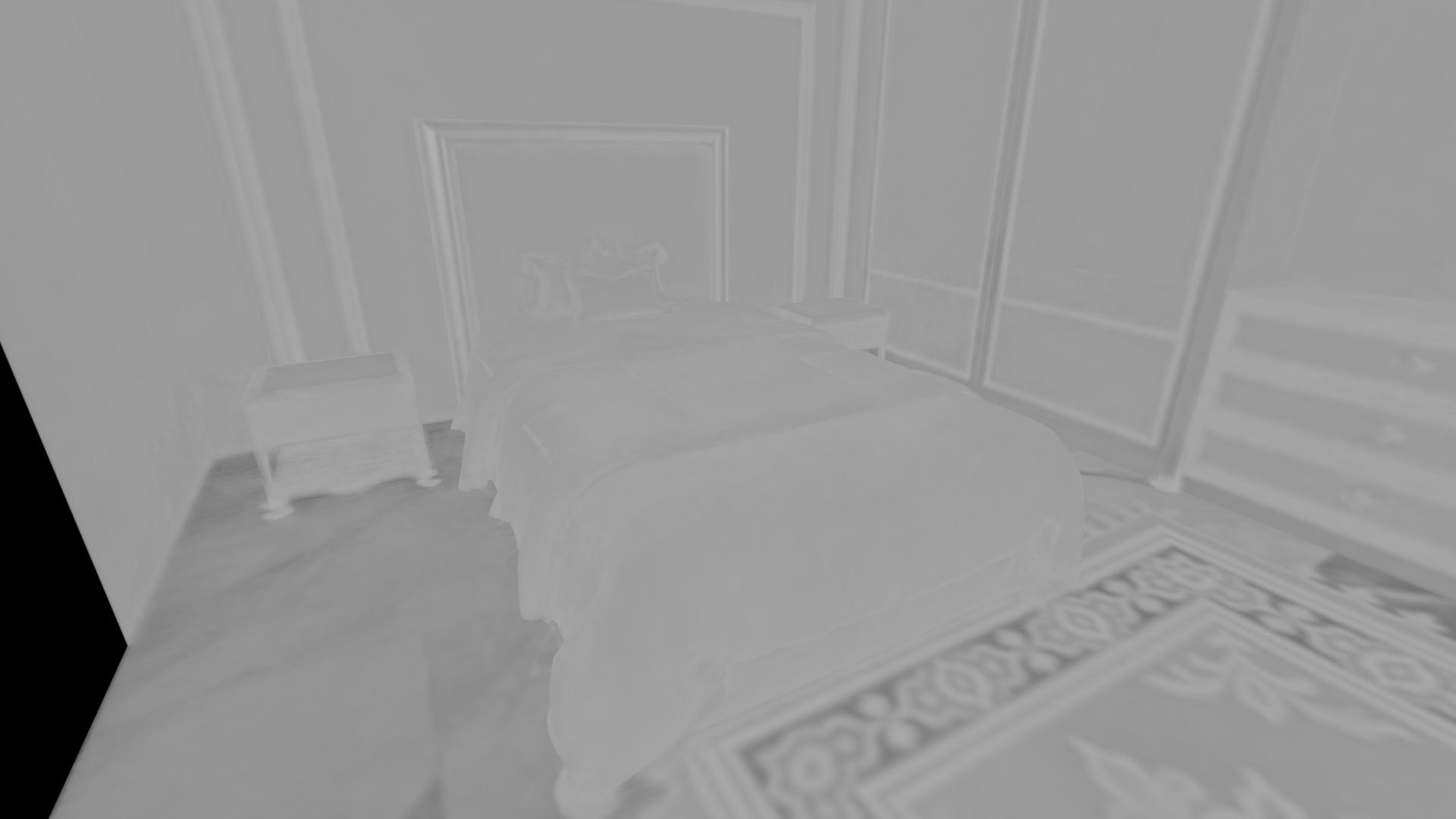}} 
        &
        \fbox{\includegraphics[width=0.15\textwidth,trim={10cm 0 10cm 0},clip]{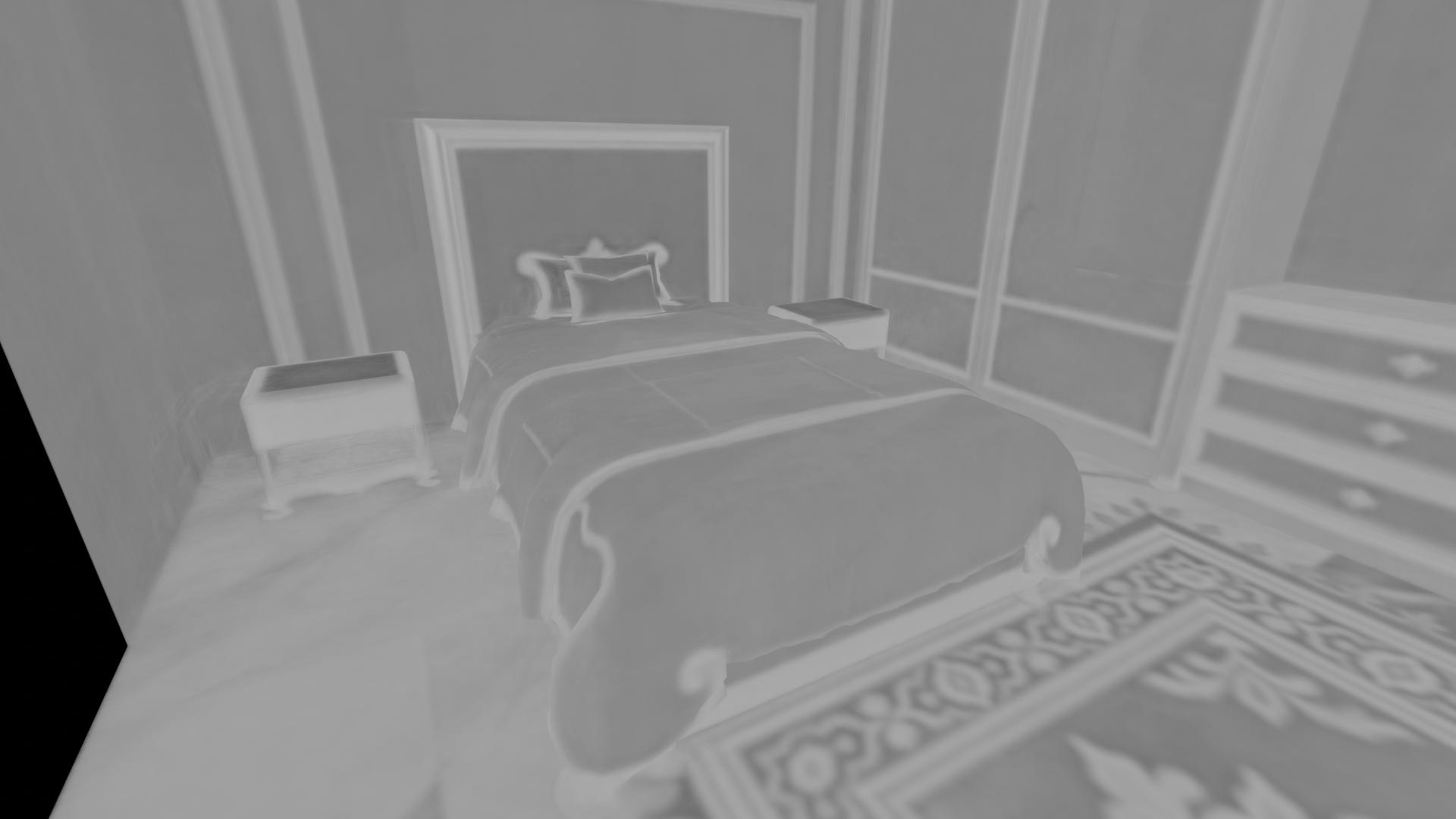}} 
        &
        \fbox{\includegraphics[width=0.15\textwidth,trim={10cm 0 10cm 0},clip]{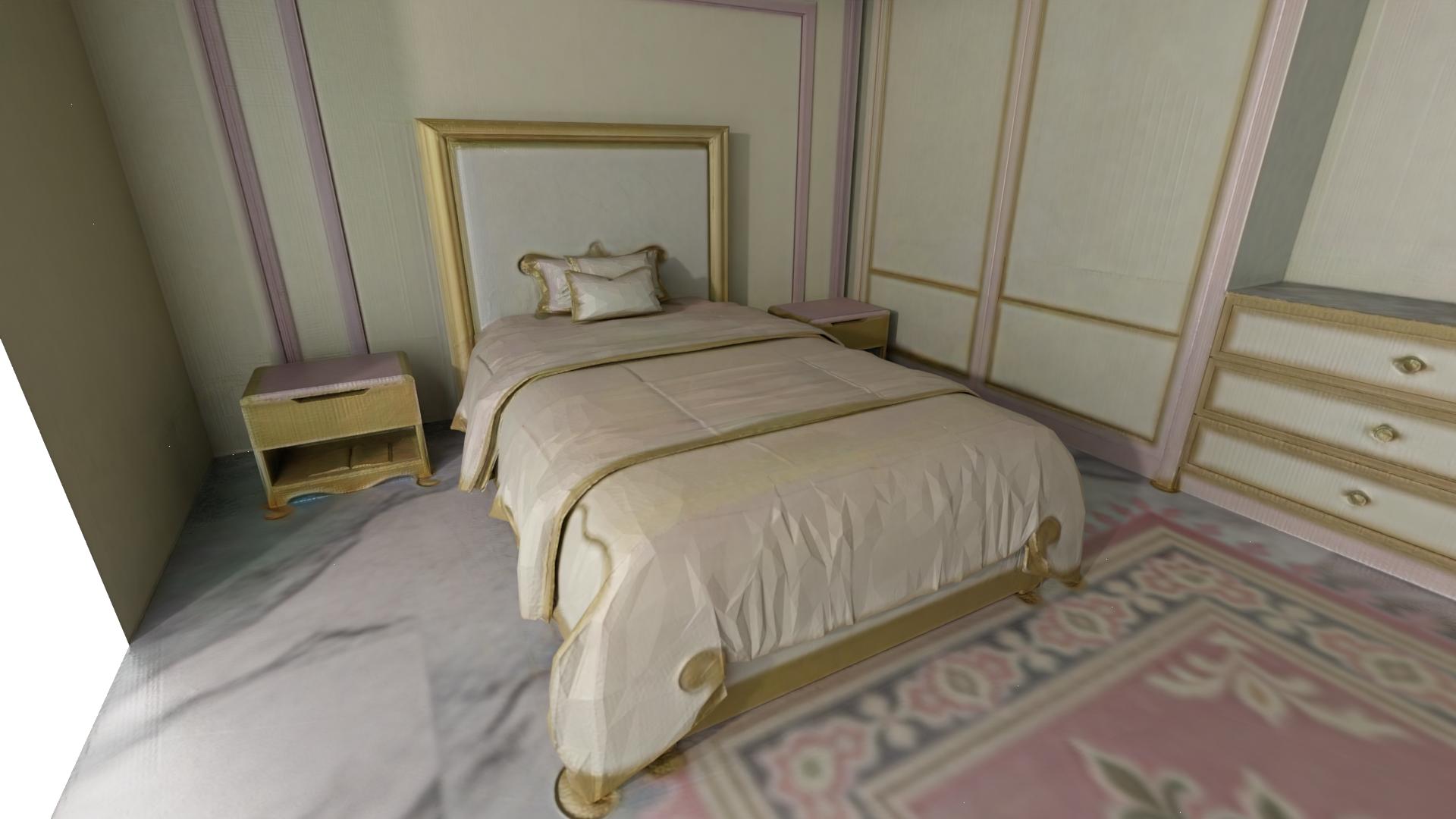}} 
        \\

        &
        {\footnotesize Normal} &
        {\footnotesize Albedo} &
        {\footnotesize Roughness} &
        {\footnotesize Metallic} &
        {\footnotesize Rendering} \\

        \midrule
        
        \rotatebox{90}{{\footnotesize View 1}}
        &
        \fbox{\includegraphics[width=0.15\textwidth,trim={10cm 0 10cm 0},clip]{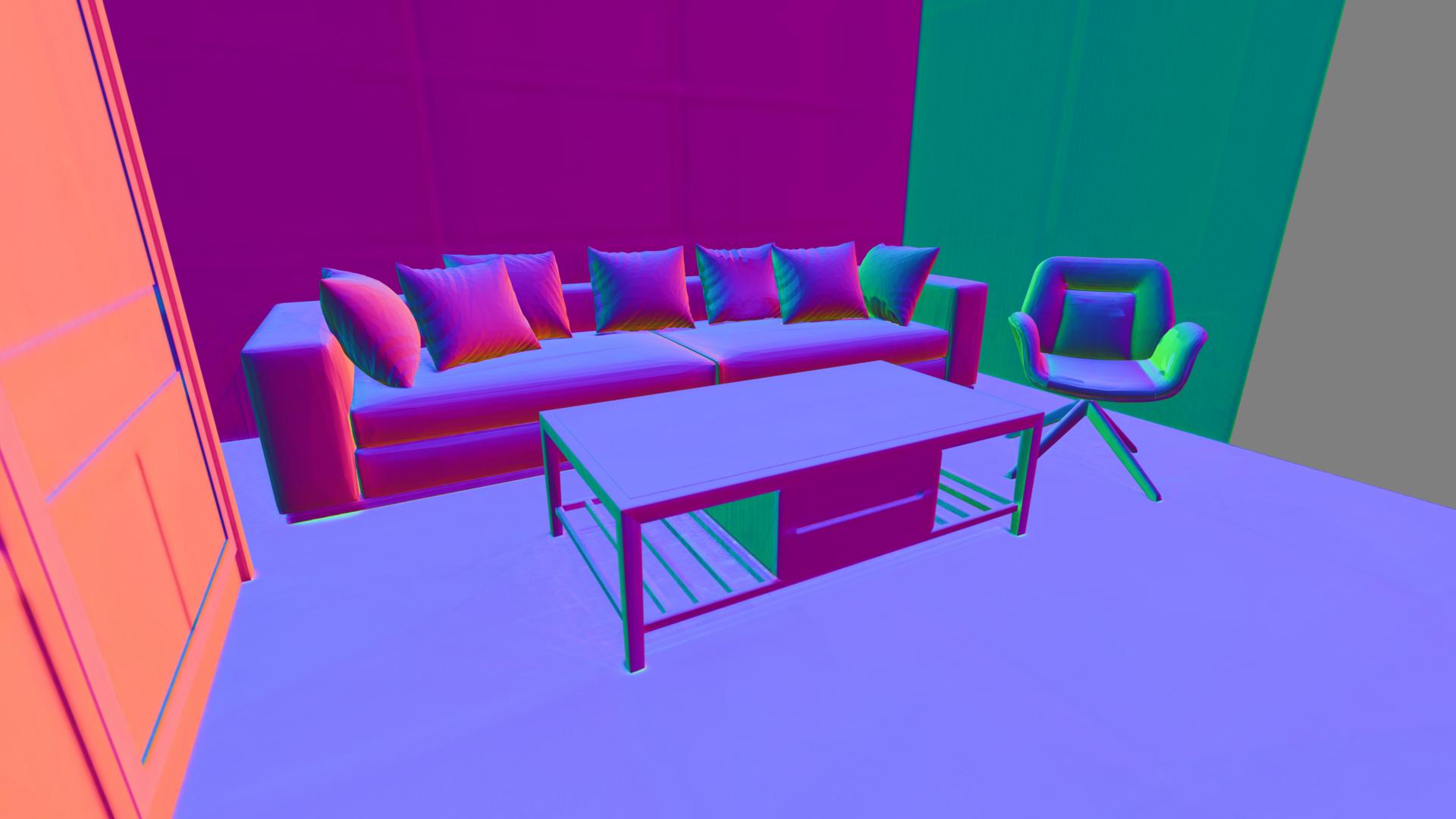}}
        &
        \fbox{\includegraphics[width=0.15\textwidth,trim={10cm 0 10cm 0},clip]{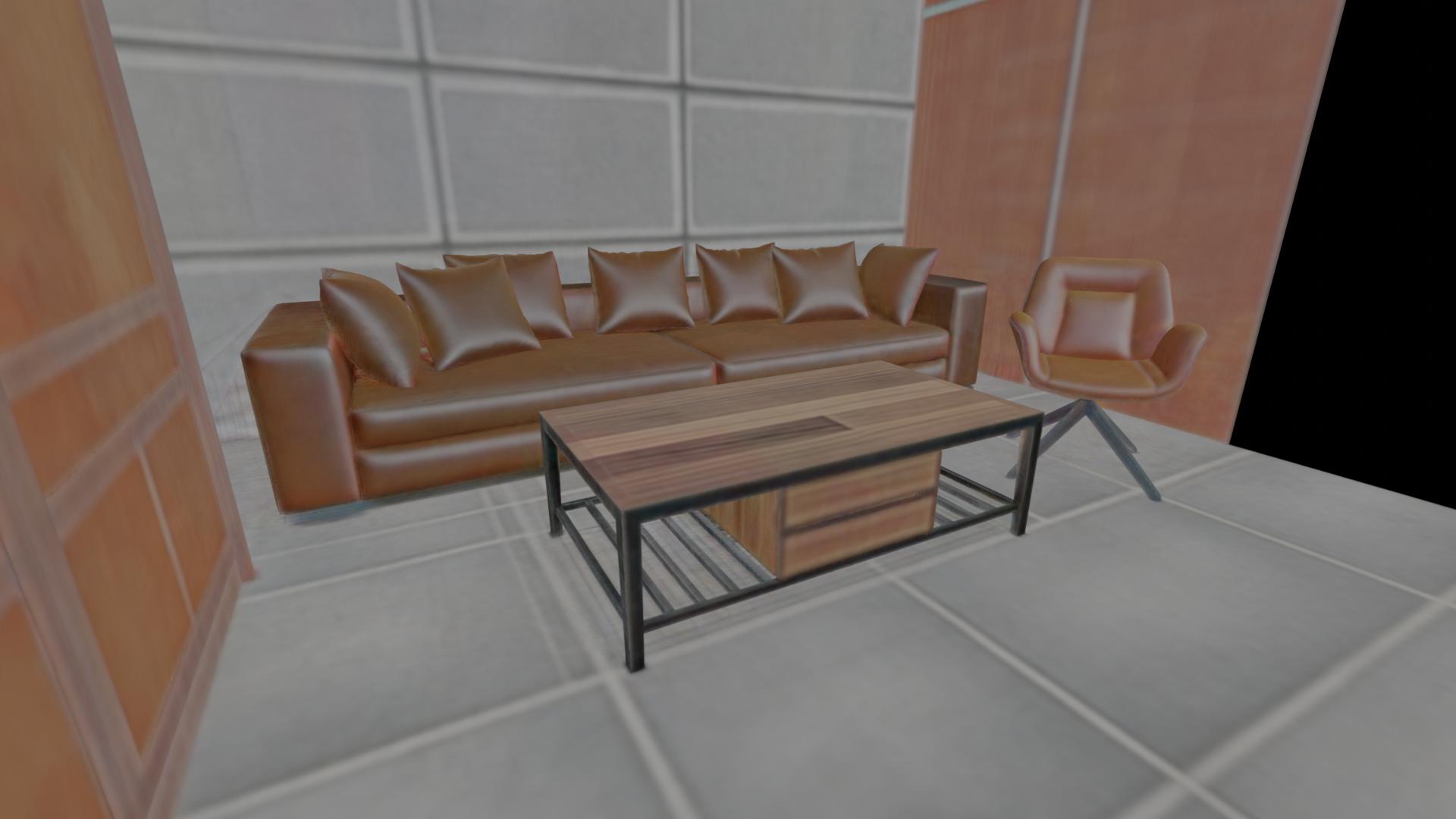}} 
        &
        \fbox{\includegraphics[width=0.15\textwidth,trim={10cm 0 10cm 0},clip]{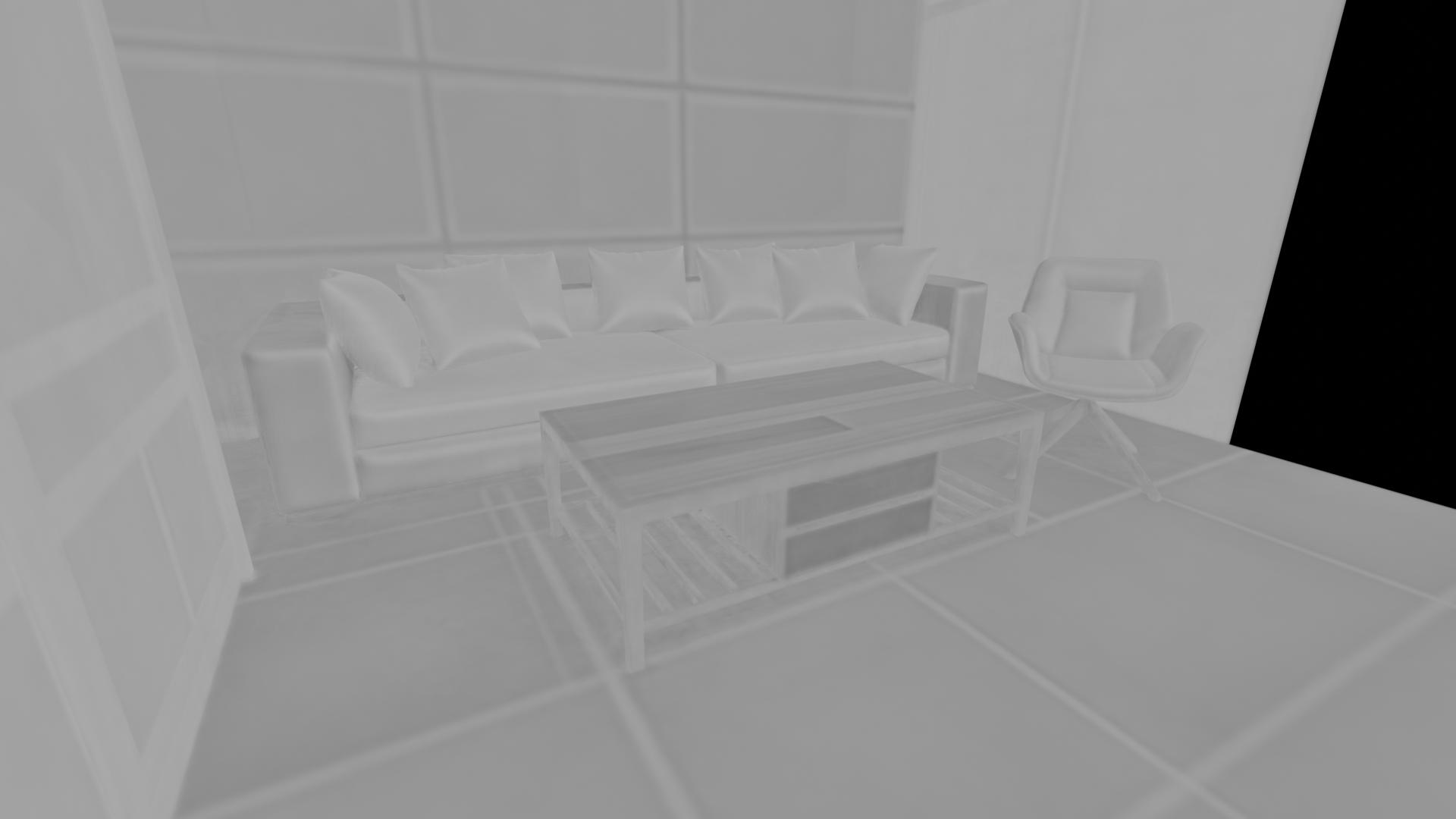}} 
        &
        \fbox{\includegraphics[width=0.15\textwidth,trim={10cm 0 10cm 0},clip]{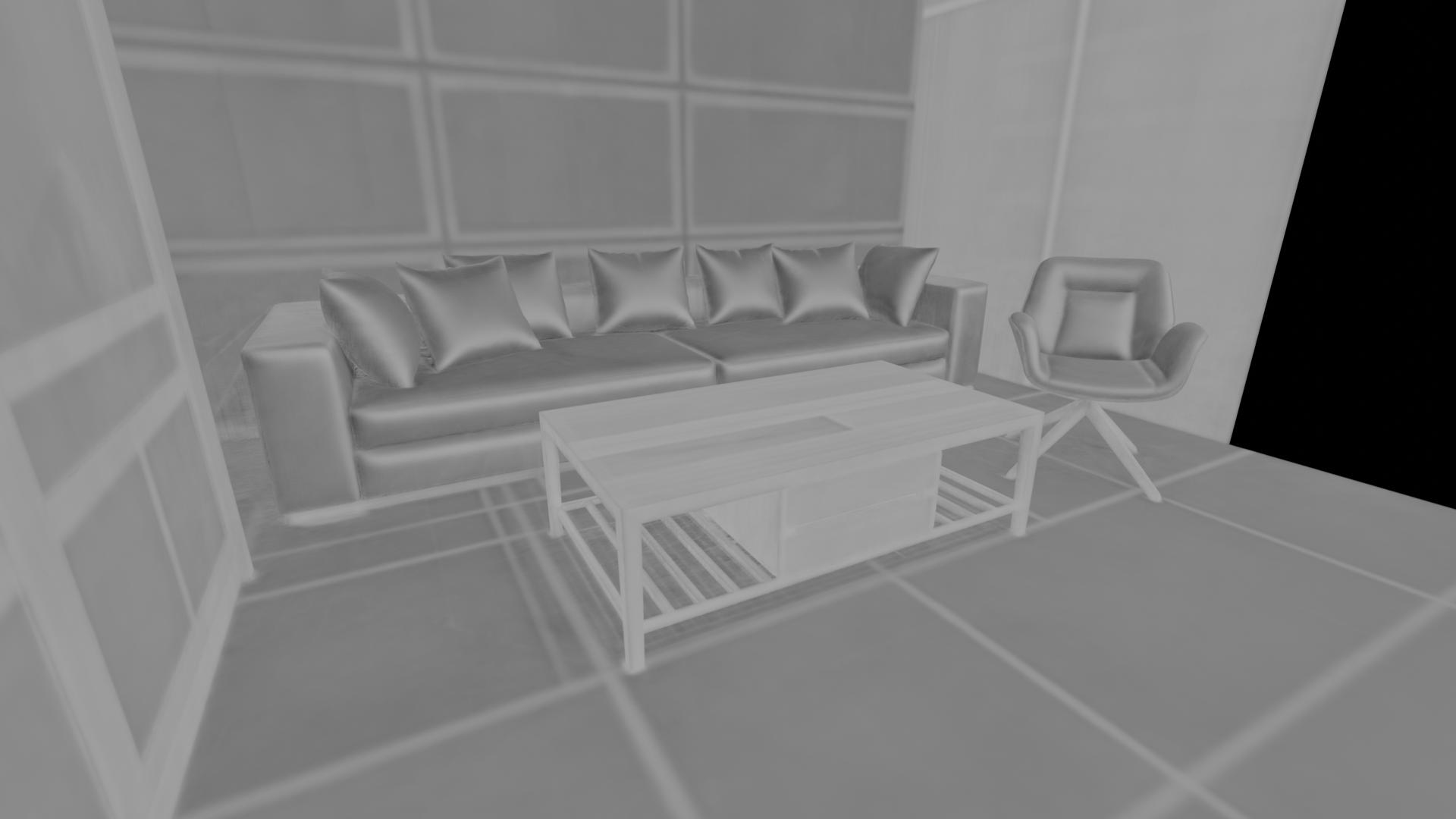}} 
        &
        \fbox{\includegraphics[width=0.15\textwidth,trim={10cm 0 10cm 0},clip]{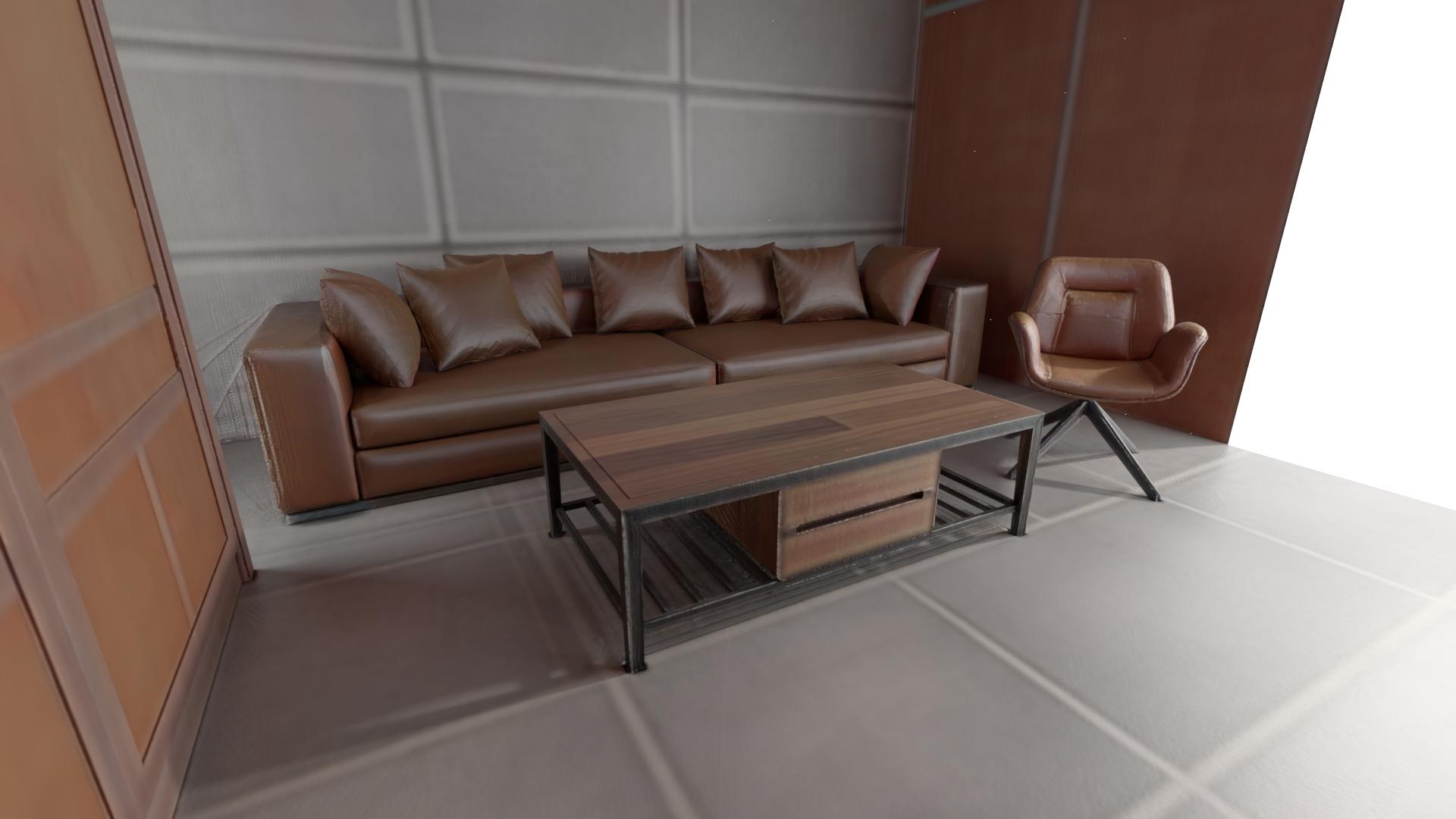}} 
        \\
        
        
        \rotatebox{90}{{\footnotesize View 3}}
        &
        \fbox{\includegraphics[width=0.15\textwidth,trim={10cm 0 10cm 0},clip]{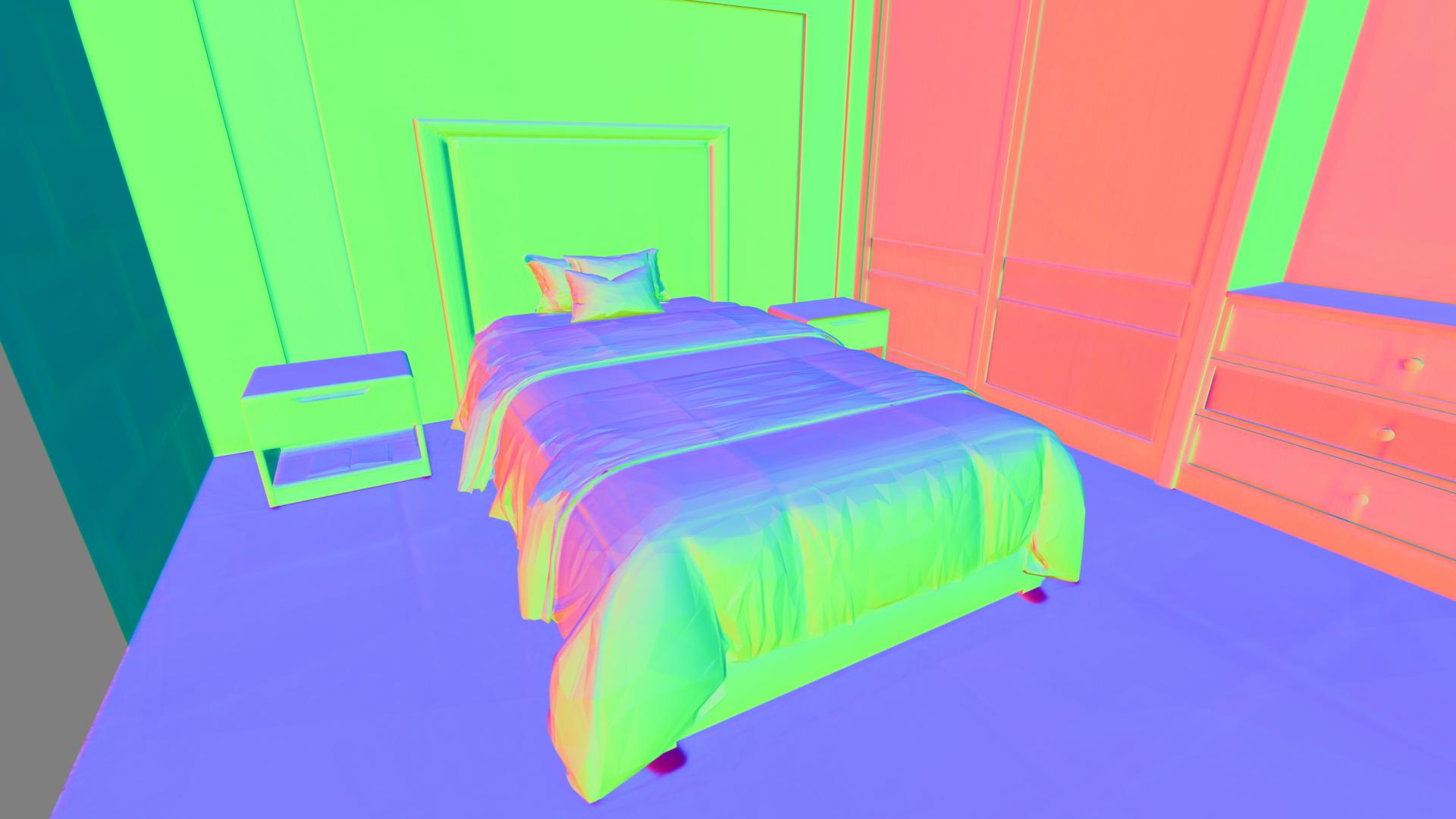}} 
        &
        \fbox{\includegraphics[width=0.15\textwidth,trim={10cm 0 10cm 0},clip]{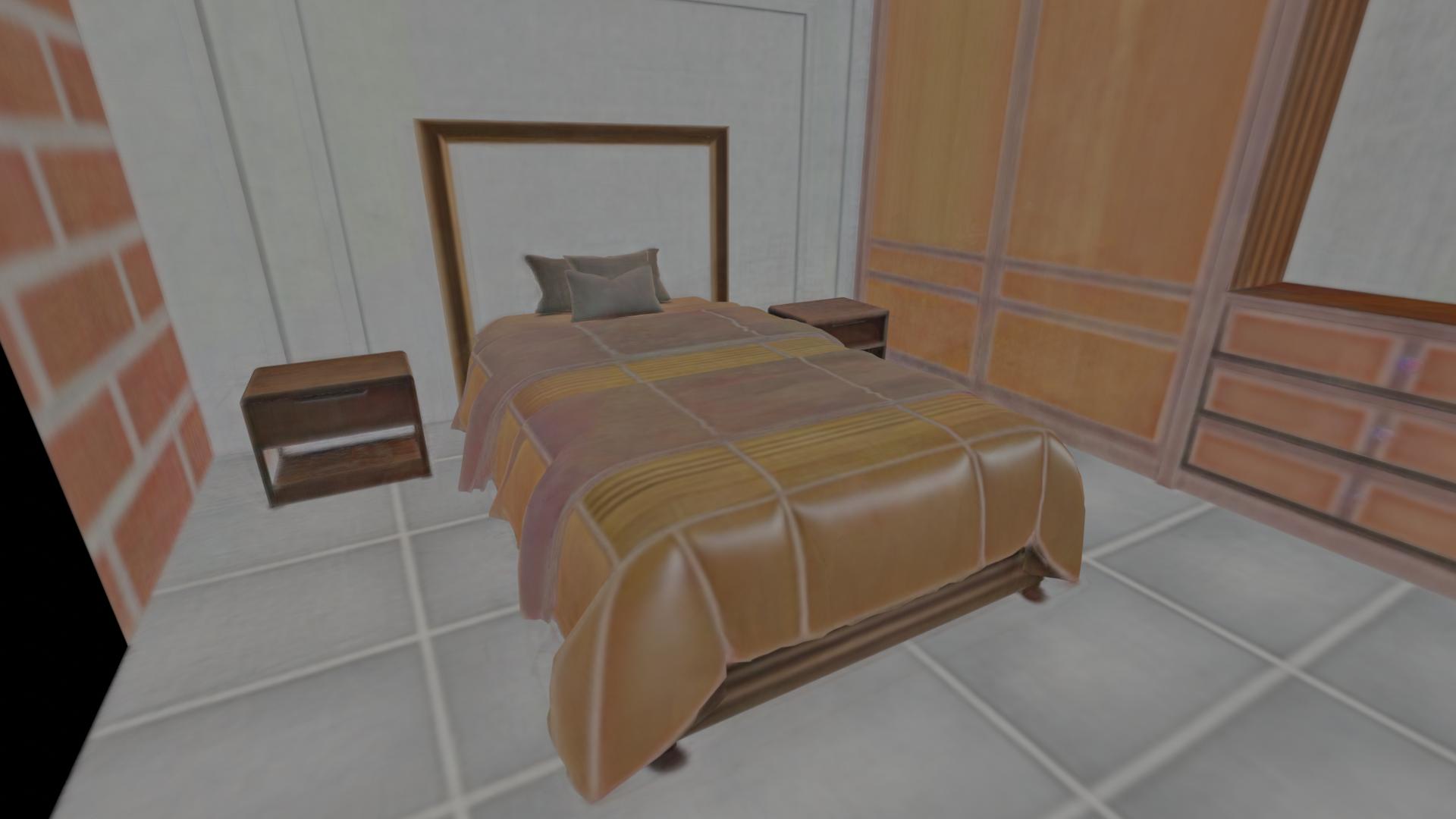}} 
        &
        \fbox{\includegraphics[width=0.15\textwidth,trim={10cm 0 10cm 0},clip]{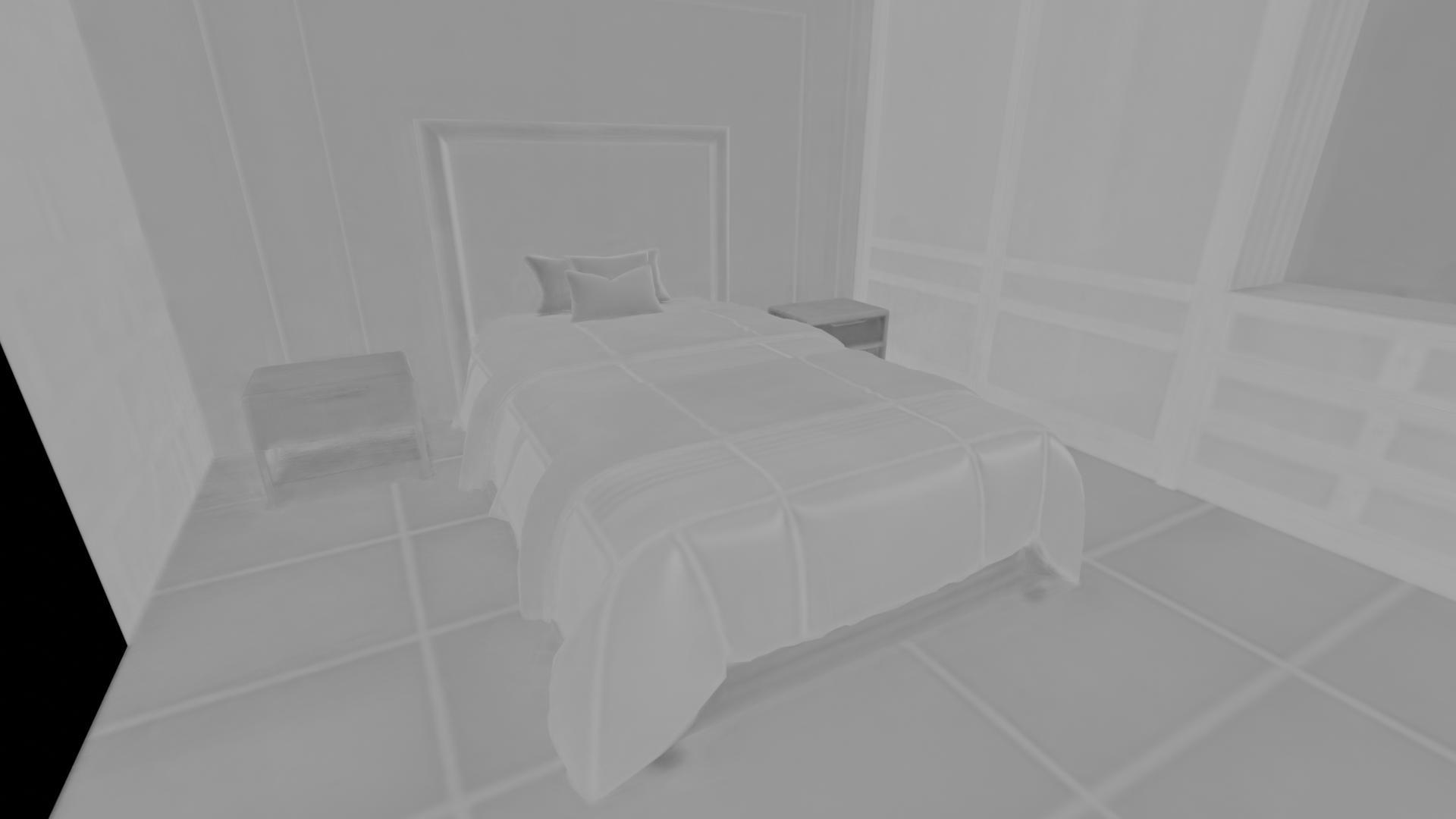}} 
        &
        \fbox{\includegraphics[width=0.15\textwidth,trim={10cm 0 10cm 0},clip]{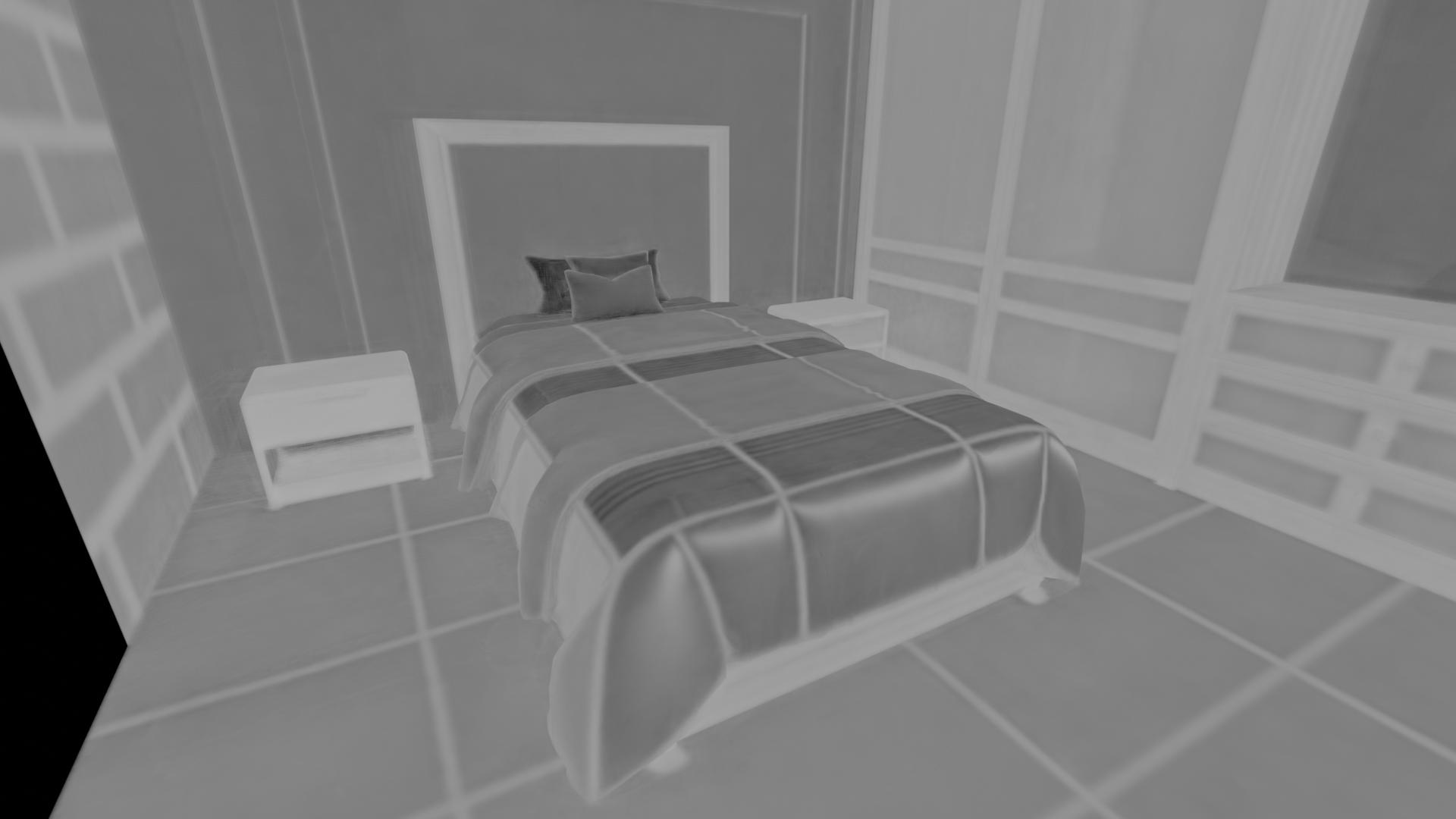}} 
        &
        \fbox{\includegraphics[width=0.15\textwidth,trim={10cm 0 10cm 0},clip]{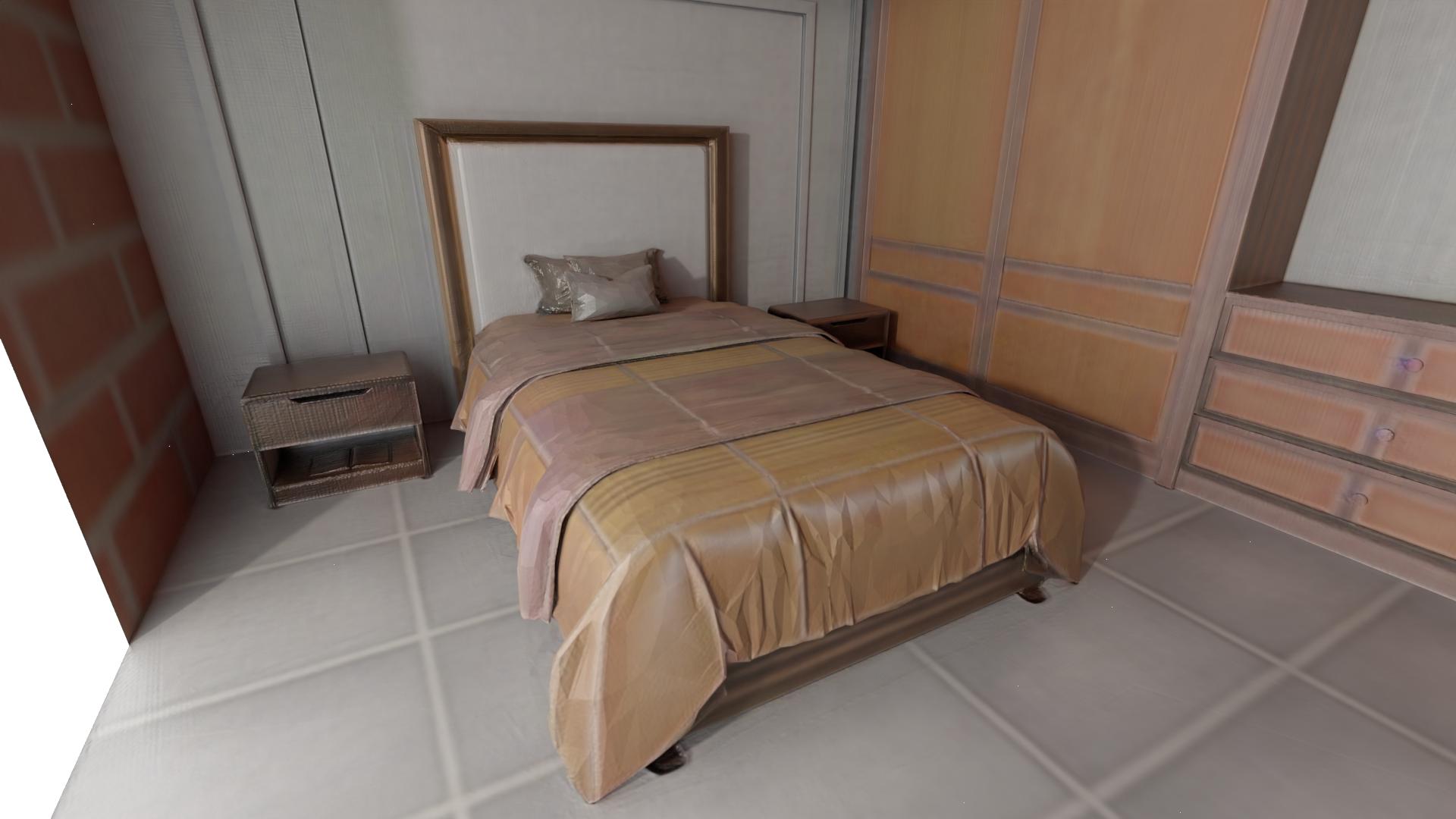}} 
        \\

        &
        {\footnotesize Normal} &
        {\footnotesize Albedo} &
        {\footnotesize Roughness} &
        {\footnotesize Metallic} &
        {\footnotesize Rendering} \\

        \midrule
        
        \rotatebox{90}{{\footnotesize View 1}}
        &
        \fbox{\includegraphics[width=0.15\textwidth,trim={10cm 0 10cm 0},clip]{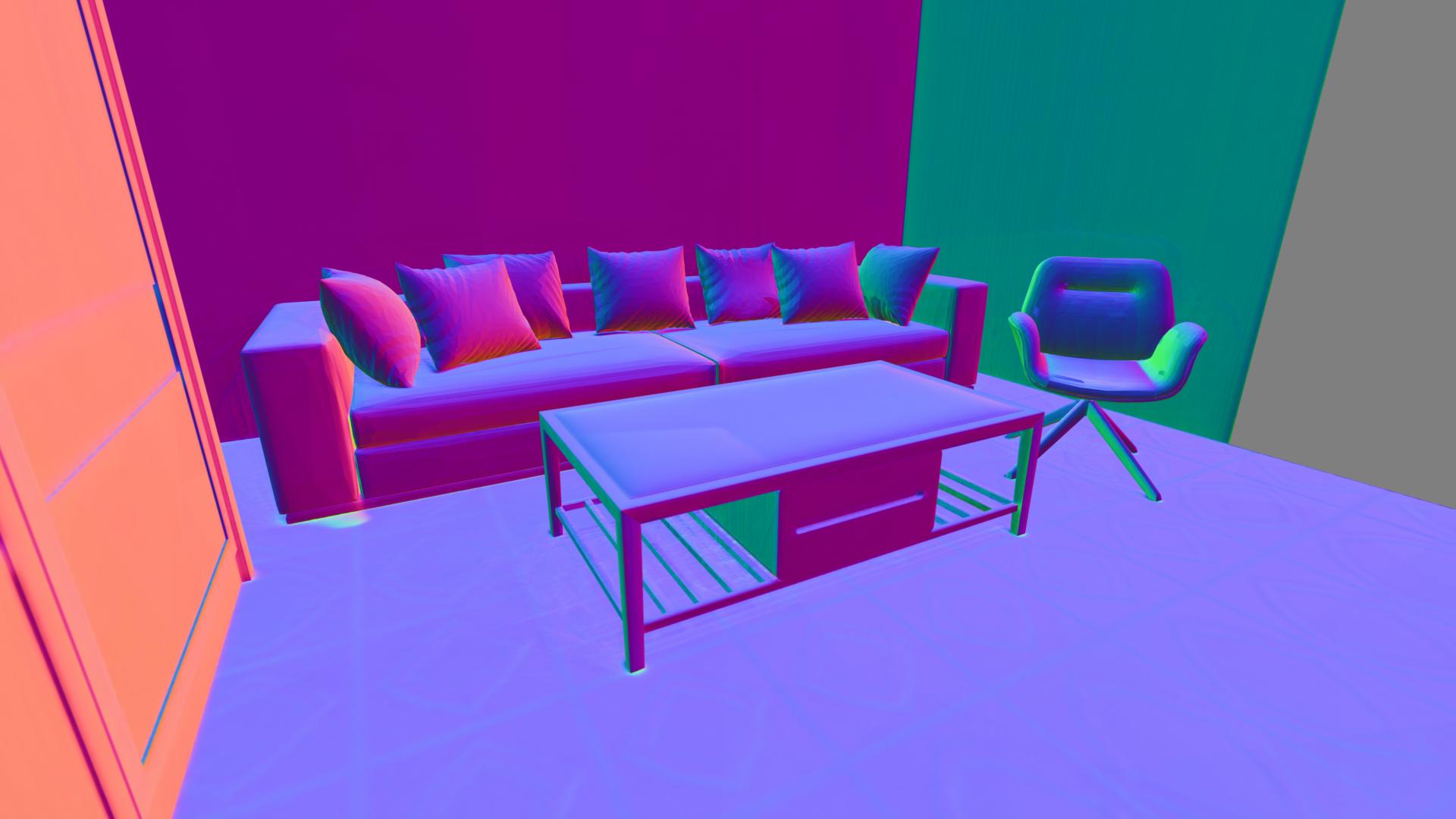}}
        &
        \fbox{\includegraphics[width=0.15\textwidth,trim={10cm 0 10cm 0},clip]{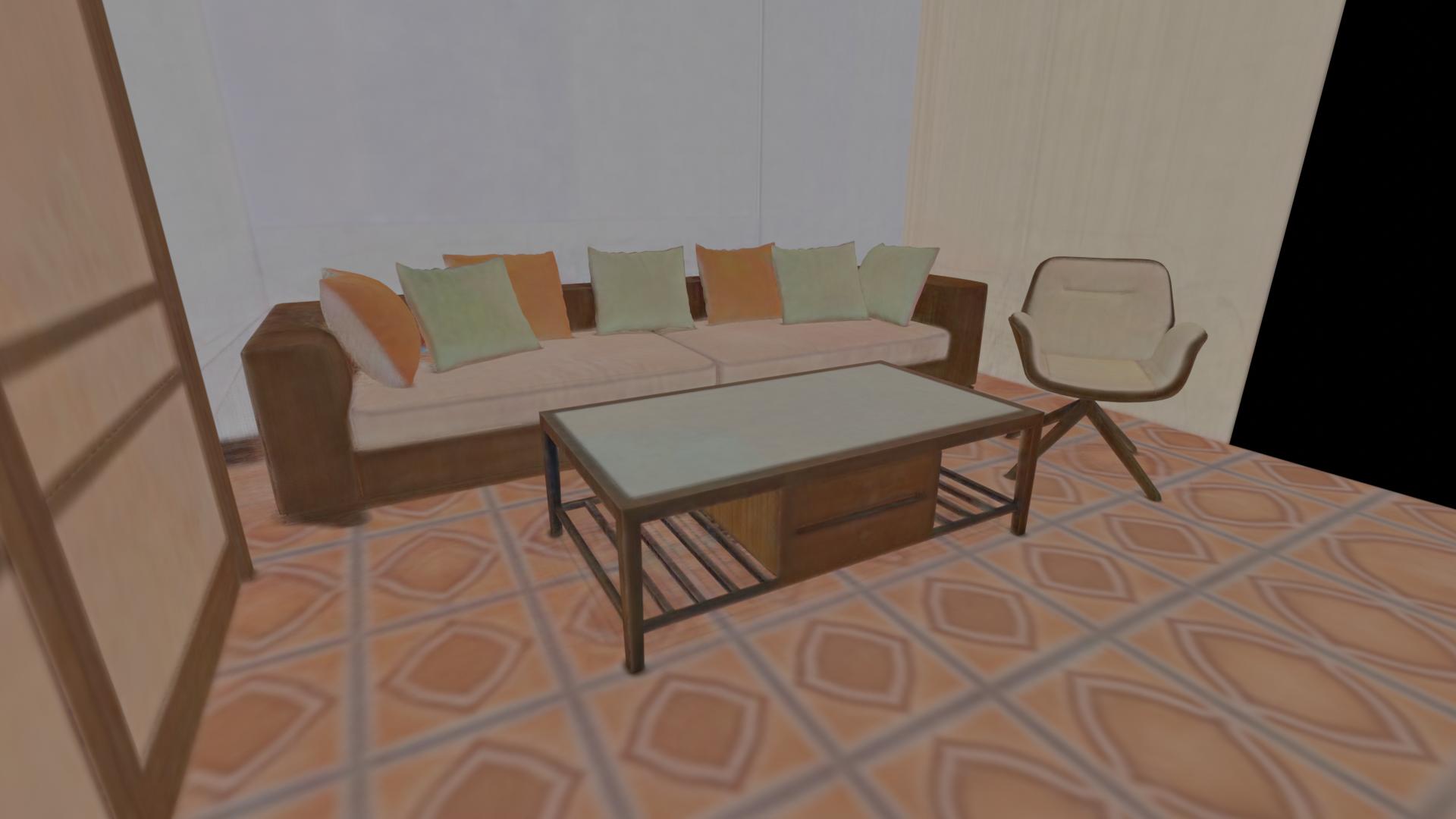}} 
        &
        \fbox{\includegraphics[width=0.15\textwidth,trim={10cm 0 10cm 0},clip]{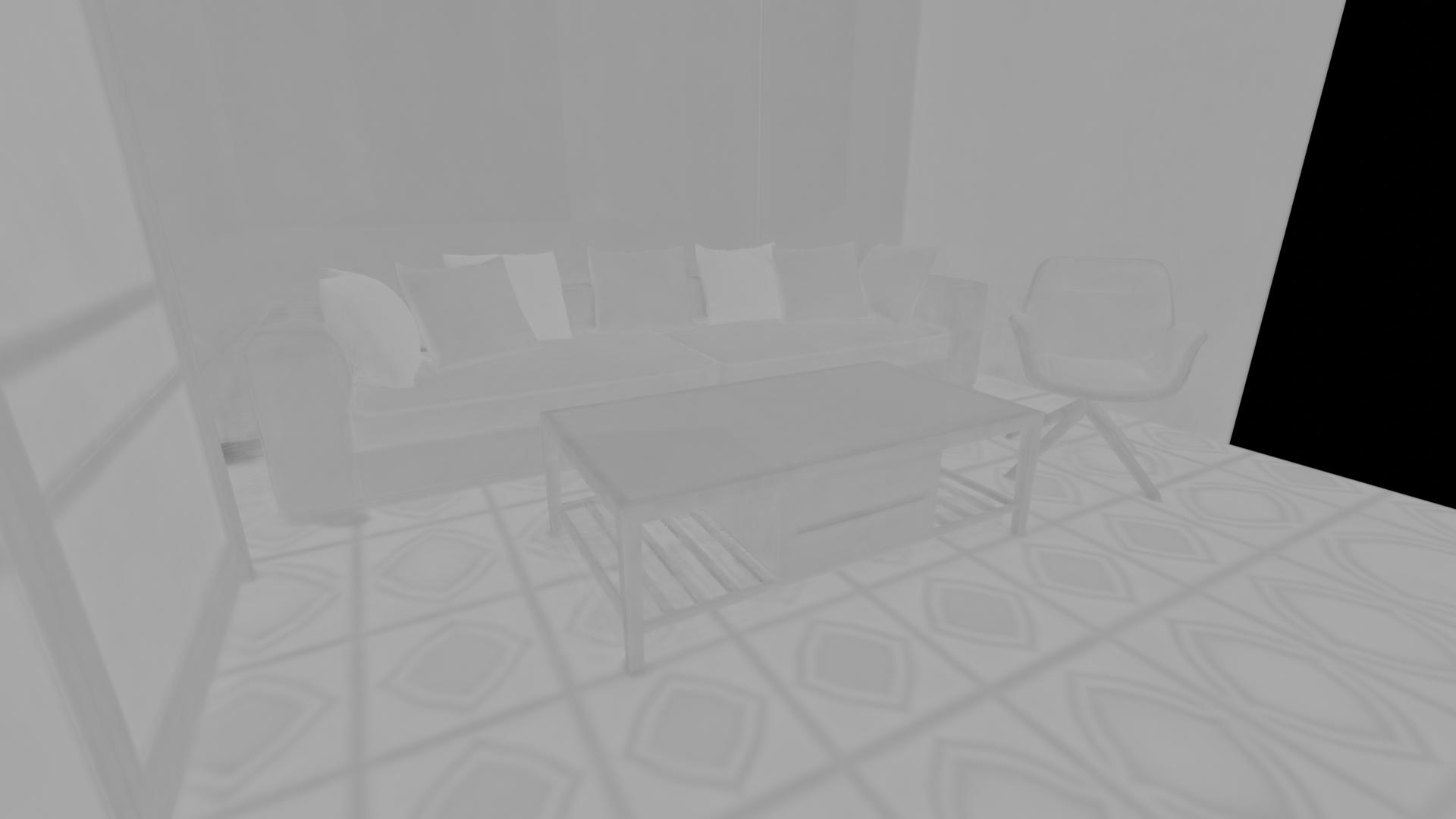}} 
        &
        \fbox{\includegraphics[width=0.15\textwidth,trim={10cm 0 10cm 0},clip]{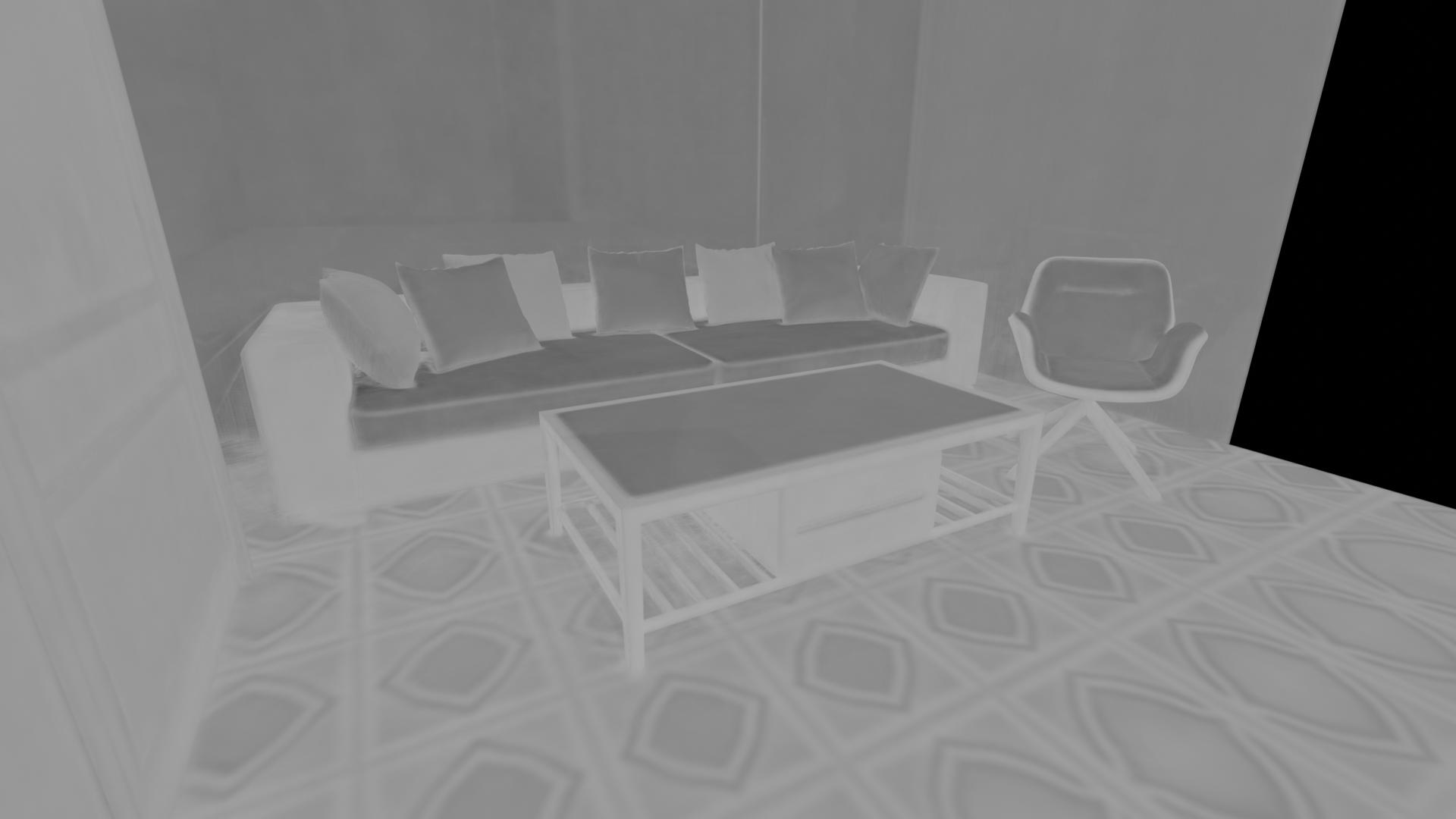}} 
        &
        \fbox{\includegraphics[width=0.15\textwidth,trim={10cm 0 10cm 0},clip]{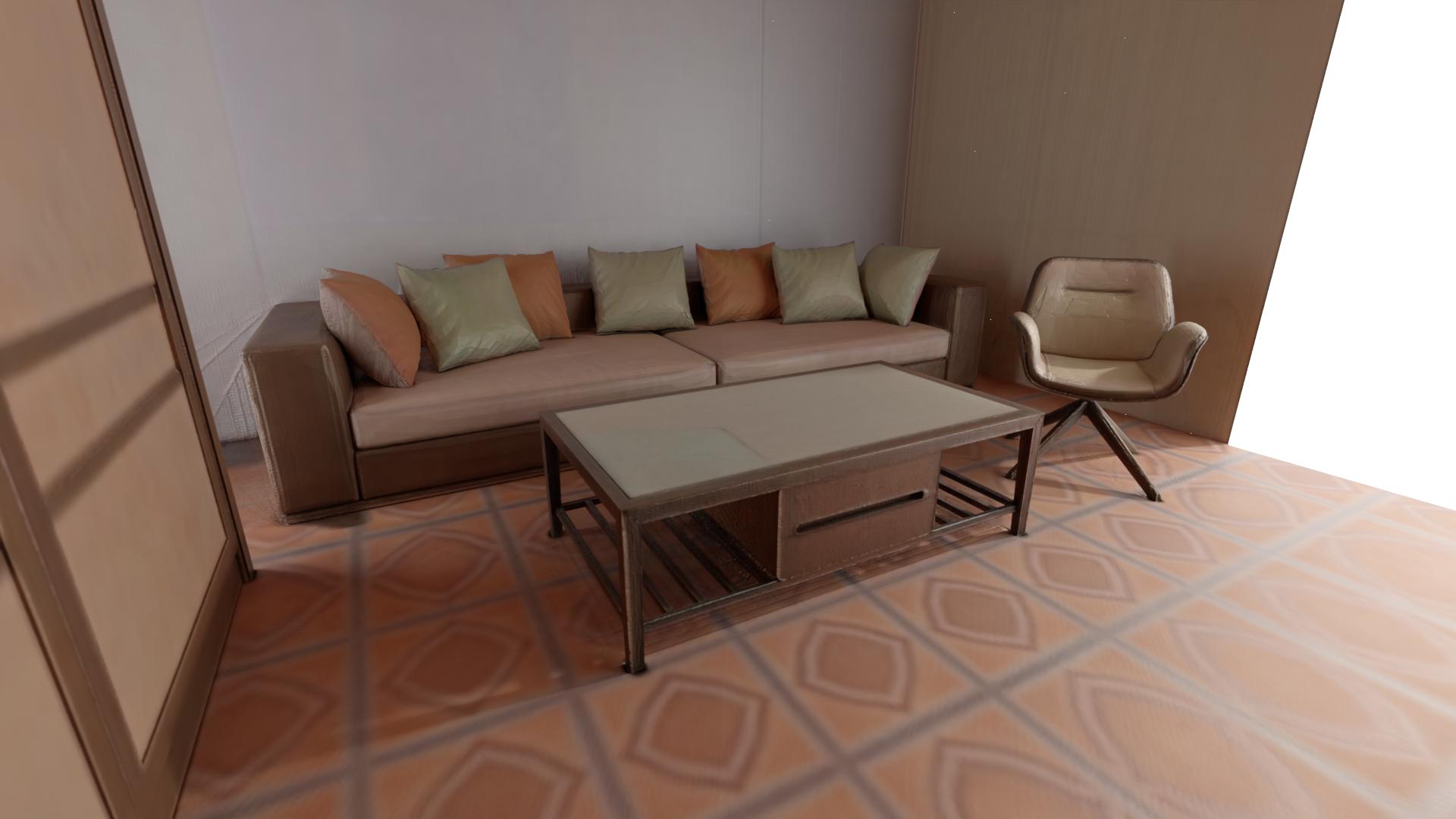}} 
        \\
        
        
        \rotatebox{90}{{\footnotesize View 3}}
        &
        \fbox{\includegraphics[width=0.15\textwidth,trim={10cm 0 10cm 0},clip]{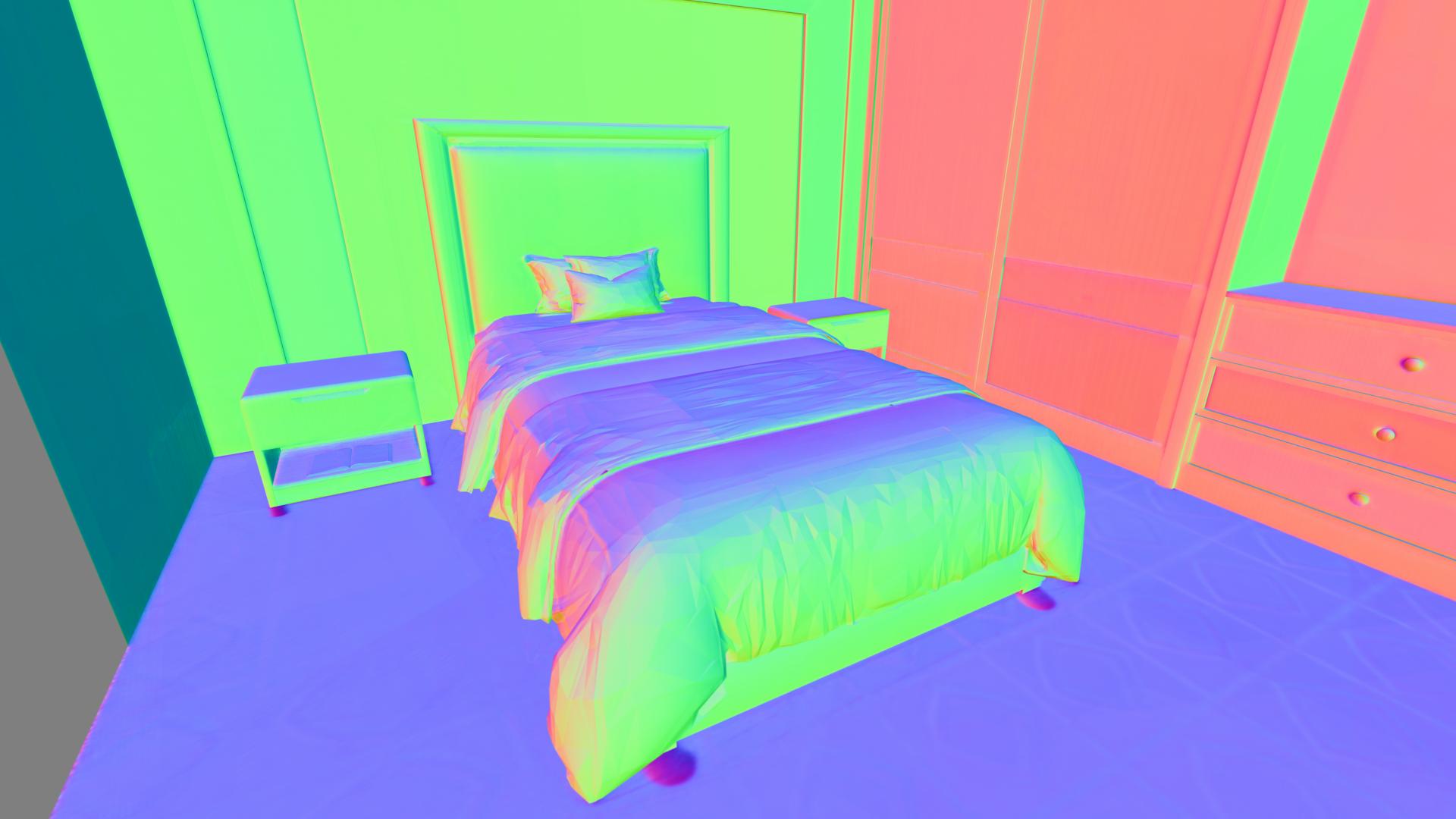}} 
        &
        \fbox{\includegraphics[width=0.15\textwidth,trim={10cm 0 10cm 0},clip]{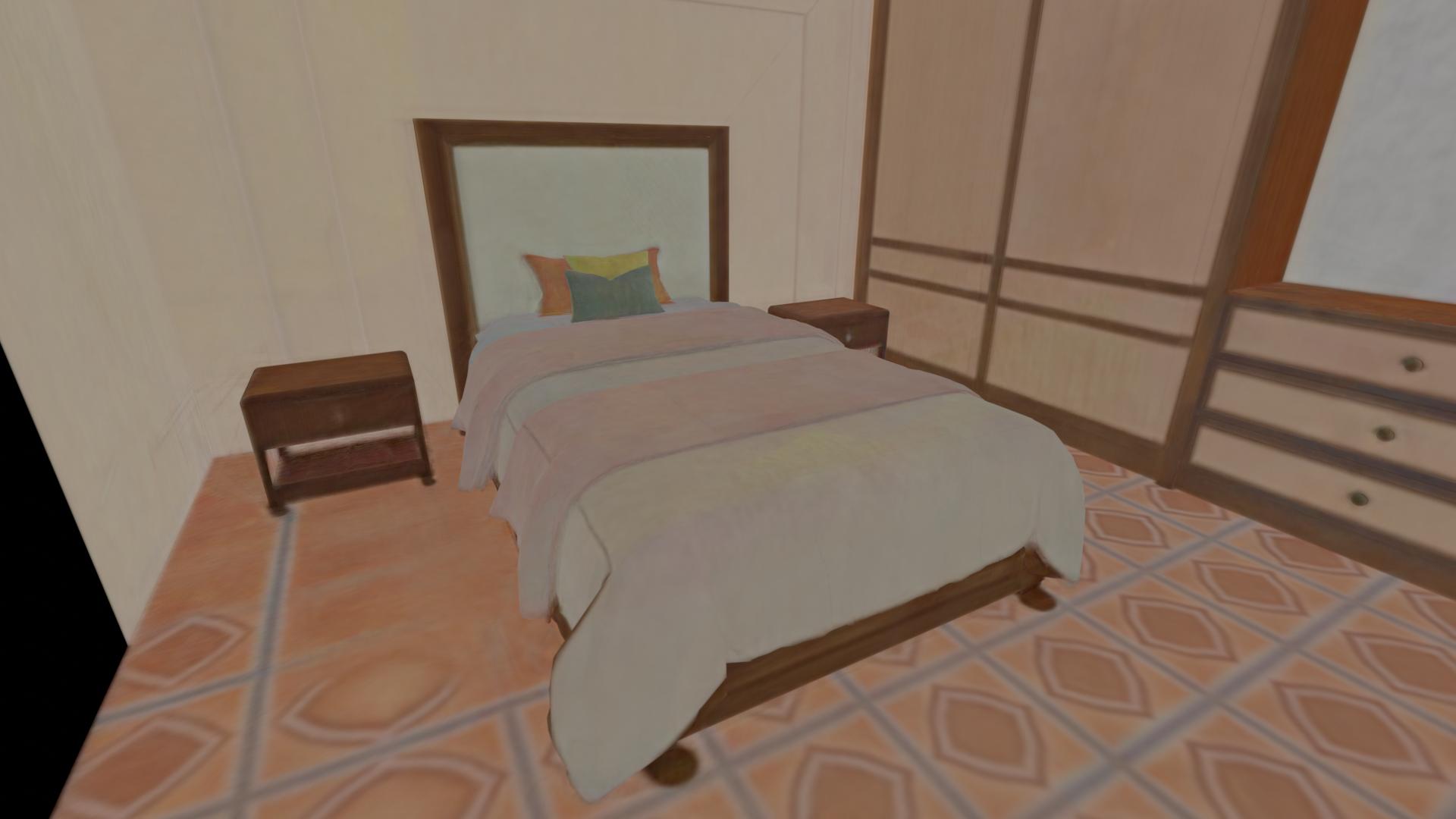}} 
        &
        \fbox{\includegraphics[width=0.15\textwidth,trim={10cm 0 10cm 0},clip]{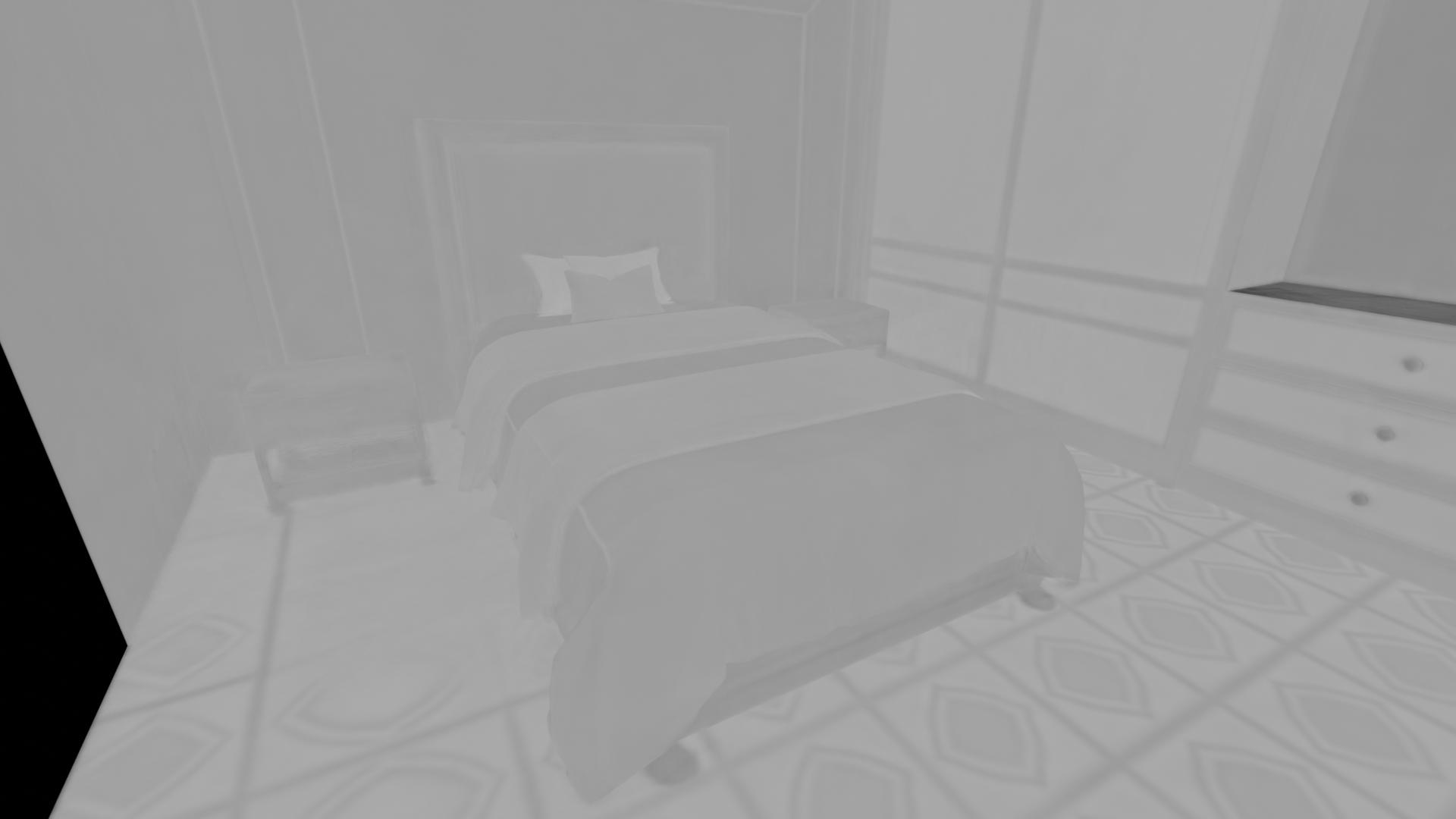}} 
        &
        \fbox{\includegraphics[width=0.15\textwidth,trim={10cm 0 10cm 0},clip]{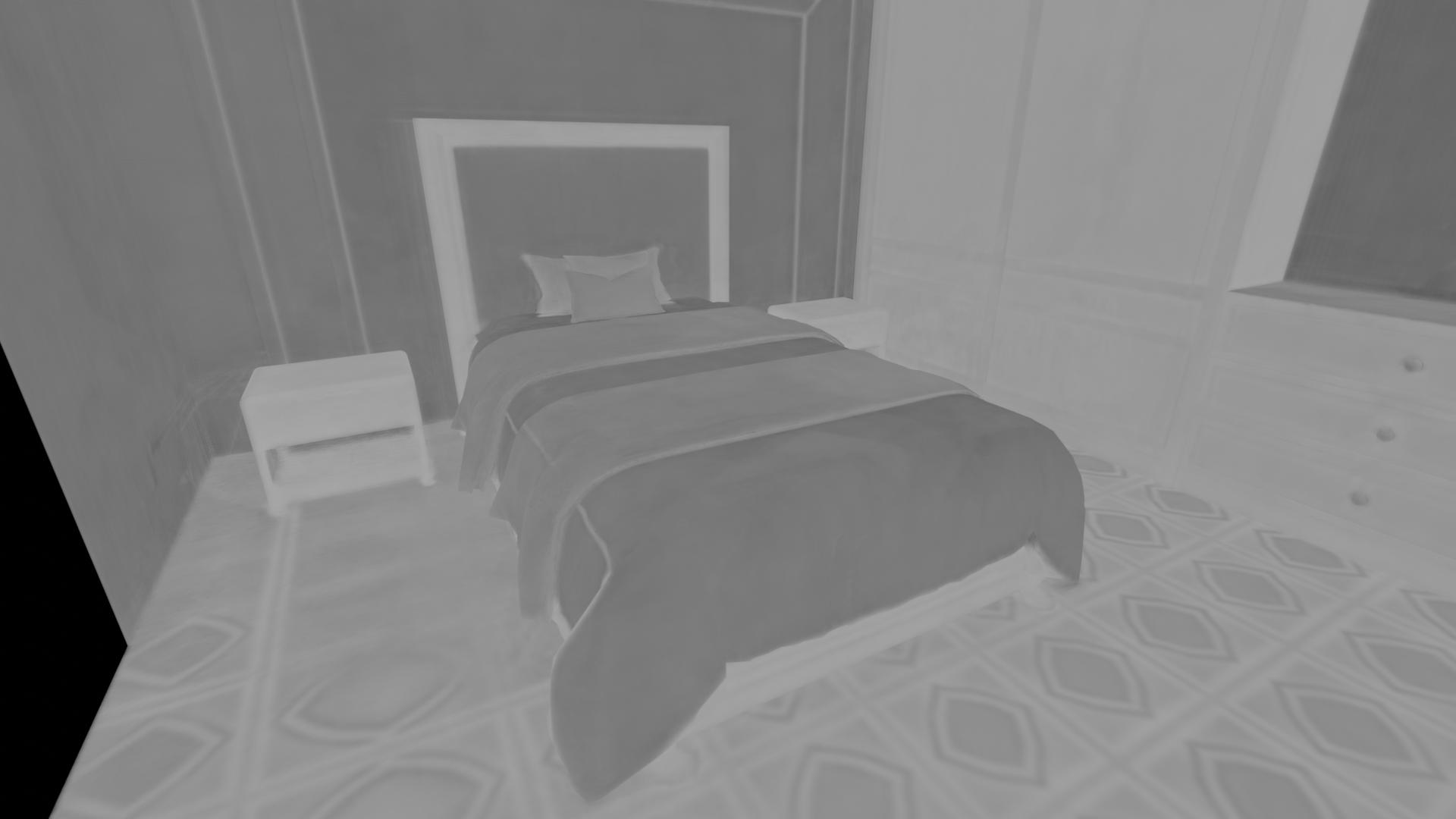}} 
        &
        \fbox{\includegraphics[width=0.15\textwidth,trim={10cm 0 10cm 0},clip]{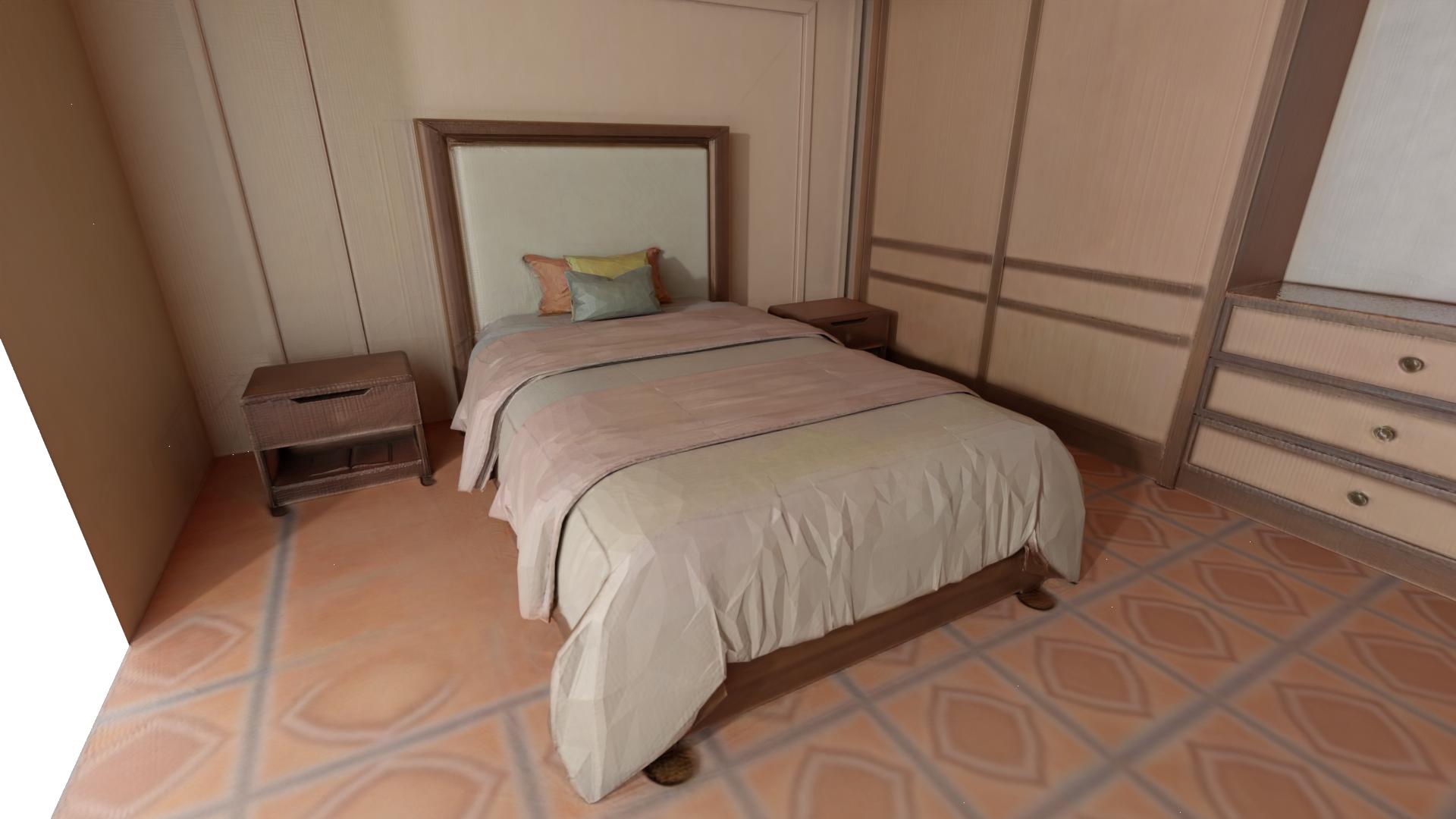}} 
        \\

        &
        {\footnotesize Normal} &
        {\footnotesize Albedo} &
        {\footnotesize Roughness} &
        {\footnotesize Metallic} &
        {\footnotesize Rendering} \\

        \midrule
        
        \rotatebox{90}{{\footnotesize View 1}}
        &
        \fbox{\includegraphics[width=0.15\textwidth,trim={10cm 0 10cm 0},clip]{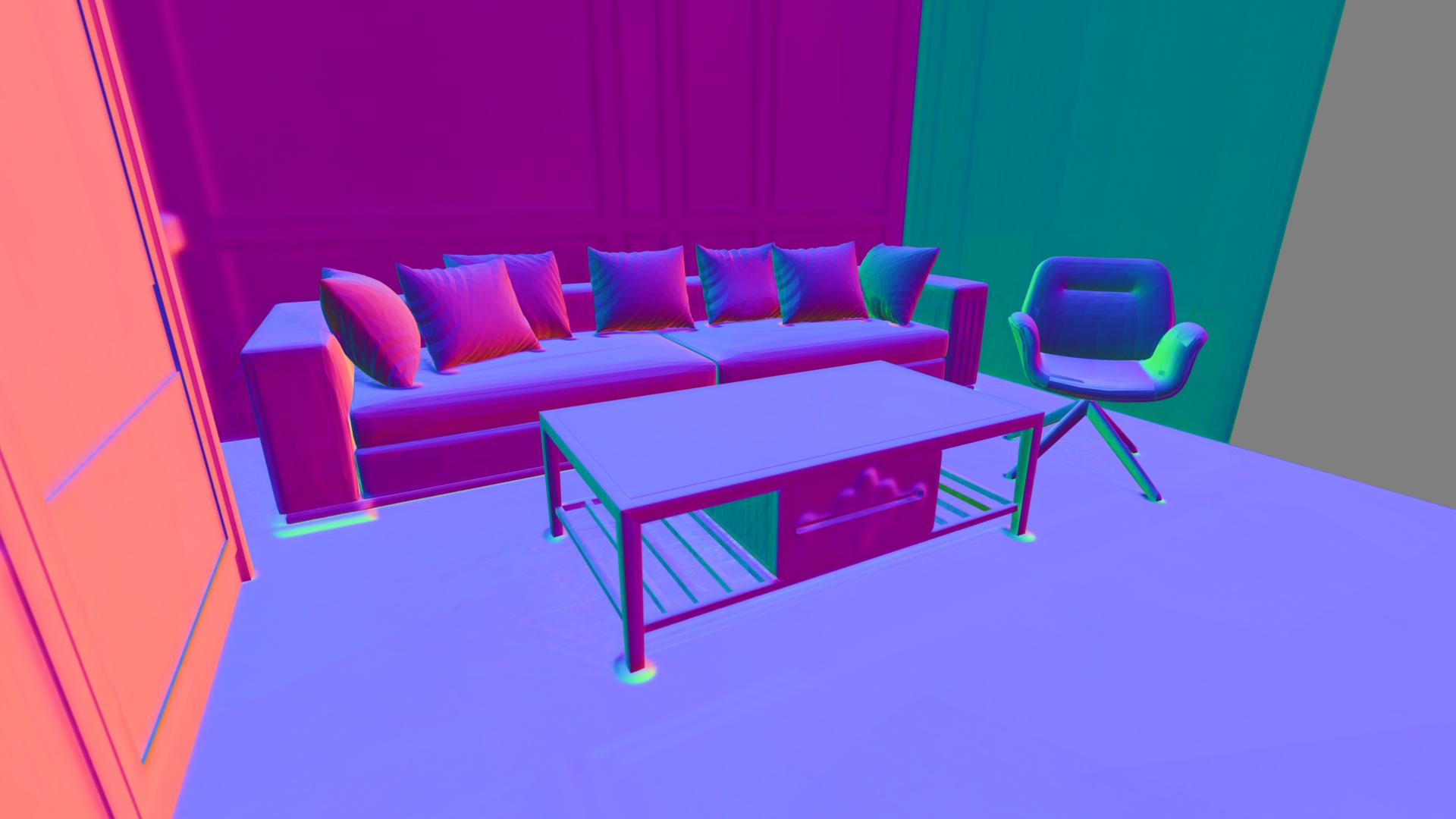}}
        &
        \fbox{\includegraphics[width=0.15\textwidth,trim={10cm 0 10cm 0},clip]{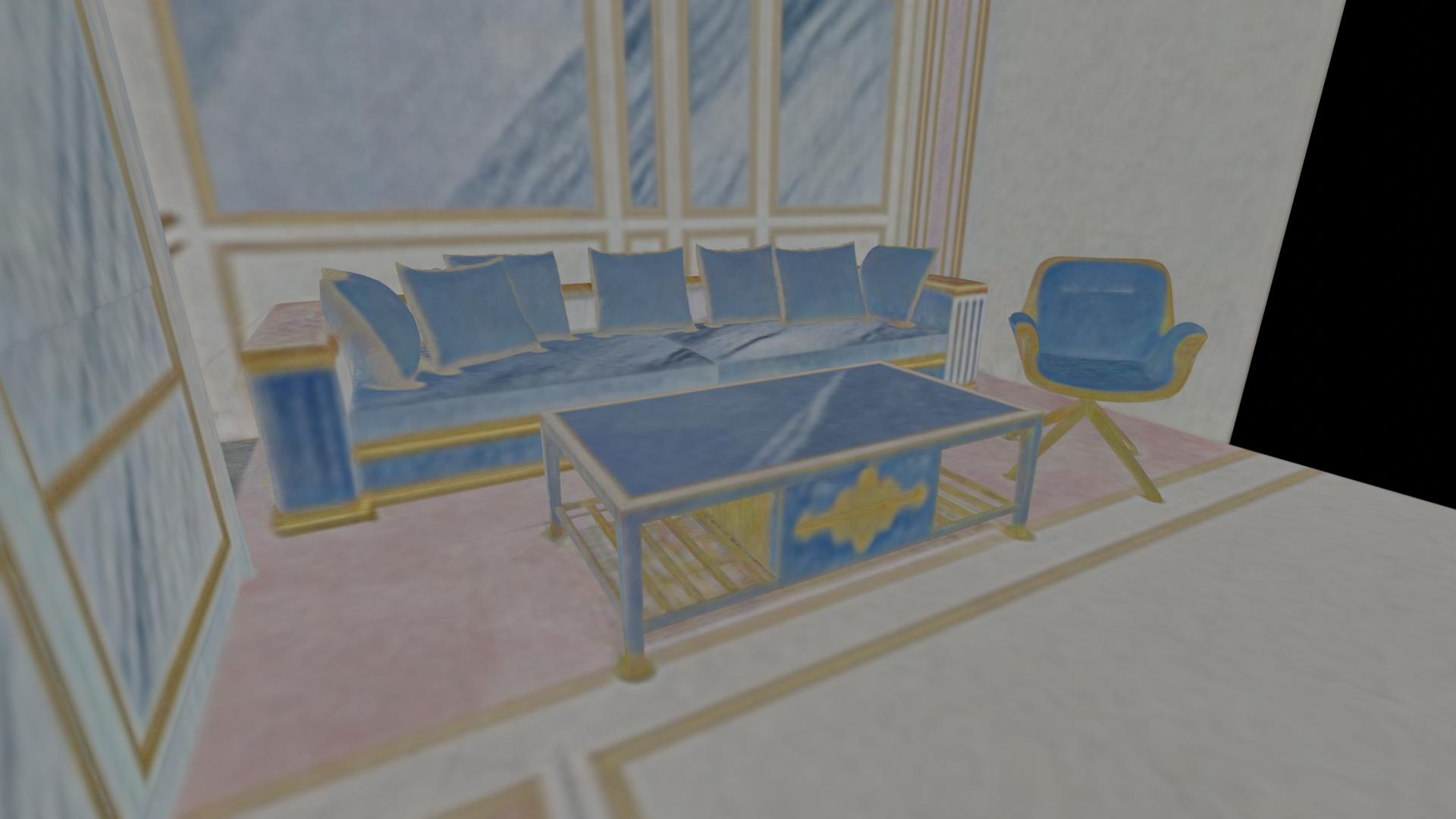}} 
        &
        \fbox{\includegraphics[width=0.15\textwidth,trim={10cm 0 10cm 0},clip]{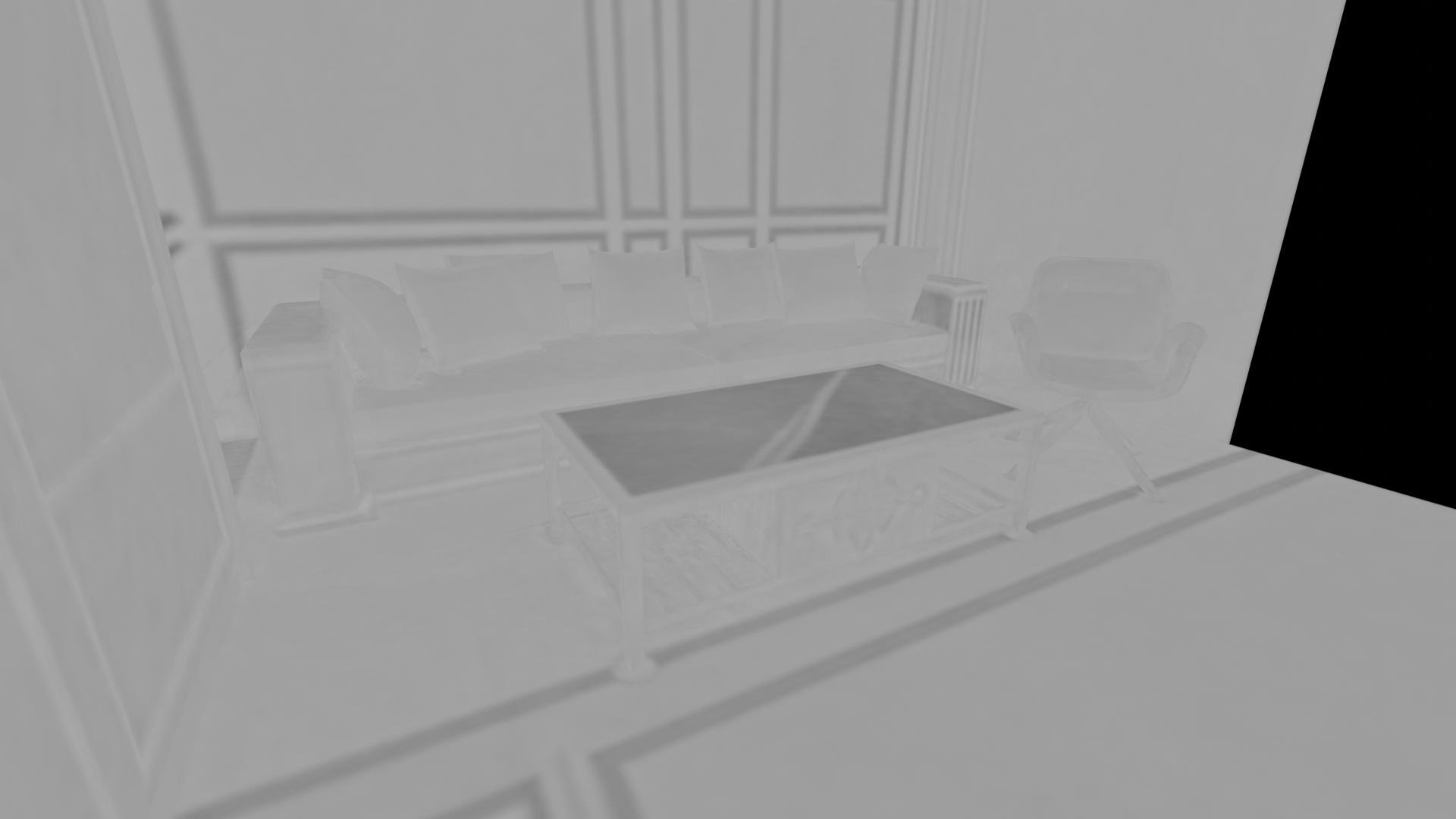}} 
        &
        \fbox{\includegraphics[width=0.15\textwidth,trim={10cm 0 10cm 0},clip]{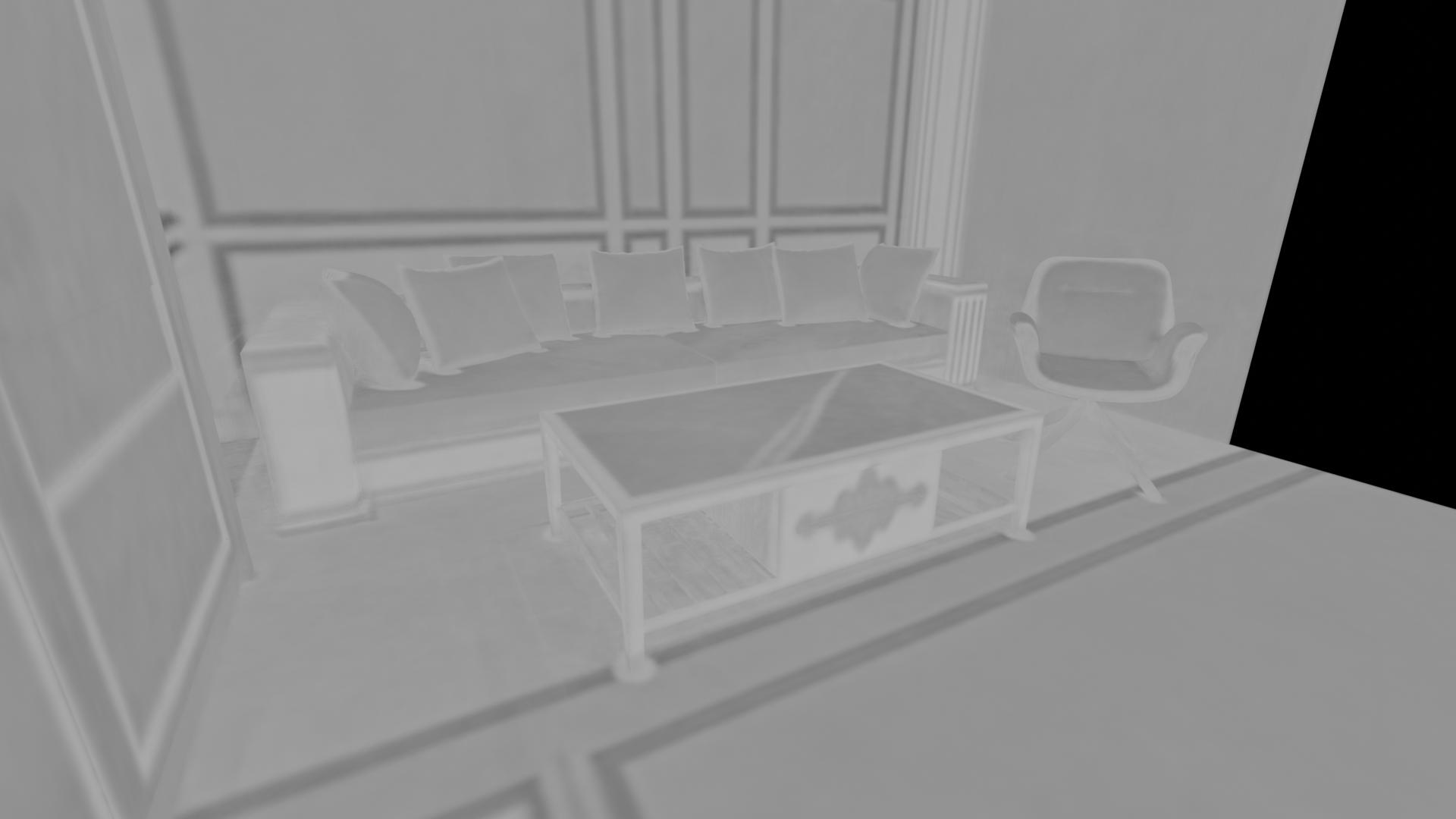}} 
        &
        \fbox{\includegraphics[width=0.15\textwidth,trim={10cm 0 10cm 0},clip]{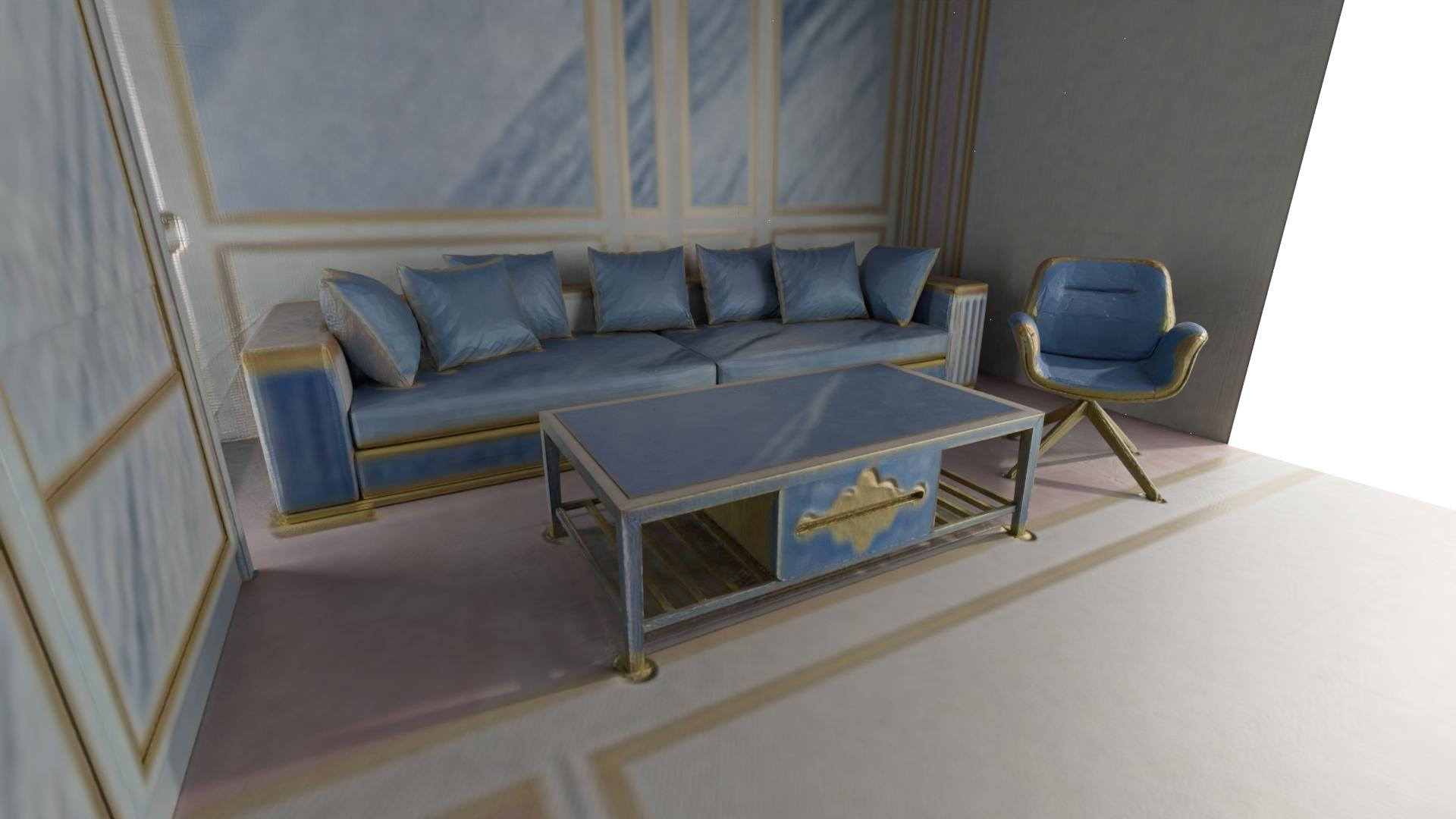}} 
        \\
        
        
        \rotatebox{90}{{\footnotesize View 3}}
        &
        \fbox{\includegraphics[width=0.15\textwidth,trim={10cm 0 10cm 0},clip]{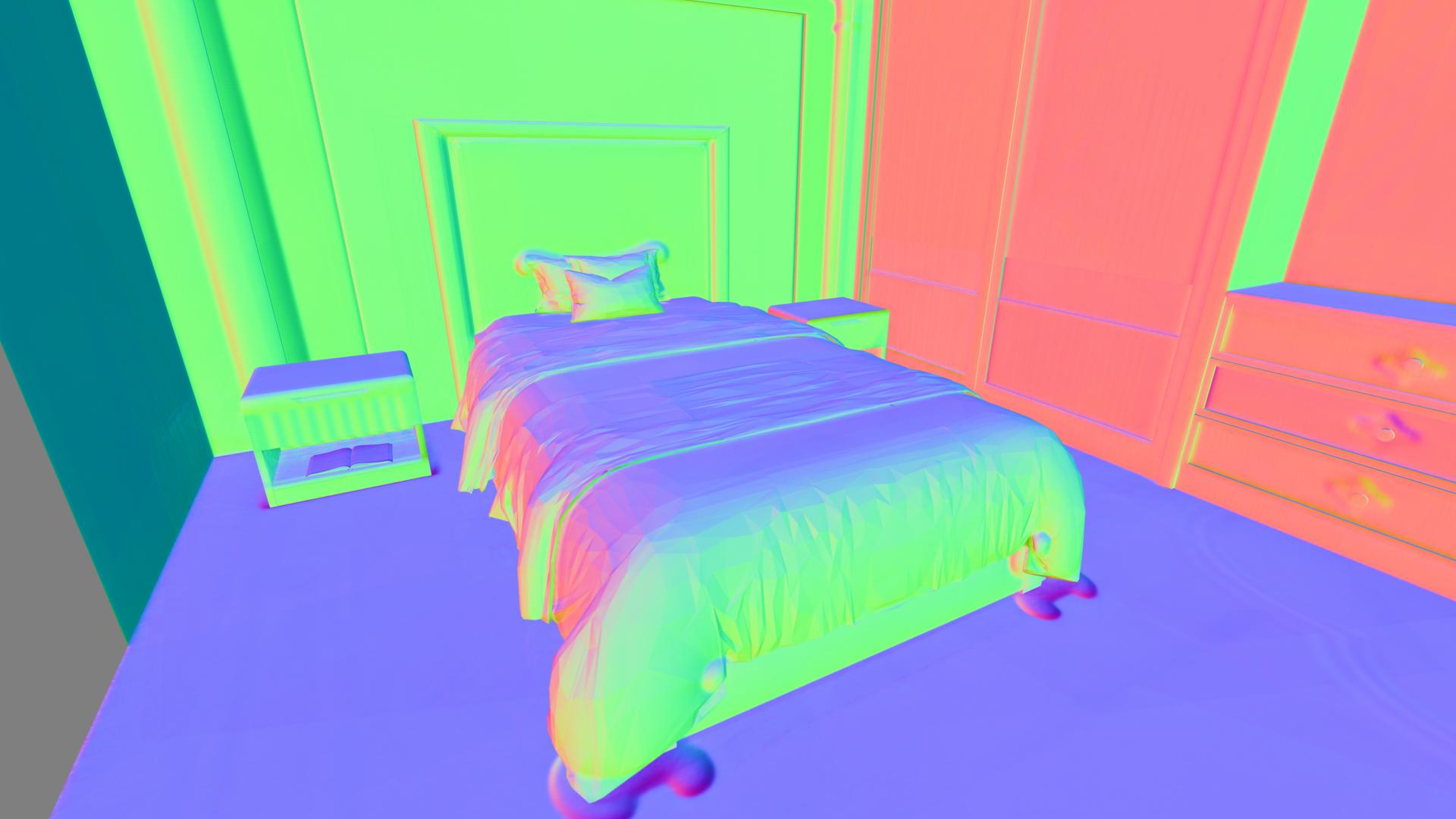}} 
        &
        \fbox{\includegraphics[width=0.15\textwidth,trim={10cm 0 10cm 0},clip]{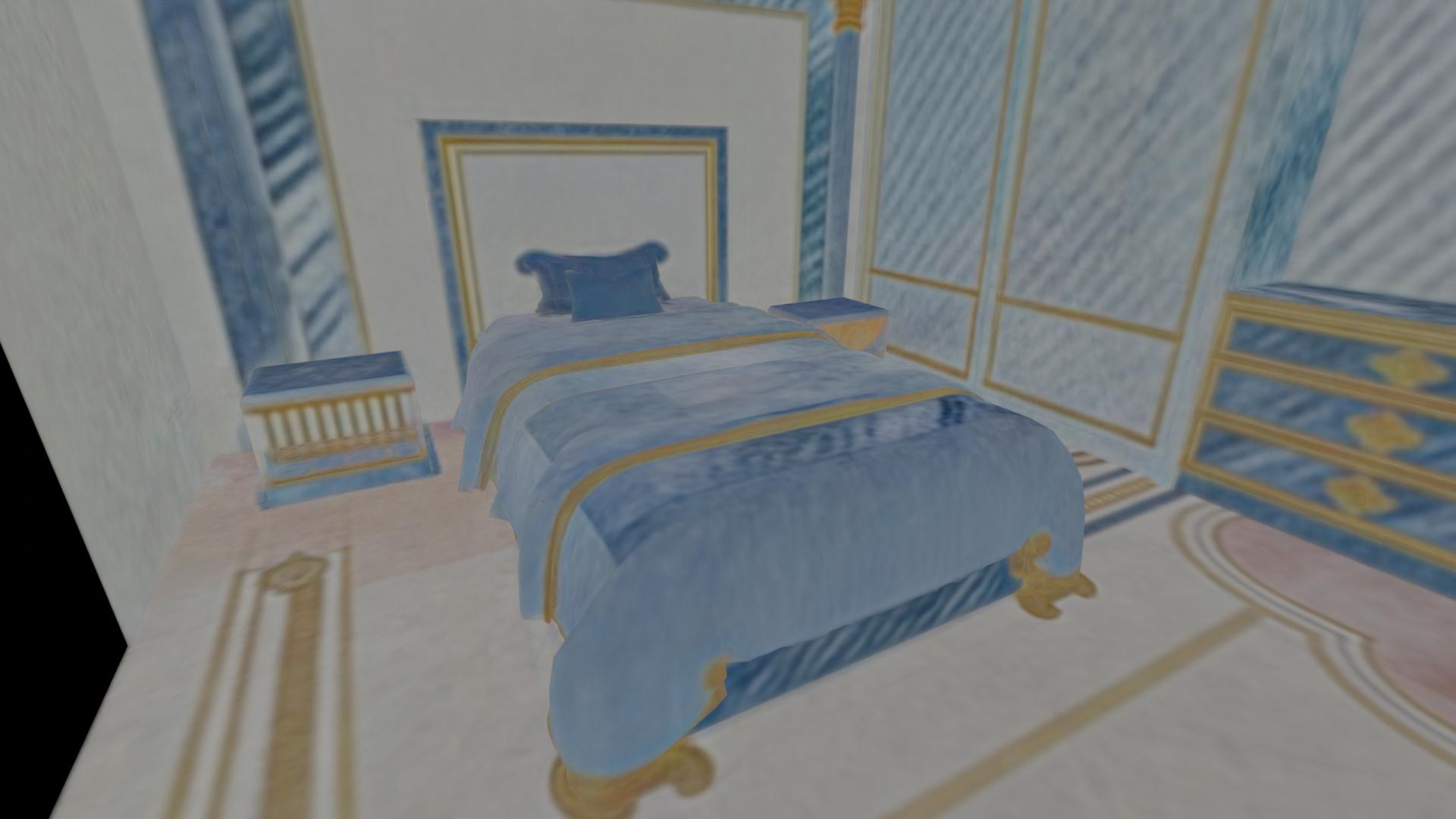}} 
        &
        \fbox{\includegraphics[width=0.15\textwidth,trim={10cm 0 10cm 0},clip]{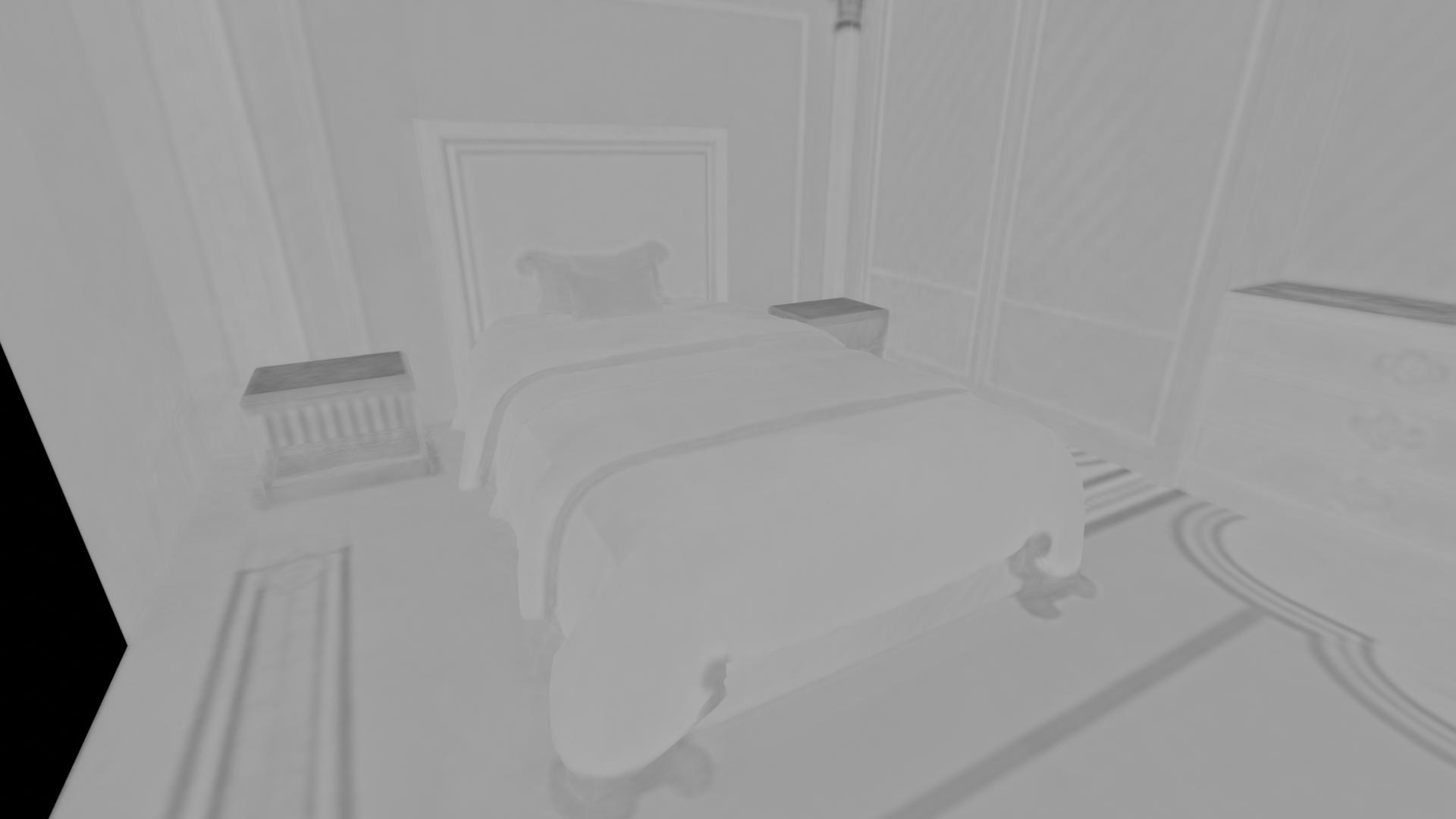}} 
        &
        \fbox{\includegraphics[width=0.15\textwidth,trim={10cm 0 10cm 0},clip]{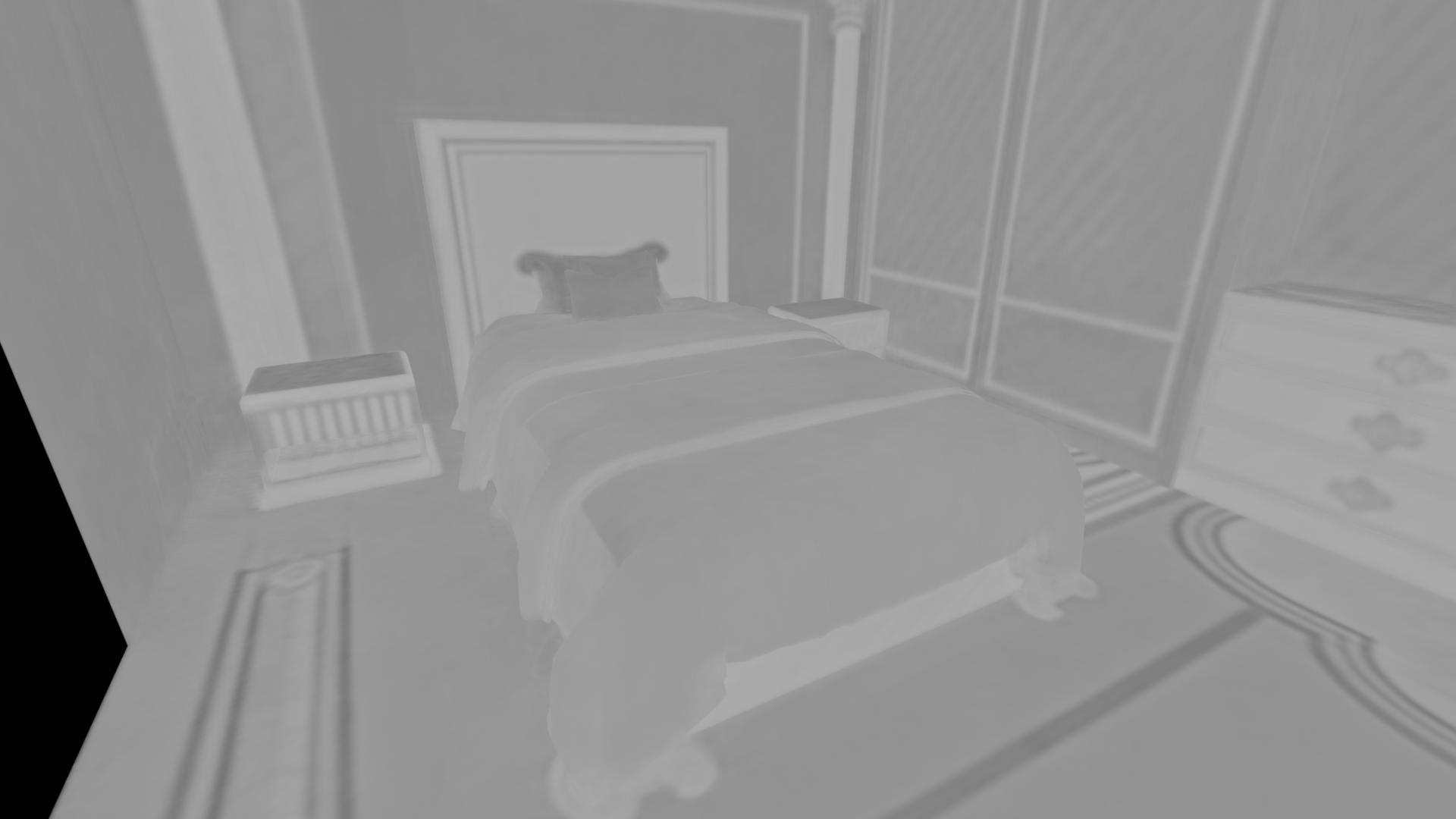}} 
        &
        \fbox{\includegraphics[width=0.15\textwidth,trim={10cm 0 10cm 0},clip]{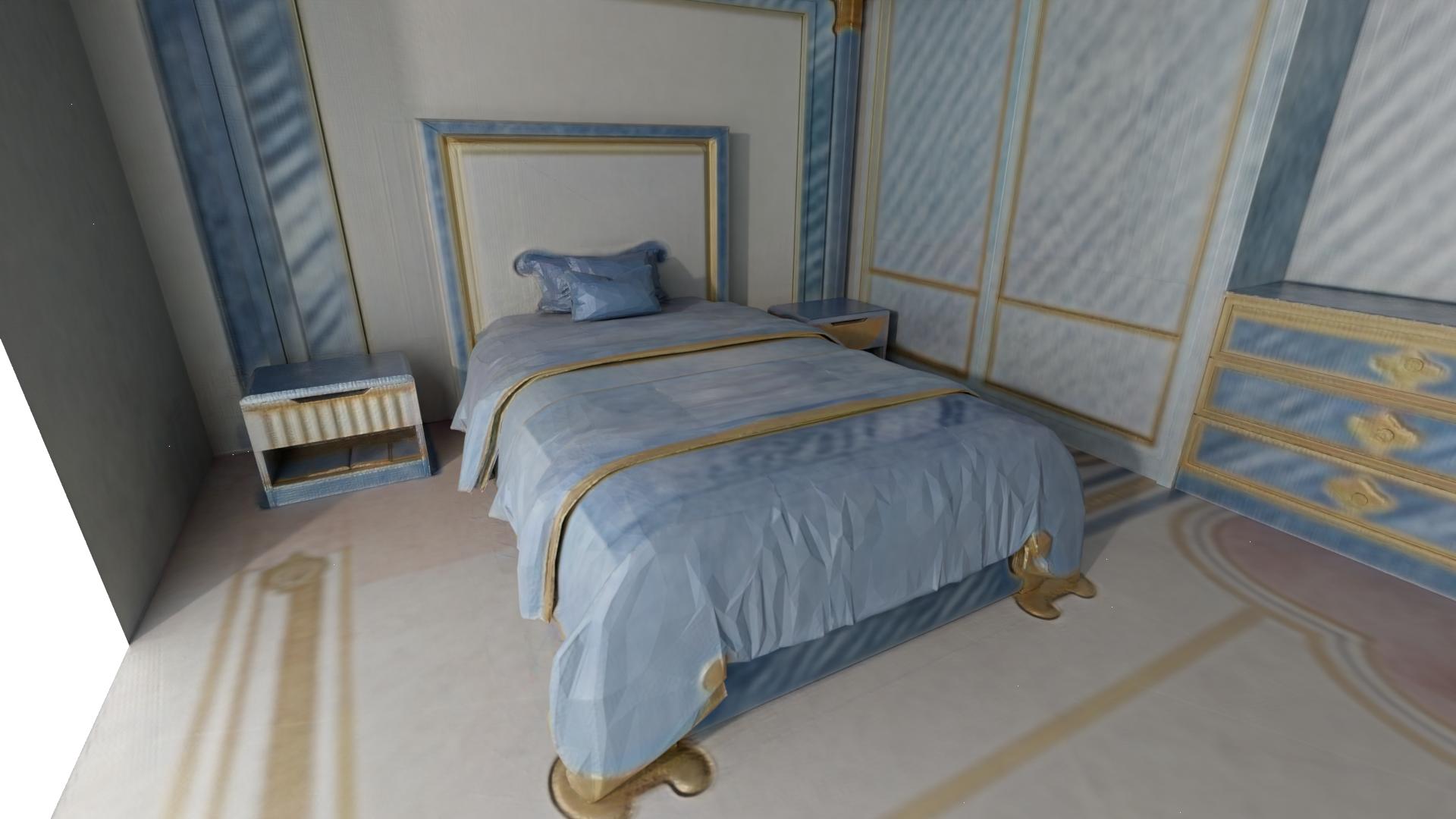}} 
        \\

        &
        {\footnotesize Normal} &
        {\footnotesize Albedo} &
        {\footnotesize Roughness} &
        {\footnotesize Metallic} &
        {\footnotesize Rendering} \\
    \end{tabular}}
    \caption{\textbf{Scene Texturing}. 
    We show more scene texturing results on multiple 3D-Front scenes \cite{Front3d} with multiple prompts. Continues on the next page.
    }
\end{figure*}\begin{figure*}[p]
    \ContinuedFloat
    \centering
    \setlength\tabcolsep{1.25pt}
    \resizebox{\textwidth}{!}{
    \fboxsep=0pt
    \begin{tabular}{ccccc|c}
        \rotatebox{90}{{\footnotesize View 1}}
        &
        \fbox{\includegraphics[width=0.15\textwidth,trim={10cm 0 10cm 0},clip]{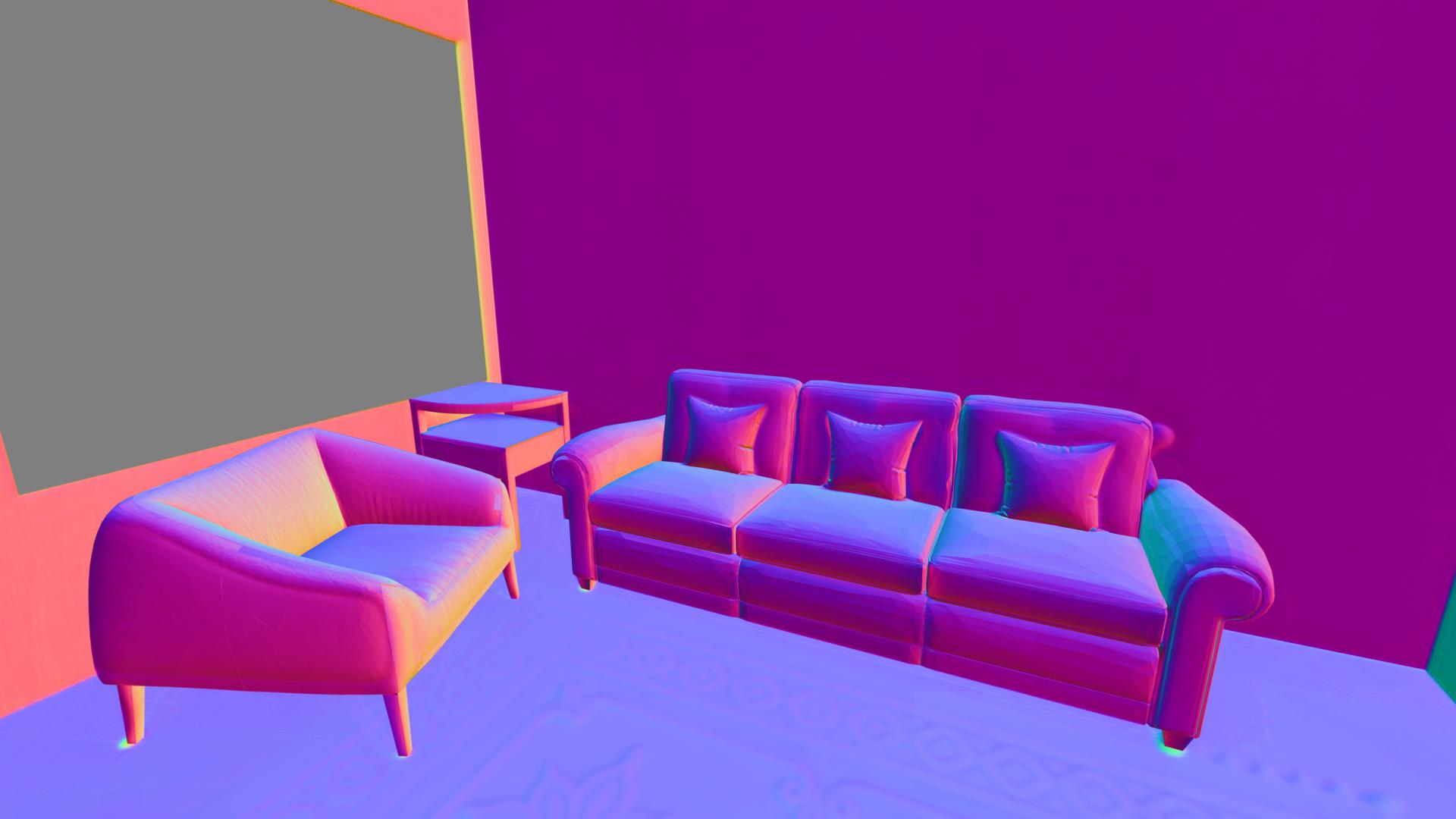}}
        &
        \fbox{\includegraphics[width=0.15\textwidth,trim={10cm 0 10cm 0},clip]{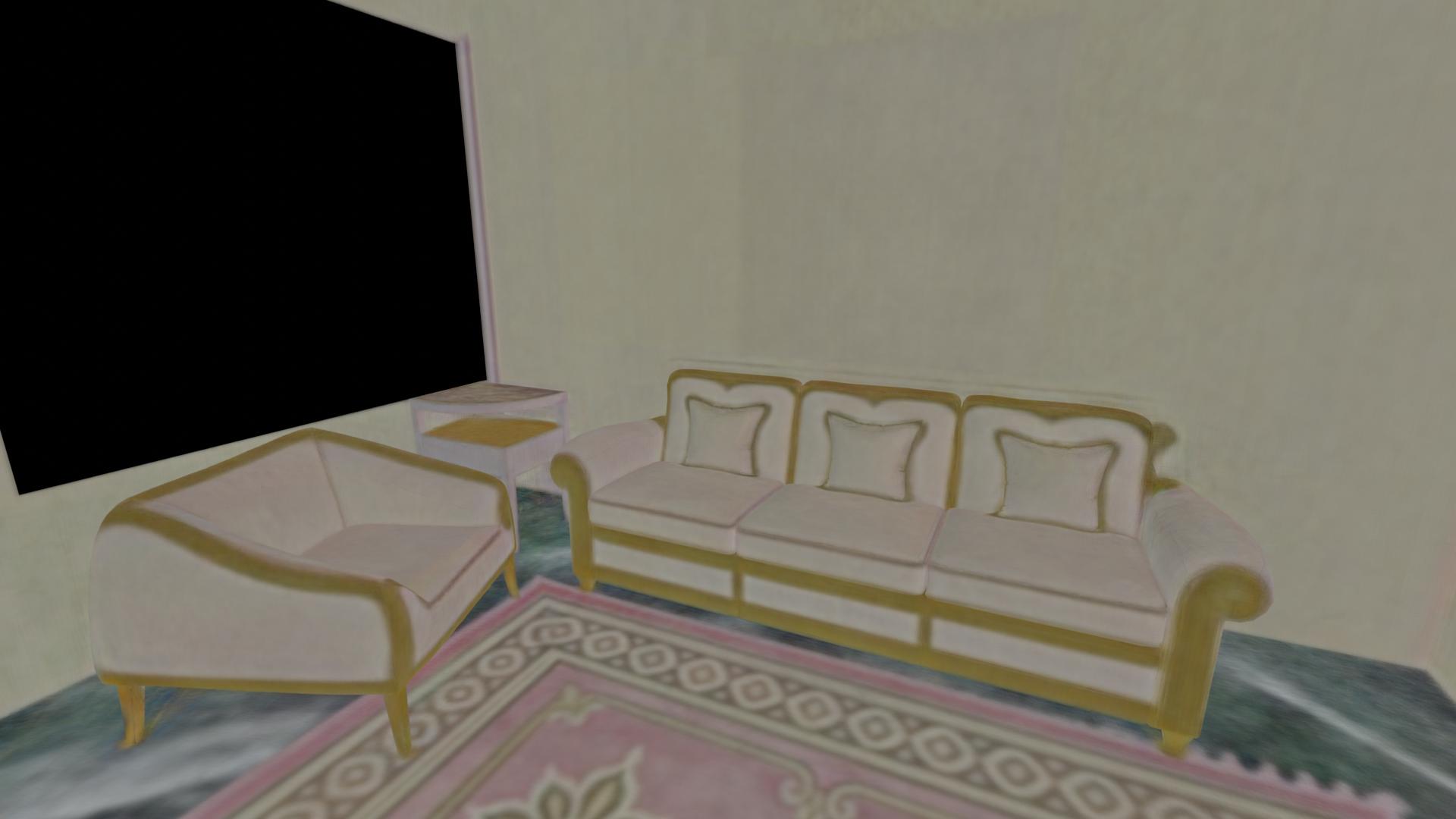}} 
        &
        \fbox{\includegraphics[width=0.15\textwidth,trim={10cm 0 10cm 0},clip]{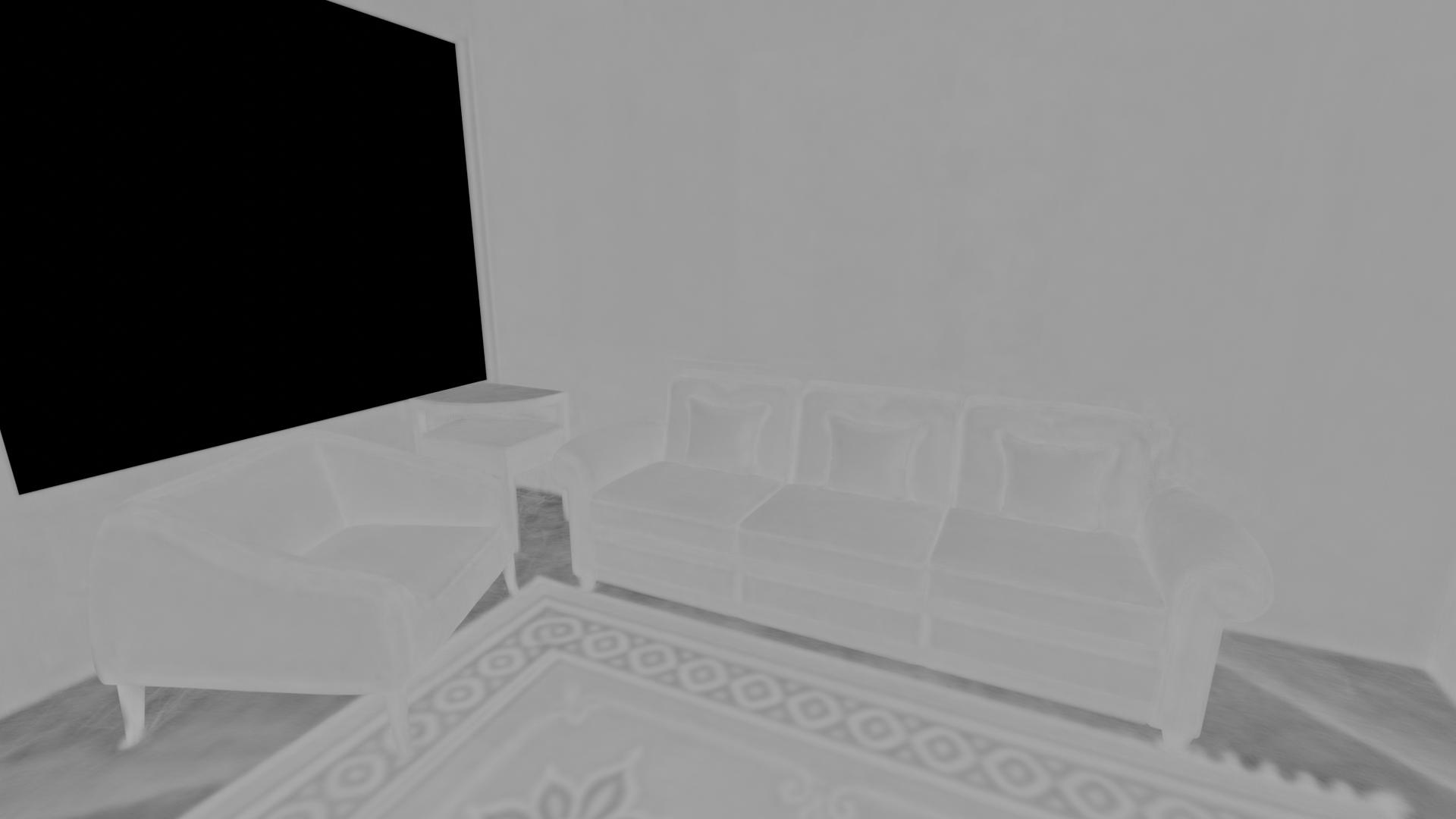}} 
        &
        \fbox{\includegraphics[width=0.15\textwidth,trim={10cm 0 10cm 0},clip]{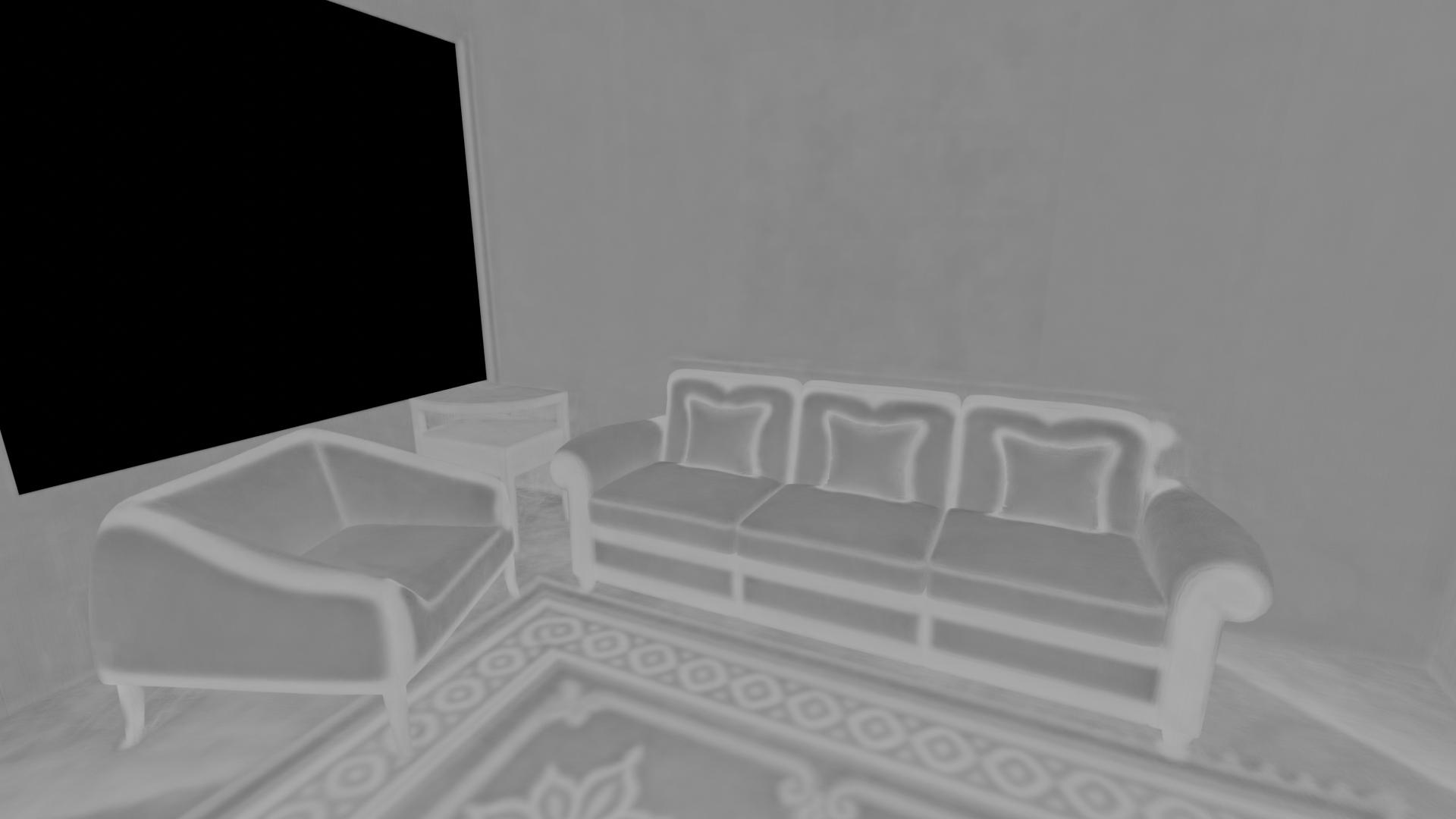}} 
        &
        \fbox{\includegraphics[width=0.15\textwidth,trim={10cm 0 10cm 0},clip]{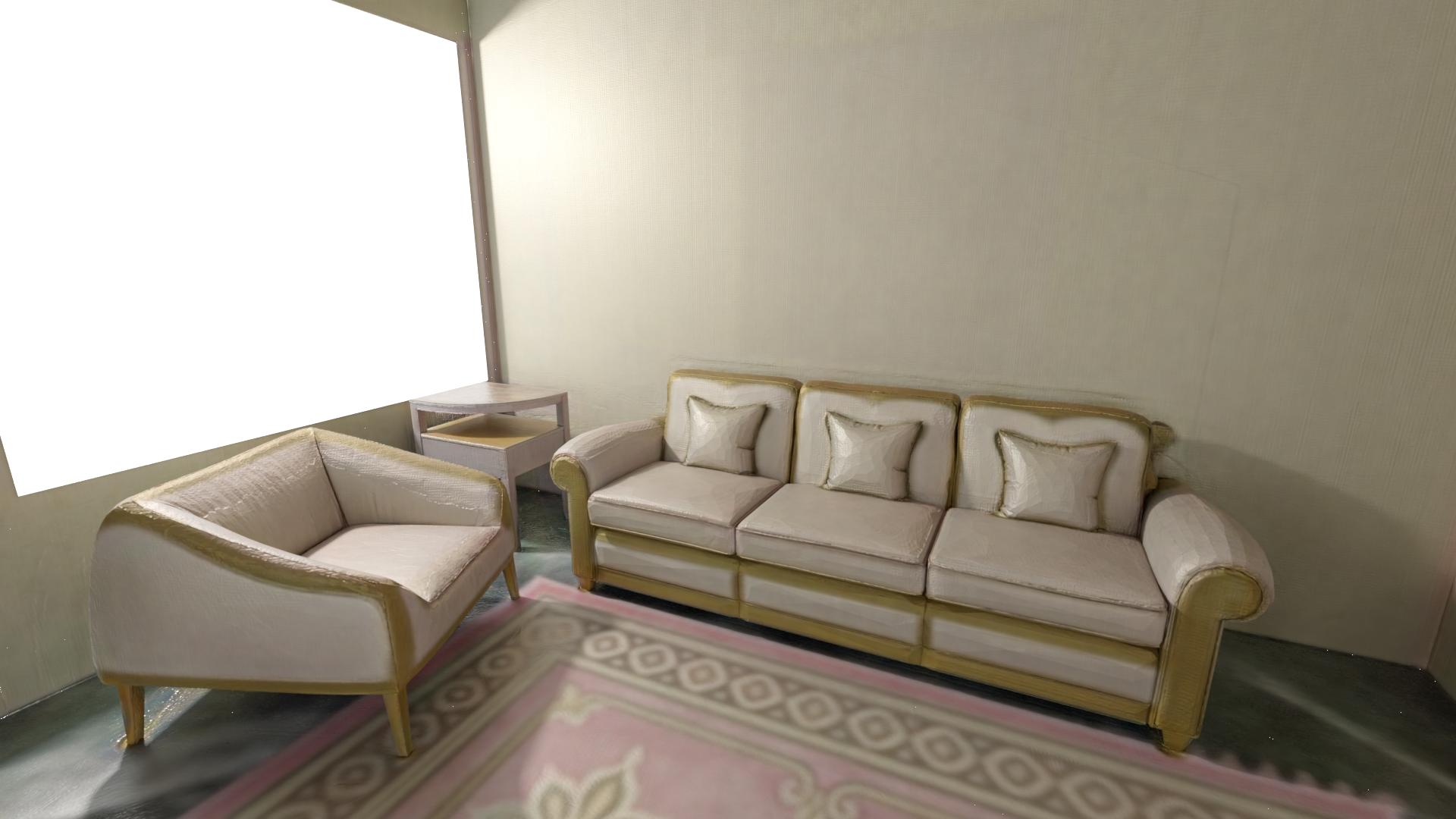}} 
        \\
        
        
        \rotatebox{90}{{\footnotesize View 3}}
        &
        \fbox{\includegraphics[width=0.15\textwidth,trim={10cm 0 10cm 0},clip]{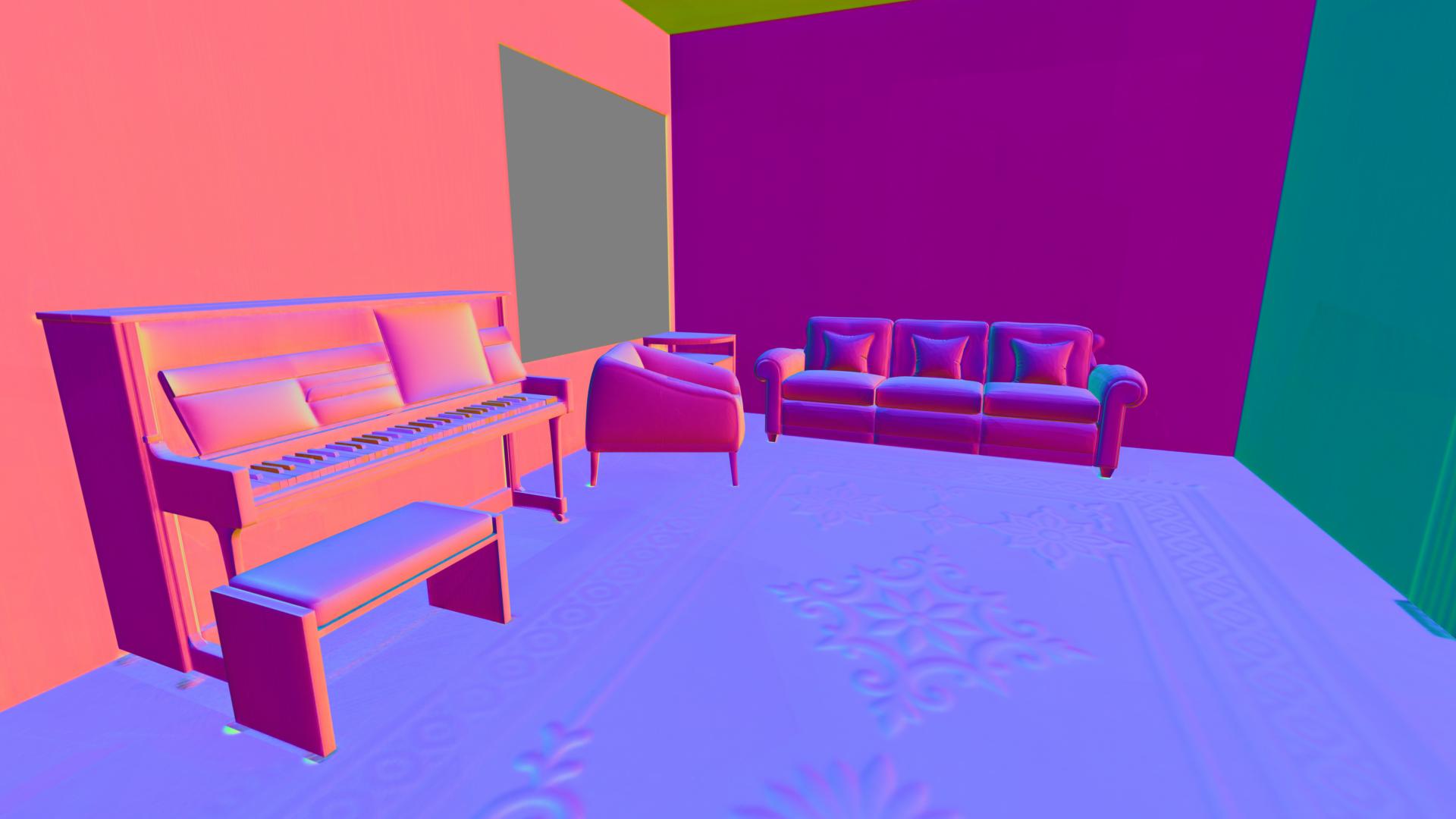}} 
        &
        \fbox{\includegraphics[width=0.15\textwidth,trim={10cm 0 10cm 0},clip]{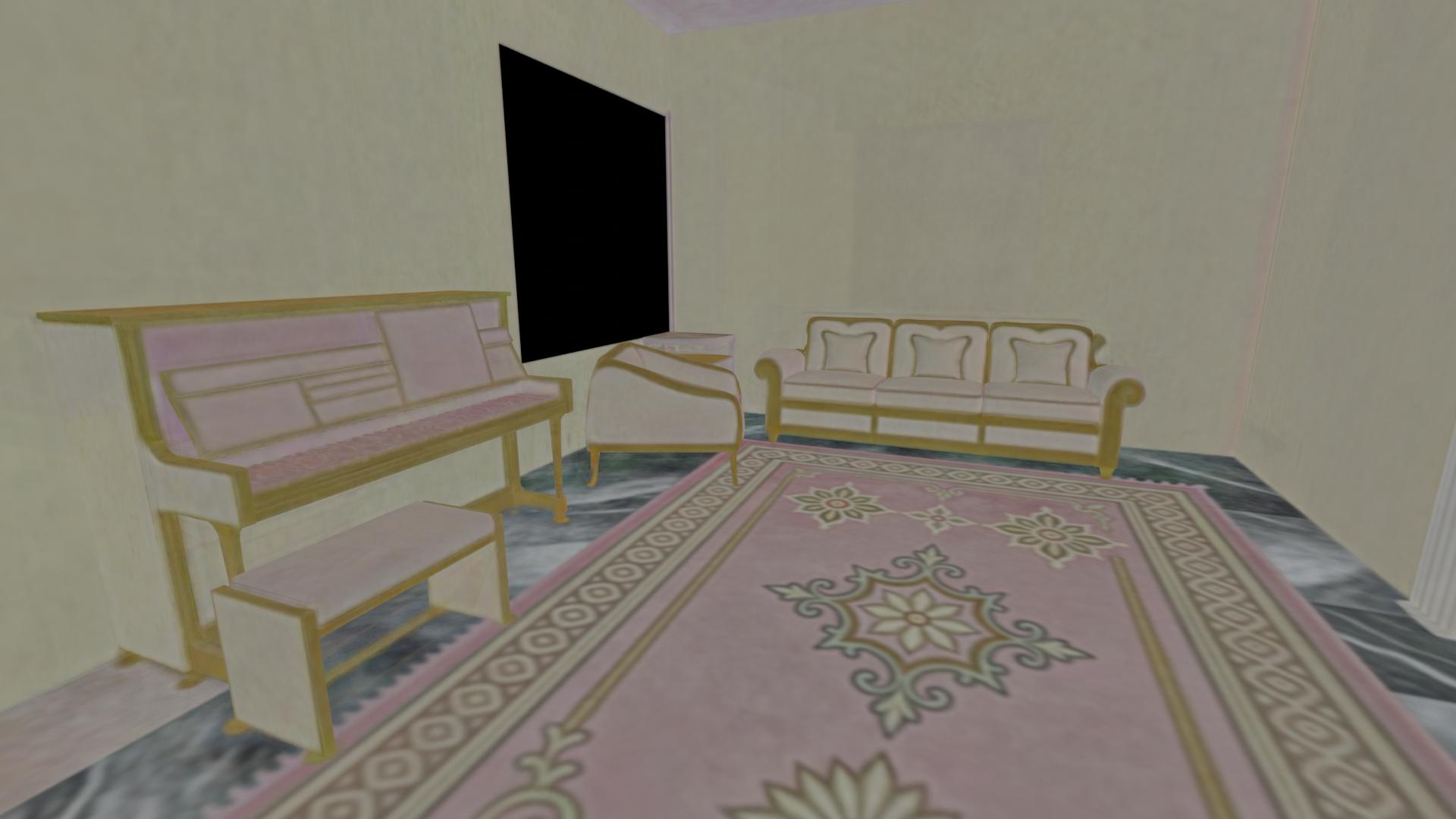}} 
        &
        \fbox{\includegraphics[width=0.15\textwidth,trim={10cm 0 10cm 0},clip]{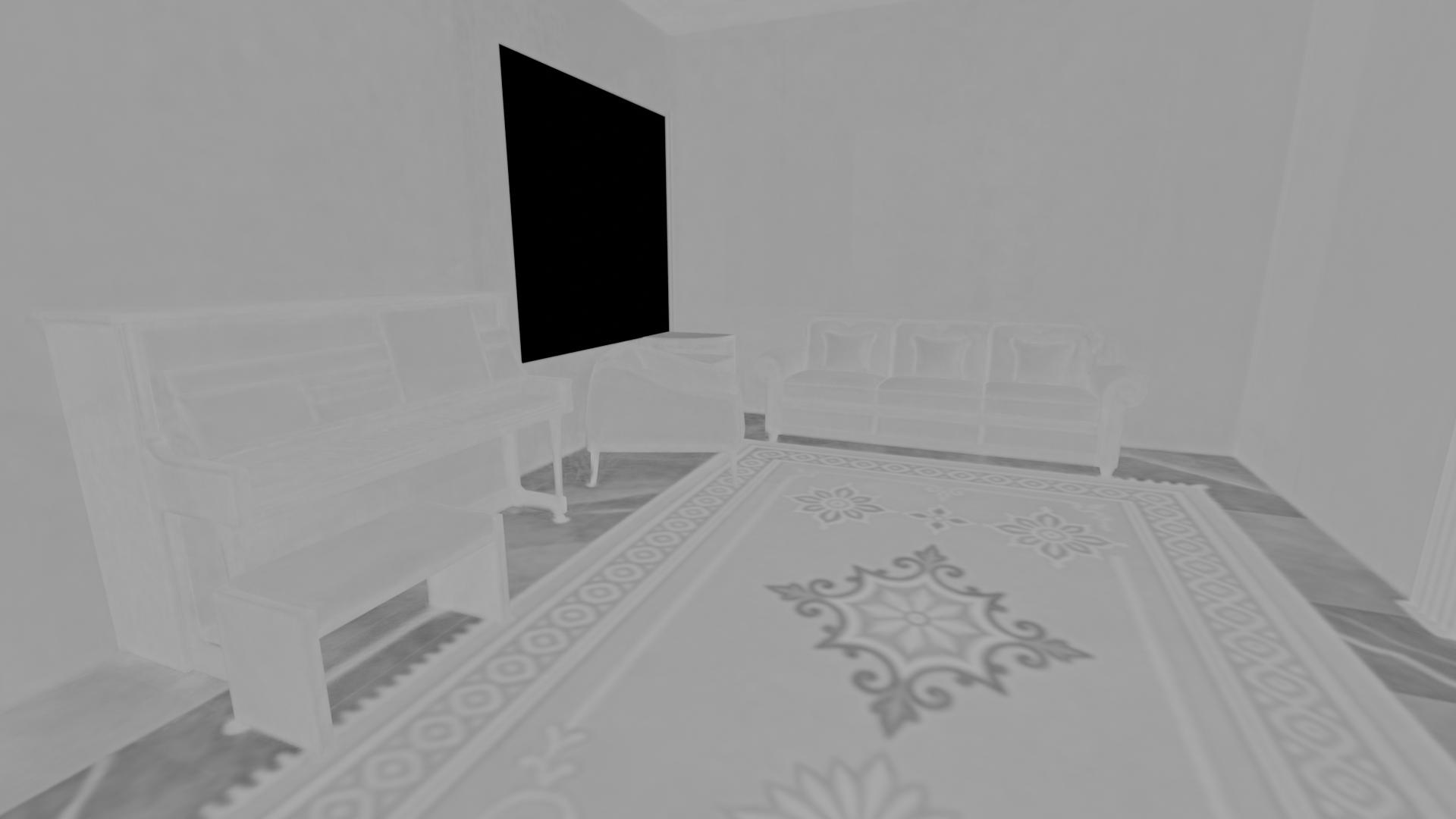}} 
        &
        \fbox{\includegraphics[width=0.15\textwidth,trim={10cm 0 10cm 0},clip]{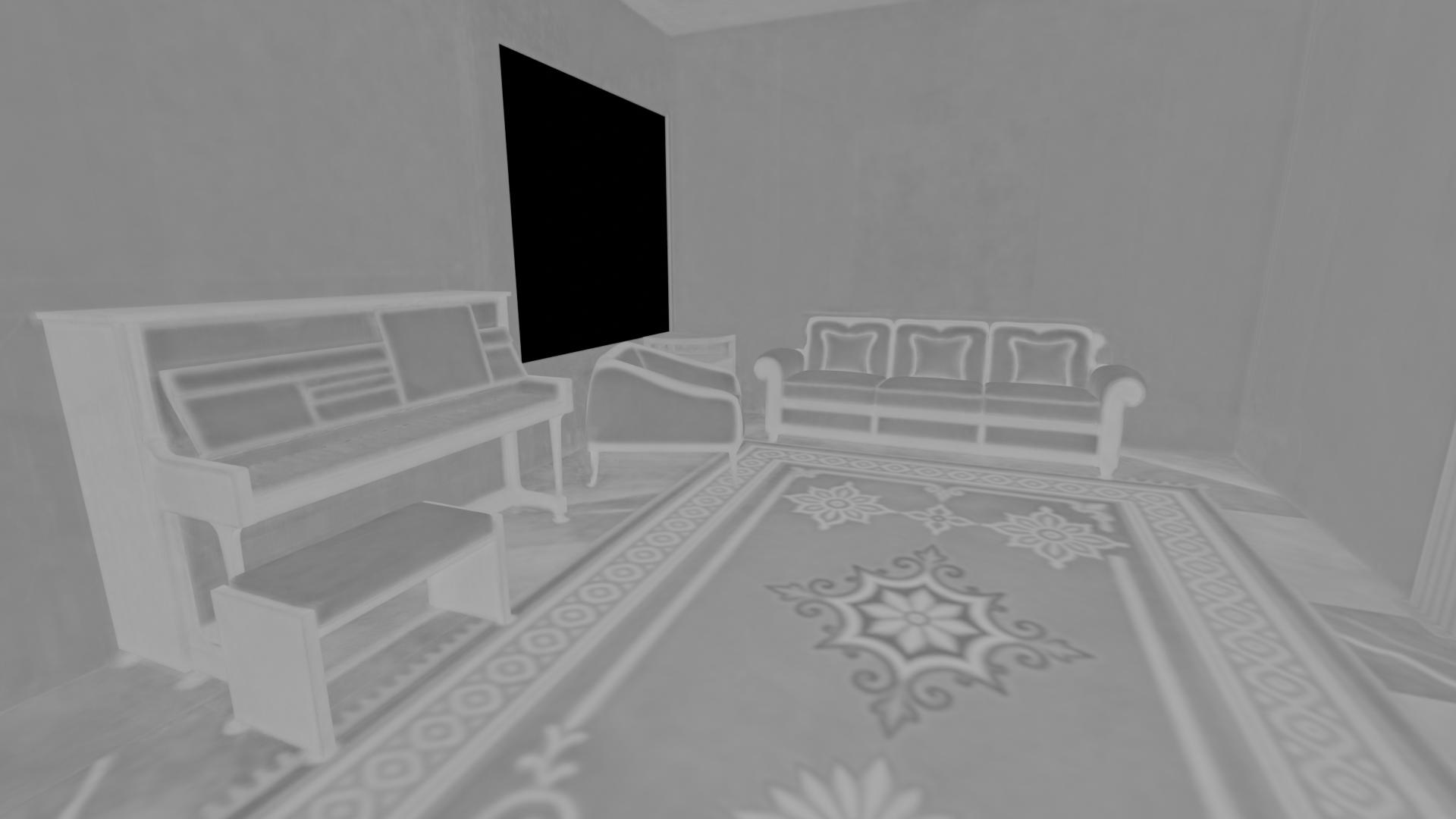}} 
        &
        \fbox{\includegraphics[width=0.15\textwidth,trim={10cm 0 10cm 0},clip]{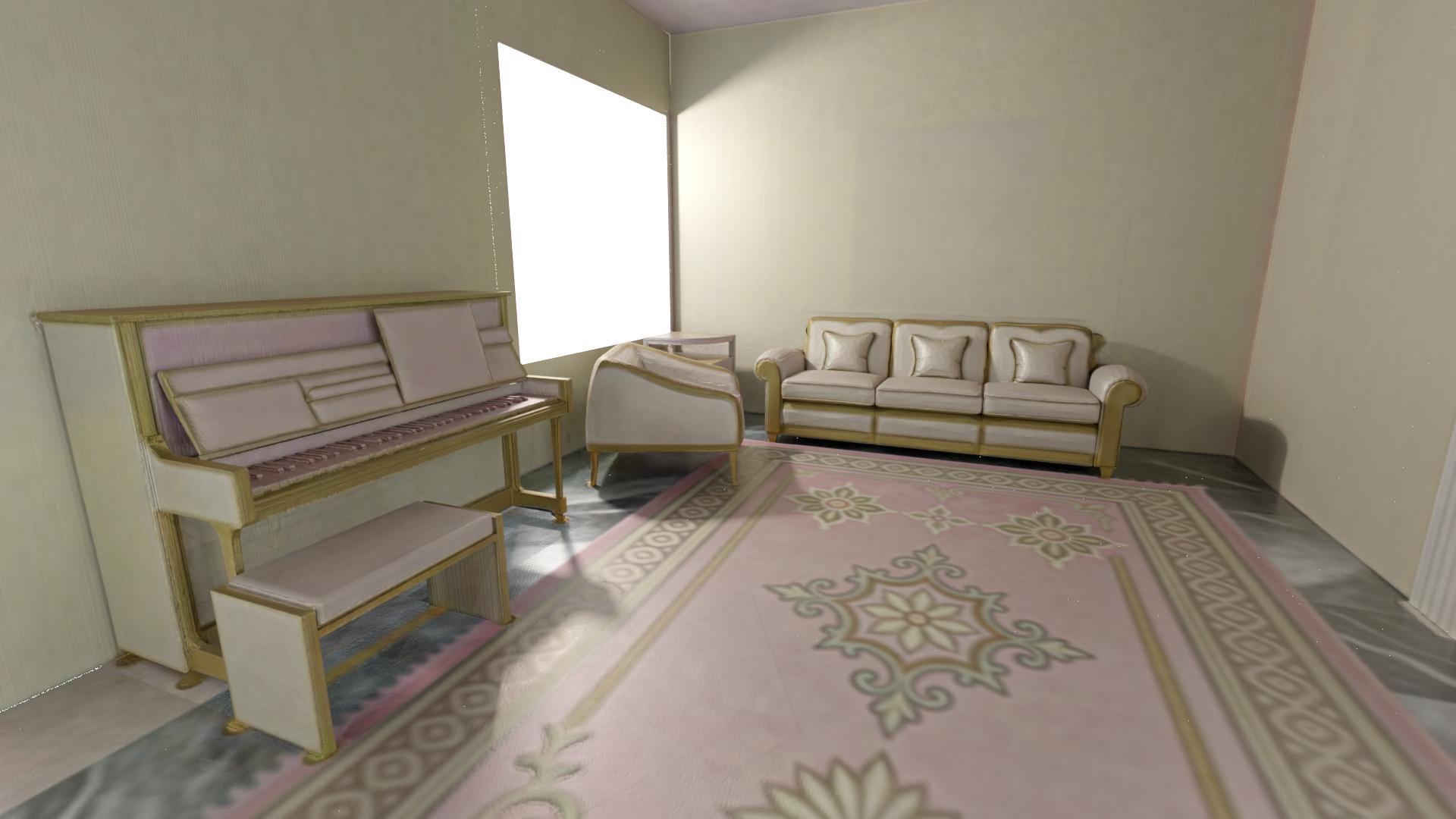}} 
        \\

        &
        {\footnotesize Normal} &
        {\footnotesize Albedo} &
        {\footnotesize Roughness} &
        {\footnotesize Metallic} &
        {\footnotesize Rendering} \\

        \midrule
        
        \rotatebox{90}{{\footnotesize View 1}}
        &
        \fbox{\includegraphics[width=0.15\textwidth,trim={10cm 0 10cm 0},clip]{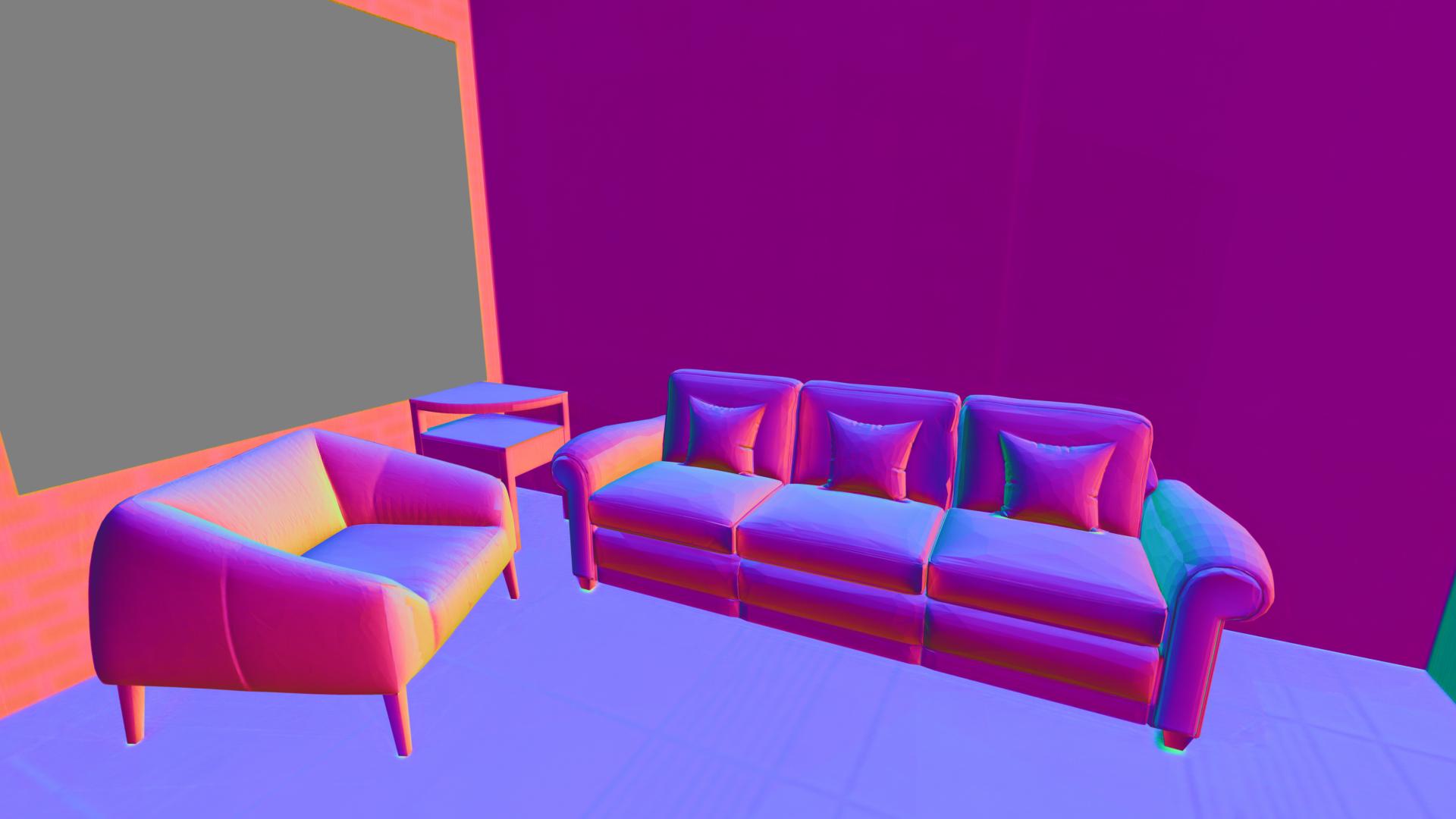}}
        &
        \fbox{\includegraphics[width=0.15\textwidth,trim={10cm 0 10cm 0},clip]{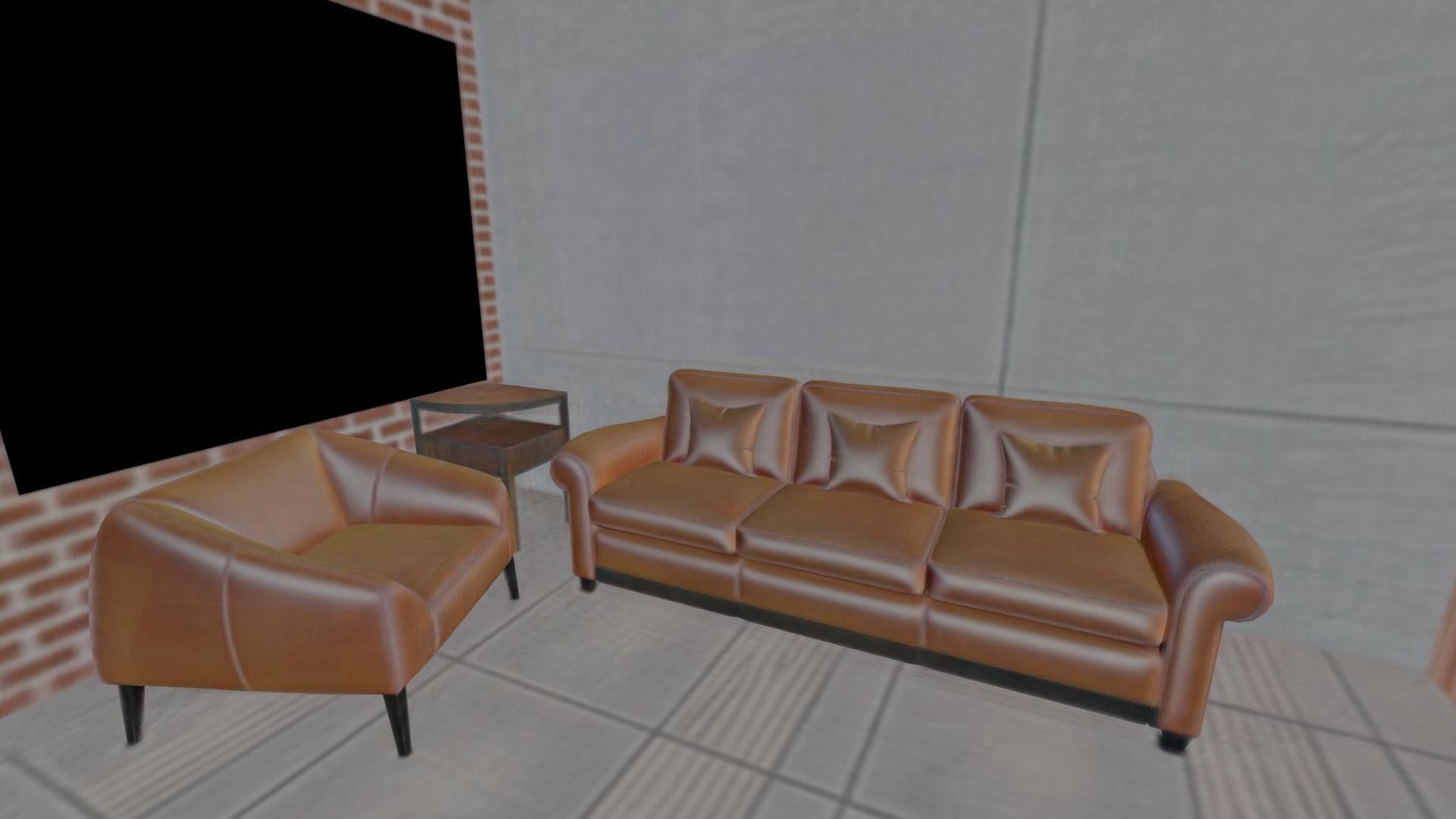}} 
        &
        \fbox{\includegraphics[width=0.15\textwidth,trim={10cm 0 10cm 0},clip]{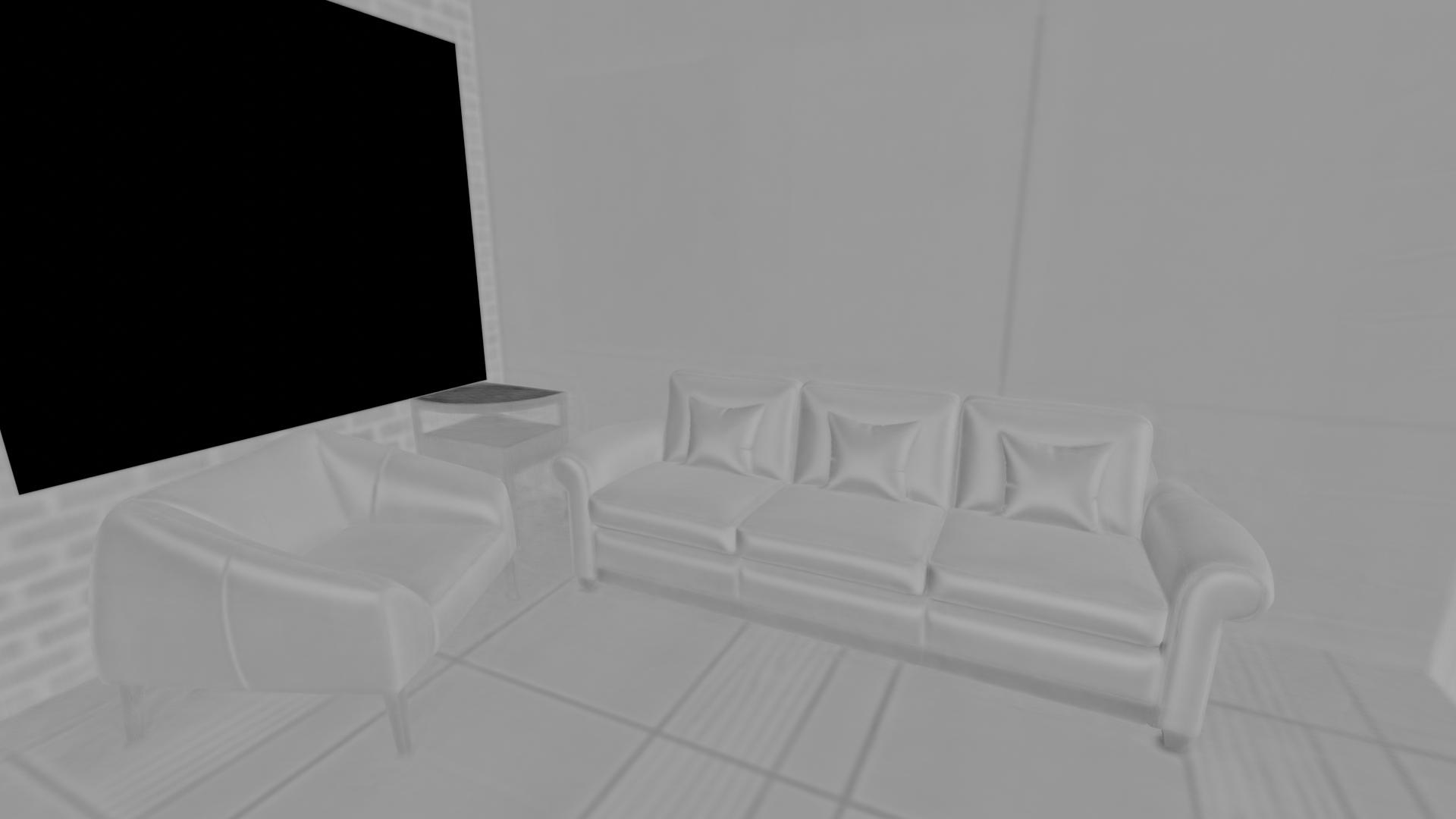}} 
        &
        \fbox{\includegraphics[width=0.15\textwidth,trim={10cm 0 10cm 0},clip]{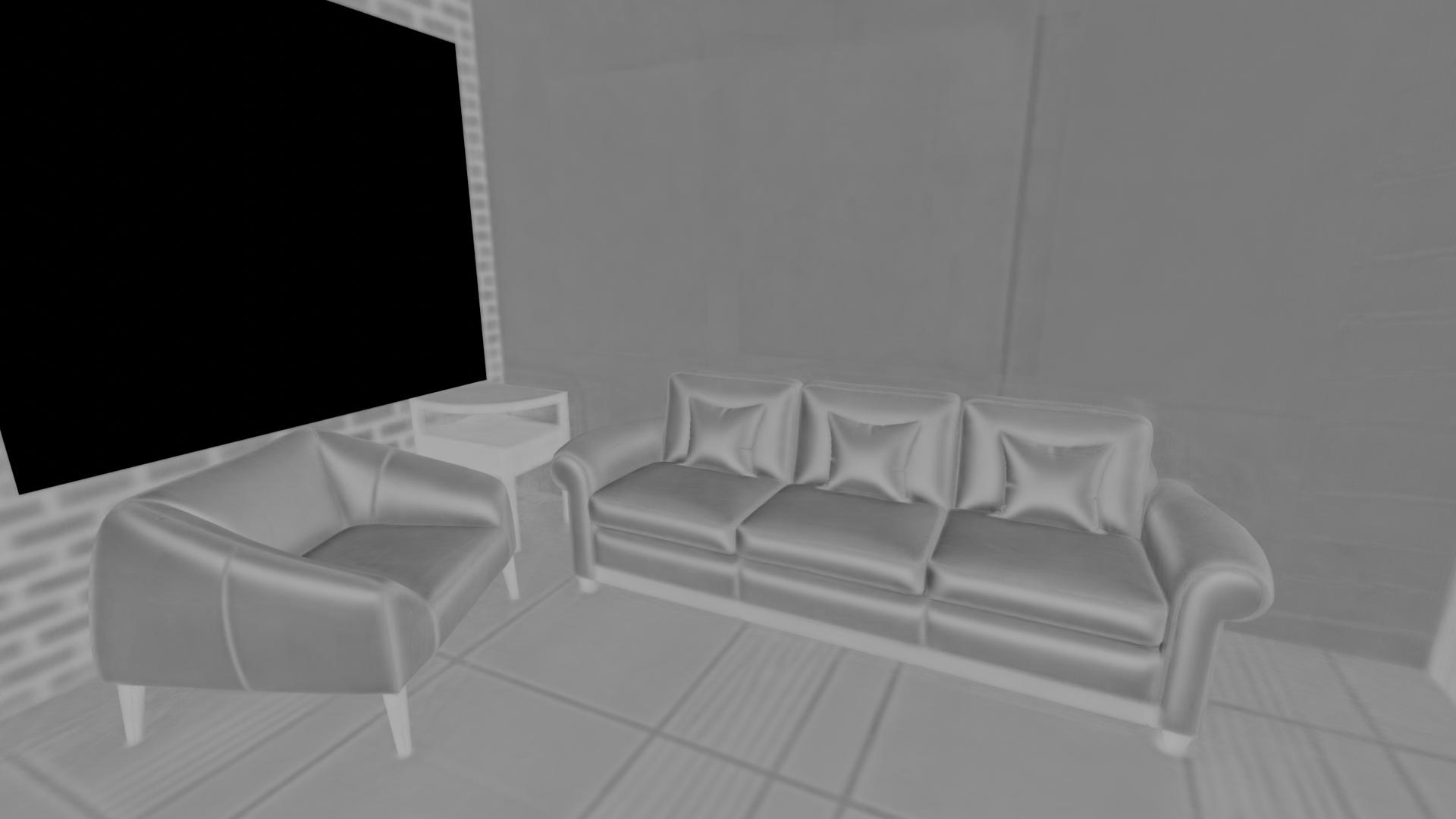}} 
        &
        \fbox{\includegraphics[width=0.15\textwidth,trim={10cm 0 10cm 0},clip]{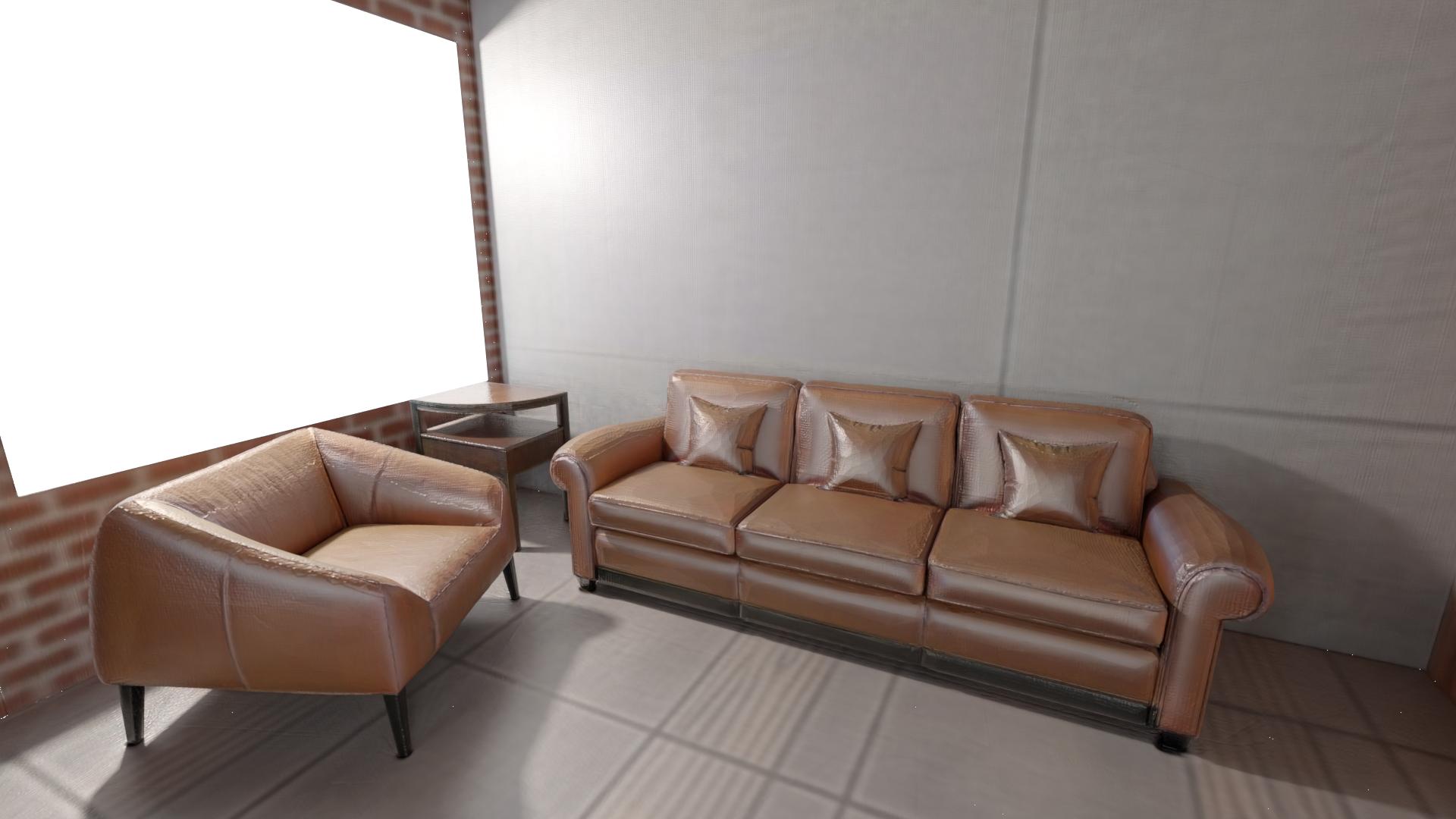}} 
        \\
        
        
        \rotatebox{90}{{\footnotesize View 3}}
        &
        \fbox{\includegraphics[width=0.15\textwidth,trim={10cm 0 10cm 0},clip]{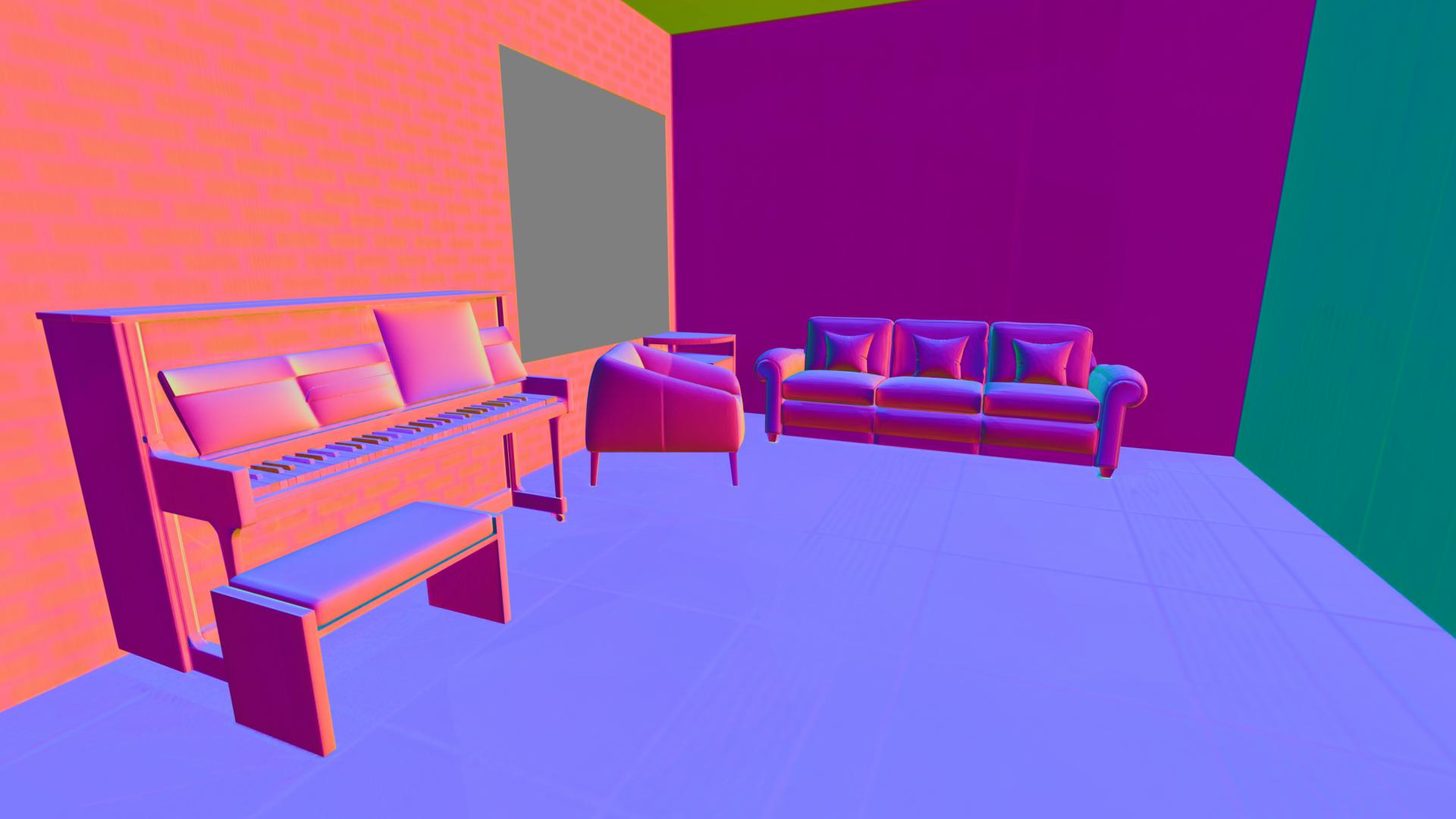}} 
        &
        \fbox{\includegraphics[width=0.15\textwidth,trim={10cm 0 10cm 0},clip]{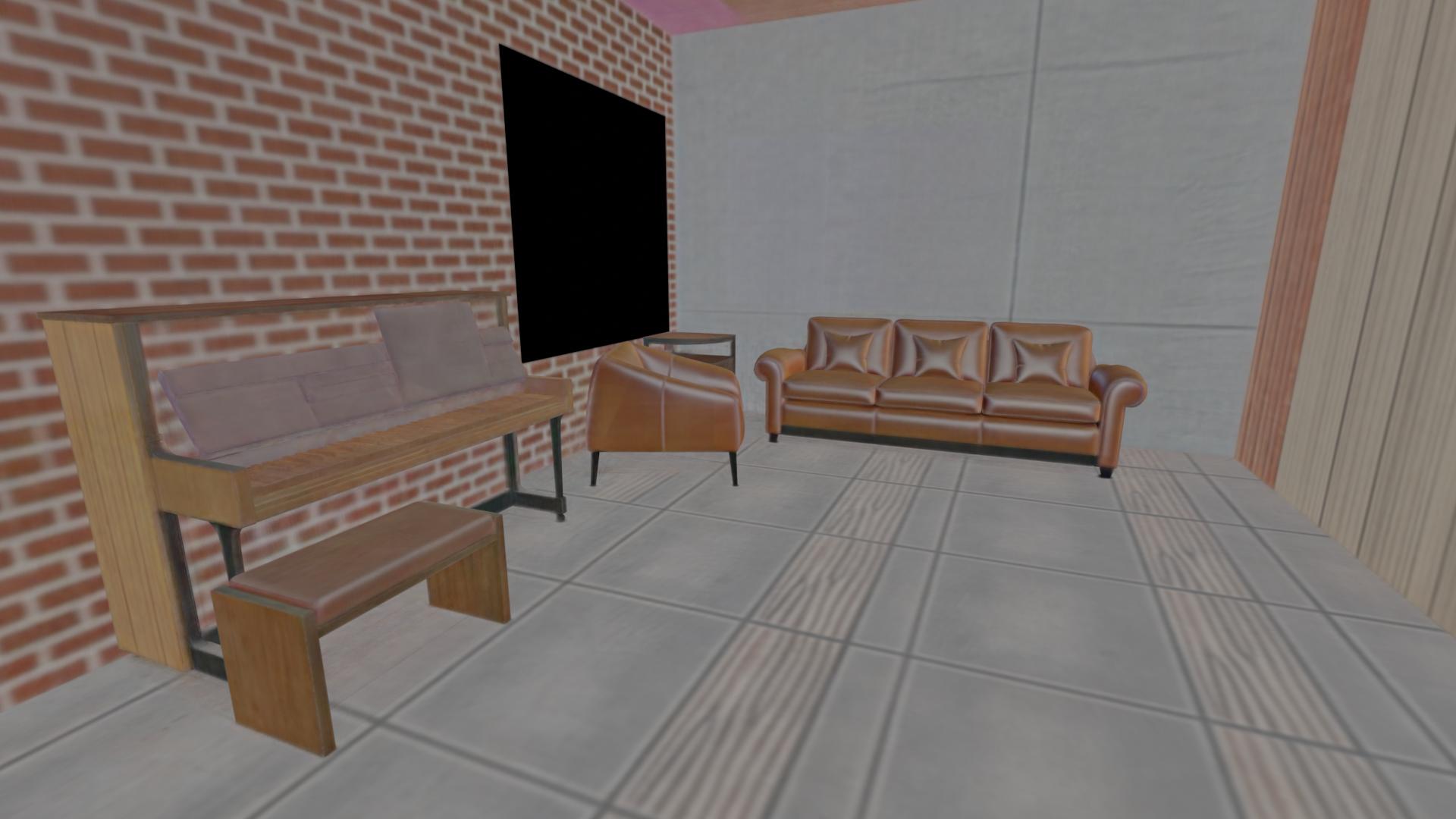}} 
        &
        \fbox{\includegraphics[width=0.15\textwidth,trim={10cm 0 10cm 0},clip]{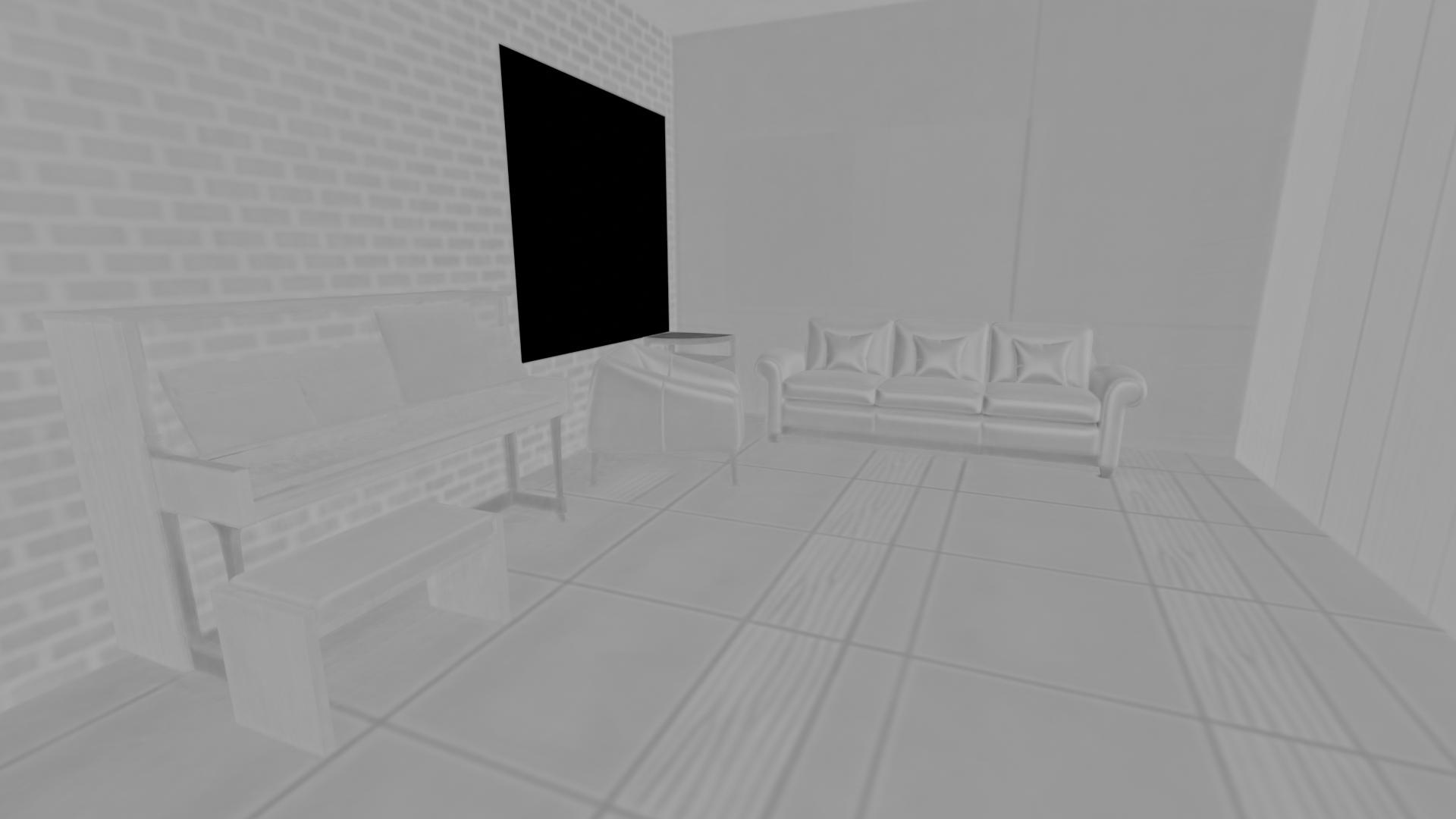}} 
        &
        \fbox{\includegraphics[width=0.15\textwidth,trim={10cm 0 10cm 0},clip]{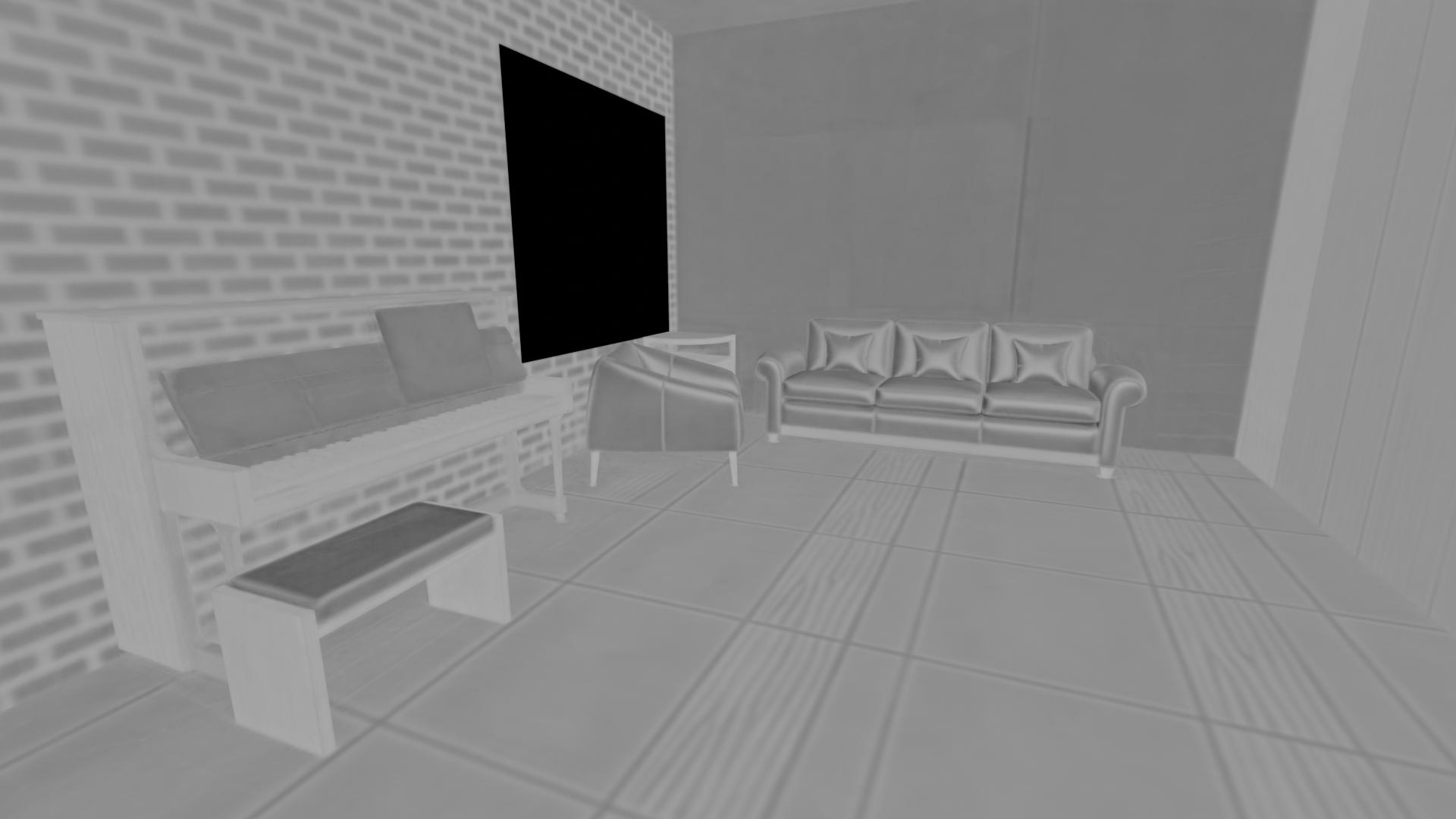}} 
        &
        \fbox{\includegraphics[width=0.15\textwidth,trim={10cm 0 10cm 0},clip]{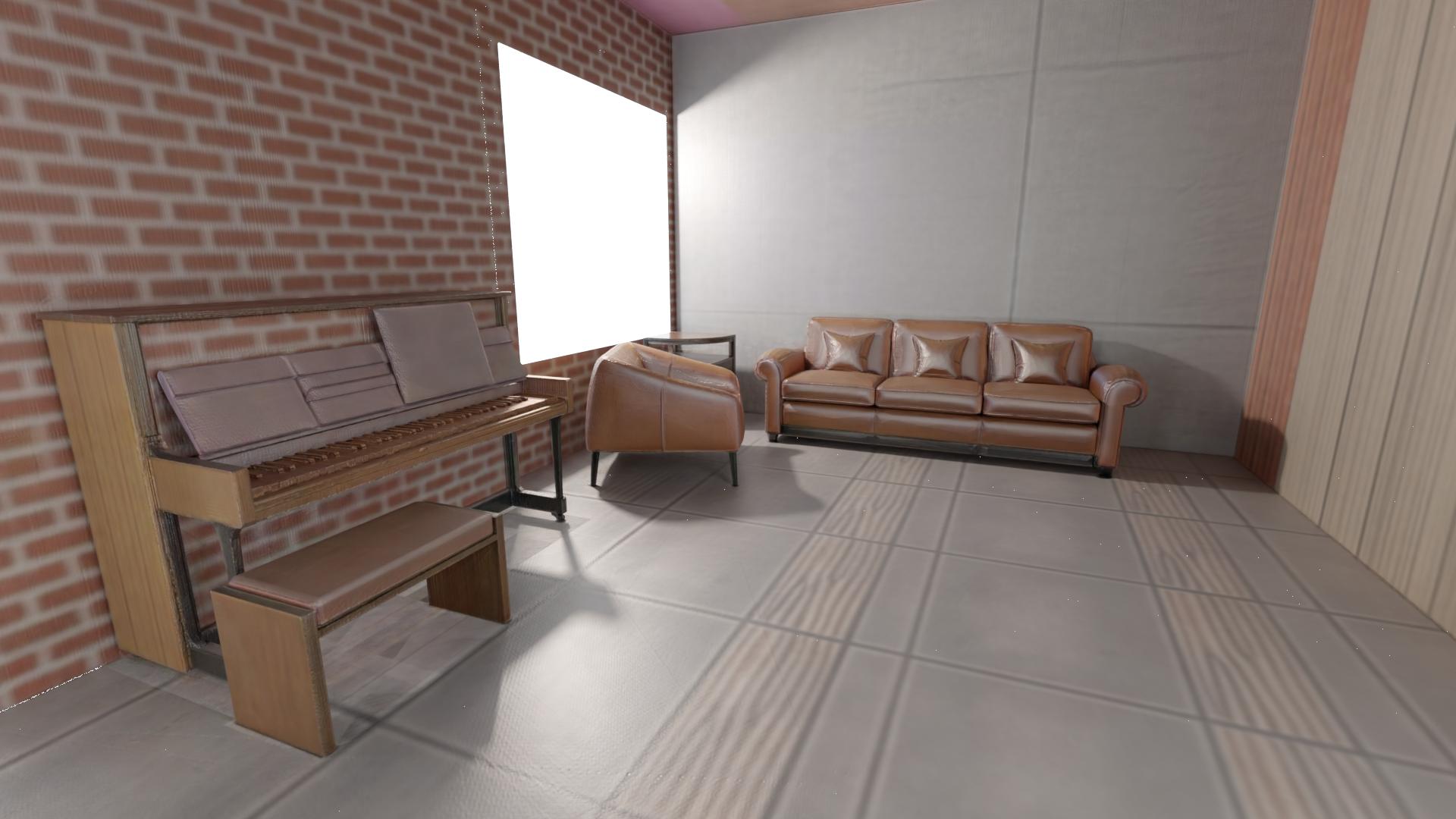}} 
        \\

        &
        {\footnotesize Normal} &
        {\footnotesize Albedo} &
        {\footnotesize Roughness} &
        {\footnotesize Metallic} &
        {\footnotesize Rendering} \\

        \midrule
        
        \rotatebox{90}{{\footnotesize View 1}}
        &
        \fbox{\includegraphics[width=0.15\textwidth,trim={10cm 0 10cm 0},clip]{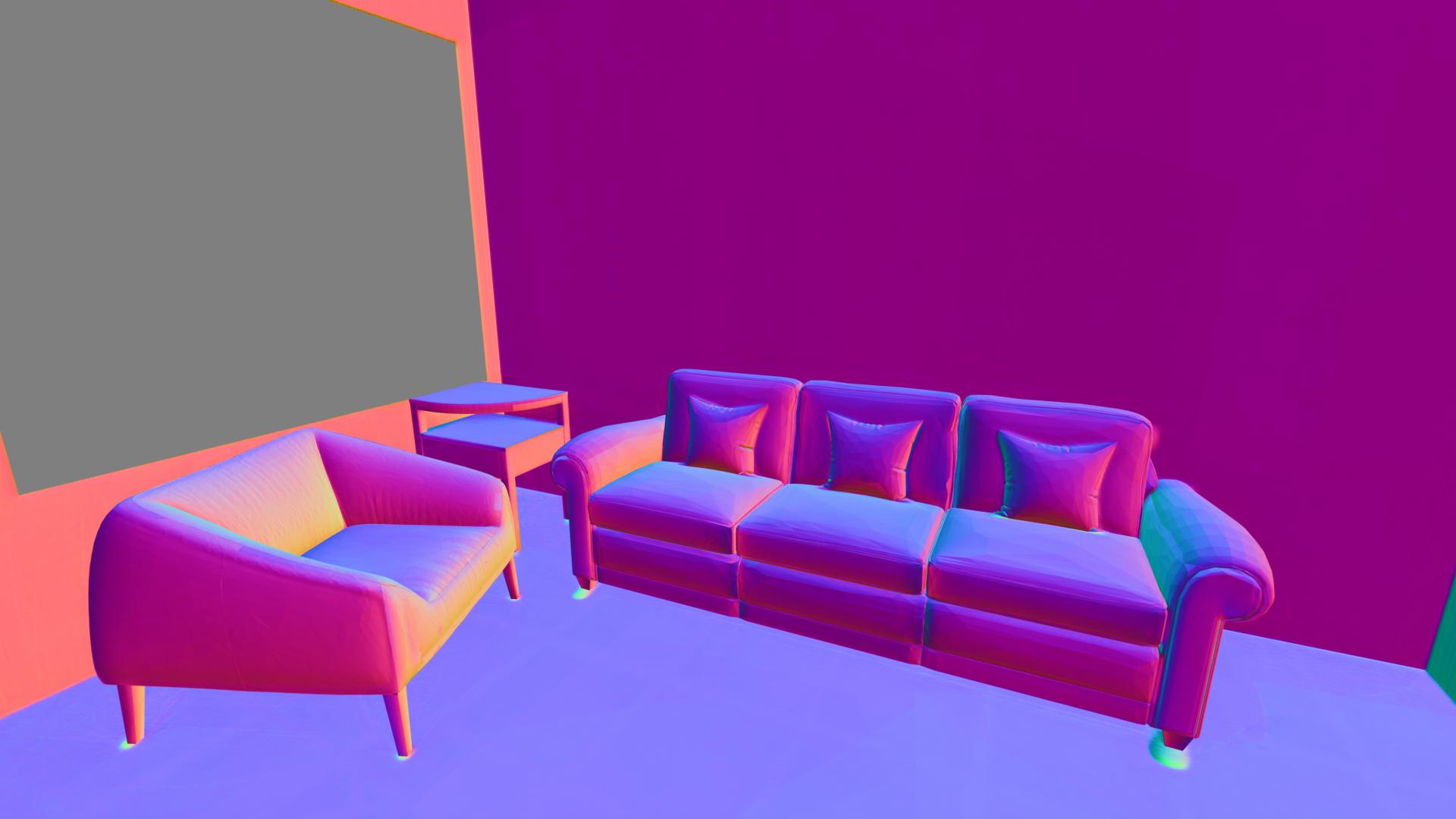}}
        &
        \fbox{\includegraphics[width=0.15\textwidth,trim={10cm 0 10cm 0},clip]{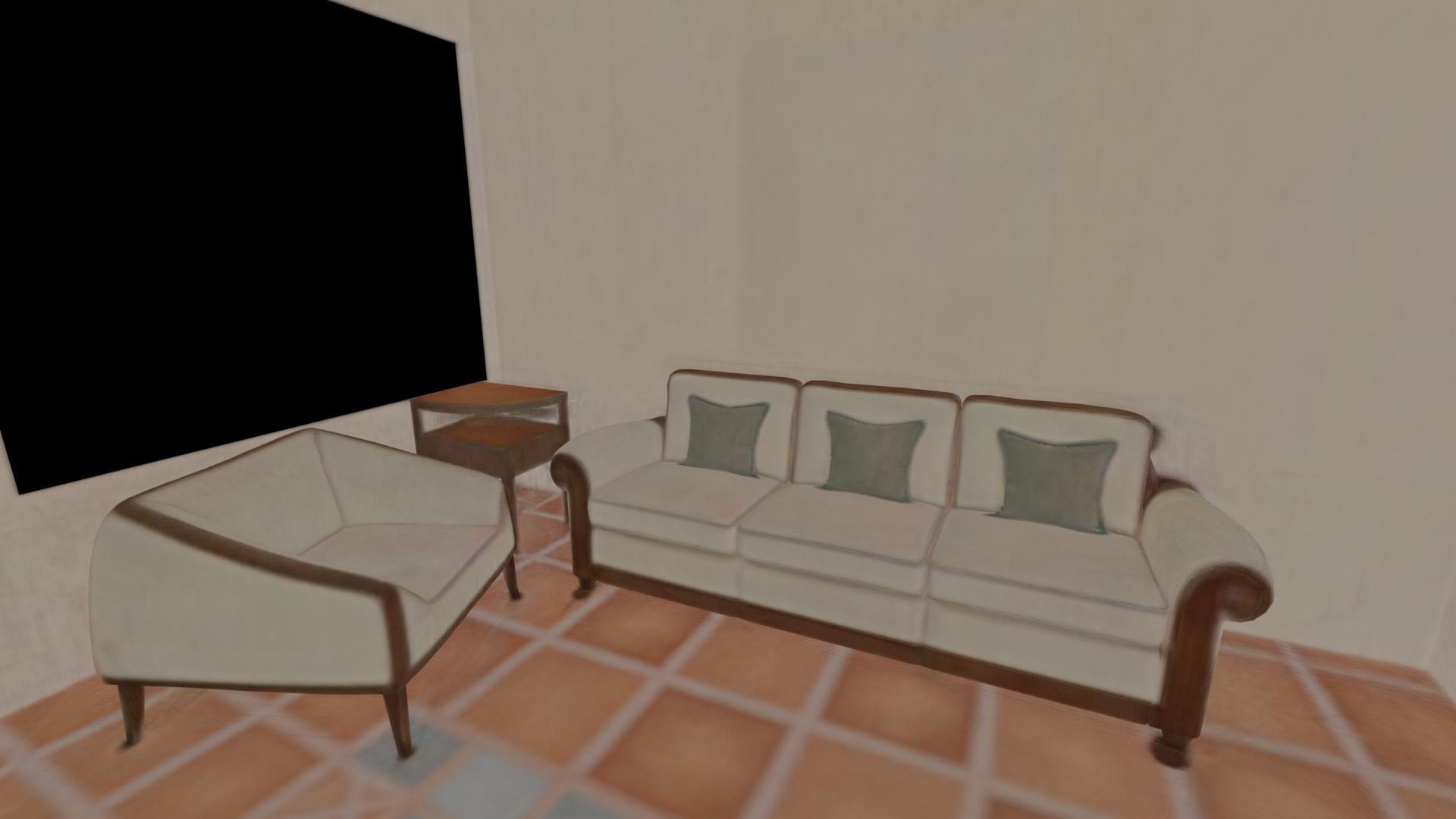}} 
        &
        \fbox{\includegraphics[width=0.15\textwidth,trim={10cm 0 10cm 0},clip]{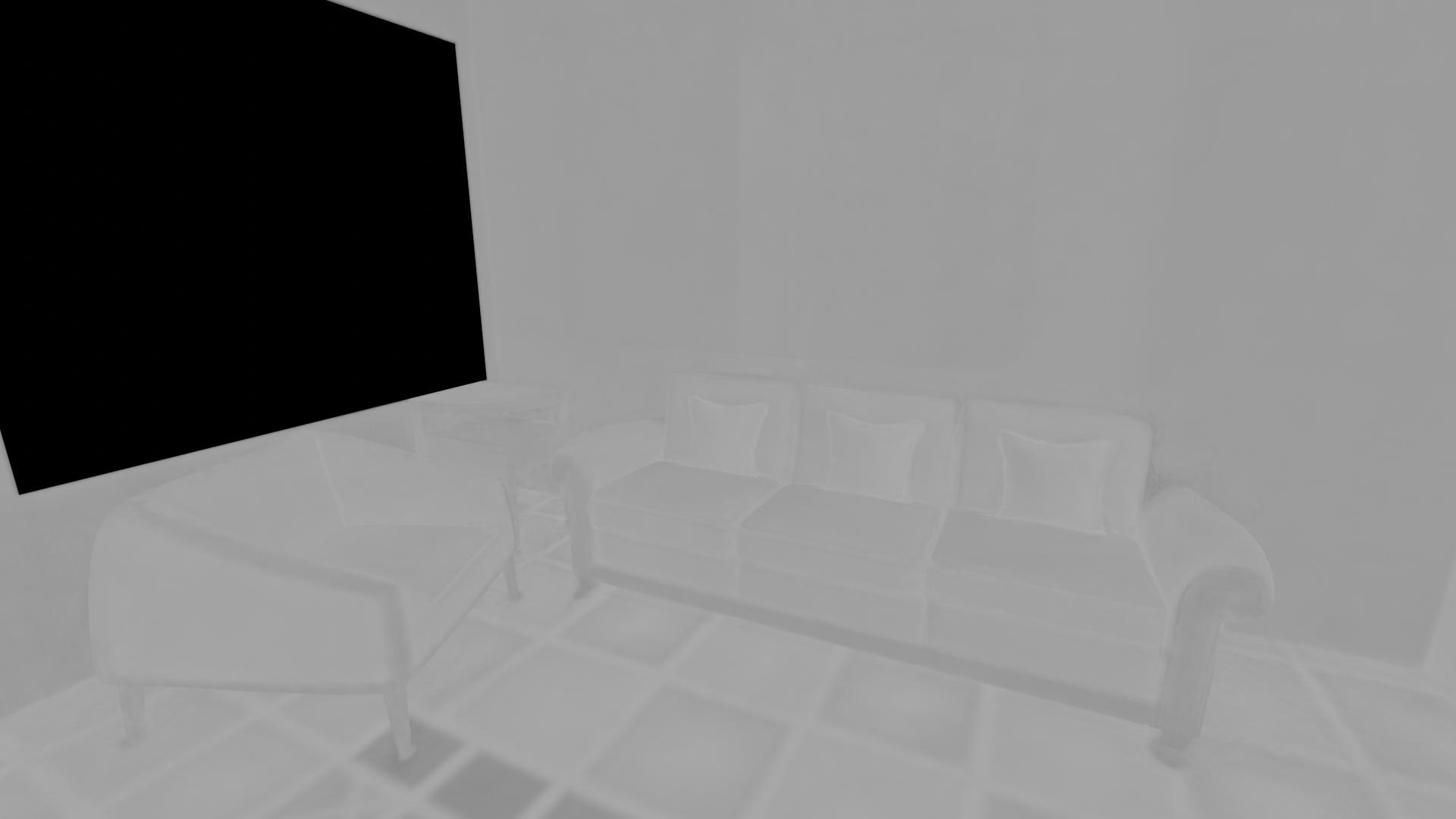}} 
        &
        \fbox{\includegraphics[width=0.15\textwidth,trim={10cm 0 10cm 0},clip]{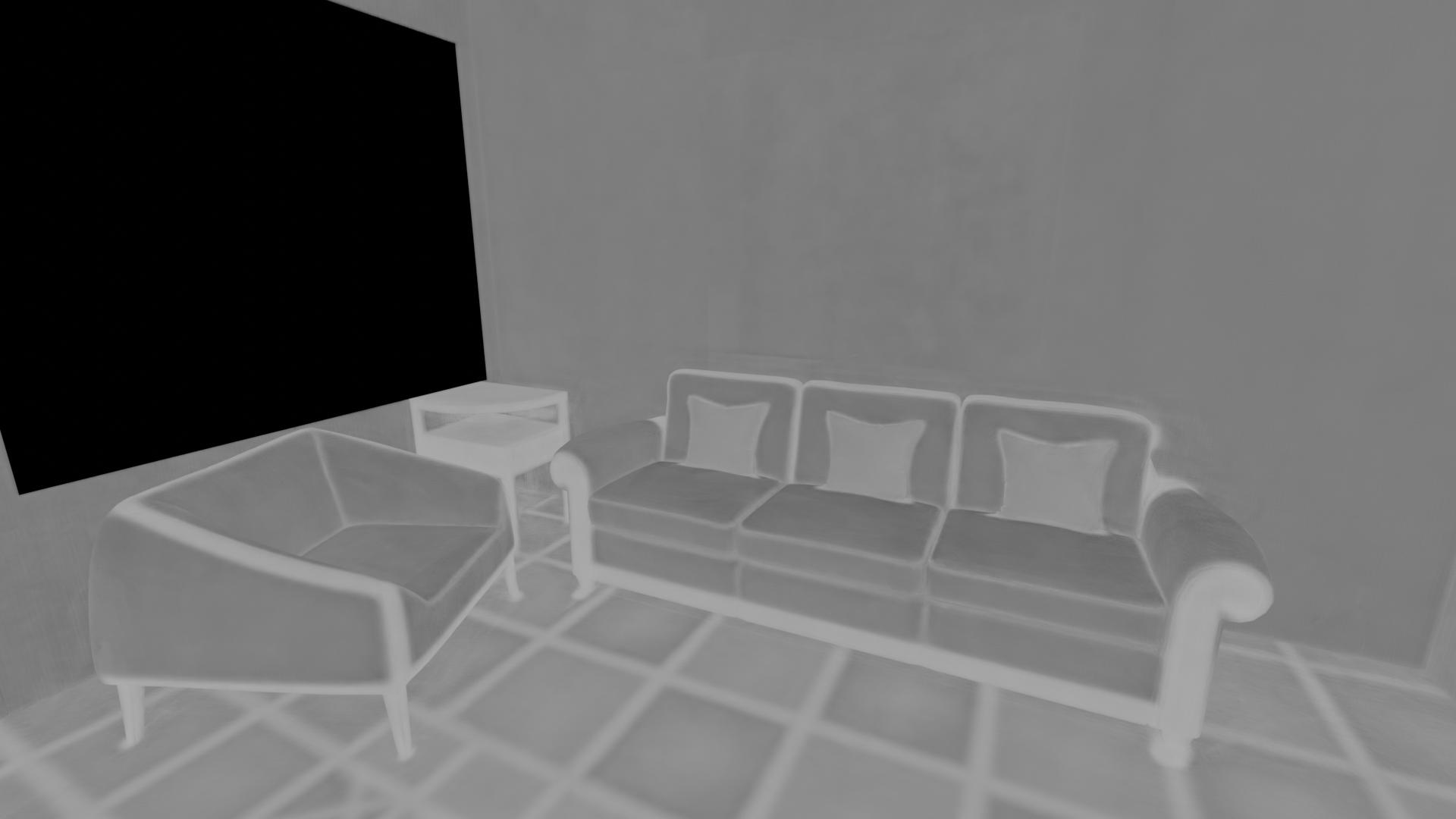}} 
        &
        \fbox{\includegraphics[width=0.15\textwidth,trim={10cm 0 10cm 0},clip]{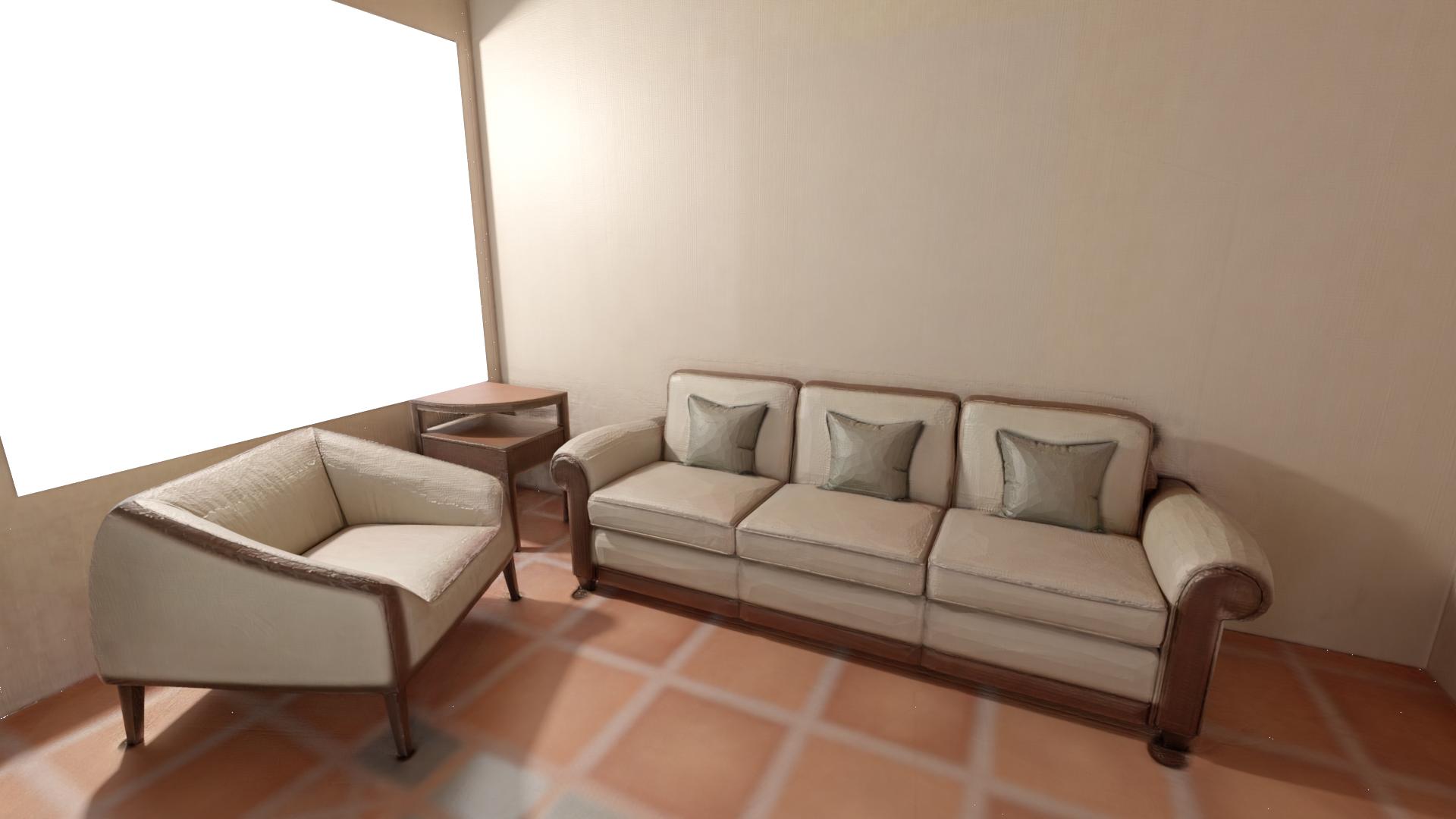}} 
        \\
        
        
        \rotatebox{90}{{\footnotesize View 3}}
        &
        \fbox{\includegraphics[width=0.15\textwidth,trim={10cm 0 10cm 0},clip]{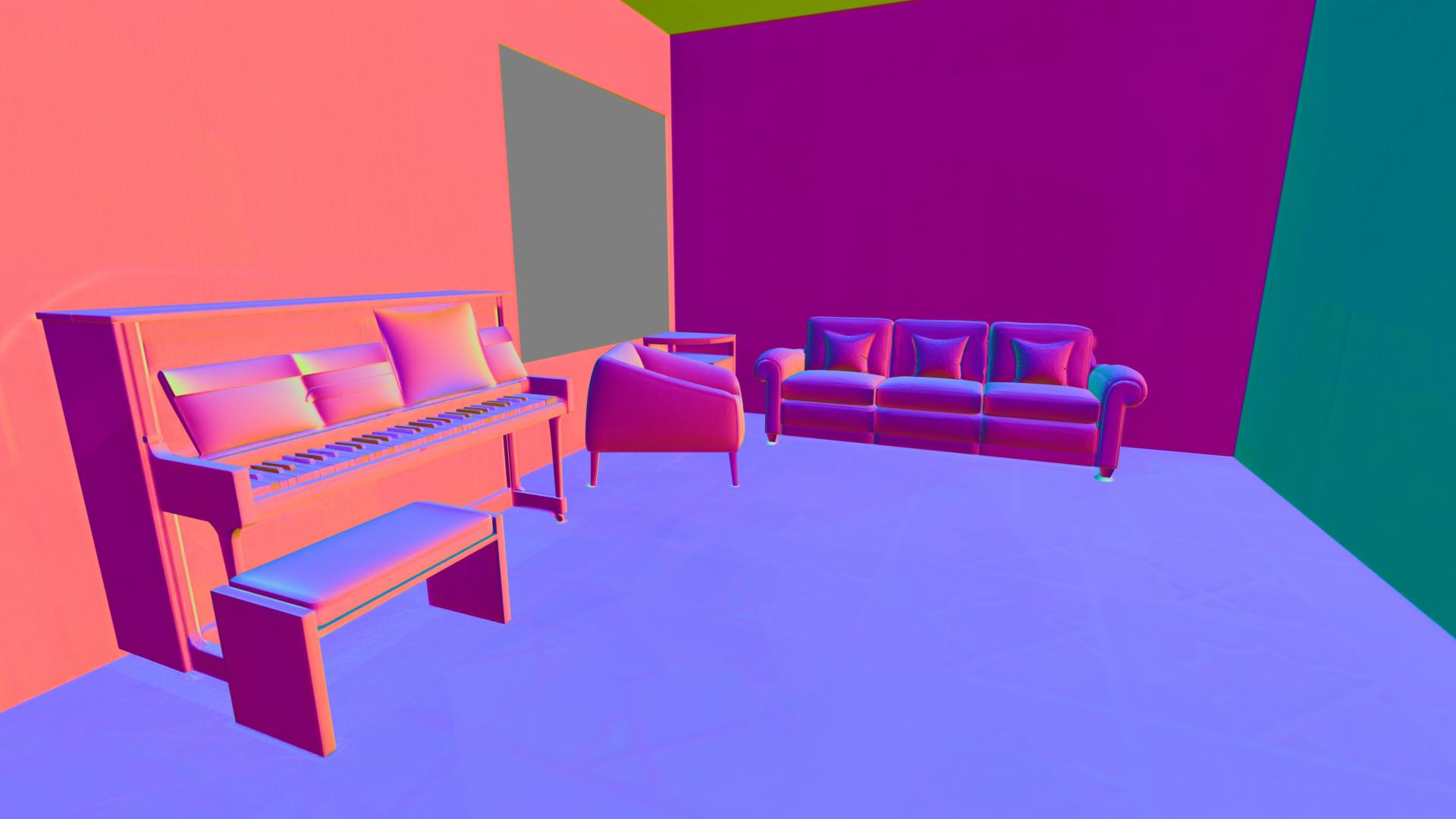}} 
        &
        \fbox{\includegraphics[width=0.15\textwidth,trim={10cm 0 10cm 0},clip]{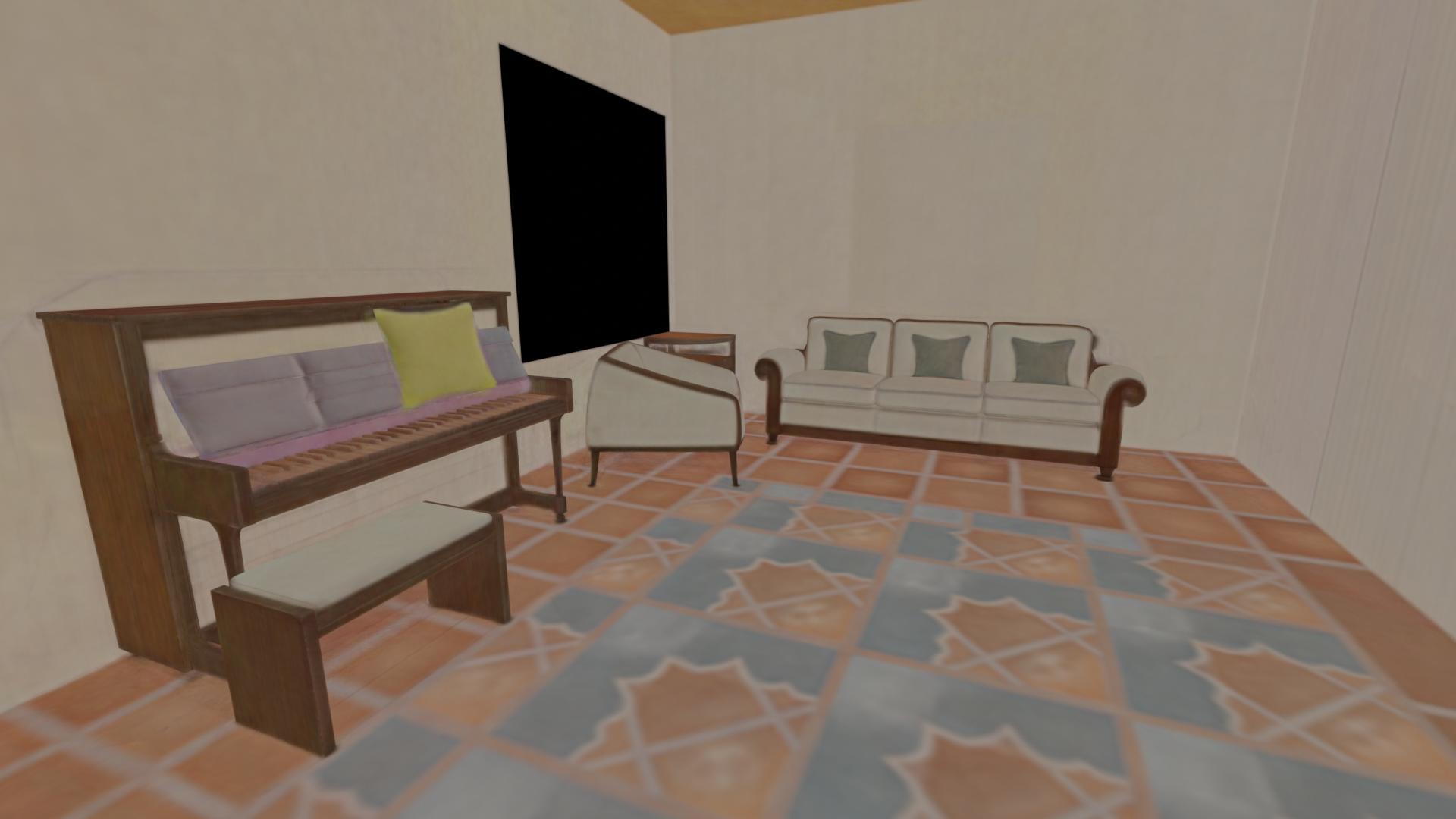}} 
        &
        \fbox{\includegraphics[width=0.15\textwidth,trim={10cm 0 10cm 0},clip]{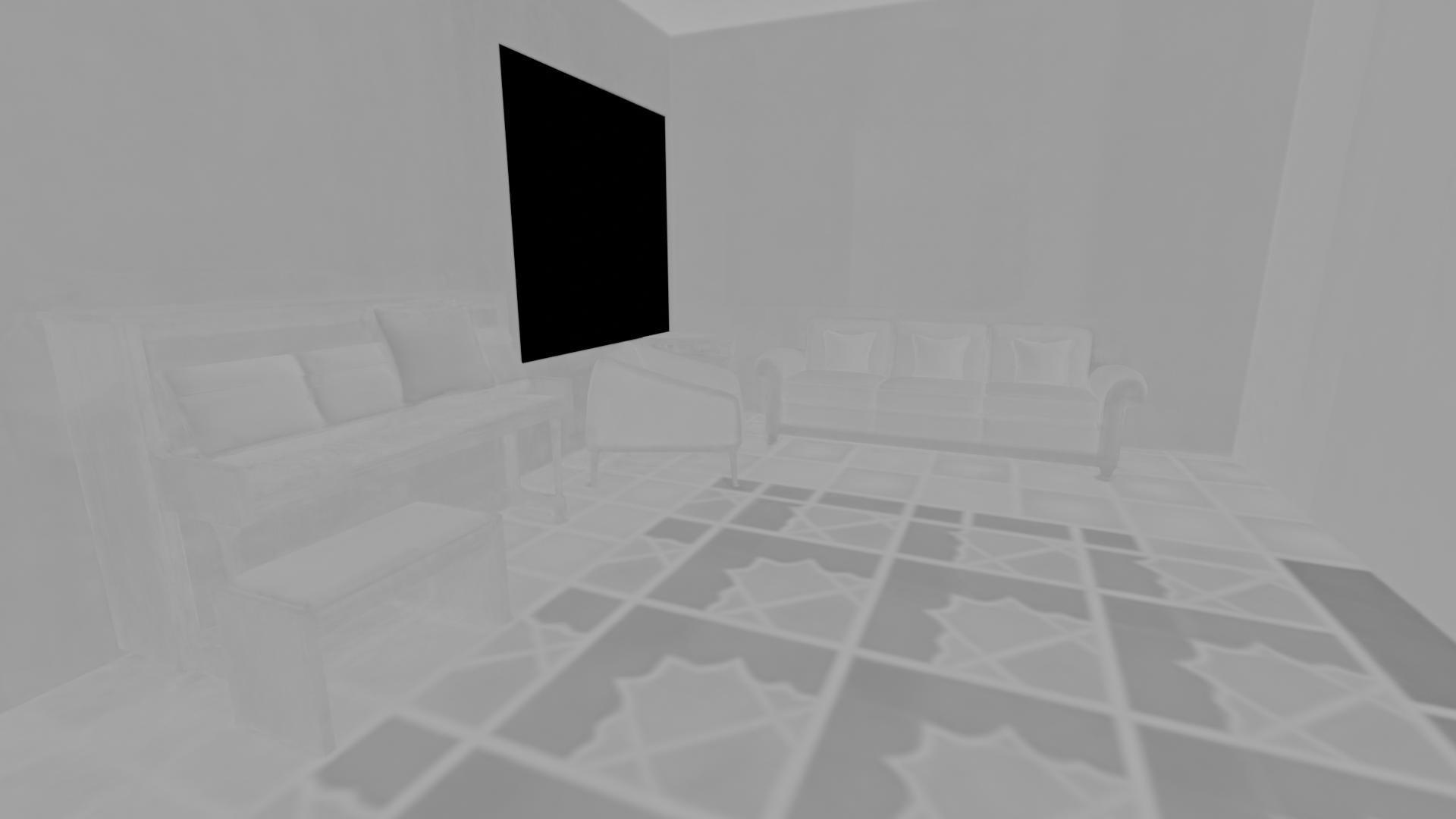}} 
        &
        \fbox{\includegraphics[width=0.15\textwidth,trim={10cm 0 10cm 0},clip]{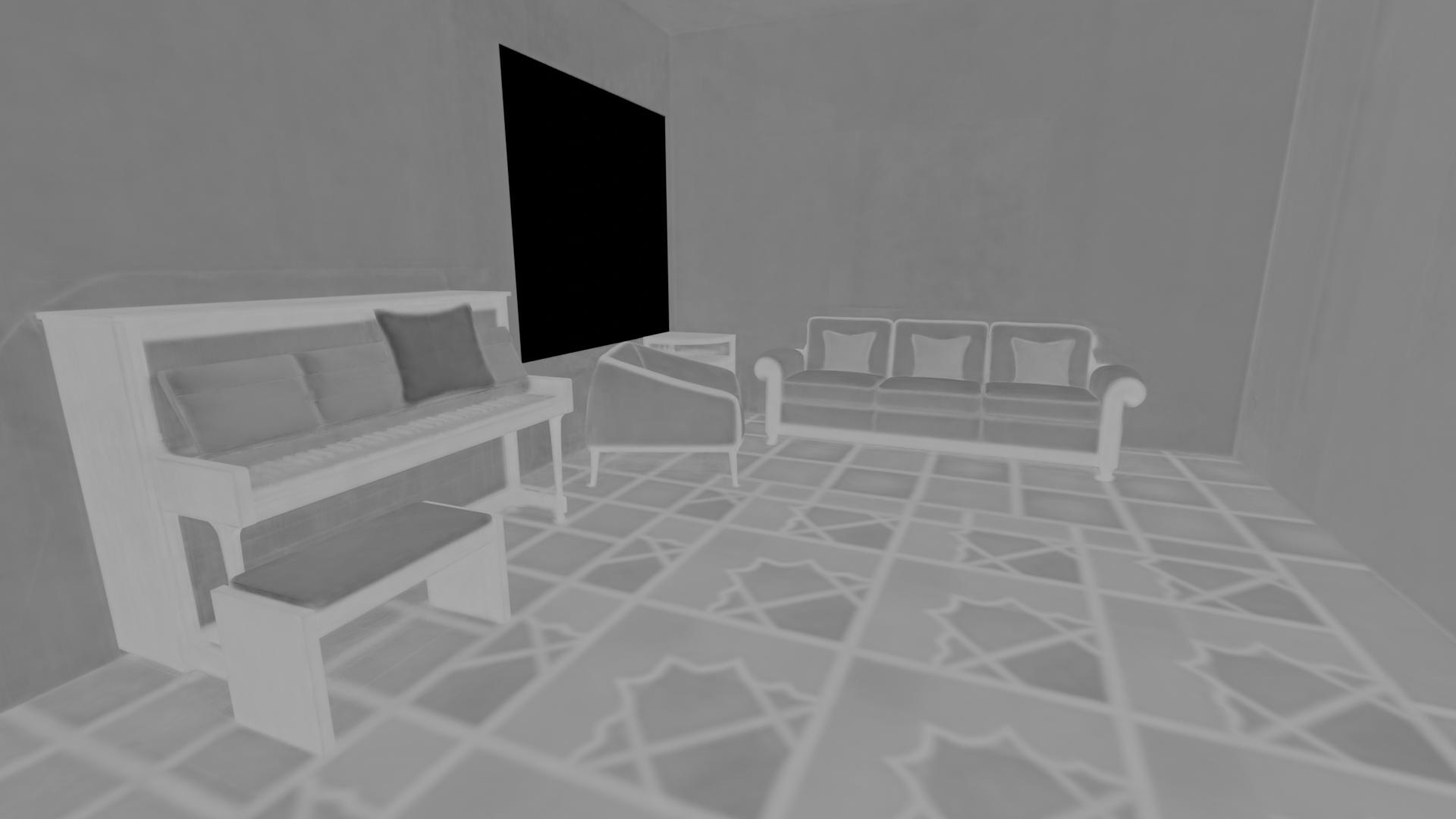}} 
        &
        \fbox{\includegraphics[width=0.15\textwidth,trim={10cm 0 10cm 0},clip]{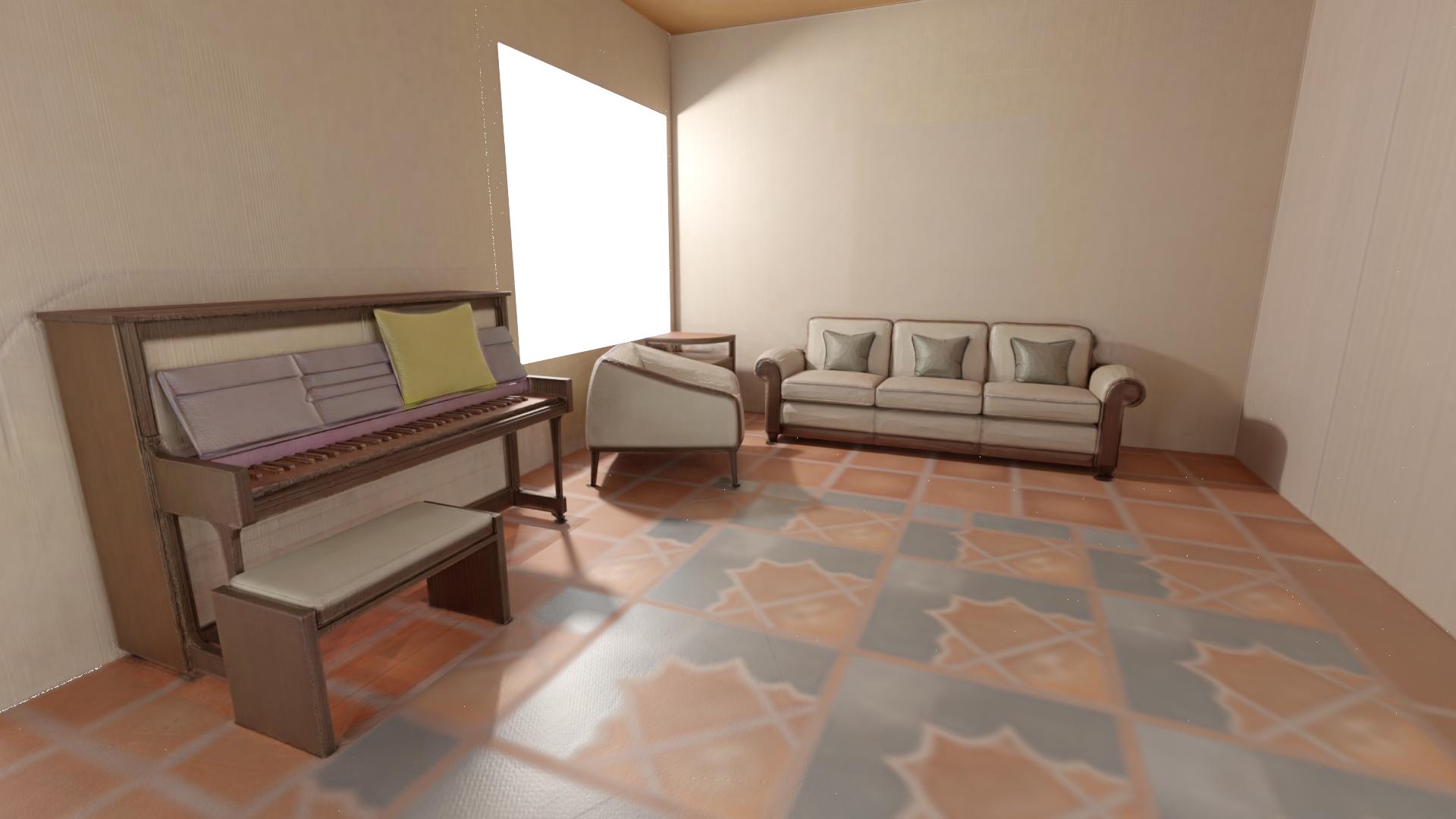}} 
        \\

        &
        {\footnotesize Normal} &
        {\footnotesize Albedo} &
        {\footnotesize Roughness} &
        {\footnotesize Metallic} &
        {\footnotesize Rendering} \\

        \midrule
        
        \rotatebox{90}{{\footnotesize View 1}}
        &
        \fbox{\includegraphics[width=0.15\textwidth,trim={10cm 0 10cm 0},clip]{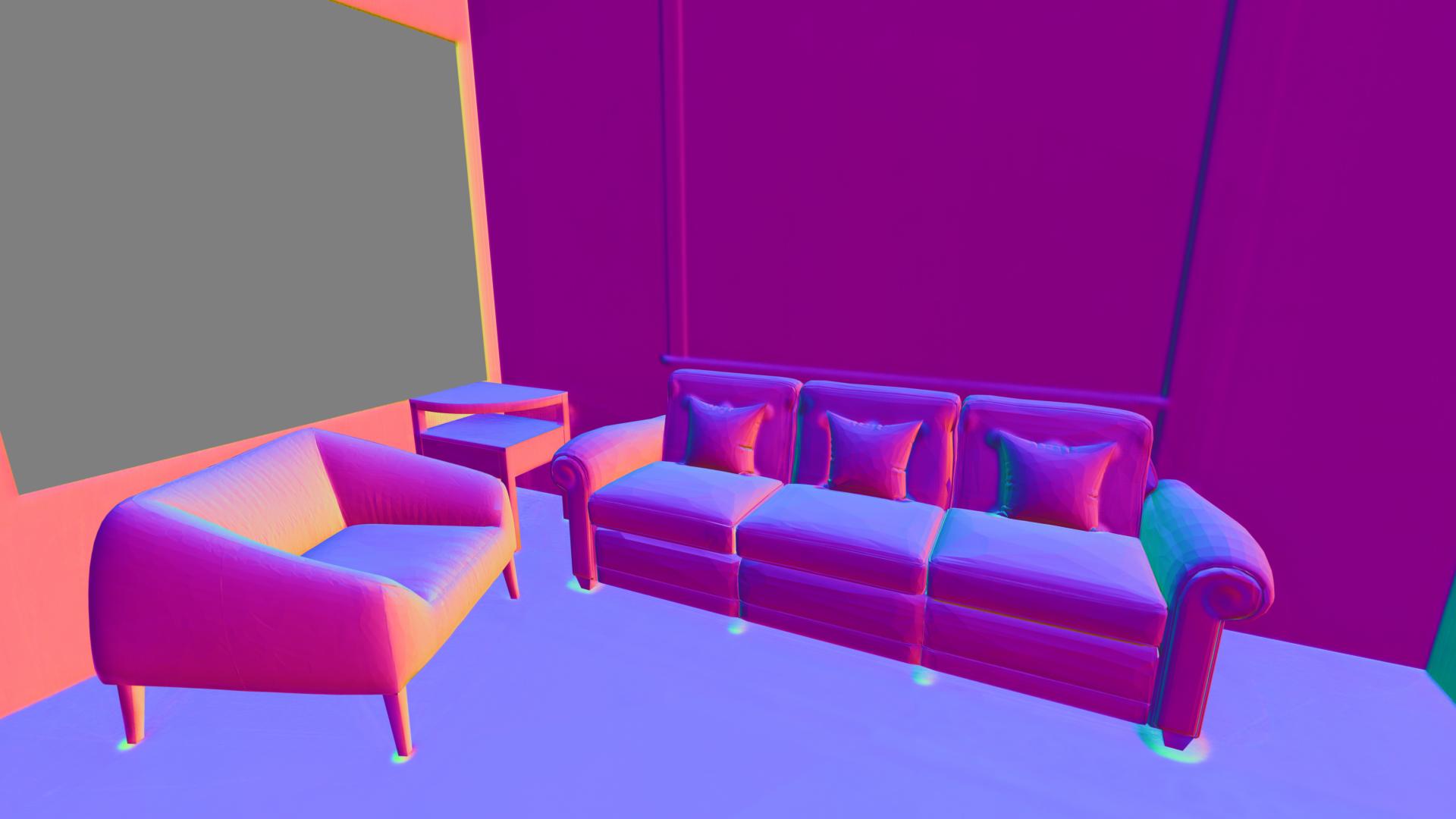}}
        &
        \fbox{\includegraphics[width=0.15\textwidth,trim={10cm 0 10cm 0},clip]{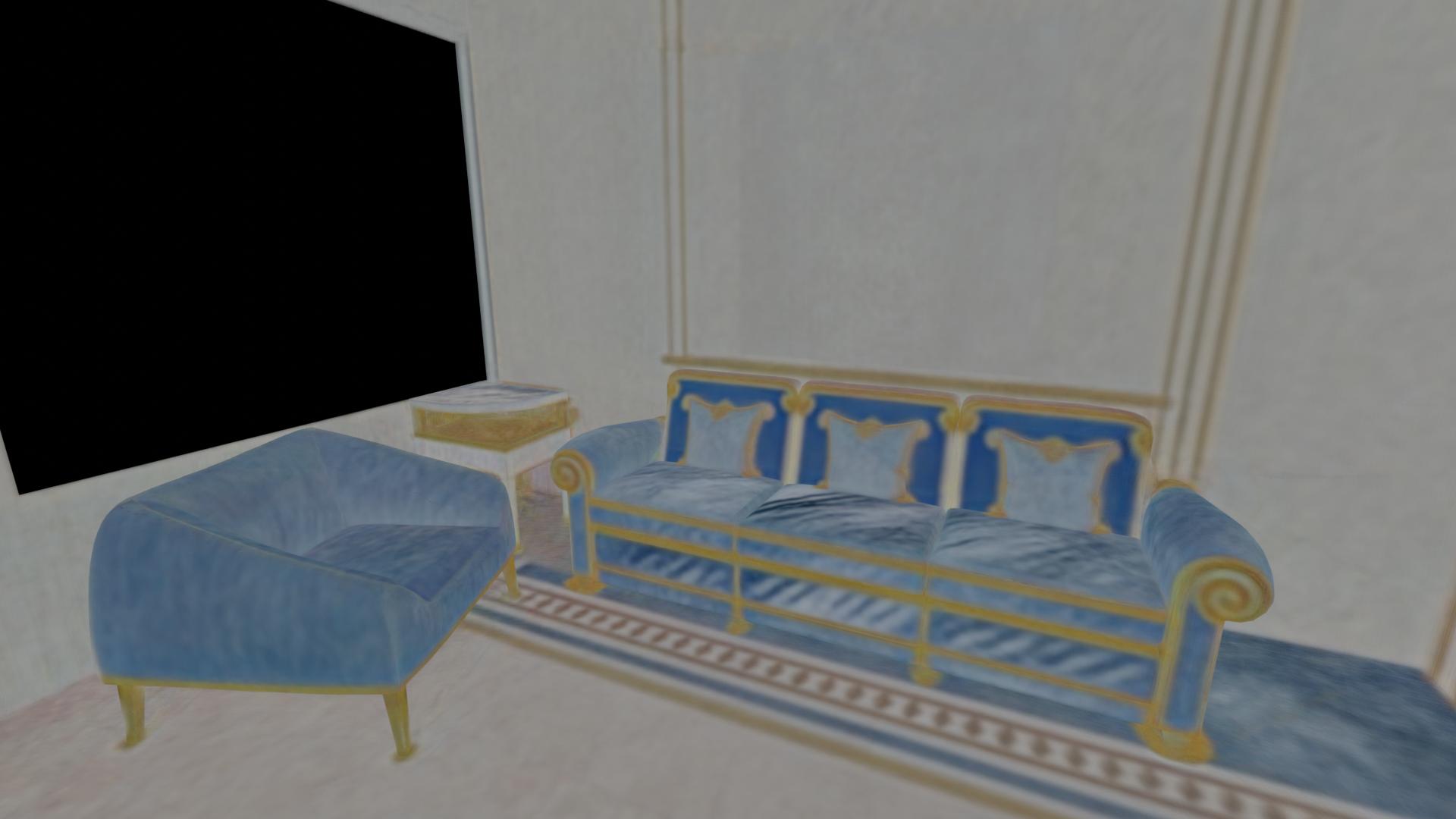}} 
        &
        \fbox{\includegraphics[width=0.15\textwidth,trim={10cm 0 10cm 0},clip]{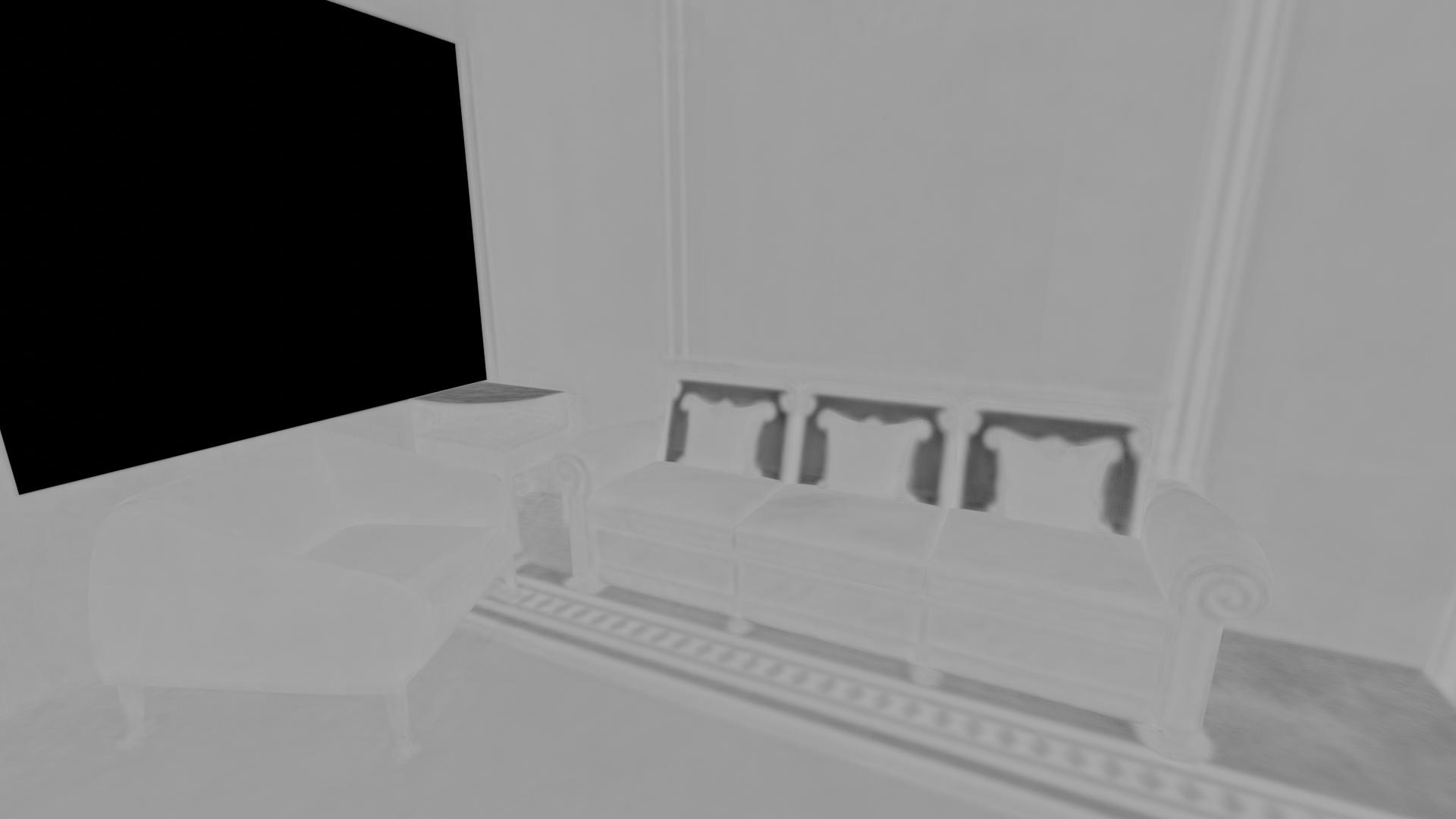}} 
        &
        \fbox{\includegraphics[width=0.15\textwidth,trim={10cm 0 10cm 0},clip]{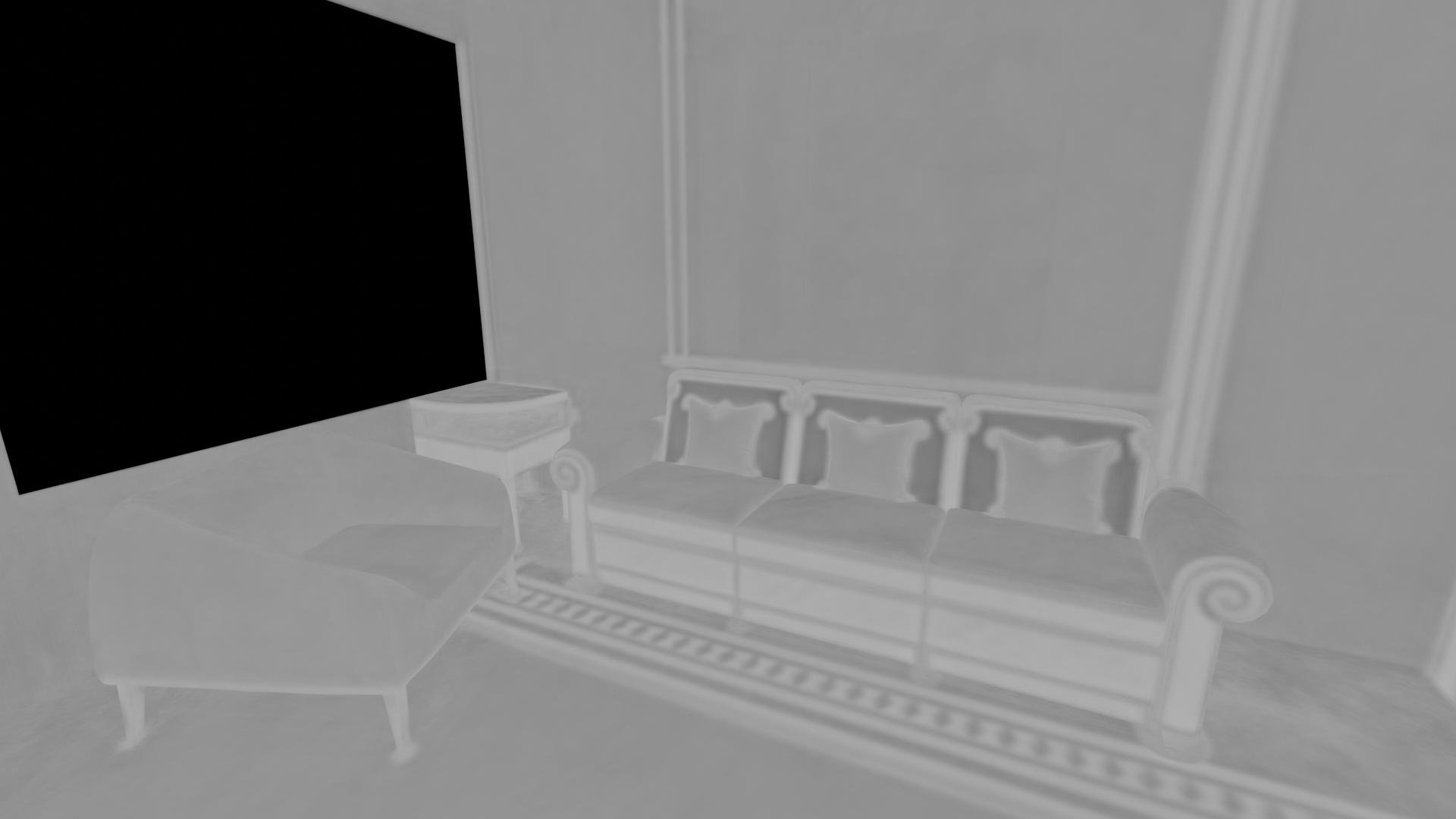}} 
        &
        \fbox{\includegraphics[width=0.15\textwidth,trim={10cm 0 10cm 0},clip]{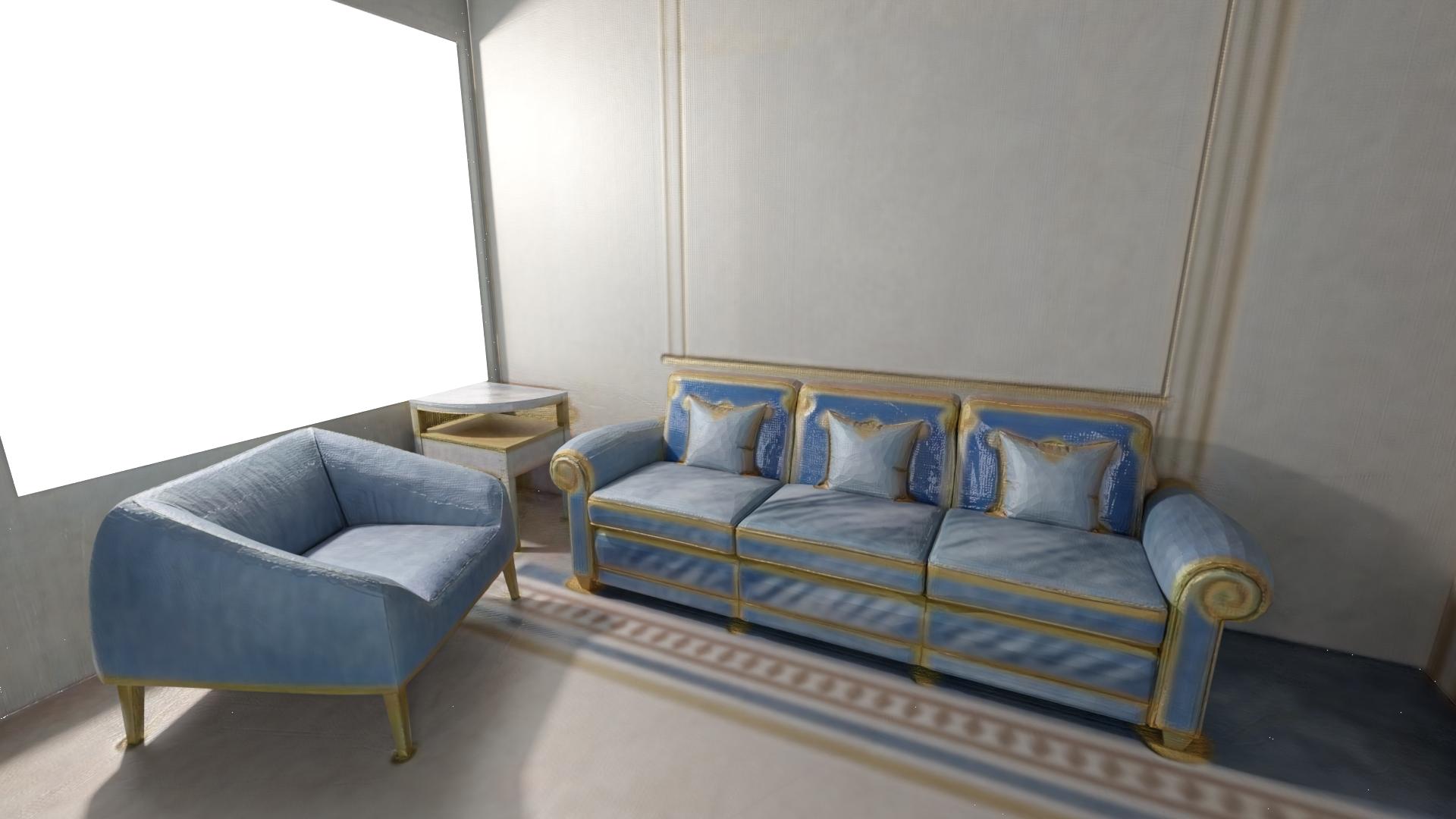}} 
        \\
        
        
        \rotatebox{90}{{\footnotesize View 3}}
        &
        \fbox{\includegraphics[width=0.15\textwidth,trim={10cm 0 10cm 0},clip]{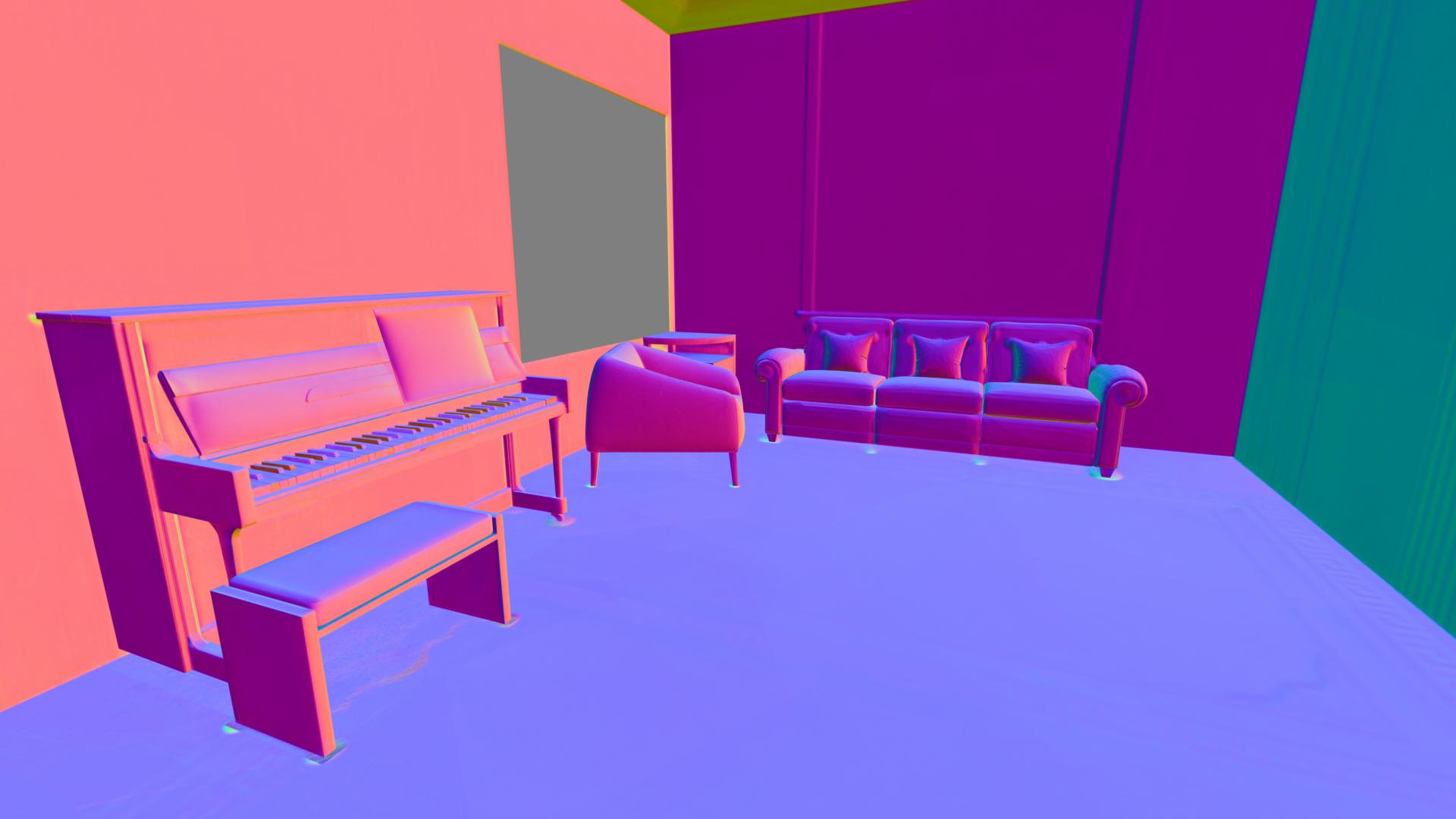}} 
        &
        \fbox{\includegraphics[width=0.15\textwidth,trim={10cm 0 10cm 0},clip]{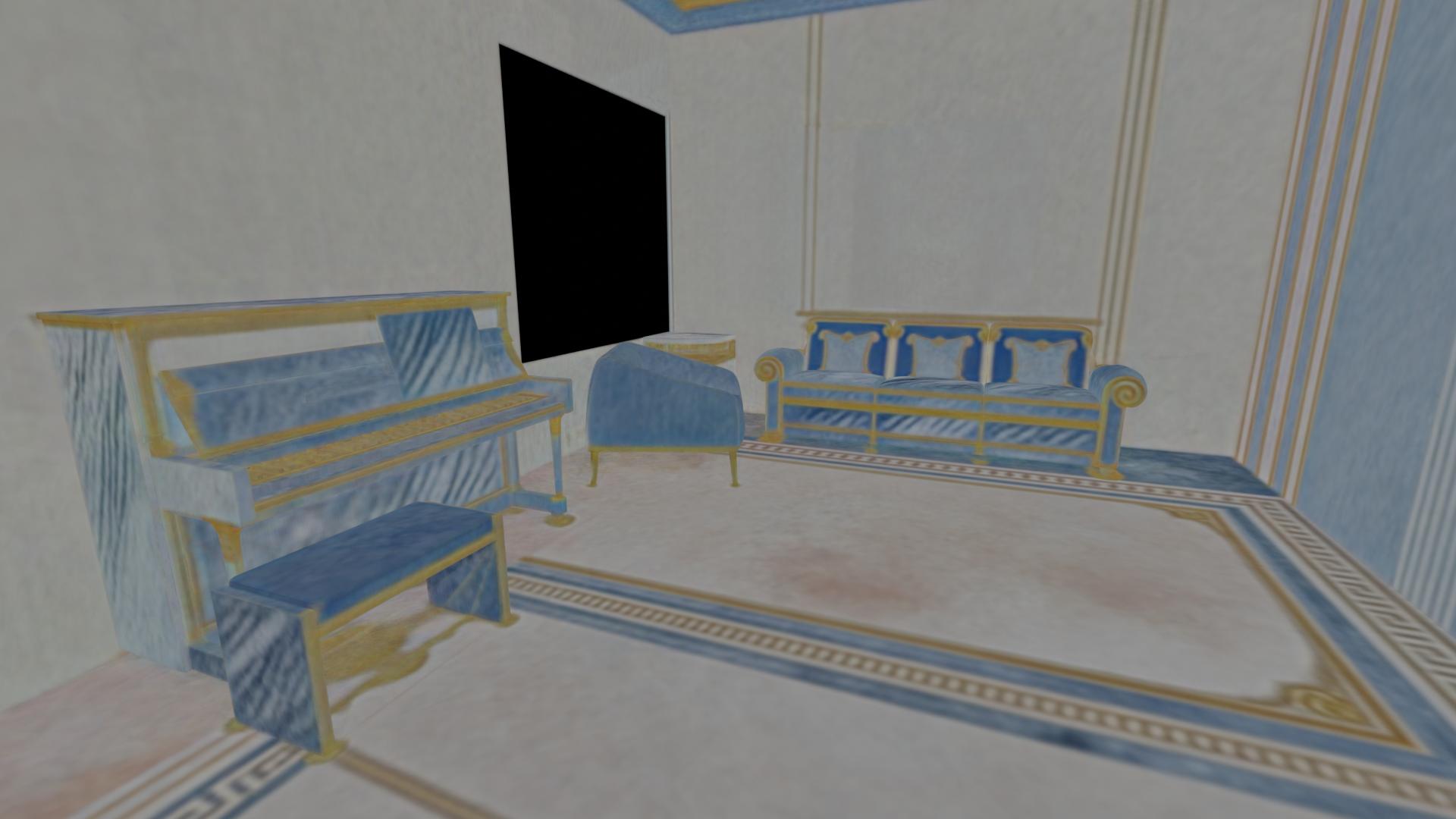}} 
        &
        \fbox{\includegraphics[width=0.15\textwidth,trim={10cm 0 10cm 0},clip]{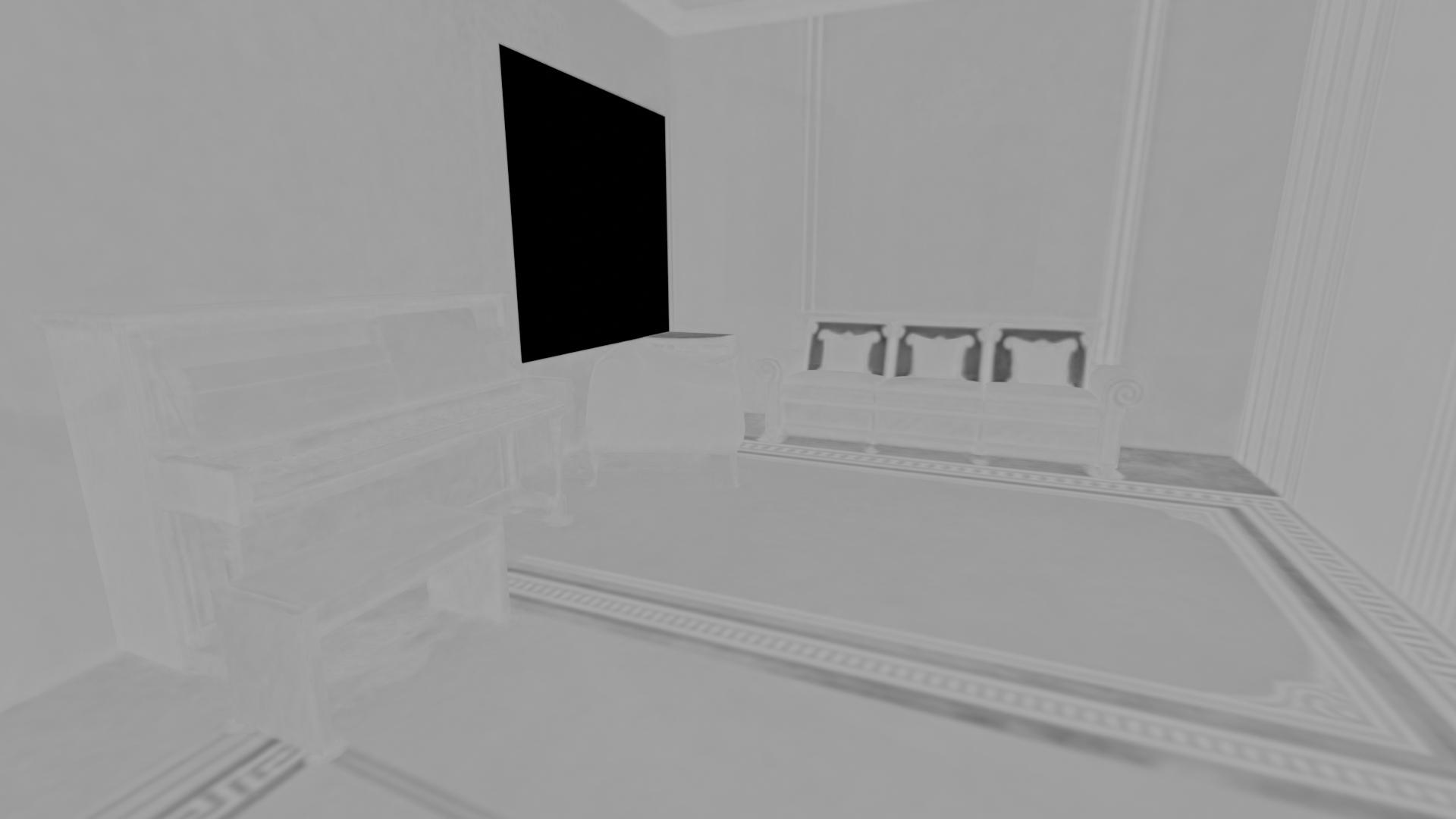}} 
        &
        \fbox{\includegraphics[width=0.15\textwidth,trim={10cm 0 10cm 0},clip]{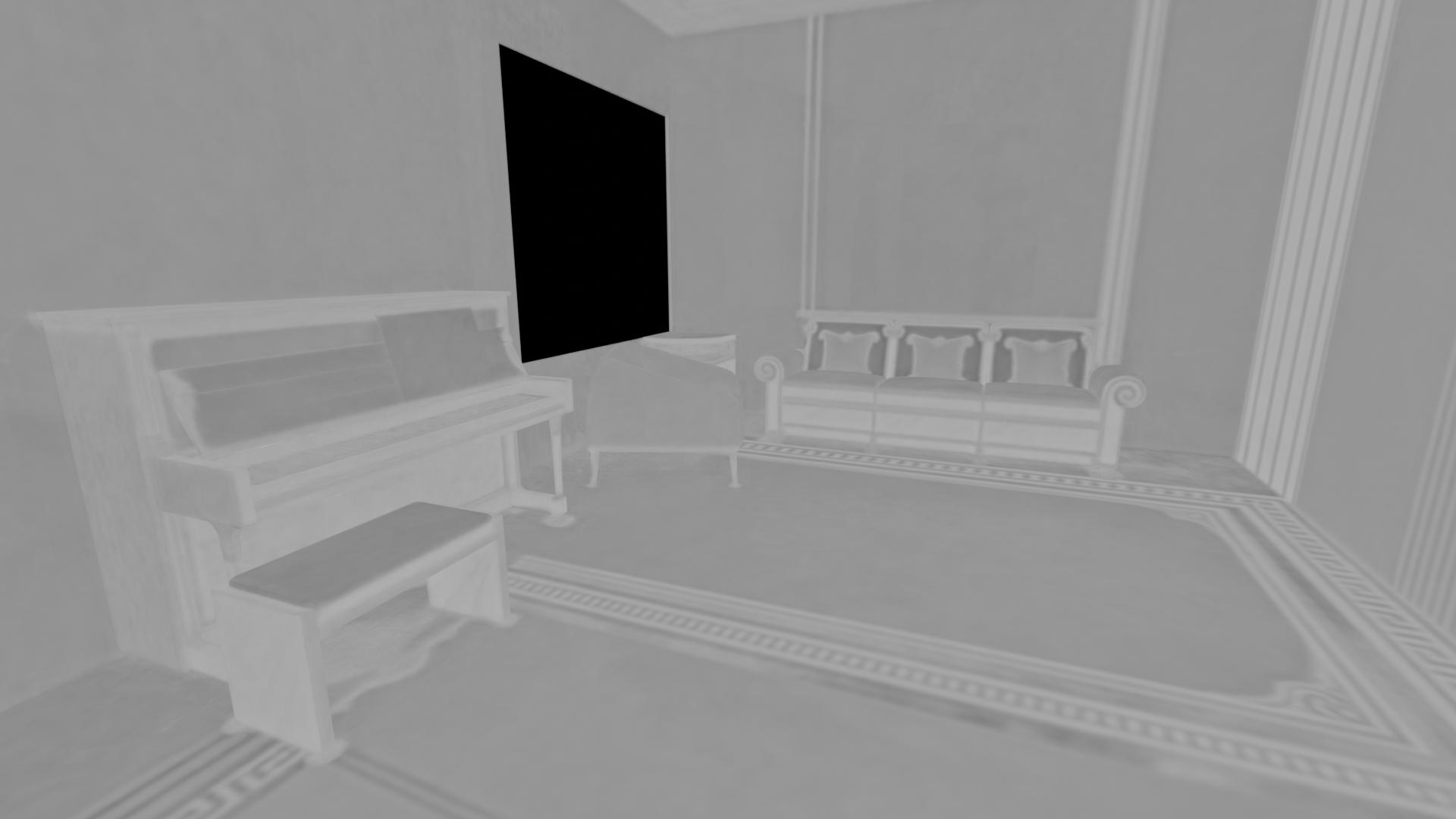}} 
        &
        \fbox{\includegraphics[width=0.15\textwidth,trim={10cm 0 10cm 0},clip]{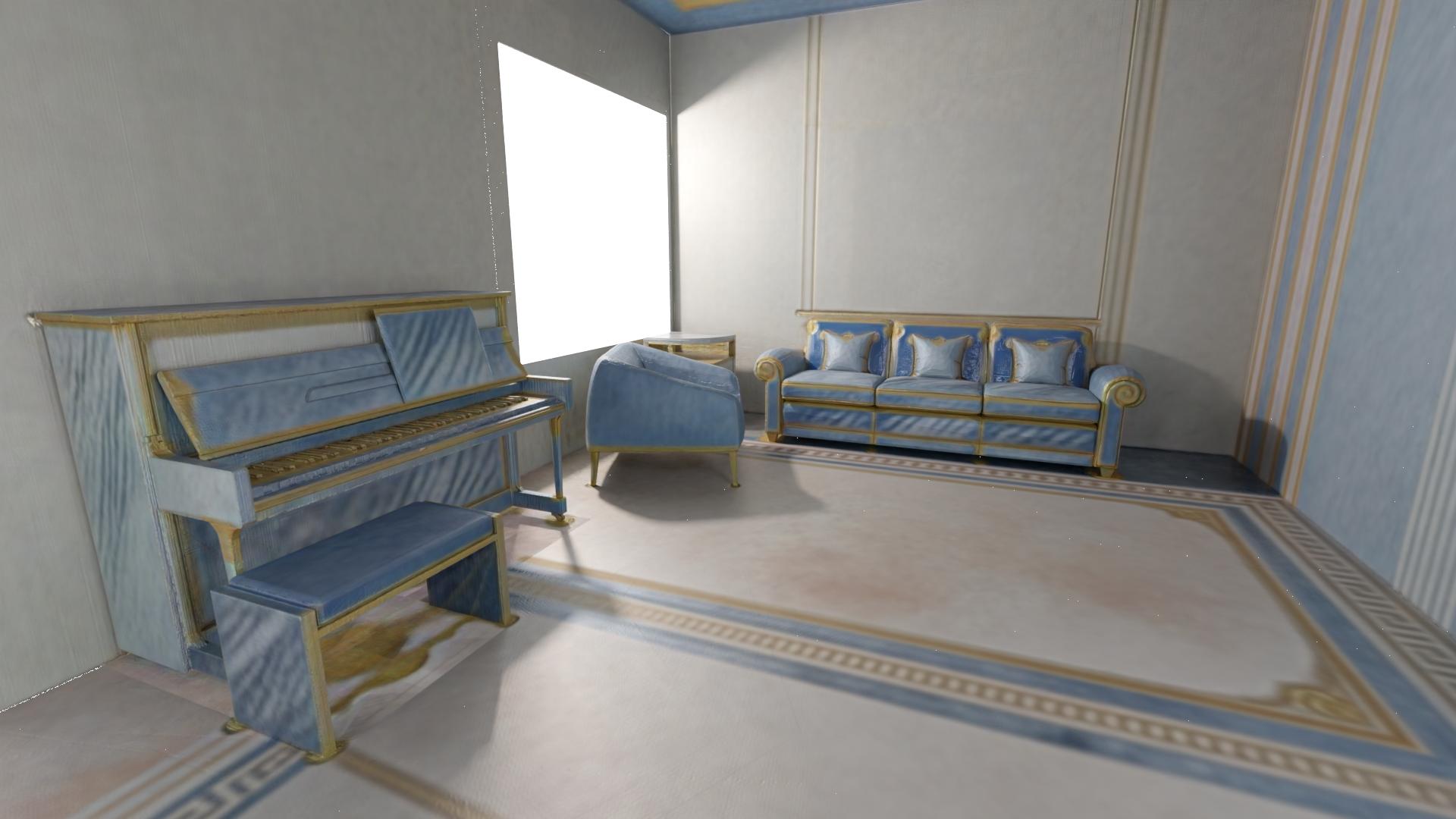}} 
        \\

        &
        {\footnotesize Normal} &
        {\footnotesize Albedo} &
        {\footnotesize Roughness} &
        {\footnotesize Metallic} &
        {\footnotesize Rendering} \\
    \end{tabular}}
    \caption{\textbf{Scene Texturing}. 
    We show more scene texturing results on multiple 3D-Front scenes \cite{Front3d} with multiple prompts. Continues on the next page.
    }
\end{figure*}
\begin{figure*}[p]\ContinuedFloat
  \centering
    \setlength\tabcolsep{1.25pt}
    \resizebox{\textwidth}{!}{
    \fboxsep=0pt
    \begin{tabular}{ccccc|c}
        \rotatebox{90}{{\footnotesize View 1}}
        &
        \fbox{\includegraphics[width=0.15\textwidth,trim={10cm 0 10cm 0},clip]{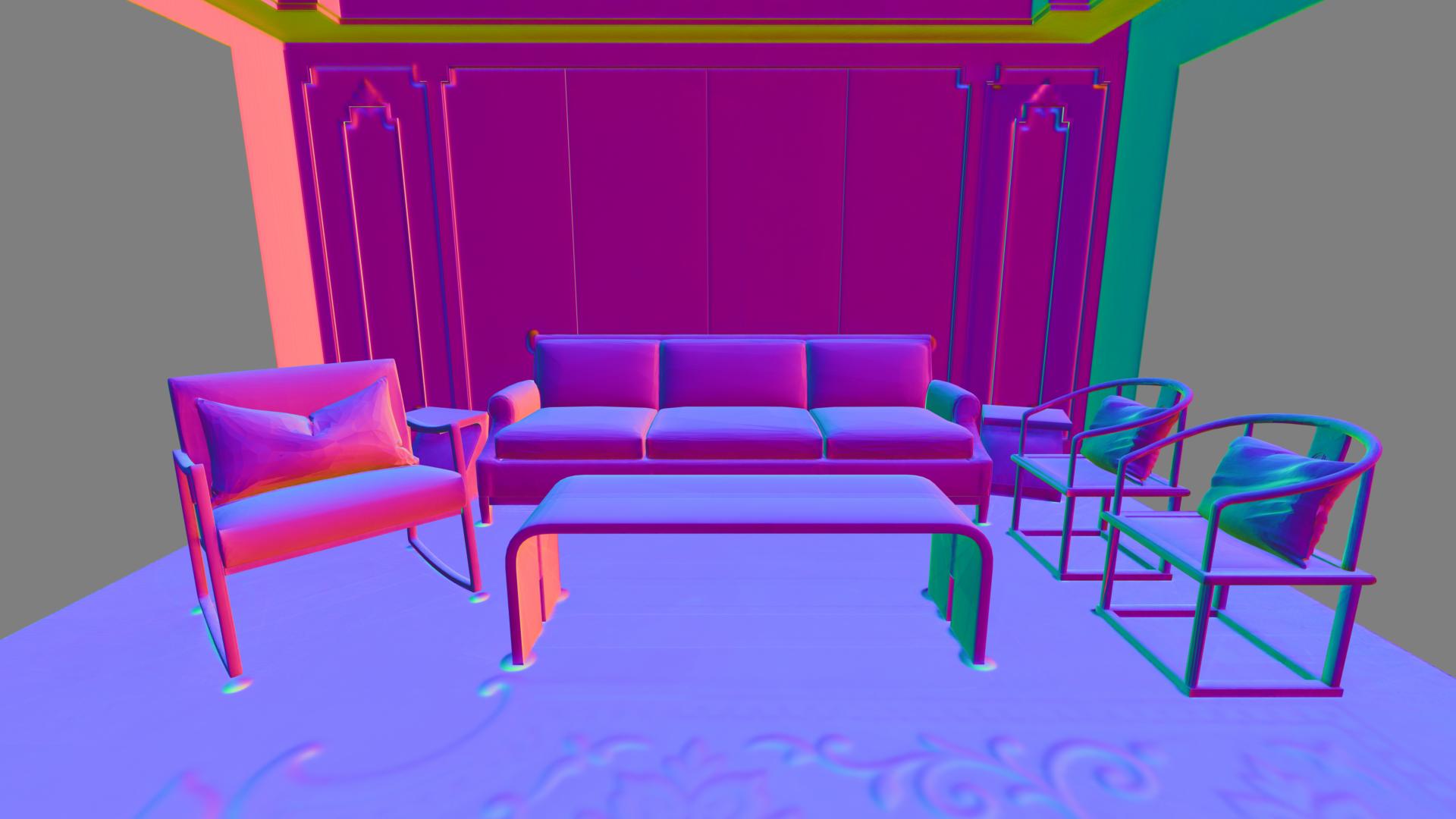}}
        &
        \fbox{\includegraphics[width=0.15\textwidth,trim={10cm 0 10cm 0},clip]{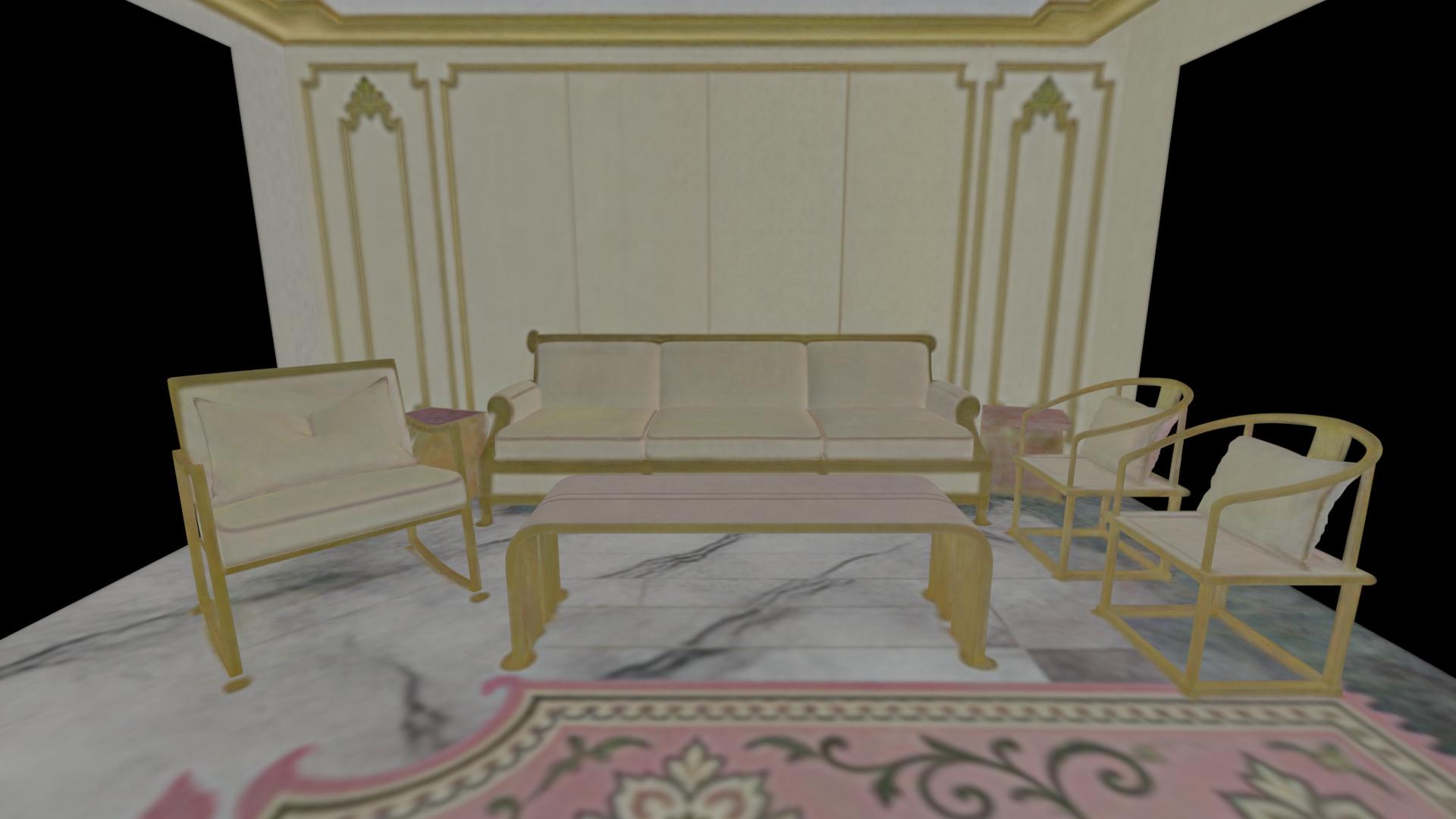}} 
        &
        \fbox{\includegraphics[width=0.15\textwidth,trim={10cm 0 10cm 0},clip]{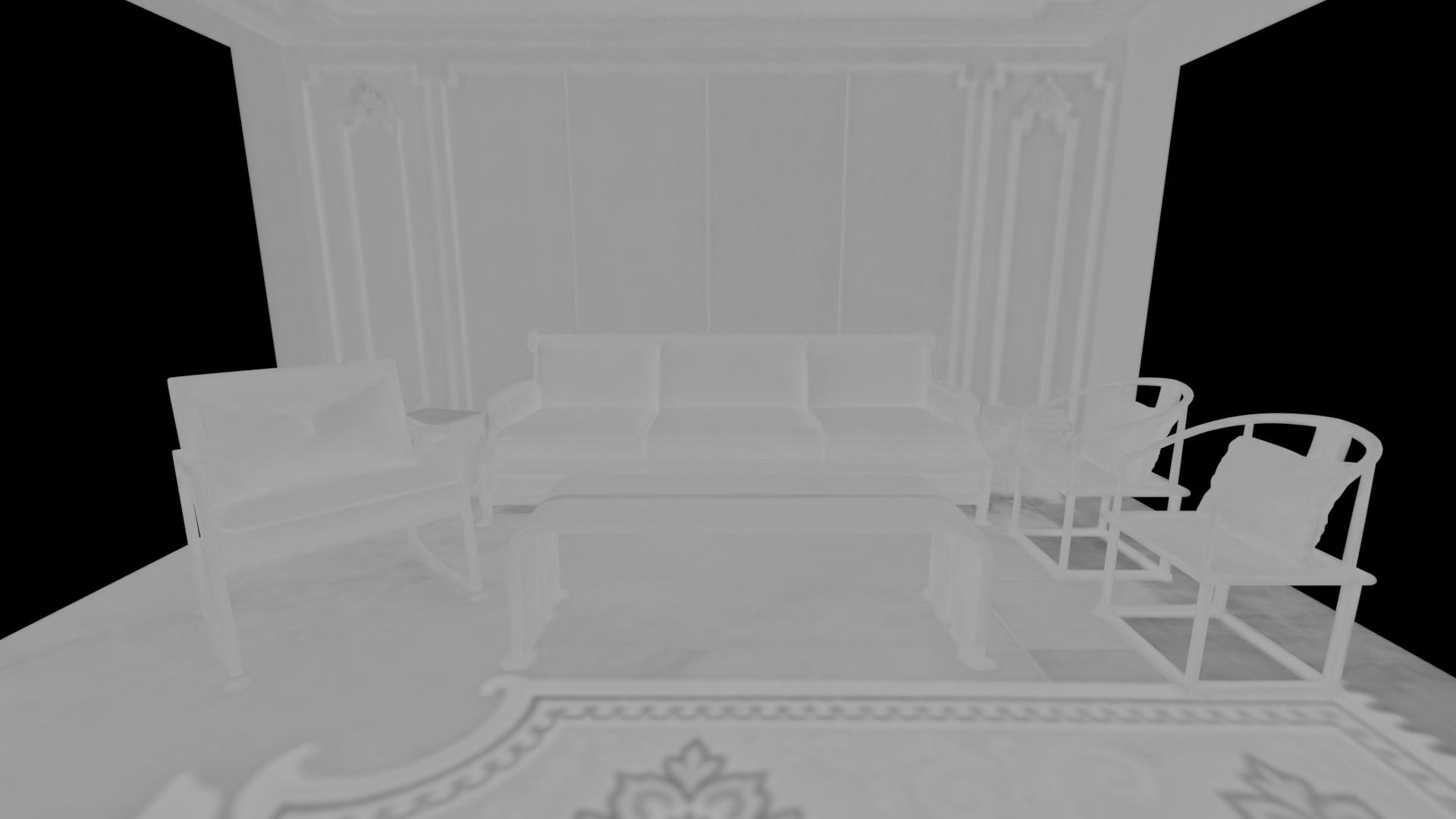}} 
        &
        \fbox{\includegraphics[width=0.15\textwidth,trim={10cm 0 10cm 0},clip]{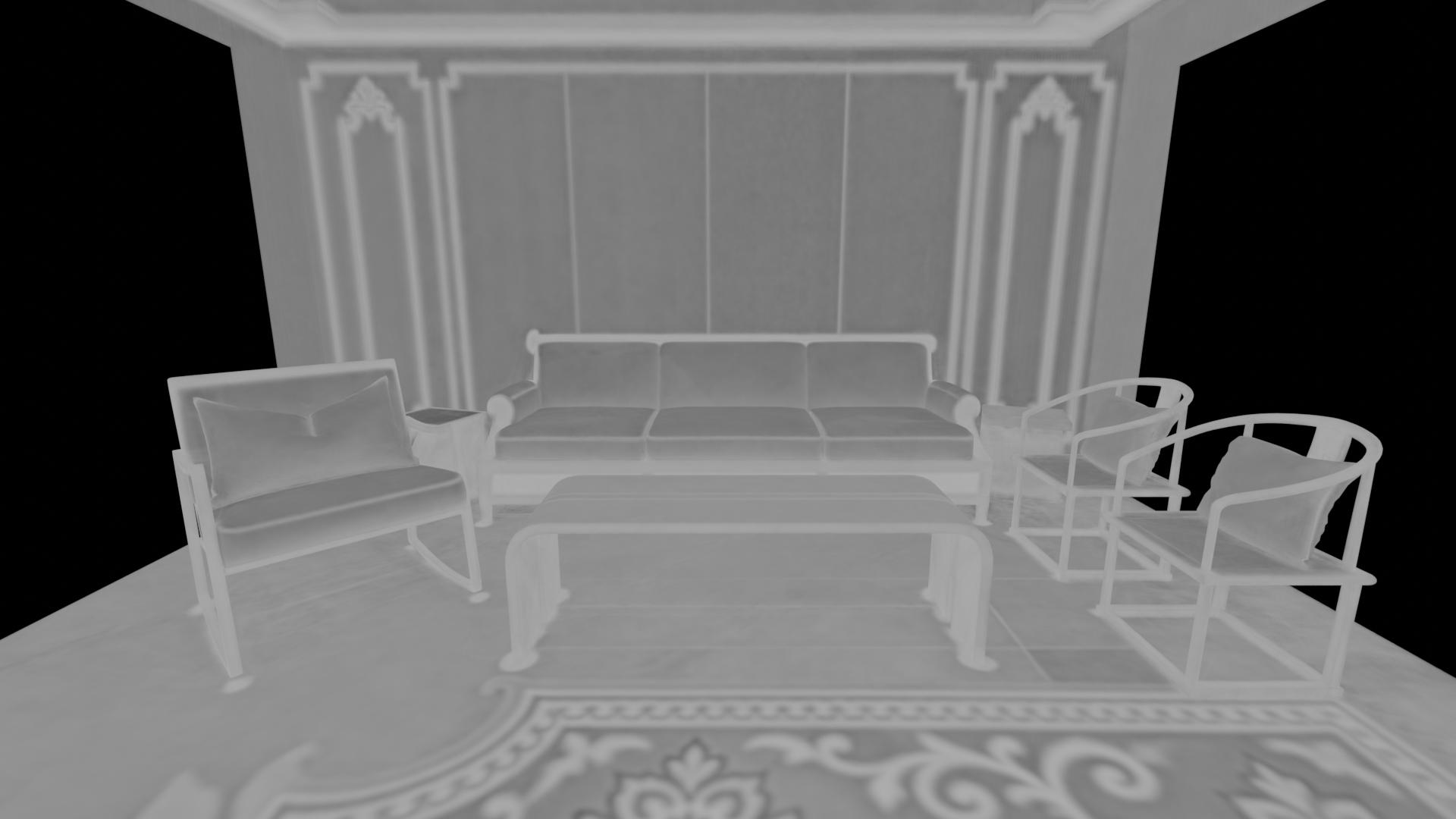}} 
        &
        \fbox{\includegraphics[width=0.15\textwidth,trim={10cm 0 10cm 0},clip]{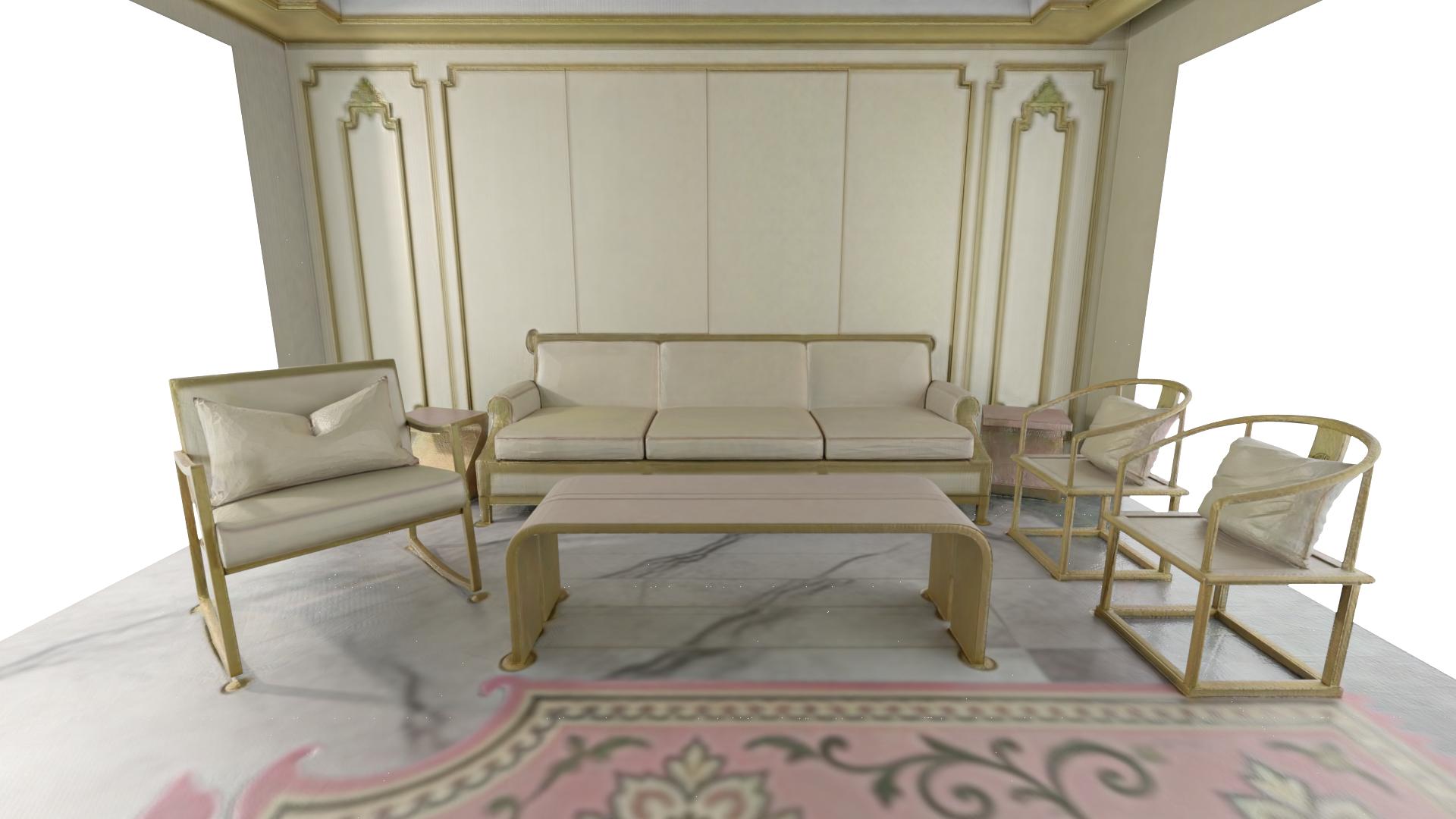}} 
        \\
        
        
        \rotatebox{90}{{\footnotesize View 3}}
        &
        \fbox{\includegraphics[width=0.15\textwidth,trim={10cm 0 10cm 0},clip]{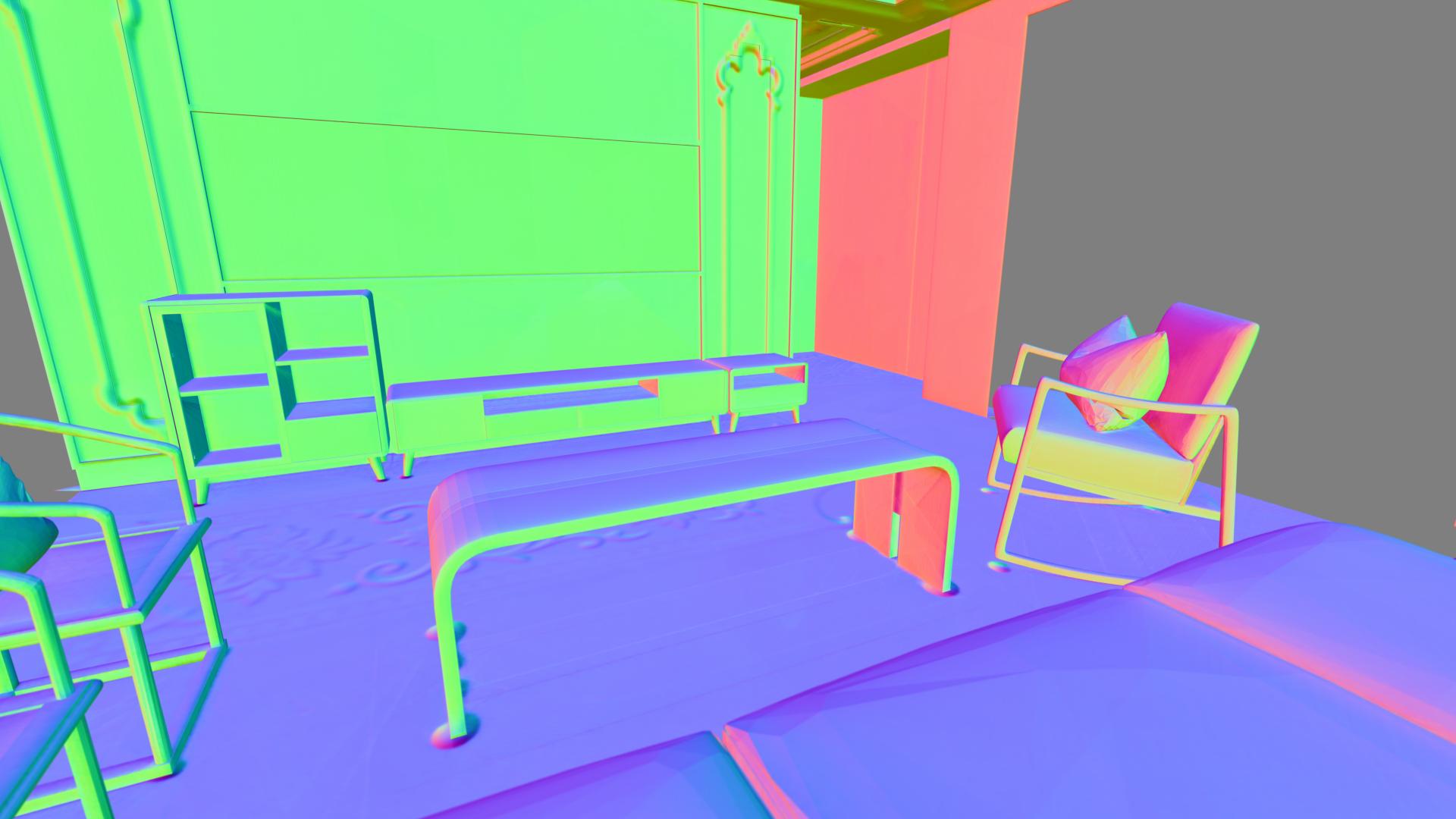}} 
        &
        \fbox{\includegraphics[width=0.15\textwidth,trim={10cm 0 10cm 0},clip]{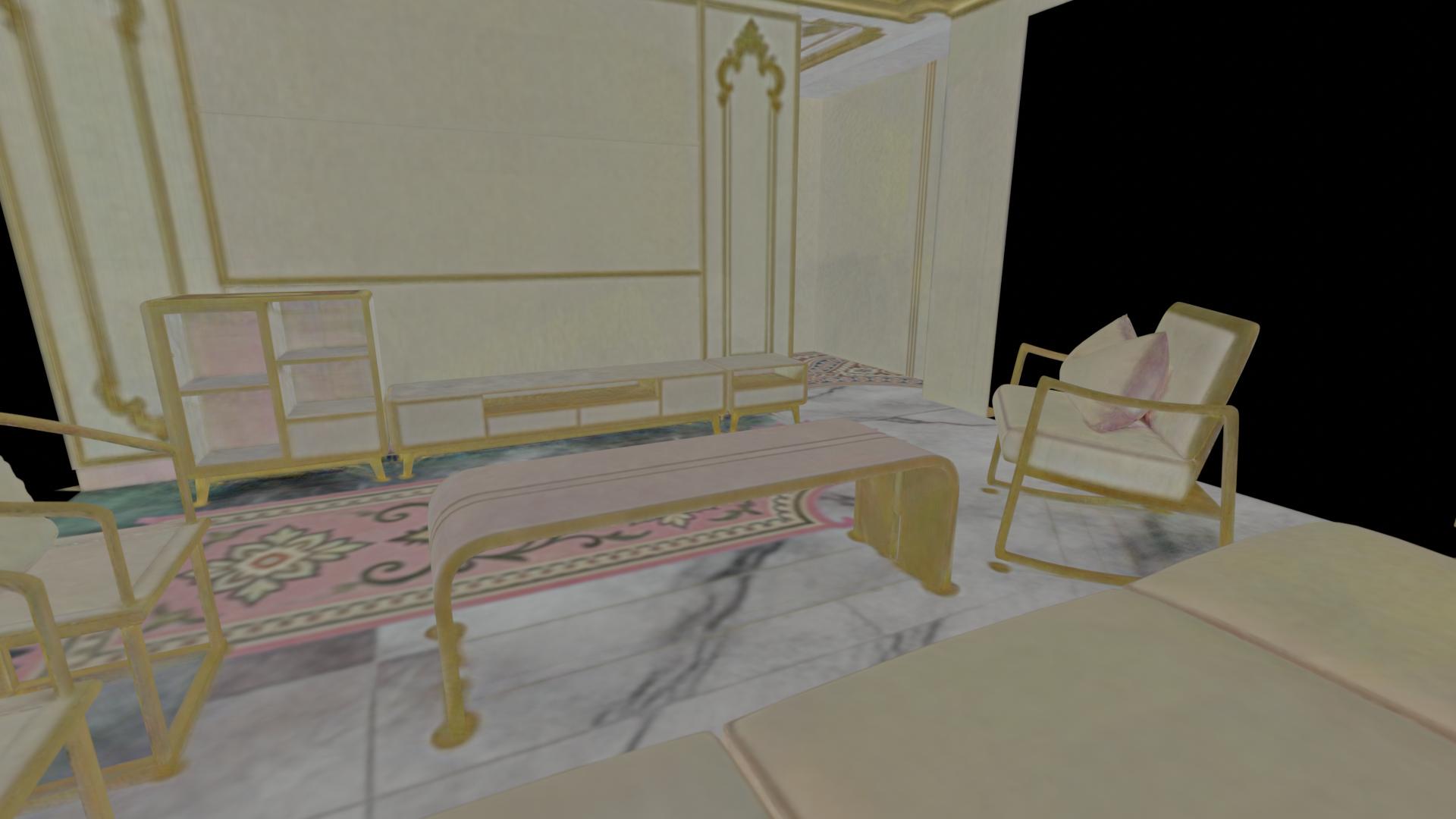}} 
        &
        \fbox{\includegraphics[width=0.15\textwidth,trim={10cm 0 10cm 0},clip]{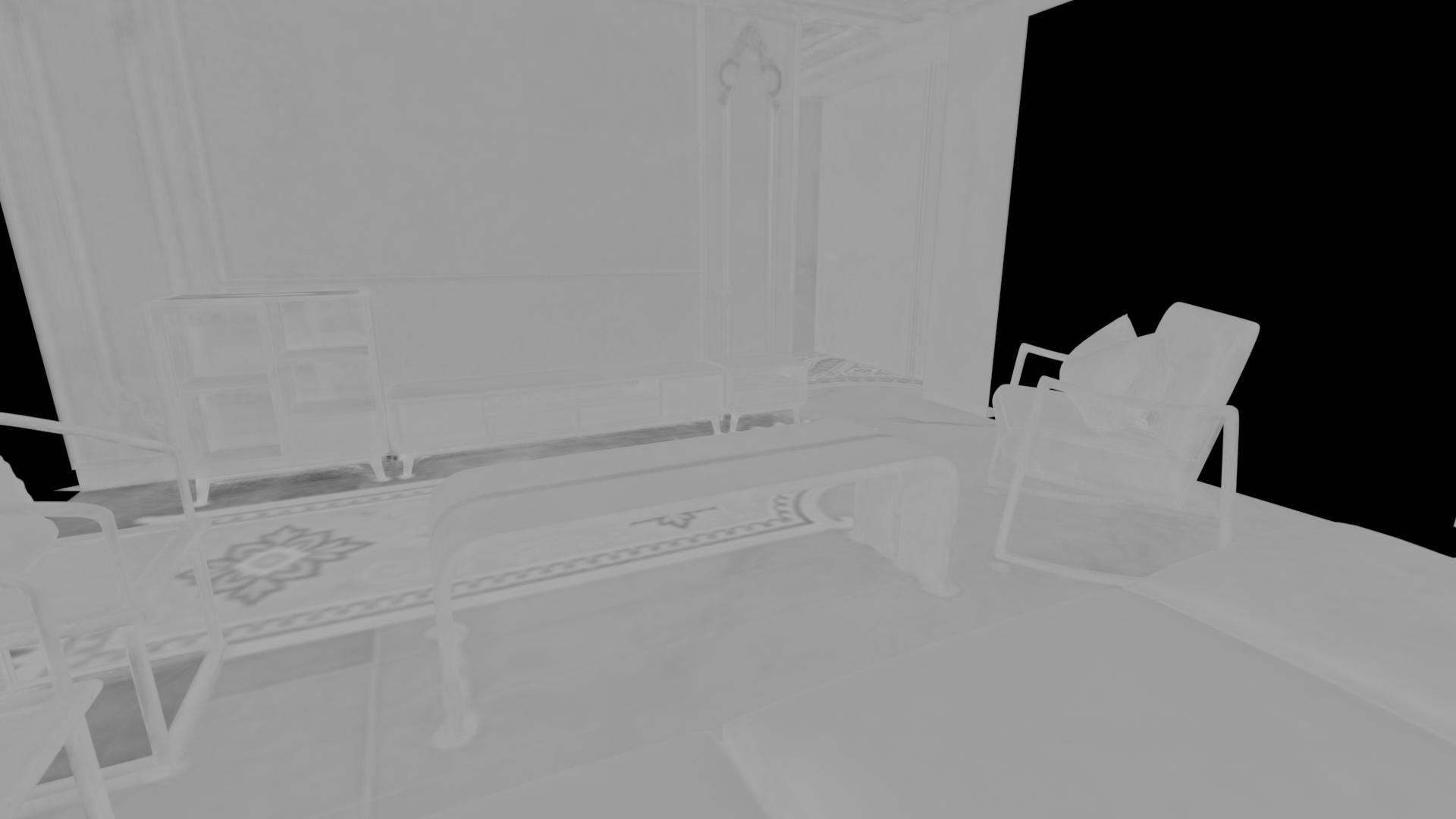}} 
        &
        \fbox{\includegraphics[width=0.15\textwidth,trim={10cm 0 10cm 0},clip]{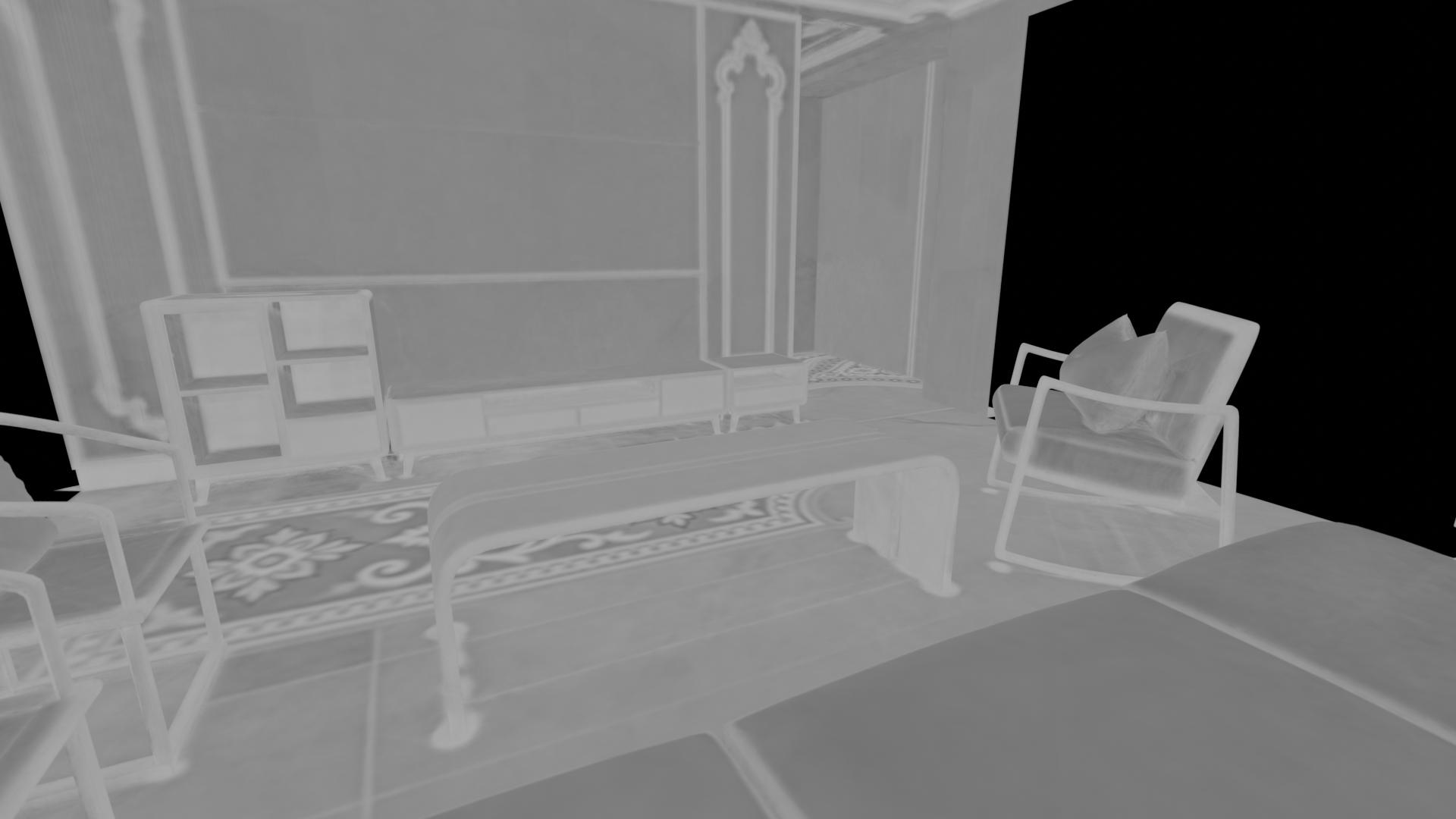}} 
        &
        \fbox{\includegraphics[width=0.15\textwidth,trim={10cm 0 10cm 0},clip]{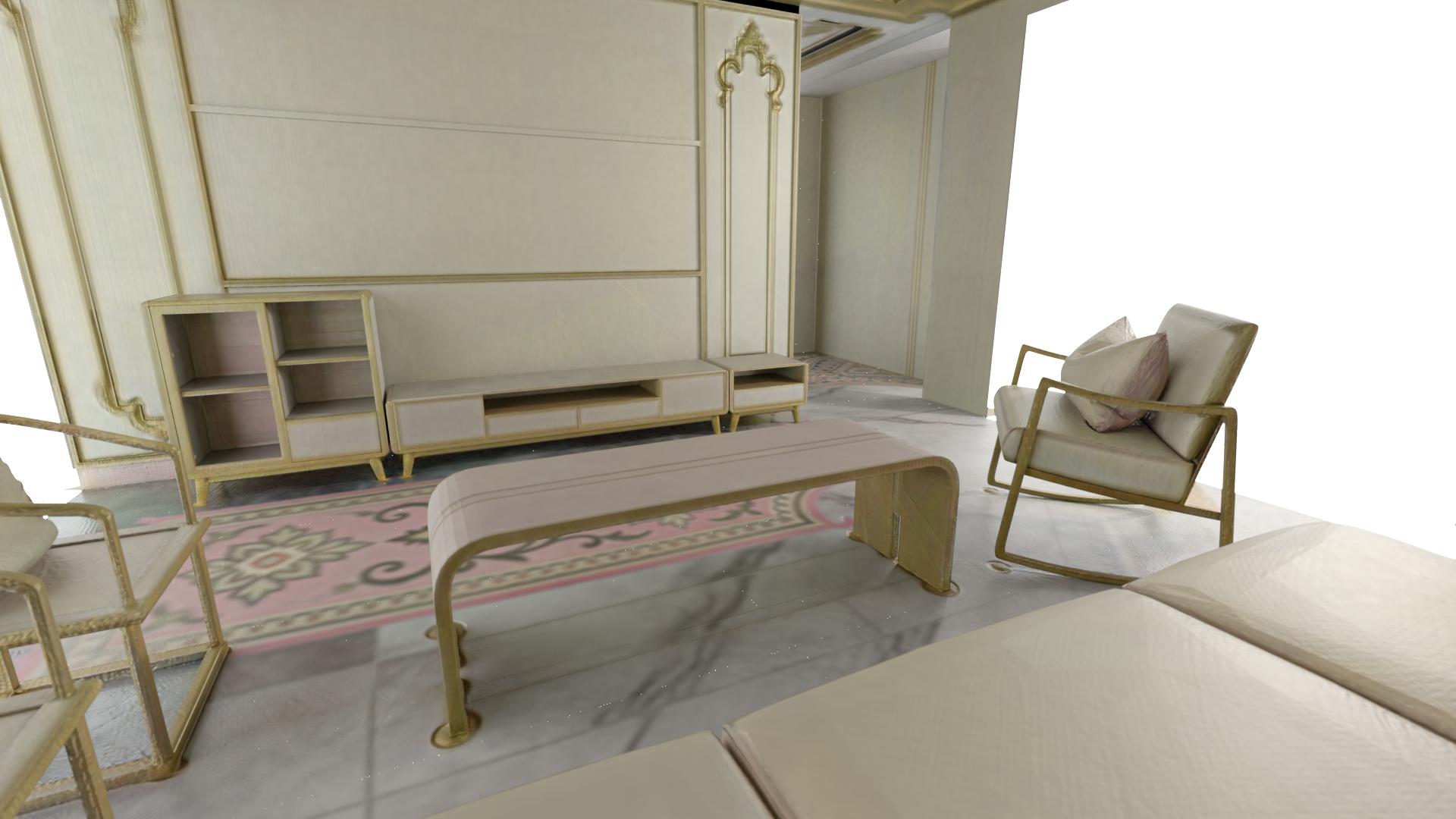}} 
        \\

        &
        {\footnotesize Normal} &
        {\footnotesize Albedo} &
        {\footnotesize Roughness} &
        {\footnotesize Metallic} &
        {\footnotesize Rendering} \\

        \midrule
        
        \rotatebox{90}{{\footnotesize View 1}}
        &
        \fbox{\includegraphics[width=0.15\textwidth,trim={10cm 0 10cm 0},clip]{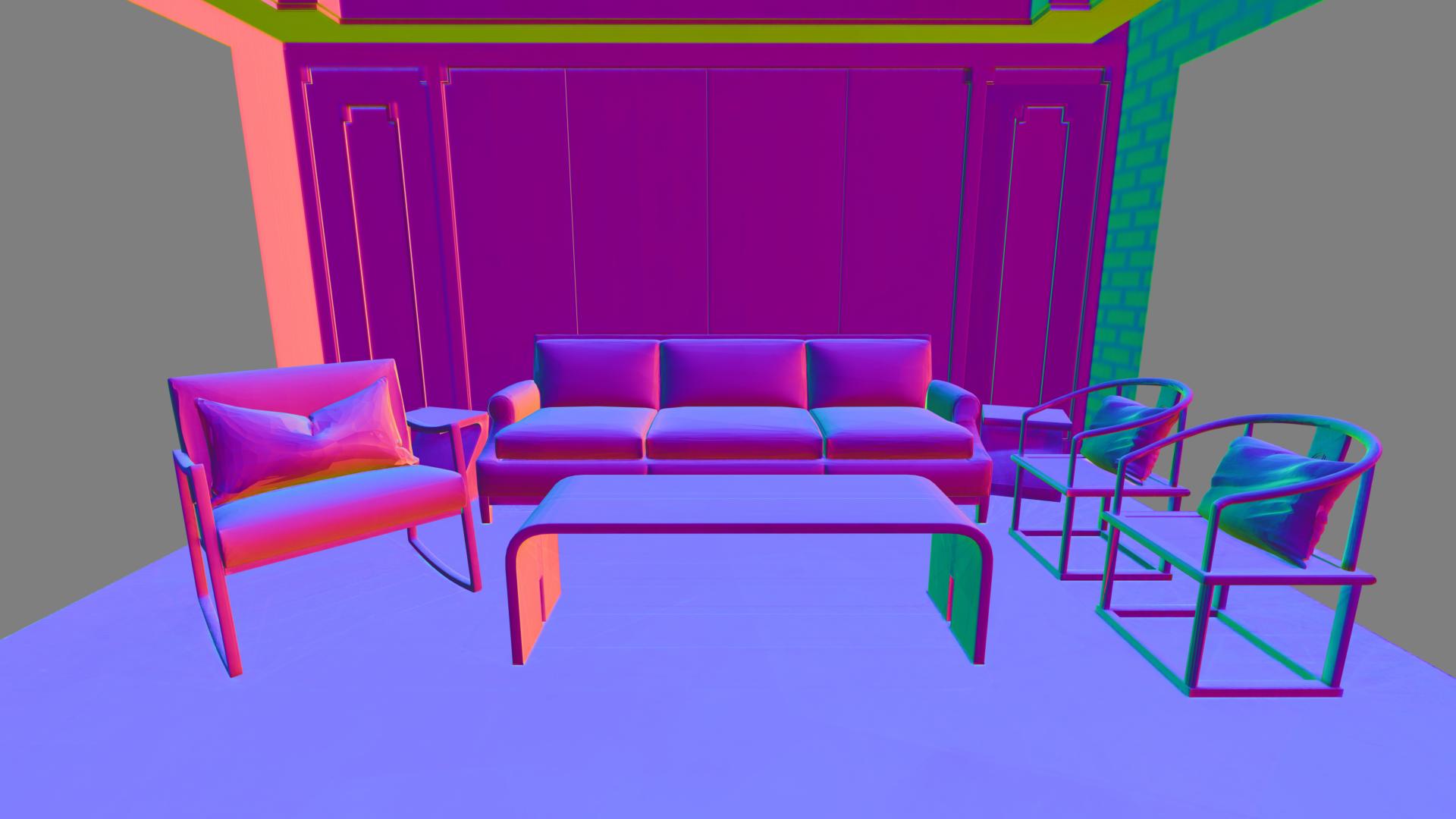}}
        &
        \fbox{\includegraphics[width=0.15\textwidth,trim={10cm 0 10cm 0},clip]{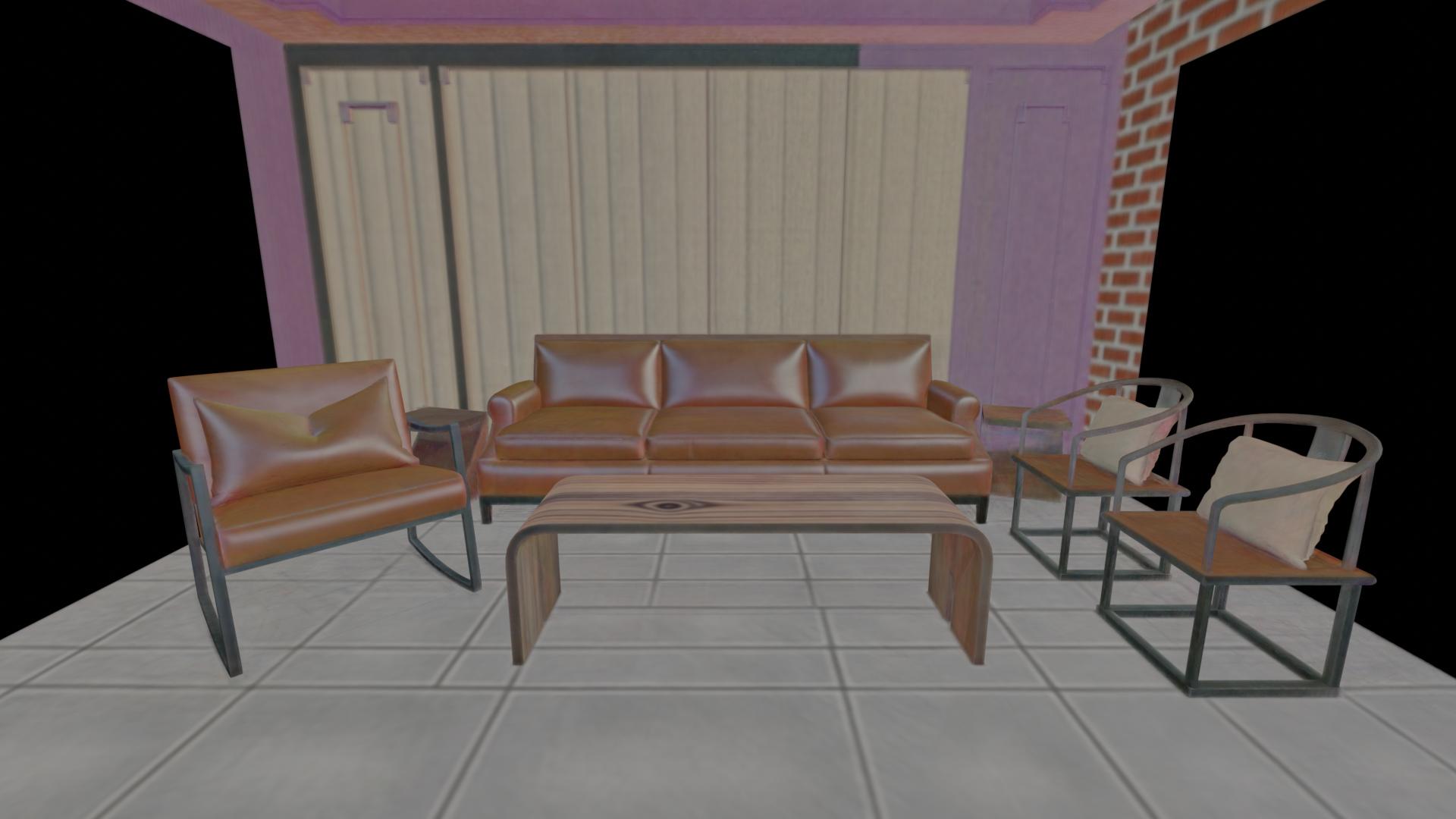}} 
        &
        \fbox{\includegraphics[width=0.15\textwidth,trim={10cm 0 10cm 0},clip]{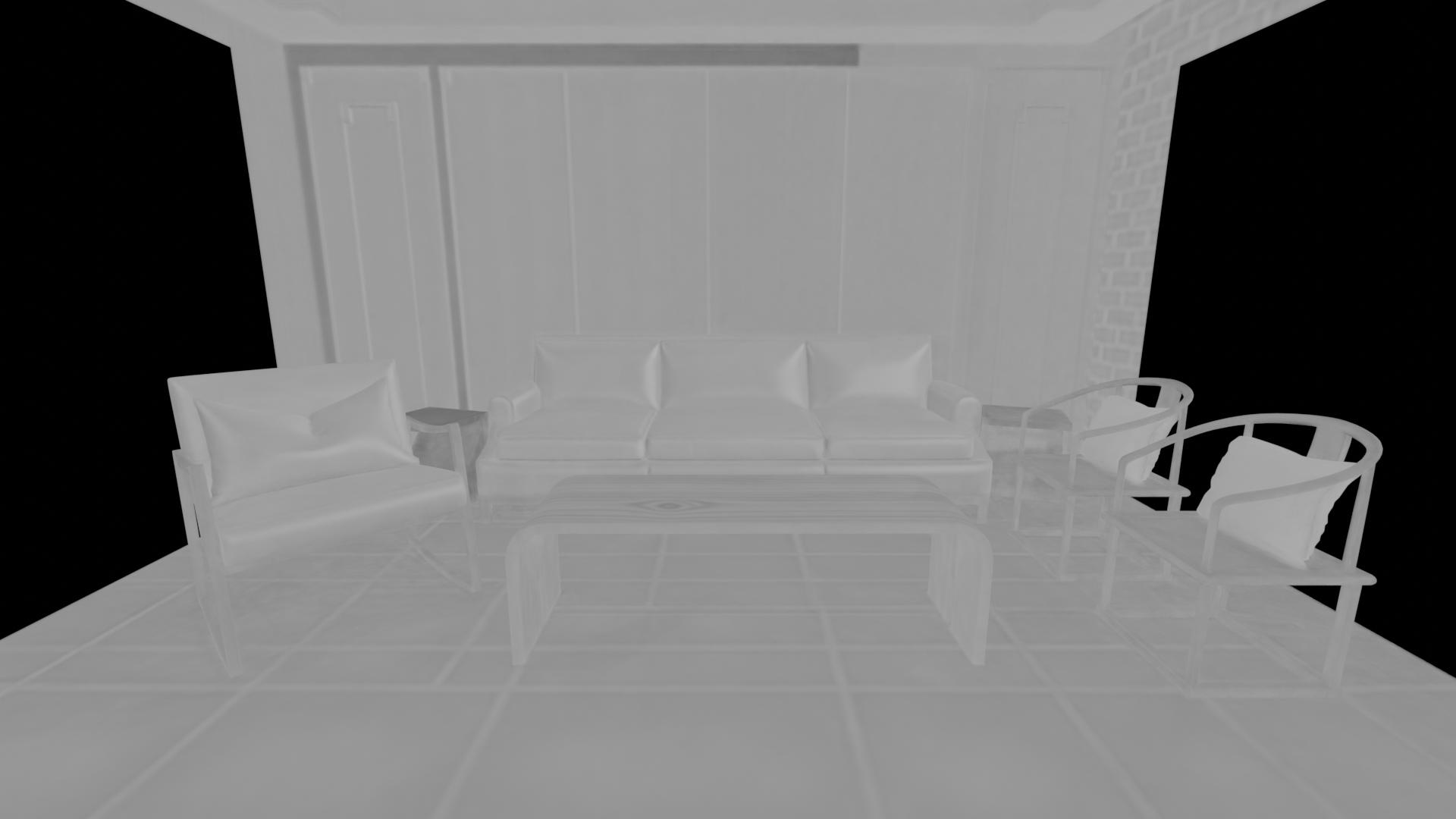}} 
        &
        \fbox{\includegraphics[width=0.15\textwidth,trim={10cm 0 10cm 0},clip]{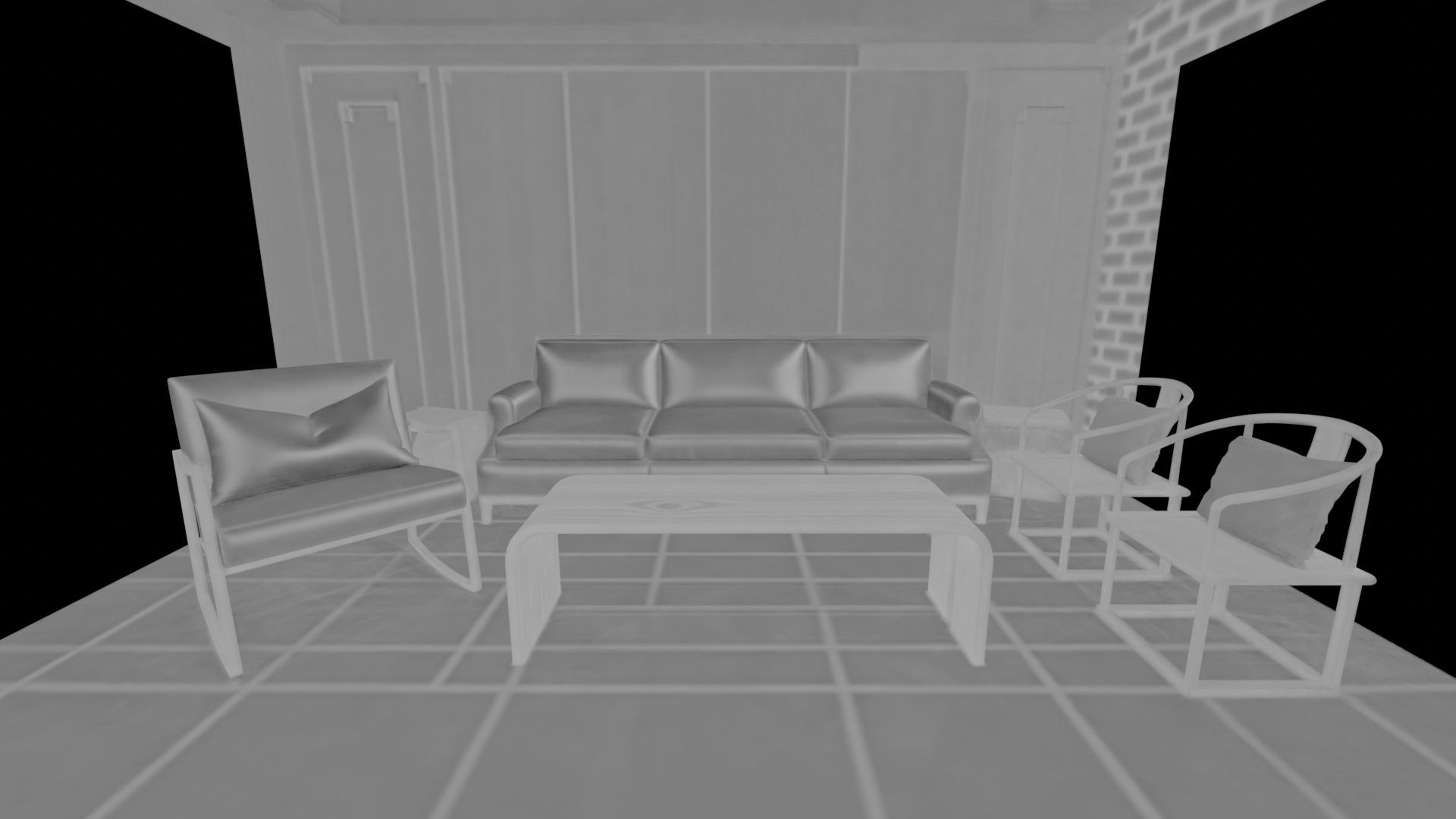}} 
        &
        \fbox{\includegraphics[width=0.15\textwidth,trim={10cm 0 10cm 0},clip]{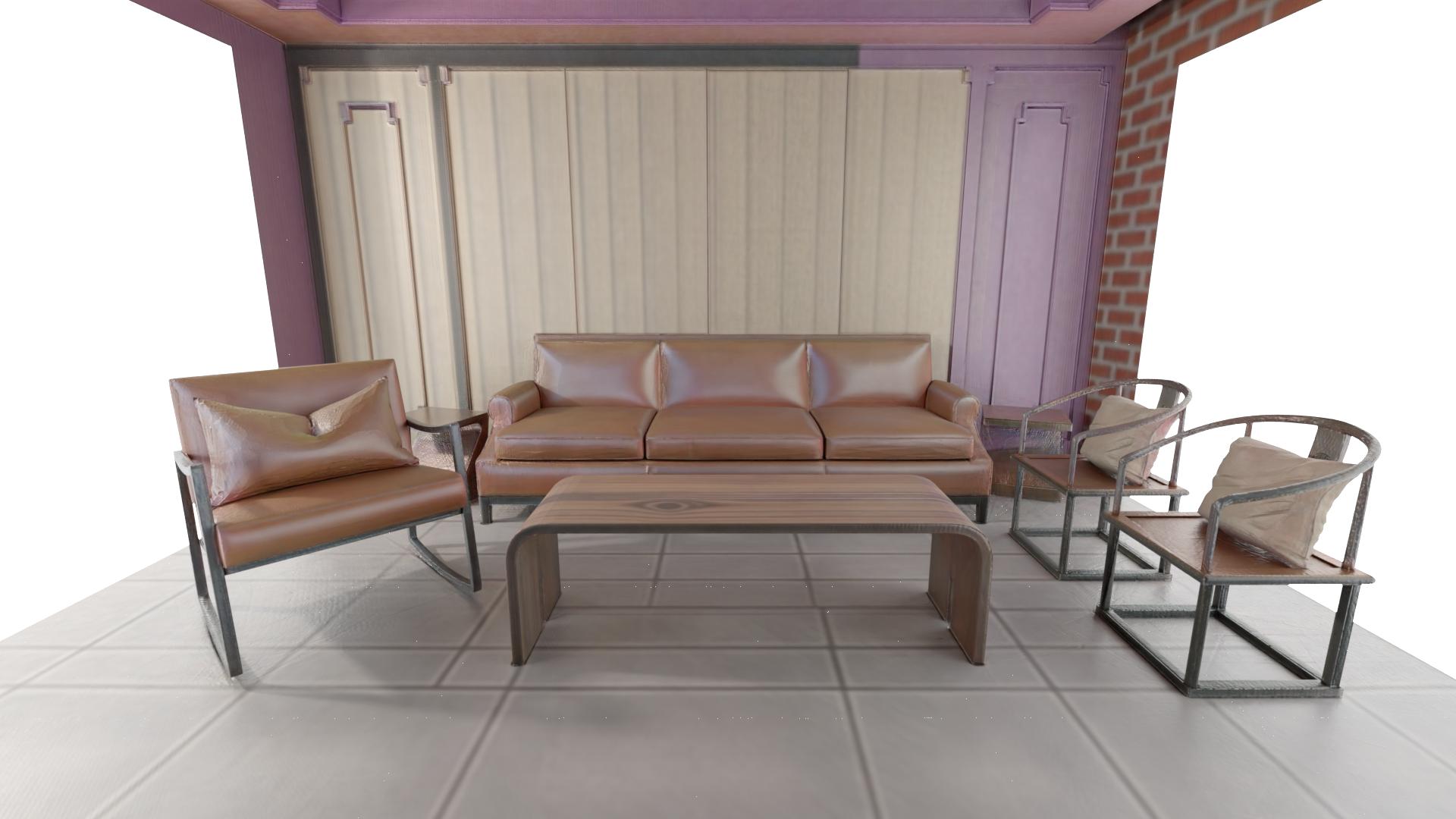}} 
        \\
        
        
        \rotatebox{90}{{\footnotesize View 3}}
        &
        \fbox{\includegraphics[width=0.15\textwidth,trim={10cm 0 10cm 0},clip]{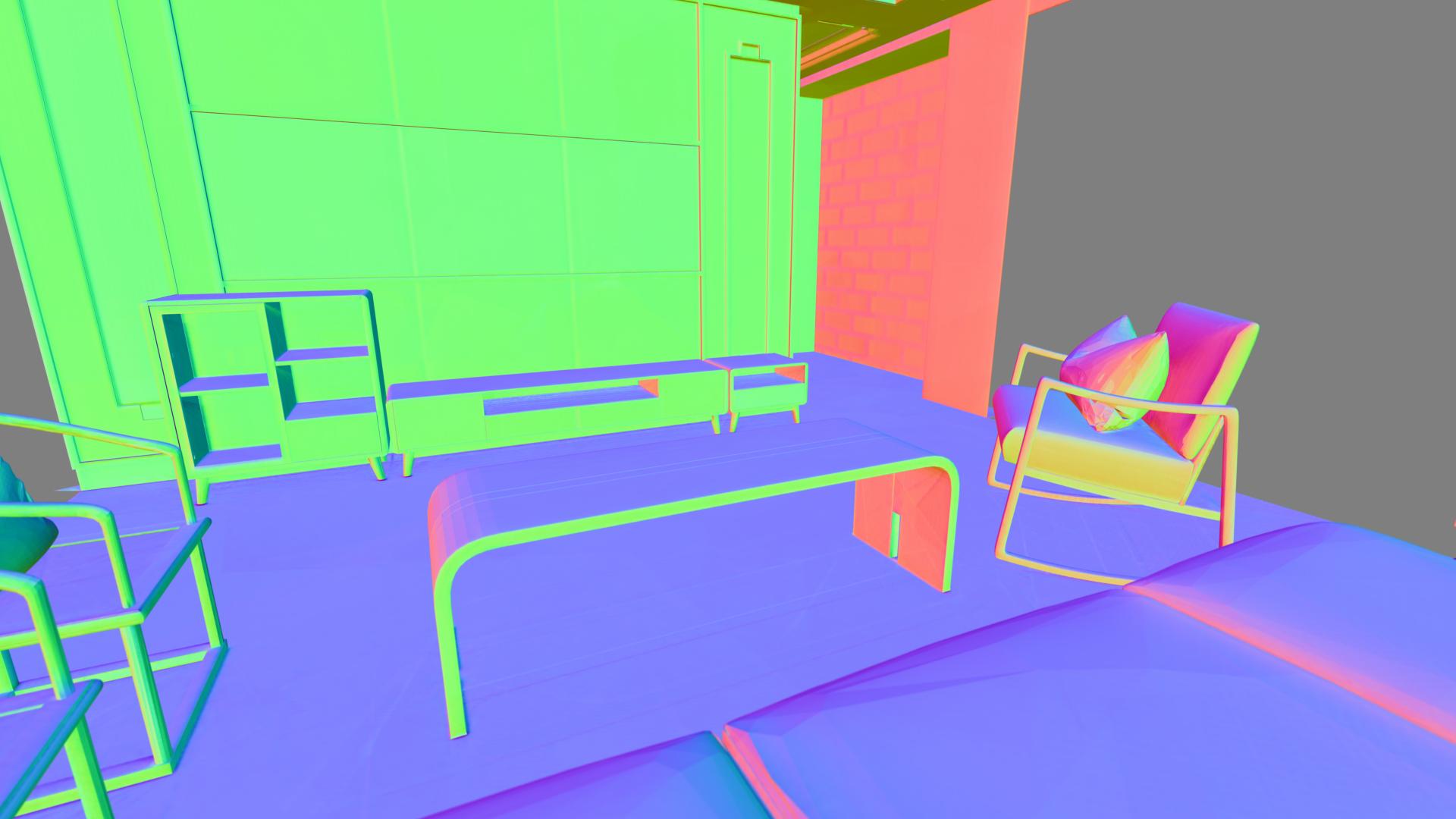}} 
        &
        \fbox{\includegraphics[width=0.15\textwidth,trim={10cm 0 10cm 0},clip]{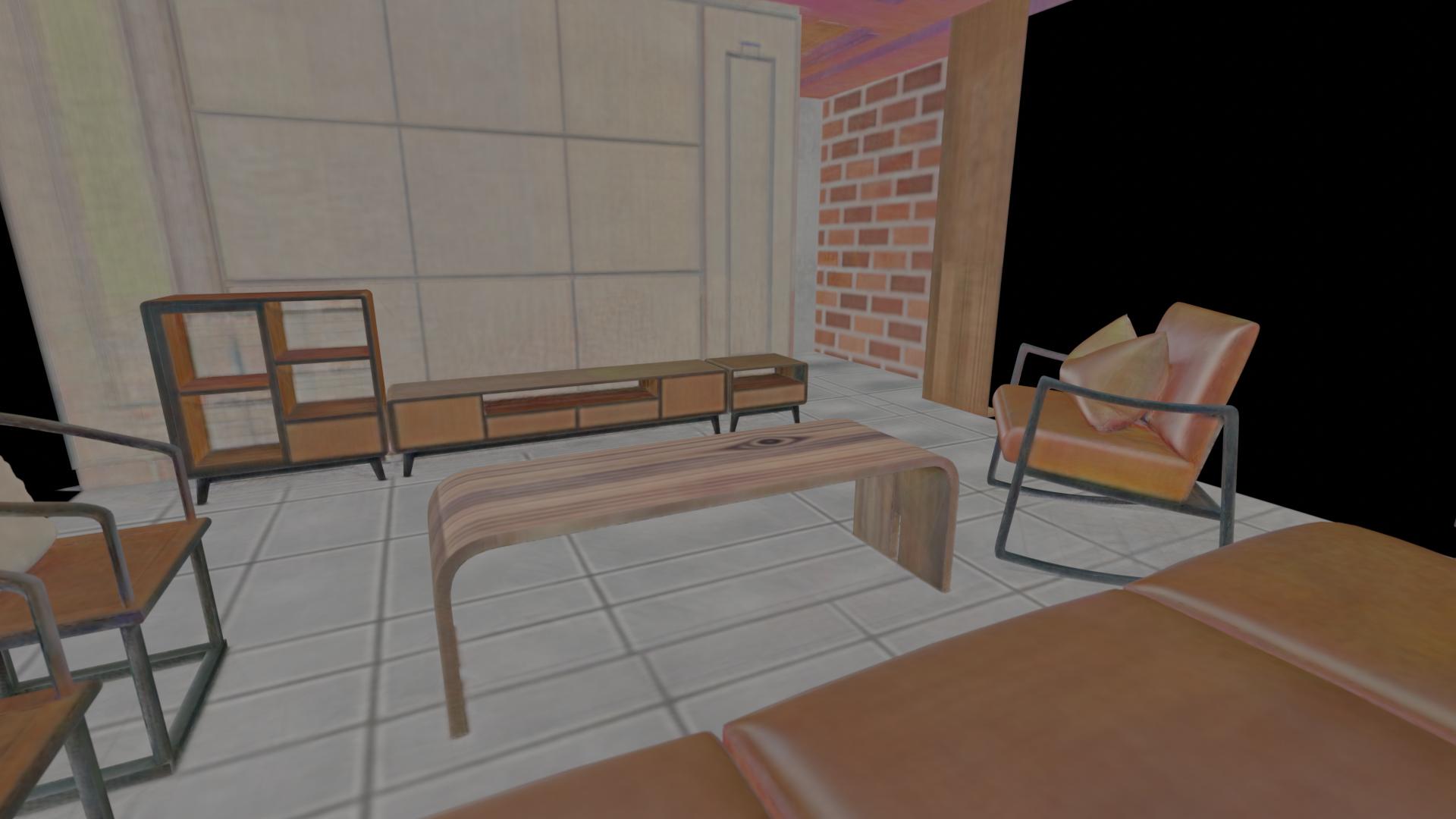}} 
        &
        \fbox{\includegraphics[width=0.15\textwidth,trim={10cm 0 10cm 0},clip]{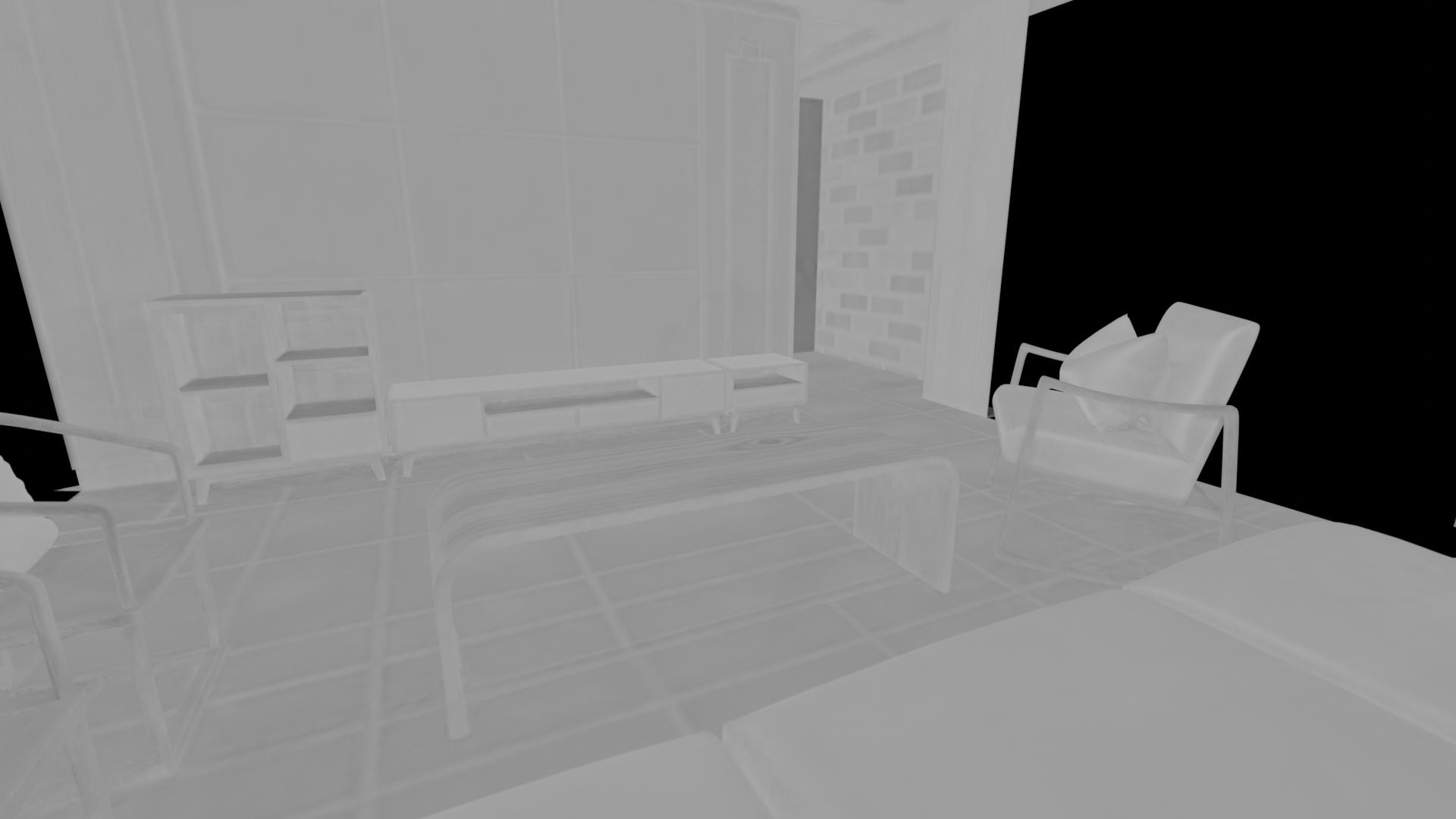}} 
        &
        \fbox{\includegraphics[width=0.15\textwidth,trim={10cm 0 10cm 0},clip]{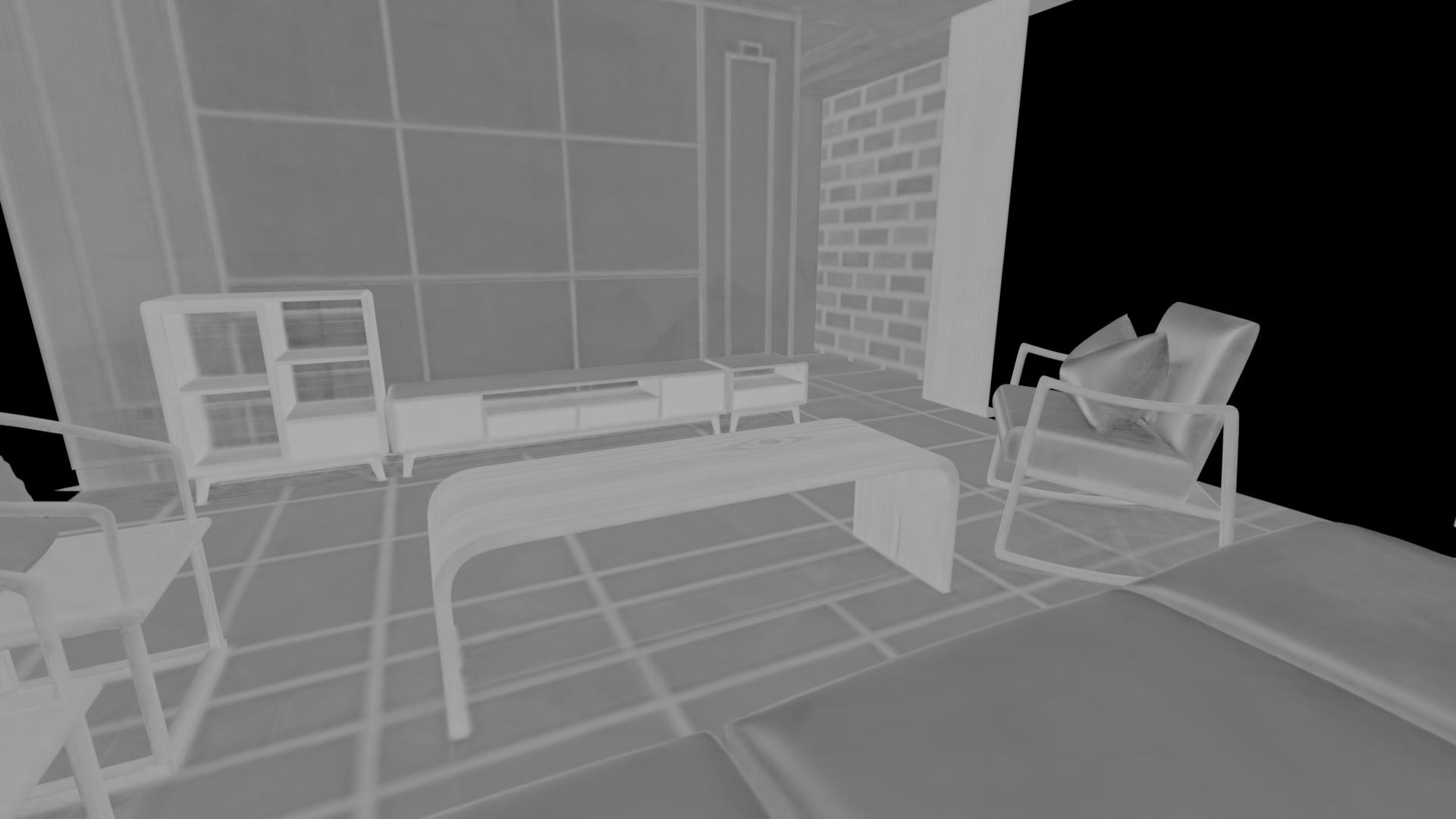}} 
        &
        \fbox{\includegraphics[width=0.15\textwidth,trim={10cm 0 10cm 0},clip]{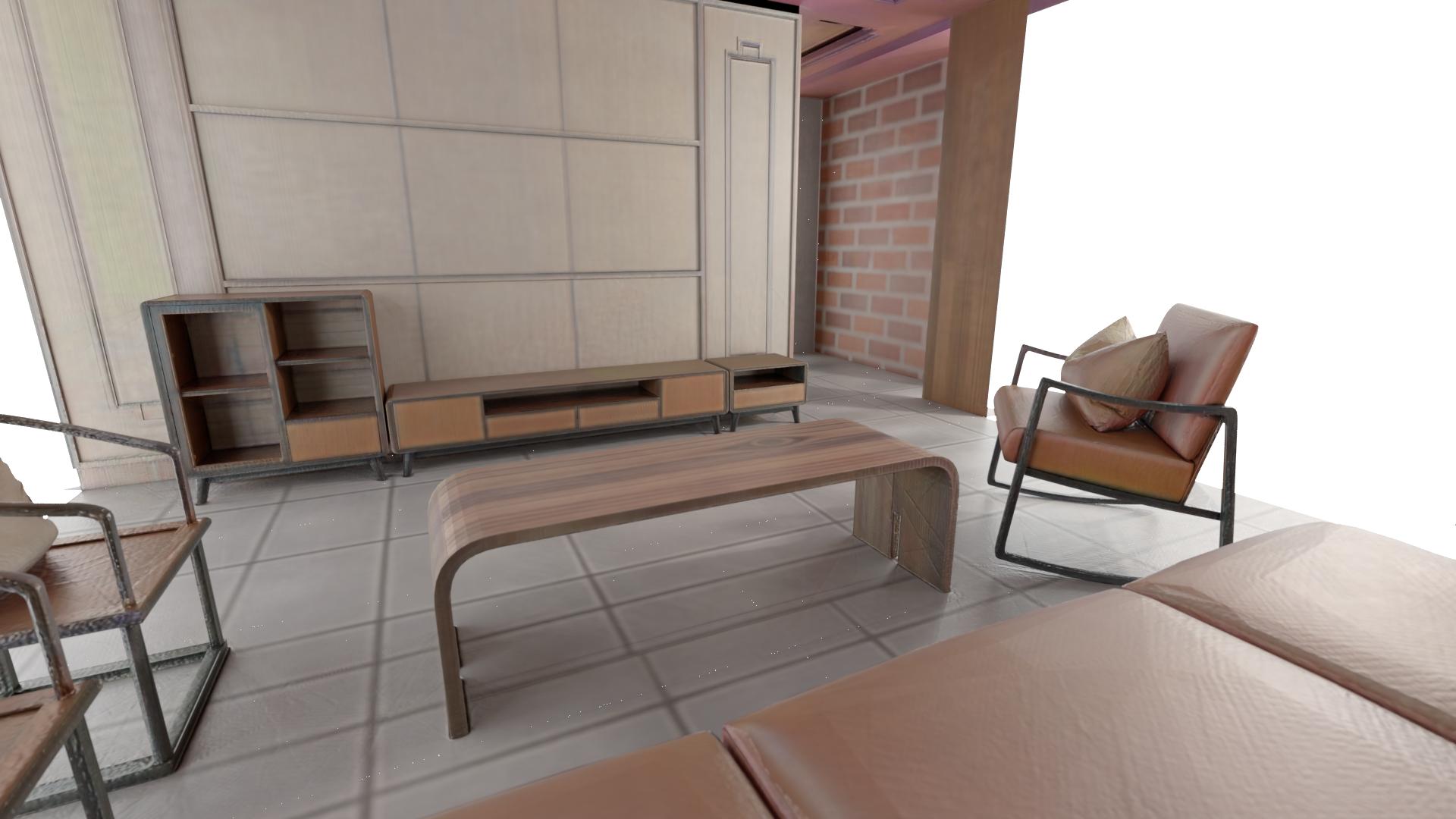}} 
        \\

        &
        {\footnotesize Normal} &
        {\footnotesize Albedo} &
        {\footnotesize Roughness} &
        {\footnotesize Metallic} &
        {\footnotesize Rendering} \\

        \midrule
        
        \rotatebox{90}{{\footnotesize View 1}}
        &
        \fbox{\includegraphics[width=0.15\textwidth,trim={10cm 0 10cm 0},clip]{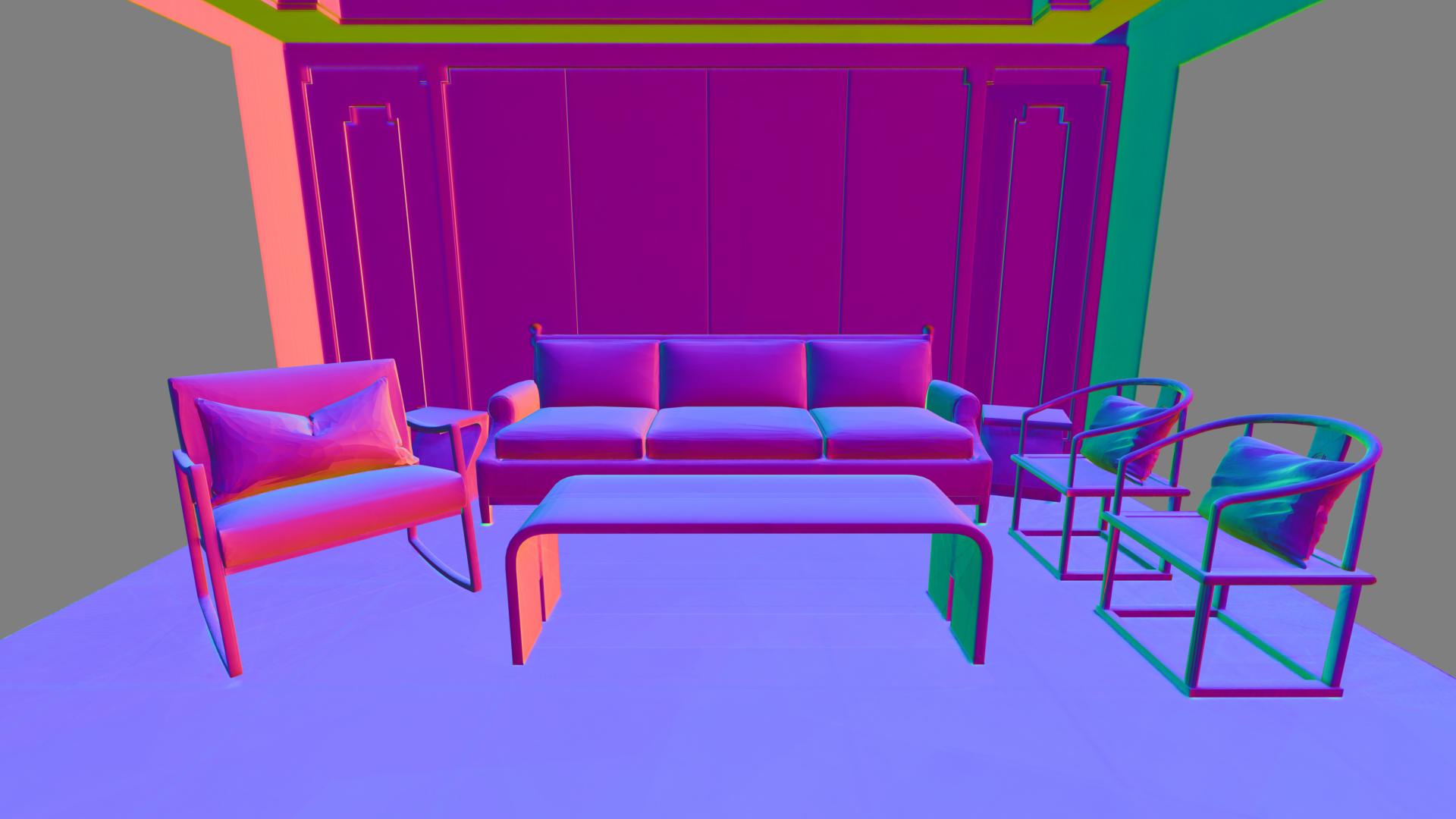}}
        &
        \fbox{\includegraphics[width=0.15\textwidth,trim={10cm 0 10cm 0},clip]{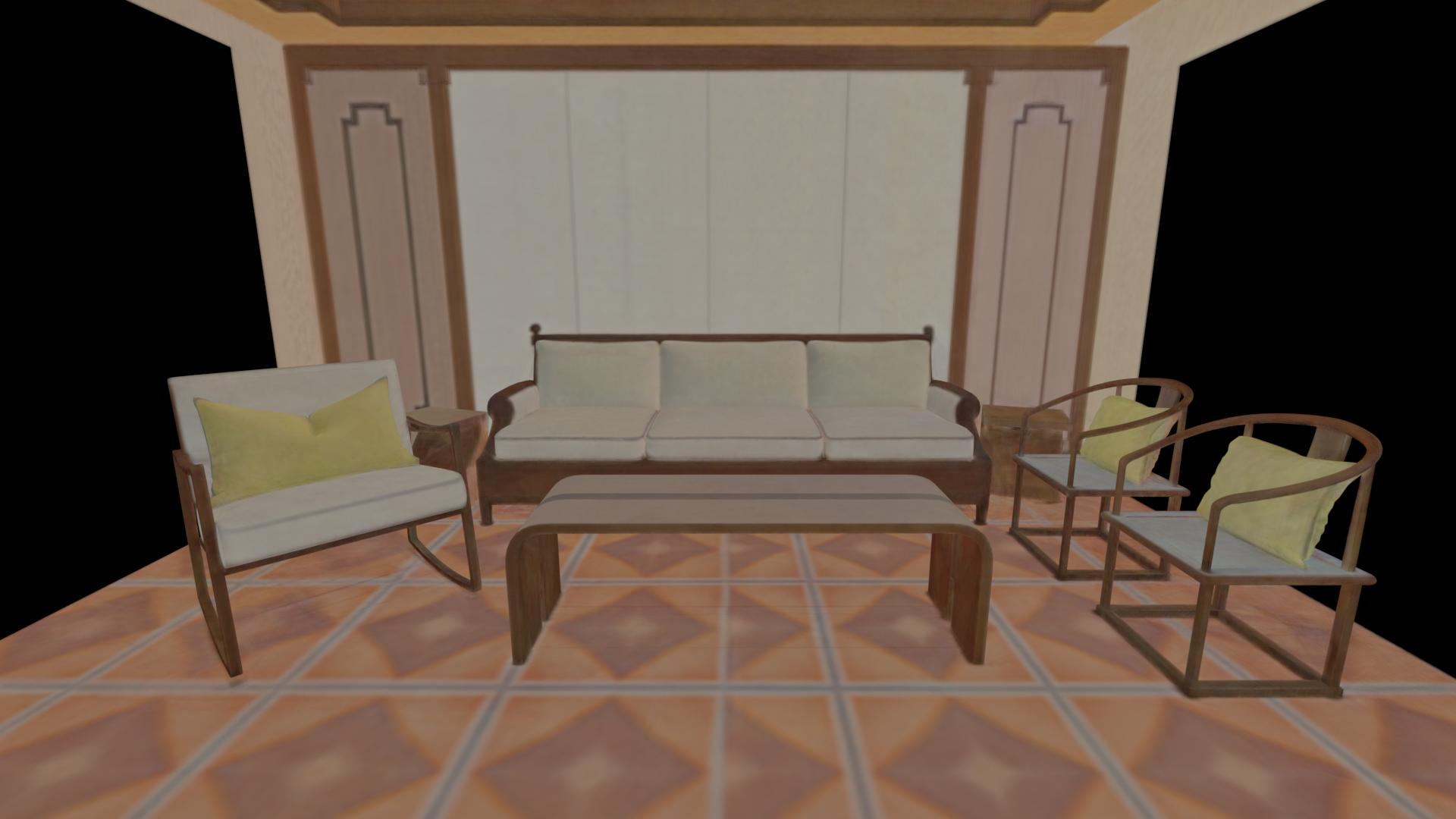}} 
        &
        \fbox{\includegraphics[width=0.15\textwidth,trim={10cm 0 10cm 0},clip]{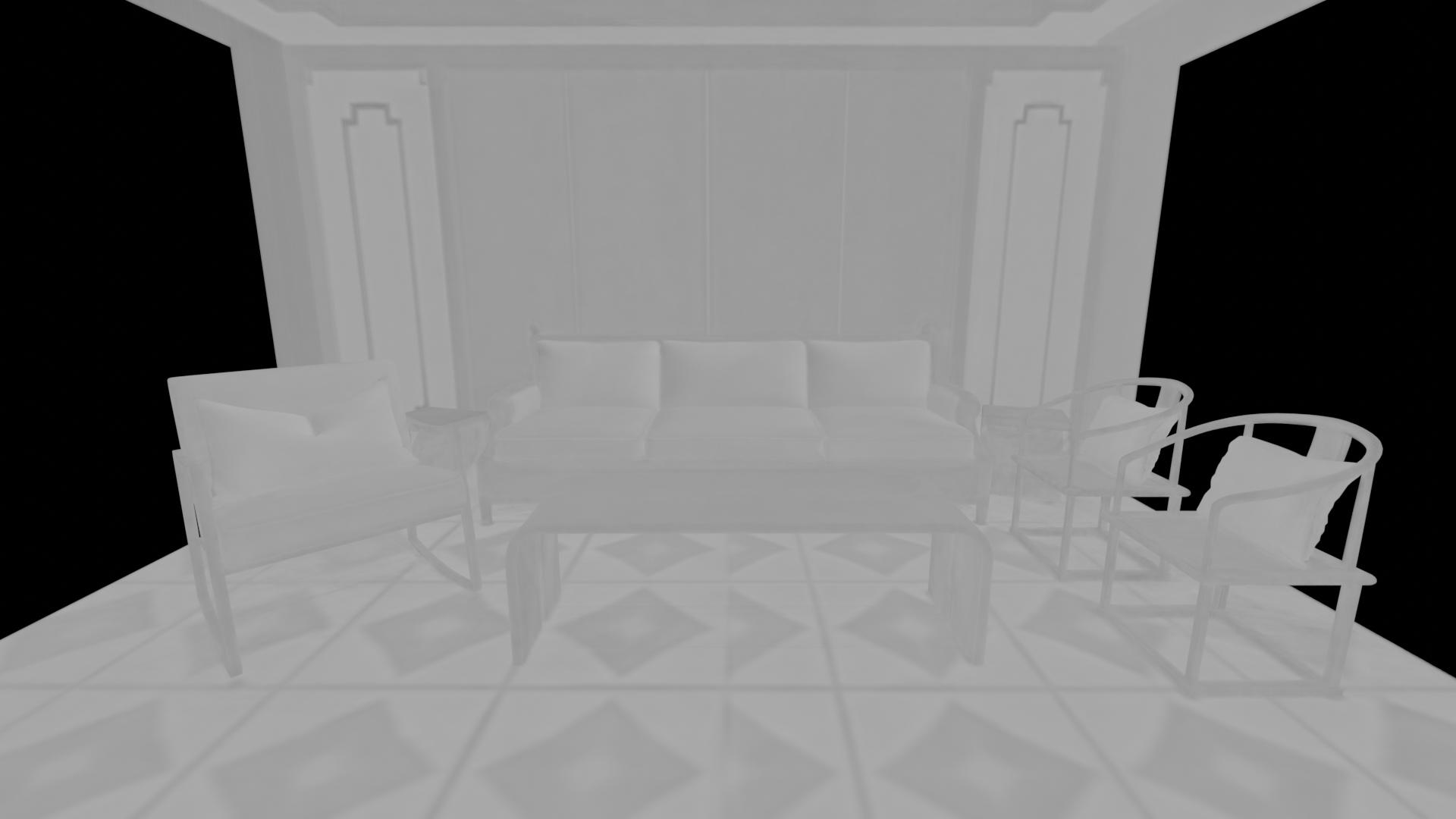}} 
        &
        \fbox{\includegraphics[width=0.15\textwidth,trim={10cm 0 10cm 0},clip]{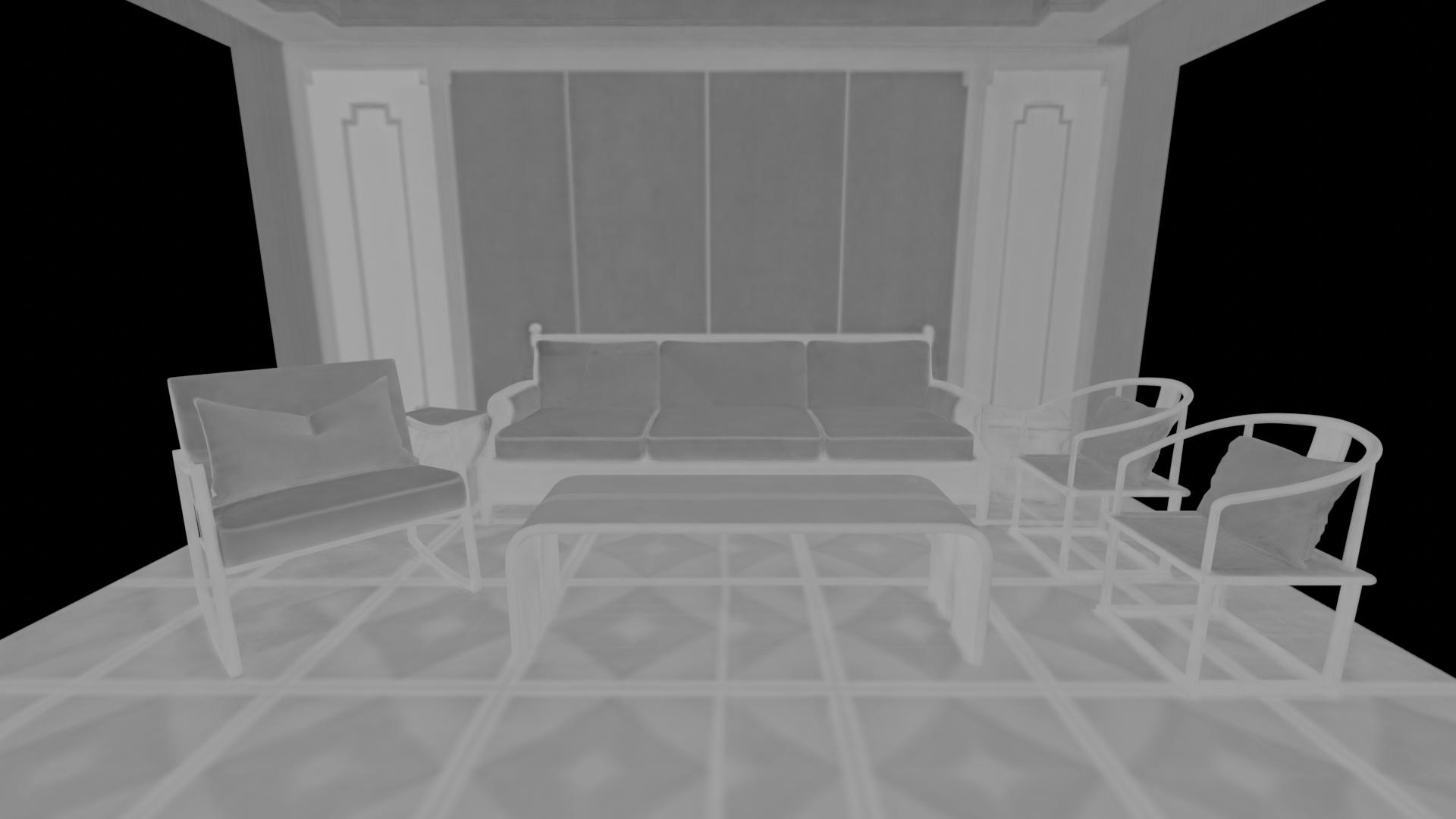}} 
        &
        \fbox{\includegraphics[width=0.15\textwidth,trim={10cm 0 10cm 0},clip]{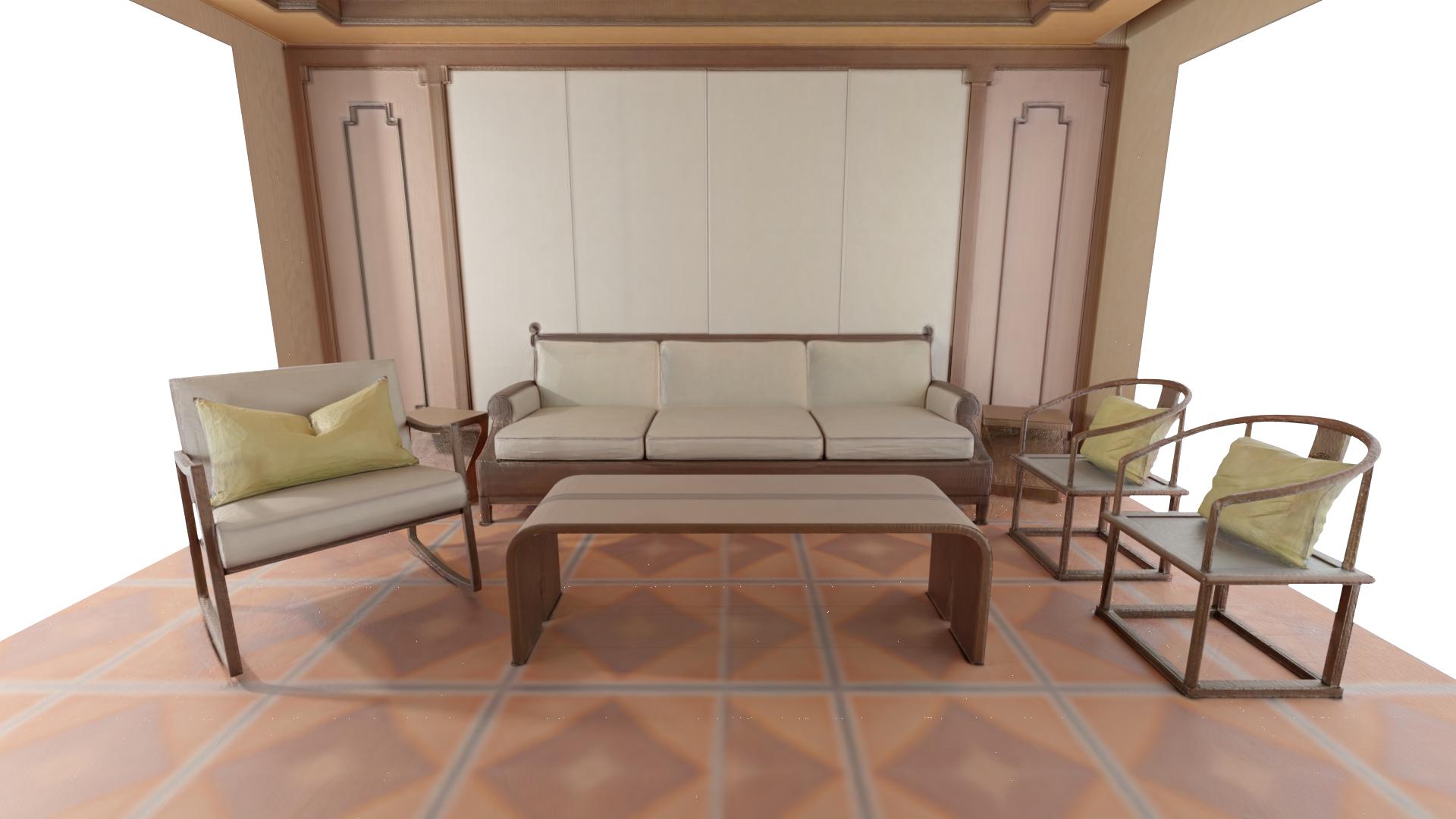}} 
        \\
        
        
        \rotatebox{90}{{\footnotesize View 3}}
        &
        \fbox{\includegraphics[width=0.15\textwidth,trim={10cm 0 10cm 0},clip]{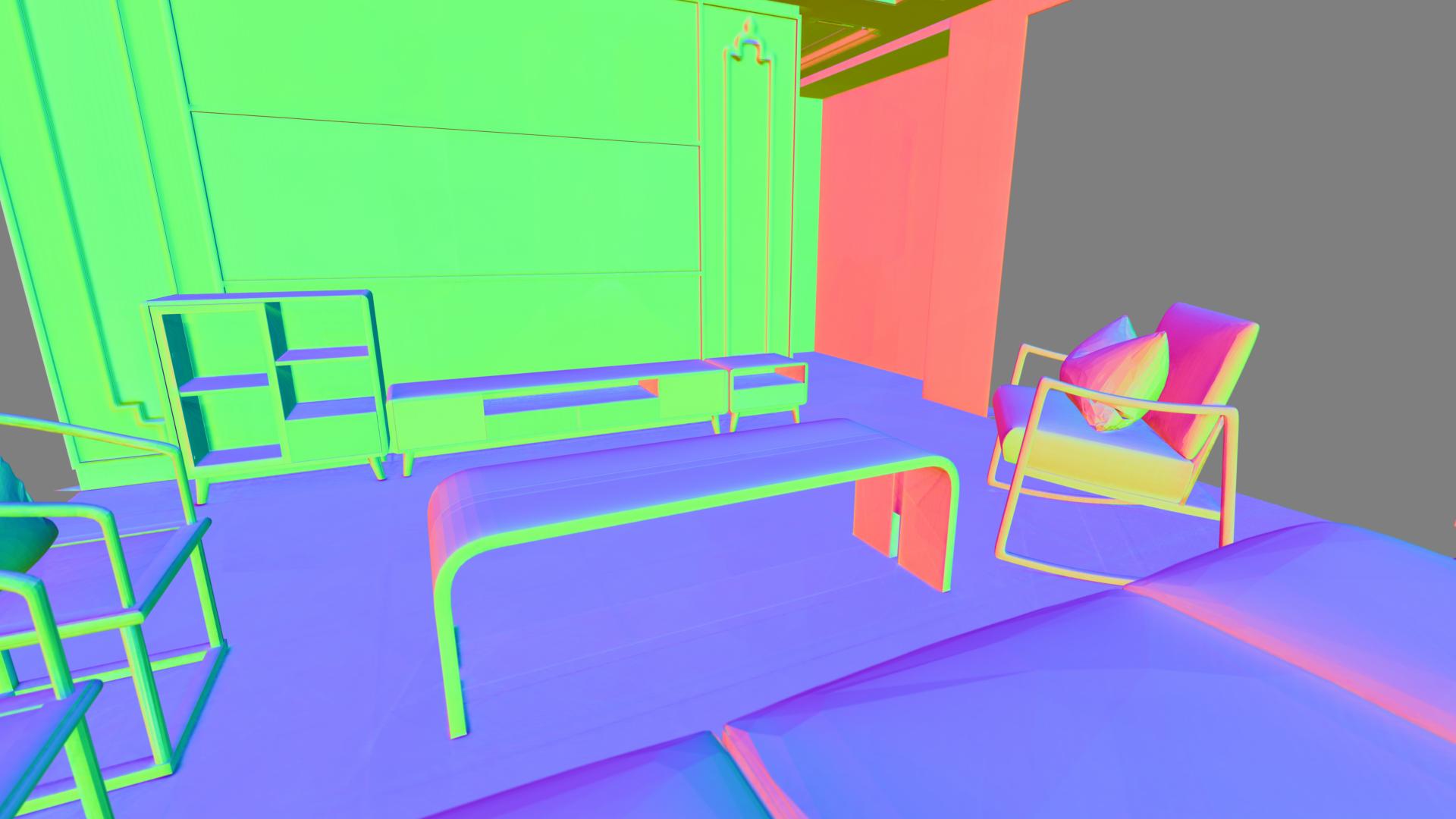}} 
        &
        \fbox{\includegraphics[width=0.15\textwidth,trim={10cm 0 10cm 0},clip]{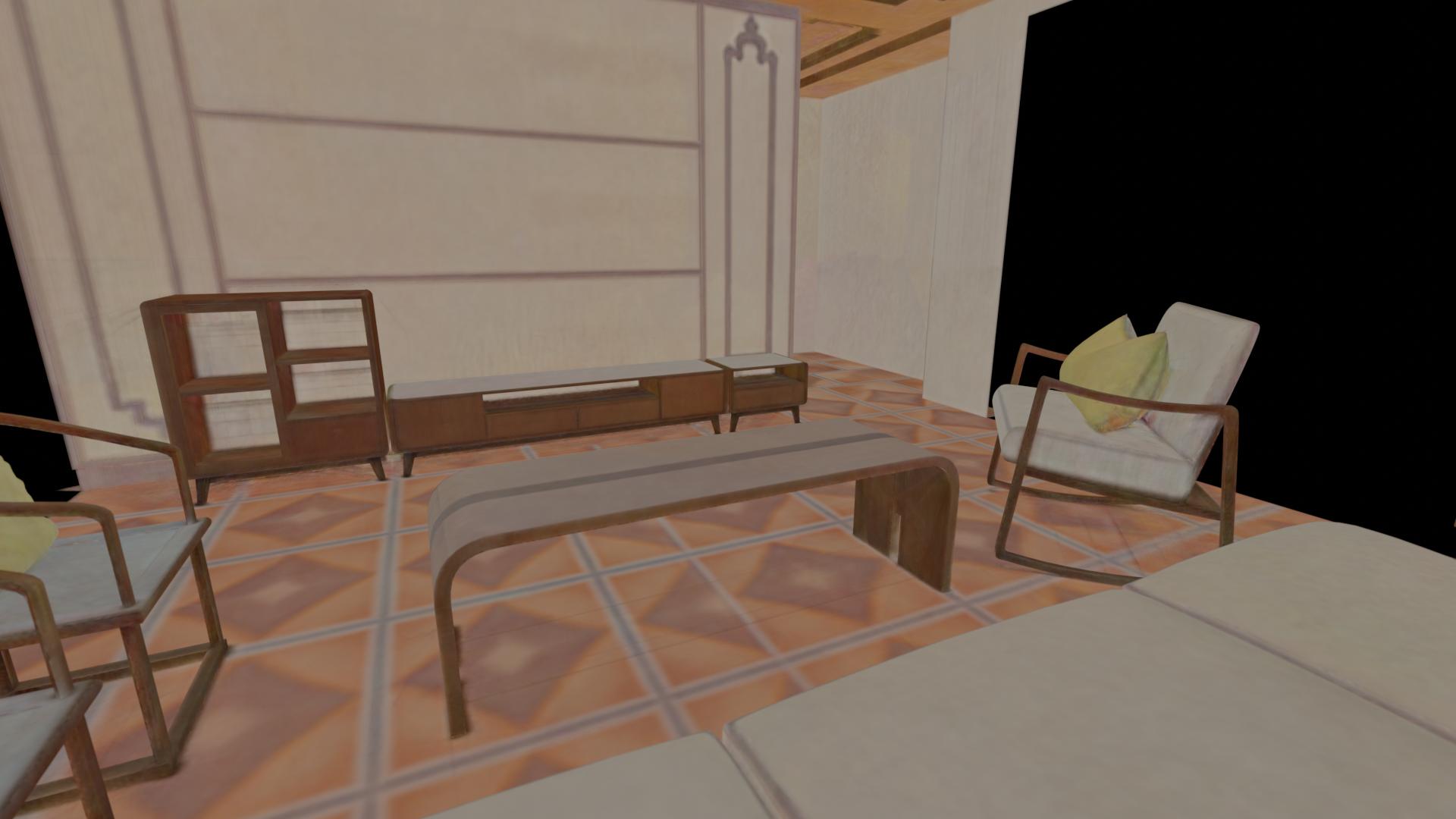}} 
        &
        \fbox{\includegraphics[width=0.15\textwidth,trim={10cm 0 10cm 0},clip]{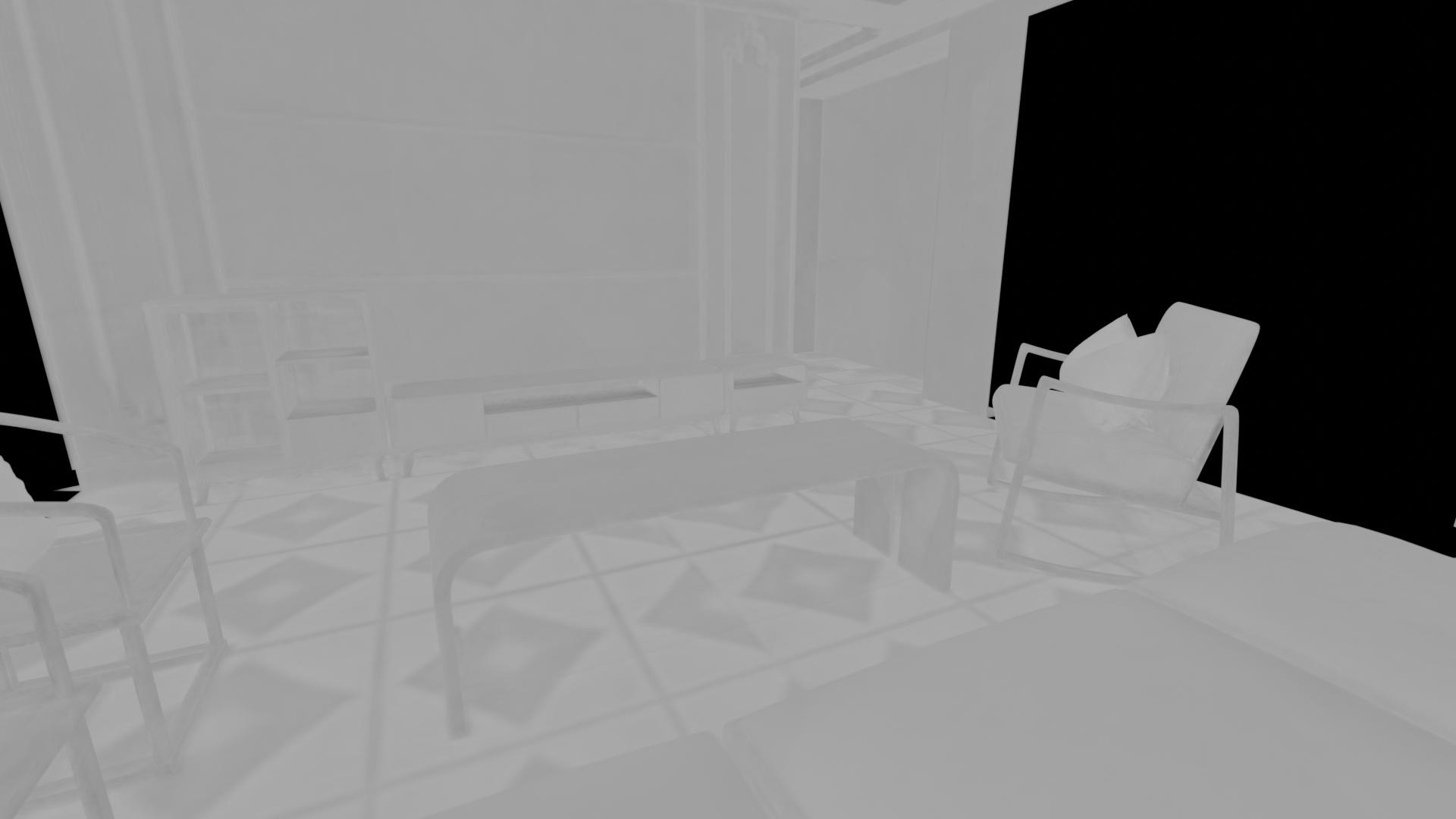}} 
        &
        \fbox{\includegraphics[width=0.15\textwidth,trim={10cm 0 10cm 0},clip]{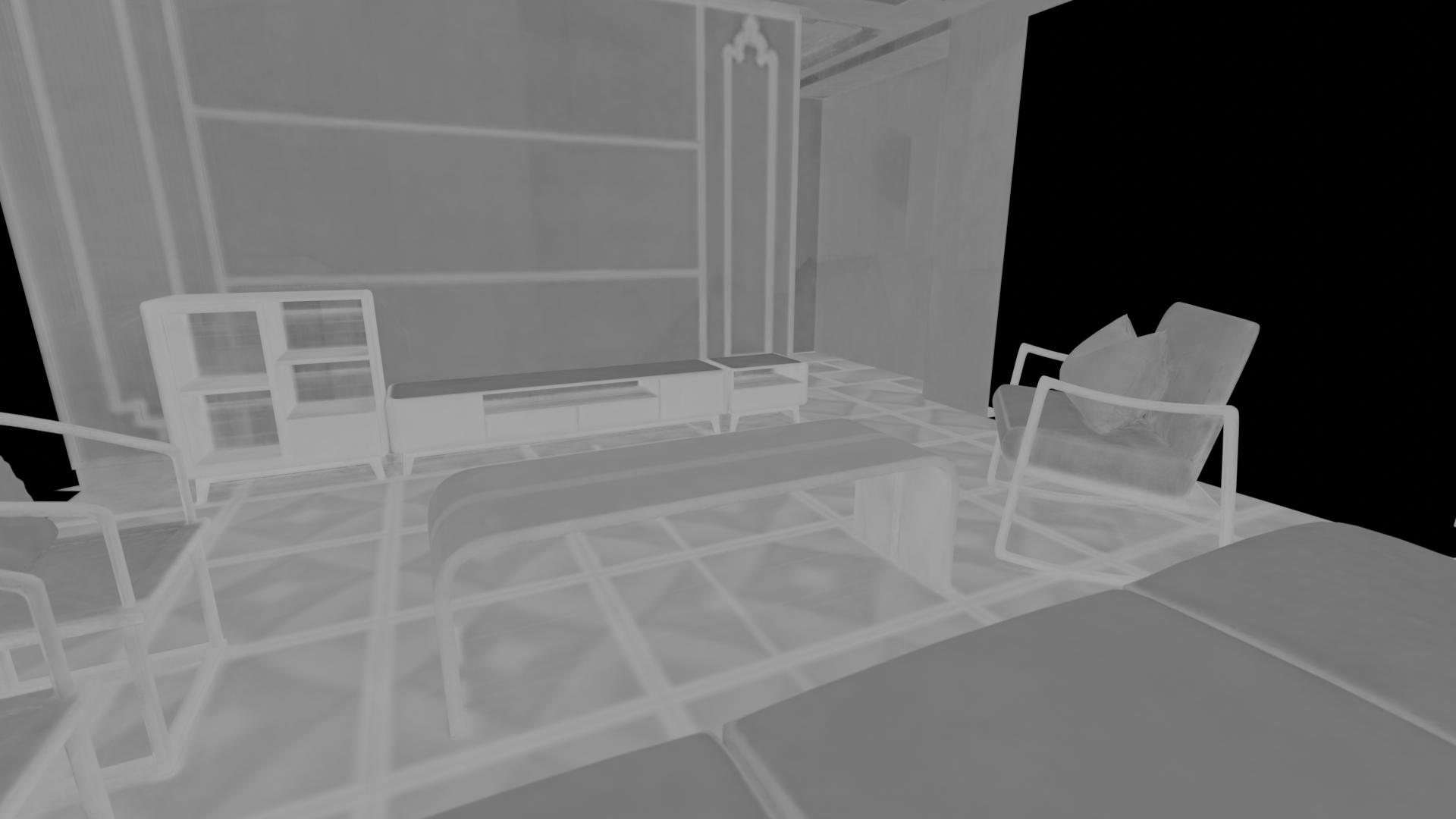}} 
        &
        \fbox{\includegraphics[width=0.15\textwidth,trim={10cm 0 10cm 0},clip]{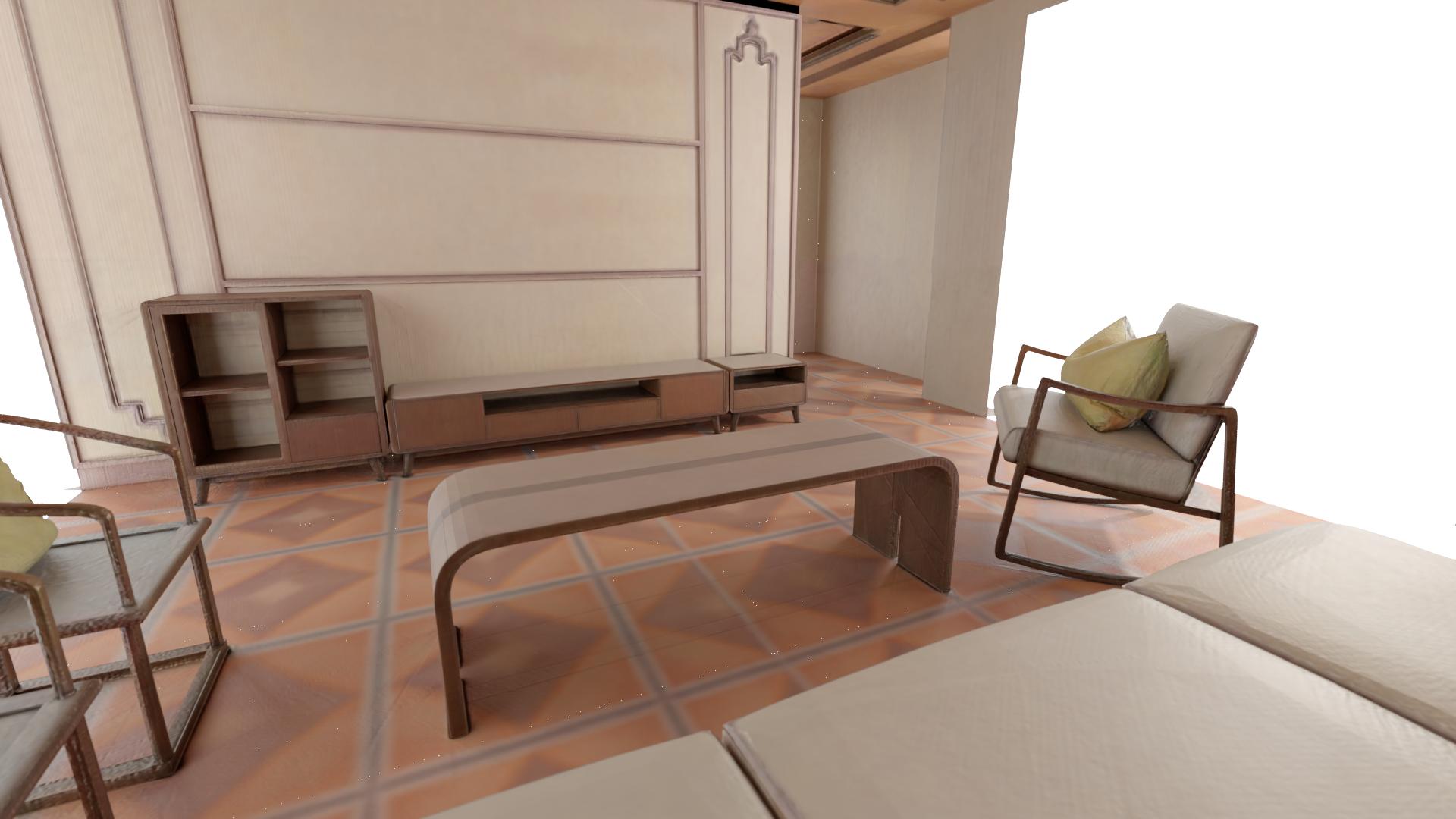}} 
        \\

        &
        {\footnotesize Normal} &
        {\footnotesize Albedo} &
        {\footnotesize Roughness} &
        {\footnotesize Metallic} &
        {\footnotesize Rendering} \\

        \midrule
        
        \rotatebox{90}{{\footnotesize View 1}}
        &
        \fbox{\includegraphics[width=0.15\textwidth,trim={10cm 0 10cm 0},clip]{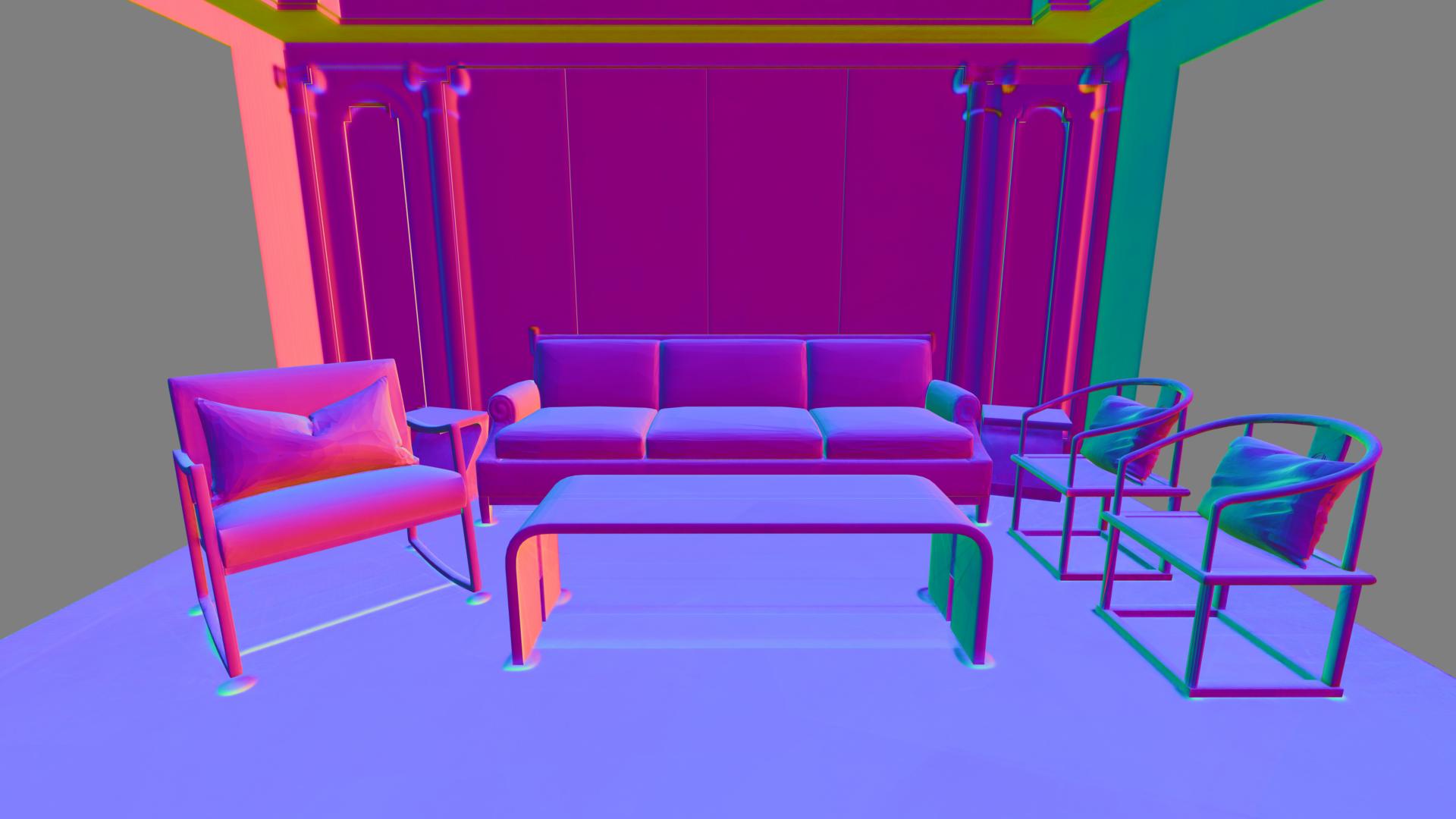}}
        &
        \fbox{\includegraphics[width=0.15\textwidth,trim={10cm 0 10cm 0},clip]{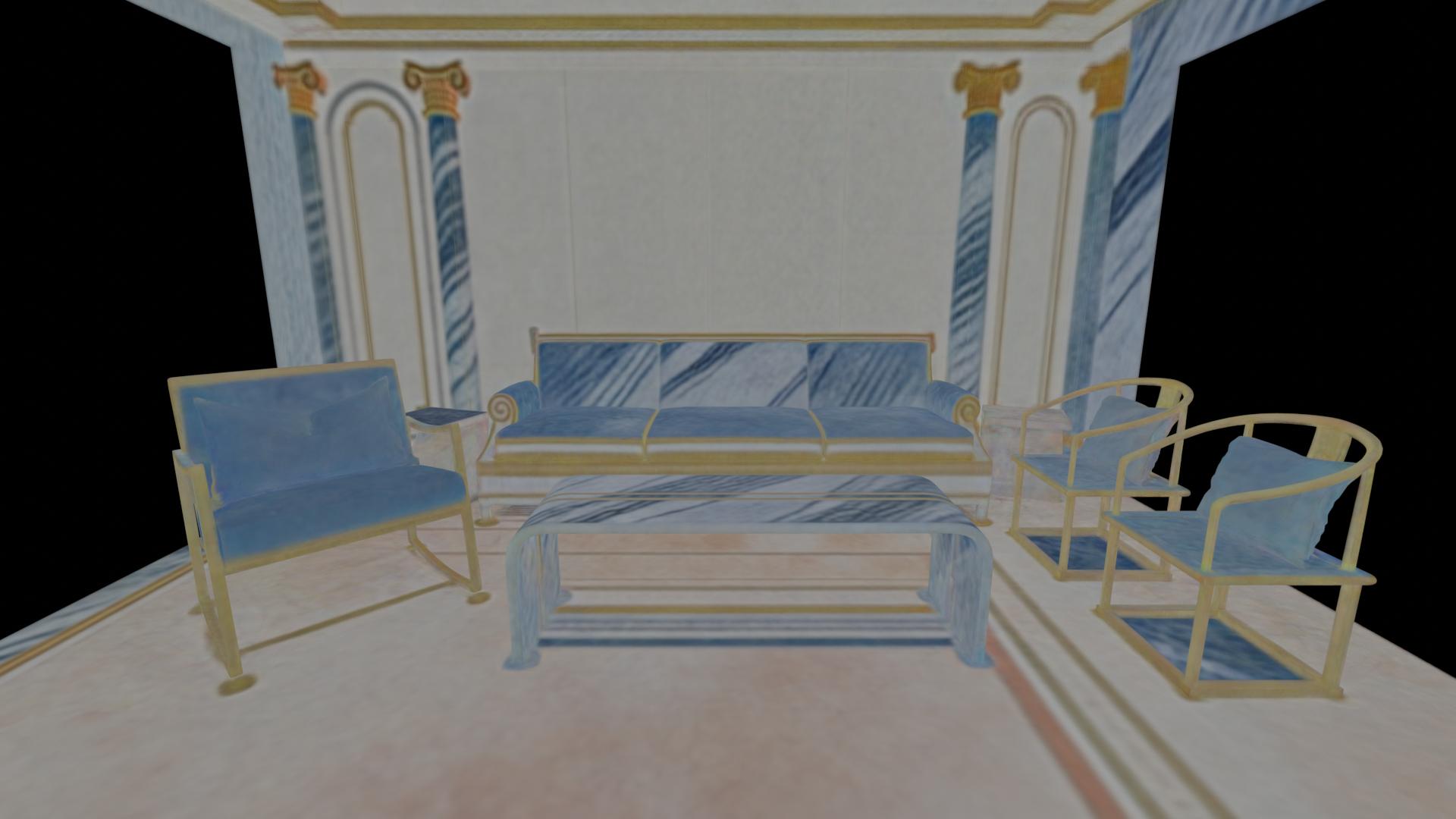}} 
        &
        \fbox{\includegraphics[width=0.15\textwidth,trim={10cm 0 10cm 0},clip]{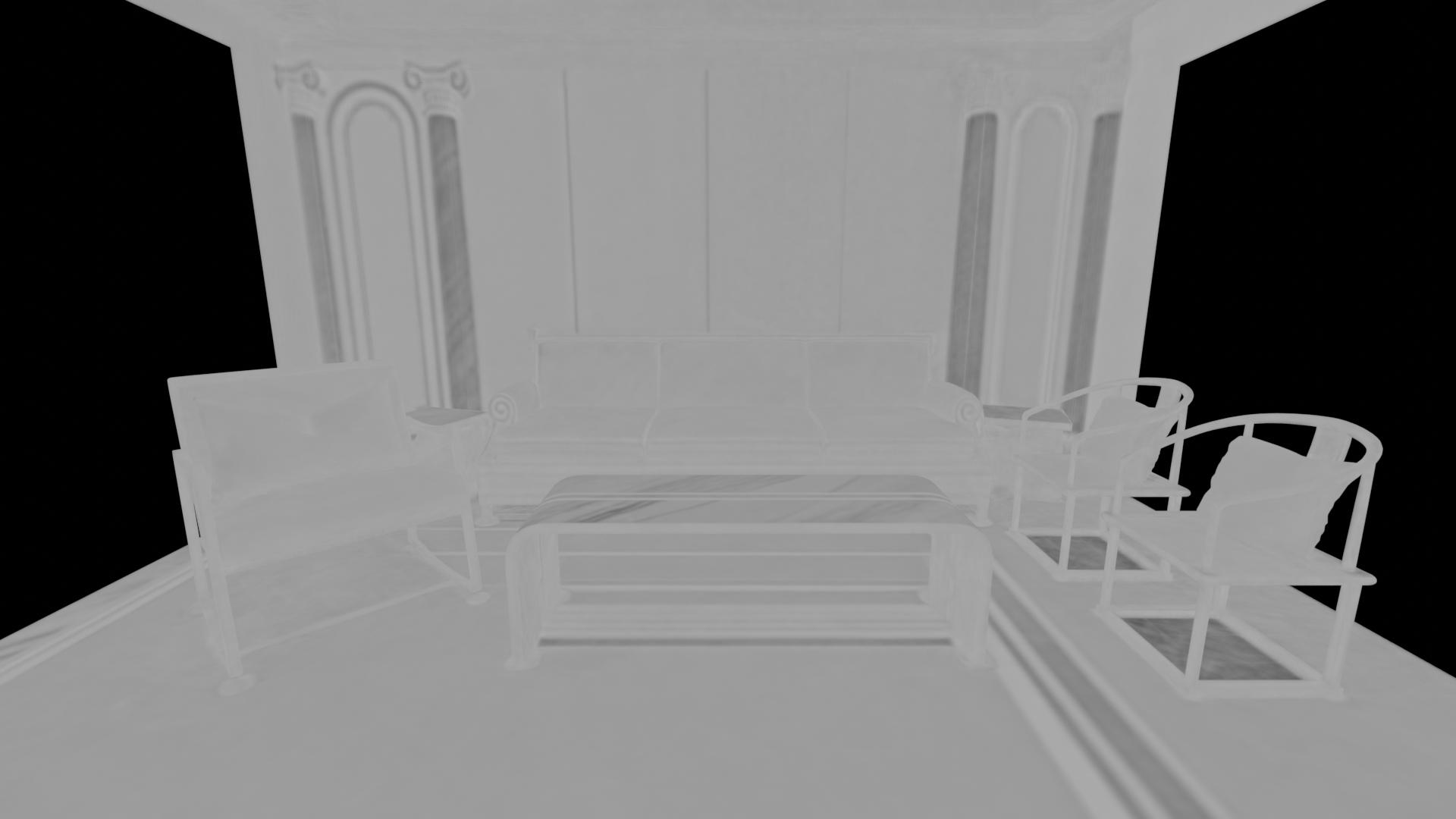}} 
        &
        \fbox{\includegraphics[width=0.15\textwidth,trim={10cm 0 10cm 0},clip]{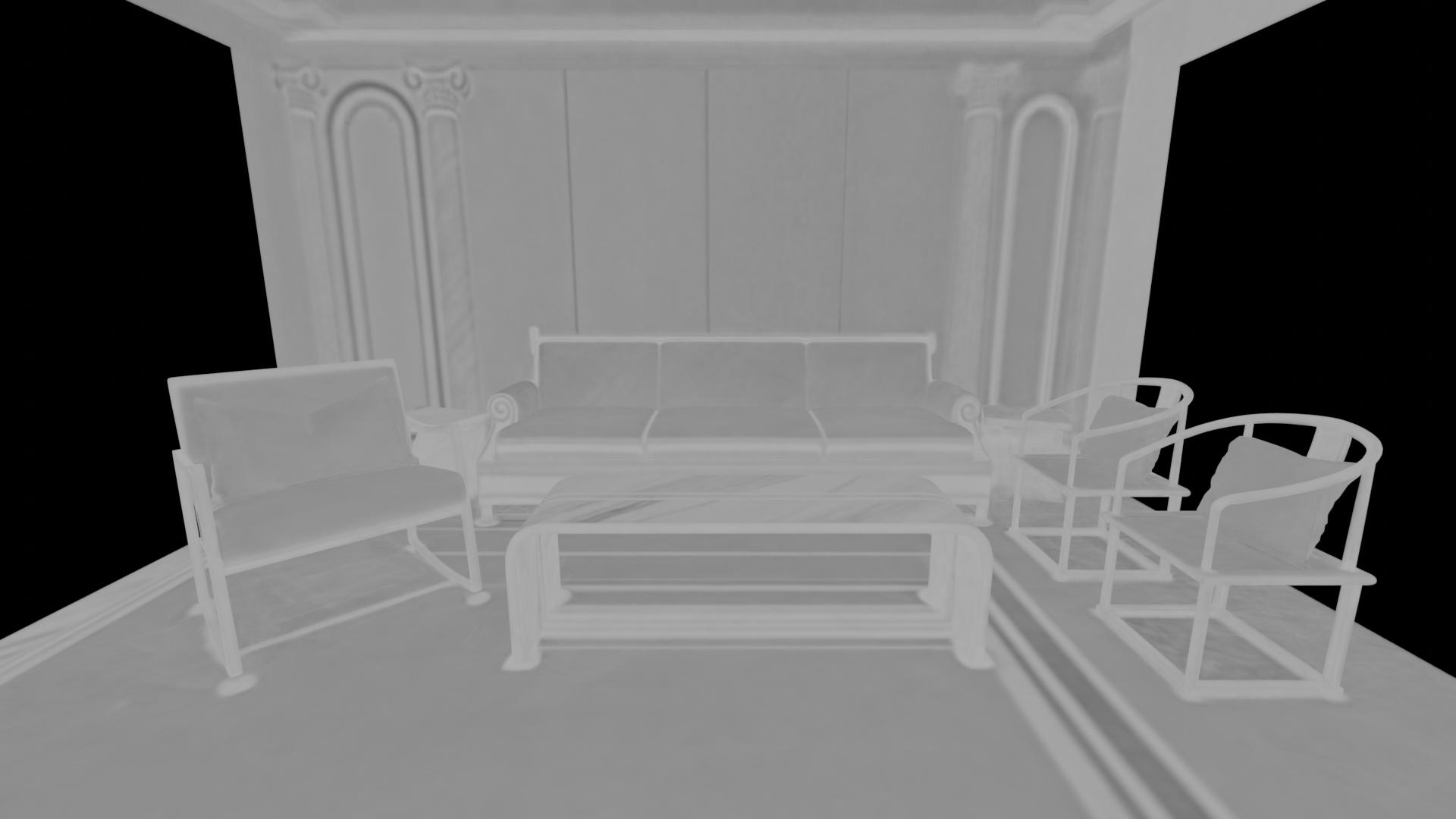}} 
        &
        \fbox{\includegraphics[width=0.15\textwidth,trim={10cm 0 10cm 0},clip]{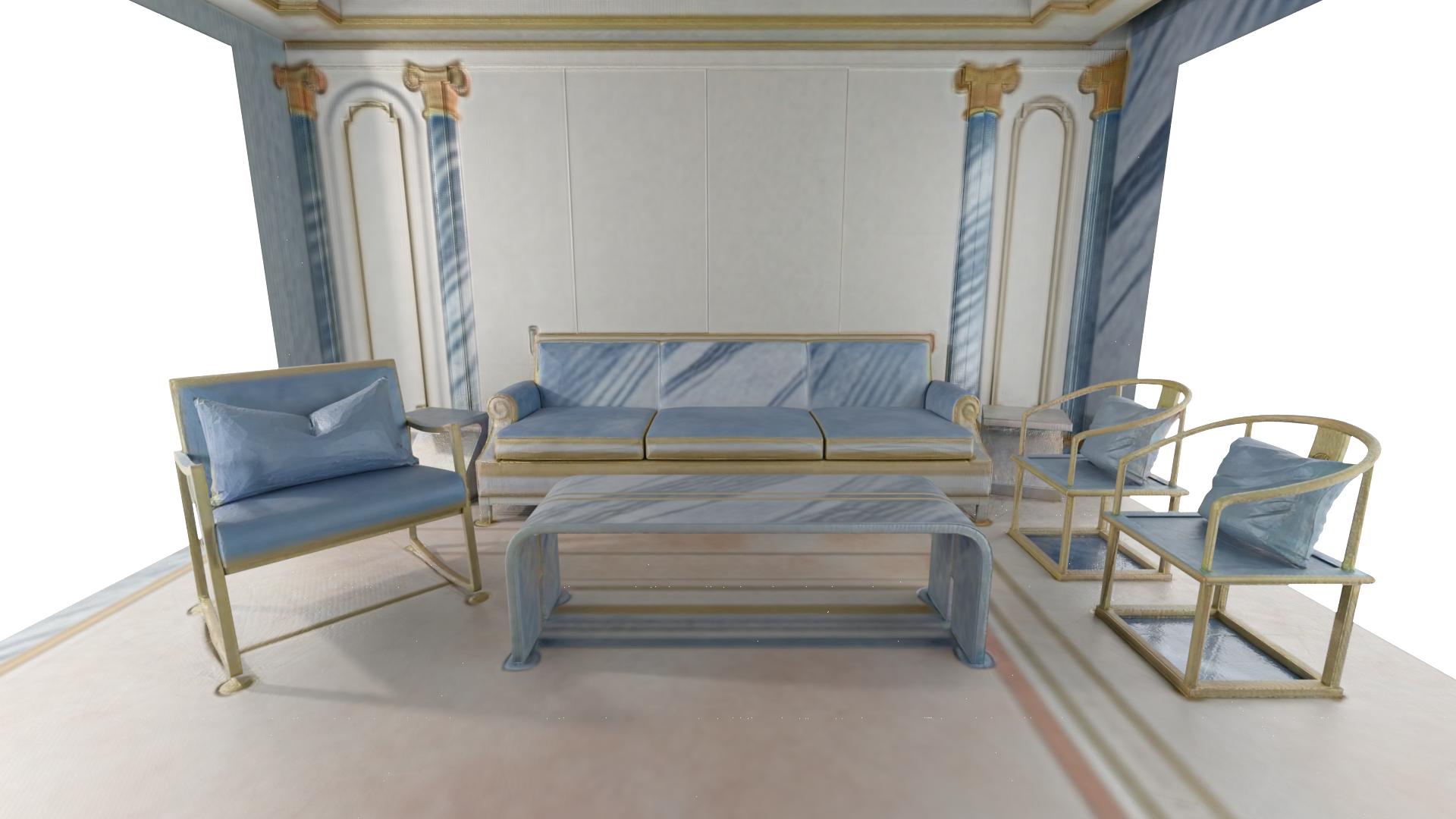}} 
        \\
        
        
        \rotatebox{90}{{\footnotesize View 3}}
        &
        \fbox{\includegraphics[width=0.15\textwidth,trim={10cm 0 10cm 0},clip]{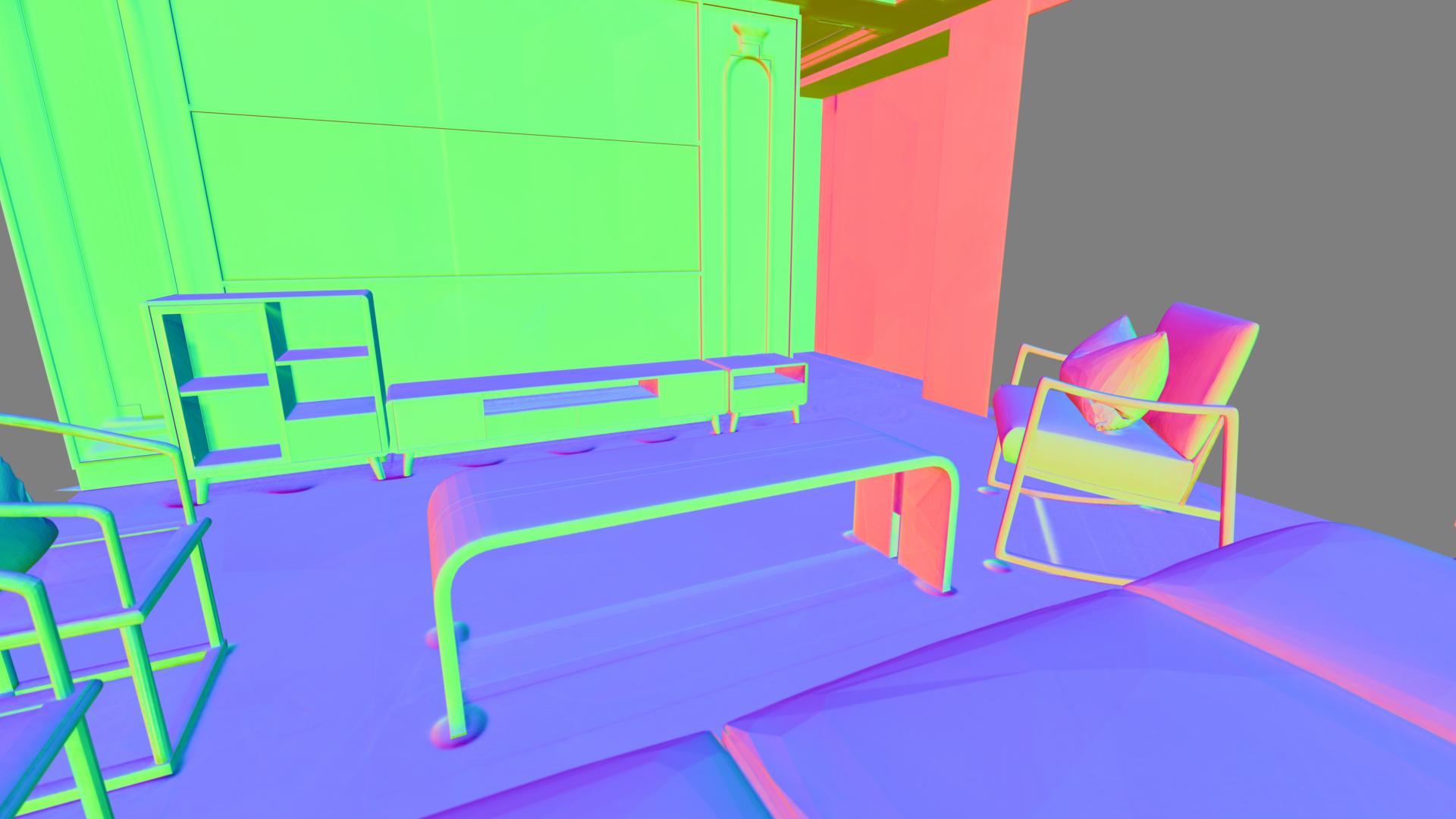}} 
        &
        \fbox{\includegraphics[width=0.15\textwidth,trim={10cm 0 10cm 0},clip]{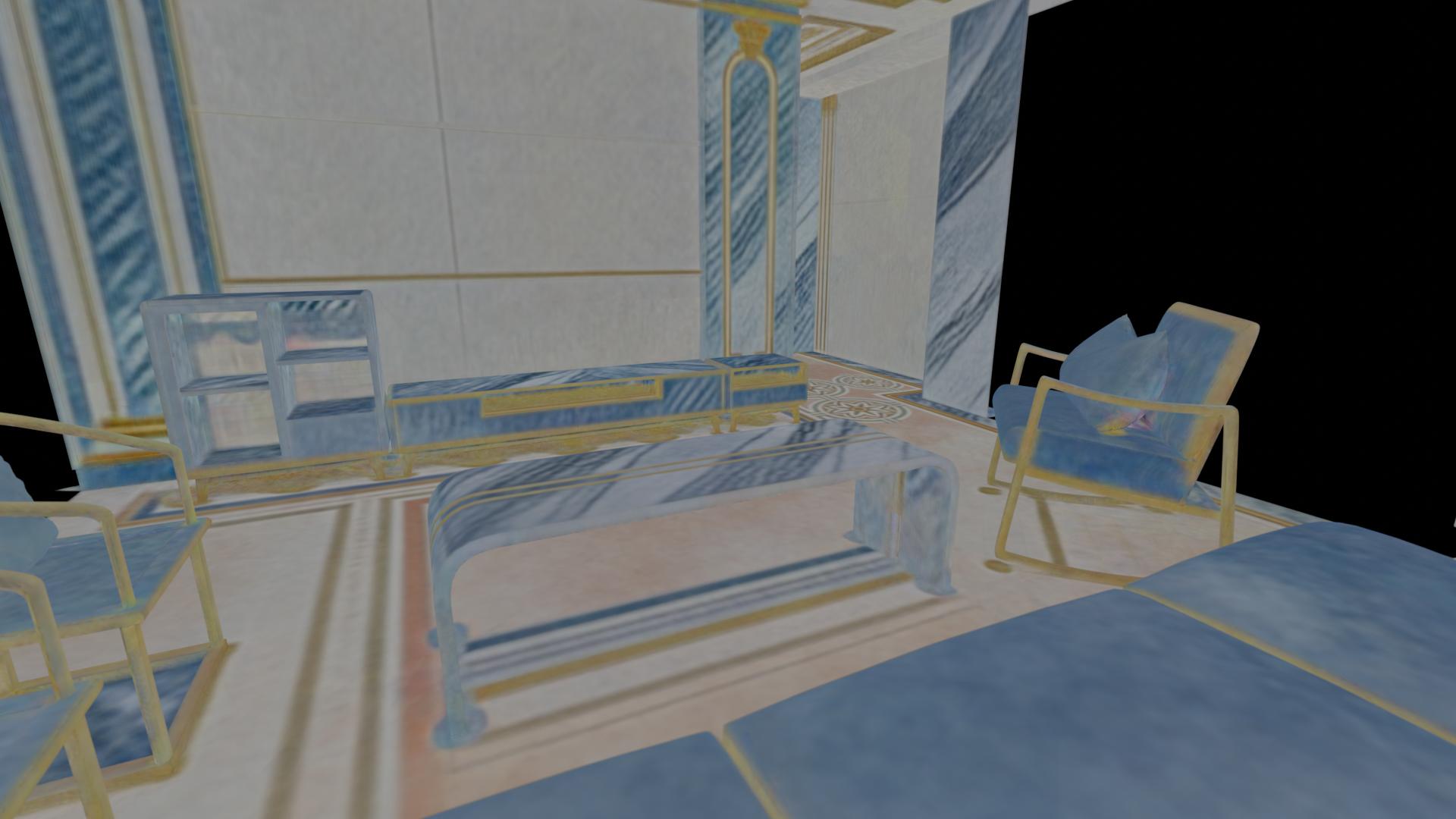}} 
        &
        \fbox{\includegraphics[width=0.15\textwidth,trim={10cm 0 10cm 0},clip]{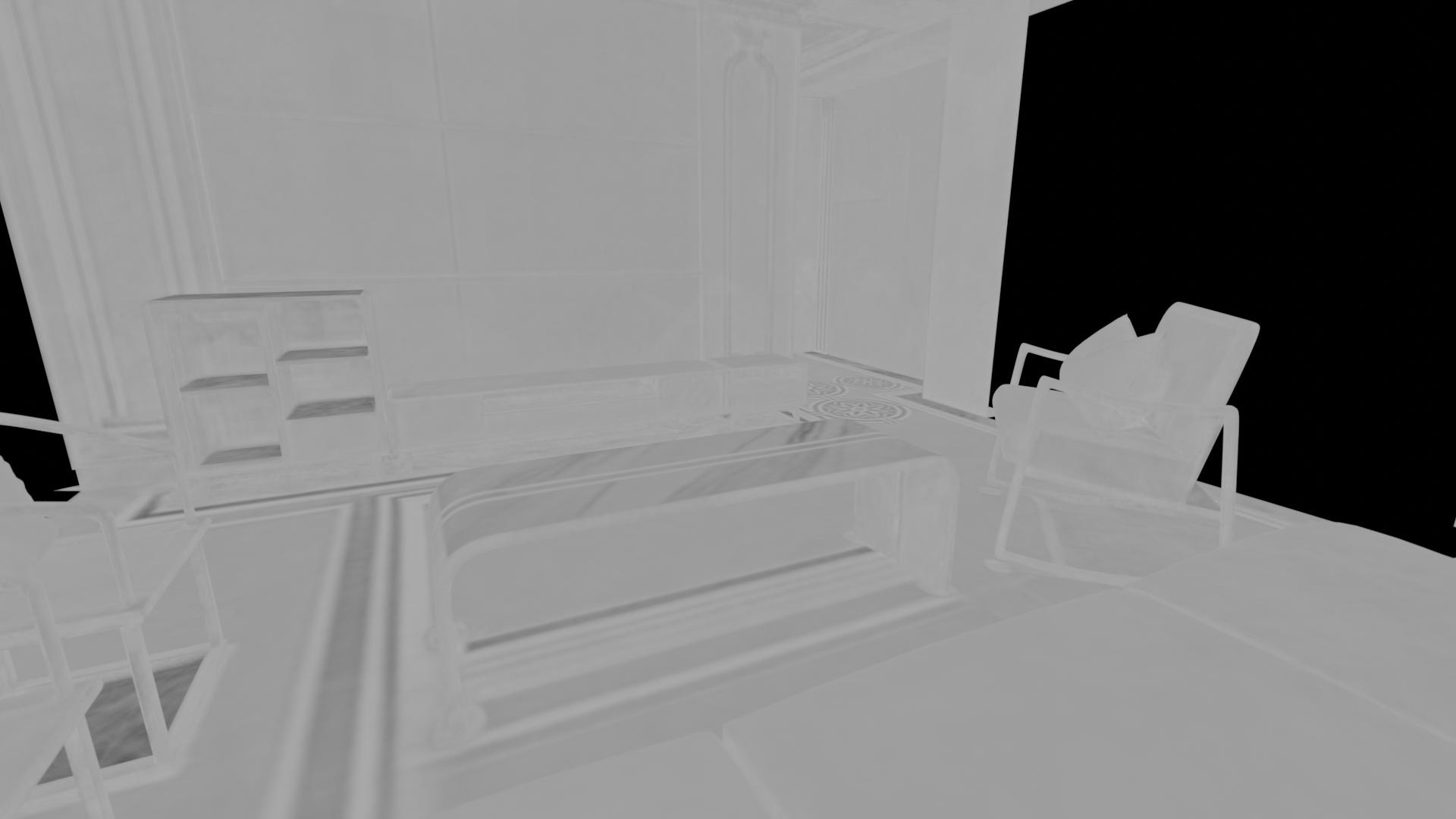}} 
        &
        \fbox{\includegraphics[width=0.15\textwidth,trim={10cm 0 10cm 0},clip]{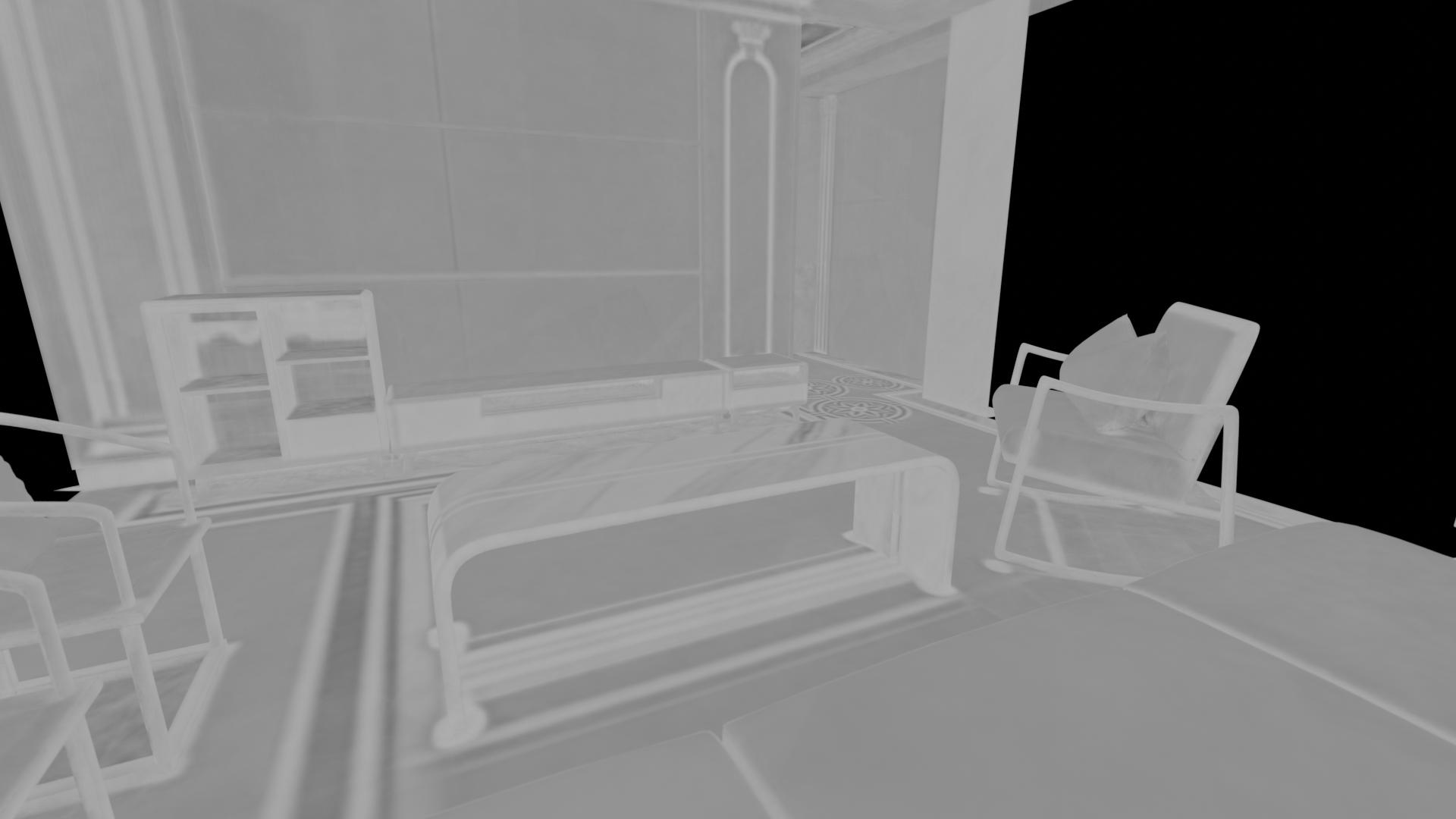}} 
        &
        \fbox{\includegraphics[width=0.15\textwidth,trim={10cm 0 10cm 0},clip]{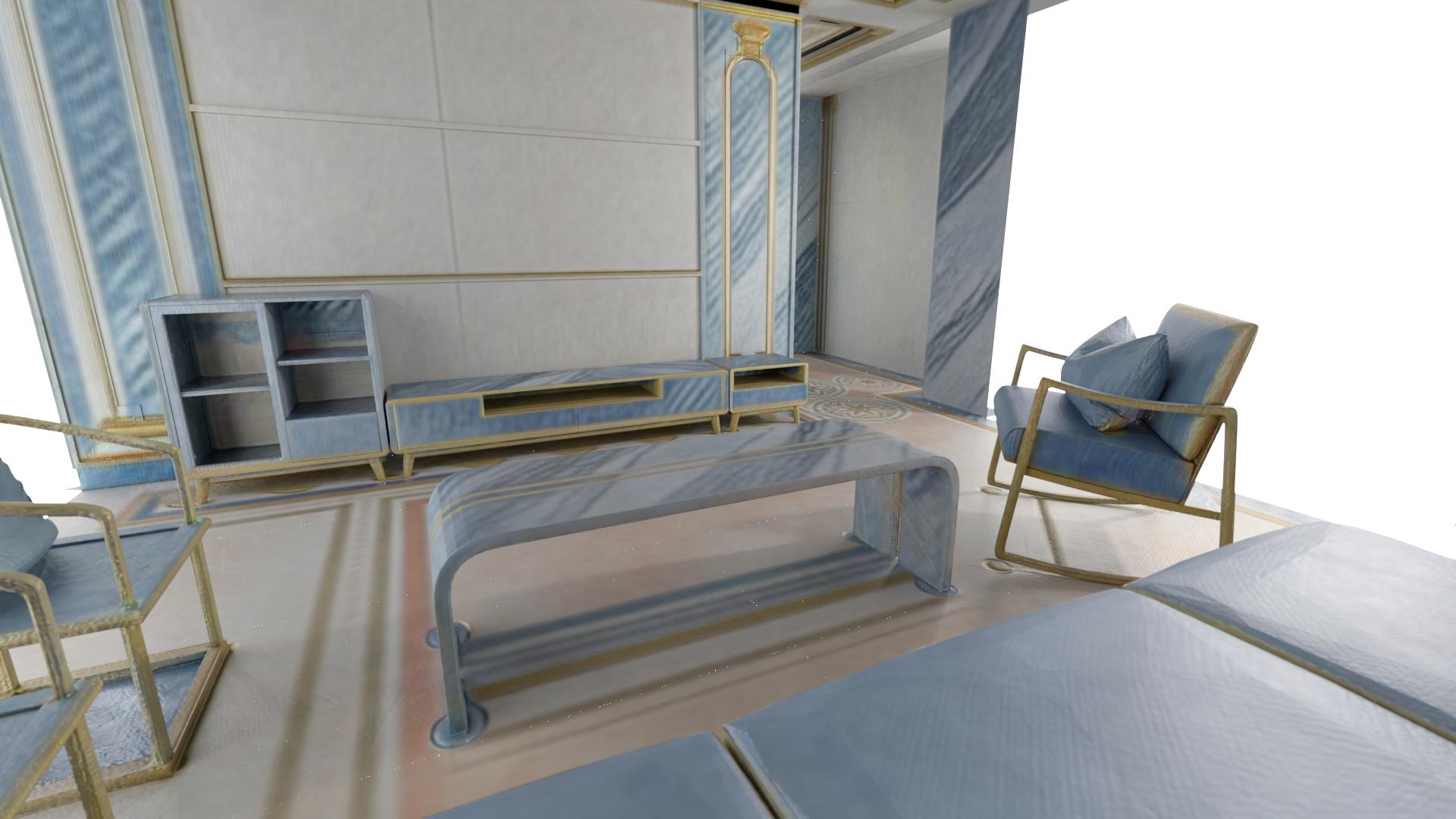}} 
        \\

        &
        {\footnotesize Normal} &
        {\footnotesize Albedo} &
        {\footnotesize Roughness} &
        {\footnotesize Metallic} &
        {\footnotesize Rendering} \\
    \end{tabular}}
    \caption{\textbf{Scene Texturing}. 
    We show more scene texturing results on multiple 3D-Front scenes \cite{Front3d} with multiple prompts. 
    }
    \label{fig:supp:scenetex_more}
\end{figure*}

%% file: figures/experiments/limitations.tex
\begin{wrapfigure}{r}{0.5\textwidth}
    \centering
    \setlength\tabcolsep{1.25pt}
    \vspace{-60pt}
    \resizebox{0.5\textwidth}{!}{
    \fboxsep=0pt
        \begin{tabular}{ccc}        \fbox{\includegraphics[width=0.15\columnwidth]{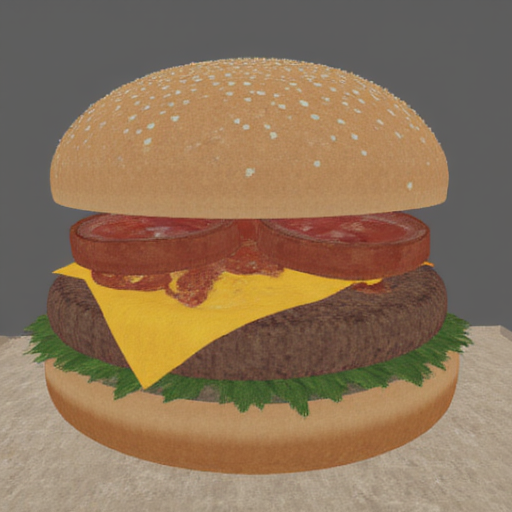}} 
        &
        \fbox{\includegraphics[width=0.15\columnwidth]{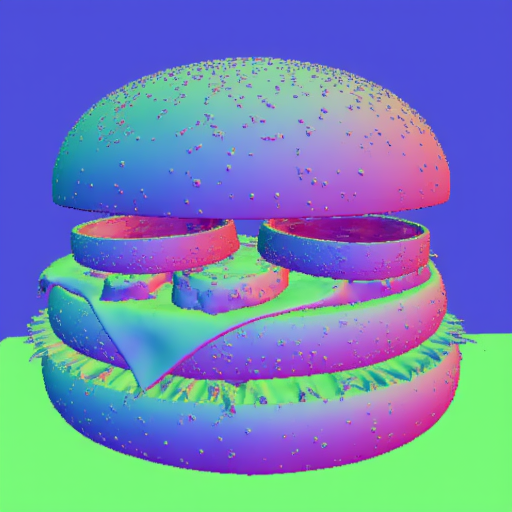}} 
        &
        \fbox{\includegraphics[width=0.15\columnwidth]{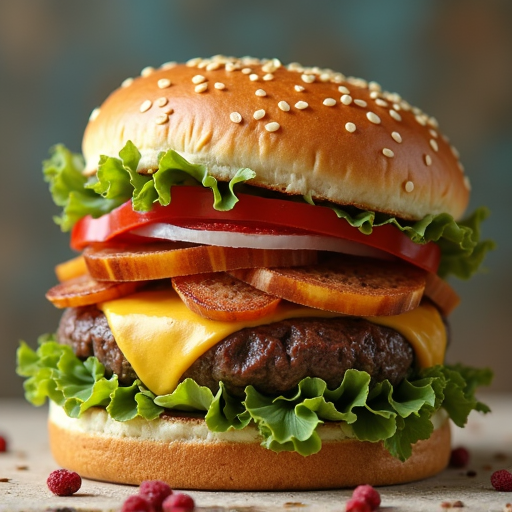}} 
        \\

        Albedo & 
        Normal & 
        FLUX \cite{flux2023}
    \end{tabular}}
    \caption{\textbf{Limitations.}
    Since FLUX \cite{flux2023} does not inherently know about the intrinsic properties and we cannot train on a similarly large dataset as the model was originally trained, we sacrifice details during the fine-tuning. Therefore, our PBR maps do not contain as much details as an image generate by FLUX and sometimes the generated properties can be incorrect (e.g. normals). 
    }
    \label{fig:limitations}
\end{wrapfigure}

%% file: tables/licenses.tex
\begin{table}[b]
    \begin{center}
    \caption{\textbf{Licenses.}
    We provide a summary about the licenses of the used resources. 
    }
    \label{tab:supp:licenses}
    \vspace{9pt}
    \begin{tabular}{lcc}
      \toprule
       Type & Source & License  \\
       \midrule 
       Code & \tt\href{https://github.com/black-forest-labs/flux}{Flux 1.0 Dev} & Apache v2.0 \\
       Code & \tt\href{https://github.com/daveredrum/SceneTex}{SceneTex} & CC BY-NC-SA 3.0 \\
       Data & \tt\href{https://github.com/jingsenzhu/IndoorInverseRendering}{InteriorVerse} & MIT License \\
       Data & \tt\href{https://tianchi.aliyun.com/specials/promotion/alibaba-3d-scene-dataset}{3D-FRONT} & Custom, research-only \\
       Data &\tt\href{https://github.com/modelscope/richdreamer/tree/main/dataset/gobjaverse}{GObjaVerse} & Apache v2.0\\
      \bottomrule
    \end{tabular}
    \end{center}
\end{table}

%% file: figures/user_study/sample_questions.tex
\begin{figure*}[t]
    \centering
    \setlength\tabcolsep{1.25pt}
    \resizebox{\textwidth}{!}{
    \fboxsep=0pt
        \begin{tabular}{cc}
        \rotatebox{90}{Albedo Preference}
        &
        \fbox{\includegraphics[width=0.9\textwidth]{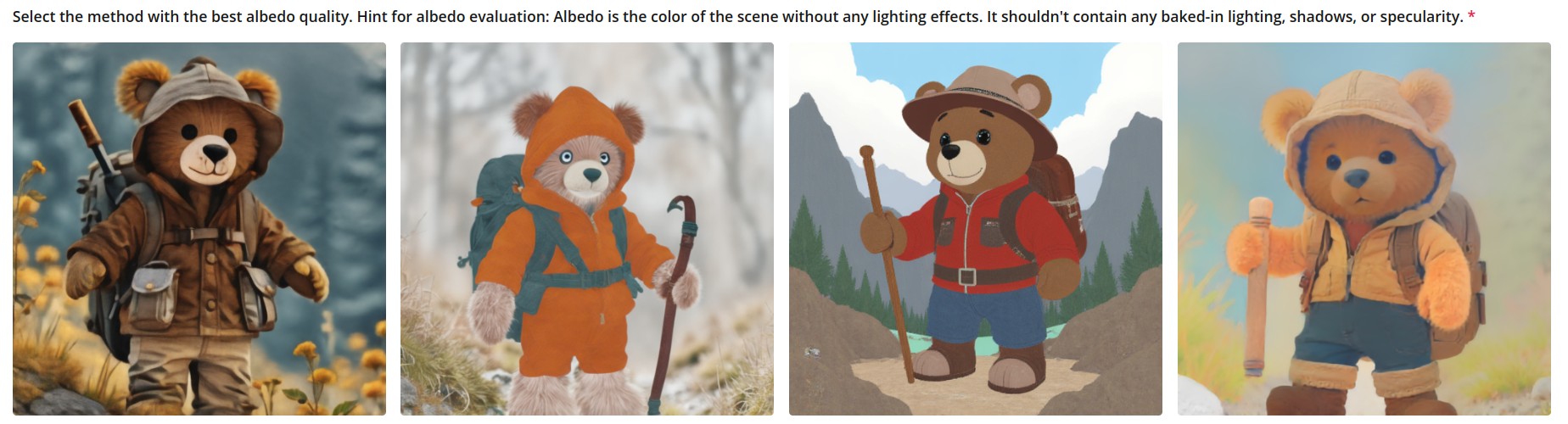}} 
        \\
        \rotatebox{90}{Rendering/Lighting Quality, Prompt Coherence}
        &
        \fbox{\includegraphics[width=0.9\textwidth]{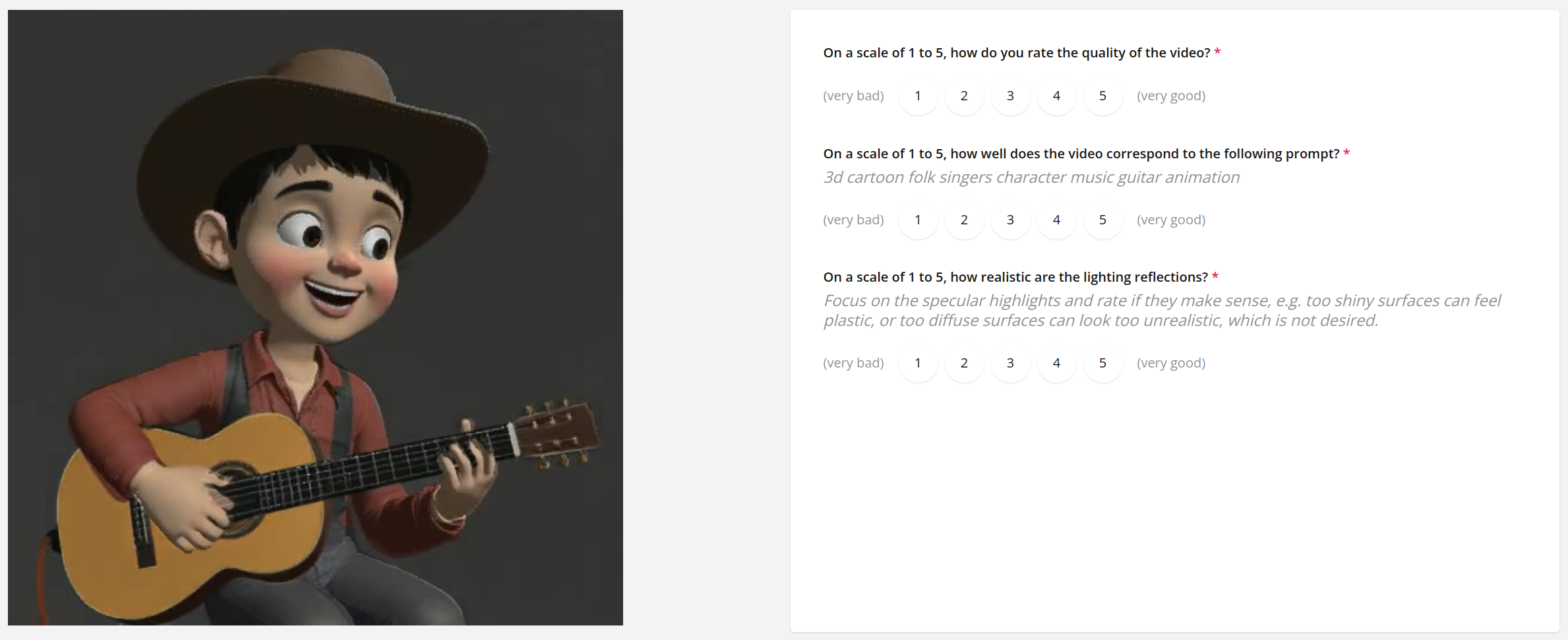}} 
    \end{tabular}}

    \vspace{-6pt}
    \caption{\textbf{Sample questions in the user study}. 
    Users are presented with two types of questions.
    \textbf{Top:} users select the best albedo among all methods.
    \textbf{Bottom:} users rate the specular and rendered quality as well as the prompt coherence on a scale of 1-5 for a rendered video example.
    }
    \label{fig:user-study:sample-questions}
    \vspace{-6pt}
\end{figure*}